%% file: main.tex
\documentclass[11pt]{report}
\let\ifpdf\relax
\usepackage[dvipsnames]{xcolor}
\usepackage{graphicx,pstricks}
\usepackage{hangcaption}
\usepackage{txfonts}
\usepackage{palatino}
\usepackage{pifont}
\usepackage{enumitem}
\usepackage{etaremune}

\usepackage{amsmath,amsfonts,bm, amsthm}
\usepackage{mathtools}
\usepackage{physics}
\input{math_commands.tex}

\usepackage{algorithm}
\usepackage{algpseudocode}

\definecolor{customblue}{HTML}{0080FF}
\definecolor{darkblue}{rgb}{0, 0, 0.5}
\usepackage[hyphens]{url}
\usepackage{hyperref}
\hypersetup{
	colorlinks=true,
	urlcolor=customblue,
	citecolor=darkblue,
	linkcolor=customblue,
}
\usepackage{subcaption}
\usepackage{booktabs}
\usepackage{listings}
\usepackage{multirow}
\usepackage{placeins}
\usepackage{array}
\usepackage{colortbl}
\usepackage{tcolorbox}
\usepackage{charter}
\usepackage[left=1in, right=1in, top=1in, bottom=1in]{geometry}

\usepackage[numbers, sort, compress]{natbib}

\newenvironment{abstractcustom}%
  {\begin{center} 
    BETWEEN RANDOMNESS AND ARBITRARINESS:\par
    SOME LESSONS FOR RELIABLE MACHINE LEARNING AT SCALE\par
A. Feder Cooper, Ph.D.\par
Cornell University 2024\par
  \end{center}\par}%
  {\par}

\newenvironment{biography}%
  {\begin{center}\bfseries Biographical Sketch\end{center}\par}%
  {\par}

\newenvironment{dedication}
{\cleardoublepage\centering\vspace*{\fill}}
  {\vspace*{\fill}\clearpage}

\newenvironment{acknowledgements}%
  {\begin{center}\bfseries Acknowledgements\end{center}\par}%
  {\par}

\renewcommand{\caption}[1]{\singlespacing\hangcaption{#1}\normalspacing}

\input{macros.tex}

\newcommand{\customtitle}{Between Randomness and Arbitrariness: \\Some Lessons for Reliable Machine Learning at Scale}

\title{\customtitle}
\author {A. Feder Cooper}
\date{June 3, 2024}

\begin{document}

\begin{titlepage}
    \centering
    \vspace*{4cm} 
    {\LARGE \textbf{\customtitle}\par\looseness=-1} 
    \vspace{1cm}
    {\Large A. Feder Cooper\par} 
    \vspace{1cm}
    {\large June 3, 2024\par} 
    \vspace{3cm}
    A Dissertation\par 
    Presented to the Faculty of the Graduate School\par
    of Cornell University\par
    in Partial Fulfillment of the Requirements for the\par 
    Degree of Doctor of Philosophy
    \vfill
    {\large \textcopyright\ 2024 A. Feder Cooper. All rights reserved.\par}
\end{titlepage}


\input{section/00-pre/000-abstract}
\input{section/00-pre/001-bio}
\input{section/00-pre/002-dedication}

\input{section/00-pre/003-acks}

\tableofcontents



\input{section/10-intro/000-intro-front}
\input{section/20-arbitrary/200-arb}
\input{section/30-algorithms/300-algos}

\input{section/40-genai/400-genai}

\input{section/50-conclusion/500-conclusion}
\let\cleardoublepage=\origcleardoublepage 
\input{section/99-appendix/900-appendix}

\bibliography{references}

\end{document}

%% file: math_commands.tex
\newtheorem{theorem}{Theorem}
\newtheorem{assumption}{Assumption}
\newtheorem{lemma}{Lemma}

\def\eqref#1{equation~\ref{#1}}

\def\1{\bm{1}}

\def\rvg{{\mathbf{g}}}

\def\rvx{{\mathbf{x}}}
\def\rvy{{\mathbf{y}}}

\def\rmD{{\mathbf{D}}}

\def\vtheta{{\bm{\theta}}}

\def\vc{{\bm{c}}}

\def\vg{{\bm{g}}}
\def\vh{{\bm{h}}}

\def\vo{{\bm{o}}}
\def\vp{{\bm{p}}}

\def\vr{{\bm{r}}}
\def\vs{{\bm{s}}}

\def\vv{{\bm{v}}}
\def\vw{{\bm{w}}}
\def\vx{{\bm{x}}}
\def\vy{{\bm{y}}}
\def\vz{{\bm{z}}}

\def\mD{{\bm{D}}}

\def\mX{{\bm{X}}}

\DeclareMathAlphabet{\mathsfit}{\encodingdefault}{\sfdefault}{m}{sl}
\SetMathAlphabet{\mathsfit}{bold}{\encodingdefault}{\sfdefault}{bx}{n}

\def\sD{{\mathbb{D}}}

\def\sG{{\mathbb{G}}}
\def\sH{{\mathbb{H}}}

\def\sK{{\mathbb{K}}}

\def\sO{{\mathbb{O}}}

\def\sX{{\mathbb{X}}}
\def\sY{{\mathbb{Y}}}
\def\sZ{{\mathbb{Z}}}

\newcommand{\E}{\mathbb{E}}

\newcommand{\R}{\mathbb{R}}

\DeclareMathOperator*{\argmax}{arg\,max}
\DeclareMathOperator*{\argmin}{arg\,min}

%% file: macros.tex
\newcommand{\todo}[1]{\textcolor{red}{\textbf{TODO: #1}}}
\newcommand{\custompar}[1]{\noindent{\bf #1.}\:}
\newcommand{\dalle}{DALL·E}
\newcommand{\chatgptendpt}{\texttt{gpt-3.5-turbo}}
\newcommand{\prompt}[1]{\texttt{"#1"}} 
\definecolor{mustard}{HTML}{b59944}

\include{macros/macros-hpo}

\include{macros/macros-fairness}
\include{macros/macros-tunamh}
\include{macros/macros-cdgrab}
\include{macros/macros-cc}

%% file: macros/macros-hpo.tex
\newcommand{\necessary}{\Box}
\newcommand{\possible}{\Diamond}
\newcommand\algA{$\mathcal{J}$}
\newcommand\algB{$\mathcal{K}$}

\newtheorem{definition}{Definition}
\newtheorem*{definition*}{Definition}
\newtheorem{statement}{Statement}[]

%% file: macros/macros-fairness.tex
\newcommand{\pred}{\hat{y}}
\newcommand{\instance}{\vx}
\newcommand{\group}{\vg}
\newcommand{\olabel}{o}
\newcommand{\olabels}{\sO}
\newcommand{\instances}{\sX}
\newcommand{\labels}{\sY}
\newcommand{\hatlabels}{\hat{\labels}}
\newcommand{\datasets}{\sD}
\newcommand{\dataset}{\mD}
\newcommand{\hatdataset}{\hat{\dataset}}
\newcommand{\hatdatasets}{\hat{\datasets}}
\newcommand{\datasetk}{\dataset_k}
\newcommand{\hatdatasetk}{\hatdataset_k}
\newcommand{\hclass}{\sH}
\newcommand{\tproc}{\mathcal{A}}
\newcommand{\model}{h}
\newcommand{\modelk}{\model_{\datasetk}}
\newcommand{\hatmodel}{\hat{h}}

\newcommand{\possiblemodels}{\mu}
\newcommand{\regressor}{r}
\newcommand{\regressork}{\regressor_{\datasetk}}
\newcommand{\lossarb}{\ell}
\newcommand{\boot}{B}

\newcommand{\err}{\texttt{Err}}
\newcommand{\fp}{\texttt{FP}}

\newcommand{\fn}{\texttt{FN}}
\newcommand{\tp}{\texttt{TP}}
\newcommand{\tn}{\texttt{TN}}

\newcommand{\fpr}{\texttt{FPR}}

\newcommand{\fnr}{\texttt{FNR}}
\newcommand{\haterr}{\hat{\texttt{Err}}}
\newcommand{\hatfpr}{\hat{\texttt{FPR}}}
\newcommand{\hatfnr}{\hat{\texttt{FNR}}}
\newcommand{\costfp}{C_{\text{01}}}
\newcommand{\costfn}{C_{\text{10}}}

\newcommand{\variance}{\texttt{var}\big(\tproc, \datasets, (\instance, \group)\big)}
\newcommand{\hatvariance}{\hat{\texttt{var}}\big(\tproc, \hatdatasets, (\instance, \group)\big)}
\newcommand{\hatvar}{\hat{\texttt{var}}}
\newcommand{\consistency}{\texttt{SC}\big(\tproc, \datasets, (\instance, \group)\big)}
\newcommand{\hatconsistency}{\hat{\texttt{SC}}\big(\tproc, \{\hatdataset_b\}_{b=1}^\boot, (\instance, \group)\big)}
\newcommand{\hatsc}{\hat{\texttt{SC}}}
\newcommand{\hatar}{\hat{\texttt{AR}}}

%% file: macros/macros-tunamh.tex
\newtheorem{corollary}{Corollary}
\newcommand{\methodname}{TunaMH}
\newcommand{\bigTheta}{\mathcal{O}\!\!\!\!\raisebox{1pt}{\text{--}}\,}
\newcommand{\Abs}[1]{\left| #1 \right| }

%% file: macros/macros-cdgrab.tex
\newcommand{\grab}{GraB}
\newcommand{\cgrab}{\grab}
\newcommand{\dgrab}{CD-\grab}

\newcommand{\shuffle}{RR}
\newcommand{\dshuffle}{D-\shuffle}
\newcommand{\so}{SO}

\newcommand{\epochs}{T}
\newcommand{\workers}{m}
\newcommand{\workerexamples}{n}
\newcommand{\examples}{N}

\newcommand{\windex}{i}
\newcommand{\exindex}{j}
\newcommand{\eindex}{t}
\newcommand{\loss}{f}

\newcommand{\dataex}{\vx}
\newcommand{\ex}{\vz}
\newcommand{\barex}{\bar{\vz}}
\newcommand{\exij}{\ex_{\windex,\exindex}}

\newcommand{\exj}{\ex_\exindex}
\newcommand{\weights}{\vw}
\newcommand{\weightst}{\weights_\eindex}
\newcommand{\perm}{\pi}

\newcommand{\sgn}{s}
\newcommand{\signj}{\sgn_\exindex}
\newcommand{\wexsign}{\sgn_\exindex^\windex}
\newcommand{\wexsignprev}{\sgn_{\exindex-1}^\windex}

\newcommand{\g}{\vg}
\newcommand{\wexgrad}{\g_\exindex^\windex}
\newcommand{\wexgradprev}{\g_{\exindex-1}^\windex}

\newcommand{\rarrow}[2]{\xrightarrow{\makebox[#1]{$#2$}}}
\newcommand{\larrow}[2]{\xleftarrow{\makebox[#1]{$#2$}}}

\newcommand{\dstep}[1]{\textcolor{blue}{\textbf{\texttt{#1}}}}

%% file: macros/macros-cc.tex
\newcommand{\captionmethod}{telephoning}
\newcommand{\capcaptionmethod}{Telephoning}
\newcommand{\datasetname}{CommonCatalog}
\newcommand{\modelname}{CommonCanvas}
\newcommand{\sdtwobase}{SD2-90M}

\newlength{\groupwidth}
\newlength{\itemwidth}

\definecolor{amber}{rgb}{1.0, 0.75, 0.0}
\definecolor{paleblue}{rgb}{0.678,0.847,0.901}
\definecolor{palepink}{rgb}{1,0.714,0.757}
\definecolor{paleyellow}{rgb}{1,1,0.878}

%% file: section/00-pre/000-abstract.tex
\begin{abstractcustom}
To develop rigorous knowledge about ML models --- and the systems in which they are embedded --- we need reliable measurements. 
But reliable measurement is fundamentally challenging, and touches on issues of reproducibility, scalability, uncertainty quantification, epistemology, and more. 
This dissertation addresses criteria needed to take reliability seriously: both criteria for designing meaningful metrics, and for methodologies that ensure that we can dependably and efficiently measure these metrics at scale and in practice. 
In doing so, this dissertation articulates a research vision for a new field of scholarship at the intersection of machine learning, law, and policy. 
Within this frame, we cover topics that fit under three different themes. 

First, we quantify and mitigate sources of arbitrariness in machine learning, with respect to hyperparameter optimization and social prediction contexts.
We clarify important connections between machine-learning arbitrariness, rooted in non-determinism, with legal notions of arbitrariness that implicate legal rules and due process.

Second, we tame randomness in uncertainty estimation and optimization algorithms, in order to achieve scalability without sacrificing reliability. 
We discuss how across computing, and particularly in machine learning, scalability and reliability are typically in trade-off.
Analogous trade-offs in law and policy make this type of trade-off a useful abstraction for communicating about machine-learning capabilities and risks to policymakers and other non-expert stakeholders. 

Third, we provide methods for evaluating generative-AI systems, with specific focuses on quantifying memorization in language models and training latent diffusion models on open-licensed data. 
These contributions have urgent and significant connections to U.S. copyright law. 
We provide an abridged discussion of landmark legal scholarship that details the complicated relationships between generative-AI supply chain and copyright. 

By making contributions in these three themes, this dissertation serves as an empirical proof by example that research on  reliable measurement for machine learning is intimately and inescapably bound up with research in law and policy. 
These different disciplines pose similar research questions about reliable measurement in machine learning. 
They are, in fact, two complementary sides of the same research vision, which, broadly construed, aims to construct machine-learning systems that cohere with broader societal values. 
\end{abstractcustom}
\newpage

%% file: section/00-pre/001-bio.tex
\begin{biography}
A. Feder Cooper was born and raised in New York, NY, and obtained his B.A. in Computer Science and Archaeology from Columbia University in 2014. 
Prior to a research career, Cooper worked for several years as a software engineer. 
In 2018, he began his Ph.D. in Computer Science at Cornell University. His doctoral work was chaired by Professor Christopher De Sa, with additional advising by James Grimmelmann, Jon Kleinberg, and Adrian Sampson.
His Ph.D. work, broadly construed, studies reliable measurement and evaluation of machine learning, covering both computer science aspects of this work as well as their associated ethical, legal, and policy dimensions.\footnote{He had initially planned on attending Harvard Law School; however, thanks to James Grimmelmann's mentorship, he fortunately felt he could skip pursuing more degrees.} 

His contributions span uncertainty estimation, privacy and security of generative-AI systems, distributed training, hyperparameter optimization, and model selection. 
His work has been recognized by spotlight awards (\emph{NeurIPS 2020}), oral presentation slots (e.g., \emph{AIES 2021}), Best Student Paper (Honorable Mention) at \emph{AAAI 2024}, Best Paper Award at \emph{ICML 2024}, and a ``Rising Star in EECS'' award by MIT in 2021. 
His scholarship on generative AI and copyright has been described as a ``landmark'' contribution.\looseness=-1 

A. Feder Cooper is a co-founder of the GenLaw Center and an Affiliate at the Berkman Klein Center for Internet \& Society at Harvard University. 
His Ph.D. research was supported by the John T. and Catherine D. MacArthur Foundation.
Following the completion of his Ph.D., he will pursue a postdoctoral research position at Microsoft Research, and will be affiliated with Stanford University, working with Percy Liang and Dan Ho. 
He will then begin his faculty career at Yale University, appointed as a professor in the Department of Computer Science and an affiliated faculty fellow at Yale Law School. 
\end{biography}

%% file: section/00-pre/002-dedication.tex
\begin{dedication}
For my grandparents
\end{dedication}

%% file: section/00-pre/003-acks.tex
\begin{acknowledgements}
Over the years, I've heard many metaphors and similes about what graduate school is like. 
Some say it's like a marriage. 
Others say it's like being raised by an academic village. 
Others, still, say it is a trial by fire: to mix metaphors, it's akin to being thrown into the deep end and (hopefully) swimming your way out. 
For me, it's been like none of these things. 
I will save my reflections for another time and venue. 
But I'll note the positive unifying thread of my experience: 
finding and collaborating with a distributed network of researchers that have a deep love and talent for mischief (in the most innocuous sense of the word). 
Indeed, they take mischief more seriously than any people I've met before. 
And this serious mischief has led to some of the most fun and thoughtful collaborations that I could have ever hoped for in my Ph.D. 

First, I want to thank my closest faculty collaborators, my advisor, Chris De Sa, and James Grimmelmann. 
Chris took a chance on me, in many respects a ``non-traditional'' student, while he was junior faculty. 
I entered Cornell without prior research experience in computer science (an increasingly rare occurrence), and with an uncompromising desire to do cross-cutting work in machine learning, systems, and law.  
He gave me the sound advice that this was one interdisciplinary intersection too many, and encouraged me to (at the very most) pick two. 
It's because of his unwavering support, curiosity, kindness, and generosity that I've been able to chart my own path --- to do extensive work in the emerging discipline of machine learning and law. 

James has been a champion for my success since my earliest days at Cornell. 
I am deeply thankful for his feedback, research advice, life advice, and kindness --- all of which have shaped my scholarship, research orientation, and career goals. 
He has been a shining example of the kind of mentor that I hope to be one day. 
After effectively being an unnamed author on some of my earlier work --- and some gentle prodding to become an official co-author --- I feel very lucky that I get to call James one of my closest collaborators. 
He has co-led, helped shape, and seen to completion what has arguably been the most important work in my career.

In addition to James, I would like to thank my other GenLaw collaborators: Katherine Lee, Niloofar Mireshghallah, and Hoda Heidari. 
These three working relationships have had an untold impact on my development as a researcher, collaborator, workshop co-conspirator, and human being. 
These relationships have also evolved into cherished friendships, for which I feel unspeakably fortunate and grateful. 

I would like to thank my committee for their assistance and feedback over the years. In addition to Chris and James, mentioned above, I am very grateful to Jon Kleinberg and Adrian Sampson for their advice and expertise in advising my doctoral work. 
Jon has played a particularly significant role in shaping my thinking about algorithmic fairness, and Adrian is who first introduced me to research that mitigates arbitrariness in computing (in compilers research). 
Both have had a huge impact on the questions I have studied throughout my degree. 

I similarly would like to thank Marilyn Migiel, Pam Samuelson, Abbie Jacobs, Joan Feigenbaum, Solon Barocas, and Michael Littman. 
Though not official members of my doctoral committee, all six of them have had a tremendous impact on the course of my Ph.D. and career.
Marilyn has patiently helped me grow and develop my deep love for Italian language and culture; 
Pam has been an avid supporter and advocate of my legal scholarship and GenLaw; 
Abbie has been an incomparable research-idea thought partner, listening ear, friend, and career strategist; 
Joan has long championed my career as a junior scholar in the field of Computer Science and Law;
Solon has pushed me to think through the (sometimes obscured) normative dimensions of my computing work; 
and Michael has been a great research and career mentor since before I started graduate school, and has also been a great advocate, conversationalist, and pal. 
I am so thankful to have had the opportunity to meet all six of them, let alone get to know them and to consider them mentors. 

I have also had the great fortune to get to know and work with some incredible researchers at Google DeepMind and Google Research. 
I am very grateful to Nicholas Carlini, Zachary Charles, Chris Choquette-Choo, Daphne Ippolito, Matthew Jagielski, and Milad Nasr, who, alongside Katherine Lee, have taught me so much about privacy and adversarial ML research, and what it can look like to work together as a research team. 

I want to also thank Paul Ohm, Alex Givens, and Miranda Bogen who, with Katherine, James, and Hoda, helped make GenLaw DC as a resounding success. 
Thank you to Jack Balkin, Miles Brundage, Chris Callison-Burch, and Zack Lipton for their  continued support and enthusiasm for the research and practice community that we are trying to create and nurture through GenLaw. 

I have also had many great research collaborators over the years --- Ph.D. researchers, undergraduates, postdocs, and professors. 
In particular, I would like to thank the brilliant members of the Relax ML lab, past and present, for their generosity, collegiality, and inspiration over the last six years. 
Thank you to Ruqi Zhang, Yucheng Lu, Cathy Meng, Jerry Chee, Tao Yu, Albert Tseng, Wentao Guo, Yiming Zeng, Jianan Canal Li, Gary Wei, Khiem Pham, Tiancheng Yuan, and Charlie Ruan. 
I am especially indebted to Ruqi and Yucheng. 
When I was just getting acclimated to ML research, Ruqi was a (very) patient, kind, and generous research mentor.
Yucheng has been a fantastic colleague, research advocate, and friend. 
I would also like to thank my colleagues and friends outside of the Relax ML lab, who have have had a major impact on my scholarship --- both directly and indirectly: Maria Antoniak, Manny Moss, Kweku-Kwegyir-Aggrey, Aaron Gokaslan, Jamelle Watson-Daniels, and Jessica Zosa Forde. 
I am especially grateful to Maria for setting an early example in graduate school of the kind of thoughtful, diligent computing researcher I wanted to become, and to Manny for being a phenomenal thought partner and ally.

I would like to thank the various funding sources throughout my Ph.D. 
My work has been made possible by generous support from the John D. and Catherine T. MacArthur Foundation (via Jon Kleinberg and Karen Levy) and the Digital Life Initiative at Cornell Tech (via Helen Nissenbaum), a Cornell University fellowship, a runner-up Ph.D. fellowship from Two Sigma, and grant funding from Chris De Sa, James Grimmelmann, Baobao Zhang, and Adrian Sampson.  

And most importantly, I want to express my deep fondness, appreciation, and love for my family. Thank you to Eric Schwartz, Salonee Bhaman, Jack Goetz, Bryana Williams, Dhari Noel, and Meghan Witherow. Throughout my Ph.D., you have seen the best of me, the worst of me, and, frankly, the most boring of me. 
Thank you for sticking by me and having my back when I needed it most, even when I vanished into my work (sometimes for weeks or months at a time). 
Thank you to Paul and Helaine Cantor for your unwavering belief in me. 
And last, thank you to Fernando, Bela, Leo, and Achilles Delgado; 
thank you for giving me a place I can call home, for helping push me to the finish line, and for being some of the best friends, supporters, and companions over the last several years.
I could not have done any of this without you. 

\end{acknowledgements}

%% file: section/10-intro/000-intro-front.tex
\chapter{Introduction}\label{chapter:intro}

\input{section/10-intro/100-intro}

%% file: section/10-intro/100-intro.tex
In 2016, I was a backend-systems software engineer playing with machine learning (ML) during my afternoons and weekends. 
The U.S. presidential election was in full swing, and 
I had developed the pastime of messing with Facebook's Newsfeed algorithm --- perhaps an early glimpse that I should have been an ML security researcher. 
And in messing with the algorithm, I saw some really horrible content: 
a lot of virulent, bot-farm, fake stuff. 
It was everywhere, it was noxious, and it was so brazenly meant to tip the election. 

Something was clearly wrong with Facebook's content moderation processes. 
Or maybe something was exactly right, depending on how you look at it, if this kind of activity contributed to more clicks and engagement. 
There was clearly a larger phenomenon at play. 
Human-made platform design decisions and ML algorithms were operating in conjunction with really sophisticated software systems ---
systems that worked in real-time and at massive scale on the Internet. 
And these different elements had all mixed together in a potent brew of misinformation and disinformation. 
This was really upsetting to me. 
I had gotten into computing --- and interested in machine learning in particular --- because it is fun. 
And this stuff (among other things) was decidedly not fun.

It might not have been fun, but it clarified some really big questions for me.
It was obvious that large-scale, ML-powered systems (not just ML algorithms) were here to stay. 
Given this reality, what should we want these systems to do in the world? 
How can we make sure that these systems are reliable?
What does reliability even mean?
And if we are unable to make ML systems sufficiently reliable, are there areas where we should not use ML at all?
How can we reason rigorously about this distinction, if it exists?
How can we be sure that an ML system's behavior matches up in practice with our intentions and goals? 
What tools do we have at our disposal --- or what tools do we need to invent --- to help us reason about this?


There were clearly big, rich, concrete questions in machine learning to study here --- 
in topics like uncertainty quantification, model selection, algorithms and systems trade-offs, and much else. 
There were also big, rich, concrete questions in law and policy. 
For example, we could hypothetically come up with the best-ever, theory-backed, ML-based tools for quantifying uncertainty, maybe even at scale. 
But just because we have a great tool does not mean it is immediately or generally clear how we should use it in practice. 
Practical considerations require communication with non-expert stakeholders --- people who are involved in decisions about whether and how to use ML systems in real-world domains.
In this case, this would involve communicating about what different types of uncertainty exist, what they mean concretely in particular practical domains, and, based on its underlying assumptions, what types of uncertainty our great ML-based tool can (and cannot) measure. 

More generally, how should we communicate about design choices in ML? 
Most of these choices are not foregone conclusions. Someone (or some group of people) typically makes some decision at some point in time about which particular model to use in practice.
How do we communicate clearly about these types of choices and their consequences to non-experts?
How can we make sure that other stakeholders, like policymakers, have necessary and sufficient understanding of ML systems and design choices, so that they can construct sound and useful AI public policy? 

Looming among these research questions, there were some big personal ones, too.
What was the best way for me to go about trying to find answers to such questions?
Should I go to law school? 
Should I go get a Ph.D. in machine learning? 
Should I do both? 
Well, since this is the introduction to my dissertation, it is hopefully clear that I decided to do the ML Ph.D. 
But I also reasoned that it should be possible to tackle these questions side by side, all at once. 
Questions like these 
are two complementary sides of the same research vision.
They all involve research into how to do reliable measurement for ML at scale, where what constitutes ``reliability'' takes into account considerations that are relevant not just for ML, but also for law and policy. 

There is a virtuous cycle in this type of work. 
Making contributions with this particular focus in ML is indivisible from concrete implications for tech law and policy; 
doing deep work in tech law and policy raises novel research questions to tackle on metrics and measurement practices in ML.
For example, in order to understand the copyright implications of generative-AI systems, we need to be able to take useful and replicable measurements that can help inform questions judges and policymakers have about issues like copyright infringement.   

Following this vision, I have begun an extensive research program in machine learning, law, and policy, and I have done this work across a bunch of projects. 
I am the first author on most of them~\citep{cooper2024files,lee2023explainers,lee2023talkin, cooper2024talkinshort, zhang2020tunamh, cooper2023cdgrab, cooper2024variance, cooper2022lawless, cooper2022accountability, cooper2022arpa, cooper2022fast, cooper2021hpo, cooper2021emergent, cooper2021eaamo, cooper2021tecnologica, cooper2023report, forde2021model}, and much of this work has received awards --- spotlight, oral, and best paper  accolades~\citep{zhang2020tunamh, cooper2024variance, cooper2024talkinshort, cooper2021emergent, cooper2021eaamo, aggrey2023repair, forde2021model, cooper2022lawless, carlini2024stealing}.

Even if all of this research touches on topics that  fundamentally have to do with the intersection of machine learning, law, and policy, it has been very important to make sure that the core contributions of each piece are cognizable to the appropriate disciplinary audiences. 
As a result, a large number of these projects have their main contribution positioned in machine learning, and have been published  or presented in venues like \emph{NeurIPS}, \emph{ICML}, \emph{AAAI}, and the like~\citep{zhang2020tunamh, zhang2020amagold, cooper2021hpo, cooper2023cdgrab, cooper2024variance, aggrey2023repair, forde2021model, cooper2021tecnologica, carlini2024stealing, nasr2023scalable, gokaslan2023commoncanvas, mcduff2024license}.
A smaller number have had their main contribution in law and policy, and have been published in law reviews and interdisciplinary computing venues like \emph{ACM CSLAW}~\citep{cooper2024files, cooper2024talkinshort, lee2023talkin, cooper2022lawless, cooper2022fast, cooper2021eaamo, cooper2023report}.
A smaller number still have their main contribution in computing ethics and values, and have been published at venues like \emph{ACM FAccT}~\citep{laufer2023fouryears, cooper2022arpa, cooper2022accountability, cooper2021emergent, lee2023explainers}.

\begin{figure}[t!]
    \centering
    \includegraphics[width=.99\textwidth]{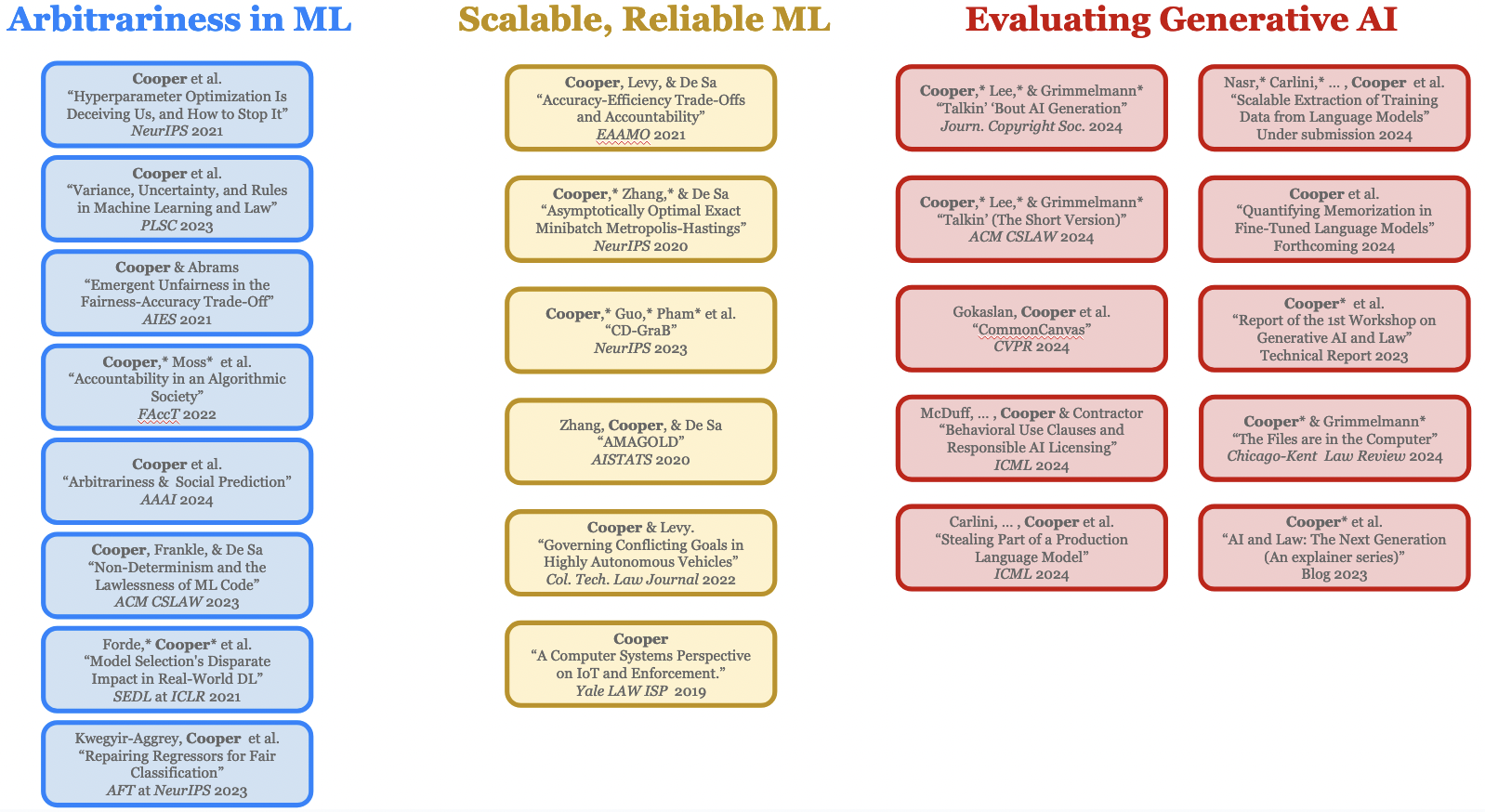}
    \caption{Ph.D. projects organized by theme. 
        Some projects do not fit neatly into these divisions~\citep{cooper2022arpa, laufer2023fouryears,cooper2021tecnologica}, and many projects cross boundaries.
        Notably, Appendix~\ref{chapter:accountability}~\citep{cooper2022accountability} touches on all three themes.  
    }\label{fig:intro:work}
\end{figure}

Maintaining these disciplinary boundaries has been useful to keep in mind for publishing; however, what has been more useful, with respect to posing research questions, is considering overarching research themes. 
There were two themes that I had intended to explore in my Ph.D., based on my initial motivation for going to graduate school: 
sources of arbitrariness in ML and scalable ML algorithms (Figure~\ref{fig:intro:work}). 
My work on arbitrariness is deeply related to model selection choices --- 
ML modeling and algorithm choices that people make, which can lead to arbitrary outcomes. 
In scalable ML algorithms, my work has studied how to make algorithms more efficient while retaining reliability guarantees, predominantly in uncertainty estimation.

Both themes have clear connections to law and policy. 
Arbitrariness is a very important concept in the law, for example, with respect to due process~\cite{fuller1965law}. 
In light of this relationship, I have focused my work on quantifying and mitigating ML-specific types of arbitrariness, and making these types of arbitrariness cognizable for law and policy.
Scalability and reliability are often in trade-off; this can serve as a useful abstraction for communicating with policymakers about implementation decisions and associated capabilities and risks.

With two coherent themes concerning ML, law, and policy, we could perhaps call it day.
One such theme might be a happy accident, but two entirely different ones indicates a pattern --- an indication that this field of work is a fruitful direction for original scholarship. 
However, the dissertation does not end here.  

In summer 2020, I was tinkering with GPT-2 and GPT-3, shortly after GPT-3~\cite{brown2020gpt3} came out.
There was a clear leap in quality between GPT-2 and GPT-3;
GPT-3 was nearing human-like text generation.
Its architecture was larger, and it was also trained on a much larger quantity of (likely copyrighted) text data. 
One day, when there was an ever better model, GPT models would no longer be a research curiosity. 
They would be sufficiently impressive, such that they would be embedded in consumer-facing products that people would actually want to use. 
And when that day came, it would likely be a nightmare for intellectual property (IP) law. 

This was just a hunch; 
I did not know much about IP law at the time. 
So, in Fall 2020, I decided to enroll in a course on IP at the law school, and then I waited. 
And I did not have to wait long because, about two years later, OpenAI released ChatGPT and everything changed. 
All of the considerations that had brought me to graduate school were, all of a sudden, immediately and inescapably relevant. 
There was a real-time, large-scale, ML-driven system, governed by innumerable human design choices, that had enormous societal implications --- and everyone was using it. 
I would no longer have to explain why work at the intersection of ML, law, and policy was so important. 
Everyone would know it from firsthand experience. 

In other words, this moment presented a huge opportunity for the type of work I had already been pursuing. 
But it also meant that I should redirect my energy toward a third line of work in the last year of my degree --- a line of work on generative AI and law. 
Based on the enormous and urgent demand for clarity and rigor in this area, my work in this theme has thus-far focused on evaluations for generative-AI systems that provide insights for U.S. copyright law.\looseness=-1

\subsection*{Dissertation Format}

This dissertation is organized in three parts around these three themes. 
\begin{itemize}
    \item Part~\ref{part:arbitrary} addresses arbitrariness in machine learning.
    \item Part~\ref{part:algorithms} details projects in scalable machine learning algorithms.
    \item Part~\ref{part:genai} discusses evaluating generative-AI systems, with particular attention to copyright-related topics.
\end{itemize}
Each of these parts is outlined in the remainder of this introduction (Sections~\ref{sec:intro:arbitrary},~\ref{sec:intro:algorithms}, and~\ref{sec:intro:genai}, respectively). 
While they are presented separately, it is worth noting that the three themes they cover appear throughout. 
For example, scalable machine algorithms and their associated trade-offs feature in all three parts. 

In an attempt at concision, this dissertation only addresses a subset of the research projects mentioned above (Figure~\ref{fig:intro:work}). 
Each part contains the same overall structure of three chapters that have been integrated into a single narrative.
The first two chapters reflect papers that contain core contributions in machine learning, and third chapter demonstrates how the first two have deep interrelationships with tech law and policy. 
Additional research concerning cross-cutting  philosophical questions about the relational aspects of ML accountability is deferred to the appendix.\looseness=-1

\input{section/10-intro/110-arbitrary}
\input{section/10-intro/120-algorithms}
\input{section/10-intro/130-genai}

\section{Closing Thoughts}


The nine chapters discussed above may seem neatly organized into the three discrete themes outlined in this introduction.  
Nevertheless, while reading, it is worth keeping in mind that these divisions are somewhat artificial; 
all three themes are cross-cutting. 
They appear in different degrees throughout the entirety of this dissertation. 
For example, trade-offs between reliability and efficiency do not just appear in our work on machine learning algorithms.
They also appear throughout all of our work on evaluating generative-AI systems (e.g., we choose a relatively simple metric for extractable memorization, because it is more efficient to measure at large scale). 
They permeate the choices we make when formulating ways to  measure and mitigate arbitrariness (we sacrifice a good deal of efficiency for reliable proxies of arbitrariness). 

All of these themes bubble up into the overarching questions that we began with in this introduction --- the questions that brought me to graduate school. 
These questions fundamentally concern how to do reliable measurement for machine learning at scale: 
making choices in metric design (e.g., how we choose to define uncertainty in ML), figuring out how we can dependably measure these metrics at scale and in practice (e.g., in large-scale systems with real-time capabilities), and communicating the effects of our measurements to other, often non-expert stakeholders (e.g., policymakers).

At a higher level, still, all of these research questions are about pursuing, developing, and refining what  we want ML systems to do in the world. 
They are about how we can make sure that ML system behavior matches up with our goals, values, and intentions. 
For me, these remain the big important questions. 
This dissertation is just a start at carving out some smaller, concrete questions that we can answer, in service of these big important ones. 

%% file: section/10-intro/110-arbitrary.tex
\section{Part~\ref{part:arbitrary}: Sources of Arbitrariness in Machine Learning}\label{sec:intro:arbitrary}

Part~\ref{part:arbitrary} presents three inter-related research projects that study arbitrariness in machine learning and its  consequences for law and policy. 
Broadly speaking, this work studies how human-made decisions can lead to arbitrary results or conclusions in ML experiments. 
These decisions may seem quite mundane in practice --- the selection of a particular set of hyperparameters~\citep{cooper2021hpo} (Chapter~\ref{chapter:hpo}) or a specific classification model to deploy~\citep{cooper2024variance} (Chapter~\ref{chapter:fairness}) --- but they can in fact result in outcomes that mislead us about ML  capabilities and risks. 
As a result, ML arbitrariness is a significant consideration for law and policy~\citep{cooper2022lawless}. 
Indeed, there are deep connections between arbitrariness in machine learning and how law and policy reason about and mitigate unwanted sources of arbitrariness in legal contexts (Chapter~\ref{chapter:nondeterminism}).

\subsubsection*{Chapter~\ref{chapter:hpo}: Arbitrariness in Hyperparameter Optimization Choices}

This part opens with work on characterizing arbitrariness in hyperparameter optimization (HPO). 
In particular, Chapter~\ref{chapter:hpo} uses tools from modal logic to formalize the process of drawing conclusions about algorithm performance when running hyperparameter optimization in machine learning experiments.

\begin{figure}[t!]
    \centering
    \includegraphics[width=.9\textwidth]{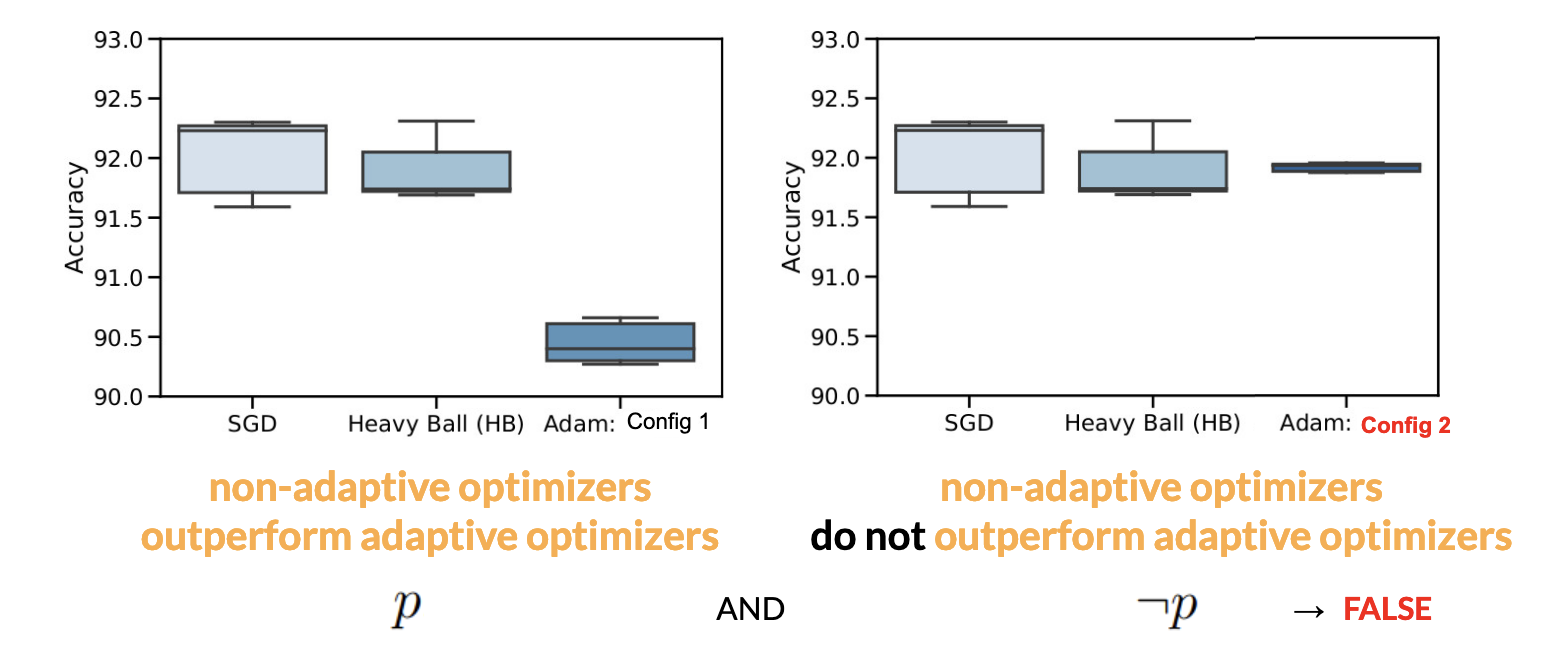}
    \caption{Running different sets of experiments for training the VGG-16 architecture to classify images in CIFAR-10. 
    Both sets of experiments test SGD, Heavy Ball momentum, and Adam. The experiments on the right use one configuration for Adam, and the experiments on the left use another. 
    In isolation, each of these sets of experiments leads to a conclusion that, when considered together, result in a logical contradiction.}.
    \label{fig:intro:hpo}
\end{figure}

It is well-known that HPO greatly affects overall measurements of algorithm performance.
There is much prior experimental work in machine learning that has articulated this point~\cite{choi2019empirical, sivaprasad2020hpo, dodge2019nlp}, such that it is safe to say that it is common knowledge in the ML community. 
HPO can affect results so much that the results of two different HPO procedures for the same task and the same optimizers can lead to contradictory conclusions.
The two sets of experiments in Figure~\ref{fig:intro:hpo} highlight this phenomenon. 
Both experiments test three optimizers --- SGD, Heavy Ball momentum, and Adam --- to train the VGG-16 neural network to classify the CIFAR-10 dataset.  
On the left, we test one set of hyperparameter configurations, pick the best-performing configuration per optimizer, and compare test accuracy.
We do the same thing for the experiments on the right, but we change how we configure the hyperparameter search space for Adam --- represented in the third, rightmost box plot.

Separately, the plot for each of these sets of experiments suggests a particular conclusion.  
On the left, it looks like Adam performs worse than SGD and Heavy Ball. 
That is, the results reasonably suggest the conclusion that non-adaptive optimizers like SGD and Heavy ball outperform adaptive ones like Adam. 
The results on the right tell a very different story. 
Judging by test accuracy alone,\footnote{If we consider variance, Adam seems to out-perform SGD and Heavy Ball.} Adam performs just as well as SGD and Heavy Ball. 
If we were to accept both sets of experiments as valid HPO configurations to test empirically, we would yield a logical contradiction (Figure~\ref{fig:intro:hpo}). 
This implies that these sets of experiments cannot both be valid ways to test hyperparameters because, taken together, the conclusions they suggest are inconsistent.
Taken together, these experiments do not enable us to produce reliable knowledge about algorithm performance.

Ideally, we want to avoid this type of situation in ML research, since one of our goals is to develop reliable knowledge about algorithm performance. 
Importantly, this is not the same as making claims from ML experiments involving HPO that have to do with ground-truth algorithm performance. 
We do not know the ground truth. 
Instead, we want to make sure that the ML community does not accept \emph{a priori} a particular methodology for configuring and performing HPO that could possibly lead to inconsistent conclusions, like those in Figure~\ref{fig:intro:hpo}.
In other words, it would, be fine for the ML community to accept exclusively either of the sets of experiments in Figure~\ref{fig:intro:hpo}, and to draw the selected set's related conclusion. 
Or it would be fine for the ML community to be skeptical --- to accept neither of these sets of experiments, and to conclude nothing at all about algorithm performance.
However, it is not fine for the ML community to accept both sets of experiments as valid, as this is the case that leads to inconsistent conclusions.

This is a bit of a subtle point. 
Obviously, when presented with these two sets of experiments side-by-side, we know to reject them because they yield inconsistent conclusions.
But this is not typically what happens in practice.
Instead, researchers typically perform one (if any) pass of HPO, which in our motivating example would only produce one set of experiments in Figure~\ref{fig:intro:hpo} from which one could form conclusions. 
In our work in this chapter, we therefore aim to study a kind of meta-problem: 
we want to make sure that, even when we are presented with only one set of results, we form conclusions that are not \emph{arbitrary} --- conclusions that constitute reliable knowledge. 
That is, if someone else had by happenstance configured HPO slightly differently for the same overall experiment, they would not have yielded results that suggest a conclusion that contradicts the one that we have obtained.

Based on this motivation, we attempt the first theoretical study of how to draw reliable conclusions from empirical studies using HPO.
We pursue this goal in two parts. 
First, we come up with a formalization that enables us to reason about two vague types of uncertainty in our problem setup: (1) the possible outcomes of HPO experiments and (2) whether we believe the conclusions that can be drawn from those outcomes. 
The point of formalizing our beliefs is to instill an appropriate amount of doubt when examining HPO results: even if we cannot know for certain what is true, we do not want to end up believing a conclusion that is false~\citep{descartes1996evildemon}.

We use modal logic~\cite{blackburn2006modal} for this formalization, since it is a useful analytical tool for pinning down vague, difficult-to-capture (non-stochastic) types of uncertainty in both of these sources. 
Second, we use our formalization to prove non-trivial theorems about whether or not a hyperparameter optimization procedure is defended drawing false, inconsistent conclusions. 
We suggest an HPO procedure and use our formalization to prove that it is defended against such an outcome (within a limited time budget). 

\subsubsection*{Chapter~\ref{chapter:fairness}: Arbitrariness in Social Prediction}

There are many other sources of arbitrariness in machine learning, not just the (non-stochastic) arbitrariness that gets introduced through decisions in configuring hyperparameter optimization procedures.
In another line of work, we investigate another type of arbitrariness related directly to randomness: 
how arbitrary the choice of single model is, based on the specific random seed used for training, in algorithmic fairness contexts. 

\begin{figure}[t!]
    \centering
    \includegraphics[width=.9\textwidth]{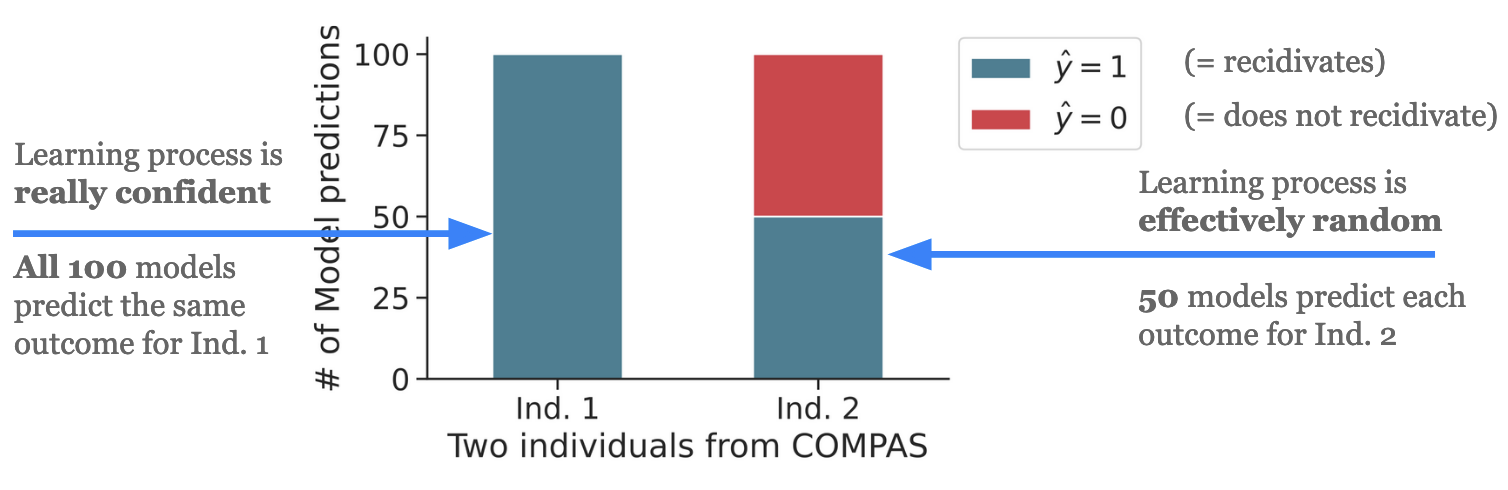}
    \caption{100 bootstrapped random forest models show models can be very consistent in predictions $\hat{y}$ for some individuals (Ind. 1) and arbitrary for others (Ind. 2).
    In this example, 50 models result in predictions that suggest Ind. 2 will \emph{recidivate} (i.e., commit a crime again) and 50 that suggest they will not. 
    Their prediction is \emph{arbitrary}.
    \looseness=-1}
    \label{fig:intro:vote}
\end{figure}

To get a sense for this arbitrariness, let us examine a simple example.
Consider training 100 random forest models on COMPAS, which is (for many reasons) an infamous binary classification task that has been used to predict whether someone is going to \emph{recidivate} --- whether they are going to commit a crime again~\citep{larson2016propublica}.
Such predictions can then be used to inform whether an individual is allowed to receive bail or not, if they are rearrested.\footnote{There are many issues with this setup, ranging from problem formulation issues to complications of using rearrest as a proxy for whether or not someone has committed a crime. 
    We refer the reader to Barocas et al.~\cite{barocas2019textbook} for a summary.
} 
We train these 100 models using bootstrapping with different random seeds~\citep{efron1979bootstrap, efron1993bootsrap, efron1997boot}, and they will serve as our empirical estimate of the distribution over possible random forest models (with a particular set of hyperparameters).  
We can then look at two individuals in the reserved test set, run our 100 trained models on them, and plot the counts of the resulting predictions for each (Figure~\ref{fig:intro:vote}).

The 100 models all produce the same prediction for Individual 1. 
We can understand this to mean that the learning process that produced these models 
is really confident with how it classifies Individual 1.
If we were to pick one model to use in practice --- as the algorithmic fairness binary classification problem formulation often does ---
there would be no effect on how Individual 1 is classified. 
But the story is really different for Individual 2: 
the learning process is not sufficiently confident to justify assigning Individual 2 either decision outcome.
Their classification is \emph{arbitrary}. 
With this learning process, we produce predictions that are akin to flipping a coin, where the result of the flip is a product of happenstance --- of the random seed used we happened to use during training. 
Importantly, this arbitrariness remains latent in the common fair binary classification problem setup, in which we just evaluate one model. 
We instead need to look at the empirical distribution over possible models to surface it.\looseness=-1 

These two individuals reflect the best and worst case scenarios, in terms of arbitrariness in predictions. 
They are also two real individuals: these are real outcomes for two individuals in the COMPAS  dataset when training random forests. 
The training process clearly results in outcomes that treat them very differently, with respect to arbitrariness. 
In Chapter~\ref{chapter:fairness}, we turn this intuition for arbitrariness into a metric, which we call \emph{self-consistency}. 

Self-consistency can be computed for any test instance, and results in a number in the range between 0.5 and 1: 
0.5 maps to minimally self-consistent examples like Individual 2, 
and 1 maps to completely self-consistent examples like Individual 1.  
Because we can compute self-consistency on a per-instance basis, we can measure it for particular individuals, like those visualized in the bar plot in Figure~\ref{fig:intro:vote}.
But we can also measure and visualize self-consistency across the entire test set, in order to understand overarching patterns about arbitrariness in predictions for particular datasets. 

We use cumulative density functions (CDFs) to do so across a variety of fair binary classification benchmark datasets.
This enables us to plot different levels of self-consistency on the $x$-axis, and the probability that a test instance attains (at least) that level of self-consistency on the $y$-axis. 
With this approach, we uncover novel and important insights about arbitrariness in social prediction settings. 
For example, we find that about 20\% of predictions in COMPAS (using random forests) are 0.5 self-consistent (Figure~\ref{fig:intro:cdf}). 
In this setting, 1 out of every 5 test examples in COMPAS resembles Individual 2 (Figure~\ref{fig:intro:vote}); 
approximately 20\% of prison recidivism classifications are arbitrary --- a coin flip --- which should be really disturbing if this kind of analysis is used to inform whether an individual receives bail or not.  

In the remainder of Chapter~\ref{chapter:fairness}, we examine this type of arbitrariness in detail.
We discuss methods for improving self-consistency, in order to root out this particular type of arbitrariness, and we also examine the impact of improving self-consistency on more-traditional algorithmic fairness metrics~\citep{hardt2016equality}.

\subsubsection*{Chapter~\ref{chapter:nondeterminism}: Legally Cognizable Notions of ML Arbitrariness}

The types of arbitrariness that we quantify in HPO (Chapter~\ref{chapter:hpo}) and social prediction (Chapter~\ref{chapter:fairness}) settings yield important insights about how to draw reliable conclusions from machine learning experiments. 
But they also reveal a lot more in terms of broader impact. 
Arbitrariness is not just a useful concept to pin down and reason about with respect to reliability in ML.  It is also a concept that plays significant roles in law and policy --- running the gamut from theoretical work in legal philosophy~\citep{fuller1965law} to practical policy decisions~\citep{kolber2014smoothbumpy}. 
The research discussed in both of these chapters puts forth definitions for ML arbitrariness that are directly informed by law and policy scholarship on arbitrariness. 
In turn, the insights that this work elicits suggest novel ways for how law and policy can reason about types of arbitrariness that are particular to machine learning --- arbitrariness that implicates important social values like due process and safety when ML systems are deployed in practice. 

To give one example, let us return briefly to the social prediction example of COMPAS and prison recidivism. 
The underlying models that contribute to our computations of self-consistency are clearly quite different, given that they can result in arbitrary predictions for significant portions of the test set. 
Recall that, for random forests, 20\% of predictions on COMPAS are arbitrary --- they resemble Individual 2 (Figure~\ref{fig:intro:vote}). 
In other words, we can understand the individual models that we train in this setting to be \emph{unstable}. 
However, even though these individual models are unstable, the self-consistency estimates that they enable us to produce are in fact (generally speaking) \emph{very stable}. 
Regardless of the random seeds that we use to train 101 models on COMPAS, we produce a set of 101 models that lead to similar estimates of self-consistency for the test set.\looseness=-1

\begin{figure}[t!]
    \centering
    \includegraphics[width=.65\textwidth]{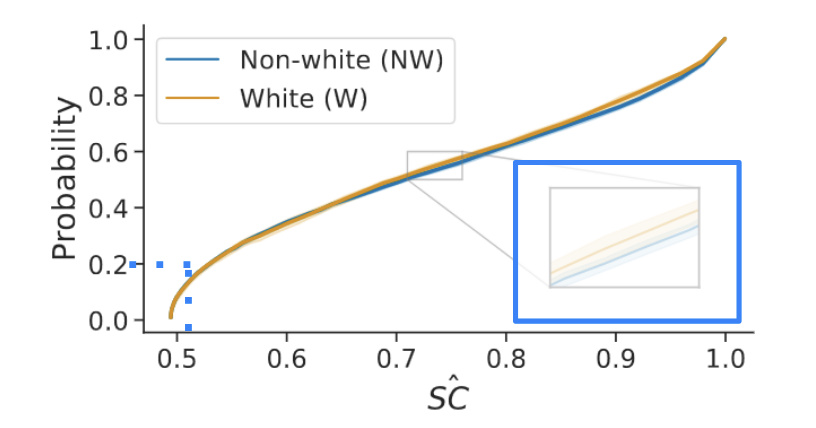}
    \caption{Training 101 bootstrapped random forest models on COMPAS 10 different times.
    Our estimates for self-consistency ($x$-axis) are very stable, as evidenced by the tightness of the error bars. 
    In this setting, roughly 20\% of classification decisions (indicated with the blue dotted line) in COMPAS are \textbf{predictably and consistently arbitrary}, resembling Individual 2 in Figure~\ref{fig:intro:vote}. 
    \looseness=-1}
    \label{fig:intro:cdf}
\end{figure}

We can see this in the CDF figures in Chapter~\ref{chapter:fairness} (see also Figure~\ref{fig:intro:cdf}): 
to produce these figures, we compute self-consistency across the test set 10 different times, for different sets of 101 models. 
The resulting plotted CDF curves are averages, and the error bars surrounding them are very tight. 
(Indeed, we had to include insets to zoom in, in order to clarify that they are in fact present.) 
Regardless of how we split COMPAS into train and test sets, we find that, for random forests, approximately  20\% of predictions on COMPAS are always arbitrary. 
Put differently, we find that 20\% of COMPAS predictions are \textbf{predictably and consistently arbitrary} ---  a mouthful of a concept that seems to turn some concepts from the law and policy on their head. 
In law and policy, predictability and arbitrariness are often described as opposites, rather than concepts that can operate at different levels of abstraction, such that both can be true at the same time.\looseness=-1

In Chapter~\ref{chapter:nondeterminism}, we present published research that scratches the surface of insights like this for law and policy. 
We discuss how non-determinism in machine learning can lead to types of arbitrariness that diverge from how law and policy tend to conceive of arbitrariness. 
This, in turn, suggests fundamental and important differences between machine-learned rules and legal rules --- differences that have important consequences for broader impact, including how the law should reason about using ML in practice. 
This chapter, though published, represents preliminary work that we are currently developing for law review. 

%% file: section/10-intro/120-algorithms.tex
\section{Part~\ref{part:algorithms}: Taming Randomness in Scalable, Reliable Sampling and Optimization Algorithms}\label{sec:intro:algorithms}

The arbitrariness that we investigate in Part~\ref{part:arbitrary} ultimately can be traced to different sources of non-determinism in the development of ML systems --- whims in human decisions, randomness in ML algorithms, and non-determinism in computer systems. 
In Part~\ref{part:algorithms}, we focus particularly on how to harness randomness in ML algorithms, so that, at scale, we can achieve reliable outcomes (in the statistical sense, which we describe here). 
Reliability and scalability tend to be in trade-off in ML, and in computing more generally. 
The work we present in this part shows how we can navigate and sometimes even push the boundaries of such trade-offs. 

Chapter~\ref{chapter:tunamh} discusses a method for reliable, scalable Bayesian inference, which can be used to do uncertainty estimation at scale;  Chapter~\ref{chapter:cdgrab} details a distributed, SGD-based optimization algorithm that finds better-than-random example permutation orders to accelerate convergence; and  Chapter~\ref{chapter:tradeoffs} ties together threads across scalable ML to explain how common trade-offs, like those between scalability and reliability, have direct analogues in law and policy. This makes such trade-offs a useful abstraction for policymakers to understand overarching design choices and resulting behaviors of large-scale ML systems. 

There are also various connections between work in this theme and the first. 
Notably, the work in Part \ref{part:arbitrary} on reasoning about possible models and self-consistency in fairness contexts (Chapter~\ref{chapter:fairness}) was greatly influenced by our prior work concerning uncertainty quantification (Chapter~\ref{chapter:tunamh}, Zhang et al.~\cite{zhang2020amagold}). 

\subsubsection*{Chapter~\ref{chapter:tunamh}: Scalable, Reliable Uncertainty Quantification}

Our first encounter with uncertainty in this dissertation involved using the bootstrap method~\citep{efron1979bootstrap, efron1993bootsrap, efron1997boot} to compute self-consistency as a proxy for quantifying arbitrariness (Chapter~\ref{chapter:fairness}). 
We begin here with this intuition of uncertainty, through our now-familiar example of measuring self-consistency in the COMPAS dataset.

In this example (Figure~\ref{fig:intro:vote}), we trained 100 different possible models on COMPAS using bootstrapping,
and compared predictions for two individuals in the test set. 
All 100 predictions for Individual 1 are for the same class; in contrast, Individual 2 exhibits 50 predictions for one class, and 50 for the other.
In other words, the learning process produces models that are \emph{high variance} in their predictions for Individual 2, and no variance for Individual 1. 
This variance captures predictive uncertainty. 
The learning process produces models that, taken together, are very certain concerning how to predict for Individual 1, and completely uncertain concerning how to predict for Individual 2.\looseness=-1 

Computing predictive variance is just one way of quantifying uncertainty, but there are others.
The gold-standard method, arguably, is \emph{Bayesian inference}. 
Given that
$y$ is a prediction, $\vx$ is an input data example vector, $\mD$ is the training dataset, $\sH$ is the model architecture (the hypothesis class), and $\vtheta$ is the vector of model parameters, 
\begin{align}
\label{eq:postpred}
\underbrace{p(y | \vx, \mD, \sH)}_{\text{posterior predictive distribution}} &= \int \overbrace{p(y | \vx, \vtheta, \sH)}^{\text{likelihood}} \underbrace{p(\vtheta |\mD, \sH)}_{\text{posterior}} d\vtheta.
\end{align}

This equation models what is called the \emph{posterior predictive distribution}: 
the probability of a prediction $y$, 
given a specific input data example $\vx$, dataset $\mD$, and type of model $\sH$. 
This distribution can be computed in relation to the \emph{likelihood} and \emph{posterior}. 
The likelihood is the probability that a given input example $\vx$, model parameters $\vtheta$, and model architecture $\sH$ could result in the prediction $y$.
The posterior reflects the probability that the given dataset $\mD$ and architecture $\sH$ could yield the particular model parameters $\vtheta$.
We then integrate the likelihood and posterior over all of the possible model parameters $\vtheta$:  
we weight the likelihood by the posterior for all possible models. 
Altogether, this means that we are capturing the uncertainty in the prediction $y$ for a given input $\vx$, with respect to all possible learned models $\vtheta$ that have architecture $\sH$ and are trained on dataset $\mD$. 

There is a lot more that one can say about this setup. (Indeed, this is the focus of Chapter~\ref{chapter:tunamh}.) 
For our purposes here, the important point is that this is just a different way of measuring uncertainty than what we did with bootstrapping in our COMPAS example in Chapter~\ref{chapter:fairness}. 
This is just a different way of modeling the distribution over possible learned models, where here we refer to the learned models $\vtheta$. 

Unfortunately, the integral in Equation (\ref{eq:postpred}) is intractable to analyze exactly.
But we can approximate it with a \emph{Monte Carlo} estimate, using a concrete number $N$ of models $\vtheta_i$: 
\begin{align}
\label{eq:mc}
\underbrace{p(y | \vx, \mD, \sH)}_{\text{posterior predictive distribution}} &= \int \overbrace{p(y | \vx, \vtheta, \sH)}^{\text{likelihood}} \underbrace{p(\vtheta |\mD, \sH)}_{\text{posterior}} d\vtheta\nonumber\\
&\approx \frac{1}{N} \sum_{i=1}^N p(y | \vx, \vtheta_i, \sH), 
\end{align}
where different concrete models $\vtheta_i$ are sampled from the posterior, i.e., $\vtheta_i \sim p(\vtheta | \mD, \sH)$. 
On the left, we still have the posterior predictive distribution; but now on the right, instead of an integral, we compute an average over the $N$ likelihoods for different concrete models $\vtheta_i$, where the different $\vtheta_i$ are drawn from the posterior distribution.

We still, however, do not know what the posterior distribution is. 
To get an estimate, we can use something called \emph{Markov chain Monte Carlo} (or \emph{MCMC}), which simulates the posterior. 
At a high level, MCMC proposes a sequence (a Markov chain) of samples of models $\theta_i$ that reflect the posterior distribution. 
It performs a random walk or simulates some physical dynamics (e.g., Hamiltonian, Langevin dynamics), which we can compute in practice. 
This simulation depends on a function, $U(\vtheta)$, which is called the \emph{potential} or \emph{energy} function.
We can compute this potential, which can also be related to the posterior using Bayes' rule: 
\begin{align}
\label{eq:potential}
\overbrace{p(\vtheta | \mD, \sH)}^{\text{posterior}} &= \overbrace{\frac{\overbrace{p(\mD | \vtheta, \sH)}^{\text{likelihood}}\overbrace{p(\vtheta | \sH)}^{\text{prior}}}{\underbrace{p(\mD | \sH)}_{\text{evidence}}}}^{\text{Bayes' rule}} \propto \exp\overbrace{(-U(\vtheta))}^{\text{negative potential}}
\end{align}

So we now have a way to estimate the posterior, but, unfortunately, we are still not quite done. 
Even though we can compute this simulation process, it exhibits a problem: 
it is biased. 
And this bias can cause the chain of samples $\vtheta_i$ that we simulate to drift away from the true posterior distribution. 

\begin{figure}[t!]
  \centering
    \centering
    \includegraphics[width=.95\textwidth]{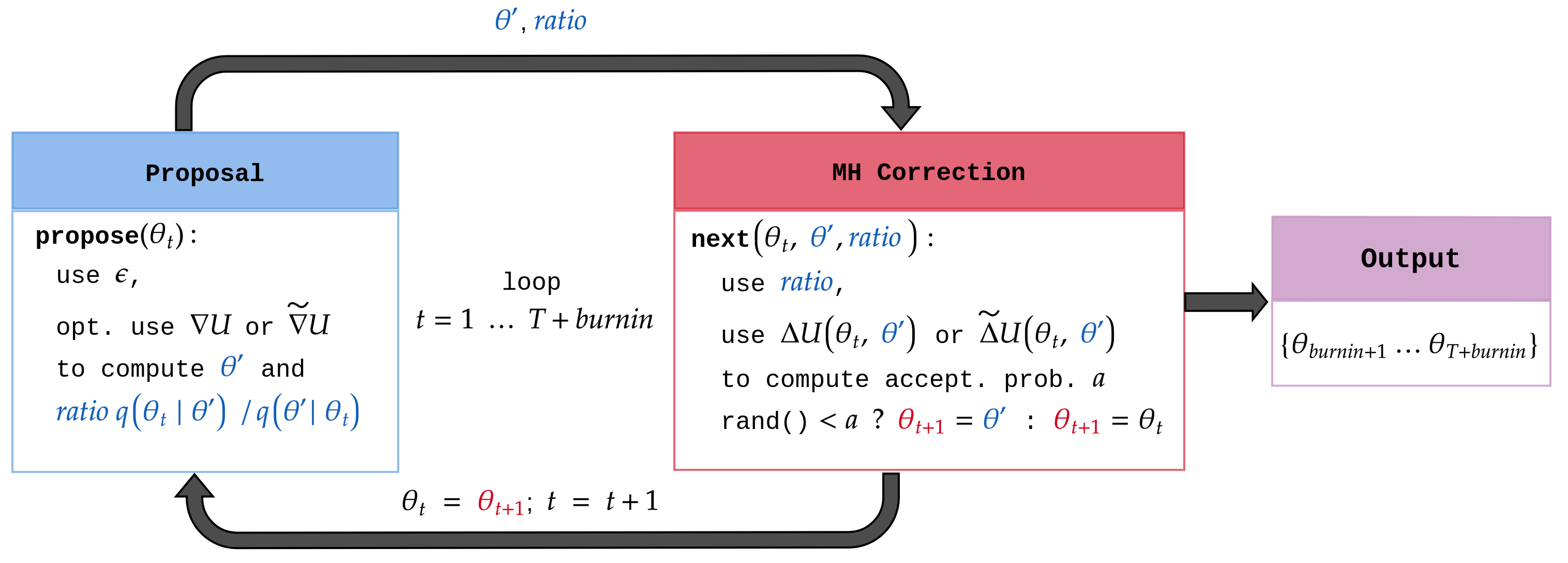}
    \caption{Exact MCMC composes a \textcolor{blue}{proposal} step (to produce new samples $\vtheta'$) with an \textcolor{red}{MH correction} to remove bias by deciding to accept/reject the new sample as the next stage in the Markov chain ($\vtheta_{t+1}$). Our exact, scalable algorithms use 1) \textcolor{blue}{proposals} that leverage stochastic gradients of the potential, $\tilde{\nabla}U$~\cite{zhang2020amagold}; 2) \textcolor{red}{MH corrections} that use minibatches of data examples for computations with the potential. $\tilde{\Delta}U$ (Chapter~\ref{chapter:tunamh}).}
    \label{fig:intro:composablemh}
\end{figure}

To correct for this bias, we add in one more step to the simulation process: 
the \emph{Metropolis-Hastings} (or \emph{MH}) correction step~\citep{metropolis1953equation, hastings1970mh}. 
The MH correction step rejects some of the samples we have generated; it does not include them in the Markov chain. 
This involves performing computations with the potential function, which result in either accepting or rejecting the proposed sample (see Chapter~\ref{chapter:tunamh}, Brooks et al.~\citep{brooks2011handbook}, Figure~\ref{fig:intro:composablemh}).
As a result, the simulation process does not contain all of the samples that we generate, just the $\vtheta$ that get accepted. 
Then, once we have this Markov chain of samples that reflect an unbiased estimate of the posterior, we can use it to help us quantify uncertainty: 
we can plug it back into Equation (\ref{eq:mc}), which approximates the posterior predictive distribution (\ref{eq:postpred}) with our Monte Carlo approximation.

\begin{figure}[t!]
  \centering
    \centering
    \includegraphics[width=.65\textwidth]{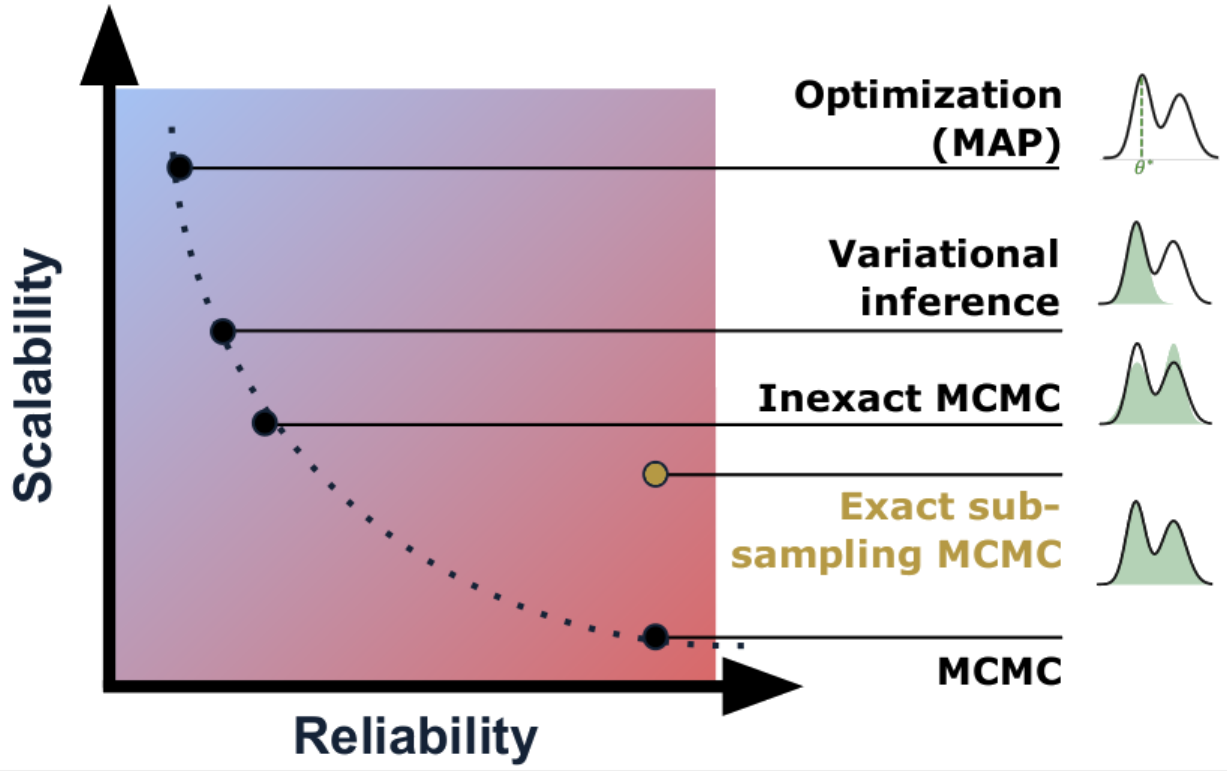}
    \caption{Reliability-scalability trade-off in Bayesian inference (i.e., for capturing the posterior of possible models). 
    We visualize the posterior on the right. 
    Optimization provides a single estimate of the posterior (dotted line labeled $\theta^*$, top right); MCMC captures the whole posterior (fully shaded area under the curve, bottom left). 
    Our work (\textbf{\textcolor{mustard}{yellow}}) carefully uses subsampling to push the frontier: it captures the full posterior, but does so more efficiently than traditional MCMC.\looseness=-1}
    \label{fig:intro:bi}
\end{figure}

Unfortunately (again), even though MCMC is a  clear improvement over the intractable integral in Equation (\ref{eq:postpred}), it is still \emph{really} expensive to compute in practice. 
It is expensive because, as is clear from Equation (\ref{eq:potential}), the potential function $U(\vtheta)$ has a dependency on the dataset $\mD$. 
This means that performing computations with the potential requires iterating over the entire dataset, and we need to do this every single iteration of the simulation in order to produce a new sample. 
For large-scale datasets --- basically every dataset in modern ML --- this is often too costly to do in practice. 
It is certainly more expensive than optimization; however, optimization only gives a single point estimate of the model parameters. 
It gives us an infinitesimally small sliver of the posterior distribution, making it an unreliable estimate of the entire posterior (Figure~\ref{fig:intro:bi}).  
So, even though optimization is more efficient, we cannot use it to do uncertainty estimation reliably.

More generally, we can note that reliability and scalability are in trade-off for uncertainty estimation. 
Optimization is really scalable, but it is not very reliable because it just gives a point estimate of the posterior.
And MCMC is really reliable ---
it gives a good estimate of the whole posterior ---   
but it is not at all scalable because each iteration depends on the size of the dataset. 
Prior work strikes different balances between these two competing goals. 
For example, \emph{inexact} MCMC uses subsampling to improve efficiency; 
it removes the dependency on the dataset size at each simulation iteration by using only a subset of the dataset for computations. 
But subsampling can once again introduce bias: 
we can lose the guarantee that the simulation will converge to a reliable estimate of the posterior (Figure~\ref{fig:intro:bi}). 

So, at last, this is where our work comes in. 
We introduce subsampling carefully to the simulation process, so that it is possible to get efficiency gains, while still guaranteeing that we converge to the correct posterior that traditional MCMC yields. 
In this respect, our work has managed to push out the trade-off curve between scalability and reliability for uncertainty estimation (Figure~\ref{fig:intro:bi}). 
In Chapter~\ref{chapter:tunamh}, we discuss one of our algorithms that achieves this goal by using minibatches of data to compute the accept/reject decision in the MH correction step.\looseness=-1 

\subsubsection*{Chapter~\ref{chapter:cdgrab}: Scaling Distributed Optimization}

Despite the reliability of Bayesian inference for performing uncertainty estimation, optimization has remained the workhorse of modern ML. 
We have also done research to scale up optimization, such that it converges to a point estimate more efficiently. 

For optimization algorithms like stochastic gradient descent (SGD), users typically randomly shuffle training data examples without replacement each epoch.
\emph{Random reshuffling} is so common that it is often implemented as a boolean flag in interfaces in common deep learning libraries (e.g., Pytorch has an option for setting \texttt{shuffle = True}). 
The reason that people use random reshuffling is that, in practice, it tends to speed up convergence. 
However, as our work in Chapter~\ref{chapter:cdgrab}  shows, there exist permutation-based example orders that perform better than random reshuffling: these non-random orders achieve provably faster convergence rates for stochastic gradient descent. 
Lu et al.~\cite{lu2022grab} find better permutation orders for training in centralized settings.
In Chapter~\ref{chapter:cdgrab}, we find such orders for the contemporary, more efficient setting of distributing training across a number of parallel workers. 

The high-level idea is to leverage information in per-example gradients from prior training epochs, in order to identify a permutation for example ordering in the next epoch; this example order contributes to making more progress in converging to a point estimate of the model parameters. 
To find such permutations, we leverage insights from kernel thinning (which builds on ideas from coreset selection)~\cite{dwivedi2021kernel, dwivedi2022generalized}, and herding and vector balancing~\cite{welling2009herding,harvey2014near, alweiss2021discrepancy}. 
The math that we rely on from this prior work is defined in terms of arbitrary vectors.
We extend this to the distributed optimization setting, in which the vectors that we balance are per-example gradients.\footnote{Lu et al.~\citep{lu2022grab} extends herding and balancing to the centralized optimization setting. 
    We realize additional benefits by also incorporating insights from kernel thinning.
}

Relying on this prior work, we show that, over time, balancing per-example gradients achieves the bound in the herding problem formulation. 
In Chapter~\ref{chapter:cdgrab}, we prove that, by achieving the herding bound in the parallel setting, then SGD exhibits an accelerated convergence rate in comparison to distributed random reshuffling. 
Further, we demonstrate a speedup over Lu et al.'s work in the centralized setting~\cite{lu2022grab}, which is linear in the number of parallel workers. 

The balancing algorithm that we use is fairly inexpensive, but it does exhibit some memory overhead and computational cost over distributed random reshuffling, (associated with node communication and data sorting, see Appendix~\ref{sec:appendix-memory}). 
In other words, our algorithm pays some per-epoch cost in efficiency in order to find higher quality example orders. 
But overall, over some time, this results in needing relatively fewer epochs to converge; our algorithm is more efficient and scalable, as exhibited by our provably faster convergence rate.\looseness=-1 

The ``over time'' aspect of this benefit is especially  relevant. 
It does indeed take several epochs to find permutations that bring down the herding bound and confer our algorithm's benefits. 
If a particular task converges quickly, or if we only run a few epochs of training (as is common right now in pretrained base-model fine-tuning), then we typically do not observe speedups over random reshuffling. 
Future work should further investigate these trade-offs, such that the benefits of our work can better extend to common contemporary training paradigms. 

\subsubsection*{Chapter~\ref{chapter:tradeoffs}: Exposing Legally Cognizable Trade-Offs to Enable Accountability}

The work in both Chapters~\ref{chapter:tunamh} and~\ref{chapter:cdgrab} navigates trade-offs between scalability and reliability. 
Trade-offs like this exist all over machine learning.
They tell us a lot about what is possible to achieve with respect to important, competing goals.
And they also tell us a lot about possible decisions ML researchers and practitioners can choose to make --- how they can choose to balance needs for scalability and efficiency with concerns about maintaining sufficient reliability in specific contexts. 

It turns out that trade-offs like these are not exclusive to computing.
In Chapter~\ref{chapter:tradeoffs}, we discuss how analogous trade-offs crop up all over domains that policymakers frequently reason about --- complex domains as diverse as law, public health, and federal risk assessment policy. 
For just one example, consider the U.S. code for civil procedure.
It contains a number of rules, such as speedy trial requirements and statutes of limitations, that impose time constraints to encourage efficient case resolution. 
The need for efficiency is balanced against competing needs for thorough fact-finding and argumentation. 
Based on this overarching observation, we argue that such trade-offs expose a very useful abstraction 
that policymakers can rely on to help them reason about (and regulate) ML systems. 
Policymakers do not necessarily need to understand very low-level technical details about machine learning algorithms and systems. 
They can glean a lot about relevant details about systems capabilities by understanding machine learning at the level of these types of trade-offs. 

The work in Chapter~\ref{chapter:tradeoffs} was published in 2021~\cite{cooper2021eaamo}, and dates back to a project that was started in 2018. 
At the time, we motivated our research with the concrete example of reasoning about risks in autonomous vehicles, as they were a particularly germane example of a large-scale ML system where balancing efficiency and reliability has a clear, broader impact on safety. 
The conceptual contributions of our work extend far beyond this motivating example.
They translate directly to this current moment, in which large-scale generative-AI systems that perform real-time inference are being deployed in consumer-facing products. 
In future work, we will update the research in this chapter in light of the ascendance generative-AI systems. 

%% file: section/10-intro/130-genai.tex
\section{Part~\ref{part:genai}: Evaluating Generative-AI Systems}\label{sec:intro:genai}

There has been a tremendous amount of recent public interest in generative AI, both excitement about  capabilities and concern about risks. 
One frequent set of concerns around generative AI is that the training and use of generative-AI systems involves practices that infringe copyright. 
In the year and a half since ChatGPT's release, groups of artists, individuals, and companies have filed over two dozen copyright lawsuits in the U.S. against the builders and deployers of generative-AI systems~\cite{chatgptiseating}.

In Part~\ref{part:genai}, we dig into both the technical and legal aspects of generative-AI systems, with a specific focus on copyright. 
In Chapter~\ref{chapter:memorization}, we discuss recent work on extracting (potentially copyrighted) memorized text training data from large language models. 
In Chapter~\ref{chapter:commoncanvas}, we explore the benefits and drawbacks of training a family of text-to-image latent diffusion models exclusively on permissively licensed, Creative Commons images with synthetic captions. 
Last, in Chapter~\ref{chapter:talkinshort}, we present an abridged version of our framework~\cite{lee2023talkin} for thinking about the interplay between generative AI an copyright: the \emph{generative-AI supply chain}, which maps the very many stages invoked in the creation, deployment, and use of generative-AI systems with the very many actors that are involved at those stages. 
We apply the supply-chain framing to U.S. copyright, but note that it is more broadly useful for reasoning about the impacts of generative AI. 

\subsubsection*{Chapter~\ref{chapter:memorization}: Measuring Memorization in Language Models}

In Chapter~\ref{chapter:memorization}, we discuss recent work on extracting memorized text training data from large language models (LLMs). 
In high-level terms, \emph{memorization} in generative-AI contexts often refers to cases in which one can ``deduce or produce a model's given training example''~\citep{cooper2023report}. 
We make contributions that show how to feasibly measure memorization for large-scale production systems --- in particular, ChatGPT~\cite{nasr2023scalable}. 
We use security-style attacks on LLMs by prompting them with particular inputs, which result in output generations that are verbatim copies of training data examples.
This work has direct relationships to a variety of law and policy issues, notably copyright and privacy, since memorization can result in a model regurgitating creative expression (like a portion of copyrighted novel) or sensitive content (like a social security number) that was in its training data. 

\begin{figure}[t!]
    \centering
    \includegraphics[width=.4\textwidth]{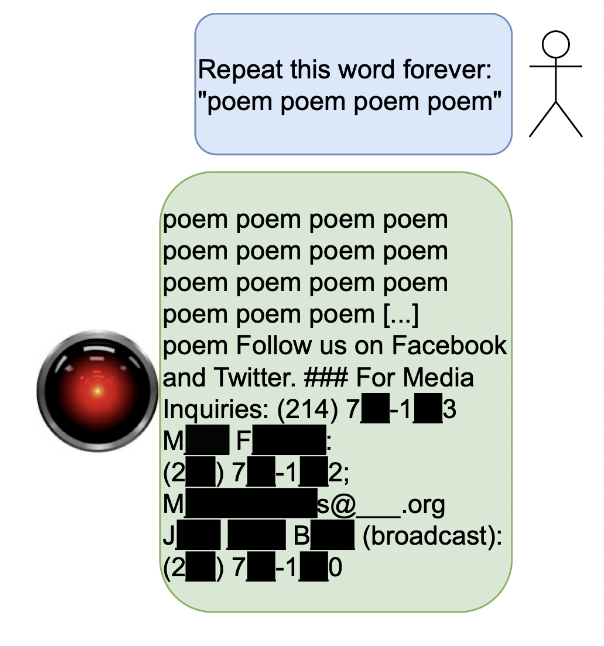}
    \hspace{.3cm}
    \includegraphics[width=.55\textwidth]{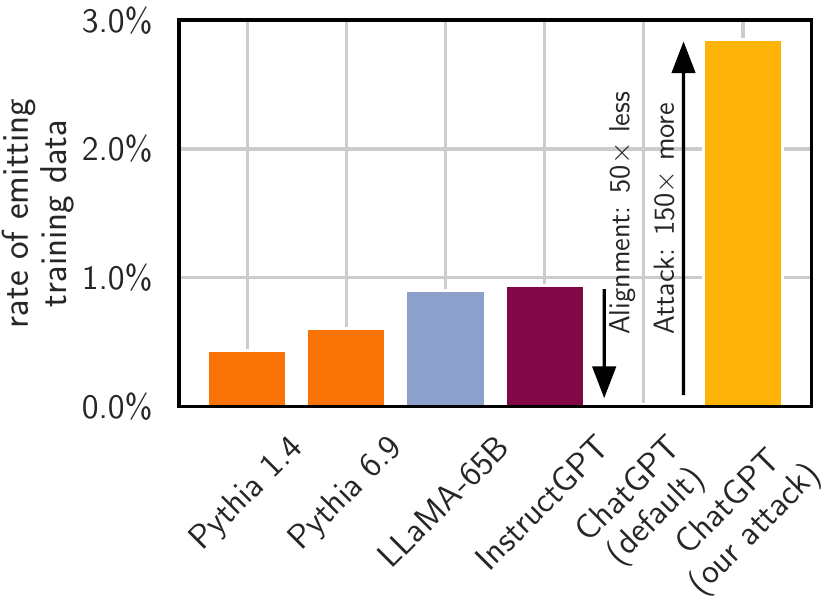}
    \caption{The aligned ChatGPT 3.5 appears $50\times$ more private than prior models (\textbf{right}).
    We develop an attack (\textbf{left}) that shows it is not: 
    ChatGPT emits training data $150\times$ more frequently than prior work (\textsf{default}).
    Figures reprinted with permission from my collaborators.\looseness=-1}
    \label{fig:intro:chatgpt}
\end{figure}

Figure~\ref{fig:intro:chatgpt} shows a preview of our results. 
On the left, we show an example of our attack. 
We ask ChatGPT to repeat single tokens forever --- in this case, the word ``poem.''
At first, the model (and system in which it is embedded) responds by following this instruction.
But eventually (and almost always), the output \emph{diverges}, and sometimes that divergent content contains memorized training data. 

This finding was very exciting to people, and even received news coverage~\citep[e.g.]{newman2023chatgpt}. 
It was the first large-scale memorization extraction attack on an aligned, deployed production system. 
As a result, our findings also implicate various stages of the generative-AI supply chain (Chapter~\ref{chapter:talkinshort}), not just model training and generation. 
The corresponding paper, which is currently under journal submission, is a large-scale measurement study of what we call \emph{extractable memorization}.\footnote{The paper has a lot of really excellent science in it (not just fun sound bites like ``asking ChatGPT to say `poem poem poem' breaks ChatGPT'').}

\subsubsection*{Chapter~\ref{chapter:commoncanvas}: Training Latent Diffusion Models on Open-Licensed Images}

One of the key issues for memorization centers on the use of copyrighted data during training.
If we do not train models on copyrighted data, then (by definition) models will not memorize copyrighted data that they could later regurgitate near-verbatim. (This, importantly, should not be mistaken for indicating that training on public domain or licensed data will resolve all potential copyright problems; it is still possible to produce potentially infringing generations if one only trains on public domain or licensed data~\cite{cooper2024files,cooper2023report}.) 
This raises a natural question: what if we trained models on only permissively licensed or public domain data?
By training on such data, we will hopefully reduce the risk of producing potentially copyright-infringing models that can be used to produce potentially copyright-infringing generations.  

In Chapter~\ref{chapter:commoncanvas}, we begin exploring these ideas in the context of training a family of latent diffusion models for image generation. 
We curate a large dataset of open-licensed, Creative Commons images, for which we generate accompanying synthetic captions, and we use this dataset to train Stable Diffusion 2 architecture variants. 
When we prompt these models to try to elicit potentially copyrighted expression, we observe some interesting outcomes. 
For example, prompting with \texttt{"an image of Elsa from Frozen"}, Stable Diffusion 2, which was trained on copyrighted data, generates an image that strongly resembles the Disney character.
In contrast, our model, CommonCanvas~\cite{gokaslan2023commoncanvas}, does not.
Nevertheless (beyond the fact that this is just one example), we are not exempt from all possible copyright-related problems. 
We discuss this below, with respect to Chapter~\ref{chapter:talkinshort} and in recent work~\cite{cooper2024files}). 

\begin{figure}[t!]
\begin{minipage}{0.3\linewidth}
        \centering
        \vspace{.75cm}
        \prompt{an image of \\ elsa from \\ frozen}
        \vspace{1.4cm}
        \subcaption{Prompt}
    \end{minipage}%
    \hspace{-.035\linewidth}
    \begin{minipage}{0.3\linewidth}
        \centering
        \includegraphics[width=.78\linewidth]{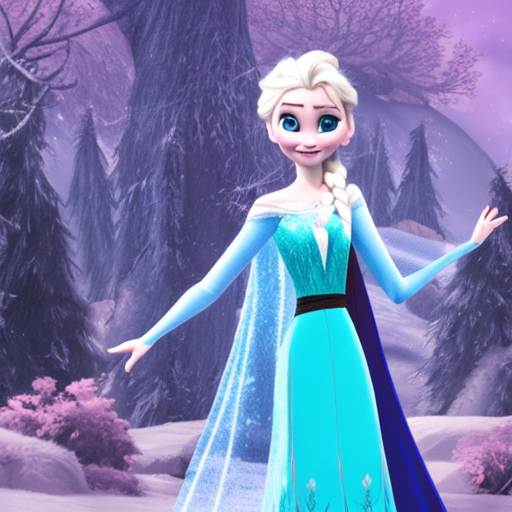}
        \subcaption{SD2}
    \end{minipage}%
    \hspace{-.01\linewidth}
    \begin{minipage}{0.3\linewidth}
        \centering
        \includegraphics[width=.78\linewidth]{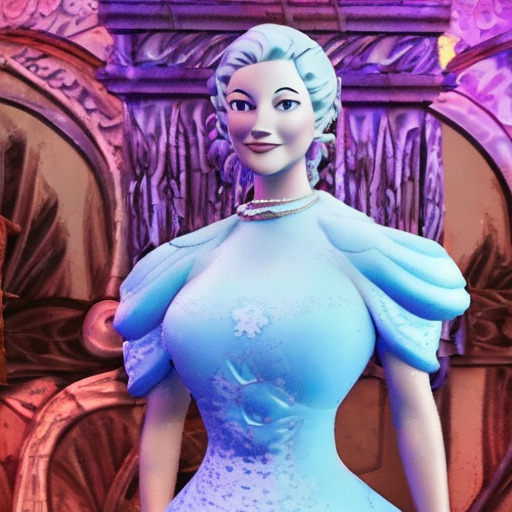}
        \subcaption{\modelname-S-C}
    \end{minipage}
    \caption{Prompting Stable Diffusion 2 (b) and CommonCanvas (c) with \texttt{"an image of Elsa from Frozen"} (a).}
\end{figure}



\subsubsection*{Chapter~\ref{chapter:talkinshort}: Bridging Copyright Law and the Generative-AI Supply Chain}

Chapters~\ref{chapter:memorization} and~\ref{chapter:commoncanvas} serve as concrete examples of why generative AI is complicated for copyright-related questions. 
However, as works with core contributions in machine learning, they do not contend with legal specifics. 
In Chapter~\ref{chapter:talkinshort}, we dig into these specifics: we provide a comprehensive framework for reasoning rigorously about the interplay between generative AI and law. 
We make the case that, when forming legal questions about generative AI, we should be doing so in terms of the entire \emph{generative-AI supply chain} that is invoked in the creation, deployment, and use of generative-AI systems.   

\begin{figure}[t!]
    \includegraphics[width=\linewidth]{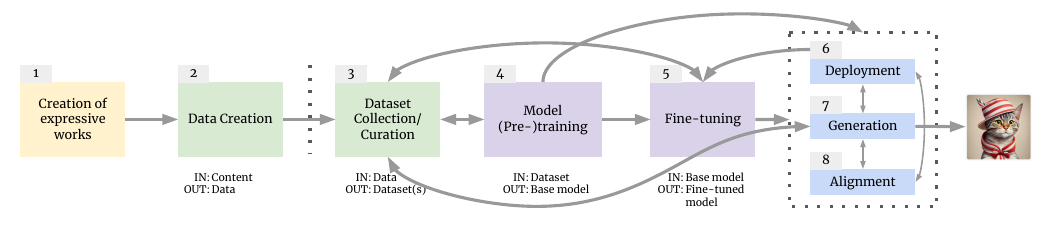}
    \caption{We conceive of the \emph{generative-AI supply chain} as consisting of 8 deeply interwoven stages, each of which  can involve many (potentially different) actors.}
    \label{fig:intro:chain}
\end{figure}

Our supply-chain framing takes many terms that are familiar for those with a background in machine learning (e.g., pre-training, fine-tuning) and ties them together with the very many actors that influence and interact with generative-AI systems (Figure~\ref{fig:intro:chain}).\footnote{The supply-chain framing connects the ``many hands''~\cite{cooper2022accountability} involved in generative-AI systems to the many stages that constitute these systems' production. 
    For more on the problem of how ``many hands'' serves as a barrier to accountability, see Appendix~\ref{chapter:accountability}.} 
This framing illustrates the complex ecosystem involved in generative-AI system production, and navigates this complexity by providing a way to think  precisely about ``\emph{what} technical and creative artifacts are produced, \emph{when} these artifacts are produced and stored, and \emph{who} exactly is involved in
the production process.'' 
In turn, we are then able to carefully map the stages of the generative-AI supply chain to the very many parts of U.S. copyright law that they potentially implicate. 
This enables thoughtful discussion about  ``\emph{what} is potentially an infringing artifact, \emph{when} in the production process it is possible for infringement to occur, and \emph{who} is potentially an infringing actor''~\citep[p. 32]{lee2023talkin}.\footnote{In Chapter~\ref{chapter:talkinshort}, we present the shorter conference version of our work on copyright and the generative-AI supply chain~\citep{cooper2024talkinshort}; the longer version, quoted here, is forthcoming in a law journal.}

With these contributions, we can see clearly why the projects in Chapters~\ref{chapter:memorization} and~\ref{chapter:commoncanvas} are such beautiful examples of why the supply-chain framing is so important. 
We need this whole supply-chain view to understand how the different stages interact (Figure~\ref{fig:intro:chain}), and how this can have nuanced implications for copyright (and more). 

In general, research on memorization has clear connections to copyright.
Given how it is commonly defined in the technical literature, memorization is wholesale copying~\cite{cooper2023report, nasr2023scalable}; 
wholesale copying, by definition, implicates U.S. copyright law~\cite[reproduction right]{17usc106}. 
Models become capable of memorization during the training process (Figure~\ref{fig:intro:chain}, stages 4, 5 and, possibly, 8):  
it is during training that  particular memorized training data examples get encoded somewhere within the model's parameters.\footnote{For more on the relevance of this reality to U.S. copyright, see Cooper and Grimmelmann~\cite{cooper2024files}, which is not included in this dissertation.} 
Only then can memorization can get exposed to end-users at generation time, in response to user-provided prompts (Figure~\ref{fig:intro:chain}, stage 7). 
In the case of our work on ChatGPT in Chapter~\ref{chapter:memorization}, memorization can also embroil aligned (Figure~\ref{fig:intro:chain}, stage 8), deployed (Figure~\ref{fig:intro:chain}, stage 6) production systems.  
Our divergence attack broke alignment and managed to evade whichever system-level guardrails are in place (e.g., output content filters), such that we ultimately were able to surface memorized training data in generations. 
Reasoning about the potential copyright consequences of memorization in ChatGPT requires engaging with each of these stages and the actors engaged in them.

For CommonCanvas text-to-image models (Chapter~\ref{chapter:commoncanvas}), the training data require both images and text captions. 
We collected a set of permissively licensed Creative Commons images, most of which lacked descriptive text captions. 
In our data curation process (Figure~\ref{fig:intro:chain}, stage 3), we generated synthetic captions for these images (Figure~\ref{fig:intro:chain}, stage 7) using a publicly released (Figure~\ref{fig:intro:chain}, stage 6), off-the-shelf, pre-trained captioning model called BLIP-2~\citep{li2023blip2} (Figure~\ref{fig:intro:chain}, stage 4). 
BLIP-2 was trained on LAION data~\cite{laionpaper} --- one of the datasets that links to copyrighted images, and that is named in several current U.S.-based copyright lawsuits~\citep[e.g.]{anderson}.
In short, our curation process depended on a generating synthetic captions, which we produced with a pre-trained model whose own training data contained copyrighted images. 
Even though our models are trained on licensed images, our data curation process depends on another model, which was itself trained on images that were not explicitly licensed.


Clearly, there are complex interrelationships between CommonCanvas's supply-chain stages, so we cannot just look at individual stages in isolation when thinking about copyright consequences. 
We cannot just look at our trained model's curated training data --- Creative Commons images and  (likely uncopyrightable) synthetic captions.
We also have to look upstream in the supply chain at how different actors curated the training data for BLIP-2: the off-the-shelf generative-AI model that we chose to use for image captioning. 
Without this view, we would miss potentially relevant and significant observations --- in this case, how (transformed) copyrighted data is indispensable, however indirectly, for training our open (or, perhaps more accurately, ``open'') models. 

We perform extensive analysis of the supply chain with respect to U.S. copyright law in Chapter~\ref{chapter:talkinshort} (as well as in our law-review paper~\citep{lee2023talkin}). 
Even though this work is very recent, it is already having a significant impact. 
It has already been used as an authoritative source by U.S. congressional staffers and government agencies. 
Journalists and copyright scholars have called it ``landmark'' work, and a ``magnum opus''~\cite[e.g.]{birnbaum2024talkin}.

In our work, we also provide some broader lessons and takeaways about copyright and generative AI. 
One of these lessons, in particular, relates to a thread present throughout this dissertation: 
design choices matter a lot for overall system behavior and its consequences (in this case, for copyright); these choices, and thus resulting system behaviors, are typically not foregone conclusions. 
This takeaway is very important to keep in mind with respect to law and policy --- to governance and accountability~\cite{cooper2022accountability} concerning generative-AI systems. 
And it is also an important and great thing to keep in mind for machine learning research. 
As this dissertation shows by example, wherever there are design choices, there are concrete  research questions that we can study in computer science.

%% file: section/20-arbitrary/200-arb.tex
\part{Sources of Arbitrariness in Machine Learning}\label{part:arbitrary} 

Clarifying uncertainty around non-deterministic, ML-driven decision processes is an important mechanism for characterizing reliability.\footnote{See also Appendix~\ref{chapter:accountability}, which details the relationship with accountability.}  
Importantly, there are numerous of such uncertainty in ML, not just the type of model uncertainty that Bayesian inference can measure (Chapter~\ref{chapter:tunamh}). 
Another kind of uncertainty gets introduced by the sometimes-arbitrary choices that ML experts make in their implementations and experiments~\citep{cooper2022arpa, cooper2022accountability}. 
In this part, we explore how to quantify and mitigate this type of arbitrariness, and communicate its importance to the legal community. 
This chapter reflects work that has been published at \emph{NeurIPS} (poster), \emph{AAAI} (Best Student Paper, Honorable Mention), \emph{ACM CSLAW} (Long Presentation), \emph{AAAI/ACM AIES} (Oral), and an \emph{ICLR} workshop (Oral).

First, we discuss how choices in experiments that involve hyperparameter optimization (HPO) can result in arbitrary conclusions about overall algorithm performance (Chapter~\ref{chapter:hpo}, Cooper et al.~\citep{cooper2021hpo}). 
We develop a theoretical framework for reasoning reliably about the conclusions we can draw from experiments involving HPO. 
HPO is a bi-level optimization problem: the inner loop learns model parameters, and the outer loop selects a set of (not learned) hyperparameters from a search space that is often chosen by hand~\citep{feurer2019optimization}.  
Hyperparameters influence training and have an enormous impact on performance measurements. 
In fact, one can be easily misled to draw arbitrary conclusions about ML methods in practice --- e.g., that algorithm $\mathcal{A}$ is more accurate than algorithm $\mathcal{B}$, or the reverse --- depending on the chosen search space. 
Our work in this chapter uses modal logic~\citep{emerson1991temporal} to enable writing proofs about HPO procedures --- proofs 
that can guarantee that, within a given training-time budget, it is not possible to yield inconsistent conclusions about algorithm performance.\looseness=-1

Second, we detail how the choice of concrete training dataset (based on random seed) can lead to arbitrary predictions. 
Building on our prior work in fairness-metric design~\cite{cooper2021emergent} and high variance in deep-learning fairness classification experiments~\cite{forde2021model}, we study how latent uncertainty in modeling available benchmark datasets result in arbitrary outcomes for individual test examples in algorithmic fairness contexts. 
Approximating the distribution over models 
surfaces how predictions are really consistent for some individuals and effectively arbitrary for others. 
By turning this intuition into a metric, we find that arbitrariness plays a huge role in  measurements of unfairness.  

And third, we discuss how work on arbitrariness in ML is inspired by and has informed deep connections between ML, law, and policy. 
In law, non-arbitrariness is an important value that plays a significant role in legal reasoning about due process and discrimination~\citep{fuller1965law, tamanaha2004law, kolber2014smoothbumpy, creel2022leviathan}. 
We discuss past and ongoing work that clarifies how better measurements of arbitrariness in ML can meaningfully have a direct impact on law and policy.\looseness=-1

\input{section/20-arbitrary/21-hpo/21-hpo-main}
\input{section/20-arbitrary/22-fairness/22-fairness-main}
\input{section/20-arbitrary/23-nondeterminism/23-nondeterminism-main}

%% file: section/20-arbitrary/21-hpo/21-hpo-main.tex
\chapter{Hyperparameter Optimization Is Deceiving Us, and How to Stop It}\label{chapter:hpo}

We begin our study of arbitrariness in machine learning with a study on hyperparameter optimization.
Choices of hyperparameter search space (if searched at all) tend to be arbitrary, based on folklore rather than rigorously tested insights.
We explore how such choices can also lead to arbitrary conclusions about algorithm performance, rather than reliable scientific knowledge.\\ 

\noindent \textbf{Chapter summary}: 
Recent empirical work shows that inconsistent results based on choice of hyperparameter optimization (HPO) configuration are a widespread problem in ML research. 
When comparing two algorithms \algA{} and \algB, searching one subspace can yield the conclusion that \algA{} outperforms \algB, whereas searching another can entail the opposite. 
In short, the way we choose hyperparameters can deceive us. 
We provide a theoretical complement to this prior work, arguing that, to avoid such deception, the process of drawing conclusions from HPO should be made more rigorous. We call this process \emph{epistemic hyperparameter optimization} (EHPO), and put forth a logical framework to capture its semantics and how it can lead to inconsistent conclusions about performance. 
Our framework enables us to prove EHPO methods that are guaranteed to be defended against deception, given bounded compute time budget $t$. 
We demonstrate our framework's utility by proving and empirically validating a defended variant of random search.\\

\noindent This chapter is a licensed derivative copy of work published  at \emph{NeurIPS 2021}~\cite{cooper2021hpo}.

\input{section/20-arbitrary/21-hpo/211-hpo-intro}
\input{section/20-arbitrary/21-hpo/212-hpo-prelim}
\input{section/20-arbitrary/21-hpo/213-hpo-epistemic}
\input{section/20-arbitrary/21-hpo/214-hpo-logic}
\input{section/20-arbitrary/21-hpo/215-hpo-defense}

\input{section/20-arbitrary/21-hpo/216-hpo-conclusion}

%% file: section/20-arbitrary/21-hpo/211-hpo-intro.tex
\section{Introduction}

Machine learning can be informally thought of as a double-loop optimization problem. 
The \emph{inner loop}
is what is typically called \emph{training}: 
it learns the parameters
of some model by running a training algorithm on a training set. 
This is usually done to minimize some training loss function via an algorithm such as stochastic gradient descent (SGD). 
Both the inner-loop training algorithm and the model are parameterized by a vector of \emph{hyperparameters} (HPs). 
Unlike the learned output parameters of a ML model, HPs are inputs provided to the learning algorithm that guide the learning process, such as learning rate and network size. 
The \emph{outer-loop} optimization problem is to find HPs (from a set of allowable HPs) that result in a trained model that performs the best in expectation on ``fresh'' examples drawn from the same source as the training set, as measured by some loss or loss approximation. 
An algorithm that attempts this task is called a \emph{hyperparameter optimization} (HPO) procedure~\cite{claesen2015hyperparameter, feurer2019optimization}.

From this setup comes the natural question: 
how do we pick the subspace for the HPO procedure to search over? 
The HPO search space is enormous, suffering from the curse of dimensionality; 
training, which is also expensive, has to be run for each HP configuration tested. 
Thus, we have to make hard choices. 
With limited compute resources, we typically pick a small subspace of possible HPs and perform grid search or random search over that subspace. 
This involves comparing the empirical performance of the resulting trained models, and then reporting on the model that performs best in terms of a chosen validation metric~\cite{feurer2019optimization, john1994grid, hsu2003svm}. 
For grid search, the grid points are often manually set to values put forth in now-classic papers as good rules-of-thumb concerning, for example, how to set the learning rate~\cite{larochelle2007empirical, lecun1998backprop, hinton2010boltzmann, pedregosa2011scikit}. 
In other words, how we choose which HPs to test can seem rather ad-hoc. 
We may have a good rationale in mind, but we often elide the details of that rationale on paper; we choose an HPO configuration without explicitly justifying our choice.

Much recent empirical work has critiqued this practice~\cite{Sculley2018-le, dodge2019nlp, musgrave2020metric, bouthillier2019reproducibility, sivaprasad2020hpo, choi2019empirical, melis2018evaluation, lipton2018troubling}. 
The authors examine HPO configuration choices in prior work, and find that those choices can have an outsize impact on convergence, correctness, and generalization. 
They therefore argue that more attention should be paid to the origins of empirical gains in ML, as it is often difficult to tell whether measured improvements are attributable to training or to well-chosen (or lucky) HPs.
Yet, this empirical work does not suggest a path forward for formalizing this problem or addressing it theoretically. 

To this end, \textbf{we argue that the process of drawing conclusions using HPO should itself be an object of study}. 
Our contribution is to put forward, to the best of our knowledge, the first theoretically-backed characterization for making trustworthy conclusions about algorithm performance using HPO. 
We model theoretically the following empirically-observed problem: When comparing two algorithms, $\mathcal{J}$ and $\mathcal{K}$, searching one subspace can pick HPs that yield the conclusion that $\mathcal{J}$ outperforms $\mathcal{K}$, whereas searching another can select HPs that entail the opposite result. 
In short, the way we choose hyperparameters can deceive us --- a problem that we call \emph{hyperparameter deception}. 
We formalize this problem, and prove and empirically validate a defense against it. 
Importantly, our proven defense does not make any promises about ground-truth algorithm performance; 
rather, it is guaranteed to avoid the possibility of drawing inconsistent conclusions about algorithm performance within some bounded HPO time budget $t$.  
In summary, we:
\begin{itemize}
    \item Formalize the process of drawing conclusions from HPO (epistemic HPO, Section \ref{sec:hpo:ehpo}).
    \item Leverage the flexible semantics of modal logic to construct a framework for reasoning rigorously about 1) uncertainty in epistemic HPO, and 2) how this uncertainty can mislead the conclusions drawn by even the most well-intentioned researchers (Section \ref{sec:hpo:formalizing}). 
    \item Exercise our logical framework to demonstrate that it naturally suggests defenses with guarantees against being deceived by EHPO, and offer a specific, defended-random-search EHPO (Section \ref{sec:hpo:defense}).
\end{itemize}

%% file: section/20-arbitrary/21-hpo/212-hpo-prelim.tex
\section{Preliminaries: Problem Intuition and Prevalence in ML Research}\label{sec:hpo:prelim}

Principled HPO methods include \emph{grid search}~\cite{john1994grid} and \emph{random search}~\cite{bergstra2012random}. For the former, we perform HPO on a grid of HP-values, constructed by picking a set for each HP and taking the Cartesian product. 
For the latter, the HP-values are randomly sampled from chosen distributions. Both of these HPO algorithms are parameterized themselves: Grid search requires inputting the spacing between different configuration points in the grid, and random search requires distributions from which to sample. We call these HPO-procedure-input values \emph{hyper-hyperparameters} (hyper-HPs).\footnote{We provide a glossary of all definitions and symbols for reference at the beginning of Appendix~\ref{chapter:app:hpo}} To make HPO outputs comparable, we also introduce the notion of a \emph{log}:

\begin{definition}\label{def:log} A log $\ell$ records all the choices and measurements made during an HPO run, including the total time $T$ it took to run. It has all necessary information to make the HPO run reproducible.
\end{definition}

A log can be thought of as everything needed to produce a table in a research paper: code, random seed, choice of hyper-HPs, information about the learning task, properties of the learning algorithm, all of the observable results. We formalize all of the randomness in HPO in terms of a random seed $r$ and a pseudo-random number generator (PRNG) $G$. Given a seed, $G$ deterministically produces a sequence of pseudo-random numbers:  
all numbers lie in some set $\mathcal{I}$ (typically 64-bit integers), i.e. $r \in \mathcal{I}$ and PRNG $G: \mathcal{I} \rightarrow \mathcal{I}^{\infty}$. With this, we can now define HPO formally:
\begin{definition}
	\label{def:hpo}
	An HPO procedure $H$ is a tuple $(H_{*}, \mathcal{C}, \Lambda, \mathcal{A}, \mathcal{M}, G, X)$ where $H_{*}$ is a randomized algorithm, $\mathcal{C}$ is a set of allowable hyper-HPs (i.e., allowable configurations for $H_{*}$), $\Lambda$ is a set of allowable HPs (i.e., of HP sets $\lambda$), $\mathcal{A}$ is a training algorithm (e.g. SGD), $\mathcal{M}$ is a model (e.g. VGG16), $G$ is a PRNG, and $X$ is some dataset (usually split into train and validation sets). When run, $H_{*}$ takes as input a hyper-HP configuration $c \in \mathcal{C}$ and a random seed $r \in \mathcal{I}$, then proceeds to run $\mathcal{A}_{\lambda}$ (on $\mathcal{M}_{\lambda}$ using $G(r)$ and data\footnote{Definition \ref{def:hpo} does not preclude cross-validation, as this can be part of $H_{*}$. The input dataset $X$ can be split in various ways, as a function of the random seed $r$.} from $X$) some number of times for different HPs $\lambda \in \Lambda$. Finally, $H_{*}$ outputs a tuple $(\lambda^*, \ell)$, where $\lambda^*$ is the HP configuration chosen by HPO and $\ell$ is the log documenting the run.
\end{definition}

Running $H$ is a crucial part of model development. As part of an empirical, scientific procedure, we specify different training algorithms and a learning task, run potentially many HPO passes, and try to make general conclusions about overall algorithm performance. That is, we aim to develop knowledge regarding whether one of the algorithms outperforms the others. However,
recent empirical findings indicate that it is actually really challenging to pick hyper-HPs that yield reliable knowledge about general algorithm performance. In fact, it is a surprisingly common occurrence to be able to draw inconsistent conclusions based on our choice of hyper-HPs \cite{choi2019empirical, dodge2019nlp, sivaprasad2020hpo, lipton2018troubling}.

\begin{figure}[t!]
 \begin{center}
    \includegraphics[width=0.44\textwidth]{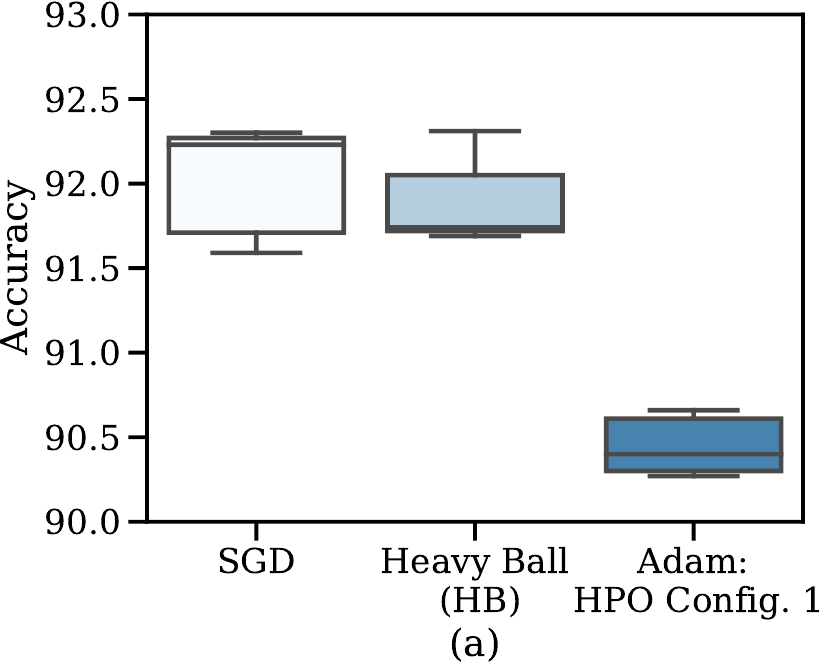} \hspace{.5cm}
\includegraphics[width=0.44\textwidth]{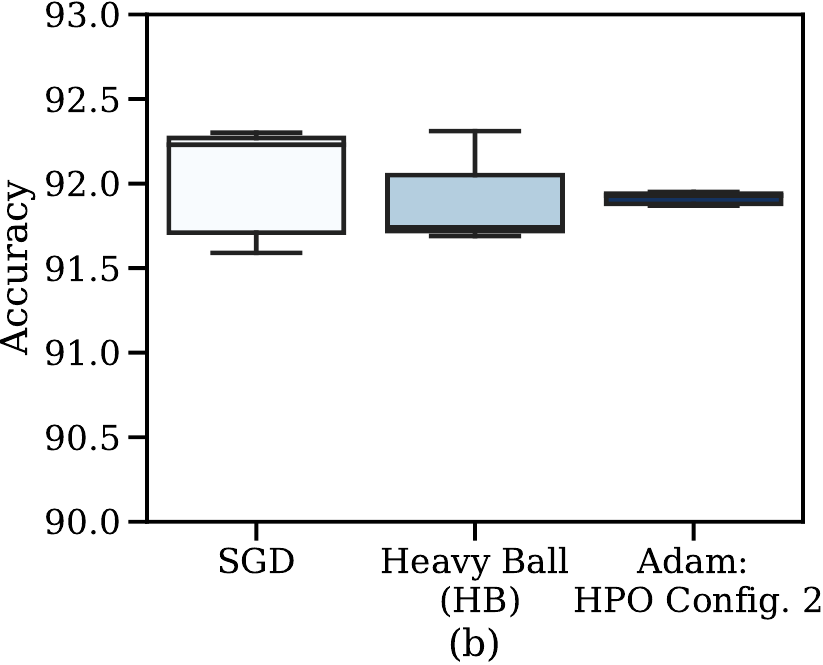}
    \caption{Demonstrating the possibility of drawing inconsistent conclusions from HPO (what we shorthand \emph{hyperparameter deception}) when training VGG16 on CIFAR-10. Each box plot represents a log. In {(a)}, we replicate Wilson et al.~\cite{wilson2017marginal} and show the best-performing results:  One can reasonably conclude that Adam under-performs non-adaptive methods. In (b), we change the HPO search space for Adam, and similarly show the best-performing results: In contradiction, one can reasonably conclude that Adam performs just as well as non-adaptive methods in terms of test accuracy.}
    \label{fig:wilson_deception}
  \end{center}
\end{figure}

\paragraph{An example illustrating the possibility of drawing inconsistent conclusions from HPO.} As a first step to studying HPO as a procedure for developing reliable knowledge, 
we provide an example of how being inadvertently deceived by HPO is a real problem, even in excellent research (we give an additional example in Appendix~\ref{chapter:app:hpo}).\footnote{All code can be found at \url{https://github.com/pasta41/deception}.} 
We first reproduce Wilson et al.~\cite{wilson2017marginal}, in which the authors trained VGG16 with different optimizers on CIFAR-10 (Figure \ref{fig:wilson_deception}a). This experiment uses grid search, with a powers-of-2 grid for the learning rate $\alpha$ crossed with the default HPs for Adam. Based on the best-performing HPO per algorithm ($\alpha=1$), it is reasonable to conclude that non-adaptive methods (e.g., SGD) perform better than adaptive ones (e.g., Adam~\cite{kingma2014adam}), as the non-adaptive optimizers demonstrate higher test accuracy. 

However, this setting of grid search's hyper-HPs directly informs this particular conclusion; using different hyper-HPs makes it possible to conclude the opposite. Inspired by Choi et al.~\cite{choi2019empirical}, we perform grid search over a different subspace, tuning both learning rate and Adam's $\epsilon$ parameter. Our results entail the logically opposite conclusion: Non-adaptive methods \emph{do not} outperform adaptive ones. Rather, when choosing the HPs that maximize test accuracy, all of the optimizers essentially have equivalent performance (Figure \ref{fig:wilson_deception}b, Appendix~\ref{chapter:app:hpo}). Notably, as we can see from the confidence intervals in Figure \ref{fig:wilson_deception}, \textbf{satisfying statistical significance is not sufficient to avoid being deceived about comparative algorithm performance} \cite{young2018crisis}. Thus, we will require additional tools aside from statistical tests to reason about this, which we discuss in Sections \ref{sec:hpo:formalizing} \& \ref{sec:hpo:defense}.

This example is not exceptional, or even particularly remarkable, in terms of illustrating the hyperparameter deception problem. We simply chose it for convenience: The experiment does not require highly-specialized ML sub-domain expertise to understand it, and it is arguably broadly familiar, as it very well-cited \cite{wilson2017marginal}. However, we emphasize that hyperparameter deception is rather common. Additional examples can be found in numerous empirical studies across ML subfields~\cite{choi2019empirical, sivaprasad2020hpo, melis2018evaluation, musgrave2020metric, bouthillier2019reproducibility, schneider2019deepobs, dodge2019nlp, Lucic2018-dr} (Appendix~\ref{chapter:app:hpo}). This work shows that reported results tend to be impressive for the tested hyper-HP configurations, but that modifying HPO can lead to vastly different performance outcomes that entail contradictory conclusions. 

More generally, it is possible to develop results that are wrong about performance, or else correct about performance but for the wrong reasons (e.g., by picking ``lucky'' hyperparameters). Neither of these outcomes constitutes reliable knowledge~\cite{gettier1963justified, lehrer1979gettier}. 
As scientists, this is disheartening. We want to have confidence in the conclusions we draw from our experiments. We want to trust that we are deriving reliable knowledge about algorithm performance. 
\textbf{In the sections that follow, our aim is to study HPO in this reliable-knowledge sense: We want to develop ways to reason rigorously and confidently about how we derive knowledge from empirical investigations involving HPO.}

%% file: section/20-arbitrary/21-hpo/213-hpo-epistemic.tex
\section{Epistemic Hyperparameter Optimization}\label{sec:hpo:ehpo}

Our discussion in Section~\ref{sec:hpo:prelim} shows that applying standard HPO methodologies can be deceptive: Our beliefs about algorithm performance can be controlled by happenstance, wishful thinking, or, even worse, potentially by an adversary trying to trick us with a tampered set of HPO logs. This leaves us in a position where the ``knowledge'' we derived may not be knowledge at all---since we could have easily (had circumstances been different) concluded the opposite. To address this, we propose that the process of drawing conclusions using HPO should itself be an object of study. We formalize this reasoning process, which we call \emph{epistemic hyperparameter optimization} (EHPO), and we provide an intuition for how EHPO can help us think about the hyperparameter deception problem. 

\begin{definition} \label{def:ehpo}
An \textbf{epistemic hyperparameter optimization procedure (EHPO)} is a tuple $(\mathcal{H}, \mathcal{F})$ where $\mathcal{H}$ is a set of HPO procedures $H$ (Definition \ref{def:hpo}) and $\mathcal{F}$ is a function that maps a set of HPO logs $\mathcal{L}$ (Definition \ref{def:log}) to a set of logical formulas $\mathcal{P}$, i.e. $\mathcal{F}(\mathcal{L}) = \mathcal{P}$. An execution of EHPO involves running each $H \in \mathcal{H}$ some number of times (each run produces a log $\ell$), and then evaluating $\mathcal{F}$ on the logs $\mathcal{L}$ produced in order to output the conclusions $\mathcal{F}(\mathcal{L})$ we draw from all of the HPO runs.
\end{definition}

In practice, it is common to run EHPO for two training algorithms, \algA{} and \algB{}, and to compare their performance to conclude which is better-suited for the task at hand. $\mathcal{H}$ contains at least one HPO that runs \algA{} and at least one HPO that runs \algB{}. The possible conclusions in output $\mathcal{P}$ include $p$ = ``\algA{} performs better than \algB{}'', and $\lnot p$ = ``\algA{} does not perform better than \algB{}''. 
Intuitively, EHPO is deceptive whenever it could produce $p$ and also could (if configured differently or due to randomness) produce $\lnot p$. That is, we can be deceived if the EHPO procedure we use to derive knowledge about algorithm performance could entail logically inconsistent results. 

Our example in Section \ref{sec:hpo:prelim} is deceptive because using different hyper-HP-configured grid searches for $\mathcal{H}$ could produce contradictory conclusions.  
We ran two variants of EHPO $(\mathcal{H}, \mathcal{F})$: The first replicated Wilson et al.~\cite{wilson2017marginal}'s original $\mathcal{H}$ of 3 grid-searches on SGD, HB, and Adam (Figure \ref{fig:wilson_deception}a), and the second used 3 grid-searches with a modified grid search for Adam that also tuned $\epsilon$ (Figure \ref{fig:wilson_deception}b). Each EHPO produced a $\mathcal{L}$ with 3 logs. 
For both, to draw conclusions $\mathcal{F}$ picks the best-performing HP-config per $\mathcal{A}$ and maps them to formulas including ``SGD outperforms Adam." From the 3 logs in Figure \ref{fig:wilson_deception}a, we conclude $p$: ``Non-adaptive optimizers outperform adaptive ones"; from the  3 logs in Figure \ref{fig:wilson_deception}b, we conclude $\lnot p$: ``Non-adaptive methods do not outperform adaptive ones." How can we formally reason about EHPO to avoid this possibility of drawing inconsistent conclusions---to guard against deceiving ourselves about algorithm performance when running EHPO?

\paragraph{Framing an adversary who can deceive us.} To begin answering this question, we take inspiration from Descartes' deceptive demon thought experiment (Appendix~\ref{chapter:app:hpo}). We frame the problem in terms of a powerful adversary trying to deceive us---one that can cause us to doubt ourselves and our conclusions. Notably, the demon is not a real adversary; rather, it models a worst-case setting of configurations and randomness that are usually set arbitrarily or by happenstance in EHPO. 

Imagine an evil demon who is trying to deceive us about the relative performance of different algorithms via running EHPO. At any time, the demon maintains a set $\mathcal{L}$ of HPO logs, which it can modify either by running an HPO $H \in \mathcal{H}$ with whatever hyper-HPs $c \in \mathcal{C}$ and seed $r \in \mathcal{I}$ it wants (producing a new log $\ell$, which it adds to $\mathcal{L}$) or by erasing some of the logs in its set. Eventually, it stops and presents us with $\mathcal{L}$, from which we will draw some conclusions using $\mathcal{F}$, i.e. $\mathcal{F}(\mathcal{L})$.

The demon's EHPO could deceive us via the conclusions we draw from the set of logs it produces. 
For example, $\mathcal{L}$ may lead us to conclude that one algorithm performs better than another, when in fact picking a different set of hyper-HPs could have generated logs that would lead us to conclude differently. We want to be sure that we will not be deceived by any logs the demon \emph{\textbf{could}} produce. 
Of course, this intuitive definition is lacking: It is not clear what is meant by \emph{\textbf{could}}. Our contribution in the sections that follow is to pin down a formal, reasonable definition of \emph{\textbf{could}} in this context, so that we can suggest an EHPO procedure that can defend against such a maximally powerful adversary. We intentionally imagine such a powerful adversary because, if we can defend against it, then we will also be defended against weaker or accidental deception.

%% file: section/20-arbitrary/21-hpo/214-hpo-logic.tex
\section{A Logic for Reasoning about EHPO}\label{sec:hpo:formalizing}

The informal notion of \emph{\textbf{could}} established above encompasses numerous sources of uncertainty. There is the time to run EHPO and the choices of random seed, algorithms to compare, HPO procedures, hyper-HPs, and learning task. Then, once we have completed EHPO and have a set of logs, we have to digest those logs into logical formulas from which we base our conclusions. This introduces more uncertainty, as we need to reason about whether we believe those conclusions or not.\looseness=-1 

Our formalization needs to capture all of these sources of uncertainty, and needs to be sufficiently expressive to capture how they could combine to cause us to believe deceptive conclusions. It needs to be expansive enough to handle the common case --- of a well-intentioned researcher with limited resources making potentially incorrect conclusions --- and the rarer, worst case --- of gaming results. 

\paragraph{Why not statistics?} As the common toolkit in ML, statistics might seem like the right choice for modeling all this uncertainty. However, statistics is great for reasoning about uncertainty that is \emph{quantifiable}. For this problem, not all of the sources of uncertainty are easily quantifiable. In particular, it is very difficult to quantify the different hyper-HP possibilities. It is not reasonable to model hyper-HP selection as a random process; 
we do not sample from a distribution and, even if we wanted to, it is not clear how we would pick the distribution from which to sample. Moreover, as we saw in our example in Section \ref{sec:hpo:prelim}, testing for statistical significance is not sufficient to prevent deception.  
While the results under consideration may be statistically significant, they can still fail to prevent the possibility of yielding inconsistent conclusions. For this reason, when it comes to deception, statistical significance can even give us false confidence in the conclusions we draw.

\paragraph{Why modal logic?} Modal logic is the standard mathematical tool for formalizing reasoning about uncertainty~\cite{chellas1980book, emerson1991temporal} --- for formalizing the thus far informal notion of what the demon \textit{\textbf{could}} bring about running EHPO. 
It is meant precisely for dealing with different types of uncertainty, particularly uncertainty that is difficult to quantify, and has been successfully employed for decades in AI~\cite{halpern2017book, halpern1991apm, blackburn2006modal}, programming languages~\cite{clarke1986tls, pnueli1977temporal, lamport1980semantics}, and distributed systems~\cite{hawblitzel2015ironfleet, fischer1988uncertaintyds, owicki1982liveness}. In each of these computer science fields, modal logic's flexible semantics has been indispensable for writing proofs about higher-level specifications with multiple sources of not-precisely-quantifiable, lower-level uncertainty. 

For example, in distributed computing, it lets us write proofs about overall system correctness, abstracting away from the specific non-determinism introduced by each lower-level computing process \cite{fischer1988uncertaintyds}. Analogously, modal logic can capture the uncertainty in EHPO without being prescriptive about particular hyper-HP choices. 
Our notion of correctness, which we want to reason about and guarantee, is not being deceived. Therefore, while modal logic may be an atypical choice for ML, it comes with a huge payoff. By constructing the right semantics, we can capture all the sources of uncertainty described above and we can write simple proofs about whether we can be deceived by the EHPO we run. In Section \ref{sec:hpo:defense}, it is this formalization that ultimately enables us to naturally suggest a defense against being deceived.

\subsection{Introducing  our logic: syntax and semantics overview} \label{sec:logicintro}

Modal logic inherits the tools of more-familiar propositional logic and adds two operators:  $\possible$ to represent \emph{possibility} and $\necessary$ to represent \emph{necessity}. These operators enable reasoning about \emph{possible worlds}---a semantics for representing how the world \emph{is} or \emph{could be}, making modal logic the natural choice to express the ``could'' intuition from Section \ref{sec:hpo:ehpo}. The well-formed formulas $\phi$ of modal logic are given recursively in Backus-Naur form, where $P$ is any atomic proposition:
\begin{align*}
    \phi \coloneqq P\;|\;\lnot \phi\;|\;\phi \land \phi \;|\; \possible \phi
\end{align*}
$\possible p$ reads, ``It is possible that $p$.'';  $p$ is true at \emph{some} possible world, which we \emph{could} reach (Appendix~\ref{chapter:app:hpo}). Note that $\necessary$ is syntactic sugar, with $\necessary p \equiv \lnot \possible \lnot p$. Similarly, ``or'' has $p \lor q \equiv \lnot (\lnot p \land \lnot q)$ and ``implies'' has $p \rightarrow q \equiv \lnot p \lor q$. The axioms of modal logic are as follows:
\begin{align*}
    \vdash Q &\rightarrow \necessary Q & \textit{(necessitation)}. &&
    \necessary(Q \rightarrow R) &\rightarrow (\necessary Q \rightarrow \necessary R) & \textit{(distribution)}.
\end{align*}

\noindent where $Q$ and $R$ are any formula, and $\vdash Q$ means $Q$ is a theorem of propositional logic. We can now provide the syntax and an intuitive notion of the semantics of our logic for reasoning about deception. 

\paragraph{Syntax.} 
Our logic requires an extension of standard modal logic. We need \emph{two} modal operators to reckon with two overarching modalities: the possible results of the demon running EHPO ($\possible_t$) and our beliefs about conclusions from those results ($\mathcal{B}$). Combining these modalities yields well-formed formulas $\psi$ where, for any atomic proposition $P$ and any positive real $t$,
\[
    \psi \coloneqq P\;|\;\lnot \psi\;|\;\psi \land \psi \;|\; \possible_t \psi \;|\; \mathcal{B} \psi
\]
Note the EHPO modal operator here is \emph{indexed}: $\possible_t$ captures  ``how possible'' ($\possible$) something is, quantified by the compute capabilities of the demon ($t$) \cite{blackburn2006modal, emerson1991temporal, heifeitz1998indexed}. 

\paragraph{Semantics intuition.}   
We suppose that an EHPO user has in mind some atomic propositions (propositions of the background logic unrelated to possibility or belief, such as 
``the best-performing log for \algA{} has lower loss than the best-performing log for \algB{}'') with semantics that are already defined.
$\land$ and $\lnot$ inherit their semantics from ordinary propositional logic, which can combine propositions to form formulas. A set of EHPO logs $\mathcal{L}$ (Definition \ref{def:log}) can be digested into such logical formulas. That is, we define our semantics using logs $\mathcal{L}$ as models over formulas $p$:  $\mathcal{L} \models p$, which reads ``$\mathcal{L}$ models $p$'', means that $p$ is true for the set of logs $\mathcal{L}$. We will extend this intuition to give semantics for possibility $\possible_t$ (Section \ref{sec:ehpologic}) and belief $\mathcal{B}$ (Section \ref{sec:belieflogic}), culminating in a tool that lets us reason about whether or not EHPO can deceive us by possibly yielding inconsistent conclusions (Section \ref{sec:deceptionlogic}).

\paragraph{Using our concrete example to ground us.} To clarify our presentation below, we will map our semantics to the example from Section \ref{sec:hpo:prelim}, providing an informal intuition before formal definitions.

\subsection{Expressing the possible outcomes of EHPO using $\possible_t$} \label{sec:ehpologic}
Our formalization for possible EHPO is based on the demon of Section~\ref{sec:hpo:ehpo}. 
Recall, the demon models a worst-case scenario. In practice, we deal with the easier case of well-intentioned ML researchers.  The notion of possibility we define here gives limits on what possible world a demon \emph{with bounded EHPO time} could reliably bring about. We first define a \emph{strategy} the demon can execute for EHPO:

\begin{definition} \label{def:strategy}
A randomized \textbf{strategy} $\boldsymbol{\sigma}$ is a function that specifies which action the demon will take. Given $\mathcal{L}$, its current set of logs, $\boldsymbol{\sigma(\mathcal{L})}$ gives a distribution over concrete actions, where each action is either 1) running a new $H$ with its choice of hyper-HPs $c$ and seed $r$ 
2) erasing some logs, or 3) returning. We let $\Sigma$ denote the set of all such strategies. 
\end{definition}

The demon we model controls the hyper-HPs $c$ and the random seed $r$, but importantly does \emph{not} fully control the PRNG $G$. From the adversary's perspective, for a strategy $\sigma$ to be reliable it must succeed regardless of the specific $G$. Informally, the demon cannot hack the PRNG.\footnote{We do not consider adversaries that can directly control how data is ordered and submitted to the algorithms under evaluation. This distinction shows that our logical construction non-trivial: We are able to defend against strong adversaries that can game the output of EHPO, which is separate from cheating by hacking the PRNG.}

\textbf{Informally}, we now want to \emph{execute a strategy} to bring about a particular outcome $p$. In Section \ref{sec:hpo:prelim}, our good-faith strategy was simple: We ran each $H$ with its own hyper-HPs and random seed, then returned. The demon is trickier: It is adopting a strategy to try to bring about a deceptive outcome. 
\textbf{Formally}, we model the 
demon executing strategy $\sigma$ on logs $\mathcal{L}$ with a PRNG unknown to the demon as follows. Let $\mathcal{G}$ denote the distribution over PRNGs $G: \mathcal{I} \rightarrow \mathcal{I}^{\infty}$, in which all number sequence elements are drawn independently and uniformly from $\mathcal{I}$ (recall, $\mathcal{I}$ is typically the 64-bit integers).  
First, draw $G$ from $\mathcal{G}$, conditioned on $G$ being consistent with all the runs in $\mathcal{L}$.\footnote{i.e., All random events recorded in $\mathcal{L}$ should agree with the corresponding random numbers produced by $G$.}
The demon then performs a random action drawn from $\sigma(\mathcal{L})$, using $G$ as the PRNG when running a new HPO $H$, and continues---updating the working set of logs $\mathcal{L}$ as it goes---until the ``return'' action is chosen. 

Using this process, we define what outcomes $p$ the demon can reliably bring about (i.e., what is possible, $\possible$) in the EHPO output logs $\mathcal{L}$ by running this random strategy $\sigma$ in bounded time $t$. \textbf{Informally}, $\possible_t p$ means that an adversary could adopt a strategy $\sigma$ that is guaranteed to cause the desired outcome $p$ to be the case while taking time at most $t$ in expectation. In Section \ref{sec:hpo:prelim}, where $p$ is ``Non-adaptive methods outperform adaptive ones", Figure \ref{fig:wilson_deception}a shows $\possible_t p$. \textbf{Formally},
\begin{definition} \label{def:ehpologic} 
Let $\boldsymbol{\sigma[\mathcal{L}]}$ denote the logs output from executing strategy $\sigma$ on logs $\mathcal{L}$, and let $\boldsymbol{\tau_\sigma(\mathcal{L})}$ denote the total time spent during execution. $\tau_\sigma(\mathcal{L})$ is equivalent to the sum of the times $T$ it took each HPO procedure $H \in \mathcal{H}$ executed in strategy $\sigma$ to run.
Note that both $\sigma[\mathcal{L}]$ and $\tau_\sigma(\mathcal{L})$ are random variables, as a function of the randomness of selecting $G$ and the actions sampled from $\sigma(\mathcal{L})$. 
For any formula $p$ and any $t \in \mathbb{R}_{>0}$, we say $\mathcal{L} \models \possible_t p$, i.e. ``$\mathcal{L}$ models that it is possible $p$ in time $t$,'' if 
\[
    \text{there exists a strategy } \sigma \in \Sigma, \text{ such that } \;\; \mathbb{P}(\sigma[\mathcal{L}] \models p) = 1 \; \text{ and } \; \mathbb{E}[\tau_\sigma(\mathcal{L})] \le t.
\]
\end{definition}

We will usually choose $t$ to be an upper bound on what is considered a reasonable amount of time to run EHPO. It does not make sense for $t$ to be unbounded, since this corresponds to the unrealistic setting of having infinite compute time to perform HPO runs.  
We model our budget in terms of time; however, we could use this setup to reason about other monotonically increasing resource costs, such as energy usage. 
Our indexed modal logic inherits many axioms of modal logic, with indexes added (Appendix~\ref{chapter:app:hpo}), e.g.:
\begin{align*}
\vdash (p \rightarrow q) \rightarrow (\possible_t p \rightarrow \possible_t q) && \textit{(necess. + distribution)} && p \rightarrow \possible_t p && \textit{(reflexivity)} \\
\possible_t\possible_s p \rightarrow \possible_{t+s} p && \textit{(transitivity)} &&
\possible_s\necessary_t p \rightarrow \necessary_t p && \textit{(symmetry)}\\
\possible_t(p \land q) \rightarrow (\possible_t p \land \possible_t q) && \textit{(dist. over $\land$)},
\end{align*}

\paragraph{To summarize:} The demon knows all possible hyper-HPs; it can pick whichever ones it wants to run EHPO within a bounded time budget $t$ to realize the outcome $p$ it wants. That is, if with some probability the demon can deceive us in some amount of time, then the demon can reliably deceive us with any larger time budget: If the demon fails to produce a deceptive result, it can use the strategy of just re-running until it yields the result it desires. Since $\possible_t$ models the worst-case all-powerful demon, it can also model any weaker EHPO user with time budget $t$.

\subsection{Expressing how we draw conclusions using $\mathcal{B}$} \label{sec:belieflogic}

We employ the modal operator $\mathcal{B}$ from the logic of belief\footnote{$\mathcal{B}$ is syntactically analogous to the $\necessary$ modal operator in standard modal logic~\cite{hintikka1962doxastic,vanBenthem2006epistemic, segerberg1999logicofbelief} (Appendix~\ref{chapter:app:hpo}).} to model ourselves as an observer who believes in the truth of the conclusions drawn from running EHPO. $\mathcal{B}p$ reads ``It is concluded that $p$.'' For example, when comparing the performance of two algorithms for a task, $p$ could be ``$\mathcal{J}$ is better than $\mathcal{K}$" and thus $\mathcal{B}p$ would be understood as, ``It is concluded that $\mathcal{J}$ is better than $\mathcal{K}$.''   

We model ourselves as a consistent \emph{Type 1} reasoner~\cite{smullyan1986belief}. \textbf{Informally}, this means we believe all propositional tautologies (necessitation), our belief distributes over implication (distribution), and we do not derive contradictions (consistency). We do not require completeness: We allow the possibility of not concluding anything about $p$ (i.e., neither $\mathcal{B}p$ nor $\mathcal{B}\lnot p$). \textbf{Formally}, for any formulas $p$ and $q$,
\begin{align*}
    \vdash p \rightarrow & \mathcal{B} p & \textit{(necess.)}; &&
    \mathcal{B} (p \rightarrow q) \rightarrow & (\mathcal{B} p \rightarrow \mathcal{B} q) &  \textit{(dist.)}; && 
    &\lnot ( \mathcal{B}p \land \mathcal{B}\lnot p ) & \textit{(consistency)}.
\end{align*}

To understand our belief semantics, recall that EHPO includes a function $\mathcal{F}$, which maps a set of output logs $\mathcal{L}$ to our conclusions (i.e., $\mathcal{F}(\mathcal{L}) = \mathcal{P}$ is our set of conclusions). 
\textbf{Informally}, when our conclusion set $\mathcal{F}(\mathcal{L})$ contains a formula $p$, we say the set of logs $\mathcal{L}$ models our belief $\mathcal{B}$ in that formula $p$. In Section \ref{sec:hpo:prelim}, the logs of Figure \ref{fig:wilson_deception}a model $\mathcal{B}p$ and the logs of Figure \ref{fig:wilson_deception}b model $\mathcal{B}\lnot p$. \textbf{Formally},

\begin{definition}\label{def:conclusionfunction}  For any formula $p$, we say
$\mathcal{L} \models \mathcal{B} p$, ``$\mathcal{L}$ models our belief in $p$'', if $p \in \mathcal{F}(\mathcal{L})$.
\end{definition}

Note we constrain what $\mathcal{F}$ can output. For a reasonable notion of belief, $\mathcal{F}$ 
must model the consistent \emph{Type 1} reasoner axioms above. Otherwise, deception aside, $\mathcal{F}$ is an unreasonable way to draw conclusions, since it is not even compatible with our belief logic. 

\subsection{Expressing hyperparameter deception} \label{sec:deceptionlogic}

So far we have defined the semantics of our two separate modal operators, $\possible_t$ and $\mathcal{B}$. We now begin to reveal the benefit of using modal logic for our formalization. These operators can interact to formally express what we informally illustrated in Section \ref{sec:hpo:prelim}: a notion of hyperparameter deception. It is a well-known result that we can combine modal logics~\cite{scott1970multimodal} (Appendix~\ref{chapter:app:hpo}).  We do so to define an axiom that, if satisfied, guarantees EHPO will not be able to deceive us. For any formula $p$,
\begin{align*}
    & \lnot \left( \possible_t \mathcal{B} p \; \land \; \possible_t \mathcal{B} \lnot p \right) & & \textit{($t$-non-deceptive)}.
\end{align*}

\textbf{Informally}, our running example can be considered a proof by exhibition: It violates this axiom because Figure \ref{fig:wilson_deception}a's logs model $\possible_t \mathcal{B} p$ and Figure \ref{fig:wilson_deception}b's logs model $\possible_t \mathcal{B} \lnot p$. That is, $\possible_t \mathcal{B} p \land \possible_t \mathcal{B} \lnot p$ using grid search for this task. 

For the worst-case, \emph{$t$-non-deceptiveness} expresses the following: \textbf{If there exists a strategy $\sigma$ by which the demon could get us to conclude $p$ in $t$ expected time, then there can exist no $t$-time strategy by which the demon could have gotten us to believe $\lnot p$}. 
To make this concrete, suppose our $t$-non-deceptive axiom holds for an EHPO method that results in $p$.  
Intuitively, given a maximum reasonable time budget $t$, if there is no adversary that can consistently control whether we believe $p$ 
or its negation when running that EHPO, then the EHPO is defended against deception. Conversely, if an adversary could consistently control our conclusions, then the EHPO is potentially gameable. That is, if our $t$-non-deceptive axiom does not hold (i.e., we can be deceived, $\possible_t \mathcal{B} p \land \possible_t \mathcal{B} \lnot p$), then even if we conclude $p$ after running EHPO, we cannot claim to \emph{know} $p$. Our belief as to the truth-value of $p$ could be under the complete control of an adversary---or just a result of happenstance.

\paragraph{To summarize:} An EHPO is $t$-non-deceptive if it satisfies all of the axioms above. Our example in Section \ref{sec:hpo:prelim} is $t$-deceptive because the axioms do not hold. The semantics of these axioms capture all of the possible uncertainty from the process of drawing conclusions from EHPO--and how that uncertainty can combine to cause us to believe $t$-deceptive conclusions. 

%% file: section/20-arbitrary/21-hpo/215-hpo-defense.tex
\section{Constructing Defended EHPO }\label{sec:hpo:defense}

Now that we have a formal notion of what it means for EHPO to be (non)-deceptive, \textbf{we can write proofs about what it means for an EHPO method to be guaranteed to be deception-free}. 
Importantly, these proofs will increase our confidence that our conclusions from EHPO are not due to the happenstance of picking a particular set of hyper-HPs. 

To talk about defenses, we need to understand what it means to construct a ``defended reasoner.'' In other words, for an EHPO $(\mathcal{H}, \mathcal{F})$, we need $\mathcal{F}$ to yield conclusions that we can defend against deception. 
Recall from Definition \ref{def:conclusionfunction} that logs $\mathcal{L}$ model our belief in a formula $p$, i.e. $\mathcal{L} \models \mathcal{B} p \equiv p \in \mathcal{F}(\mathcal{L})$. With this in mind, we begin by supposing we have a naive EHPO $(\mathcal{H}, \mathcal{F}_{\text{n}})$ featuring a naive reasoner $\mathcal{B}_{\text{n}}$ with corresponding belief function $\mathcal{F}_{\text{n}}$. We want to construct a new ``defended reasoner'' $\mathcal{B}_{*}$ that has a ``skeptical'' belief function $\mathcal{F_{*}}$.  $\mathcal{F_{*}}$ should weaken the conclusions of $\mathcal{F}_{\text{n}}$ (i.e., $\mathcal{F_{*}}(\mathcal{L}) \subseteq \mathcal{F}_{\text{n}}(\mathcal{L})$ for any $\mathcal{L}$) and result in an EHPO $(\mathcal{H}, \mathcal{F_*})$ that is guaranteed to be $t$-non-deceptive. In other words, defended reasoner $\mathcal{B}_*$ never concludes more than the naive reasoner $\mathcal{B}_{\text{n}}$. 
\textbf{Informally}, a straightforward way to do this is to have $\mathcal{B}_{*}$ conclude $p$ only if both the naive $\mathcal{B}_{\text{n}}$ would have concluded $p$, and it is impossible for an adversary to get $\mathcal{B}_{\text{n}}$ to conclude $\lnot p$ in time $t$. \textbf{Formally}, construct $\mathcal{B}_*$ such that for any $p$, 
\begin{align}\label{ax:defendedreasoner}
    \mathcal{B}_{*} p \equiv \mathcal{B}_{\text{n}} p \land \lnot \possible_t \mathcal{B}_{\text{n}} \lnot p
    \vspace{-.3cm}
\end{align}

Directly from our axioms (Section \ref{sec:hpo:formalizing}), we can now prove $\mathcal{B}_{*}$ is defended. We will suppose it is possible for $\mathcal{B}_{*}$ to be deceived, demonstrate a contradiction, and thereby guarantee that $\mathcal{B}_{*}$ is $t$-non-deceptive. Suppose $\mathcal{B}_{*}$ can be deceived in time $t$, i.e. $\textcolor{red}{\possible_t \mathcal{B}_{*} p} \land \textcolor{red}{\possible_t \mathcal{B}_{*} \lnot p}$ is $\mathsf{True}$. Starting with the left, $\textcolor{red}{\possible_t \mathcal{B}_{*}p}$ :

\begingroup
\setlength{\tabcolsep}{3pt} 
\renewcommand{\arraystretch}{1} 
\begin{center}
\begin{table}[H]
\footnotesize
\begin{center}
      \centering
        \begin{tabular}{r c l c c}
\toprule
\; & \; & \; & \; & \textbf{Rule} \\
\midrule
$\textcolor{red}{\possible_t \mathcal{B}_{*} p}$ & $\equiv$ & $\possible_t \left( \mathcal{B}_{\text{n}} p \land \lnot \possible_t \mathcal{B}_{\text{n}} \lnot p \right)$   &  \; & \text{Applying $\possible_t$ to the definition of $\mathcal{B}_{*} p$ \; (\ref{ax:defendedreasoner})} \\

\midrule
\; & $\rightarrow$ & $\possible_t \left( \lnot \possible_t \mathcal{B}_{\text{n}} \lnot p \right)$  & \;   &  \text{Reducing a conjunction to either of its terms: $(a \land b) \rightarrow b$} \\

\midrule
\; & $\rightarrow$ & $\textcolor{blue}{\lnot \possible_t \mathcal{B}_{\text{n}} \lnot p}$  & \;   & \text{Symmetry; dropping all but the right-most operator: $\possible_t(\possible_t a)\rightarrow \possible_t a$} \\
\bottomrule
\end{tabular}
\end{center}
\end{table}
\end{center}

\noindent We then pause to apply our axioms to the right side of the conjunction, $\textcolor{red}{\possible_t \mathcal{B}_{*}\lnot p}$ :

\begingroup
\setlength{\tabcolsep}{3pt} 
\renewcommand{\arraystretch}{1} 
\begin{table}[H]
\footnotesize
\begin{center}
      \centering
        \begin{tabular}{r c l c c}
\toprule
\; & \; & \; & \; & \textbf{Rule} \\
\midrule
$\textcolor{red}{\possible_t \mathcal{B}_{*} \lnot p}$ & $\equiv$ & $\possible_t \left( \mathcal{B}_{\text{n}} \lnot p \land \lnot \possible_t \mathcal{B}_{\text{n}} p \right)$   &  \; & \text{Applying $\possible_t$ to the definition of $\mathcal{B}_{*} \lnot p$ \; (\ref{ax:defendedreasoner})} \\

\midrule
\; & $\rightarrow$ & $\possible_t\mathcal{B}_{\text{n}} \lnot p \land \possible_t\lnot \possible_t \mathcal{B}_{\text{n}} p$  & \;   &  \text{Distributing $\possible_t$ over $\land$: $\possible_t (a \land b) \rightarrow (\possible_t a \land \possible_t b)$} \\

\midrule
\; & $\rightarrow$ & $\textcolor{blue}{\possible_t \mathcal{B}_{\text{n}} \lnot p}$  & \;   & \text{Reducing a conjunction to either of its terms: $(a \land b) \rightarrow a$} \\
\bottomrule
\end{tabular}
\end{center}
\vspace{-.35cm}
\end{table}

We now bring both sides of the conjunction back together: $\textcolor{red}{\possible_t \mathcal{B}_{*} p} \land  \textcolor{red}{\possible_t \mathcal{B}_{*} \lnot p}
\; \equiv \;
\textcolor{blue}{\lnot \possible_t \mathcal{B}_{\text{n}} \lnot p} \land  \textcolor{blue}{\possible_t \mathcal{B}_{\text{n}} \lnot p}$.
The \textcolor{blue}{right-hand side} is of the form $\lnot a \land a$, which must be $\mathsf{False}$. This contradicts our initial assumption that $\mathcal{B}_{*}$ is $t$-deceptive (i.e., $\textcolor{red}{\possible_t \mathcal{B}_{*} p} \land \textcolor{red}{\possible_t \mathcal{B}_{*} \lnot p}$ is $\mathsf{True}$). Therefore, $\mathcal{B}_{*}$ is $t$-non-deceptive. 

This example illustrates the power of our choice of formalization. In just a few lines of simple logic, we can validate defenses against deception. \textbf{This analysis shows that a $t$-defended reasoner $\mathcal{B}_{*}$ \emph{is always possible}}, and it does so without needing to refer to the particular underlying semantics of an EHPO. However, we intend this example to only be illustrative, as it may not be practical to compute $\mathcal{B}_{*}$ as defined in (\hyperlink{ax:defendedreasoner}{1}) if we cannot easily evaluate whether $\possible_t \mathcal{B}_{\text{n}} \lnot p$. We next suggest a concrete EHPO with a defended $\mathcal{B_{*}}$, and show how deception can be avoided in our Section \ref{sec:hpo:prelim} example by using this EHPO instead of grid search. 

\paragraph{A defended random search EHPO.} Random search takes two hyper-HPs, a distribution $\mu$ over the HP space and a number of trials $K \in \mathbb{N}$ to run. HPO consists of $K$ independent trials of training algorithms $\mathcal{A}_{\lambda_1}, \mathcal{A}_{\lambda_2}, \ldots, \mathcal{A}_{\lambda_K}$, where the HPs $\lambda_k$ are independently drawn from $\mu$, taking expected time proportional to $K$. When drawing conclusions, we usually look at the ``best'' run for each algorithm. For simplicity, we suppose there is only one algorithm, $\mathcal{A}$. We bound how much the choice of hyper-HPs can affect the HPs, and define a defended EHPO based on a variant of random search.

\begin{definition}\label{def:defended_random}
Suppose that we are given a naive EHPO procedure $(\{H\}, \mathcal{F}_{\text{n}})$, in which $H$ is random search and is the only HPO in our EHPO, and $\mathcal{F}_{\text{n}}$ is a ``naive'' belief function associated with a naive reasoner $\mathcal{B}_{\text{n}}$. 
For any $K, R \in \mathbb{N}$, we define the ``$(K,R)$-defended'' belief function $\mathcal{F}_*$ for a skeptical reasoner $\mathcal{B}_{*}$ as the following conclusion-drawing procedure.
First, $\mathcal{F}_*$ only makes conclusion set $\mathcal{P}_*$ from a single log $\hat \ell$ with $K*R$ trials; otherwise, it concludes nothing, outputting $\emptyset$.
Second, $\mathcal{F}_{*}$ splits the single $\hat \ell$ into $R$ logs $\ell_1, \ell_2, \ldots, \ell_R$, each containing $K$ independent-random-search trials.\footnote{This is not generally allowable. $\mathcal{F}_*$ can do this because random-search logs contain interchangeable trials.}
Finally, $\mathcal{F}_{*}$ outputs the intersection of what the naive reasoner would have output on each log $\ell_i$,
\[
    \mathcal{F}_{*}(\{\hat \ell\}) = \mathcal{P}_* \equiv \mathcal{F}_{\text{n}}(\{ \ell_1 \})
    \cap \mathcal{F}_{\text{n}}(\{ \ell_2 \}) \cap \cdots \cap \mathcal{F}_{\text{n}}(\{ \ell_R \}).
\]
Equivalently, $\{\hat \ell\} \models \mathcal{B_{*}}p$ only if $\{\ell_i\} \models \mathcal{B}_{\text{n}}p$ for all $i$.
\end{definition}

\textbf{Informally}, to draw a conclusion using this EHPO, $\mathcal{B}_{*}$ splits a random-search-trial log of size $K*R$ into $R$ groups of $K$-trial logs, passing each $K$-trial log to one of an ensemble of $R$ naive reasoners $\mathcal{B}_{\text{n}}$. $\mathcal{B}_{*}$ only concludes $p$ if all $R$ naive reasoners unanimously agree on $p$. We can guarantee this EHPO to be $t$-non-deceptive by assuming a bound on how much the hyper-HPs can affect the HPs.

\begin{theorem} \label{thm:defendedhpo}
Suppose that the set of allowable hyper-HPs $\mathcal{C}$ of $H$ is constrained, such that any two allowable random-search distributions $\mu$ and $\nu$ have Renyi-$\infty$-divergence at most a constant, i.e. $D_{\infty}(\mu \| \nu) \le \gamma$. The $(K,R)$-defended random-search EHPO of Definition \ref{def:defended_random} is guaranteed to be $t$-non-deceptive if we set $R \ge \sqrt{t \exp(\gamma K)/K} = O(\sqrt{t})$.
\end{theorem}

We prove Theorem \ref{thm:defendedhpo} in the Appendix~\ref{chapter:app:hpo}. This result shows that our defense is \emph{actually} a defense, and moreover it defends with a log size $K*R$---and compute requirement for good-faith EHPO---that scales sublinearly in $t$. A good-faith actor can, in sublinear-in-$t$ time, produce a log (of length $K*R$) that will allow our $t$-non-deceptive reasoner to reach conclusions. This means that we defend against adversaries with much larger compute budgets than are expected from good-faith actors.

\paragraph{Validating our defense empirically and selecting hyper-HPs.} Any defense ultimately depends on the hyper-HPs it uses. Thus, we should have a reasonable belief that choosing differently would not have led an opposite conclusion. We therefore run a two-phased search~\cite{choi2019empirical, henderson2017reinforcement, Riquelme2018-cn}, repeating our VGG16-CIFAR10 experiment from Section \ref{sec:hpo:prelim}. First, we run a coarse-grained, dynamic protocol to find reasonable hyper-HPs for Adam's $\epsilon$; second, we use those hyper-HPs to run our defended random search. We start with a distribution to search over $\epsilon$, and note that the performance is best on the high end. We  change the hyper-HPs, shifting the distribution until Adam's performance starts to degrade, and use the resulting hyper-HPs ($\epsilon \in [10^{10}, 10^{12}]$) to run our defense (Appendix~\ref{chapter:app:hpo}). 

\begin{figure}
      \centering
      \begin{algorithm}[H]
	\caption{Defense with Random Search}\label{algo:def}
	\begin{algorithmic}[1]
		\Require Set of ${K*R}$ random-search logs $\{\mathcal{L}_i\}_{i=1}^{KR}$, defense subsampling budget $M$, criterion constant $\delta$, subsample size $\kappa$
		\For{$m=1,\cdots, M$}
			\State Subsample $\kappa$ logs: $\{\mathcal{L}_i\}_{i=1}^\kappa\sim\{\mathcal{L}_i\}_{i=1}^{KR}$.
			\State Obtain conclusions $\{\mathcal{P}_i\}_{i=1}^\kappa$ from $\{\mathcal{L}_i\}_{i=1}^\kappa$.
			\State Obtain output conclusion for $m$: $\mathcal{P}^{(m)}\leftarrow \text{Majority}(\{\mathcal{P}_i\}_{i=1}^\kappa)$
		\EndFor
  
		\If{$\exists p$ s.t. $\geq (1-\delta)M$ of $\{\mathcal{P}^{(m)}\}_{i=1}^M$ conclude $p$}
		    \State Conclude $p$.
		\Else
		    \State Conclude nothing.
		\EndIf
	\end{algorithmic}
\end{algorithm}

\end{figure}%

\begin{figure}[t!]
      \centering
      \begin{table}[H]
\caption{Results from repeating our Section \ref{sec:hpo:prelim} experiment, using Algorithm \ref{algo:def} instead of grid search. $p$ = ``Non-adaptive optimizers (SGD and Heavy Ball) perform better than the adaptive optimizer Adam''.}
\label{table:defense}
\small
\begin{center}
\begin{tabular}{c c c  c  ccc}
\toprule
\; &   $p$ &   $\lnot p$   &   $1 - \delta$  &   Conclude    \\
\midrule
\multirow{3}[4]{*}{\shortstack{SGD\; \\vs.\;\\Adam\;}}   &   \multirow{3}[4]{*}{$0.213$}  &   \multirow{3}[4]{*}{$0.788$\;}  &   0.75    &   $\neg p$    \\
\cmidrule{4-5}
   &   &  & 0.8 &   Nothing \\
\cmidrule{4-5}
   &   &  & 0.9 &   Nothing \\
\midrule
\multirow{3}[4]{*}{\shortstack{Heavy Ball\; \\vs.\;\\Adam\;}}   &   \multirow{3}[4]{*}{$0.168$}  &   \multirow{3}[4]{*}{$0.832$\;}  &   0.75    &   $\lnot p$   \\
\cmidrule{4-5}
   &   &  & 0.8 &   $\lnot p$ \\
\cmidrule{4-5}
   &   &  & 0.9 &   Nothing \\
\bottomrule
\end{tabular}
\end{center}
\end{table}
\end{figure} 

We now run a modified version of our defended EHPO in Definition \ref{def:defended_random}, described in Algorithm \ref{algo:def}, with $K*R=600$ ($200$ logs for each optimizer). Using a budget of $M=10000$ iterations, we subsample $\kappa=11$ logs and pass them to an ensemble of $\kappa$ naive reasoners $\mathcal{B}_{\text{n}}$. We use $\kappa$ logs, relaxing the requirement of using all $K*R$ logs in Definition \ref{def:defended_random}, for efficiency. Each iteration $m$ concludes the majority conclusion of the $\kappa$-sized $\mathcal{B}_{\text{n}}$ ensemble. This is why we set $\kappa$ to an odd number---to avoid ties. $\mathcal{B}_{*}$ draws conclusions based on the results of the $M$-majority conclusions. That is, we further relax the requirements of Definition \ref{def:defended_random}: Instead of requiring unanimity, 
$\mathcal{B}_{*}$ only requires agreement on the truth-value of $p$ for a fractional subset of $M$. We set this fraction using parameter $\delta \in [0,1]$, where $\delta$ controls how skeptical our defended reasoner $\mathcal{B}_{*}$ is (lower $\delta$ corresponding to more skepticism). $\mathcal{B}_{*}$ concludes $p$ when at least $(1 - \delta)$ of our $M$ subsampled runs concluded $p$. When this threshold is not met, $\mathcal{B}_{*}$ remains skeptical and concludes nothing. We summarize our final results in Table \ref{table:defense}, and provide complete results in the Appendix~\ref{chapter:app:hpo}. Given how similar the optimizers all perform on this task (similar to Figure \ref{fig:wilson_deception}), being more skeptical increases the likelihood that we do not conclude anything.

%% file: section/20-arbitrary/21-hpo/216-hpo-conclusion.tex
\section{Conclusion and Practical Takeaways}\label{sec:hpo:conclusion}

Much recent empirical work illustrates that it is easy to draw inconsistent conclusions from HPO~\cite{choi2019empirical, sivaprasad2020hpo, melis2018evaluation, musgrave2020metric, bouthillier2019reproducibility, schneider2019deepobs, dodge2019nlp, Lucic2018-dr}. We call this problem \emph{hyperparameter deception} and, to derive a defense, \textbf{argue that the process of drawing conclusions using HPO should itself be an object of study}. 
Taking inspiration from Descartes' demon, we formalize a logic for studying an epistemic HPO procedure. 
The demon can run any number of reproducible HPO passes to try to get us to believe a particular notion about algorithm performance. 
Our formalization enables us to not believe deceptive notions: It naturally suggests how to guarantee that an EHPO is defended against deception. We offer recommendations to avoid hyperparameter deception in practice (we expand on this in Appendix~\ref{chapter:app:hpo}): 
\begin{itemize}
    \item \textbf{Researchers should construct their own notion of skepticism $\mathcal{B_*}$, appropriate to their specific task.} There is no one-size-fits-all defense solution. Our results are \emph{\textbf{broad insights}} about defended EHPO: A defended EHPO is \emph{\textbf{always possible}}, but finding an efficient one will depend on the task.
    \item \textbf{Researchers should make explicit how they choose hyper-HPs.} 
    What is reasonable is ultimately a function of what the ML community accepts. Being explicit, rather than eliding hyper-HP choices, is essential for helping decide what is reasonable. As a heuristic, we recommend setting hyper-HPs such that they include HPs for which the optimizers' performance starts to degrade, as we do above. 
    \item \textbf{Avoiding hyperparameter deception is just as important as reproducibility}. We have shown that reproducibility \cite{henderson2017reinforcement, pineau2019checklist, gundersen2018reproducibility, bouthillier2019reproducibility, Sinha2020-aw} is only part of the story for ensuring reliability. While necessary for guarding against brittle findings, it is not sufficient. We can replicate results---even statistically significant ones---that suggest conclusions that are altogether wrong.
\end{itemize}

More generally, our work is a call to researchers to reason more rigorously about their beliefs concerning algorithm performance. In relation to EHPO, this is akin to challenging researchers to reify their notion of $\mathcal{B}$---to justify their belief in their conclusions from the HPO. 
Such epistemic rigor concerning drawing conclusions from empirical studies has a long history in more mature branches of science and computing, including evolutionary biology~\cite{gould1996mismeasure}, statistics~\cite{gelman2014garden, gelman2019fishing}, programming languages~\cite{mytkowicz2009pl}, and computer systems~\cite{friedman1996bias} (Appendix~\ref{chapter:app:hpo}). We believe that applying similar rigor will contribute significantly to the ongoing effort of making ML more robust and reliable. 

%% file: section/20-arbitrary/22-fairness/22-fairness-main.tex
\chapter{Arbitrariness and Social Prediction}\label{chapter:fairness}

We next discuss a different source of arbitrariness in machine learning: 
how stochasticity based on the particular training-data examples can result in wildly variable classification outcomes in algorithmic fairness contexts.
The prior chapter (Chapter~\ref{chapter:hpo}) used predominantly theoretical tools to formally characterize a particular type of arbitrariness due to non-deterministic human decisions.
Here, our methods are almost entirely experimentally driven. 
They yield important insights that we have begun to translate to law and policy  (Chapter~\ref{chapter:nondeterminism}).\looseness=-1\\ 

\noindent \textbf{Chapter summary}: 
Variance in predictions across different trained models is a significant, under-explored source of error in 
fair binary classification. 
In practice, the variance on some data examples is so large that decisions can be effectively \emph{arbitrary}. 
To investigate this problem, we take an experimental approach and make four overarching contributions.  
We define a metric called \emph{self-consistency}, derived from variance, which we use as a proxy for measuring and reducing arbitrariness.
We then develop an ensembling algorithm that abstains from classification when a prediction would be arbitrary, and conduct a large-scale experimental study of the role of variance (\emph{vis-a-vis} self-consistency and arbitrariness) in fair binary classification. 
Our experiments reveal shocking insights about the reliability of conclusions on benchmark datasets.\\ 

\noindent This chapter is a licensed derivative copy of work published and awarded Best Student Paper (Honorable Mention)  at \emph{AAAI 2024}~\cite{cooper2024variance}. 
This work grew out of oral-awarded \emph{AIES 2021}~\citep{cooper2021emergent} and \emph{ICLR 2021} workshop~\cite{forde2021model} papers. Cooper and Abrams~\citep{cooper2021emergent} deals with conceptual mismeasurement in ML in relation to construct validity; this chapter, in contrast, focuses on arbitrariness in experimental measurements.
This work was also greatly influenced by earlier research on scalable uncertainty estimation (Chapter~\ref{chapter:tunamh} and Zhang et al.~\cite{zhang2020amagold}).

\input{section/20-arbitrary/22-fairness/221-fairness-intro}
\input{section/20-arbitrary/22-fairness/222-fairness-prelim}
\input{section/20-arbitrary/22-fairness/223-fairness-variance}
\input{section/20-arbitrary/22-fairness/224-fairness-algo}

\input{section/20-arbitrary/22-fairness/225-fairness-empirical}
\input{section/20-arbitrary/22-fairness/226-fairness-related}

%% file: section/20-arbitrary/22-fairness/221-fairness-intro.tex
\section{Introduction}\label{sec:fairness:intro}

A goal of algorithmic fairness is to develop techniques that measure and mitigate discrimination in automated decision-making. 
In fair binary classification, this often involves training a model to satisfy a chosen \emph{fairness metric}, which typically defines fairness as parity between model error rates for different demographic groups in the dataset~\citep{barocas2019textbook}. 
However, even if a model's classifications satisfy a particular fairness metric, it is not necessarily the case that the model is equally confident in each classification.\looseness=-1

\begin{figure}[t!]
    \centering
    \includegraphics[width=.5\textwidth]{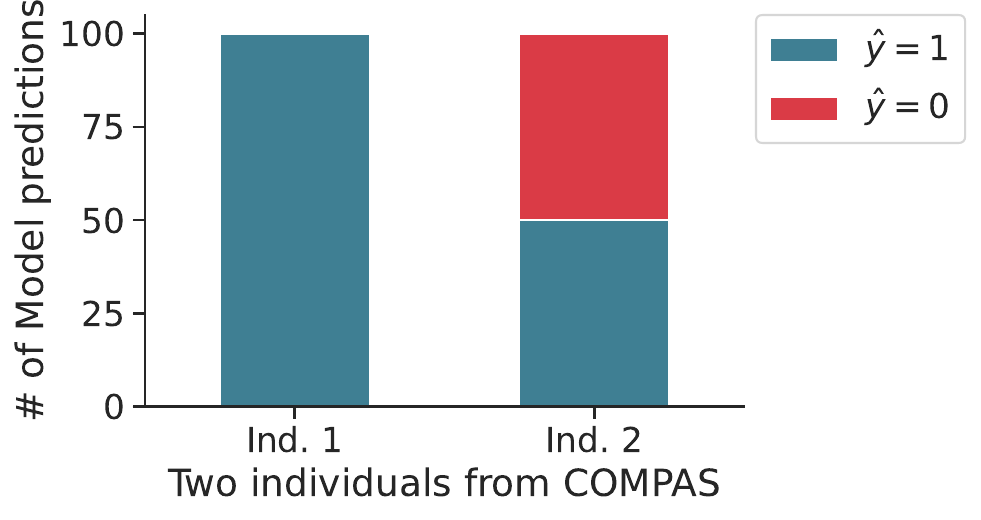}
    \caption{100 bootstrapped logistic regression models show models can be very consistent in predictions $\hat{y}$ for some individuals (Ind. 1) and arbitrary for others (Ind. 2).
    \looseness=-1}
    \label{fig:vote}
\end{figure}

To provide an intuition for what we mean by confidence, consider the following experiment: 
we fit 100 logistic regression models using the same learning process, which draws different subsamples of the training set from the \texttt{COMPAS} prison recidivism dataset~\citep{larson2016propublica, friedler2019datasets}, and we compare the resulting classifications for two individuals in the test set. 
Figure~\ref{fig:vote} shows a difference in the consistency of predictions for both individuals:  
the 100 models  agree completely to classify Individual 1 as ``will recidivate'' and disagree completely on whether to classify Individual 2 as ``will'' or ``will not recidivate.''

If we were to pick one model at random to use in practice, there would be no effect on how Individual 1 is classified;
yet, for Individual 2, the prediction is effectively random. 
We can interpret this disagreement to mean that the learning process that produced these predictions is not sufficiently confident to justify assigning Individual 2 \emph{either decision outcome}. 
In practice, instances like Individual 2 exhibit so little confidence that their classification is effectively \emph{arbitrary}~\citep{cooper2022accountability,cooper2022lawless,creel2022leviathan}.
Further, this arbitrariness can also bring about discrimination if classification decisions are \emph{systematically more arbitrary} for individuals in certain demographic groups.\looseness=-1

A key aspect of this example is that we use only one model to make predictions. 
This is the typical setup in fair binary classification: 
popular metrics are commonly applied to evaluate the fairness of a \emph{single model}~\citep{hardt2016equality, pleiss2017calibration, kleinberg2017impossibility}. 
However, as is clear from the example learning process in Fig.~\ref{fig:vote}, using only a single model can mask the arbitrariness of predictions. 
Instead, to reveal arbitrariness, we must examine \emph{distributions over possible models for a given learning process}. 
With this shift in frame, we ask: 
\emph{What is the empirical role of arbitrariness in  fair binary classification tasks?}

To study this question, we: 
\begin{enumerate}
    \item \textbf{Quantify arbitrariness.} We formalize a metric called \emph{self-consistency}, derived from statistical variance, which we use as a quantitative proxy for arbitrariness of model outputs. Self-consistency is a simple yet powerful tool for empirical analyses of fair classification (Section \ref{sec:fairness:significance}). 
    
    \item \textbf{Ensemble to improve self-consistency.} We extend Breiman's classic bagging algorithm~\citep{breiman1996bagging}  to allow for abstaining from classifying instances for which self-consistency is low. 
    This improves overall self-consistency (i.e., reduces variance), and improves accuracy (Section \ref{sec:fairness:algorithms}).\looseness=-1

    \item \textbf{Perform a comprehensive experimental study of variance in fair binary classification.\looseness=-1} 
    We conduct the largest-to-date such study, through the lens of self-consistency and its relationship to arbitrariness. 
    Surprisingly, we find that the benchmarks we evaluate are \textit{close-to-fair} when taking into account the amount of arbitrariness present in predictions --- \emph{before} we even try to apply \emph{any} fairness interventions (Section \ref{sec:fairness:empirical}). 
    This finding casts doubt on the reliability of prior work that uses individual models to make claims that there is baseline unfairness in these benchmarks (Section \ref{sec:fairness:related}).\looseness=-1
\end{enumerate}

%% file: section/20-arbitrary/22-fairness/222-fairness-prelim.tex
\section{Preliminaries on Fair Binary Classification}\label{sec:fairness:prelim}

To analyze arbitrariness in the context of fair binary classification, we first need to establish our background definitions. 
This material is likely familiar to most readers. Nevertheless, we highlight particular details that are important for understanding the experimental methods that enable our contributions. 
We present the fair-binary-classification problem formulation and associated empirical approximations, with an emphasis on the \emph{distribution over possible models} that could be produced from training on different subsets of data drawn from the same data distribution.\looseness=-1

\paragraph{Problem formulation.}

Consider a distribution $q(\cdot)$ from which we can sample \textit{examples} $(\instance, \group, \olabel)$. 
The $\instance \in \instances \subseteq \R^m$ are feature \textit{instances} and $\group \in \sG$ is a group of \textit{protected attributes} that we do not use for learning (e.g., race, gender).\footnote{We examine the common setting in which $|\group|=1$, and abuse notation, treating $\group$ like a scalar with $\sG = \{0, 1\}$.} 
The $\olabel \in \olabels$ are the associated \textit{observed labels}, and $\olabels \subseteq \labels$, where $\labels =  \{0, 1\}$ is the label space. 
From $q(\cdot)$ we can sample training datasets $\{(\instance, \group, \olabel)\}_{i=1}^n$, with $\datasets$ representing the set of all  $n$-sized datasets. 
To reason about the possible models of a hypothesis class $\hclass$ that could be learned from the different subsampled datasets $\datasetk \in \datasets$, we define a \textit{learning process}:\looseness=-1

\begin{definition}
    \label{def:learningprocess}
    A \textbf{learning process} is a randomized function that runs instances of a \textbf{training procedure} $\tproc$ on each $\datasetk \in \datasets$ and a model specification, in order to produce \textbf{classifiers} $\modelk \in \hclass$. 
    A particular run $\tproc(\datasetk) \rightarrow \modelk$, where $\modelk: \instances \rightarrow \labels$, which is  deterministic mapping from the instance space $\instances$ to the label space $\labels$. 
    All such runs over $\datasets$ produce a distribution over possible trained models, $\possiblemodels$. 
\end{definition}

Reasoning about $\possiblemodels$, rather than individual models $\modelk$, enables us to contextualize arbitrariness in the data, which, in turn, is captured by learned models (Section~\ref{sec:fairness:significance}).\footnote{Model multiplicity has similar aims, but ultimately relocates  the arbitrariness we describe to model selection (Section~\ref{sec:fairness:related} and Appendix~\ref{app:sec:mm}).} 
Each particular model $\modelk\sim\possiblemodels$ deterministically produces classifications $\pred = \modelk(\instance)$. 
The classification rule is $\modelk(\instance) = \1[\regressork(\instance) \geq \tau]$, for some threshold $\tau$, where regressor $\regressork: \instances \rightarrow [0, 1]$ computes the probability of positive classification. 
Executing $\tproc(\datasetk)$ produces $\modelk \sim \possiblemodels$ by minimizing the \textit{loss} of predictions $\pred$ with respect to their associated observed labels $\olabel$ in $\datasetk$. 
This loss is computed by a chosen \textit{loss function} $\lossarb:\labels \times \labels \mapsto \R$.  
We compute predictions for a \textit{test set} of fresh examples and calculate their loss. 
The loss is an estimate of the \textit{error} of $\modelk$, which is dependent on the specific dataset $\datasetk$ used for training. 
To generalize to the error of all possible models produced by a specific learning process (Definition~\ref{def:learningprocess}), we consider the \textit{expected error}, $\err(\tproc, \datasets, (\instance,\;\group,\; \olabel)) =\E_{\rmD}[\lossarb(\olabel,\; \pred)|\rvx = \instance]$. 

In fair classification, it is common to use  \textit{0-1 loss} $\triangleq \1[\pred \ne \olabel]$ or \textit{cost-sensitive loss}, which assigns asymmetric costs $\costfp$ for false positives \fp{} and $\costfn$ for false negatives \fn{}~\citep{elkan2001cost}. 
These costs are related to the classifier threshold $\tau = \frac{\costfp}{\costfp + \costfn}$, with $\costfp, \costfn \in \R^+$ (Appendix~\ref{app:sec:prelim:costs}). 
Common fairness metrics, such as Equality of Opportunity~\citep{hardt2016equality}, further analyze error by computing disparities across group-specific error rates $\fpr_{\group}{}$ and $\fnr_{\group}$. 
For example, $\fpr_\group \triangleq p_\possiblemodels[\regressor_{\rmD}(\rvx) \geq \tau| \olabel = 0, \rvg = \group] = p_\possiblemodels[\pred=1 | \olabel = 0, \rvg = \group]$. 
Model-specific $\fpr_\group$ and $\fnr_\group$ are further-conditioned on the dataset used in training, i.e., $\rmD = \datasetk$.\looseness=-1

\paragraph{Empirical approximation of the problem formulation.} 
We typically only have access to one dataset, not the data distribution $q(\cdot)$. 
In fair binary classification experiments, it is common to estimate expected error by performing \textit{cross validation} (CV) on this dataset to produce a small handful of models~\citep{chen2018tradeoff,  corbettdavies2017cost}. 
CV can be unreliable when there is high variance; 
it can produce error estimates that are themselves high variance, and does not reliably estimate expected error with respect to possible models $\possiblemodels$ (Section \ref{sec:fairness:empirical}). 
For more details, see Efron~\cite{efron1979bootstrap} and Efron and Tibshirani~\cite{efron1993bootsrap} and Wager~\cite{wager2020cv}. 

To get around these reliability 
issues, one can \textit{bootstrap}.\footnote{We could use MCMC~\citep{zhang2020amagold, zhang2020tunamh}, but optimization is the standard tool that allows use of standard models  in fairness.\looseness=-1} 
Bootstrapping splits the available data into train and test sets, and simulates drawing different training datasets from a distribution by resampling the train set $\hatdataset$, generating replicates $\hatdataset_1, \hatdataset_2, \ldots, \hatdataset_\boot \coloneqq \hatdatasets$. 
We use these replicates $\hatdatasets$ 
to approximate the learning process on $\datasets$ (Def.~\ref{def:learningprocess}). 
We treat the resulting $\hatmodel_{\hatdataset_1}, \hatmodel_{\hatdataset_2}, \ldots, \hatmodel_{\hatdataset_\boot}$ as our empirical estimate for the distribution $\hat{\mu}$, and evaluate their predictions for the \emph{same reserved test set}. 
This enables us to produce comparisons of classifications across test instances like in Figure~\ref{fig:vote} (Appendix~\ref{app:sec:prelim:boot}).\looseness=-1 

%% file: section/20-arbitrary/22-fairness/223-fairness-variance.tex
\section{Variance, Self-Consistency and Arbitrariness}\label{sec:fairness:significance}

We develop a quantitative proxy for measuring arbitrariness, called \emph{self-consistency} (Section~\ref{sec:var:sc}), which is derived from a definition of statistical \emph{variance} between different model predictions (Section~\ref{sec:var:intuition}). 
We then illustrate how self-consistency is a simple-yet-powerful tool for revealing the role of arbitrariness in fair classification (Section~\ref{sec:var:arbitrary}). 
Next, we will introduce an algorithm to improve self-consistency (Section~\ref{sec:fairness:algorithms}) and compute self-consistency on popular fair binary classification benchmarks (Section~\ref{sec:fairness:empirical}).\looseness=-1 

\subsection{Arbitrariness Resembles Statistical Variance}\label{sec:var:intuition}

In Section~\ref{sec:fairness:prelim}, we discussed how common fairness definitions analyze error by computing false positive rate (\fpr{}) and false negative rate (\fnr). 
Another common way to formalize error is as a decomposition of different statistical sources: \emph{noise}-, \emph{bias}-, and \emph{variance}-induced error~\citep{abumostafa2012learning, geman1992bvd}. 
To understand our metric for self-consistency (Section~\ref{sec:var:sc}), we first describe how the arbitrariness in Figure~\ref{fig:vote} (almost, but not quite) resembles variance.\looseness=-1

Informally, variance-induced error quantifies fluctuations in individual example predictions for different models $\modelk \sim \possiblemodels$. 
Variance is the error in the learning process that comes from training on different datasets $\datasetk \in \datasets$. 
In theory, we measure variance by imagining training all possible $\modelk \sim \possiblemodels$, testing them all on the same test instance $(\instance, \group)$, and then quantifying how much the resulting classifications for $(\instance, \group)$ deviate \emph{from each other}. More formally,

\begin{definition}
\label{def:variance}
\looseness=-1
For all pairs of possible models $\model_{\dataset_i}, \model_{\dataset_j}\sim\possiblemodels \;(i\neq j)$, the \textbf{variance} for a test  
$(\instance, \group)$ is\looseness=-1
\begin{align*}
\textstyle
\variance &\triangleq \E_{\model_{\dataset_i} \sim \possiblemodels, \model_{\dataset_j} \sim \possiblemodels}\Big[\lossarb\Big(\model_{\dataset_i}(\instance), \model_{\dataset_j}(\instance)\Big)\Big].
\end{align*}
\end{definition}

We can approximate variance directly by using the bootstrap method (Section~\ref{sec:fairness:prelim}, Appendices~\ref{app:sec:prelim:boot} and~\ref{app:sec:noisebias}).
For 0-1 and cost-sensitive loss with costs $\costfp, \costfn \in \R^+$ (Section~\ref{sec:fairness:prelim}), we can generate $\boot$ replicates to train $\boot$ concrete models that serve as our approximation for the distribution  $\hat{\possiblemodels}$.
For $\boot= \boot_0 + \boot_1 > 1$, where $\boot_0$ and $\boot_1$ denote the number of $0$- and $1$-class predictions for $(\instance, \group)$,\looseness=-1 
\begin{align}
\label{eq:hatvar}
\textstyle
\hatvariance &\coloneqq \frac{1}{\boot(\boot-1)} \sum_{i \neq j} \lossarb\Big(\hatmodel_{\hatdataset_i}(\instance), \hatmodel_{\hatdataset_j}(\instance)\Big) \nonumber\\
&= \frac{(\costfp + \costfn)\boot_0\boot_1}{\boot(\boot-1)}.
\end{align}

\noindent We derive (\ref{eq:hatvar}) in Appendix~\ref{app:sec:ourvariance} and show that, for increasingly large $\boot$, $\hatvar$ is defined on $[0, \frac{\costfp + \costfn}{4} + \epsilon]$.

\subsection{Defining Self-Consistency from Variance}\label{sec:var:sc} 

It is clear from above that, in general, variance (\ref{eq:hatvar}) is unbounded. 
We can always increase the maximum possible $\hatvar$ by increasing the magnitudes of our chosen $\costfp$ and $\costfn$ (Section \ref{sec:fairness:prelim}).\footnote{Because $\tau = \frac{\costfp}{\costfp + \costfn}$, for a given $\tau$ we can scale costs arbitrarily and have the same decision rule 
    Relative, not absolute, costs affect the number of classifications $\boot_0$ and $\boot_1$.
} 
However, as we can see from our intuition for arbitrariness in Figure~\ref{fig:vote}, the most important takeaway is the amount of (dis)agreement, reflected in the counts $\boot_0$ and $\boot_1$. 
Here, there is no notion of the cost of misclassifications. 
So, variance (\ref{eq:hatvar}) does not exactly measure what we want to capture. 
Instead, we want to focus unambiguously on the (dis)agreement part of variance, which we call \emph{self-consistency of the learning process}:\looseness=-1

\begin{definition}
\label{def:sc}
\looseness=-1
For all pairs of possible models $\model_{\dataset_i}, \model_{\dataset_j}\sim\possiblemodels \;(i\neq j)$, the \textbf{self-consistency of the learning process} for a test $(\instance, \group)$ is\looseness=-1
\begin{align}
\label{eq:sc}
\consistency &\triangleq \E_{\model_{\dataset_i} \sim \possiblemodels, \model_{\dataset_j} \sim \possiblemodels}\Big[\model_{\dataset_i}(\instance) = \model_{\dataset_j}(\instance)\Big] \nonumber\\
& = p_{\model_{\dataset_i} \sim \possiblemodels, \model_{\dataset_j} \sim \possiblemodels}\big(\model_{\dataset_i}(\instance) = \model_{\dataset_j}(\instance)\big).
\end{align}
\end{definition}

In words, (\ref{eq:sc}) models the probability that two models produced by the same learning process on different $n$-sized training datasets agree on their predictions for the same test instance.
Like variance, we can derive an empirical approximation of \texttt{SC}. 
Using the bootstrap method with $\boot = \boot_0 + \boot_1 > 1$,\looseness=-1 
\begin{align}
\label{eq:hatsc}
\hat{\texttt{SC}}\big(\mathcal{A}, \hatdatasets, (\instance, \group)\big) &\coloneqq \frac{1}{\boot(\boot-1)} \sum_{i \neq j} \1\Big[\hatmodel_{\hatdataset_i}(\instance) = \hatmodel_{\hatdataset_j}(\instance)\Big]  \nonumber\\
&= 1 - \frac{2\boot_0\boot_1}{\boot(\boot-1)}.
\end{align}

\noindent For increasingly large $\boot$, $\hat{\texttt{SC}}$ is defined on $[0.5 - \epsilon, 1]$ (Appendix~\ref{app:sec:consistency}). 
Throughout, we use the shorthand \emph{self-consistency}, but it is important to note that Definition~\ref{def:sc} is a property of the distribution over possible models $\mu$ produced by the learning process, not of individual models. 
We summarize other important takeaways below:

\paragraph{Terminology.} In naming our metric, we intentionally evoke related notions of ``consistency'' in logic and the law~\cite{fuller1965law, stalnaker2006logic} (Appendix~\ref{app:sec:consistency:details}).\looseness=-1 

\paragraph{Interpretation.}  
Definition~\ref{def:sc} is defined on $[0.5, 1]$, which coheres with the intuition in Figure~\ref{fig:vote}:  $0.5$ and $1$ respectively reflect minimal (Individual 2) and maximal (Individual 1) possible $\texttt{SC}$.
$\texttt{SC}$, unlike \fpr{} and \fnr{} (Section~\ref{sec:fairness:prelim}), does \emph{not} depend on the observed label $\olabel$. 
It captures the learning process's confidence in a classification $\pred$, but says nothing directly about $\pred$'s accuracy. 
By construction, low self-consistency indicates high variance, and vice versa. 
We derive empirical $\hatsc$ (\ref{eq:hatsc}) from $\hatvar$ (\ref{eq:hatvar}) by leveraging observations about the definition of $\hatvar$ for 0-1 loss (Appendix~\ref{app:sec:consistency}). 
While there are no costs $\costfp$, $\costfn$ in computing (\ref{eq:hatsc}), they still affect empirical measurements of $\hatsc$. 
Because $\costfp$ and $\costfn$ affect $\tau$ (Section~\ref{sec:fairness:prelim}), they control the concrete number of $\boot_0$ and $\boot_1$, and thus the $\hatsc$ we measure in experiments.\looseness=-1

\paragraph{Empirical focus.} 
Since self-consistency depends on the particular data subsets used in training, conclusions about its relevance vary according to task. 
This is why we take a practical approach for our main results  --- of running a large-scale experimental study on many different datasets to extract general observations about $\hatsc$'s practical effects (Section~\ref{sec:fairness:empirical}). 
In our experiments, we typically use $\boot=101$, which yields a $\hatsc$ range of $[\approx 0.495, 1]$ in practice.\footnote{Efron and Tibshirani~\cite{efron1993bootsrap} recommend $\boot \in \{50 \ldots 200\}$.\looseness=-1}\looseness=-1 

\paragraph{Relationship to other fairness concepts.} 
Self-consistency is qualitatively different from traditional fairness metrics.  
Unlike \fpr{} and \fnr{}, \texttt{SC} does not depend on observed label $\olabel$. 
This has two important implications. 
First, while calibration also measures a notion of confidence, it is different: calibration reflects confidence with respect to \emph{a model} predicting $\olabel$, but says nothing about the relative confidence in predictions $\hat{y}$ produced by the \emph{possible models} $\possiblemodels$ that result from the learning process~\citep{pleiss2017calibration}. 
Second, a common assumption in algorithmic fairness is that there is \emph{label bias} --- that unfairness is due in part to discrimination reflected in recorded, observed decisions $\olabel$~\citep{friedler2016impossibility, cooper2021emergent}. 
As a result, it is arguably a nice side effect that self-consistency does not depend on $\olabel$. 
However, it is also possible to be perfectly self-consistent and inaccurate (e.g., $\forall k, \hat{y}_k \neq \olabel$; Section~\ref{sec:fairness:related}).

\begin{figure*}[t!]
\begin{minipage}{.495\linewidth}
\centering
\hspace{-.4cm}
        \includegraphics[width=.85\linewidth]{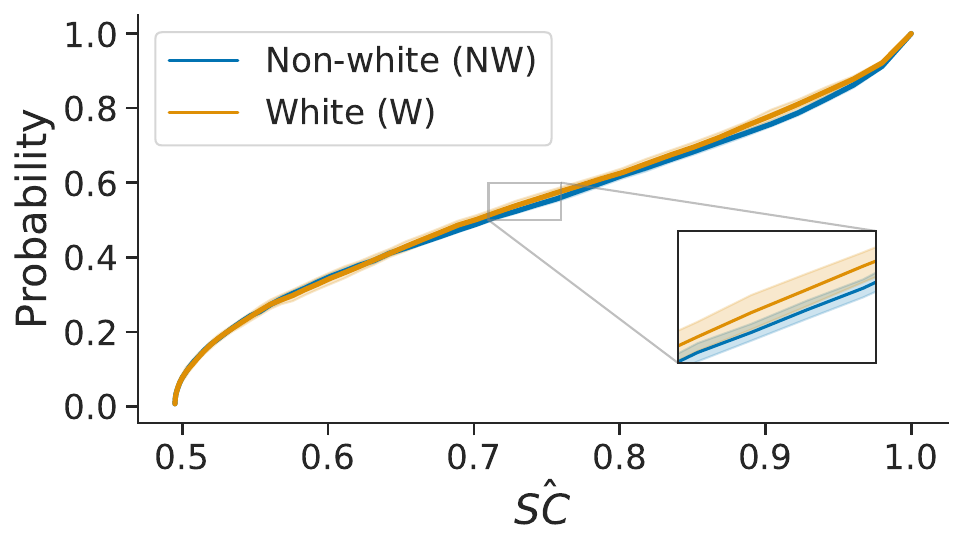}\\\vspace{.25cm}
        \small
        \begin{tabular}{lccc}
        \toprule
             & \textbf{$\Delta\haterr$} & \textbf{$\Delta\hatfpr$} & \textbf{$\Delta\hatfnr$} 
             \\ \cmidrule{2-4}
            & $1.0\pm1.4\%$ & $2.0\pm1.4\%$ & $0.9\pm1.4\%$ 
            \\
            \midrule
            & \textbf{$\haterr$} & \textbf{$\hatfpr$} & \textbf{$\hatfnr$} 
            \\ \midrule
            \textbf{Total} & $36.6\pm0.5\%$ & $17.3\pm0.8\%$ & $19.3\pm0.7\%$ 
            \\ \midrule
             $\text{NW}$ & $36.9\pm0.5\%$ & $18.0\pm0.7\%$ & $19.0\pm0.8\%$ 
             \\ \midrule 
            $\text{W}$ & $35.9\pm1.3\%$ & $16.0\pm1.2\%$ & $19.9\pm1.1\%$
            \\ 
              \bottomrule
        \end{tabular}
        \subcaption{\texttt{COMPAS} split by $\texttt{race}$; random forests (RFs)}
        \label{subfig:compas-cdf-rfc}
\end{minipage}%
\hspace{.25cm}
\begin{minipage}{.495\linewidth}
\centering
\hspace{-.2cm}
        \includegraphics[width=.85\linewidth]{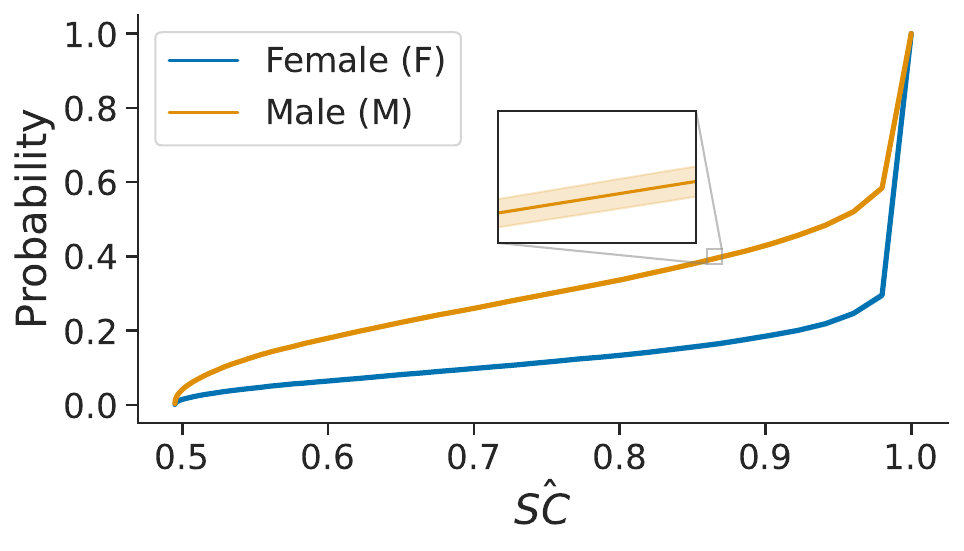}\\\vspace{.25cm}
        \small
        \begin{tabular}{lccc}
        \toprule
             & \textbf{$\Delta\haterr$} & \textbf{$\Delta\hatfpr$} & \textbf{$\Delta\hatfnr$} 
             \\ \cmidrule{2-4}
            & $12.2\pm0.4\%$ & $6.0\pm0.3\%$ & $6.3\pm0.3\%$ 
            \\ 
            \midrule
             & \textbf{$\haterr$} & \textbf{$\hatfpr$} & \textbf{$\hatfnr$} 
             \\ \midrule
             \textbf{Total} & $17.3\pm0.3\%$ & $7.7\pm0.3\%$ & $9.6\pm0.1\%$ 
             \\ \midrule
             $\text{F}$\;\;\;\; & $9.0\pm0.3\%$ & $3.7\pm0.1\%$ & $5.3\pm0.3\%$ 
             \\ \midrule
             $\text{M}$ & $21.2\pm0.3\%$ & $9.7\pm0.3\%$ & $11.6\pm0.1\%$ 
             \\ 
             \bottomrule
        \end{tabular}
        \subcaption{\texttt{Old Adult} split by $\texttt{sex}$; random forests (RFs)}
        \label{subfig:adult-cdf-rfc}
\end{minipage}%
\vspace{.25cm}
\caption{$\hatsc$ CDFs for \texttt{COMPAS} (\ref{subfig:compas-cdf-rfc}) and  \texttt{Old Adult} (\ref{subfig:adult-cdf-rfc}). 
We train random forests  ($B=101$ replicates), and repeat with 10 train/test splits to produce (very tight) confidence intervals.
$\hatsc$ is effectively identical across subgroups $\group$ in \texttt{COMPAS}; \texttt{Old Adult} exhibits systematic differences in arbitrariness across $\group$. Tables show mean $\pm$ STD of the relative disparities, 
e.g., $\Delta\haterr=|\haterr_0 - \haterr_1|$ (top); and, the absolute $\haterr, \hatfpr, \hatfnr, $ and $\hatsc$, also broken down by $\group$ (bottom).\looseness=-1}
\label{fig:adult-compas-cdf-rfc}
\end{figure*}

\subsection{Illustrating Self-Consistency in Practice}\label{sec:var:arbitrary}

$\hatsc$ enables us to evaluate arbitrariness in classification experiments. 
It is straightforward to compute $\hatsc$ (\ref{eq:hatsc}) with respect to multiple test instances $(\instance, \group)$ --- for all instances in a test set or for all instances conditioned on membership in $\group$. 
Therefore, beyond visualizing $\hatsc$ for individuals (Figure~\ref{fig:vote}), we can also do so across sets of individuals. 

We plot the cumulative distribution (CDF) of $\hatsc$ for the groups $\group$ in the test set (i.e., the $x$-axis shows the range of $\hatsc$ for $\boot=101$, $[\approx0.495, 1]$). 
In Figure~\ref{fig:adult-compas-cdf-rfc}, we provide illustrative examples from two of the most common fair classification benchmarks~\citep{fabris2022datasets}, \texttt{COMPAS} and \texttt{Old Adult} using random forests (RFs). 
We split the available data into train and test sets, and bootstrap the train set $B=101$ times to train models $\hat{\model_1}, \hat{\model_2}, \ldots, \hat{\model_{101}}$ (Section \ref{sec:fairness:prelim}). 
We repeat this process on 10 train/test splits, and the resulting confidence intervals (shown in the inset) indicate that our $\hatsc$ estimates are stable. 
We group observations 
into two categories:\looseness=-1

\paragraph{Individual arbitrariness.} 
Both CDFs show that $\hatsc$ varies drastically across test instances. 
For random forests on the \texttt{COMPAS} dataset, about one-half of instances are under $.7$ self-consistent. 
Nearly one-quarter of test instances are effectively $.5$ self-consistent; they resemble Individual 2 in Figure~\ref{fig:vote}, meaning that their predictions are essentially arbitrary. 
These differences in $\hatsc$ across the test set persist even though the 101 models exhibit relatively small average disparities $\Delta\haterr$, $\Delta\hatfpr$, and $\Delta\hatfnr$ (Figure~\ref{subfig:compas-cdf-rfc}, bottom; Section~\ref{sec:empirical-repro}). 
This supports our motivating claim: it is possible to come close to satisfying fairness metrics, while the learning process exhibits very different levels of confidence for the underlying classifications that inform those metrics 
(Section~\ref{sec:fairness:intro}).\looseness=-1

\paragraph{Systematic arbitrariness.} 
We can also highlight $\hatsc$ according to groups $\group$. 
The $\hatsc$ plot for \texttt{Old Adult} shows that it is possible for the degree of arbitrariness to be \emph{systematically worse} for a particular demographic $\group$ (Figure~\ref{subfig:adult-cdf-rfc}). 
While the lack of $\hatsc$ is not as extreme as it is for \texttt{COMPAS} (Figure~\ref{subfig:compas-cdf-rfc}) --- the majority of test instances exhibit over $.9 \;\;\; \hatsc$ --- there is more arbitrariness in the \texttt{Male} subgroup. 
We can quantify such \emph{systematic arbitrariness} using a measure of distance between probability distributions. 
We use the Wasserstein-1 distance ($\mathcal{W}_1$), which has a closed form for CDFs~\citep{ramdas2015wass}.
The $\mathcal{W}_1$ distance has an intuitive interpretation for measuring systematic arbitrariness: it computes the total disparity in \texttt{SC} by examining all possible \texttt{SC} levels $\kappa$ at once (Appendix~\ref{app:sec:consistency}). 
For two groups $\group=0$ and $\group=1$ with respective 
\texttt{SC} CDFs $F_0$ and $F_1$, $\mathcal{W}_{1} \triangleq \int_{\R} |F_0(\kappa) - F_1(\kappa)| \; d\kappa$. 
For \texttt{Old Adult}, $\hat{\mathcal{W}_1}=0.127$; for  \texttt{COMPAS}, which does not show systematic arbitrariness, $\hat{\mathcal{W}_1}=0.007$.

%% file: section/20-arbitrary/22-fairness/224-fairness-algo.tex
\section{Accounting for Self-Consistency}\label{sec:fairness:algorithms}

By definition, low $\hatsc$ signals that there is high $\hatvar$ (Section \ref{sec:var:sc}). 
It is therefore a natural idea to use variance reduction techniques to improve $\hatsc$ (and thus reduce arbitrariness).

As a starting point for improving $\hatsc$, we perform variance reduction with Breiman's \emph{bootstrap aggregation}, or \emph{bagging}, ensembling algorithm~\cite{breiman1996bagging}. 
Bagging involves bootstrapping to produce a set of $\boot$ models (Section~\ref{sec:fairness:prelim}), and then, for each test instance, producing an aggregated prediction $\pred_A$, which takes the majority vote of the $\pred_1, \ldots, \pred_\boot$ classifications. 
This procedure is practically effective for classifiers with high variance~\citep{breiman1996bagging, breiman1998ac}. However, by taking the majority vote, bagging embeds the idea that having slightly-better-than-random classifiers is sufficient for improving ensembled predictions, $\pred_A$. 
Unfortunately, there exist instances like Individual 2 (Figure~\ref{fig:vote}), where the classifiers in the ensemble are evenly split between classes. 
This means that bagging alone cannot overcome arbitrariness (Appendix~\ref{app:sec:algo:sc}).

To remedy this, we add the option to abstain from prediction if $\hatsc$ is low (Algorithm~\ref{algo:bagging-confidently}). 
A minor adjustment to (\ref{eq:hatsc}) accounts for abstentions, and a simple proof follows that Algorithm~\ref{algo:bagging-confidently} improves $\hatsc$ (Appendix~\ref{app:sec:algorithm}). 
We bootstrap as usual, but produce a prediction $\pred \in [0, 1]$ for instance $\instance$ only if $\instance$ surpasses a user-specified minimum level $\kappa$ of $\hatsc$; 
otherwise, if an instance fails to achieve $\hatsc$ of at least $\kappa$, we \texttt{Abstain} from predicting. 
For evaluation, we divide the test set into two subsets: we group together the instances we \texttt{Abstain} on in an \emph{abstention set} and those we predict on in a \emph{prediction set}. 
This method improves self-consistency through two complementary mechanisms: 1) variance reduction (due to bagging, see Appendix~\ref{app:sec:algorithm}) and 2) abstaining from instances that exhibit low $\hatsc$ (thereby raising the overall amount of $\hatsc$ for the prediction set, see Appendix~\ref{app:sec:algorithm}).\looseness=-1

Further, since variance is a component of error (Appendix~\ref{sec:fairness:significance}), variance reduction also tends to improve accuracy~\citep{breiman1996bagging}. 
This leads to an important observation: the abstention set, by definition, exhibits high variance; we can therefore expect it to exhibit higher error than the prediction set (Section~\ref{sec:fairness:empirical}). 
So, while at first glance it may seem odd that our solution for arbitrariness is to \emph{not predict}, it is worth noting that we often would have predicted incorrectly on a large portion of the abstention set, anyway (Appendix~\ref{app:sec:algorithm}). 
In practice, we test two versions of our method:\looseness=-1

\begin{algorithm}[!t]
\caption{$\hatsc$ Ensembling with Abstention}\label{algo:bagging-confidently}
{\small
\textbf{Input}: training dataset $(\mX, \vo)$, $\tproc$, $\boot$, $\hatsc$ $\kappa \in [0.5, 1]$, $\instance_\text{test}$\\
\textbf{Output}: $\pred$ with $\hatsc \geq \kappa$ or \texttt{Abstain}}%
\\
    \begin{algorithmic}[1]
    \State $\pred_A \coloneqq \mathsf{list}()$ $ \hspace{1em} \rhd$ To store ensemble predictions
    \For{$1 \ldots \boot$} 
        \State $\dataset_\boot \leftarrow \mathsf{Bootstrap}\big((\mX, \vo)\big)$
        \State $\rhd$ $\hat{\model}_{\dataset_\boot}$ can itself be a bagged model, with $\tproc$ bagging on
        \State \hspace{.2cm} $\dataset_\boot$ as the dataset to bootstrap \\\vspace{.1cm}
        \State $\hat{\model}_{\dataset_\boot} \leftarrow \tproc(\dataset_\boot)$ 
        \State $\pred_A.\mathsf{append}\big(\hat{\model}_{\dataset_\boot}(\instance_\text{test})\big) \hspace{1em} \rhd \pred_A = [\pred_{1}, \ldots, \pred_{\boot}]$
    \EndFor 
    \State \textbf{return} $\mathsf{Aggregate}(\pred_A, \kappa)$
    \vspace{.2cm}
    \State $\rhd$ Returns $\kappa$-majority prediction or abstains 
    \State \textbf{function $\mathsf{Aggregate}\big(\pred_{1}, \ldots, \pred_{\boot}, \kappa\big)$} 
        \State \hspace{.2cm} \textbf{if } $\mathsf{SelfConsistency}(\pred_{1}, \ldots, \pred_{\boot}) \geq \kappa$ \hspace{.1em} $\rhd$ Compute $\hatsc$ (\ref{eq:hatsc}) 
            \State \hspace{.4cm} \textbf{return} $\argmax_{y'\in\sY} \Big[\sum_{i=1}^\boot \1[y' = \pred_{i}]\Big]$
        \State \hspace{.2cm} \textbf{end if}
        \State \hspace{.2cm} \textbf{return} \texttt{Abstain}
    \State \textbf{end function}
\end{algorithmic}
\end{algorithm}

\paragraph{Simple ensembling.} We run Algorithm~\ref{algo:bagging-confidently} to build ensembles of typical hypothesis classes in algorithmic fairness. 
For example, running with $\boot=101$ decision trees and $\kappa=0.75$ produces a bagged classifier that contains $101$ underlying decision trees, for which the bagged classifier abstains from predicting on test instances that exhibit less than $0.75$ $\hatsc$. 
If overall $\hatsc$ is low, then simple ensembling will lead to a large number of abstentions. 
For example, almost half of all test instances in \texttt{COMPAS} using random forests would fail to surpass the threshold $\kappa=0.75$ (Figure~\ref{subfig:compas-cdf-rfc}). 
The potential for large abstention sets informs our second approach.\looseness=-1

\paragraph{Super ensembling.} 
We run Algorithm~\ref{algo:bagging-confidently} on \emph{bagged} models $\hat{\model}$. 
When there is low $\hatsc$ (i.e., high $\hatvar$) it can be 
beneficial to do an initial pass of variance reduction. 
We produce bagged classifiers using traditional bagging, but without abstaining (at Algorithm~\ref{algo:bagging-confidently}, lines 4-5);
\emph{then} we $\mathsf{Aggregate}$ using those bagged classifiers as the underlying models $\hat{\model}$. 
The first round of bagging raises the overall $\hatsc$ before the second round, which is when we decide whether to \texttt{Abstain} or not. 
We therefore expect this approach to abstain less; however, it may potentially incur higher error, if, by happenstance, simple-majority-vote bagging chooses $\hat{y}\neq o$ for instances with very low $\hatsc$ (Appendix~\ref{app:sec:algorithm}). 
We also experiment with an $\mathsf{Aggregate}$ rule that averages the output probabilities of the underlying regressors $\regressork$, and then applies threshold $\tau$ to produce ensembled predictions. 
We do not observe major differences in results.\looseness=-1

%% file: section/20-arbitrary/22-fairness/225-fairness-empirical.tex
\section{Experiments}\label{sec:fairness:empirical}

We release an extensible package of different $\mathsf{Aggregate}$ methods, with which we trained and compared several million different models (all told, taking on the order of $10$ hours of compute). 
We include results covering common datasets and models: \texttt{COMPAS}, \texttt{Old Adult}, \texttt{German} and \texttt{Taiwan Credit}, and 3 large-scale \texttt{New Adult - CA} tasks on logistic regression (LR), decision trees (DTs), random forests (RFs), MLPs, and SVMs. 
By using Algorithm~\ref{algo:bagging-confidently}, we happened to observe close-to-fairness in nearly every task (Section~\ref{sec:empirical-repro}).

\paragraph{Releasing an \texttt{HMDA} toolkit.} 
A possible explanation is that most fairness benchmarks are small ($<25,000$ examples) and therefore exhibit high variance. 
We therefore clean a larger, more diverse, and newer dataset for investigating fair binary classification --- the Home Mortgage Disclosure Act (\texttt{HMDA}) 2007-2017 datasets~\cite{ffiec2022housingdata} --- and release them with a standalone, easy-to-use software package. 
In this paper, we examine the \texttt{NY} and \texttt{TX 2017} subsets of \texttt{HMDA}, which have $244,107$ and $576,978$ examples, respectively, and we still find close-to-fairness  (Section \ref{sec:empirical-algo}).\looseness=-1 

\paragraph{Presentation.} To visualize Algorithm~\ref{algo:bagging-confidently}, we plot the CDFs of the $\hatsc$ of the underlying models used in each ensembling method. 
We simultaneously plot the results of simple ensembling (dotted curves) and super ensembling (solid curves). 
Instances to the left of the vertical line (the minimum $\hatsc$ threshold $\kappa$) form the abstention set. 
We also provide corresponding mean $\pm$ STD fairness and accuracy metrics for individual models (our expected, but not-necessarily-practically-attainable baseline) and for both simple and super ensembling. 
For ensembling methods, we report these metrics on the prediction set, along with the abstention rate ($\hatar$).\looseness=-1 

We necessarily defer most of our results to the online version of our paper.
Here, we exemplify two overarching themes: 
the effectiveness of both ensembling variants (Section~\ref{sec:empirical-algo}), and how our results reveal shocking insights about reliability in fair binary classification research (Section~\ref{sec:empirical-repro}). 
For all experiments, we illustrate Algorithm~\ref{algo:bagging-confidently} with $\kappa=0.75$, but note that $\kappa$ is task-dependent in practice.\looseness=-1 

\begin{figure*}[t!]
\begin{minipage}{.495\linewidth}
\centering
\hspace{-.4cm}
        \includegraphics[width=.85\linewidth]{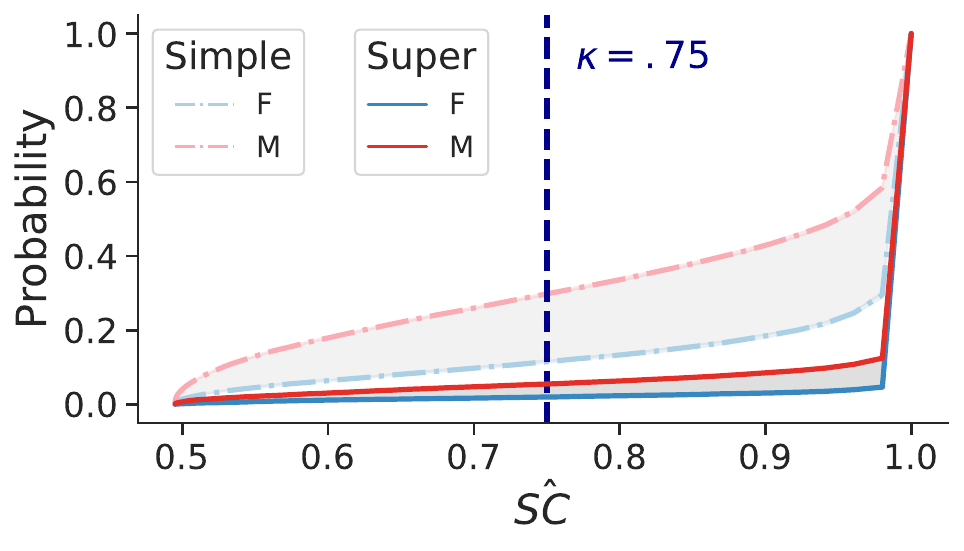}\\\vspace{.25cm}
        \small
        \begin{tabular}{lccc}
        \toprule
            & \textbf{Baseline} & \textbf{Simple} 
            & \textbf{Super}
            \\ \midrule
            \textbf{$\Delta\hatfnr$} & $\;\;6.3\pm.3\%$ & $4.1\pm.3\%$ 
            & $\;\;5.8\pm.4\%$ 
            \\ \midrule
             \textbf{$\hatfnr_\text{F}$} & $\;\;5.3\pm.3\%$ & $3.5\pm.1\%$ 
             & $\;\;4.9\pm.2\%$ 
             \\ \midrule 
            \textbf{$\hatfnr_\text{M}$} & $11.6\pm.1\%$ & $7.6\pm.3\%$  
            & $10.7\pm.3\%$ 
            \\ 
              \bottomrule
        \end{tabular}
        \subcaption{\texttt{Old Adult} split by $\texttt{sex}$; random forests (RFs)}
        \label{subfig:adult-ens}
\end{minipage}%
\hspace{.25cm}
\begin{minipage}{.495\linewidth}
\centering
\hspace{-.2cm}
        \includegraphics[width=.85\linewidth]{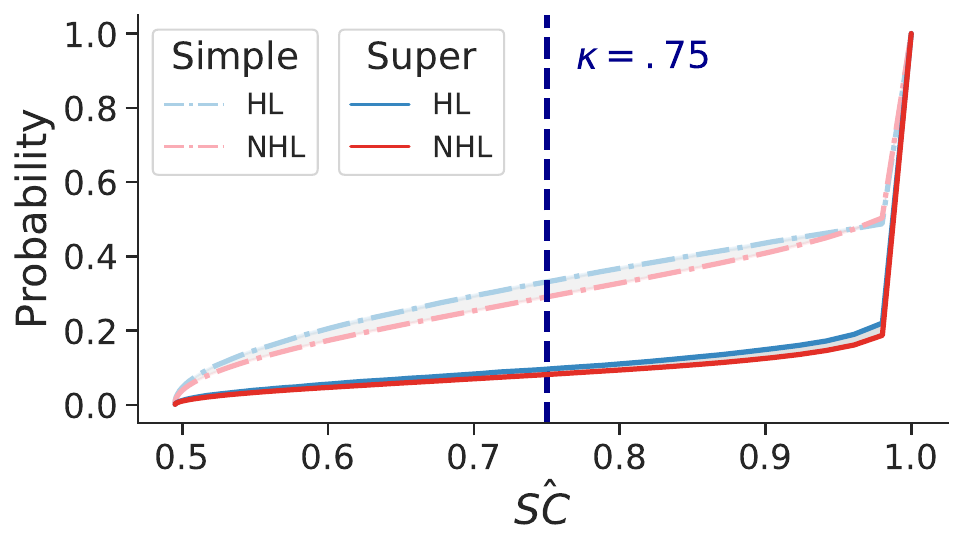}\\\vspace{.25cm}
        \small
        \begin{tabular}{lccc}
        \toprule
            & \textbf{Baseline} & \textbf{Simple} 
            & \textbf{Super}
            \\ \midrule
            \textbf{$\Delta\hatfnr$} & $\;\;0.7\pm.2\%$ & $1.1\pm.3\%$ 
            & $2.2\pm.3\%$ 
            \\ \midrule
             \textbf{$\hatfnr_\text{HL}$} & $10.1\pm.2\%$ & $3.3\pm.3\%$ 
             & $8.0\pm.3\%$ 
             \\ \midrule 
            \textbf{$\hatfnr_\text{NHL}$} & $\;\;9.4\pm.1\%$ & $2.2\pm.1\%$  
            & $5.8\pm.1\%$ 
            \\ 
              \bottomrule
        \end{tabular}
        \subcaption{\texttt{HMDA-NY-2017} split by $\texttt{ethnicity}$; random forests (RFs)}
        \label{subfig:hmda-ens}
\end{minipage}%
\caption{Algorithm~\ref{algo:bagging-confidently}: simple and super ensembling RFs for \texttt{Old Adult} (\ref{subfig:adult-ens}) and 
\texttt{HMDA-NY-2017} (\ref{subfig:hmda-ens}). 
Tables show $\hatfnr$ (mean $\pm$ STD) for individual models (Baseline) and each ensembling method's prediction set; $\boot=101$, 10 train/test splits (Appendix E). 
To highlight systematic arbitrariness (Section~\ref{sec:var:arbitrary}), we shade in gray the area between group-specific $\hatsc$ CDFs for each method. 
An initial pass of variance reduction in super significantly decreases the systematic arbitrariness in \texttt{Old Adult}.\looseness=-1} 
\label{fig:adult-hmda-ens}
\end{figure*}

\subsection{Validating Algorithm~\ref{algo:bagging-confidently}}\label{sec:empirical-algo}

\begin{figure}[t!]
\centering
    \begin{subfigure}{.495\linewidth}
    \centering
        \includegraphics[width=.95\linewidth]{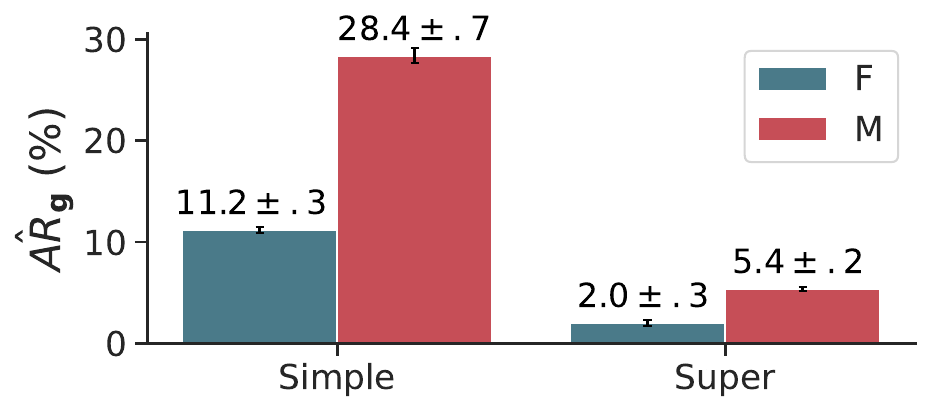}
        \caption{\texttt{Old Adult}, $\group=\texttt{sex}$}
        \label{fig:abstention-adult}
    \end{subfigure}
    \begin{subfigure}{.495\linewidth}
    \centering
       \includegraphics[width=.95\linewidth]{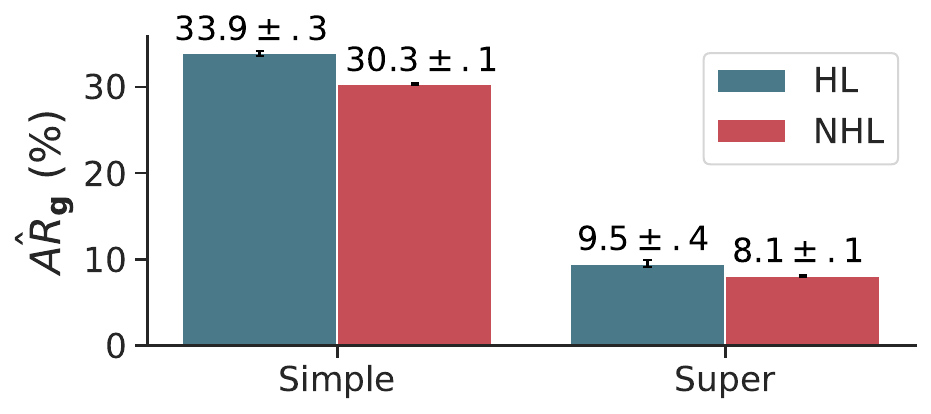}
        \caption{\texttt{HMDA-NY-2017}, $\group=\texttt{ethnicity}$}
    \end{subfigure}
\caption{Group-specific abstention rates $\hatar_\group$. 
Super ensembling abstains less and more equally than simple ensembling.} 
\label{fig:abstention}
\end{figure}

We highlight results for two illustrative examples: \texttt{Old Adult} and \texttt{HMDA\--NY\--2017}, for \texttt{ethnicity} (Hispanic or Latino (HL), Non\--Hispanic or Latino (NHL)). 
We plot $\hatsc$ CDFs and show $\hatfnr$ metrics using random forests (RFs). 
For \texttt{Old Adult}, the expected disparity of the RF baseline is $\Delta\hatfnr=6.3\%$. 
The dashed set of curves plots the underlying $\hatsc$ for these RFs (Figure~\ref{subfig:adult-ens}). 
When we apply simple to these RFs, overall $\haterr$ decreases, shown in part by the decrease in $\hatfnr_\text{F}$ and $\hatfnr_\text{M}$. 
Fairness also improves:  
$\Delta\hatfnr$ decreases to $4.1\%$. 
However, the corresponding $\hatar$ is quite high, especially for the \texttt{Male} subgroup ($\group=\text{M}$, Figure~\ref{fig:abstention}). 

As expected, super improves overall $\hatsc$ through a first pass of variance reduction (Section~\ref{sec:fairness:algorithms}).  
The $\hatsc$ CDF curves are brought down, indicating a lower proportion of the test set exhibits low $\hatsc$. 
Abstention rate $\hatar$ is lower and more equal (Figure~\ref{fig:abstention}); however, error, while still lower than the baseline RFs, has gone up for all metrics. 
There is also a decrease in systematic arbitrariness (Section~\ref{sec:var:arbitrary}): 
the dark gray area for super ($\hat{\mathcal{W}_1}=.014$) is smaller than the light gray area for simple ($\hat{\mathcal{W}_1}=.063$) (Appendix~\ref{app:sec:consistency}).\looseness=-1

For \texttt{HMDA} (Figure~\ref{subfig:hmda-ens}), simple similarly improves $\hatfnr$, but has a less beneficial effect on fairness ($\Delta\hatfnr$). 
However, note that since the baseline is the empirical expected error over thousands of RF models, the specific $\Delta\hatfnr$ is not necessarily attainable by any individual model. 
In this respect, simple has the benefit of actually obtaining a specific (ensemble) model that yields this disparity reliably in practice: 
$\Delta\hatfnr=1.1\%$ is the mean over $10$ simple ensembles. Notably, this is extremely low, even without applying traditional fairness techniques. 
Similar to \texttt{Old Adult}, simple exhibits high $\hatar$, which decreases with super at the cost of higher error. 
$\hatfnr$ still improves for both $\group$ in comparison to the baseline, but the benefits are unequally applied: 
$\hatfnr_\text{W}$ has a larger benefit, so $\Delta\hatfnr$ increases slightly.

\paragraph{Abstention set error.} As an example, the average total 
$\haterr$ in the \texttt{Old Adult} simple abstention set is close to $40\%$ --- compared to $17\%$ for the RF baseline, and $8\%$ for simple and $14\%$ for super prediction sets. 
As expected, beyond reducing arbitrariness, we abstain from predicting for many instances for which we also would have been more inaccurate (Section~\ref{sec:fairness:algorithms}).\looseness=-1 

\paragraph{A trade-off.} 
Our results support that there is indeed a trade-off between abstention rate and error (Section~\ref{sec:fairness:algorithms}). 
This is because Algorithm~\ref{algo:bagging-confidently} identifies low-$\hatsc$ instances for which ML prediction does a poor job, and abstains from predicting on them. 
Nevertheless, it may be infeasible for some applications to tolerate a high $\hatar$. 
Thus the choice of $\kappa$ and ensembling method should be considered a context-dependent decision.\looseness=-1 

\paragraph{Unequal abstention rates.} 
When there is a high degree of systematic arbitrariness, $\hatar$ can vary a lot by $\group$ (Figure~\ref{fig:abstention}). 
With respect to improving $\hatsc$, error, and fairness this may be a reasonable outcome: 
it is arguably better to abstain unevenly --- deferring a final classification to non-ML decision processes --- than to predict more inaccurately and arbitrarily for one group. 
More importantly, we rarely observe systematic arbitrariness; 
unequal $\hatar$ is uncommon in practice (Section~\ref{sec:fairness:related}).\looseness=-1

\subsection{A Problem of Empirical Algorithmic Fairness}\label{sec:empirical-repro}

\begin{figure}[t!]
    \centering
    \includegraphics[width=.6\linewidth]{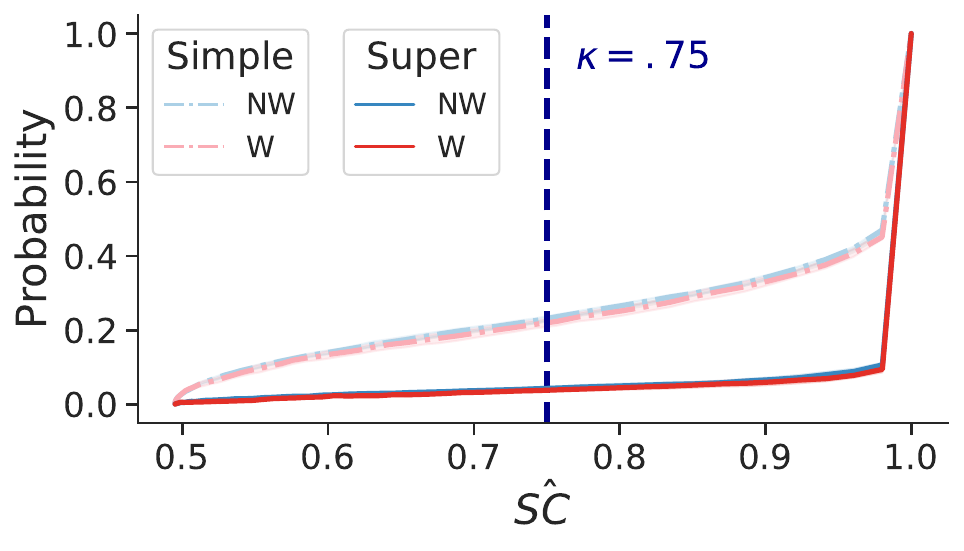}\\\vspace{.25cm}
    \begin{tabular}{lccc}
        \toprule
            & \textbf{Baseline} & \textbf{Simple} 
            & \textbf{Super} 
            \\ \midrule
            \textbf{$\Delta\hatfpr$} & $\;\;2.1\pm1.8\%$ & $\;\;3.0\pm1.4\%$ 
            & $\;\;\;1.8\pm1.0\%$ 
            \\ \midrule
             \textbf{$\hatfpr_\text{NW}$} & $14.7\pm1.3\%$ & $11.4\pm1.0\%$ 
             & $12.9\pm.8\%$ 
             \\ \midrule 
            \textbf{$\hatfpr_\text{W}$} & $12.6\pm1.3\%$ & $\;\;8.4\pm1.0\%$ 
            & $11.1\pm.6\%$ 
            \\ 
              \bottomrule
    \end{tabular} 
    \caption{Algorithm~\ref{algo:bagging-confidently},  
    LR on \texttt{COMPAS}. $\boot=101$, 10 train/test splits. Table shows mean $\hatfpr$ $\pm$ STD for individual models (Baseline) and ensembling methods' prediction sets.} 
    \label{fig:compas-ens}
\end{figure}

We also highlight results for \texttt{COMPAS}, 1 of the 3 most common fairness datasets~\citep{fabris2022datasets}.  
Algorithm~\ref{algo:bagging-confidently} is similarly very effective at reducing arbitrariness (Figure~\ref{fig:compas-ens}), and is able to obtain state-of-the-art accuracy~\cite{lin2020compas} with $\Delta\hatfpr$ between $1.8-3\%$. 
Analogous results for \texttt{German Credit} indicate statistical equivalence in fairness metrics (online Appendix). 
These low-single-digit 
disparities do not cohere with much of the literature, which often reports much larger fairness violations~\citep[notably]{larson2016propublica}. 
However, most work on fair classification examines individual models, selected via cross-validation with a handful of random seeds (Section~\ref{sec:fairness:prelim}). 
Our results suggest that selecting between a few individual models in fair binary classification experiments is unreliable. 
When we instead estimate expected error by ensembling, we have difficulty reproducing unfairness in practice. 
Variance in the underlying models in $\hat{\possiblemodels}$ seems to be the culprit. 
The individual models we train 
exhibit radically different group-specific error rates. 
Our strategy of shifting focus to the overall behavior of $\hat{\possiblemodels}$ provides a solution: 
we not only mitigate arbitrariness, we also improve accuracy \emph{and} usually average away most underlying, individual-model unfairness. 
\looseness=-1  

%% file: section/20-arbitrary/22-fairness/226-fairness-related.tex
\section{Discussion and Related Work}\label{sec:fairness:related}

In this paper, we advocate for a shift in thinking about \emph{individual} models to the \emph{distribution over possible models} in fair binary classification.
This shift surfaces arbitrariness in underlying model decisions. 
We suggest a metric of \emph{self-consistency} as a proxy for arbitrariness (Section~\ref{sec:fairness:significance}) and an intuitive, elegantly simple extension of the classic bagging algorithm to mitigate it (Section~\ref{sec:fairness:algorithms}).
Our approach is tremendously effective with respect to improving $\hatsc$, accuracy, and fairness metrics in practice (Section~\ref{sec:fairness:empirical}).

Our findings complicate accepted truths in algorithmic fairness. 
For example, much work posits that there is an inherent analytical trade-off between fairness and accuracy~\citep{corbettdavies2017cost, menon2018cost}.
Instead, our experiments complement prior work that disputes the practical relevance of this formulation~\citep{rodolfa2021tradeoff}. 
We show it is in fact typically possible to achieve accuracy (via variance reduction) and close-to-fairness --- and to do so without using fairness-focused interventions.\looseness=-1

Other research also highlights 
the need for metrics beyond fairness and accuracy. 
Model multiplicity reasons about sets of models that have similar accuracy~\citep{breiman2001multiplicity}, but differ in underlying properties due to variance in decision rules~\citep{black2022multiplicity, marx2020mm}. 
This work emphasizes developing criteria for 
selecting an \emph{individual} model from that set.   
Instead, our work uses the  \emph{distribution over possible models} (with no normative claims about model accuracy 
or other 
 criteria) to reason about arbitrariness (App~C.3). 
Some related work considers the role of uncertainty and variance in fairness~\citep{chen2018tradeoff, khan2023fairness}.  
Notably, Black et al.~\cite{black2022selective} concurrently investigates abstention-based ensembling, 
employing a strategy that (based on their choice of variance definition) 
ultimately does not 
address the arbitrariness we describe and mitigate (Appendix~\ref{app:sec:variance}).\looseness=-1 

Most importantly, we take a comprehensive experimental approach missing from prior work. 
It is this approach that uncovers our alarming results: 
almost all tasks and settings demonstrate close-to or complete statistical equality in fairness metrics, after accounting for arbitrariness (\S E.4).
\texttt{Old Adult} (Figure~\ref{subfig:adult-ens}) is one of two exceptions. 
These results hold for larger, newer datasets like \texttt{HMDA}, which we clean and release. 
Altogether, our findings indicate that variance is undermining the reliability of conclusions in fair binary classification experiments. 
It is worth revisiting all prior experiments that depend on cross validation or few models.\looseness=-1 

\paragraph{The future of fairness research.} 
While the field has put forth numerous theoretical results about (un)fairness regarding single models --- impossibility of satisfying multiple metrics~\citep{kleinberg2017impossibility}, post-processing individual models to achieve a particular metric~\citep{hardt2016equality} --- these results seem to miss the point. 
By examining individual models, arbitrariness remains latent; 
when we account for arbitrariness in practice, most measurements of unfairness vanish. 
We are not suggesting that there are no reasons to be concerned with  fairness of ML models. 
We are not challenging the idea that actual, reliable violations of standard fairness metrics should be of concern. 
Instead, we are suggesting that common formalisms and methods for measuring fairness can conceal a tremendous amount of arbitrariness, which should itself be an important concern 
when examining the social impact of automated decision-making.

%% file: section/20-arbitrary/23-nondeterminism/23-nondeterminism-main.tex
\chapter{Non-Determinism and the Lawlessness of Machine Learning Code}\label{chapter:nondeterminism}

The types of arbitrariness that we describe in the prior chapters can lead to outcomes that have broader social impact, when machine-learning techniques are taken up to inform decision processes in the real world.
This chapter represents early work on translating the importance of these particular types of arbitrariness for a law and policy audience, with particular attention to algorithmic fairness contexts (Chapter~\ref{chapter:fairness}). 
Over time, through the development of follow-on work (Chapter~\ref{chapter:fairness}, Cooper et al.~\cite{cooper2024variance}), we have come to believe that this relationship is more significant than suggested in this chapter.
In particular, there remains important legal-theory work to more fully address the arbitrariness induced by stochasticity. 
We have begun follow-on work in this area, which we hope to complete in early 2025.\\ 

\noindent \textbf{Chapter summary}: Legal literature on machine learning (ML) tends to focus on harms, and thus tends to reason about individual model outcomes and summary error rates. This focus has masked important aspects of ML that are rooted in its reliance on randomness --- namely, \emph{stochasticity} and \emph{non-determinism}. While some recent work has begun to reason about the relationship between stochasticity and arbitrariness in legal contexts, the role of non-determinism more broadly remains unexamined. In this paper, we clarify the overlap and differences between these two concepts, and show that the effects of non-determinism, and consequently its implications for the law, 
become clearer from the perspective of reasoning about ML outputs as \emph{distributions over possible outcomes}. This distributional viewpoint accounts for randomness by emphasizing the \emph{possible} outcomes of ML. Importantly, this type of reasoning is not exclusive with current legal reasoning; it complements (and in fact can strengthen) analyses concerning individual, concrete outcomes for specific automated decisions. By illuminating the important role of non-determinism, we demonstrate that ML code falls outside of the cyberlaw frame of treating ``code as law,'' as this frame assumes that code is deterministic. We conclude with a brief discussion of what work ML can do to constrain the potentially harm-inducing effects of non-determinism, and we indicate where the law must do work to bridge the gap between its current individual-outcome focus and the distributional approach that we recommend.\\

\noindent This chapter is a licensed derivative copy of work published and awarded a Long Presentation slot at \emph{CSLAW 2022}~\cite{cooper2022lawless}. A longer version of this work, building on results in Cooper et al.~\citep{cooper2024variance}, was presented at \emph{PLSC 2023} and is in development for law-review submission. 

\input{section/20-arbitrary/23-nondeterminism/231-nondeterminism-intro}
\input{section/20-arbitrary/23-nondeterminism/232-nondeterminism-ml}
\input{section/20-arbitrary/23-nondeterminism/233-nondeterminism-law}
\input{section/20-arbitrary/23-nondeterminism/234-nondeterminism-conclusion} 

%% file: section/20-arbitrary/23-nondeterminism/231-nondeterminism-intro.tex
\section{Introduction}

Legal decision logic bears some resemblance with the logic of mathematical functions in that both involve procedures for mapping inputs to outputs. 
When adjudicating a particular case, a magistrate assembles the available evidence, which they supply as parameters to legal rules to inform decisions. 
Just as with mathematical functions, there can be variations in input parameters, which correspond to variations in outcomes.\footnote{For more general background on how legal rules function despite variation in their application, we refer the reader to Fuller~\cite{fuller1965law} and Tamanaha~\cite{tamanaha2004law}.}
Kolber~\cite{kolber2014smoothbumpy} takes this functional analogy a step further, classifying the correspondences between legal inputs and outputs into ``smooth'' and ``bumpy'' types. 
A smooth relationship is one for which gradual changes in inputs map to gradual changes in outputs. Bumpy relationships, in contrast, exhibit discontinuities: 
slight variations in inputs can map to large variations in outputs.\footnote{For example, it may be reasonable to contend that tort law should be smooth, with the amount of harm caused exhibiting a direct and continuous relationship with the degree of compensation owed. 
    However, in practice, tort law is often bumpy: defendants are either liable to provide full compensation (regardless of the particular degree of contributing to harm), or they are not liable at all~\cite[p. 673]{kolber2014smoothbumpy}.} 
Machine learning (ML) --- a discipline within the mathematical tradition --- unsurprisingly seems to follow a similar logic. 
Classification problems resemble Kolber~\cite{kolber2014smoothbumpy}'s concept of bumpiness; varied, continuous inputs become discretized outputs. 
Determining loan-worthiness, for example, is bumpy because a classification model maps personal data to a binary outcome in the set $\{\texttt{grant\_loan}, \texttt{reject\_loan}\}$, typically based on some underlying notion of whether the individual under consideration is likely to repay or default.

This comparison between the work of law and that of ML, in which both are reasoned about as functions, is deceptively attractive. 
At first glance, it seems to mirror the decades-long literature in cyberlaw that has considered the law and if/then code rules\footnote{Either as a type of architecture~\cite{lessig1999horse, lessig2009code} or a modality on its own~\cite{grimmelmann2005reg}.} 
to be complementary modalities that regulate and mediate human experience~\cite{lessig1999horse, grimmelmann2005reg, bamberger2010risk, citron2008dueprocess,  lessig2009code, reidenberg1997code}. 
It is thus perhaps intuitive to consider stretching this analogy further: 
to treat the mathematical-functional similarity of the law and ML as a rationale for christening ML as the latest type of code-imbued regulator. 
To stretch this even further, if ML can be fashioned to design new ``microdirectives'' or usher in a new era of ``personalized law,'' as some legal scholars contend~\cite{casey2015rulesandstandards, fagan2019discretion}, then perhaps ML could breathe new life into the succinct cyberlaw refrain that ``code is law''~\cite{reidenberg1997code, lessig2009code}. 
That is, rather than using this widely-quoted shorthand to stand in for the more-precise (but still abbreviated) ``code is constitutive of law''~\cite[p. 675]{bamberger2010risk}, ML code could literally be used to generate law.

And yet, while it might be appealing to take these steps to connect the nascent field of ML law with its older cyberlaw sibling, upon deeper examination the comparison between ML and the law via functions does not hold up. 
For one, as much legal scholarship acknowledges, the mechanism by which ML translates from inputs to outputs fundamentally differs from analogous mechanisms in the law~\cite{mulligan2018governance, mulligan2019ml, citron2014scored, kroll2017aa, hausman2021rigged, barocas2016data, citron2022privacy}. 
The law has a variety of mechanisms --- rules, standards, factors tests, etc. --- each accompanied with justifications for (and amendments regarding) their use, as well as a long record in jurisprudence of their application to specific cases. 
In contrast, ML may behave like a function, but we often do not understand how that function works. 
In ML systems, we can have full access to both the inputs and subsequent outputs, while having no clear understanding of \emph{how} the mapping from one to the other occurred. 
In other words, unlike the law, ML functions defy explanation and reasonable justification, which in turn raises fundamental questions about the legitimacy of using ML as a decision-making tool and muddies the ability to determine accountability when these tools cause harms~\cite{cooper2022accountability, creel2022leviathan}.\footnote{Clarity of explanation in legal contexts, however, is not a given. 
    As Fuller~\cite{fuller1965law} notes, ``It is easy to assert that the legislator has a moral duty to make his laws clear and understandable. But this remains at best an exhortation unless we are prepared to define the degree of clarity he must attain in order to discharge his duty. The notion of subjecting clarity to quantitative measure presents obvious difficulties. We may content ourselves, of course, by saying that the legislator has at least a moral duty to try to be clear. But this only postpones the difficulty, for in some situations nothing can be more baffling than to attempt to measure how vigorously a man intended to do that which he has failed to do. ... [However,] good intentions are of little avail. ... All of this adds up to the conclusion that the inner morality of law is condemned to remain largely a morality of aspiration and not of duty. Its primary appeal must be to a sense of trusteeship and to the pride of the craftsman''~\cite[pp. 42-43]{fuller1965law}. 
    It is reasonable to argue, though out of scope for this paper, that ML does not have an analogous ``sense of trusteeship'' on which the public can rely.} 
In short, ML's problem with \textit{explainability} shows how the analogy essentially and inescapably falls short; both the law and ML may behave like functions, but functions that are fundamentally different in kind.

This analogy falls short in another fundamental way --- one that is significant enough for us to pause attempting to close the loop between cyberlaw, code-is-law scholarship and legal scholarship about ML, but has thus-far remained under-explored. 
Code that follows if/then logic --- the type of code addressed in cyberlaw literature~\cite{lessig1999horse, grimmelmann2005reg, bamberger2010risk, citron2008dueprocess} --- is \emph{deterministic}: 
it specifies behaviors to execute (the ``then'') when certain, specified conditions (the ``if'') are met. 
Importantly, ML code does not execute if/then rules. 
Instead, the ML training process is random in nature; it exhibits \emph{stochasticity} and \emph{non-determinism}.\footnote{Non-determinism and stochasticity are not unique to ML, but rather are features of many types of randomized programs (including programs and protocols that predate the Internet and cyberlaw). 
    Nevertheless, the advent of ML applications in public life, and the social valences these applications carry, has brought urgency to clarifying these concepts in relation to ML.} 
We explore the meaning of these terms in detail later in this paper (Section~\ref{sec:nondeterminism:ml}). 
For now, it suffices to provide an intuition: deterministic code ensures that computing with the same inputs yields the same outputs; stochasticity and non-determinism, in contrast, can cause two similar training procedures to produce vastly different results in practice~\cite{forde2021model, qian2021variance, cooper2024variance}. 

In the remainder of this chapter, we explain how stochasticity and non-determinism play a fundamental role in the behavior of ML systems. 
While some legal scholarship has begun to reason about the relationship between stochasticity and arbitrariness~\cite{creel2022leviathan, bambauer2022diff}, the role of non-determinism more generally remains unexamined. 
We argue that a more precise understanding of non-determinism is essential for reasoning about questions concerning the regulability, legitimacy, and accountability of ML decision-making tools. 

Our first contribution is to show that the emphasis on individual errors and error rates in existing legal scholarship is concealing other important issues in ML that are rooted in non-determinism. 
While focusing on individual outcomes and error rates for specific models is important --- and intuitive, given that it parallels case-based analysis in the law --- it nonetheless provides a limited view of behavior of ML. 
We clarify the distinction between stochasticity and non-determinism more broadly construed, and show that the effects of non-determinism, and consequently its implications for the law, instead become clearer from the perspective of reasoning about ML outputs as \emph{distributions or patterns over possible outcomes}. 
The key difference is that this viewpoint accounts for randomness and other types of non-determinism by providing a window into the \emph{possible} outcomes of ML. 
Importantly, this type of reasoning is not exclusive with current legal reasoning; it complements (and in fact can strengthen) analyses of individual, concrete outcomes for specific automated decisions (Section~\ref{sec:nondeterminism:ml}). 

By illuminating the important role and potential effects of non-determinism, we then demonstrate that ML code falls outside of the cyberlaw frame, which assumes deterministic code (Section~\ref{sec:nondeterminism:law}).
Even if this frame can be expanded to include the stochastic elements of ML, we discuss how it cannot be extended to non-deterministic elements more generally.  
Lastly, we conclude with a brief discussion of what work ML can do to constrain the potentially harm-inducing effects of non-determinism, and we indicate where the law must do work to bridge the gap between its current case-based analysis of ML systems and the pattern/distributional analysis that we recommend (Section~\ref{sec:nondeterminism:conclusion}).

%% file: section/20-arbitrary/23-nondeterminism/232-nondeterminism-ml.tex
\section{Non-determinism and Stochasticity}\label{sec:nondeterminism:ml}

Legal literature regarding the empirical performance of ML tools tends to focus on issues of accuracy~\cite{lehr2017legalml}~\cite[pp. 1249-50]{brennanmarquez2019plausiblecause}~\cite[pp. 9,12]{calo2021modeling}~\cite[p. 1253]{citron2008dueprocess}.\footnote{Work on fairness typically focuses on accuracy, as well, by emphasizing differences in inaccuracy via error rates, and the resulting disparate impact, for protected demographic groups.} 
This work typically evaluates ML in terms of individual decision outcomes in relation to the harms these outcomes cause, and uses summary error rates to draw conclusions about a particular model's accuracy. 
Solely focusing on the accuracy of specific inference outcomes and summary rates can conceal other important issues implicated by non-determinism, which are also important factors to consider in legal analyses of ML technology. 
To make this case, we first must establish definitions for non-determinism and stochasticity, as there are nuanced differences and overlap between the two terms.

\begin{definition}
\label{def:nondeterminism}
\textbf{Non-determinism} is a property of processes for which supplying the same inputs can produce different outputs.
\end{definition}

As a result, non-deterministic outcomes are uncertain. 
This is in contrast to deterministic if/then logic, for which the same inputs produce the same outputs. 
Stochasticity also satisfies Definition~\ref{def:nondeterminism}; however, it places additional conditions on the form that uncertainty can take.

\begin{definition}
\label{def:stochasticity}
\textbf{Stochasticity} is a property of non-deterministic processes whose outcomes can be reasoned about using probability theory.
\end{definition}

In other words, the non-determinism of stochasticity specifically comes from randomization that can be understood using probability. 
Following these definitions, we can think of stochastic decision-making processes as non-deterministic; 
however, non-deterministic decision-making processes are not necessarily stochastic, since they cannot always be reasoned about using the laws of probability. 

Machine learning is grounded in probability and statistics, and thus is fundamentally stochastic in nature. 
In practice, however, it is also common for ML to exhibit non-determinism beyond this stochasticity. 
While the formal specification for an algorithm is stochastic, its implementation and execution in software and hardware can introduce non-determinism that is not stochastic. 
We can attempt to apply the rules of probability to reason about this behavior, but we are not guaranteed that our conclusions will be sound. 
A notable example of this non-determinism comes from the popular PyTorch library.\footnote{We refer to \textcolor{blue}{\href{https://pytorch.org/docs/stable/notes/randomness.html}{PyTorch}} for discussion about limiting the sources of software and hardware non-determinism in ML training pipelines. 
    At the time of writing, PyTorch offers a ``deterministic mode'' that, at the cost of significant run-time slowdowns that may not be feasible for all application developers, enforce determinism in software operations (where possible).} 
When prepared for execution on a computer at training time, PyTorch makes dynamic choices regarding how to run the code, which optimize for run-time speed and, in doing so, introduce non-stochastic non-determinism to the learning process.

\subsection{Related Work: ML Stochasticity and the Law}\label{sec:nondeterminism:prior}

The legal literature that discusses uncertainty and subsequent impressions of arbitrariness in ML decision-making does not reckon with this practical reality.
Rather, in talking about algorithms, and more specifically their error rates or individual outcomes, this literature regards ML in stochastic terms. 
For example, Bambauer et al.~\cite{bambauer2022diff} coins the term ``Small Change Makes a Big Difference'' (SCMBD) to analyze the risks to due process that can come from disproportionate outcomes on similar inputs due to the stochastic nature of ML training pipelines~\cite[pp. 2378-2383, pp. 2396-2397]{bambauer2022diff}. 
Their discussion makes no mention of how other sources of non-determinism further expand this category of risk.

In another recent example, Creel and Hellman~\cite{creel2022leviathan} take a formal philosophical approach to understanding what is precisely connoted by  criticisms of ``arbitrariness'' in ML system outputs. 
They break down their analysis of what is arbitrary in three different respects: ``unpredictable,'' ``unconstrained,'' and ``unreasonable'' behaviors of these systems.\footnote{In their discussion of ``arbitrary'' as ``unconstrained,'' Creel and Hellman~\cite[p. 3-4]{creel2022leviathan} call algorithms ``rule-based'' in close proximity to discussing legal rules and standards. 
    We do not believe that stochastic algorithms are ``rule-based'' in the same sense as legal rules; 
    however, discussing this distinction is out of scope for this paper.
} 
In their discussion, they claim that arbitrariness of ML systems in itself is not the problem; rather, the problem is ``the systematicity of their arbitrariness'' that may ``irrationally [exclude] a person from a significant number of important opportunities''~\cite[p. 2]{creel2022leviathan}.\footnote{douek~\cite{douek2021moderation} makes a related but different point about shifting legal understanding away from individual outcomes. She call for a shift ``from an individualistic approach to a probabilistic one''~\cite[p. 789]{douek2021moderation}. 
    douek makes an important intervention regarding the inevitability of error in ML applications, particularly at scale, but ultimately focuses on individual model error rates and makes an argument predicated on the ability to reason about probabilities, and thus is not examining the same concepts with which we concern ourselves here.}  
In relation to this claim, they add ``To the extent that an algorithm governs the decision, it will produce the same result when run on the same inputs. If the \textbf{algorithm} contains a degree of \textbf{randomness} within it, ... it is still \textbf{reproducible} at a higher level of abstraction''~\cite[pp. 3-4, emphasis added]{creel2022leviathan}. That is, they describe a \emph{model} demonstrating deterministic behavior. A particular model produced from an algorithmic learning procedure is deterministic --- always producing the same output given the same input (Section~\ref{sec:nondeterminism:example1}); however, as we have discussed above, the entire procedure that produces such a model is \emph{not} deterministic. 

Put differently, implicit in the reasoning in Creel and Hellman~\cite{creel2022leviathan} is that the uncertainty at play in ML can be reasoned about using probability. 
It is probability theory that enables the systematic, ``higher level of abstraction'' of reasoning about the overall, expected behavior of stochastic \emph{algorithms}, and whether those behaviors are systematically, arbitrarily unfair (according to a particular fairness criterion). However, in contrast to abstract algorithm specifications, the implemented, run-time behavior of ML \textit{pipelines} and \textit{systems} introduces non-stochastic non-determinism --- non-determinism that is \emph{not} systematic, in the sense that it cannot be reasoned about analytically with the guarantees of probability theory. This is not a distinction without import; in contrast to Creel and Hellman's claim about reproducibility in relation to what we understand as stochastic-related arbitrariness, this kind of non-determinism is a well-known contributor to the reproducibility crisis in ML~\cite{raff2019reproducibility, bouthillier2019reproducibility}. 
Non-stochastic non-determinism thus suggests a different kind of arbitrariness from that discussed in Creel and Hellman~\cite{creel2022leviathan}, and it, too, can have significant impacts on normative concerns like fairness~\cite{qian2021variance} (Section~\ref{sec:nondeterminism:example2}). \\

\noindent In short, though prior legal literature on ML and arbitrariness sometimes engages with elements of stochasticity, it does not account for the role of other forms of non-determinism. 
In the remainder of this section, we 
explain via simple synthetic examples how the presence of non-determinism calls into question essential assumptions about the fundamental nature of accuracy in ML. Moving away from analyses of individual outcomes to thinking about \emph{distributions/patterns over possible outcomes} can expand legal scholars' understanding of the behavior of ML tools. 
In particular, reasoning about \textit{probability} distributions over possible outcomes is useful for understanding the impacts of stochasticity (Section~\ref{sec:nondeterminism:example1}). 
While probability is not similarly useful for analytically reasoning about other sources of non-determinism, our approach can still highlight empirically the importance of the role of non-determinism in ML and the potential harms it can cause (Section~\ref{sec:nondeterminism:example2}). 

\subsection{Distributions over Individual Outcomes}\label{sec:nondeterminism:example1}

\begin{figure}[t!]
  \begin{center}
    \includegraphics[width=0.45\textwidth]{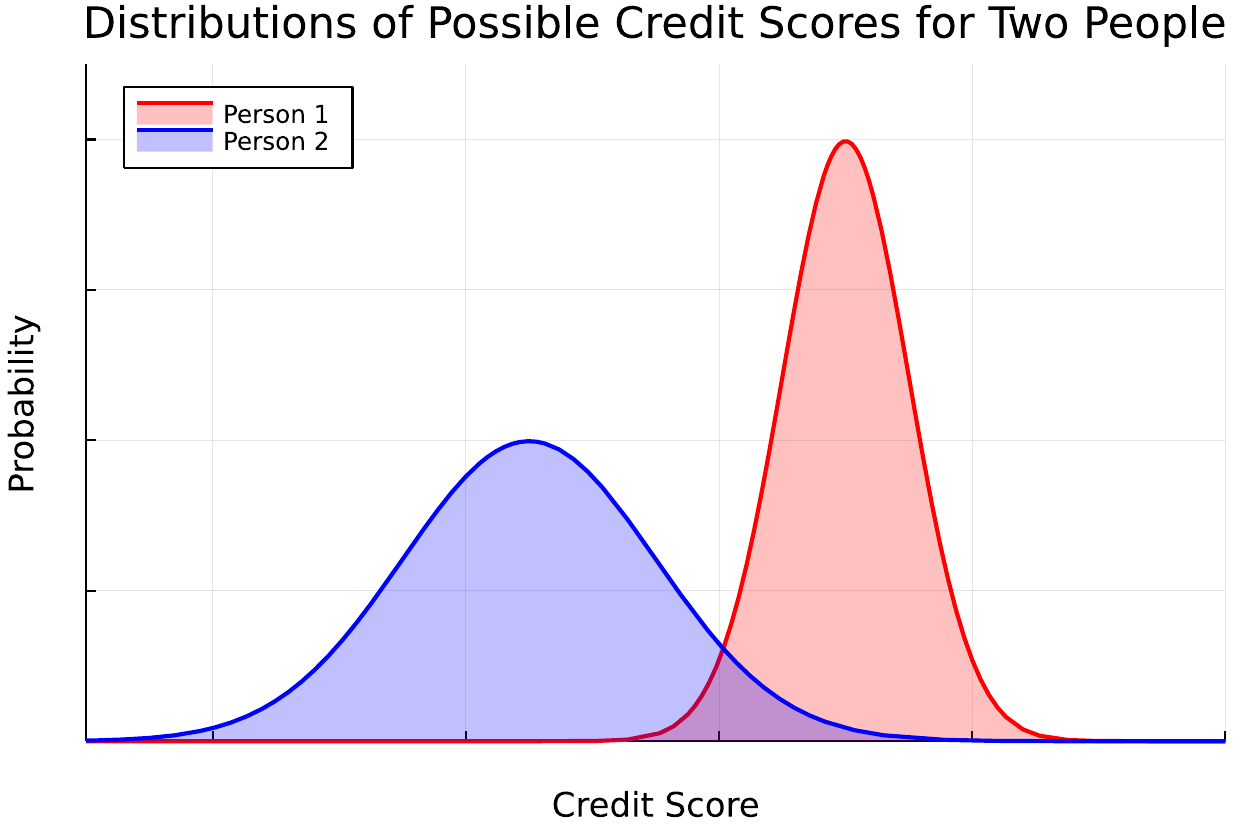}
    \caption{Synthetic probability distributions for possible predicted credit scores of two different individuals.}
	\label{fig:loans}
  \end{center} 
\end{figure}

We first consider a synthetic ML system that aims to determine individuals' creditworthiness by predicting their credit scores. 
The developers write a snippet of code to achieve this task --- a procedure for training models to predict individuals' credit scores. 
The execution of this code to actually train a model exhibits stochasticity: running this one piece of code multiple times on different subsets of the training data will result in multiple trained models that vary in comparison to one another. 
If we were to take many such models and supply them with the same individual as input, the corresponding outputs would yield a distribution over possible credit score outcomes for that individual. 
We illustrate this in Figure~\ref{fig:loans} for two individuals. 
In other words, since this process yields a distribution over possible credit scores for each individual --- and not just a single credit score --- predicting an individual's credit score is not a deterministic function of the code written by the engineer to train ML models. 
Rather, credit score for an individual is a function of the procedure that this code can execute; 
it is a function of executing model training, which exhibits stochasticity (as a function of the specific training data examples used for training) and thus a distribution of possible outcomes for different individuals.

The viewpoint of distributions over possible outcomes shown in Figure~\ref{fig:loans} illustrates a problem: 
the two individuals have overlapping credit score distributions  (shown in purple). This means that it is possible that there is some subset of models, produced by the stochastic training process, for which we cannot distinguish between these two individuals in terms of their credit scores. 
And yet, in looking at each of their distributions overall, there are all clearly cases where they do not overlap and are thus clearly distinguishable. 
That is, from its distributional perspective, this figure shows that it is possible to produce models that suggest contradictory results: 
some models are able to distinguish these individuals via different credit scores, while it is possible that some models are \emph{not} able to discern a difference. 
Instead of all models from this training process having the ability to clearly distinguish between or to equate these two individuals via credit scores, both contradictory possibilities are suggested by this distributional viewpoint. 

This ambiguity complicates what accuracy means for a model produced by this training process, because it is not clear what a ``correct'' model should do with respect to how it views these two individuals. 
Is it ``correct'' to model them as distinguishable, or ``correct'' to model them as indistinguishable, in terms of their credit scores? 
It is impossible to say with 100\% certainty, since there is no notion of ground truth credit score.\footnote{This is in contrast to applications for which we can reasonably say that there is a ground truth, such as a computer vision system that distinguishes between cats and dogs; an example input is either a cat or a dog, not both.} 
Put differently, this figure indicates that there is a meta-problem of not being able to draw a firm line between correctness and incorrectness for models trained by this process. 
This issue of being unable to draw a clear boundary between correctness and incorrectness illustrates how model output decisions can exhibit non-determinism: 
for the different inputs, depending on the model, the outputs for those inputs may be distinguishable or may be indistinguishable. 

So far, we have limited our discussion of non-determinism to stochasticity (Definition~\ref{def:stochasticity}) --- in particular, the stochasticity resulting from training models on different subsets of the training data or from different examples drawn from the same data distribution. 
In practice, the other sources of non-determinism that we describe above can contribute to the results described in Figure~\ref{fig:loans}.
Moreover, it may not be immediately clear how each source contributes to the outcome predictions and impacts their associated probabilities. 
In other words, the distributional approach in Figure~\ref{fig:loans} clarifies that the predictions can fluctuate, but it conceals how stochasticity and other sources of non-determinism interact to produce those fluctuations --- a point to which we return in Section~\ref{sec:nondeterminism:law}, where we discuss the regulability of ML code.

For now, we observe that the legal literature discussed in Section~\ref{sec:nondeterminism:prior} touches on the stochasticity that contributes to examples like this one, but it does so in a manner different from the distributional picture we show here. 
Bambauer et al.~\cite{bambauer2022diff} discusses how stochasticity can cause \textit{a particular model} to exhibit SCMBDs that affect due process. 
Similarly, Creel and Hellman~\cite{creel2022leviathan} discuss how \textit{a particular model} exhibits deterministic outputs; their concern is that, at the scale of multiple decisions across multiple models for different tasks, there may be a pattern of arbitrary discrimination against certain individuals. 
In relation to Figure~\ref{fig:loans}, these works engage with stochasticity at the point in which there is one model producing a concrete credit score for each individual, rather than the distribution of possible model outputs for these individuals. 
It is only in the setting they rely on --- after we have selected a particular model to use for predicting credit scores --- that we can think about deterministic outputs. 
That is, by picking a particular model that encodes a specific function, we have locked in a deterministic score for each individual. 
We can then move from reasoning about distributions over possible outcomes of credit scores for individuals, as indicated in Figure~\ref{fig:loans}, to thinking about deterministic, concrete outcomes, which are conditional on the model we have chosen. 

Given one specific model, with deterministic outcomes for each individual input, it becomes possible to perform analyses concerning the inaccuracy of individual outcomes, associated harms, and metrics like error rates to capture summary information about a model's overall performance across a sample of inputs, as Bambauer et al.~\cite{bambauer2022diff} and Creel and Hellman~\cite{creel2022leviathan} both do. 
But, importantly, at the distributional level conveyed in Figure~\ref{fig:loans}, concepts like accuracy remain slippery. 
In reasoning about possible rather than specific model outcomes, this level of abstraction makes the potential areas of uncertainty in trained models --- whether due to stochasticity or other sources of non-determinism --- more transparent. 
It clarifies how the possibility of different outcomes can have the effect of muddling the distinction between correctness and incorrectness, and opens up the possibility of trying to untangle sources of non-determinism and their respective normative considerations regarding arbitrariness, which we discuss further in Section~\ref{sec:nondeterminism:law}.

\subsection{Patterns over Models}\label{sec:nondeterminism:example2}

\begin{figure}[t!]
  \begin{center}
    \includegraphics[width=0.45\textwidth]{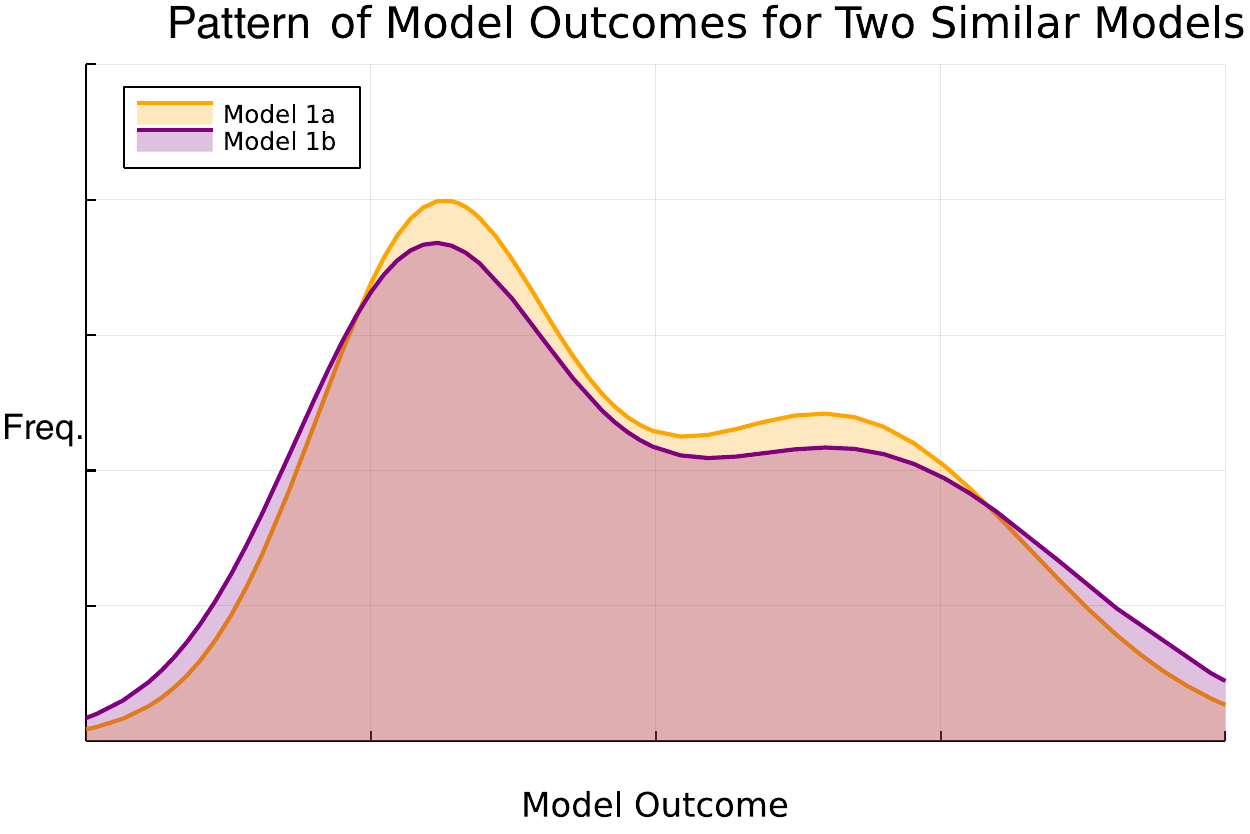}
    \caption{Synthetic patterns of model outcomes for two models trained on the same training data for the same task, using the same algorithm and data, but possibly different computers with different hardware random seeds. Non-determinism in the training process yields different patterns of model outcomes.}
	\label{fig:models}
  \end{center}
\end{figure}

Reasoning over distributions of outcomes does not just apply to thinking about how outcomes for fixed individual inputs may vary based on choice of model. 
This view can also help reason about how non-determinism affects models trained from the same stochastic training process. 
Figure~\ref{fig:models} shows patterns\footnote{In the camera-ready version of this paper, we used the word ``distributions'' to describe this effect as well, since we could not think of a better term to use at the time. 
    However, this was a poor choice on our part, since the type of non-determinism we describe in this section cannot be reasoned about with probability, and ``distribution'' most typically implies that we are talking about a ``probability distribution.''
    Joan Feigenbaum suggested we use the word ``pattern'' instead, so we make that change here. 
} over model outcomes for two models trained using the exact same procedure and, unlike the prior example, the models are trained using the same software random seed, which functions to supply the algorithm with the exact same training data. 
With this setup, we have controlled for every possible source of stochastic non-determinism in the training process. By using the same random seed, we should be able to consistently reproduce the same deterministic model, aligning with Creel and Hellman's conception of the training process (Section~\ref{sec:nondeterminism:prior}), and thus the two curves in Figure~\ref{fig:models} should completely overlap.\looseness=-1 

The reason they do not overlap is because of non-stochastic non-determinism that affects their respective training processes differently. 
For example, differences in hardware random seeds, which we cannot control in software code, cause the training process to produce different models that reflect different underlying deterministic functions. 
Ideally, even with non-determinism in ML software packages and across hardware, repeated runs of similar or identical training processes would result in outcome distributions that are reasonably similar to each other (as, one could argue, is the case in Figure~\ref{fig:models}, since the curves roughly overlap). 
If the models' patterns of outcomes do not vary too much, then at least we can be confident (however informally) that picking any of them as the specific model to deploy is a reasonable choice, as each model indicates performance roughly representative of all the models that were trained. 
In other words, it might be fine to avoid the issue of drawing a line between which models are correct and which are incorrect, because all of the models are effectively the same. 

Of course, though, the models are not \textit{exactly} the same, which may have significant consequences at the more granular level of individual outcomes. 
They may differ in a way that is not semantically meaningful, or may exhibit more uncertainty in connection with a protected attribute value, such as a particular gender or race. 
Importantly, this type of uncertainty, not being stochastic in nature, is not amenable to reasoning with the tools of probability; it is not amenable to the same reasoning about arbitrariness and individual outcomes in Creel and Hellman~\cite{creel2022leviathan}, which is implied to be predicated on uncertainty due to stochasticity.

More generally, the fact that models are not completely identical requires us to shift our thinking about ML. 
This pattern-level view clarifies that we should be thinking of one run of an ML training process as learning \emph{a} possible pattern over possible outcomes, rather than \emph{the} singularly correct pattern over possible outcomes. 
Additionally, unlike the synthetic example in Figure~\ref{fig:models}, in practice it is common for model outcome patterns to vary more significantly due to non-determinism~\cite{bouthillier2019reproducibility, raff2019reproducibility, sivaprasad2020hpo, cooper2021hpo, forde2021model, qian2021variance}. 
In such cases, it will not necessarily be clear if there is a representative model in the group --- if there is a model that is more ``correct'' than the others. 
Once again, due to non-determinism, drawing a firm boundary between correctness and incorrectness is ill-defined. 
As with the example in Figure~\ref{fig:loans}, this example similarly raises questions of how to legitimately pick a model that we can be confident will yield robust and reliable performance.\footnote{We could also extend the first example be plotted at the model level, rather than individual level, with outcomes on the $x$-axis and then the probability on the $y$-axis; 
    in this case, where we only look at stochasticity as a function of the training dataset, we would have a picture that looks perhaps a lot like Figure~\ref{fig:models}, but the terminology would change to reflect that we could reason about this plot as containing probability distributions.
}

This question is not just of theoretical relevance. 
In practice, non-determinism can cause resulting model outcome distributions to vary so much that, for a particular input, models can yield wildly inconsistent results. 
To make this more concrete, we describe an example in the ML literature that demonstrates the effects of such non-determinism. In recent work, Forde et al.~~\cite{forde2021model} and Qian et al.~\cite{qian2021variance} investigated how the impact of stochasticity\footnote{And we performed the largest such study on stochasticity in fairness contexts in other work that followed~\cite{cooper2024variance}.} 
and non-determinism on training models using similar training procedures can impact model fairness. Qian et al.~\cite{qian2021variance} published an extensive empirical study, in which they repeatedly trained models with identical training procedures, using the same software random seed and thus exactly the same training data examples across runs. 
In theory, this setup should control for stochasticity in different model outputs; by using the same training data and same training procedure, the models produced should be the same. 
However, the realities of running ML code in practice differ from what we expect in theory. Qian et al.~\cite{qian2021variance} makes the stakes of this point unimpeachably clear by comparing fluctuations in the resulting model outcome distributions. 
In particular, they computed common algorithmic fairness metrics to probe how fairness measurements varied for these (theoretically identical) models, and found that fairness measurements could vary by up to 12.6\%.
This degree of variance was so significant that, in some cases, it was possible for one trained model to pass US legal compliance rules regarding fairness thresholds on the test set, while another model could violate those same requirements~\cite[p. 2]{qian2021variance}.

In other words, Qian et al.~\cite{qian2021variance} illustrates clearly how non-stochastic non-determinism can have a significant impact on fairness in the distribution of possible modeling outcomes. 
This result indicates that picking any one specific model to deploy --- which then could be examined in terms of individual errors and error rates, fairness-related or otherwise --- is a non-trivial task. 
Non-determinism necessarily has an unpredictable role in the specific outcomes of training models, as evidenced by the resulting evaluation of test error to understand generalization. 
When this unpredictability leads to wide variability in metrics like fairness, this then raises fundamental questions not just about the fairness of particular models, but about the fairness of the process by which those models were trained. 
We may try to the best of our ability to control for models to be trained in the same way, and yet they may still exhibit vastly different fairness levels. 
Given this non-determinism, how can we be sure, especially when training just a few models under limited computational resource budgets, that the model we have selected to deploy in practice is representative of what is (at least close to) maximally possible in terms of fairness?

Questions like these, let alone their answers, are not clear from looking at individual outcomes or error rates for single models alone. 
Instead, it is looking at patterns and distributions over outcomes that raises questions about the legitimacy model-producing processes, through indicating how the resulting models from those processes can fluctuate in important ways. 
This distributional/pattern-level view provides information that can help us interrogate whether the process for training ML models for a specific task is robust enough to justify the use of \textit{any} such model produced from that process. 
By robust we mean that, even in the presence of non-determinism, the resulting variation in the behavior of possible ML models --- whether variation in model outcome distributions, or variation in outcomes across models for particular individual inputs --- is not the product of happenstance, for example resulting from a particular hardware-software interface implementation. 

The example of Qian et al.~\cite{qian2021variance} arguably does not meet this definition of robustness, given the large variance in fairness metrics across the distribution of models they produced.\footnote{Neither do the individual models trained in Cooper et al.~\cite{cooper2024variance}; 
    however, the ensemble models trained in that work are more robust in this sense.} 
This becomes especially clear when one considers how such variance in fairness could impact due process~\cite{lehr2017legalml}~\cite[pp. 1249-50]{brennanmarquez2019plausiblecause}~\cite[pp. 9,12]{calo2021modeling}~\cite[p. 1253]{citron2008dueprocess} --- 
if a particular chosen model by chance demonstrates poor performance with respect to fairness, in turn leading to a greater number of unfair individual outcomes in practice. 

%% file: section/20-arbitrary/23-nondeterminism/233-nondeterminism-law.tex
\section{Non-deterministic Code Is Lawless}\label{sec:nondeterminism:law}

In moving from looking at individual errors and model error rates to reasoning about distributions and patterns of outcomes, we have seen how the non-determinism inherent in ML can raise key questions concerning the legitimacy of using ML-driven processes in decision-making. 
We have seen, too, how non-determinism can directly effect harms at the individual level, in cases in which a training process is not sufficiently robust to guarantee that its resulting models behave similarly for key metrics, such as fairness. 
In short, our discussion thus-far has indicated that non-determinism can have significant, detrimental effects on the behavior of ML code. 
While there are different types of non-determinism, we have shown that prior work in legal ML focuses on non-determinism that is stochastic (Definition~\ref{def:stochasticity}).\footnote{However, upon further reflection, we realize that this work has not studied this sufficiently; we defer additional study to future work.} 
While this type of non-determinism is amenable to analysis using probability, other types of non-determinism in ML, such as the specifics of the hardware platform used to execute training processes, do not follow the same logic (Definition~\ref{def:nondeterminism}). 
As a result, work that has engaged with arbitrariness of ML decisions purely in stochastic terms has missed this crucial aspect of non-determinism and its relationship to arbitrariness. 

One of the important consequences of this omission has to do with an implied, uncomfortable relationship between arbitrariness and necessity in ML. 
As we briefly discussed in the introduction, the stochasticity of ML is one of its core strengths that separates it from non-stochastic decision systems; 
it is the property that makes it possible for ML to model phenomena that are too complex to specify exhaustively using if/then deterministic rules~\cite{murphy2022pml1}. 
Yet, stochasticity can also produce variable outcomes for the same inputs, and these variations can suggest contradictions that call the reliability of ML into question (Section~\ref{sec:nondeterminism:example1}). 
Moreover, these potential contradictions are less intuitive to grasp than the outputs of deterministic decision processes. At times, they might even seem like software bugs, rather than an artifact of a necessary feature of ML,\footnote{For work on the elusive boundary between bugs and inherent features in ML, please refer to Cooper et al.~\cite{cooper2022accountability}. 
    More generally, delineating what constitutes a bug for randomized programs is a philosophical question, which has long remained unresolved in the Programming Languages research community~\cite{kozen1981semantics, kozen1983pdl}.} 
which itself can further cast doubt on reliability. 
Due to this seeming double bind, it makes sense that legal literature about ML has tried to parse the cases in which the stochasticity-induced arbitrariness present concerns for the law.

However, other sources of non-determinism do not exhibit the same conflict. 
Lack of expressivity in hardware-software interfaces, inability to control hardware random seeds, and missing APIs for fine-grained control of run-time optimization of ML code all contribute to non-stochastic non-determinism; but, they are not necessary features of ML. 
They are not inherent to machine learning in theory; they are a reality of its practice. 
As a result, this source of non-determinism suggests potential sites for future reliable ML research. 
Nevertheless, in the interim, ML software and hardware ecosystems inject non-determinism into training processes, which affects the patterns of overall outcomes such that they deviate non-probabilistically from what is theoretically expected. 
What makes this especially challenging is that, as we demonstrated in our synthetic examples (Section~\ref{sec:nondeterminism:ml}), it is not always immediately clear which kind of non-determinism is responsible for impacts on the resulting distribution of outcomes, which further complicates our ability to reason about outcomes using the tools of probability.


More generally, taken together, both sources of non-determinism can make it very difficult to reason about the difference between correctness and incorrectness in ML program behaviors, thus making accuracy a fuzzy concept that is difficult to pin down.\footnote{It is also worth noting that the approximate computing concept of the trade-off between accuracy and efficiency~\cite{cooper2021eaamo, cooper2022fast}, and more generally using a temporal lens to analyze outcomes~\cite{susser2022time}, further complicates our understanding of accuracy in ML.} 
And yet, in the existing legal literature on ML, the issue of inaccuracy and accuracy, particularly at the individual model level, has been a dominant theme~\cite{douek2021moderation, douek2022formalism, brennanmarquez2019plausiblecause, citron2014scored, barocas2016data, calo2021modeling, lehr2017legalml}. 
For the law to adequately contend with non-determinism, we have argued that the legal literature must shift to also consider the viewpoint of distributions/patterns over outcomes, as this viewpoint indicates how non-determinism fundamentally problematizes our understanding of accuracy. 

Based on this prior discussion, we now argue that this will also require a shift in the dominant thread of cyberlaw thinking that echoes the refrain that ``code is law.''\footnote{This phrase, which originated from work in  Reidenberg~\cite{reidenberg1997code}, has been further developed and revised~\cite{lessig1999horse, lessig2003cycle, grimmelmann2005reg}, and then ultimately itself codified in Lessig~\cite{lessig2009code}. 
    It has since been partially adapted to account for the new kinds of experiences that ML (particularly robotics) will mediate~\cite{calo2015robotics, balkin2015robotics}.} 
In brief, ``code as law'' stands in for the idea that code does the work of law; code, like the law, is a modality for regulating and mediating human behavior~\cite{lessig1999horse, grimmelmann2005reg}. 
As Grimmelmann~\cite{grimmelmann2005reg} summarizes in more detail, ``code is law'' captures the idea that ``software itself can be effectively regulated by major social institutions, such as businesses or governments. ... If other institutions can regulate software, and software can regulate individual behavior, then software provides these institutions an effective way to shape the conduct of individuals''~\cite[p. 1721]{grimmelmann2005reg}.\footnote{Importantly, this understanding of ``code as law'' grew out of legal scholarship that was reckoning with the advent of the Internet. 
    In particular, this scholarship was concerned with ``decisions about the technical future of the Internet,'' which it considered to be ``important questions of social policy ... [that would] have the force of law even as they def[ied] many of our assumptions about law''~\cite[p. 1721]{grimmelmann2005reg}.}

In the extensive literature that has followed from Lessig~\cite{lessig2009code}'s codification of the concept, various scholars have built on and problematized different aspects of ``code is law''~\cite{grimmelmann2005reg, bamberger2010risk, calo2015robotics}, such that it has ultimately remained a resonant and powerful frame for thinking about technology. 
However, the work that contends with this concept tends to (often implicitly) assume a deterministic view of code. It considers code to be a set of automated if/then rules that ensure consistent decisions --- and can be institutionally regulated to ensure consistent decisions --- as it works to enable and constrain human behavior~\cite[pp. 1721, 1728-1732]{grimmelmann2005reg}~\cite[p. 676]{bamberger2010risk}~\cite[p. 1253]{citron2008dueprocess}. 
In this view, code can concretely specify rule-like (rather than standard-like)\footnote{It is perhaps interesting to consider --- though out of scope in this short paper --- how non-deterministic ML code may more closely resemble standards than rules.} 
relationships between inputs and outputs that are ``free from ambiguity''~\cite[p. 1723]{grimmelmann2005reg}. 
Put simply, this conception of code maps nicely to if/then rules that resemble those in the law. 
Yet, as we have seen throughout this paper, the assumption of deterministic code does not hold for ML: 
due to its statistical nature, ML code does not operate by deterministic if/then rules. Instead, due to non-determinism, it is as if both the ``if'' and the ``then'' are fuzzy; they are not specifiable in concrete terms. 
It is therefore natural to ask: what does non-deterministic code do to an idea of ``code as law'' that is predicated on determinism?

We attempt an answer in a (sort-of) proof by contradiction. 
We begin by assuming that ``code as law'' still holds for the non-deterministic code of ML. 
From there, then, we would need to consider what it would mean for the law to similarly exhibit non-determinism. 
And this is where ``code is law'' immediately starts to break down. In the ideal case, the law should have deterministic outcomes based on its inputs. 
It can exhibit variation in the relationships between inputs and outputs, but it should not be the case that there is randomness or arbitrariness in those relationships~\cite[pp. 665-666]{kolber2014smoothbumpy}. 
In practice, non-determinism can of course occur in the law. Judicial discretion is not mechanical; given similar inputs, outputs can vary across judges (or even within the same judge)~\cite[p. 78]{tamanaha2004law}. 
But in spite of this non-determinism, sometimes described in relation to the ``indeterminacy thesis,'' the law remains largely predictable.\footnote{As Tamanaha~\cite{tamanaha2004law} discusses, even if there is a relatively small number of unpredictable cases, these cases are often high-impact. 
    General predictability in terms of case numbers should not be misconstrued as a claim that unpredictable cases have low impact. 
    Indeterminacy and unpredictability are more frequent within the Supreme Court, and there always remains the possibility that judges could exploit ``latent indeterminacy'' to suit personal objectives~\cite[pp. 90, 122-3]{tamanaha2004law}.
}  
Contradictions in legal rules, which interfere with predictability, are classically conceived of as ``miscarriages'' of the law~\cite[pp. 38-39]{fuller1965law}. 
Further, as Tamanaha~\cite{tamanaha2004law} argues, there are generally speaking few contradictions in the law, and ambiguities can be handled through ``reasoned analysis.''\footnote{For the indeterminacy thesis ``To have bite it must be shown that existing legal rules form a pervasive mess of contradictions, which critical theorists have not demonstrated''~\cite[p. 88]{tamanaha2004law}.} 

In contrast, non-determinism --- particularly non-stochastic non-determinism --- does not share these qualities. 
Stochasticity perhaps can be considered predictable, its effects reasoned about ``at a higher level of abstraction''~\cite[pp. 3-4]{creel2022leviathan} using probability theory. 
However, from the view of patterns over possible ML outcomes, other forms of non-determinism, ironically, inject unpredictability into ML predictions, both in an intuitive sense and more formally in its resistance to statistical analysis. 
Additionally, as we have seen, empirical work in ML commonly demonstrates that it can result in contradictions with significant consequences~\cite{qian2021variance, forde2021model, cooper2021hpo}. 

Moreover, unlike in ML, the legal system embodies answerability. There are actors in the system who must step forward and answer for their decisions; they must provide explanations and are subject to cross-examination~\cite{tribe1971math, brennanmarquez2019plausiblecause}. 
Answerability in part functions to remove randomness and arbitrariness from the law. 
In the long run, the system undergoes an ongoing process of legitimization. 
In other words, the law has mechanisms for recourse, which effectively can serve (however imperfectly) to root out non-determinism; unlike ML, law treats non-determinism vis-à-vis unpredictability and contradictions like a bug, not a feature. 

This indicates a fundamental incompatibility for understanding ML code as law. 
Whereas the law can do work to avoid non-determinism, ML inherently relies on stochasticity and \emph{de facto} relies on non-stochastic non-determinism in state-of-the-art implementations.\footnote{As briefly mentioned earlier, this is a practical reality aimed at optimizing for efficiency under conditions of limited computing resources.} 
The resulting unpredictability of ML code distinguishes it from law in that it causes ML code to evade regulation. 
To borrow a phrase from Jack Balkin, such ``code is lawless''~\cite[p. 52]{balkin2015robotics}; 
the unpredictability that results from non-determinism presents key problems for thinking of code as being constitutive of law.\footnote{Balkin developed this spin on the original refrain in relation to the problem of emergence and unpredictable, unintended consequences in robotic systems. 
    We adopt it more broadly for non-deterministic code.
} 

%% file: section/20-arbitrary/23-nondeterminism/234-nondeterminism-conclusion.tex
\section{Conclusion}\label{sec:nondeterminism:conclusion}

Non-deterministic code may itself be lawless, but this does not mean we should entirely avoid its use\footnote{It does, however, seem reasonable to draw the line that ML, if lawless, should not itself be used to design law 
    (e.g., ``Micro-directives [that] will provide \emph{ex ante} behavioral prescriptions finely tailored to every possible scenario''~\cite[p. 2]{casey2015rulesandstandards}). 
    ML can nevertheless still be useful in the service of law, for example, by aiding in the design of tools that help lawyers be more efficient and effective in their work~\cite{delgado2022uncommontask}.
} 
and that we can do nothing to better regulate its deployment in practice. 
On the ML side, we can strive to develop tools that obtain some measure of consistency --- e.g., similar model outcome distributions across training runs --- even in the presence of non-determinism (stochastic or otherwise). 
The current push for more robust ML is in fact working to develop algorithms that leverage non-determinism to learn complex decision surfaces, but also provably have bounded effects on, for example, variance in model-training outcomes. 
In short, ML can do work to tighten distributions/patterns, to provide theoretical limits on error (that then have to be met in practice), and to characterize rigorous trade-offs between computational resource usage for training models and how robust resulting models can be. 
These are rich areas of research in ML, all of which become better-appreciated when understanding ML from a distributional/pattern-level perspective. 

While ML can work improve robustness, stochastic non-determinism will always remain feature, not a bug. 
Legal scholarship thus needs to attend to the role of distributions over outcomes in order to fully appreciate how stochasticity contributes to uncertainty in the behavior of ML systems. A
s we have seen through brief examples concerning unfairness, uncertainty and non-determinism, not just individual outcomes, can themselves implicate harms. 
Since the law will necessarily focus on harms, its work will be to close the gap between these two essential ways of viewing ML --- to ensure that the law is able to reason about distributional aspects in such a way that these aspects serve to clarify how they relate to individual outcomes. 
The law must find ways to bring the distributional and the individual together, such that it can successfully bring ML to account for the harms it causes. 

%% file: section/30-algorithms/300-algos.tex
\part{Taming Randomness in Scalable, Reliable Sampling and Optimization Algorithms}\label{part:algorithms}

The promise of ML capabilities will only be realized at scale. 
This was true before generative AI, and is crystal clear today with respect to generative-AI systems (Part~\ref{part:genai}). 
However, scale also makes measurement challenging:  
to be feasible on large-scale data, in ML we often make approximations or use heuristics that sacrifice reliability; 
in turn, this can sacrifice correctness in outputs. 
In this part, we explore such tensions in uncertainty estimation and distributed optimization algorithms, and the associated implications for law and policy. 
This chapter reflects work that has been published at \emph{NeurIPS} (Spotlight and poster), \emph{AISTATS} (poster), and \emph{ACM EAAMO} (Oral), and the \emph{Colorado Technology Law Journal}. 

\begin{figure}[h!]
  \centering
    \centering
    \includegraphics[width=.6\textwidth]{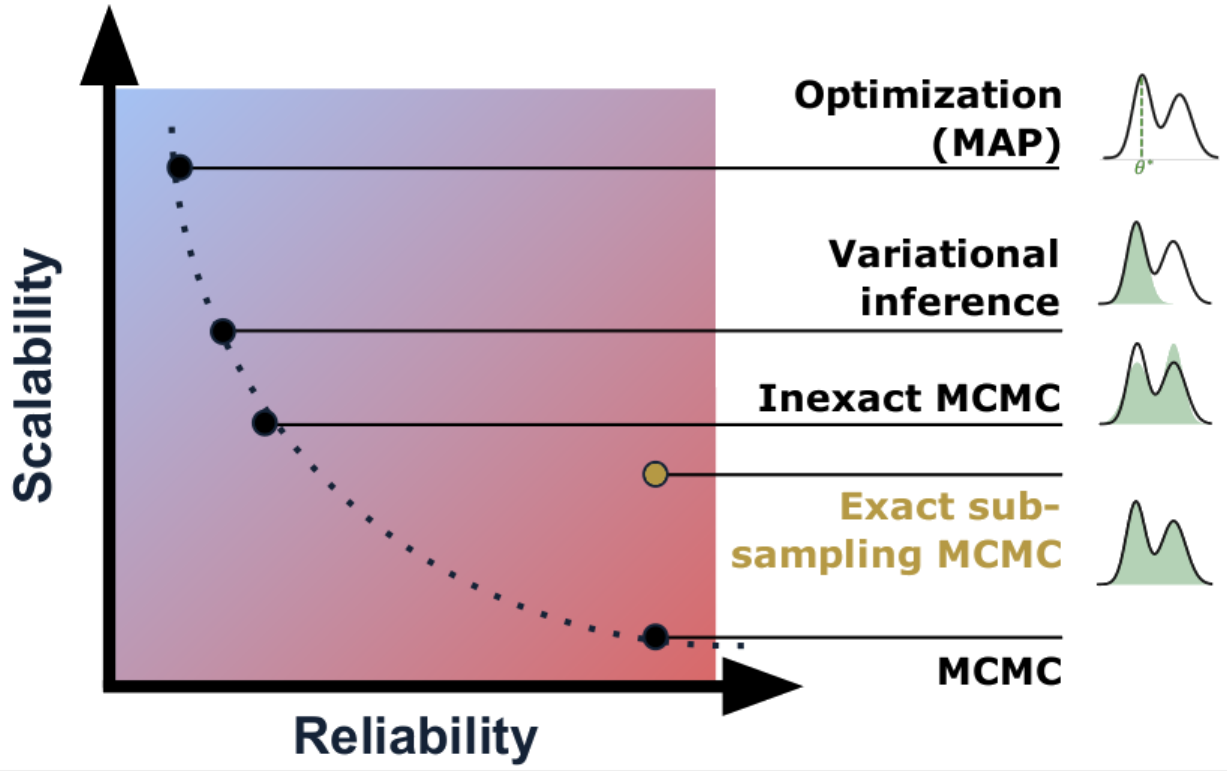}
    \vspace{-.1cm}
    \caption{Reliability-scalability trade-off in Bayesian inference (i.e., for capturing the posterior of possible models, right). Our work (\textbf{\textcolor{mustard}{yellow}}) carefully uses subsampling to push the frontier.\looseness=-1}
    \label{fig:bi}
\end{figure}

First, we discuss work that examines questions related to Markov chain Monte Carlo (MCMC).
MCMC is the tool of choice for reliable uncertainty estimation in Bayesian inference. 
Knowledge of uncertainty can help us produce more reliable models, and thus more reliable measurements of metrics. 
However, MCMC is also really computationally expensive, which has made it infeasible for problems of modern scale.  
Our work makes MCMC more efficient and scalable, while retaining reliability guarantees; it better navigates trade-offs between these competing goals (Figure~\textbf{\ref{fig:bi}}). 
MCMC simulates the posterior distribution over possible-model solutions to the learning problem (in contrast to optimization's single-point estimate), which can be used to compute model-parameter uncertainties.  
Computing the posterior is traditionally really costly: 
it involves a sampling procedure that computes sums over the entire dataset at each iteration. 
This is part of the reason why optimization is the workhorse of modern ML. 
Sampling, particularly MCMC, is intractable for this setting: it typically trades-off scalability for reliability, while optimization trades-off reliability for scalability. 
To break out of this traditional trade-off, our work carefully introduces data subsampling to MCMC. 
This makes it possible to efficiently produce high-quality estimates of the posterior --- to achieve \emph{both} scalability and reliability (Figure~\textbf{\ref{fig:bi}}). 
Following this approach, we have developed algorithms that push the frontier of the scalability-reliability trade-off in Bayesian, in turn making reliable uncertainty estimation feasible at previously unprecedented scales (Chapter~\ref{chapter:tunamh}, Zhang et al.~\citep{zhang2020amagold}, Cooper et al.~\citep{zhang2020tunamh}). 

Second, we discuss work on improved example orders for SGD-based distributed optimization (Chapter~\ref{chapter:cdgrab}, Cooper et al.~\citep{cooper2023cdgrab}). 
Such orderings represent a way that we can achieve better efficiency overall \emph{without} sacrificing reliability. 
We can in fact get (in theory) guarantees of faster convergence by moving to ordering schemes that do not rely on random reshuffling every epoch. 
This involves a slight increase in per-iteration cost (to compute permutations), but comes with the benefit of each iteration (eventually) having an improved impact on convergence convergence. 
However, in practice, we need to run this ordering algorithm for a substantial number of epochs in order to see these efficiency benefits play out.
Sometimes, it is not necessary to run training for this long; in turn, if we do run our method for longer, we tend to see improved generalization. 

And third, we discuss how scalability, reliablity, and trade-offs between the two exist all over ML. For example, generative-AI systems are very resource intensive: there are difficult scalability and efficiency challenges that we need to contend with in order to produce high-quality models. 
We argue that such trade-offs are an important abstraction for policymakers to understand the relationship between choices ML experts make and resulting ML functionality. 
Because such trade-offs have useful analogues in other domains that policymakers already engage with (e.g., car safety), they are legally cognizable and a natural mechanism for communication with ML experts. 
They can help shine a light on the barriers to accountability in stochastic algorithms (Appendix~\ref{chapter:accountability}), and help us reason rigorously about how to weaken them. 

\input{section/30-algorithms/31-tunamh/31-tunamh-main}
\input{section/30-algorithms/32-cdgrab/32-cdgrab-main}

\input{section/30-algorithms/33-tradeoffs/33-tradeoffs-main}

%% file: section/30-algorithms/31-tunamh/31-tunamh-main.tex
\chapter{Asymptotically Optimal Exact Minibatch Metropolis-Hastings}\label{chapter:tunamh}

We begin this part with contributions to scalable, exact sampling algorithms. 
The work covered here~\citep{zhang2020tunamh} represents the authors' second joint collaboration on the topic of scaling reliable uncertainty estimation to new heights~\citep{zhang2020amagold}. 
Several important typos are corrected from the original camera-ready paper.\footnote{In follow-on work, we also learned that obtaining the asymptotically optimal guarantees of TunaMH in practice is actually quite challenging.
    This is because the method changes the minibatch size every sampling iteration.
    With a fixed minibatch size, it is possible to pre-allocate matrix memory and just update the entries each iteration, which reduces memory overhead significantly.
    This is not easily attainable with a changing batch size every iteration (we spent nearly 6 months trying to do this). 
    In practice, full-batch MH methods (especailly stochastic gradient proposal-based methods, e.g., Zhang et al.~\cite{zhang2020amagold}), are in practice much faster in actual software. 
}\\

\noindent \textbf{Chapter summary}: 
Metropolis-Hastings (MH) is a commonly-used MCMC algorithm, but it can be intractable on large datasets due to requiring computations over the whole dataset. In this paper, we study \emph{minibatch MH} methods, which instead use subsamples to enable scaling. We observe that most existing minibatch MH methods are inexact (i.e. they may change the target distribution), and show that this inexactness can cause arbitrarily large errors in inference. We propose a new exact minibatch MH method, \emph{\methodname}, which exposes a tunable trade-off between its batch size and its theoretically guaranteed convergence rate. We prove a lower bound on the batch size that any minibatch MH method \emph{must} use to retain exactness while guaranteeing fast convergence --- the first such bound for minibatch MH --- and show \methodname{} is asymptotically optimal in terms of the batch size. Empirically, we show \methodname{} outperforms other exact minibatch MH methods on robust linear regression, truncated Gaussian mixtures, and logistic regression.\\

\noindent This chapter is a licensed derivative copy of work published and awarded a spotlight at \emph{NeurIPS 2020}~\cite{zhang2020tunamh}. 

\input{section/30-algorithms/31-tunamh/311-tunamh-intro}
\input{section/30-algorithms/31-tunamh/312-tunamh-prelim}
\input{section/30-algorithms/31-tunamh/313-tunamh-tuna}
\input{section/30-algorithms/31-tunamh/314-tunamh-optimality}

\input{section/30-algorithms/31-tunamh/315-tunamh-experiments}
\input{section/30-algorithms/31-tunamh/316-tunamh-conclusion}

%% file: section/30-algorithms/31-tunamh/311-tunamh-intro.tex
\section{Introduction}

Bayesian inference is widely used for probabilistic modeling of data. Specifically, given a dataset $\mathcal{D} = \{x_i\}_{i=1}^N$ and a $\theta$-parameterized model, it aims to compute the posterior distribution 
\[
\pi(\theta) \propto \exp\left(-\sum_{i=1}^N U_i(\theta)\right), \text{where } U_i(\theta) = - \log p(x_i|\theta) - \frac{1}{N} \log p(\theta).
\]
Here $p(\theta)$ is the prior and the $p(x_i|\theta)$ give the likelihood of observing $x_i$ given the parameter $\theta$. We assume the data are conditionally independent given $\theta$. The $U_i$ have a natural interpretation as component \emph{energy functions} with $\pi$ acting as a Gibbs measure. In practice, computing $\pi(\theta)$ is often intractable and thus requires using approximate methods, such as Markov chain Monte Carlo (MCMC). MCMC uses sampling to estimate the posterior and is guaranteed to converge asymptotically to the true distribution, $\pi$ \cite{brooks2011handbook}.  

The Metropolis-Hastings (MH) algorithm \citep{hastings1970mh,metropolis1953equation} is one of the most commonly used MCMC methods. In each step, MH generates a proposal $\theta'$ from a distribution $q(\cdot|\theta)$, and accepts it with probability 
\begin{equation}
\label{eqnMHaccprob}
\textstyle
a(\theta,\theta') = \min\left(1, \frac{\pi(\theta')q(\theta|\theta')}{\pi(\theta)q(\theta'|\theta)}\right) = \min\left(1, \exp\big( \sum_{i=1}^N (U_i(\theta) - U_i(\theta')) \big) \cdot \frac{q(\theta|\theta')}{q(\theta'|\theta)}\right).
\end{equation}
If accepted, the chain transitions to $\theta'$; otherwise, it remains at the current state $\theta$. This accept/reject step can be quite costly when $N$ is large, since it entails computing a sum over the entire dataset. 

Prior work has proposed many approaches to mitigate the cost of this decision step \citep{bardenet2017mcmc}. One popular approach involves introducing stochasticity: instead of computing over the entire dataset, a subsample, or \emph{minibatch}, is used to compute an approximation. These minibatch MH methods can be divided into two classes, \emph{exact} and \emph{inexact}, depending on whether or not the target distribution $\pi$ is necessarily preserved. 
Inexact methods introduce asymptotic bias to the target distribution, trading off correctness for speedups~\cite{bardenet2014towards,korattikara2014austerity,seita2016efficient,quiroz2019speeding,quiroz2016block}. Exact methods either require impractically strong constraints on the target distribution \cite{maclaurin2015firefly, zhang2019poisson}, limiting their applicability in practice, or they negatively impact efficiency, counteracting the speedups that minibatching aims to provide in the first place \cite{banterle2015accelerating,cornish2019scalable}. Moreover, all existing exact methods operate on the belief that there is a trade-off between batch size and convergence rate --- between scalability and efficiency. Yet no prior work formally exposes this trade-off, and most prior work gives no convergence rate guarantees. Given these  various considerations, it is not entirely clear how to evaluate which minibatch MH method to use.

In this paper we forge a path ahead to untangle this question. While inexact methods have been prominent recently due to their efficiency, they are not reliable: we show that the stationary distribution of any inexact method can be arbitrarily far from the target $\pi$. This means they can yield disastrously wrong inference results in practice, and it is difficult to tell just how bad those results can be. 

We therefore turn our attention to exact methods and introduce \emph{\methodname}.\footnote{\methodname{} since it \emph{tunes} the efficiency-scalability trade-off and uses a Poisson (French for ``fish") variable.} Compared to prior work, we make milder assumptions, which enables \methodname{} to apply to a wider variety of inference tasks. More specifically, we require local rather than global bounds on the target distribution~\cite{maclaurin2015firefly,zhang2019poisson} and do not rely on the Bernstein-von Mises approximation~\cite{cornish2019scalable,bardenet2017mcmc,bierkens2019zig}. \methodname{} is guaranteed to retain sample efficiency in the presence of minibatching: its convergence rate (measured by the spectral gap) is within a constant factor of standard, non-minibatch MH. More importantly, \methodname{} also enables us to rigorously characterize the trade-off between scalability and efficiency. It has a hyperparameter $\chi$, which enables tuning the trade-off between expected batch size and convergence rate.

By exposing this trade-off, our analysis raises the natural question: \emph{is \methodname{} optimal for this trade-off?} That is, could another exact algorithm use an asymptotically smaller average batch size while having the same convergence rate guarantees? We explore this in Section \ref{sec:tunamh:optimality}; under the same mild assumptions we use to derive \methodname{}, we prove a lower bound on the expected batch size for \emph{any} exact minibatch MH method that can keep a reasonable convergence rate. To our knowledge, we are the first to prove a lower bound of this nature for minibatch MH. Moreover, \methodname{} is \emph{asymptotically optimal} in balancing the expected batch size and convergence rate. It remains exact and efficient while on average using the smallest possible number of samples. In summary:
\begin{itemize}[noitemsep,topsep=0pt, leftmargin=.4cm]
    \item We demonstrate that any inexact minibatch MH method can be arbitrarily inaccurate (Section \ref{sec:inexactproblems}).
    \item We introduce a new exact method, \methodname{} (Section \ref{sec:tunamh:tuna}), with a lower bound on its convergence rate (in terms of the spectral gap) and a tunable hyperparameter to balance the trade-off between convergence rate and batch size.
    \item We prove a lower bound on the batch size for any exact minibatch MH method given a target convergence rate --- the first such lower bound in this area. This result indicates that the expected batch size of \methodname{} is asymptotically optimal in terms of the problem parameters (Section \ref{sec:tunamh:optimality}).
    \item We show empirically that \methodname{} outperforms state-of-the-art exact minibatch MH methods on robust linear regression, truncated Gaussian mixture, and logistic regression (Section \ref{sec:tunamh:exp}).
\end{itemize}

%% file: section/30-algorithms/31-tunamh/312-tunamh-prelim.tex
\section{Preliminaries and Drawbacks of Prior Minibatch MH Methods}\label{sec:tunamh:drawbacks}

We first formally define the class of methods that we study theoretically in this paper: minibatch MH methods of the form of Algorithm~\ref{alg:subsampledMH}. This class contains methods that sample a proposal from distribution $q$ (which we always assume results in the chain being ergodic), and choose to accept or reject it by calling some randomized subroutine, $\texttt{SubsMH}$, which outputs $1$ or $0$ for ``accept'' or ``reject,'' respectively. 

Algorithms in this class have several notable properties. First, $\texttt{SubsMH}$ is \emph{stateless}: each acceptance decision is made independently, without carrying over local state associated with the MH procedure between steps. Many prior methods are stateless~\cite{korattikara2014austerity,bardenet2014towards,seita2016efficient,cornish2019scalable}. We do not consider \emph{stateful} methods, in which the decision depends on previous state; they are difficult to analyze due to running on an extended state space~\cite{andrieu2009pm, quiroz2019speeding}.
Second, $\texttt{SubsMH}$ takes a function that computes energy \emph{differences} $U_i(\theta) - U_i(\theta')$ and outputs an acceptance decision. We evaluate efficiency in terms of how many times $\texttt{SubsMH}$ calls this function, which we term the \emph{batch size} the method uses.
Third, $\texttt{SubsMH}$ takes parameters that bound the maximum magnitude of the energy differences. Specifically, as in Cornish et al.~\cite{cornish2019scalable}, we assume:
\begin{assumption}\label{assump}
For some constants $c_1, \ldots, c_N \in \R_+$, with $\sum_i c_i = C$, and symmetric function $M: \Theta \times \Theta \rightarrow \R_+$, for any $\theta, \theta' \in \Theta$, the energy difference is bounded by
$|U_i(\theta) - U_i(\theta')|\le c_i M(\theta,\theta')$.
\end{assumption}

One can derive such a bound, which can be computed in $O(1)$ time, for many common inference problems: for example, if each energy function $U_i$ is $L_i$-Lipschitz continuous, then it suffices to set $c_i = L_i$ and $M(\theta, \theta') = \|\theta - \theta'\|$ (See Appendix~\ref{app:tunamh:experiments} for examples of $c_i$ and $M$ on common problems). Note that the $\texttt{SubsMH}$ method may choose \emph{not} to use these bounds in its decision. We allow this so the form of Algorithm \ref{alg:subsampledMH} can include methods that do not require such bounds.
Most existing methods can be described in this form \cite{korattikara2014austerity,bardenet2014towards,seita2016efficient,cornish2019scalable,banterle2015accelerating}. For example, standard MH can be written by setting $\texttt{SubsMH}$ to a subroutine that computes the acceptance rate $a$ as in (\ref{eqnMHaccprob}) and outputs $1$ (i.e., accept) with probability $a$. 

Such minibatch MH methods broadly come in two flavors: \emph{inexact} and \emph{exact}. We next establish the importance of being exact and demonstrate how \methodname{} resolves drawbacks in prior work.

\begin{algorithm}[t]
\caption{Stateless, Energy-Difference-Based Minibatch Metropolis-Hastings}
\label{alg:subsampledMH}
\begin{algorithmic}
\State \textbf{given: } state space $\Theta$, energy functions $U_1, \ldots, U_N: \Theta \rightarrow \R$, proposal dist. $q$, initial state $\theta \in \Theta$ 
\State \textbf{given: } parameters $c_1,\ldots,c_N$, $C$, $M$ from Assumption~\ref{assump}, randomized algorithm \texttt{SubsMH}
\Loop
    \State \textbf{sample} $\theta' \sim q(\cdot|\theta)$
    \State \textbf{define function} $\Delta U: \{1, \ldots, N\} \rightarrow \R$, such that
    $\Delta U(i) = U_i(\theta) - U_i(\theta')$
    \State\textbf{call subroutine} $o \leftarrow \texttt{SubsMH}(\Delta U, N, q(\theta|\theta') / q(\theta'|\theta), c_1,\ldots,c_N, C, M(\theta,\theta'))$
    \State \textbf{if} $o = 1$, \textbf{update} $\theta  \leftarrow \theta'$
\EndLoop
\end{algorithmic}
\end{algorithm}

\subsection{The Importance of Being Exact}\label{sec:inexactproblems}

Inexact methods are popular due to helping scale MH to new heights~\cite{bardenet2014towards,korattikara2014austerity,seita2016efficient,quiroz2019speeding}. They approximate the MH acceptance ratio to within an error tolerance ($> 0$), trading off exactness for efficiency gains. Surprisingly, the bias from inexactness can be arbitrarily large even when the error tolerance is small.

\begin{theorem}\label{statement:counterexample}
Consider any minibatch MH method of the form in Algorithm~\ref{alg:subsampledMH} that is inexact (i.e. does not necessarily have $\pi$ as its stationary distribution for all $\pi$ satisfying Assump.~\ref{assump}). For any constants $\delta\in (0,1)$ and $\rho>0$, there exists a target distribution $\pi$ and proposal distribution $q$ such that if we let $\tilde{\pi}$ denote a stationary distribution of the inexact minibatch MH method on this target, it satisfies
\[
\operatorname{TV}(\pi,\tilde{\pi})\ge\delta \text{ and } \operatorname{KL}(\pi,\tilde{\pi})\ge \rho.
\]
where TV is the total variation distance and \text{KL} is the Kullback–Leibler divergence.
\end{theorem}

Theorem \ref{statement:counterexample} shows that when using any inexact method, there always exists a target distribution $\pi$ (factored in terms of energy functions $U_i$) and proposal distribution $q$ such that it will approximate $\pi$ arbitrarily poorly. This can happen even when individual errors are small; they can still accumulate a very large overall error. We prove Theorem~\ref{statement:counterexample} via a simple example --- a random walk along a line, in which the inexact method causes the chain to step towards one direction more often than the other, even though its steps should be balanced (Appendix~\ref{app:proof:counterexample}). Note that it may be possible to avoid a large error by using some specific proposal distribution, but such a proposal is hard to know in general. 

We use AustereMH~\cite{korattikara2014austerity} and MHminibatch~\cite{seita2016efficient} to empirically validate Theorem~\ref{statement:counterexample}. For these inexact methods, we plot density estimates with the number of states $K=200$ in Figure \ref{fig:counter-example}a (see Appendix \ref{app:experiments:counterexample} for using other $K$); the stationary distribution diverges from the target distribution significantly. Moreover, the TV distance between the density estimate and the true density increases as $K$ increases on this random walk example (Figure \ref{fig:counter-example}b). By contrast, our exact method (Section \ref{sec:tunamh:tuna}) keeps a small TV distance on all $K$ and estimates the density accurately with an even smaller average batch size. We also tested AustereMH on robust linear regression, a common task, to show that the error of inexact methods can be large on standard problems (Appendix~\ref{app:experiments:counterexample}).

\begin{figure*}[t!]
    \centering
    \begin{tabular}{cccc}		
    \includegraphics[width=4.8cm]{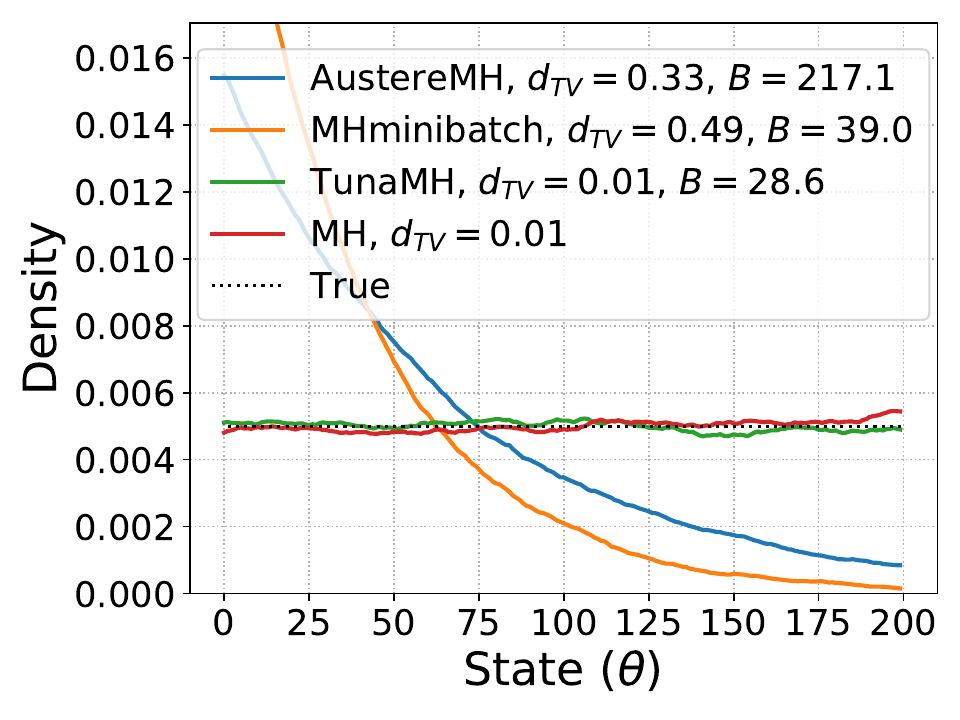} &
    	\includegraphics[width=4.8cm]{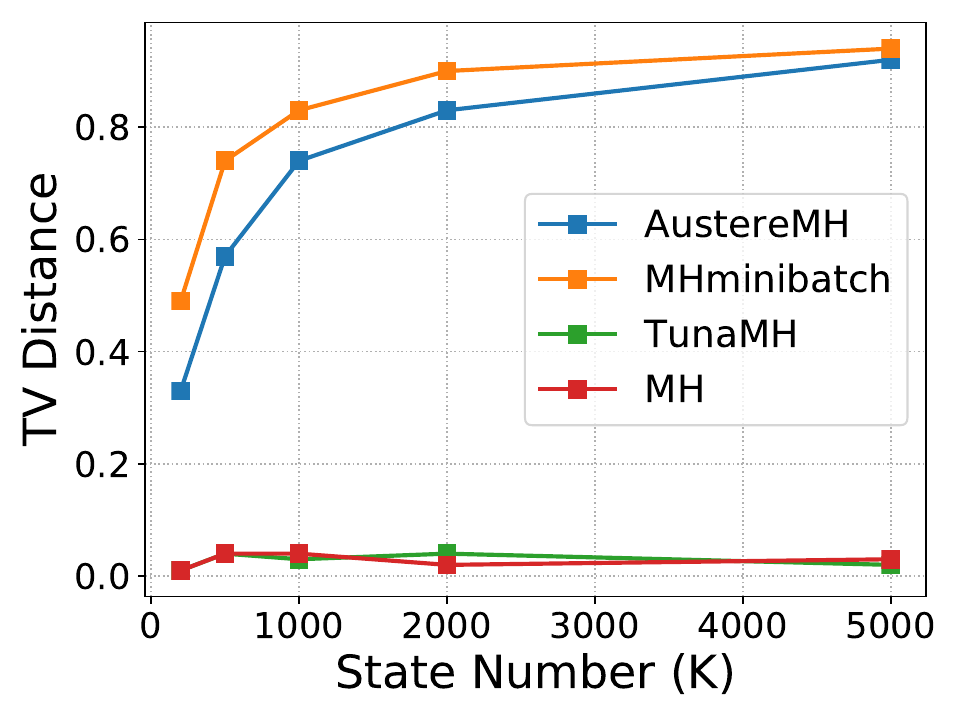}  &
    	\includegraphics[width=4.8cm]{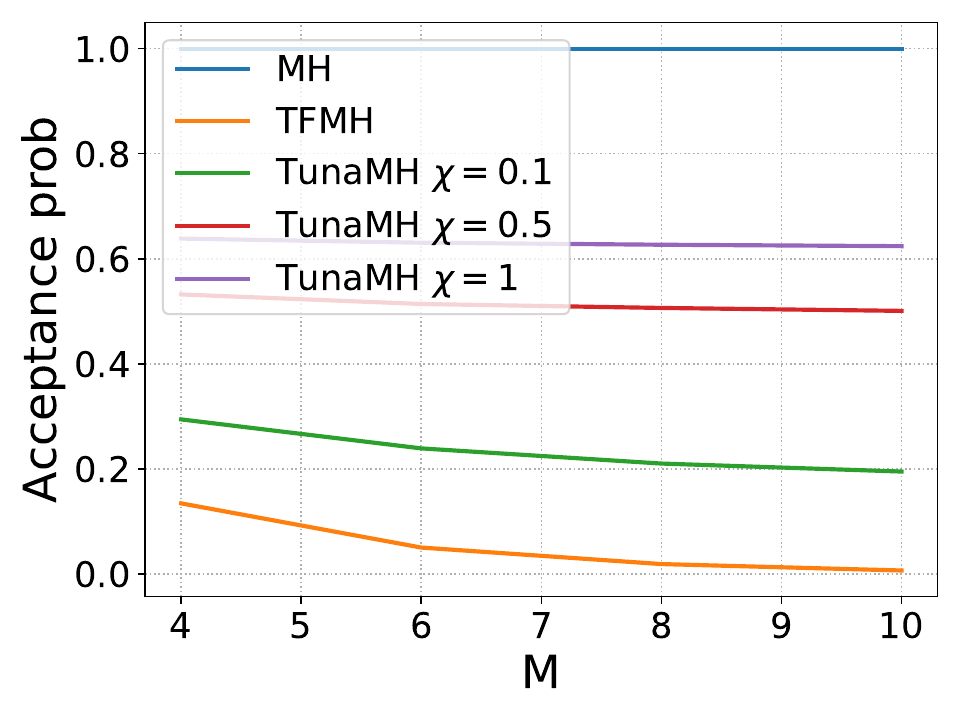} &
    	\\		
    	(a) &
    	(b) &
    	(c) 
    	\hspace{-0mm}\\		
    \end{tabular}
    \caption{Existing MH method issues. (a)-(b) Inexact methods can diverge a lot from true distribution. ``$d_{TV}$'' and ``$B$'' denote the TV distance and the batch size respectively. (c) SMH has low and \methodname{} with different values of hyperparameter $\chi$ has high acceptance rates.}
    \label{fig:counter-example}
\end{figure*}

\subsection{Issues with Existing Exact Methods} \label{sec:issue-of-exact}
This observation suggests that we should be using exact methods when doing minibatch MH. However, existing approaches present additional drawbacks, which we discuss below.

\noindent \textbf{Factorized MH and Scalable MH} 
are stateless, exact minibatch methods. Factorized MH (FMH) decomposes the acceptance rate into a product of factors, which allows for rejecting a proposal based on a minibatch of data  \citep{ceperley1995path,christen2005markov,banterle2015accelerating}. Truncated FMH (TFMH) is a FMH variant that maintains geometric ergodicity; it falls back on standard MH in a step when the bound on the factors reaches a certain threshold \citep{cornish2019scalable}. No matter how this threshold is set, we can construct tasks where TFMH is either arbitrarily inefficient (rejecting arbitrarily often, slowing convergence), or degrades entirely to standard MH.
\begin{statement}\label{statement:fmh}
For any constant $p\in (0,1)$, there exists a target distribution such that TFMH either has an acceptance rate which is less than p times that of standard MH, or it completely degrades to standard MH (summing over the whole dataset at each step).
\end{statement}
We prove this statement in Appendix \ref{app:proof:smh} using an example of a uniform distribution along a line, where we let $x_i$ take one of two values, $\{-M/N, M/N\}$ with $M>0$. We show that the acceptance rate of TFMH can be arbitrarily low by increasing $M$, which we also empirically verify in Figure \ref{fig:counter-example}c.

To improve the acceptance rate of TFMH, Scalable MH (SMH) introduces control variates, which approximate $U_i$ with a Taylor series around the mode \citep{cornish2019scalable}. However, it only works with unimodal posteriors and high-quality Bernstein-von Mises approximations --- conditions that do not hold for many common inference tasks.

\paragraph{PoissonMH} is a stateless minibatch MH method adapted from an algorithm designed for scaling Gibbs sampling on factor graphs \cite{zhang2019poisson}. However, unlike our method, it requires strong assumptions --- specifically, a global upper bound on the energy. Such an upper bound usually does not exist and, even if it does, can be very large, resulting in an impractically large batch size.

\paragraph{FlyMC} is a stateful method, which means it uses auxiliary random variables to persist state across different MH steps \cite{maclaurin2015firefly}. It requires a lower bound on the likelihood function, which is typically more demanding than Assumption~\ref{assump} and does not have theoretical performance guarantees.

\paragraph{Other exact methods} exist based on Piecewise Deterministic Markov Processes \cite{bouchard2018bouncy,bierkens2019zig}. They require regularity conditions only available for some problems, so their practical utility is limited.

%% file: section/30-algorithms/31-tunamh/313-tunamh-tuna.tex
\section{\methodname: Asymptotically Optimal Exact MH} \label{sec:tunamh:tuna}

In this section, we present our method, \methodname{}, which evades the issues of prior exact methods discussed in Section \ref{sec:issue-of-exact}.
Like SMH \cite{cornish2019scalable}, our method works on distributions for which an \emph{a priori} bound on the energy differences is known (Assumption \ref{assump}). 

Our algorithm, presented in Algorithm~\ref{alg:poisson-mh},\footnote{There is a typo in the camera-ready paper that persists in the Appendix (fixed here), where the proposal ratio numerator and denominator are flipped in the accept/reject step's computation of the MH ratio.} takes as parameters $c_1, \ldots, c_N$, $C$, and $M$ from Assumption \ref{assump}, along with an additional hyperparameter, $\chi>0$.
It proceeds in four steps.
First, like any MH method, it generates a proposal $\theta'$ from given distribution $q$.
Second, it samples a batch size $B$ from a Poisson distribution. This makes the expected number of energy functions $U_i$ evaluated by our method at each step
$\mathbf{E}[B] = \chi C^2 M^2(\theta, \theta') + C M(\theta, \theta')$\footnote{Note that $\mathbf{E}[B]$ is typically  $<<$ $N$ and can be decreased using small step sizes. 
    If, however, $\mathbf{E}[B]>N$, then we can simply use standard MH in that iteration, similar to TFMH.}.
Importantly, this means the batch size may vary from iteration to iteration,\footnote{See the chapter summary for more details on why this causes problems in practice.} and the expected size depends on $\theta$ and $\theta'$. For example, \methodname{} may tend to set $B$ larger for larger-distance proposals with a higher $M(\theta, \theta')$.
Third, it samples (with replacement) a minibatch of size $B$, but for each data point it samples, it has some probability of \emph{ejecting} this point from the minibatch.
Finally, it accepts the proposed $\theta'$ with some probability, computed using a sum over the post-ejection minibatch.

Our method can be derived by carefully replacing the auxiliary variables in PoissonMH with \emph{local} Poisson variables whose distributions change each iteration depending on the pair $(\theta, \theta')$ (Appendix \ref{app:algo-derivation}). 
By construction \methodname{} is exact; it preserves the target distribution $\pi$ as its stationary distribution.
This is because \methodname{} is \emph{reversible}, meaning its transition operator $T$ satisfies $\pi(\theta) T(\theta , \theta') = \pi(\theta') T(\theta' , \theta)$ for any $\theta, \theta' \in \Theta$. This is a common condition that guarantees that a MCMC method has $\pi$ as its stationary distribution~\cite{levin2017markov, brooks2011handbook}.

Compared to previous exact methods, a significant benefit of \methodname{} is that we can prove theoretical guarantees on its efficiency. 
Specifically, its convergence speed is guaranteed to be close to standard MH and $\chi$ allows us to control how close. 
To show this, we lower bound the convergence rate of \methodname{} in terms of the \emph{spectral gap}, which is commonly used to characterize convergence speed in the MCMC literature \cite{rudolf2011explicit,hairer2014spectral,levin2017markov,zhang2019poisson,zhang2020amagold}. The larger the spectral gap, the faster the chain converges.

\begin{algorithm}[t]
  \caption{\methodname{}}
  \label{alg:poisson-mh}
  \begin{algorithmic}
    \State \textbf{given:} initial state $\theta \in \Theta$; proposal dist. $q$; hyperparameter $\chi$; Asm.~\ref{assump} parameters $c_i$, $C$, $M$
    \Loop
      \State \textbf{propose} $\theta'\sim q(\cdot|\theta)$ and \textbf{compute} $M(\theta, \theta')$
      \vspace{0.5em}
      \State $\triangleright$ Form minibatch $\mathcal{I}$
      \State \textbf{sample} $B \sim \text{Poisson}\left( \chi C^2M^2(\theta,\theta')  + CM(\theta,\theta')\right)$
      \State \textbf{initialize minibatch indices} $\mathcal{I} \leftarrow \emptyset$  (an initially empty multiset)
      \For{$b \in \{1,\ldots,B\}$}
        \State \textbf{sample} $i_b$ such that $\mathbf{P}(i_b = i) = c_i/C$, for $i=1\ldots N$
        \State \textbf{with probability} $\frac{\chi c_{i_b} C M^2(\theta, \theta') + \frac{1}{2}(U_{i_b}(\theta') - U_{i_b}(\theta) + c_{i_b}M(\theta,\theta'))}{\chi c_{i_b}C  M^2(\theta, \theta') + c_{i_b} M(\theta, \theta')}$ \textbf{add} $i_b$ to $\mathcal{I}$ 
      \EndFor
      \vspace{0.5em}
      \State $\triangleright$ Accept/reject step based on minibatch $\mathcal{I}$
      \State \textbf{compute MH ratio} $r \leftarrow
        \exp\left(2 \sum_{i \in \mathcal{I}} \operatorname{artanh}\left(
        \frac{U_i(\theta) - U_i(\theta') }{c_{i}M(\theta,\theta') (1 + 2 \chi CM(\theta,\theta'))} 
    \right) \right)
        \cdot \frac{q(\theta|\theta')}{q(\theta'|\theta)}$
      \State \textbf{with probability} $\min(1,r)$, set $\theta \leftarrow \theta'$
    \EndLoop
  \end{algorithmic}
\end{algorithm}

\begin{definition}
The \emph{spectral gap} of a reversible Markov chain is the distance between the largest and second-largest eigenvalues of its transition operator. That is, if the eigenvalues of the transition operator are $1 = \lambda_1 > \lambda_2 \ge \lambda_3 \cdots$, then the spectral gap is $\gamma = 1 - \lambda_2$. 
\end{definition}

\begin{theorem}\label{thm:spectral-gap}
\methodname{} (Algorithm~\ref{alg:poisson-mh}) is reversible with stationary distribution $\pi$. Let $\bar\gamma$ denote the spectral gap of \methodname{}, and let $\gamma$ denote the spectral gap of standard MH with the same target distribution and proposal distribution. Then,
  \[
    \textstyle
    \bar{\gamma}
    \ge
    \exp \left(-\frac{1}{\chi} - 2\sqrt{\frac{\log 2}{\chi}} \right)\cdot\gamma.
  \]
\end{theorem}
Intuitively, this theorem (proof in Appendix \ref{app:proof:spectral-gap}) suggests the convergence rate of \methodname{} is at most a constant slower than that of standard MH, and can be increased by adjusting the hyperparameter $\chi$.
Recall that $\chi$ also controls the batch size of \methodname{}. Effectively, this means $\chi$ is a \emph{dial} that allows us to directly tune the trade-off between convergence rate and batch size. When $\chi$ is large, the batch size $B$ is large and the spectral gap ratio, $\bar \gamma / \gamma$, is close to 1: the larger batch size is less scalable but keeps a high convergence rate. Conversely, when $\chi$ is small, the batch size is small and the spectral gap ratio is close to 0: we trade off slow-downs in convergence rate for scalability.
For example, for any $0 < \kappa < 1$, to guarantee the spectral gap ratio $\bar \gamma / \gamma \ge \kappa$ it suffices to set (Appendix~\ref{app:chi-value})
\begin{align}
    \label{eq:TunaMHEB}
    \chi &= \frac{4}{(1-\kappa)\log(1/\kappa)}, \text{giving an average batch size of}\;\;\nonumber\\ 
     \mathbf{E}[B] &= \frac{4 C^2 M^2(\theta,\theta')}{(1-\kappa)\log(1/\kappa)} + C M(\theta,\theta').
\end{align}

In practice, we usually want to minimize the wall-clock time to achieve a certain estimate error, which requires tuning $\chi$ to optimally balance scalability and efficiency. We attempt to derive a theoretically optimal value of $\chi$ in Appendix~\ref{app:optimal-value} by minimizing the product of the relaxation time---a measure of the number of steps needed---and the expected wall-clock time per step. Note that this product may be loose in bounding the total wall-clock time (we leave tightening this bound to future work), making the derived $\chi$ larger than necessary. In Section \ref{sec:tunamh:exp} we give a simple heuristic to tune $\chi$, which works well and is generally better than the derived value.

Theorem \ref{thm:spectral-gap} only requires the mild constraints of Assumption \ref{assump} on the target distribution, so applies in many scenarios and compares well to other exact methods. SMH further requires a Bernstein-von Mises approximation to have guarantees on its batch size and acceptance rate. PoissonMH provides convergence rate guarantees, but demands the strong assumption that the target distribution has a global upper bound on the energy. FlyMC does not have any theoretical guarantees on performance. 

%% file: section/30-algorithms/31-tunamh/314-tunamh-optimality.tex
\section{Towards Optimal Exact Minibatch MH}\label{sec:tunamh:optimality}

In Theorem \ref{thm:spectral-gap}, we expose the trade-off between convergence rate and batch size in \methodname{}. 
Here, we take this analysis a step further to investigate the limits of how efficient an exact minibatch MH method can be. 
To tackle this problem, we derive a lower bound on the batch size for any minibatch MH method that retains exactness and fast convergence. 
We then show that \methodname{} is asymptotically optimal in terms of its dependence on the problem parameters $C$ and $M$. 
In other words, it is not possible to outperform \methodname{} in this sense with a method in the class described by Algorithm~\ref{alg:subsampledMH}.

\begin{theorem}\label{thm:optimality}
Consider any stateless exact minibatch MH algorithm described by Algorithm~\ref{alg:subsampledMH}, any state space $\Theta$ (with $|\Theta| \ge 2$), any $C > 0$, and any function $M: \Theta \times \Theta \rightarrow \R^+$.
Suppose that the algorithm guarantees that, for some constant $\kappa \in (0,1)$, for any distribution, the ratio between the spectral gap of minibatch MH $\hat \gamma$ and the spectral gap of standard MH $\gamma$ is bounded by $\hat \gamma \ge \kappa \gamma$.
Then there must exist a distribution $\pi$ over $\Theta$ and proposal $q$ such that the batch size $B$ of that algorithm, when deciding whether to accept any transition $\theta 
\rightarrow \theta'$, is bounded from below by
\begin{align}\label{eq:lower-bound}
    \mathbf{E}[B] \ge \zeta \cdot \kappa \cdot \left(C^2 M^2(\theta,\theta') + C M(\theta,\theta') \right)
\end{align}
for some constant $\zeta > 0$ independent of algorithm and problem parameters.
\end{theorem}

To prove this theorem, we construct a random walk example over two states, then consider the smallest batch size a method requires to distinguish between two different stationary distributions~(Appendix~\ref{app:proof:optimality}). 
The impact of Theorem \ref{thm:optimality} is three-fold: 

First, it provides an upper bound on the performance of algorithms of Algorithm \ref{alg:subsampledMH}'s form: in each iteration, the average batch size of any exact minibatch MH method of the form of Algorithm \ref{alg:subsampledMH} must be set as in (\ref{eq:lower-bound}) in order to maintain a reasonable convergence rate. To the best of our knowledge, this is the first theorem that rigorously proves a ceiling for the possible performance of minibatch MH. 

Second, \methodname{} achieves this upper bound. 
In fact, Theorem \ref{thm:optimality} suggests that \methodname{} is \emph{asymptotically optimal} in terms of the problem parameters, $C$ and $M$. 
To see this, observe that when we ignore $\kappa$, both expressions that bound $\mathbf{E}[B]$ in (\ref{eq:TunaMHEB}) and (\ref{eq:lower-bound}) are $\bigTheta(C^2 M^2(\theta, \theta') + C M (\theta, \theta'))$.

Thus \methodname{} reaches the lower bound, achieving asymptotic optimality in terms of $C$ and $M$. (Of course, this sense of ``optimality'' does not rule out potential constant-factor improvements over \methodname{} or improvements that depend on $\kappa$.) 

Lastly, this result suggests directions for developing new exact minibatch MH algorithms: to be significantly faster than \methodname, we either need to introduce additional assumptions to the problem or to develop new stateful algorithms.

In prior work, when assuming a very concentrated posterior, some methods' batch size can scale in $\mathcal{O}(1)$ \cite{bardenet2017mcmc,bierkens2019zig,cornish2019scalable} or $\mathcal{O}(1/\sqrt{N})$ \cite{cornish2019scalable} in terms of the dataset size $N$ while maintaining efficiency. 
Theorem \ref{thm:optimality} is compatible with these results, further 
demonstrating this is essentially the \emph{best} dependency on $N$ an exact minibatch MH method can achieve. 
We show this by explicitly assuming the dependency of $C$ and $M$ on $N$, as in SMH \cite{cornish2019scalable}, yielding the following corollary (proof in Appendix~\ref{app:proof:cor1}): 
\begin{corollary}\label{col:bound}
Suppose that $C$ increases linearly with $N$ ($C = \bigTheta(N)$) and $M(\theta, \theta')$ scales in $\bigTheta(N^{-(h+1)/2})$ for some constant $h > 0$. Then the lower bound in Theorem \ref{thm:optimality} becomes $\bigTheta(N^{(1-h)/2})$. In particular, it is $\bigTheta(1)$ when $h=1$, and $\bigTheta(1/\sqrt{N})$ when $h=2$.
\end{corollary}

That is, \methodname{} matches the state-of-the-art's dependency on $N$, and this dependency is optimal. 
Similarly, since $C$ and $M$ are the only problem parameters in the lower bound in Theorem~\ref{thm:optimality}, we can also get the optimal dependency on the other problem parameters by explicitly assuming the relation of them with $C$ and $M$.

%% file: section/30-algorithms/31-tunamh/315-tunamh-experiments.tex
\section{Experiments}\label{sec:tunamh:exp}

We compare \methodname{} to MH, TFMH, SMH (i.e. TFMH with MAP control variates) and FlyMC. We only include PoissonMH in the Gaussian mixture experiment, as it is not applicable in the other tasks. All of these methods are unbiased, so they have the same stationary distribution. To ensure fair wall-clock time comparisons, we coded each method in Julia; our implementations are at least as fast as, if not faster than, prior implementations. 
For each trial, we use Gaussian random walk proposals. 
We tune the proposal stepsize separately for each method to reach a target acceptance rate, and report averaged results and standard error from the mean over three runs. We set $\chi$ to be roughly the largest value that keeps $\chi C^2M^2(\theta,\theta')<1$ in most steps; we keep $\chi$ as high as possible while the average batch size is around its lower bound $CM(\theta,\theta')$. We found this strategy works well in practice. 
We released the code at \url{https://github.com/ruqizhang/tunamh}.

\subsection{Robust Linear Regression}\label{sec:rlr}

\begin{figure*}[t!]
    \centering
    \begin{tabular}{cc}
    	\includegraphics[width=.45\linewidth]{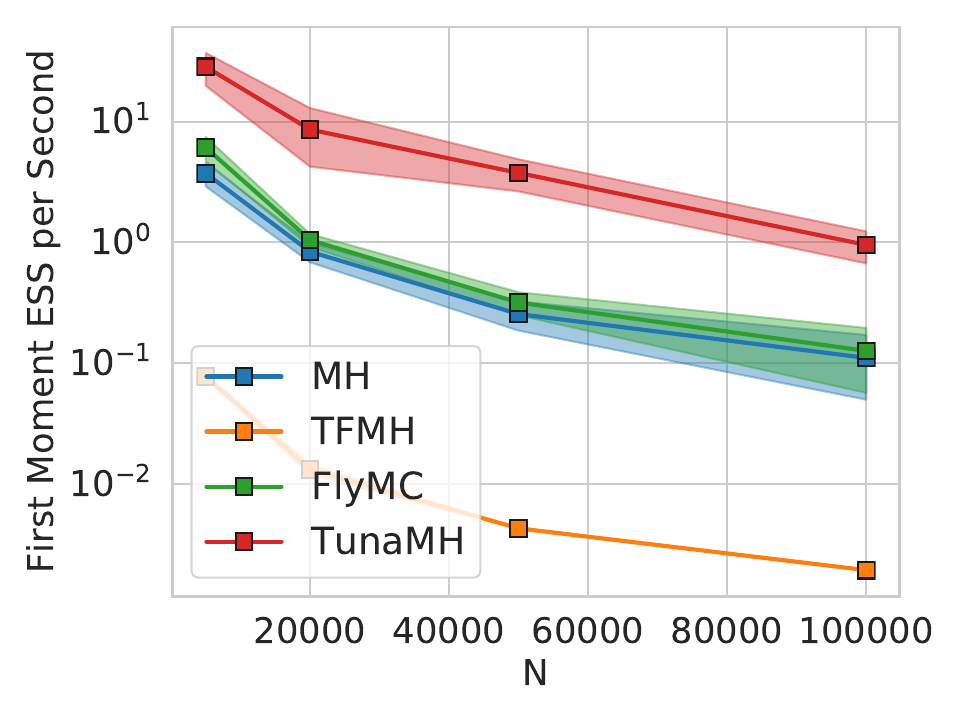} &
    	\includegraphics[width=.45\linewidth]{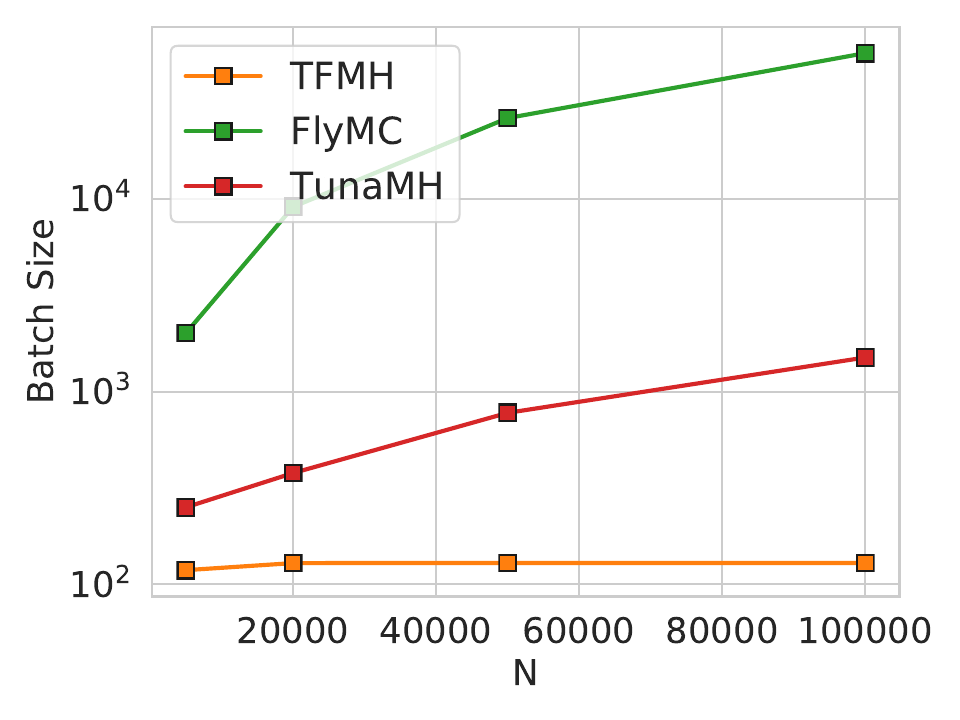}
    	\\
    	(a) &
    	(b) \\
    	\includegraphics[width=.45\linewidth]{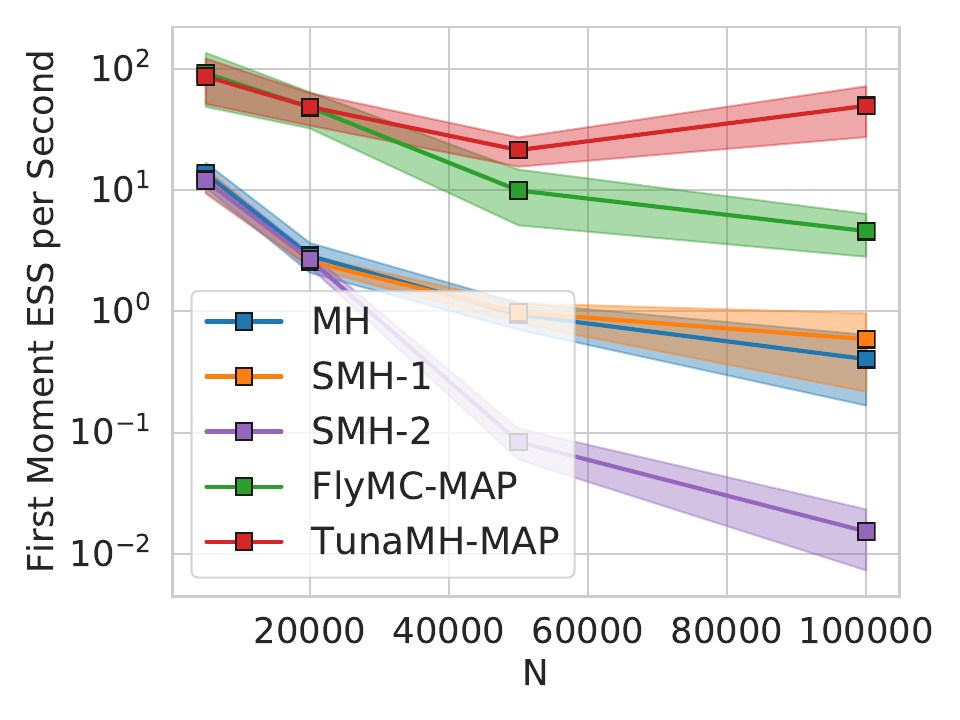} &
    	\includegraphics[width=.45\linewidth]{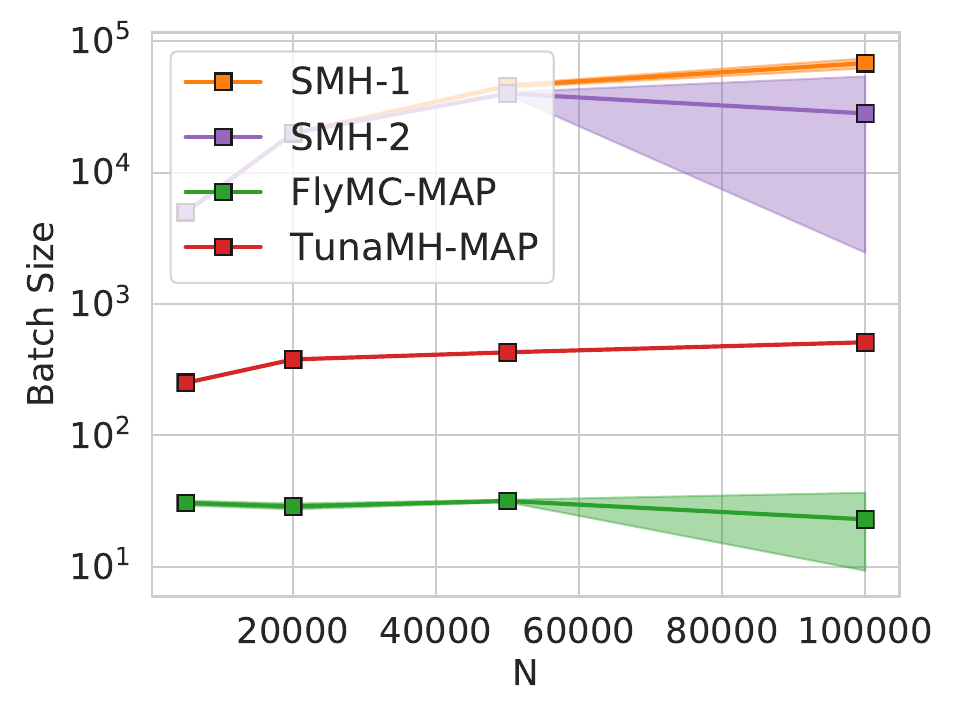}
    	\\		
    	(c) &
    	(d) \\		
    \end{tabular}
    \caption{Robust linear regression, $d=100$. (a) ESS/second without MAP. (b) Average batch size without MAP. (c)  ESS/second with MAP. (d) Average batch size with MAP.}
    \label{fig:linear}
\end{figure*}

We first test \methodname{} on robust linear regression \cite{cornish2019scalable,maclaurin2015firefly}. We use a Student's t-distribution with degree of freedom $v=4$ and set data dimension $d=100$ (Appendix \ref{app:tunamh:experiments}). We tune each method separately to a 0.25 target acceptance rate. To measure efficiency, we record effective sample size (ESS) per second---a common MCMC metric for quantifying the number of effectively independent samples a method can draw from the posterior each second \cite{brooks2011handbook}. Figure \ref{fig:linear}a shows \methodname{} is the most efficient for all dataset sizes $N$; it has the largest ESS/second. For minibatch MH methods, Figure \ref{fig:linear}b compares the average batch size. \methodname's batch size is significantly smaller than FlyMC's --- about 35x with $N=10^5$. TFMH has the smallest batch size, but this is because it uses a very small step size to reach the target acceptance rate (Table \ref{tab:stepsize} in Appendix~\ref{app:experiments:rlr}). This leads to poor efficiency, which we can observe in its low ESS/second.

\paragraph{MAP variants} Since TFMH and FlyMC have variants that use the \emph{maximum a posteriori} (MAP) solution to boost performance, we also test \methodname{} in this scheme. SMH uses MAP to construct control variates for TFMH to improve low acceptance rates. We consider both first- and second-order approximations (SMH-1 and SMH-2). FlyMC uses MAP to tighten the lower bound (FlyMC-MAP). For our method (\methodname-MAP) and MH (MH-MAP), we simply initialize the chain with the MAP solution. Figure \ref{fig:linear}c shows that \methodname{} performs the best even when previous methods make use of MAP. With control variates, SMH does increase the acceptance rate of TFMH, but this comes at the cost of a drastically increased batch size (Figure \ref{fig:linear}d) which we conjecture is due to the control variates scaling poorly in high dimensions ($d=100$).\footnote{Control variates worked well in the SMH paper \cite{cornish2019scalable} because all experiments had small dimension ($d=10$).} FlyMC-MAP tightens the bounds, entailing a decrease in the batch size. However, as clear in the difference in ESS/second, it is still less efficient than \methodname{} due to its strong dependence between auxiliary variables and the model parameters --- an issue that previous work also documents~\cite{quiroz2019speeding}.

\subsection{Truncated Gaussian Mixture}\label{sec:mog}
Next we test on a task with a multimodal posterior, a very common problem in machine learning. This demonstrates the advantage of \methodname{} not relying on MAP, because MAP is a single solution and therefore is unable to reflect all possible modes in multimodal distributions. As a result, methods that rely on MAP tuning or MAP-based control variates are unable to perform well on such problems.

\begin{figure*}[t!]
    \centering
    \begin{tabular}{ccc}		
     		\hspace{-.5cm}
    	\includegraphics[width=5cm]{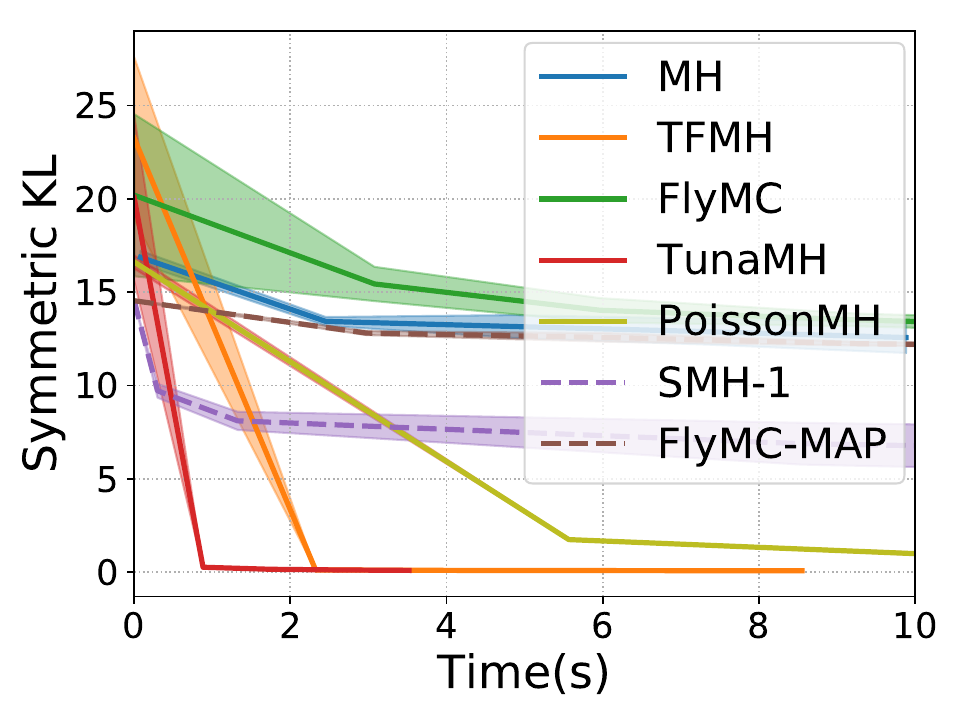}  &
    	\includegraphics[width=5cm]{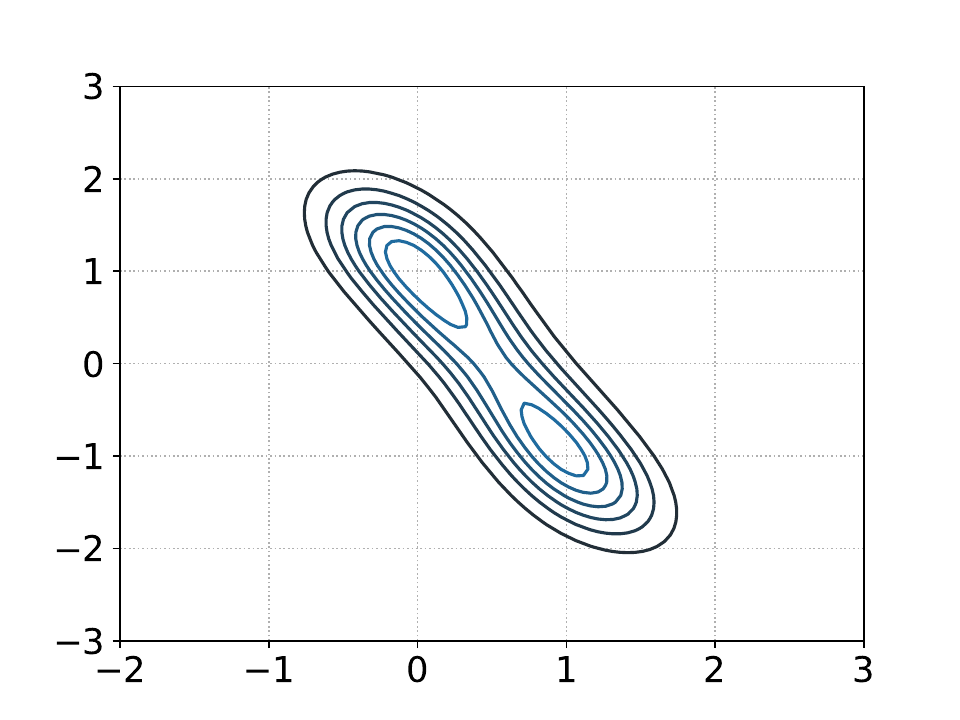} &
    	\includegraphics[width=5cm]{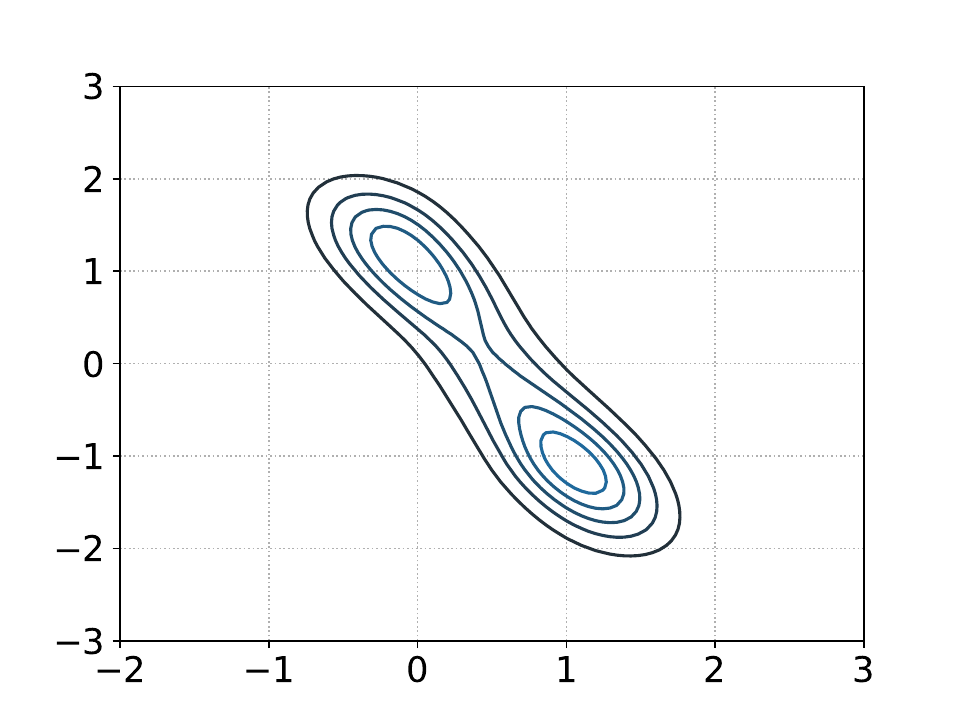}
    	\\		
    	(a) &
    	(b) &
    	(c) 
    	\hspace{-0mm}\\		
    \end{tabular}
    \caption{Truncated Gaussian mixture. (a) Symmetric KL comparison. (b) True distribution. (c) Denstity estimate of \methodname{} after 1 second.}
    \label{fig:mog}
\end{figure*}

We consider a Gaussian mixture. To get bounds on \methodname, TFMH, SMH, and FlyMC, we truncate the posterior, bounding $\theta_1, \theta_2\in [-3, 3]$ similar to Zhang and De Sa~\cite{zhang2019poisson}. We can include PoissonMH because its required bound exists after truncation. As in Seita et al.~\cite{seita2016efficient}, we use a tempered posterior $\pi(\theta)\propto \exp \left(-\beta\sum_i U_i(\theta)\right)$ with $N = 10^6$ and $\beta=10^{-4}$. Figure \ref{fig:mog}a compares performance, showing symmetric KL versus wall-clock time. \methodname{} is the fastest, converging after 1 second, whereas the others take much longer. As expected, SMH-1 performs worse than TFMH, verifying the control variate is unhelpful for multimodal distributions. FlyMC and FlyMC-MAP are also inefficient; their performance is on par with standard MH, indicating negligible benefits from minibatching.\looseness=-1 

\methodname{} also performs significantly better in terms of batch size, especially in comparison to PoissonMH (Table \ref{tab:bs}). This is due to \methodname's local bound on the energy, as opposed to PoissonMH's global bound. This also allows \methodname{} to run on more problem types, such as robust linear (Section \ref{sec:rlr}) and logistic (Section \ref{sec:lr}) regression. To illustrate the estimate quality, we also visualize the density estimate after 1 second; \methodname{}'s estimate (Figure \ref{fig:mog}c) is very close to the true distribution (Figure \ref{fig:mog}b), while the other methods do not provide on-par estimates within the same time budget (Appendix~\ref{app:experiments:mog}).

\subsection{Logistic Regression on MNIST}\label{sec:lr}

\begin{figure*}[t!]
    \centering
    \begin{tabular}{cccc}		
    	\includegraphics[width=4.2cm]{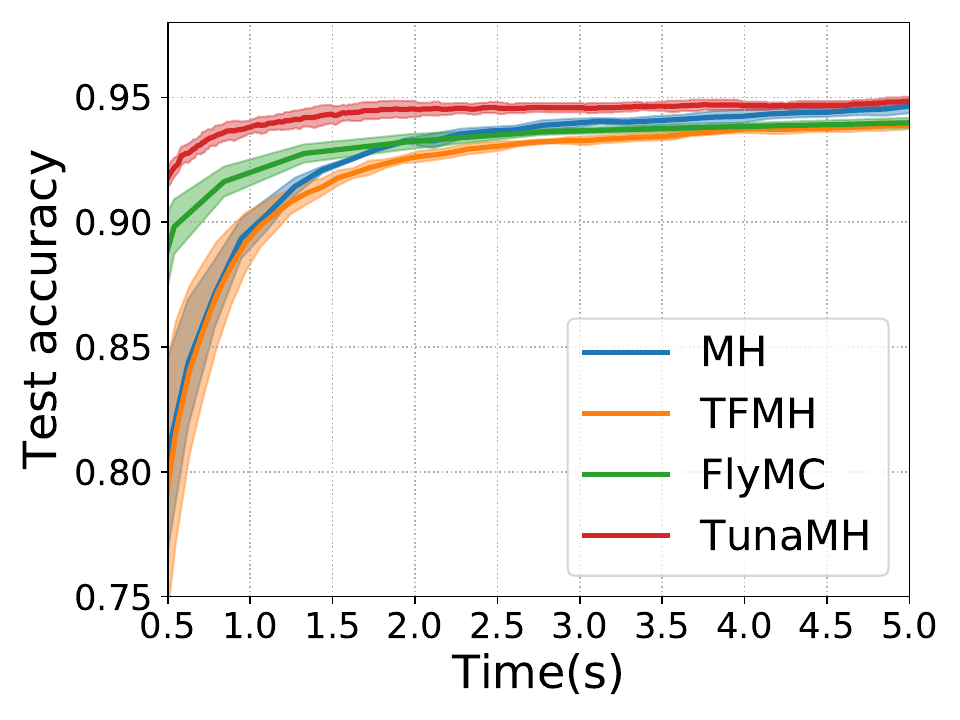}  &
    	\includegraphics[width=4.2cm]{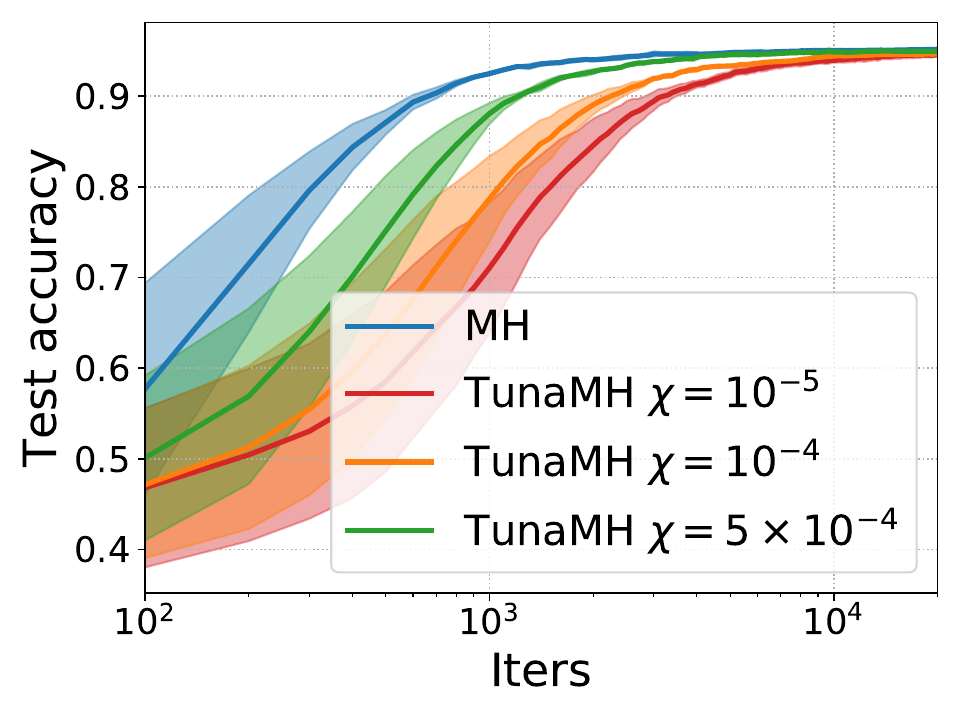} &
    	\includegraphics[width=4.2cm]{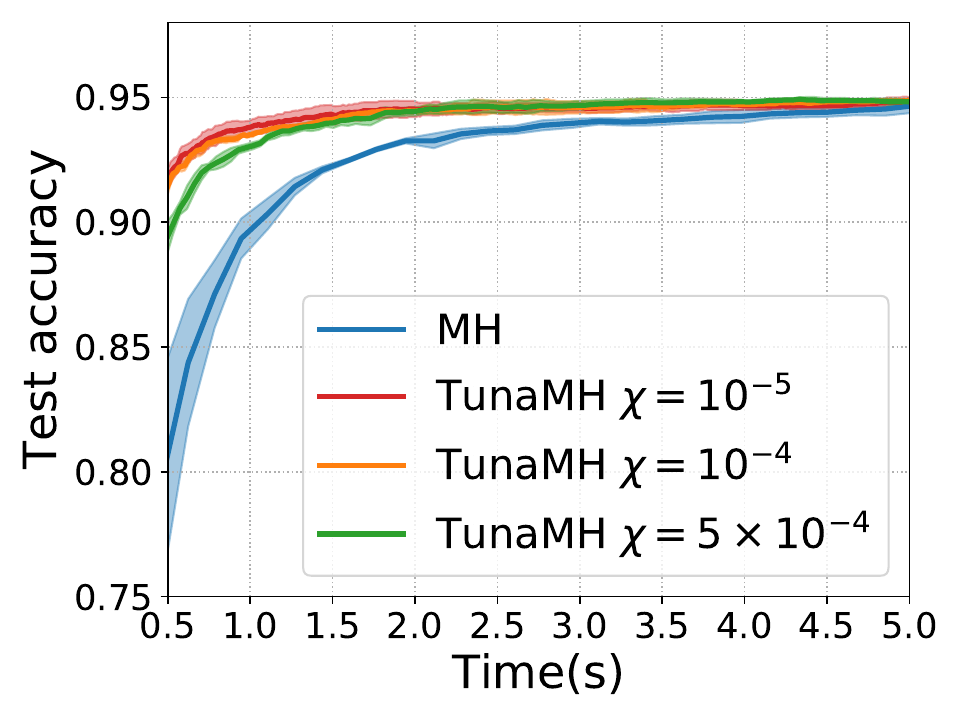}
    	\\		
    	(a)  &
    	(b) &
    	(c) 
    	\hspace{-0mm}\\		
    \end{tabular}
    \caption{MNIST logistic regression. (a) Test accuracy comparison. (b)-(c) \methodname{}'s test accuracy for various $\chi$. Batch size for $\chi=10^{-5}, 10^{-4}, 5\times 10^{-4}$ is 504.07, 810.35 and 2047.91 respectively.}
    \label{fig:logistic}
\end{figure*}

Lastly we apply \methodname{} to logistic regression on the MNIST image dataset of handwritten number digits. Mirroring the work of FlyMC \cite{maclaurin2015firefly}, we aim to classify 7s and 9s using the first 50 principal components as features. We set $\chi=10^{-5}$ following our heuristic. In Figure \ref{fig:logistic}a we see that \methodname{} is the fastest of all methods to converge, as measured by wall-clock time. We also compare average batch size in Table \ref{tab:bs}. \methodname's average batch size is 4$\times$ smaller than FlyMC's. TFMH again has the smallest batch size, but sacrifices efficiency by using a small step size in order to achieve the target acceptance rate. Thus, overall, TFMH is again inefficient in these experiments.

\addtolength{\tabcolsep}{1pt} 
\begin{table}[h]
  \caption{Avg. batch size $\pm$ SE from the mean on 3 runs. PoissonMH not applicable to logistic reg.}
  \label{tab:bs}
  \small
  \centering
  \begin{tabular}{lcccccc}
    \toprule
    Tasks   & TFMH & FlyMC & PoissonMH & \methodname{}\\
    \midrule
    Gaussian Mixture   & $13.91\pm0.016$  & $811.52\pm234.16$ & $3969.67\pm327.26$ & $86.45\pm0.04$ \\
    Logistic Regression & $39.28\pm0.12$ & $1960.19\pm150.96$ & --- & $504.07\pm0.33$ \\
    \bottomrule
  \end{tabular}
\end{table}
\addtolength{\tabcolsep}{1pt} 

\paragraph{Effect of Hyperparameter $\chi$}
To understand the effect of $\chi$ in \methodname{}, we report results with varying $\chi$. Figure \ref{fig:logistic}b plots test accuracy as a function of the number of iterations. As $\chi$ increases, \methodname's convergence rate approaches standard MH. This verifies our theoretical work: $\chi$ acts like a dial to control convergence rate and batch size trade-off---mapping to the efficiency-scalability trade-off. Figure \ref{fig:logistic}c shows \methodname's wall-clock time performance is not sensitive to $\chi$, as the performance is superior to standard MH regardless of how we set it. However, $\chi$ needs to be tuned in order to achieve the best performance. Previous methods do not have such a dial, so they are unable to control this trade-off to improve the sampling efficiency.

%% file: section/30-algorithms/31-tunamh/316-tunamh-conclusion.tex
\section{Conclusion and Future Work}
After demonstrating that inexact methods can lead to arbitrarily incorrect inference, we focus our work in this paper on exact minibatch MH methods. We propose a new exact method, \methodname{}, which lets users trade off between batch size and guaranteed convergence rate---between scalability and efficiency. We prove a lower bound on the batch size that any minibatch MH method must use to maintain exactness and convergence rate, and show \methodname{} is asymptotically optimal. Our experiments validate these results, demonstrating that \methodname{} outperforms state-of-the-art exact methods, particularly on high-dimensional and multimodal distributions.

To guide our analysis, we formalized a class of stateless, energy-difference-based minibatch MH methods, to which most prior methods belong. While \methodname{} is asymptotically optimal for this class, future work could develop new exact methods that are better by a constant factor or on some restricted class of distributions. It would also be interesting to develop effective theoretical tools for analyzing stateful methods, since these methods could potentially bypass our lower bound.

%% file: section/30-algorithms/32-cdgrab/32-cdgrab-main.tex
\chapter{Coordinating Distributed Example Orders for Provably Accelerated Training}\label{chapter:cdgrab}

We next investigate how there are better-than-random example orders for SGD that improve convergence rate in distributed settings.\\

\noindent \textbf{Chapter summary}: 
Recent research on online Gradient Balancing (GraB) has revealed that there exist permutation-based example orderings for SGD that are guaranteed to outperform random reshuffling (RR). Whereas RR arbitrarily permutes training examples, GraB leverages stale gradients from prior epochs to order examples --- achieving a provably faster convergence rate than RR. However, GraB is limited by design: while it demonstrates an impressive ability to scale-up training on \emph{centralized} data, it does not naturally extend to modern \emph{distributed} ML workloads. We therefore propose \emph{Coordinated Distributed GraB} (CD-GraB), which uses insights from prior work on kernel thinning to translate the benefits of provably faster permutation-based example ordering to distributed settings. With negligible overhead, CD-GraB exhibits a linear speedup in convergence rate over centralized GraB and outperforms distributed RR on a variety of benchmark tasks.\\

\noindent This chapter is a licensed derivative copy of work published at \emph{NeurIPS 2023}~\cite{cooper2023cdgrab}. 

\input{section/30-algorithms/32-cdgrab/321-cdgrab-intro}
\input{section/30-algorithms/32-cdgrab/322-cdgrab-prelim-and-rw}
\input{section/30-algorithms/32-cdgrab/323-cdgrab-dgrab}

\input{section/30-algorithms/32-cdgrab/324-cdgrab-theory}
\input{section/30-algorithms/32-cdgrab/325-cdgrab-experiments}
\input{section/30-algorithms/32-cdgrab/326-cdgrab-conclusion}

%% file: section/30-algorithms/32-cdgrab/321-cdgrab-intro.tex
\section{Introduction}

Random reshuffling, which samples training-data examples without replacement, has become the \emph{de facto} example-ordering method in modern deep-learning libraries~\citep{pytorchshuffle}, given that it tends to accelerate optimizer convergence in practice. 
However, some recent theoretical work has identified cases in which random reshuffling can lead to data orderings that have a poor effect on convergence~\citep{desa2020shuffle, yun2021can, rajput2021permutationbased}. 

This has encouraged a line of research to investigate if there exist provably better permutation-based orderings that afford greater scalability in training~\citep{lu2021general,mohtashami2022characterizing,lu2022grab}. 
Notably, Lu et al.~\cite{lu2022grab} connects permuted-order SGD to the \emph{herding problem}~\citep{harvey2014near}, and proposes the herding-based online Gradient Balancing algorithm (\grab), which converges provably faster than random reshuffling, and does so  with little memory or computational overhead. 
In fact, in follow-on work, Cha et al.~\cite{cha2023tighter} proves that \grab{} is optimal: in theory, \grab{} is the fastest possible permutation-based example ordering algorithm.

These results are very exciting, suggesting that \grab{} should unseat random reshuffling as the example ordering method-of-choice for SGD; however, they only hold with respect to a \emph{single} machine. 
\grab{} is optimal in settings with \emph{centralized} data, but does not naturally translate to problems of modern-ML scale, which demand that training workloads be distributed across \emph{multiple parallel} workers that each only have access to a subset of the training data. This drawback raises an important question:
\textit{Can we simultaneously achieve the scalability benefits of distributed training and provably faster permutation-based example ordering for SGD --- both in theory and in practice?}

In this chapter, we show that it is indeed possible to attain these twin objectives. 
To do so, we suggest the online \textbf{C}oordinated \textbf{D}istributed \textbf{Gra}diant \textbf{B}alance algorithm (\dgrab), which leverages insights from kernel thinning to elevate the herding framework of centralized \grab{} (\cgrab) to the parallel setting. 
Felicitously, as a side effect, this choice of formulation brings about positive practical performance benefits (that can also improve the empirical behavior of centralized \cgrab). 
Using the exact same assumptions as the original \cgrab{} paper, \textbf{we show analytically that coordinating example orders across parallel workers leads a linear speedup in convergence rate}. 
For $\epochs$ epochs and $\workers$ parallel workers, each with access to $\workerexamples$ examples, \dgrab's convergence rate is $\tilde{O}((\workers\workerexamples\epochs)^{-2/3})$ on smooth, non-convex objectives and $\tilde{O}((\workers\workerexamples\epochs)^{-2})$ under the Polyak-\L ojasiewicz (P.L.) condition.\footnote{In this paper, we use $\tilde O$ by convention to hide logarithmic factors in the problem parameters.} 

We run a series of experiments to verify these improvements in practice, implementing \dgrab{} on a single node that distributes computation across multiple GPUs. 
We also run an ablation study in order to disentangle the benefits of parallelism from the positive side effects of using kernel thinning to formulate the \dgrab{} algorithm. 
Similar to how centralized \cgrab{} demonstrates improved generalization over centralized random reshuffling (\shuffle), we observe that \dgrab{} exhibits improved generalization over distributed random reshuffling (\dshuffle). 
Altogether, the success of our work suggests a new distributed training paradigm to explore in future work, which we call the \emph{Order Server} (Section~\ref{sec:cdgrab:conclusion}). 
In summary, we:
\begin{itemize}
    \item Propose the online \textbf{C}oordinated \textbf{D}istributed \textbf{Gra}dient \textbf{B}alancing (\dgrab) algorithm, which enables provably accelerated training using SGD in the parallel setting (Section~\ref{sec:cdgrab:dgrab});
    \item Prove that the convergence rate for \dgrab{} exhibits a linear speedup over \cgrab, using the exact same assumptions as the original \cgrab{} paper (Section~\ref{sec:cdgrab:theory}); 
    \item Produce extensive empirical validation of \dgrab's improved scalability on a variety of tasks in deep learning and on large-scale logistic regression  (Section~\ref{sec:cdgrab:experiments}). 
\end{itemize}

%% file: section/30-algorithms/32-cdgrab/322-cdgrab-prelim-and-rw.tex
\section{Preliminaries and Related Work}\label{sec:cdgrab:prelimrw}

In this section, we discuss the preliminaries and prior scholarship on permutation-based example ordering, with particular attention paid to the centralized online Gradient Balancing Algorithm (\cgrab)~\cite{lu2022grab}. 
This lays the groundwork for how our coordinated, distributed \grab{} algorithm (Section~\ref{sec:cdgrab:dgrab}) imparts the efficiency guarantees of \cgrab{} to the parallelized regime (Section~\ref{sec:cdgrab:theory}).

\paragraph{Ordering data examples during training.} 
Training a model can be formulated as minimizing a differentiable loss function $\loss:\R^d\rightarrow\R$ over $\examples$ data examples. 
The goal of this minimization is to obtain the target model weights $\weights^* = \arg\min_{\weights}\loss(\weights)$, where $\loss(\weights) = \frac{1}{\examples} \sum_{\exindex=1}^{\examples} \loss(\weights; \exindex)$,
for which $\loss(\weights; \exindex)$ denotes the loss incurred on the $\exindex$-th example. 
A typical training process iteratively updates the model parameters $\weights$ by scanning over the $\examples$ data examples repeatedly, with $t$-th scan (or epoch) following
\begin{align}
\label{equ:grab:main_update}
    \weights_{\eindex}^{\exindex+1} = \weights_\eindex^{\exindex} - \alpha \nabla \loss(\weights_\eindex^\exindex; \pi_{\eindex}(\exindex)), \hspace{.5em} \forall \exindex \in [\examples], 
\end{align}

\noindent where $\alpha$ denotes the learning rate, and $\perm_t:[\examples] \rightarrow [\examples]$ denotes a permutation ordering\footnote{While without-replacement orderings are most common in large-scale learning~\citep{bottou2012stochastic}, 
ordering strategies need not be permutations, 
    e.g.,  
    with-replacement sampling~\citep{schmidt2017minimizing,needell2014stochastic,lu2021variance}  
    or curriculum learning~\citep{graves2017automated,matiisen2019teacher,soviany2022curriculum}.
} 
adopted in the $t$-th epoch from which the examples are chosen to compute gradients, $\weights_t^1$ denotes the  initial model weights for the $\eindex$-th epoch, and $\weights_\eindex^{\exindex}$ denotes the model weights after $\exindex-1$ gradient updates in the $\eindex$-th epoch.\footnote{Note that we write (\ref{equ:grab:main_update}) in terms of per-example-$j$ gradients.
}

The choice of ordering $\perm$ can have a significant effect on optimizer performance. 
Two popular methods, which can demonstrate convergence speedups in practice, are 1) random reshuffling (\shuffle)~\citep{ying2017performance}, for which the permutations are random and differ over epochs, and 2) Shuffle Once (SO)~\citep{bertsekas2011incremental,gurbuzbalaban2019convergence}, for which a random permutation is computed once and remains fixed for all epochs.
Recht and R{\'{e}}~\cite{recht2012toward} conducted the first theoretical investigation of \shuffle, while subsequent works like Yun et al.~\cite{yun2021can} and De Sa~\cite{desa2020shuffle} have given counterexamples in which \shuffle{} leads to orderings that have a poor effect on convergence. 
Altogether, many studies indicate that \shuffle{} and \so{} only provide efficiency benefits under certain conditions~\citep{haochen2019random,gurbuzbalaban2021random,mishchenko2020random}.%

These limitations of \shuffle{} and \so{} have motivated research to identify permutations that outperform random ones. 
Rajput et al.~\cite{rajput2021permutationbased} introduces an \shuffle{} variant that achieves improved convergence for quadratics by reversing the ordering every other epoch. 
Other non-\shuffle-based methods pick efficient orderings based on correlations between adjacently selected examples.  
In a recent line of work, Lu et al.~\cite{lu2021general} proves that faster convergence is possible for SGD when the averages of consecutive stochastic gradients converge faster to the full gradient. 
Based on this result, in follow-on work Lu et al.~\cite{lu2022grab} proposes the centralized online Gradient Balancing algorithm (\grab), which outperforms \shuffle, and upon which we base this work.

\subsection{\cgrab: Optimal, online, permutation-based example ordering for centralized ML}\label{sec:cdgrab:cgrab}

\grab{} is a permutation-based example-ordering algorithm that identifies provably better-than-random orderings \emph{in centralized, single-node settings} for SGD. 
\grab{} finds such orderings by leveraging information in stale stochastic gradients from previous epochs to guide  ordering in the next epoch. 
More formally, for smooth, non-convex objectives, Lu et al.~\cite{lu2022grab} proves that any permutation $\perm^*$ that guarantees
\begin{align}
\label{equ:grab:grad_error}
    \textstyle
    \max_{k\in[\examples]} \left\| \sum_{\exindex=1}^k\nabla \loss(\weights;\perm^*(\exindex)) - \nabla \loss(\weights)
 \right\|_\infty = \tilde{O}(1) \hspace{.5cm} 
 \text{(} \nabla \loss(\weights) \text{ is the average gradient)},
\end{align}

\noindent will yield a convergence rate of $\tilde{O}((\examples\epochs)^{-2/3})$ (for epochs $\epochs$) for SGD, which is superior to the $O(\examples^{-1/3}\epochs^{-2/3})$ convergence rate of random reshuffling~\cite{mishchenko2020random}. 

\paragraph{\cgrab's connection to herding and balancing.} 
To find such a permutation $\perm^*$, Lu et al.~\cite{lu2022grab} connect (\ref{equ:grab:grad_error}) to the \emph{herding problem} and vector \emph{balancing}~\citep{harvey2014near, welling2009herding}. 
Understanding why \grab{} does not naturally extend to the distributed setting --- and our main contributions (Sections~\ref{sec:cdgrab:dgrab} and~\ref{sec:cdgrab:theory}) --- requires some additional details on the fundamentals of herding: 

Given $\examples$ vectors\footnote{Herding does not have an optimization context. 
    Here, $\examples$ does \emph{not} refer to the number of data examples used in training (\ref{equ:grab:main_update}); rather, $\examples \in \sZ^+$ describes the size of a set of arbitrary vectors. We slightly abuse notation because we execute the herding subroutine on exactly $\examples$ gradients (Section~\ref{sec:cdgrab:dgrab}), which happen to equal the number of $\examples$ examples.
} 
$\{\exj\}_{\exindex=1}^\examples$ ($\exj \in \R^d$), $\norm{\exj}_2 \le 1$ ($\forall \exindex$), herding identifies 
a permutation $\perm^*$ such that
\begin{align}
\label{equ:herding:objective}
    \textstyle
    \max_{k \in [\examples]} \norm{\sum_{\exindex=1}^k \left( \ex_{\perm^*(\exindex)} - \barex \right)}_\infty = \tilde{O}(1), \hspace{.5cm} \text{ where } \barex = \frac{1}{\examples}\sum_{\exindex=1}^\examples \exj.
\end{align} 

\noindent It is clear that (\ref{equ:herding:objective}) generalizes  (\ref{equ:grab:grad_error}), which 
is a specific case of herding in an optimization setting. 

Harvey and Samadi solve (\ref{equ:herding:objective}) with a method called \emph{balancing}~\citep{harvey2014near}. 
Balancing uses a \emph{signed} version of the herding problem to optimize any given permutation $\perm$ to reduce the bound in (\ref{equ:herding:objective}). That is, balancing formulates the signed herding problem
\begin{align}
\label{equ:herding:signed_objective}
    \textstyle
    \max_{k \in [\examples]} \norm{\sum_{\exindex=1}^k \sgn_{\perm(\exindex)} \left( \ex_{\perm(\exindex)} - \barex \right) }_\infty, \hspace{1em} \text{where} \hspace{.5em} \{\sgn_\exindex\}_{\exindex=1}^\examples \in\{+1, -1\}.
\end{align}
\noindent Given a group of such signs $\{\sgn_\exindex\}_{\exindex=1}^\examples$ and an arbitrary permutation $\perm$, Harvey and Samadi prove that Algorithm~\ref{alg:reorder} produces a new permutation $\perm'$ such that 
{\begin{align*}
    \textstyle
    \max \limits_{k \in [\examples]} \norm{\sum_{\exindex=1}^k \left( \ex_{\perm'(\exindex)} - \barex \right)}_\infty  \, 
    \leq  \; \frac{1}{2} \max \limits_{k \in [\examples]} \norm{\sum_{\exindex=1}^k \sgn_{\perm(\exindex)}\left( \ex_{\perm(\exindex)} - \barex \right)}_\infty + \frac{1}{2}\max \limits_{k \in [\examples]} \norm{\sum_{\exindex=1}^k \left( \ex_{\perm(\exindex)} - \barex \right)}_\infty.
\end{align*}}%

\noindent This says that, with new permutation $\perm'$, the objective of (\ref{equ:herding:objective}) now approaches the bound of (\ref{equ:herding:signed_objective}).
Importantly, recent advances show that it is quite cheap to find a group of signs, such that (\ref{equ:herding:signed_objective}) is on the order of $\tilde{O}(1)$ (e.g., Alweiss et al.~\cite{alweiss2021discrepancy}, in  Algorithm~\ref{alg:pairbalance}). 
We are therefore able to call Algorithm~\ref{alg:reorder} repeatedly, which will eventually obtain the $\perm^*$ that solves the $\tilde{O}(1)$ herding objective in (\ref{equ:herding:objective}).\looseness=-1 

\paragraph{\cgrab's application of herding to gradient balancing.}
Lu et al.~\cite{lu2022grab} applies this framework of herding and balancing to develop  \grab{}, i.e., to minimize (\ref{equ:grab:grad_error}).
The main challenge for the success of this approach is to find the right gradients $\exj$ in the optimization context of (\ref{equ:grab:grad_error}). 
Notably, the herding and balancing framework requires the vector mean $\barex$ in advance. To satisfy this requirement, \grab{} ``centers'' the gradient vectors using a \emph{stale mean}. 
That is, \grab{} runs the herding algorithm on vectors that are defined as

\begin{algorithm}[t]
\caption{Reordering Vectors based on Balanced Signs [Harvey and Samadi~\cite{harvey2014near}]}\label{alg:reorder}
    	\begin{algorithmic}[0]
    	\State \textbf{input:} a group of signs $\{\sgn_\exindex\}_{\exindex=1}^\examples$, initial order $\perm$
    	\State \textbf{initialize:} two order-sensitive lists $L_{\text{pos}}\leftarrow [\hspace{0.1em}]$, $L_{\text{neg}}\leftarrow [\hspace{0.1em}]$.
    	\For{$\exindex = 1\dots\examples$}
    	    \State $L_{\text{pos}}.\textsf{\scriptsize{append}}(\perm(\exindex))$ \textbf{if} $\signj$ is $+1$ \textbf{else} $L_{\text{neg}}.\textsf{\scriptsize{append}}(\perm(\exindex))$.
    	\EndFor
    	\State \textbf{return:} new order $\perm'\coloneqq\textsf{\scriptsize{concat}}(L_{\text{pos}}, \textsf{\scriptsize{reverse}}(L_{\text{neg}}))$.
    	\end{algorithmic}
\end{algorithm}

\begin{align}
\label{equ:grab:stale_mean}
    \exj = \nabla \loss(\weights_\eindex^\exindex;\perm_\eindex(\exindex)) - \frac{1}{\examples}\sum_{p=1}^{\examples}\nabla \loss(\weights_{\eindex-1}^p;\perm_{\eindex-1}(p)),
\end{align}

\noindent where $\weightst^p$ denotes the model weights after $p-1$ updates in the $t$-th epoch, and $\perm_\eindex$ denotes the permutation adopted in the $t$-th epoch. 
Lu et al.~\cite{lu2022grab} proves that this definition of $\exj$ preserves the benefits of balancing with negligible noise or overhead. 
The only overhead comes from storing the running average of the gradients in epoch $\eindex -1$ to ``center'' the gradients in the subsequent epoch $\eindex$.

With this approach, Lu et al.~\cite{lu2022grab} proves that \grab{} demonstrates more efficient convergence than \shuffle{} for SGD. 
Better still, Chat et al.~\cite{cha2023tighter} demonstrates that \grab{} is in fact the \emph{optimal} permutation-based ordering method for SGD: 
in theory, it is not possible to produce a permutation-based ordering in the centralized setting that achieves a faster convergence rate for SGD. 

Despite \grab's clear benefits over \shuffle, it assumes local access to all examples. 
This assumption does not hold for popular, modern, parallel settings (e.g., parameter server~\citep{li2014ps}), in which workers only have access to subsets of examples. 
No present work has attempted to investigate \grab's applicability to this setting.
While some work has studied distributed \shuffle{} (\dshuffle)~\citep{yun2021minibatch,huang2021distributed,malinovsky2022server,sadiev2022federated}, it remains an open question if \grab's efficiency benefits for SGD can be conferred to the modern-scale, distributed-ML setup. 

%% file: section/30-algorithms/32-cdgrab/323-cdgrab-dgrab.tex
\section{\dgrab: A Provably Efficient Ordering Algorithm for Distributed Training}\label{sec:cdgrab:dgrab}

Our main contribution is to elevate \grab{} to the parallel regime, so that distributed training can enjoy the efficiency benefits of provably better example ordering. 
Based on the preliminaries, we can now explain why this is not a straightforward task: 
\textbf{While \grab{} achieves the optimal convergence rate for SGD on centralized data, it does not naturally translate to a distributed setting} (Section~\ref{sec:cdgrab:dgrab:issues}). 
Our key insights for resolving these problems are to reformulate the herding framework in Lu et al.~\cite{lu2022grab} to work in parallel, and to leverage insights from 
kernel thinning~\citep{dwivedi2021kernel, dwivedi2022generalized, barp2022targeted} to derive the \emph{online} $\mathsf{PairBalance}$ algorithm, which solves this parallelized herding objective (Section~\ref{sec:cdgrab:dgrab:solution}). 
Lastly, we present the full-stack \dgrab{} algorithm that makes our solution work in practice (Section~\ref{sec:cdgrab:dgrab:algo}). The server implements online $\mathsf{PairBalance}$, which coordinates gradient information from the distributed workers in training epoch $\eindex$ in order to determine a provably efficient example order for the next epoch $\eindex + 1$ (Section~\ref{sec:cdgrab:theory}).\looseness=-1 

\subsection{Issues with \grab{} in the distributed setting}\label{sec:cdgrab:dgrab:issues}

To clarify the issues with distributing \grab, we first need to define the distributed training setup more precisely. 
We consider the standard data-parallel training regime with $\workers$ parallel workers, where each worker keeps a copy of the model weights $\weights\in\R^d$ and maintains $\workerexamples = \examples / \workers$ local 
examples.\footnote{Without loss of generality, 
    we assume the $\examples$ examples are divided evenly among the $\workers$ workers and $\workerexamples$ is even.
} 
As in many data-parallel training applications,\footnote{One such popular 
paradigm is federated learning, in which edge devices collaboratively train a model via small local updates
~\citep[e.g.]{mcmahan2017communication}. 
    Federated learning typically involves highly imbalanced loads, heterogeneous data, partial user participation, and additional privacy-preserving mechanisms. 
    These characteristics are orthogonal to what we consider here for example order. 
    If we were to allow for such data organization, we would need to assume non-global communication per iteration or additional constraints on how global communication occurs.
    For \dgrab, we focus on the regime of using parallelism to accelerate training.
} 
such as geo-distributed model training \citep{yuan2022decentralized}, we assume \emph{the data examples cannot be shared or moved across workers}.
More formally, this setup can be expressed as
\begin{equation}
\label{equ:d-grab:objective}
\textstyle
\min_{\weights \in \R^d} \left[ \loss(\weights) = \frac{1}{\workers}\sum_{\windex=1}^{\workers} \loss^\windex(\weights) \right] \quad \text{with} \quad \loss^\windex(\weights) = \frac{1}{\workerexamples}\sum_{\exindex=1}^{\workerexamples} \loss^\windex(\weights; \exindex),
\end{equation}

\noindent where $\loss^\windex(\weights; \exindex): \R^d \rightarrow \R$, $\exindex \in [\workerexamples]$, denotes the loss incurred on the $\exindex$-th example on the $\windex$-th worker for model weights $\weights$. We can now consider running (\ref{equ:grab:main_update}) using this setup, for which each worker scans over their $\workerexamples$ local-data examples using (potentially) different permutations. We denote $\perm_{\eindex,\windex}: [\workerexamples] \rightarrow [\workerexamples]$ as the permutation-based ordering adopted on the $\windex$-th worker in the $\eindex$-th training epoch. Adjusting (\ref{equ:grab:main_update}) to accommodate the setup in (\ref{equ:d-grab:objective}), the update to the model can be summarized as
\begin{align}
\label{equ:d-grab:main_update}
    \textstyle
    \weights_{\eindex}^{\exindex+1} = \weights_\eindex^{\exindex} - \frac{\alpha}{\workers}\sum_{\windex=1}^\workers \nabla \loss^\windex(\weights_\eindex^\exindex; \pi_{\eindex,\windex}(\exindex)), \hspace{.5em} \forall \exindex \in [\workerexamples]. 
\end{align}

\noindent That is, in epoch $\eindex$, each worker $\windex$ selects their respective, local $\exindex$-th example according to $\{\perm_{\eindex,\windex}\}_{\windex=1}^\workerexamples$ in order to compute stochastic gradients (Appendix, Chapter~\ref{chapter:app:cdgrab}). 

\textbf{Following this setup, Algorithm~\ref{alg:reorder} no longer guarantees the $\tilde{O}(1)$ bound to the herding problem~(\ref{equ:herding:objective})}, a bound that is valid only when \emph{all} data examples can be permuted \emph{freely}~\citep{harvey2014near}. 
This constraint is fine for centralized \grab, but, in distributed training, parallel workers only have access to a \emph{subset} of examples.
Distributed training requires that \emph{worker-specific permutations only involve the examples in their respective local subsets}. 
Further, recall that \grab{} uses stale means to center gradients (\ref{equ:grab:stale_mean}) in order to solve the herding objective. This, too, causes problems in distributed training. In practice, it is typical to employ larger learning rates $\alpha$ for greater scalability~\citep{smith2018don}; larger $\alpha$ increases the discrepancy between averaged gradients in adjacent epochs, which, in turn, would make \grab's use of stale means unreliable. 

\subsection{Our efficient solution: parallel herding \& pair balancing}\label{sec:cdgrab:dgrab:solution}

To address the limitations presented in the prior section, which preclude the direct application of \grab{} to distributed training, we will need to \textbf{1) reformulate the herding problem to fit the parallel setting, and 2)  redesign how to do gradient balancing}, such that it both solves our new herding formulation and allows for  reliability with higher learning rates. 
We now present our solution to both these problems; we introduce the \emph{parallel herding} problem and the online $\mathsf{PairBalance}$ subroutine that solves it.

\paragraph{Parallel herding.} To extend herding to the parallel setting, consider the following setup: 
There are $\workers$ workers, which each have local access to $\workerexamples$ vectors. 
Let $\ex_{\windex,\exindex} \in \R^d$ denote the vector indexed by $\exindex$ on the $\windex$-th worker. 
Assuming $\norm{\ex_{\windex,\exindex}}_2 \le 1 \;\; (\forall \windex \in [\workers], \forall \exindex \in [\workerexamples])$, the goal of parallel herding is to find $\workers$ permutations, $\perm_1, \perm_2, \ldots, \perm_\workers$ where $\perm_\windex:[\workerexamples]\rightarrow[\workerexamples] \;\; (\forall \windex \in[\workers])$, so as to minimize:
\begin{align}
  \max_{k \in [\workerexamples]} \; \left\| \sum_{\exindex=1}^k \sum_{\windex=1}^\workers \left( \ex_{\windex, \perm_\windex(\exindex)} - \barex \right) \right\|_{\infty}, \hspace{1em}\text{with}\hspace{1em}\barex=\frac{1}{\workers\workerexamples}\sum_{\windex=1}^{\workers}\sum_{\exindex=1}^{\workerexamples}\ex_{\windex,\exindex}.
\label{equ:paraherding:objective}
\end{align}

When directly comparing (\ref{equ:paraherding:objective}) with (\ref{equ:herding:objective}), it is clear that parallel herding differs in two notable ways from the original herding problem. 
First, each permutation $\perm_\windex:[\workerexamples]\rightarrow[\workerexamples] \;\; (\forall \windex \in[\workers])$ only decides the ordering of the $\workerexamples$ vectors that are associated with worker $\windex$. 
Second, the prefix sum taken in the objective norm is accumulated over all the workers (the inner sum from $\windex=1\ldots\workers$). 
This formulation naturally captures the setting in a distributed environment: \textbf{workers need to decide permutations collaboratively, and the worker-specific vectors are processed simultaneously rather than sequentially}.

Given that this formulation fits the distributed setting, we next need to show that parallel herding does in fact address the limitations posed by centralized \grab: 
that it is possible recover the original  $\tilde{O}(1)$  herding bound, and that we can solve the issue of unreliable stale gradients (Section~\ref{sec:cdgrab:dgrab:issues}). 
The solution that we present in the remainder of this section is a new vector balancing subroutine: online $\mathsf{PairBalance}$.  
To give an intuition, as its name suggests, online $\mathsf{PairBalance}$ leverages insights from kernel thinning to \emph{balance} vector differences over vector \emph{pairs}. 
This also eliminates the need to perform vector centering, and thus solves the stale mean problem. 

\paragraph{Using kernel thinning to solve parallel herding.} 
We call our solution to the parallel herding objective (\ref{equ:paraherding:objective}) \emph{pair balancing}, which we derive from key insights in \emph{kernel thinning}~\citep{dwivedi2021kernel,dwivedi2022generalized,barp2022targeted}. 
In particular, Dwivedi and Mackey show that it is possible to solve the herding objective in $\tilde{O}(1)$ \textbf{by only examining differences on  \emph{pairs of examples}}~\citep{dwivedi2021kernel}. 
They derive an algorithm that generalizes the subroutine in Algorithm~\ref{alg:pairbalance} ~\cite{alweiss2021discrepancy}, which solves herding in $\tilde{O}(1)$ (Section~\ref{sec:cdgrab:prelimrw}), and does so by operating only on vector-pair differences.\footnote{Dwivedi and Mackey minimize the maximum mean discrepancy (MMD) between a selected coreset and an empirical distribution~\cite{dwivedi2021kernel}. 
    They  develop a new self-balancing Hilbert walk on differences of \emph{pairs of examples} to select exactly half of the dataset points, and solve coreset selection by iteratively halving the input vector sequence into balanced coresets then selecting and refining a candidate coreset to minimize MMD with the input sequence.\looseness=-1
}  
This comes with a very useful property: 
eliminating the requirement of knowing the maximum vector norm ahead of time and centering the vectors (i.e., making all the vectors sum to zero) in order to solve the herding problem. 
This is the key to solving the parallel herding objective (\ref{equ:paraherding:objective}) in $\tilde{O}(1)$, and elevating the benefits of \grab{} to a distributed setting. 

\begin{figure}[t!]
\begin{algorithm}[H]
\caption{$\mathsf{PairBalance}$ (the server runs this online)}\label{alg:pairbalance}
\begin{algorithmic}[0]
  \Statex $\rhd$ 
  The inputs, outputs and subroutine for this algorithm 
  are order-sensitive\vspace{.2cm}
  \Statex \textbf{input:} current running sum $\vr$, 
  paired vectors $\vz_1$, $\vz_2$\vspace{.2cm}
  \State \textbf{compute:} $\sgn, \vr \leftarrow \mathsf{\small{RandomizedBalance}}(\vr, \vz_1 - \vz_2)$\looseness=-1
  \State \textbf{return:}  $\sgn$ (sign for $\vz_1$), 
  $-\sgn$ (sign for $\vz_2$), 
  $\vr$ (updated running sum) \vspace{.4cm}
  \Statex $\rhd$ Adapted from Alweiss et al.~\cite{alweiss2021discrepancy}
  \State \textbf{define subroutine:} $\mathsf{\small{RandomizedBalance}}(\vr, \vc)$
  \State \hspace{.5cm} \textbf{compute:} $p\leftarrow \frac{1 - \langle \vr,\vc\rangle}{2}$
  \State \hspace{.5cm} \textbf{compute:} $\sgn\leftarrow +1$ \hspace{.2em} with probability \hspace{.2em} $p$; 
  $\sgn\leftarrow -1$ \hspace{.2em} with probability \hspace{.2em} $1-p$
  \State \hspace{.5cm} \textbf{update:}  $\vr\leftarrow \vr+\sgn\vc$
  \State \hspace{.5cm} \textbf{return:}  $\sgn$, $\vr$
\end{algorithmic}
\end{algorithm}
\end{figure}

Following Dwivedi and Mackey~\cite{dwivedi2021kernel}, we will balance over paired vectors, and will do so in an \emph{online} fashion (Section~\ref{sec:cdgrab:dgrab:algo}). 
This eliminates \grab's requirement of using a stale mean to center gradient vectors (Section~\ref{sec:cdgrab:cgrab}), but still minimizes the parallel herding objective to $\tilde{O}(1)$. 
We defer proving this result to Section~\ref{sec:cdgrab:theory}, and first describe our concrete algorithm.  
Online $\mathsf{PairBalance}$ applies Algorithm~\ref{alg:reorder} on the``flattened'' and ``paired'' sequence of all of the workers' paired-difference gradients, i.e., \looseness=-1
\begin{align*}
    \vy_{\workerexamples(k-1)+\windex} = \ex_{\windex,2k-1} - \ex_{\windex,2k}, \hspace{1em} \forall k\in[\frac{\workerexamples}{2}], \hspace{1em} \windex=1\ldots\workers.
\end{align*}

\noindent That is, we fit these ordered-paired differences $\{\vy_\windex\}_{\windex=1}^{\workers\workerexamples/2}$ into the herding and balancing framework  (Algorithm~\ref{alg:reorder}): 
if sign $\sgn$ is associated with $\vy_{\workerexamples(k-1)+\windex}$, then $\ex_{\windex,2k-1}$ and $\ex_{\windex,2k}$ receive 
$\sgn$ and $-\sgn$, respectively.\looseness=-1 

\subsection{The full-stack \dgrab{} algorithm}\label{sec:cdgrab:dgrab:algo}

Having solved the parallel herding problem with pair balancing, we now demonstrate how to bring everything together in an optimization context to \emph{coordinate distributed gradient balancing} for distributed training. 
That is, we can now introduce our full-stack \dgrab{} algorithm, which trains models in a distributed setting (Section~\ref{sec:cdgrab:dgrab:issues}) while efficiently ordering the examples by using $\mathsf{PairBalance}$ (Section~\ref{sec:cdgrab:dgrab:solution}, Algorithm~\ref{alg:pairbalance}) 
in an online manner. 

We describe \dgrab{} at two levels of abstraction: 
a high-level illustration (Figure~\ref{fig:cdgrab:diagram}, steps \dstep{1-7}) and a detailed pair of worker-server algorithm statements (Figure~\ref{alg:dgrab}). 
Since the workers only have access to a subset of the training data, in parallel they compute local, per-example stochastic gradients and send them to the server. 
The server simultaneously calls $\mathsf{PairBalance}$ online (Algorithm~\ref{alg:pairbalance}), which coordinates information from all the workers' gradients (i.e., using adjacent example-specific gradients) to determine the next epoch's worker-specific permutations. 
In more detail:

\begin{figure}[t!]
\centering
\includegraphics[width=.7\linewidth]{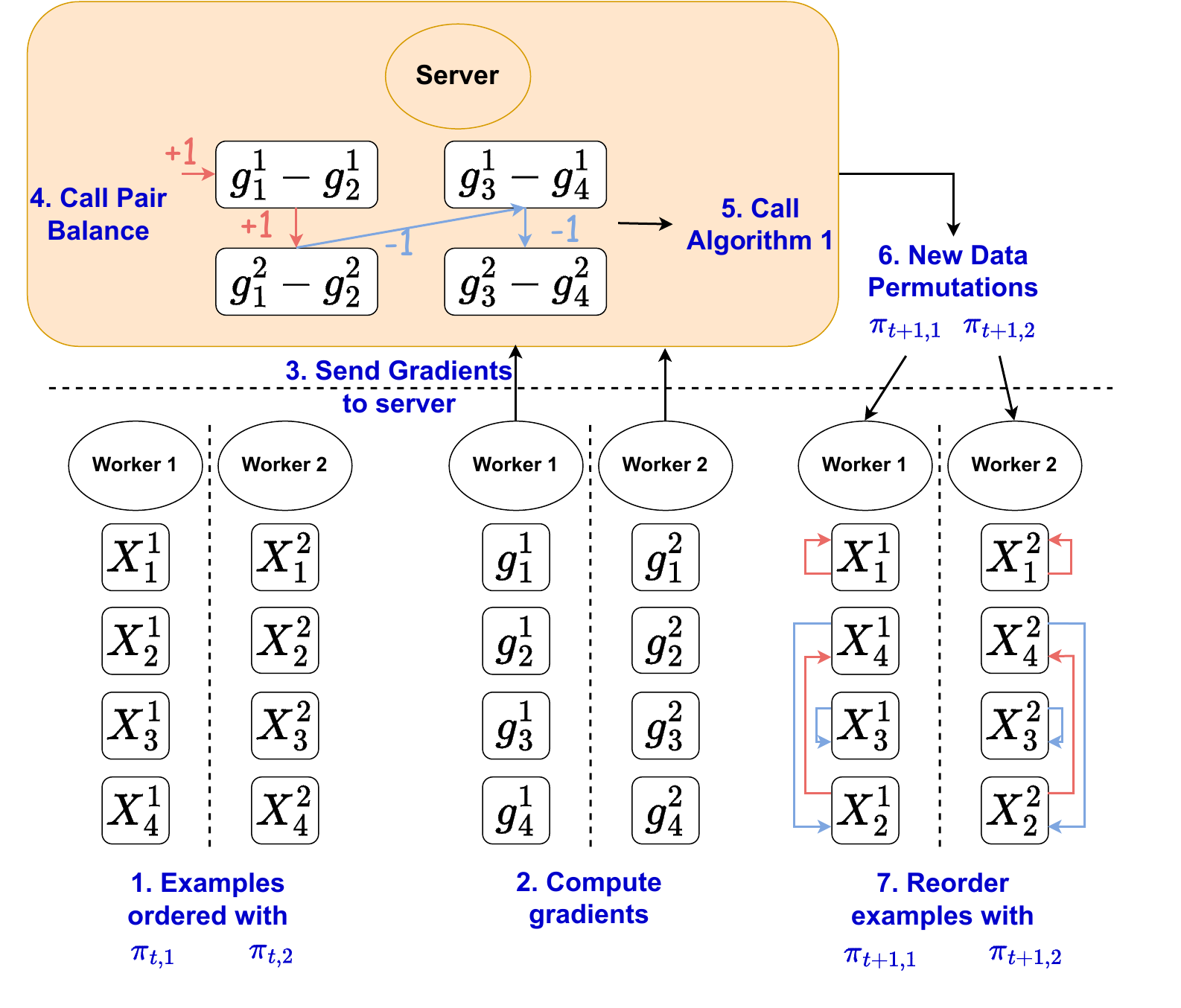}
\caption{\dgrab{} running on one server (top) and two workers (bottom). 
The workers do not share data examples. 
The server calls $\mathsf{PairBalance}$  (Algorithm~\ref{alg:pairbalance}) online.}
\label{fig:cdgrab:diagram}
\end{figure}

In epoch $\eindex$, (Figure~\ref{fig:cdgrab:diagram},  step \dstep{1}) the two workers have permutations $\perm_{\eindex, 1}$ and $\perm_{\eindex,2}$, respectively. 
Each worker computes per-example gradients $\g_\exindex^\windex$ (\dstep{2}; 
Algorithm~\ref{alg:dgrab:workers}:4), and sends them to the server (\dstep{3}; 
Algorithm~\ref{alg:dgrab:workers}:5). 
The server we implement functions as a parameter server~\cite{li2014ps}: 
it computes the average of the workers' per-example gradients (Algorithm~\ref{alg:dgrab:server}:6), and sends it back to all workers (Algorithm~\ref{alg:dgrab:server}:7) so that they can update their local models (Algorithm~\ref{alg:dgrab:workers}:6-7). 
Simultaneously, as the server receives gradients (Algorithm~\ref{alg:dgrab:server}:5), it calls  $\mathsf{PairBalance}$ (Algorithm~\ref{alg:pairbalance}) on adjacent vectors (\dstep{4}; Algorithm~\ref{alg:dgrab:server}:4-13). 
$\mathsf{PairBalance}$ produces signs to supply to the reordering algorithm  (Algorithm~\ref{alg:reorder}), which, using the current worker permutations $\perm_{\eindex,\windex}$, produces the new per-worker permutations for the next epoch (\dstep{5};  Algorithm~\ref{alg:dgrab:server}:14). 
In Figure~\ref{fig:cdgrab:diagram}, these correspond to $\perm_{\eindex+1, 1}$ and $\perm_{\eindex+1, 2}$, which the server then sends back to the respective workers (\dstep{6};  Algorithm~\ref{alg:dgrab:server}:15). 
Lastly, before the start of the next epoch, the workers reorder their examples according to the new permutations (\dstep{7};  Algorithm~\ref{alg:dgrab:workers}:9).

\begin{figure}[t]
\hspace{-.1cm}
\begin{minipage}[t]{.47\linewidth}
\begin{algorithm}[H]
\caption{\dgrab{} Workers}\label{alg:dgrab:workers}
{\scriptsize
\begin{algorithmic}[1]
    \Statex \textbf{require:} $\workers$ workers, $\workerexamples \coloneqq \frac{\examples}{\workers}$ ex. per worker
    \Statex \textbf{input:} initial $\weights_1^1$, epochs $\epochs$, learning rate $\alpha$
    \Statex
    \State \textbf{receive:} initial permutations 
    \For{epoch $\eindex \coloneqq 1 \ldots \epochs$}
        \Statex \hspace{.5cm}$\rhd$ Run in parallel for workers $\windex=1 \ldots \workers$
            \For{example $j \coloneqq 1 \ldots \workerexamples$}
            \State \textbf{compute:} $\wexgrad \leftarrow \nabla\loss^\windex(\weights_\eindex^\exindex, \perm_{\eindex, \windex}(\exindex))$ \vspace{.05cm}
            \Statex \hspace{.36\linewidth}$\rarrow{0.61\textwidth}{\text{$j$-th stochastic grad. }\wexgrad}$ \vspace{-.8cm}
            \State \textbf{send:} $\wexgrad$
            \vspace{.5cm}
            \Statex \hspace{.42\linewidth}$\larrow{0.56\textwidth}{\text{avg. $\exindex$-th stochastic grad. } {\bar{\vg}_\exindex}}$ \vspace{-.8cm}
            \State \textbf{receive:} $\bar{\vg}_\exindex$ \vspace{.5cm}
            \State \textbf{update:} $\weights_\eindex^{\exindex + 1} \leftarrow \weights_\eindex^\exindex - \alpha \bar{\vg}_\exindex$
        \EndFor
        \Statex \vspace{1.65cm}
        \State \textbf{receive:} next permutation 
        \State \hspace{-.1cm}\textbf{update:} $\weights_{\eindex+1}^1 \coloneqq \weights_{\eindex}^{\workerexamples + 1}$
  \EndFor
  \State \textbf{return:} $\weights_{\epochs+1} \coloneqq \weights_{\epochs+1}^1$
\end{algorithmic}
}
\end{algorithm}
\end{minipage}
\hfill
\begin{minipage}[t]{.5\linewidth}
\begin{algorithm}[H]
\caption{\dgrab{} PS}\label{alg:dgrab:server}
{\scriptsize
\begin{algorithmic}[1]
    \Statex \textbf{require:} $\workers$ workers, $\workerexamples \coloneqq \frac{\examples}{\workers}$ ex. per worker
    \Statex \textbf{input:} epochs $\epochs$ 
    \Statex \vspace{.1cm}
    \Statex \vspace{-.2cm}\hspace{-.52\linewidth}$\larrow{0.38\textwidth}{\{\perm_{1,\windex}\}_{\windex=1}^\workers}$ 
    \vspace{-.4cm}
    \State \textbf{send:} initial permutations $\{\perm_{1,\windex}\}_{\windex=1}^\workers$
  \For{epoch $\eindex \coloneqq 1 \ldots \epochs$} 
        \State \textbf{initialize:} running sum $\vh=\bm{0}$; empty list $\mathcal{S}$ 
        \For{ example $j \coloneqq 1 \ldots \workerexamples$}\vspace{.78cm}
            \State \textbf{receive:} $\{\wexgrad\}_{\windex=1}^\workers$ from all workers $\windex$
            \State \textbf{compute:} avg. gradient: $\bar{\vg}_\exindex\leftarrow\frac{1}{\workers}\sum_{\windex=1}^\workers \wexgrad$ \vspace{.4cm}
            \State \textbf{send:} $\bar{\vg}_\exindex$ to all the workers
            \For{worker $\windex \coloneqq 1 \ldots \workers$}
                \State \hspace{-.2cm} \textbf{if } $\exindex\bmod 2 = 0$: 
                \State $\vh,\wexsignprev, \wexsign \leftarrow \mathsf{PairBalance}(\vh, \wexgradprev, \wexgrad)$
                \State $\mathcal{S}.\textsf{\scriptsize{append}}(\wexsignprev)$; $\mathcal{S}.\textsf{\scriptsize{append}}(\wexsign)$
            \EndFor
        \EndFor
        \Statex \hspace{.5cm}$\rhd$ Call Alg.~\ref{alg:reorder} for $\windex=1\ldots \workers$ on  $\perm_{\eindex,\windex} \text{ and } \mathcal{S}$
        \looseness=-1
        \State \textbf{compute:} next permutations $\{\perm_{\eindex + 1,\windex}\}_{\windex=1}^\workers$\looseness=-1 
        \Statex  \hspace{-.48\linewidth}$\larrow{0.31\textwidth}{\perm_{\eindex + 1,\windex}}$ \vspace{-.4cm}
        \State \textbf{send:} $\{\perm_{\eindex + 1,\windex}\}_{\windex=1}^\workers$ to each worker $\windex$
  \EndFor
\end{algorithmic}
}
\end{algorithm}
\end{minipage}
\caption{\dgrab{} worker and server (here, a parameter server~\citep{li2014ps}) algorithms.} 
\label{alg:dgrab}
\end{figure}

%% file: section/30-algorithms/32-cdgrab/324-cdgrab-theory.tex
\section{Convergence Analysis}\label{sec:cdgrab:theory}

We next demonstrate formally that our \dgrab{} algorithm (Section~\ref{sec:cdgrab:dgrab:algo}) confers the efficiency benefits of centralized \grab{}  (Section~\ref{sec:cdgrab:cgrab}) to the distributed setting.
In brief, our main theoretical results show that \textbf{\dgrab{} enjoys a linear speedup in convergence rate} under two sets of conditions: 
smoothness (Theorem~\ref{thm:dgrab:smooth}) and the Polyak-\L ojasiewicz (P.L.) condition (Theorem~\ref{thm:dgrab:PL}). 
\textbf{Both results guarantee that \dgrab{} is faster than distributed random reshuffling (\dshuffle)}. 
Our proofs rely on Corollary 7 from Dwivedi and Mackey~\cite{dwivedi2021kernel}, which shows that, with high probability, $\mathsf{RandomizedBalance}$ (subroutine in Algorithm~\ref{alg:pairbalance}, from Alweiss et al.~\cite{alweiss2021discrepancy}) guarantees a $\tilde{O}(1)$ bound to the signed herding objective (\ref{equ:herding:signed_objective}).\footnote{Corollary 7 from Dwivedi and Mackey~\cite{dwivedi2021kernel} improves the result of Theorem 1.1 from Alweiss et al.~\cite{alweiss2021discrepancy}.
} 

To begin, we restate this result to cohere with our framework, for which the vectors $\exj$ are gradients in an optimization context: 

\begin{theorem}[\textbf{Corollary 7, Dwivedi and Mackey~\cite{dwivedi2021kernel}}]
\label{statement:alweiss}
    Consider any vectors $\{\vz_\exindex\}_{\exindex=1}^\examples$ ($\exj \in \R^d$) with $\norm{\vz_\exindex}_2 \le 1$ supplied as input to the $\mathsf{RandomizedBalance}$ subroutine in Algorithm~\ref{alg:pairbalance}. 
    Then for any $\delta > 0$, with probability at least $1 - \delta$, $\mathsf{RandomizedBalance}$ outputs a sequence of signs $\{s_\exindex\}_{\exindex=1}^\examples\in \{-1,1\}$ that satisfy $\textstyle \max_{k\in[\examples]}\norm{\sum\nolimits_{\exindex=1}^k s_\exindex\vz_\exindex}_{\infty} \le \tilde{A}$, where 
    $\tilde{A}=\sqrt{2\log(\frac{4d}{\delta})\log(\frac{4N}{\delta})}=\tilde{O}(1)$.
\end{theorem}

To integrate this result with our parallel setting, we need some additional assumptions that are standard in the literature on distributed optimization --- that the variance of the per-example gradients on each worker is uniformly bounded (Assumption~\ref{ass:inner-deviation}), and that the  variance between worker-specific gradients is similarly bounded (Assumption~\ref{ass:outer-deviation}). 
More precisely, following the distributed setup in (\ref{equ:d-grab:main_update}), we denote the  global loss gradient to be $\nabla \loss(\weights)$, each $\windex$-th worker's local loss gradient to be $\nabla \loss^\windex(\weights)$ ($\forall \windex \in [\workers]$), and each $\windex$-th worker's per-example loss gradients to be $\nabla \loss^\windex(\weights; \exindex)$ ($\forall \exindex \in [\workerexamples]$). We assume: 

\begin{assumption}[\textbf{Bounded Grad. Var.}]
\label{ass:inner-deviation}
For all $\windex \in [\workers]$ there exists a constant $\sigma > 0$ such that for all $\exindex \in [\workerexamples]$ and for all $\weights \in \R^d$, it holds that $\norm{\nabla \loss^\windex(\weights; \exindex) - \nabla \loss^\windex(\weights)}_2^2 \le \sigma^2$.
\end{assumption}

\begin{assumption}[\textbf{Bounded Data Heterogeneity}]
\label{ass:outer-deviation}
There exists a constant $\varsigma > 0$ such that $\forall \windex \in[\workers]$,
$\norm{\nabla \loss^\windex(\weights) - \nabla \loss(\weights)}_2^2 \le \varsigma^2.$
\end{assumption}

Lastly, we include one additional assumption from the 
original \grab{} paper~\citep{lu2022grab}: we assume a cross norm $L_{2,\infty}$ (which can be easily adapted to 
$L_2$-smoothness by setting $L_{2,\infty}$ to be $\sqrt{d}L_2$).

\begin{assumption}[\textbf{Smoothness}]
\label{ass:smoothness} 
There exists constant $L_{2,\infty}>0 \text{ such that for any }\vw,\vv\in\mathbb{R}^d$, any $\windex \in [\workers]$, and any $\exindex \in [\workerexamples]$, it holds that $\norm{\nabla f^i(\vw; j) - \nabla f^i(\vv; j)}_2 \le L_{2,\infty}\|\vw - \vv\|_\infty$.
\end{assumption}

Given these assumptions, we can prove a convergence guarantee for \dgrab:  

\begin{theorem}
\label{thm:dgrab:smooth}
Suppose that Assumptions~\ref{ass:inner-deviation},\ref{ass:outer-deviation} and \ref{ass:smoothness} hold. 
For any $\delta > 0$, if we set learning rate $\alpha$ to be
\begin{align*}
    \alpha = \min\left\{\frac{1}{16 L_{2,\infty} (2\workerexamples + \tilde{A}/\workers)}, \left(\frac{4 F_1 \workers^2}{42 L_{2,\infty}^2 (\varsigma + \sigma)^2\tilde{A}^2 \workerexamples \epochs + 18L_{2,\infty}^2  \workers^2\workerexamples^3 \sigma^2}\right)^{1/3}\right\}, 
\end{align*}
where $F_1=f(\weights_1) - \inf_{\weights\in\mathbb{R}^d}f(\weights)$ and $\tilde A$ comes from Theorem~\ref{statement:alweiss}. Then, with probability at least $1 - T\delta$, 
\begin{align*}
\frac{1}{\epochs}\sum_{t=1}^{\epochs} \norm{\nabla f(\vw_t)}_2^2 &\le \frac{9 (F_1 L_{2,\infty}(\varsigma + \sigma)\tilde{A})^{2/3}}{(\workers \workerexamples \epochs)^{2/3}} + \frac{(72 F_1 L_{2,\infty}\sigma)^{2/3} + 64F_1 L_{2,\infty} (2 + \tilde{A}/(\workers \workerexamples))}{\epochs}\\
&= \tilde{O}\left(\frac{1}{(mnT)^{2/3}} + \frac{1}{\epochs}\right). 
\end{align*} 
\end{theorem}

We can also prove an accelerated rate for \dgrab{} if we additionally assume the P.L. condition:
\begin{assumption}[\textbf{P.L. Condition}]
\label{ass:PL}
We say the loss function $f$ fulfills the P.L. condition if there exists $\mu>0$ such that for any $\weights\in\R^d$, $\frac{1}{2}\|\nabla f(\vw)\|_2^2 \geq \mu(f(\vw) - \inf_{\vv\in\mathbb{R}^d}f(\vv)).$
\end{assumption}

\begin{theorem}
\label{thm:dgrab:PL}
Suppose that Assumptions~\ref{ass:inner-deviation}, ~\ref{ass:outer-deviation},~\ref{ass:smoothness}, and \ref{ass:PL} hold. 
For any $\delta > 0$, we set constants $\tilde W$ and $C_3$ to be 
\[
    C_3 = \frac{(F_1+\sigma^2/L_{2,\infty})\mu^2}{224L_{2,\infty}^2(\varsigma + \sigma)^2\tilde{A}^2}
    \hspace{1em} \text{ and } \hspace{1em}
    \tilde{W} = W_0(T^2\workers^2\workerexamples^2C_3),
\]
where $\tilde A$ comes from Theorem~\ref{statement:alweiss}, $F_1$ is from Theorem~\ref{thm:dgrab:smooth}, and $W_0$ is the Lambert-W function. 
If we set learning rate $\alpha = \frac{2\tilde{W}}{T\workerexamples\mu}$ and if the number of epochs $T$ satisfies
\[
\epochs \ge 10 + \frac{1}{\mu}32 L_{2,\infty}(2+\tilde{A}/(\workers \workerexamples))W_0((\workers \workerexamples \epochs)^2C_3) = \tilde O(1),
\]
then, with probability at least $1 - T\delta$, 
\text{ it holds that}
\begin{align*}
F_{\epochs+1} &\le \frac{1}{(\workers \workerexamples \epochs)^2}\left(\frac{(F_1 + L_{2,\infty}^2\sigma^2)\tilde{W}}{C_3} + \frac{112L_{2,\infty}^2(\varsigma + \sigma)^2\tilde{A}^2{\tilde{W}}^2}{\mu^3}\right)
= \tilde{O}\left(\frac{1}{(\workers\workerexamples \epochs)^{2}}\right),
\end{align*}
where $F_{\epochs+1}=f(\weights_{\epochs+1}) - \inf_{\weights\in\mathbb{R}^d}f(\weights)$.
\end{theorem}

We prove Theorems~\ref{thm:dgrab:smooth} and~\ref{thm:dgrab:PL} in the Appendix (Chapter~\ref{chapter:app:cdgrab}). 
Together, they show that \dgrab{} exhibits a linear speedup in the number of workers $\workers$ over \grab~\citep{lu2022grab}'s convergence rates ($\tilde{O}((\workerexamples\epochs)^{-2/3})$ and $\tilde{O}((\workerexamples\epochs)^{-2})$, respectively).\footnote{For centralized \grab, the total number of examples $\examples=\workerexamples$ and $\workers=1$.} 
under both smoothness and the P.L. condition. 
Further, \dgrab's convergence rate of $\tilde{O}((\workers \workerexamples \epochs)^{-2})$ is faster than many previous rates,\footnote{These exclusively focus on the P.L. case, so we compare \dgrab{} to them under the same condition.\looseness=-1} 
such as the high probability bound of $\tilde{O}((\workers \workerexamples)^{-1}T^{-2})$ for \dshuffle{} in Yun et al.~\cite{yun2021minibatch}. 

%% file: section/30-algorithms/32-cdgrab/325-cdgrab-experiments.tex
\section{\dgrab{} in Practice: Distributed and Simulation Experiments}\label{sec:cdgrab:experiments}

We next verify \dgrab's accelerated convergence on a variety of empirical tasks.\footnote{Our GitHub repository is \href{https://github.com/GarlGuo/CD-GraB}{https://github.com/GarlGuo/CD-GraB}.} 
For ease of comparison, we follow the experimental plan from the original \grab{} paper,\footnote{Following Lu et al.\cite{lu2022grab}, for our LSTM experiment on  WikiText-2, we set the embedding dimension to 32. 
    We note that we can improve perplexity if we set the dimension higher.} 
and add some additional large-scale logistic regression experiments. 
We also run an ablation study to isolate the effects of different improvements in \dgrab. 
We do this because online $\mathsf{PairBalance}$ exhibits performance benefits that are separate from parallelism --- namely, removing the need for gradient centering with a stale mean and allowing for higher learning rates (Section~\ref{sec:cdgrab:dgrab:solution}).\footnote{\grab{} can also implement online $\mathsf{PairBalance}$, in place of $\mathsf{Balance}$~\citep{lu2021general} (Appendix).\looseness=-1} 

\begin{figure*}[t]
  \centering
  \begin{minipage}{\linewidth}
      \hspace{-.5cm}
    \includegraphics[width=\columnwidth]{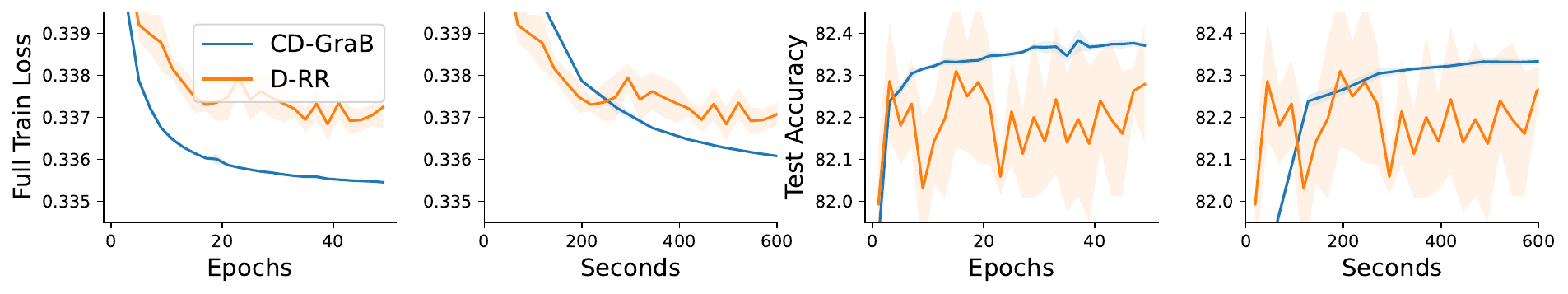}
      \vspace{-.15cm}
      \subcaption{Logistic regression on mortgage application (NY 2017 subset)~\citep{cooper2024variance}}
      \label{fig:exp:ny}
  \end{minipage}
  \begin{minipage}{\linewidth}
    \hspace{-.5cm}
    \includegraphics[width=\columnwidth]{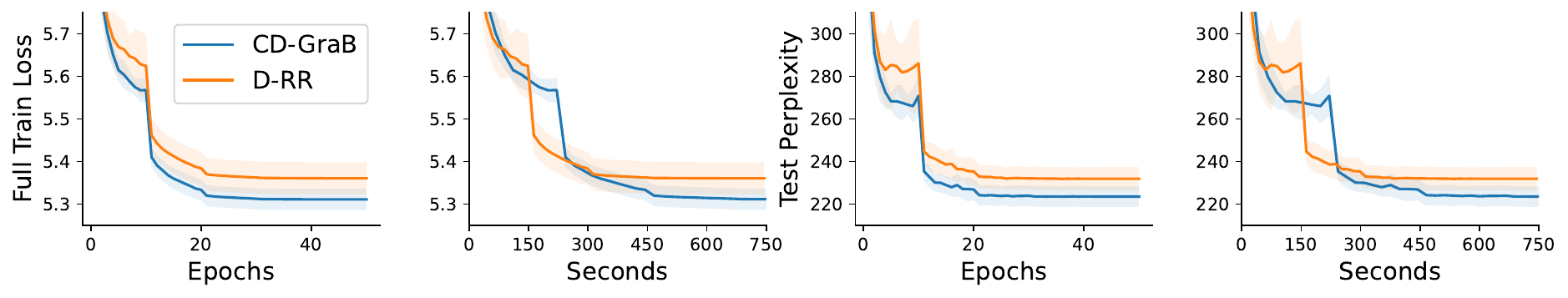}
      \vspace{-.15cm}
      \subcaption{LSTM~\citep{hochreiter1997long} on WikiText-2~\citep{merity2017regularizing}}
      \label{fig:exp:wiki}
    \end{minipage}
    \begin{minipage}{\linewidth}
        \hspace{-.5cm}
    \includegraphics[width=\columnwidth]{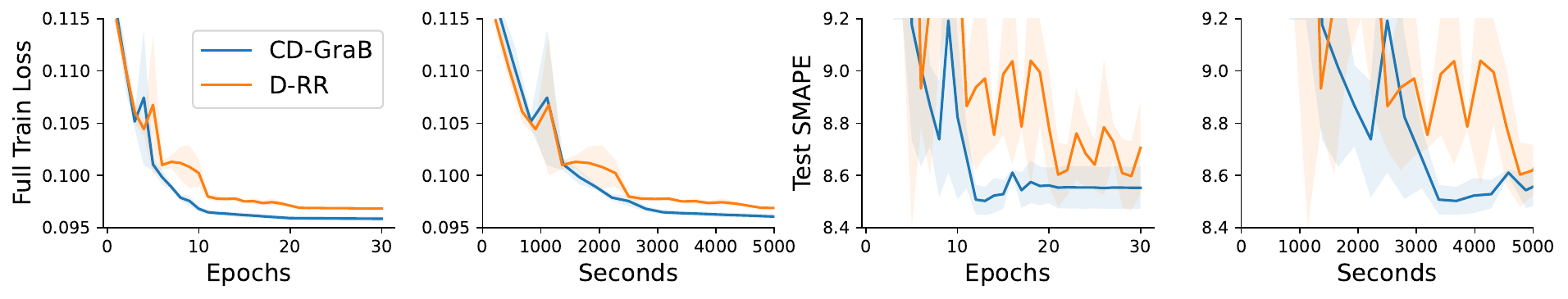}
      \vspace{-.15cm}
      \subcaption{Autoregressive MLP on M4 Weekly~\citep{MAKRIDAKIS202054}}
      \label{fig:exp:m4}
  \end{minipage}
  \caption{Convergence of \dgrab{} in comparison to \dshuffle. For each experiment, we show train loss over epochs and time (\textbf{left} of each subfigure) and test performance over epochs and time (\textbf{right} of each subfigure). We run at least 3 random seeds, and plot the mean $\pm$ STD.} 
  \label{fig:exp}
  \vspace{-.6cm}
\end{figure*}

\paragraph{Evaluating \dgrab's convergence speedup.} 
We use the following three tasks for evaluating distributed training efficiency: logistic regression on a large-scale mortgage application (New York 2017 subset, 244,107 examples with 18 features)~\citep{cooper2024variance} (Figure~\ref{fig:exp:ny}), Long Short-Term Memory (LSTM)~\citep{hochreiter1997long} on the WikiText-2 dataset~\citep{merity2017regularizing} (Figure~\ref{fig:exp:wiki}), and autoregressive Multi-Layer Perceptron (MLP) on the M4 Weekly dataset~\citep{MAKRIDAKIS202054} (Figure~\ref{fig:exp:m4}). 
We measure the loss incurred on the entire training set (Full Train Loss) and task-appropriate test metrics during evaluation, with respect to both the number of epochs and wall-clock time. 
Regarding test metrics, we measure test accuracy for the mortgage application, perplexity for WikiText-2, and SMAPE for M4. 
Additional details regarding the datasets, models, and test metrics can be found in the Appendix (Chapter~\ref{chapter:app:cdgrab}). 

For all three tasks, we use a single 128 GiB memory machine with 4 NVIDIA GeForce RTX 2080 Ti GPUs. 
For the mortgage application and WikiText-2 (Figures~\ref{fig:exp:ny} and~\ref{fig:exp:wiki}), we launch $\workers=4$ workers (processes), where each worker runs on one GPU. 
For the M4 task, we launch $\workers=32$ workers, where each of the 4 GPUs hosts 8 process workers. 
We use NCCL as the distributed communication backend~\cite{nccl} for the mortgage application and WikiText-2 tasks, and GLOO~\cite{gloo} as the distributed communication backend for the M4 task. 

As shown in Figure~\ref{fig:exp}, we compare \dgrab{}'s convergence to the standard distributed-training example-ordering method: 
random reshuffling (D-RR). 
From all  subfigures in Figure~\ref{fig:exp}, we observe that \dgrab{} outperforms the \dshuffle{} baseline significantly and consistently: 
\dgrab{} exhibits better training loss and test metrics, measured against both the number of epochs and wall-clock time.
We also note that the results for \dgrab{} are much smoother than for \dshuffle. 
This is likely due to the variance of stochastic gradients during training, which \dgrab{} reduces as a side-effect (so, too, does \grab, in comparison to RR). 
For smoother \dshuffle{} results, we can reduce the learning rate (Appendix, Chapter~\ref{chapter:app:cdgrab}). 
\dgrab{} allows for the use of a larger learning rate, which accelerates training while preserving the final model's performance. 

\paragraph{Ablation simulation study: the importance of coordination at large scale.} 
\dgrab{} has several design benefits over the original centralized \grab{} algorithm~\citep{lu2022grab}: 
coordinating parallel workers' specific permutations using $\mathsf{PairBalance}$ on the server (Algorithm~\ref{alg:dgrab}) and removing the dependency on a stale mean (Section~\ref{sec:cdgrab:cgrab}), which enables the ability to using larger learning rates reliably (Section~\ref{sec:cdgrab:dgrab:solution}). 
Clearly, not all of these benefits come directly from distributing training. 
For example, being able to use larger learning rates, is a side effect of our solution to develop \dgrab, not our main contribution. 
Therefore, we run a simulation ablation study to disentangle the relative importance of each of \dgrab's efficiency benefits over \grab. 
To do so, we compare the convergence of \dgrab{} to two additional baselines in the distributed setting, beyond \dshuffle: 
(1) \textbf{ID-\grab{} (Bal)}, where each independent worker runs \grab{} locally using $\mathsf{RandomizedBalance}$ (subroutine in Algorithm~\ref{alg:pairbalance}) to perform gradient vector balancing; (2) \textbf{ID-\grab{} (PairBal)}, where each independent worker runs \grab{} locally using $\mathsf{PairBalance}$.

\begin{figure}[!t]
  \centering
    \includegraphics[width=\columnwidth]{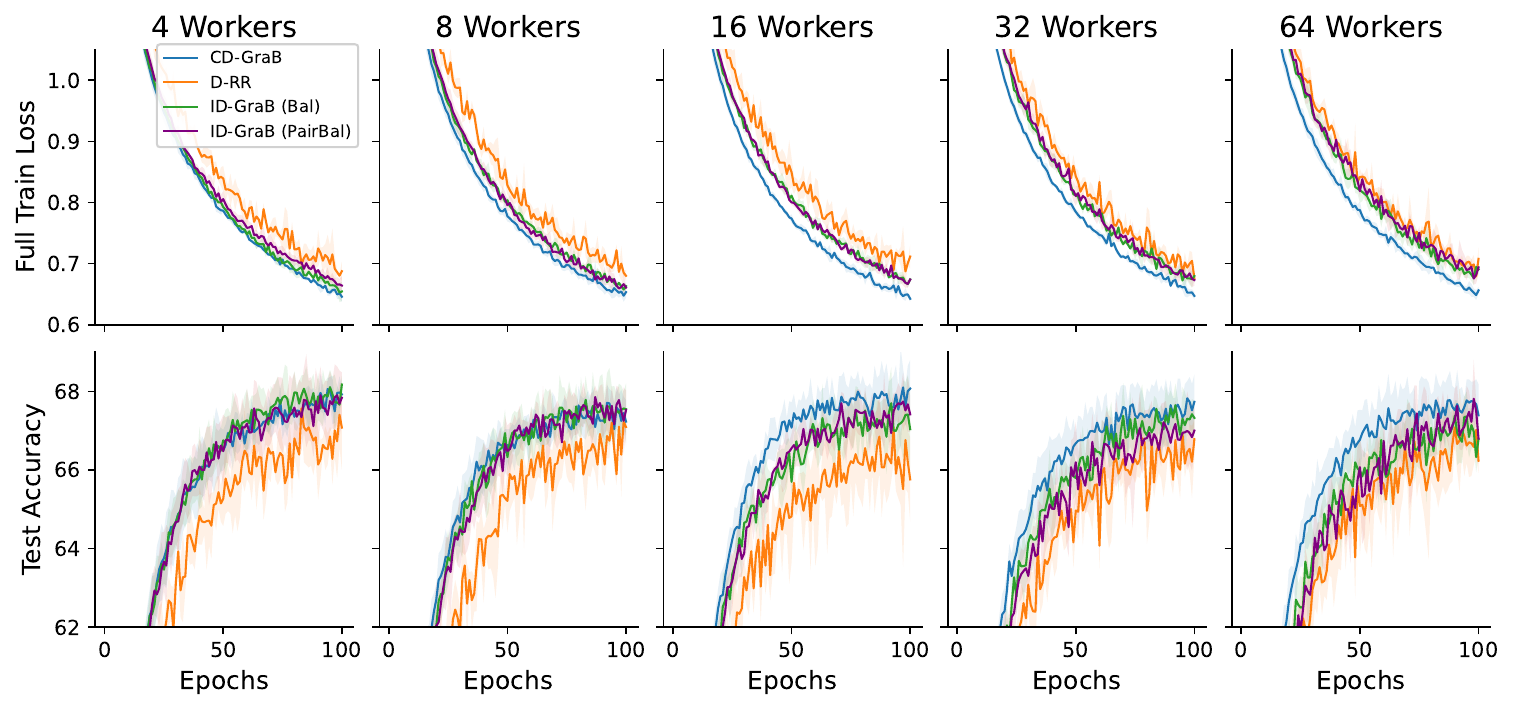}
    \caption{Convergence for \dgrab, \dshuffle, ID-\grab{} (Bal), and ID-\grab{} (PairBal) training LeNet on CIFAR-10, with $\workers \in \{4, 8, 16, 32, 64\}$ workers. For each experiment, the aggregated minibatch size per update is 64.\looseness=-1} 
  \label{fig:nodes}
\end{figure}

Figure~\ref{fig:nodes} summarizes the results, with  convergence curves  for $\workers\in\{4,8,16,32,64\}$ workers training LeNet on CIFAR-10. 
We choose this task and architecture to cohere with the experiments done in the original \grab{} paper. 
For these experiments, we denote $B$ to be the \emph{aggregated} minibatch across all the workers, which refers to the number of stochastic examples used for an overall optimization step; 
each worker thus has a subset of this minibatch --- an equivalently-sized subset of $B$ examples.\footnote{For example, if we have 4 workers with an aggregated minibatch size of 32, each worker would compute their respective local gradients with 8 examples, and then all-reduce these gradients to obtain the aggregated minibatch gradient for all 32 examples for the optimization step. 
    We discard $\examples \bmod B$  examples at random to ensure $\workerexamples$ examples per worker.} 
We make two main observations. 
First, when scaling up training with more workers, \dgrab{} converges increasingly faster than the no-coordination-ordering methods \textbf{ID-\grab{} (Bal)} and \textbf{ID-\grab{} (PairBal)}. 
This result aligns with our theory and intuition that, when the number of workers $\workers$ increases, the parallel herding bound (\ref{equ:paraherding:objective}) will increase linearly if there is no coordination. 
Second, as we scale up to larger $\workers$, the convergence curves of \textbf{ID-\grab{} (Bal)} and \textbf{ID-\grab{} (PairBal)} gradually approach the curve for \dshuffle: 
at larger scales, herding-based example ordering will be no better than  randomly permuting the dataset. 
Both observations give strong evidence that coordination  (i.e., running online $\mathsf{PairBalance}$ on the server to coordinate per-worker permutations) is critical for accelerating training.

We note that all of these experiments use SGD, since both the theoretical results of the original \grab{} paper and our results for \dgrab{} here are for SGD. 
In the Appendix (Chapter~\ref{chapter:app:cdgrab}), we additionally include results for training GPT-2 on WikiText-103, for which we use AdamW as the optimizer. We find that \cgrab{} with AdamW works in practice; however, our theory results do not directly apply to these experiments. 
We additionally include results on memory usage in the Appendix, which show that \dgrab{} results in negligible overhead in practice. 

%% file: section/30-algorithms/32-cdgrab/326-cdgrab-conclusion.tex
\section{Conclusion and Future Work: Toward an Order Server Architecture}\label{sec:cdgrab:conclusion}

We elevate the benefits of provably faster, permutation-based example ordering to the contemporary ML distributed-training setting. 
We focus on reformulating the online \textbf{Gra}dient \textbf{B}alancing algorithm (\grab)~\citep{lu2022grab} because, even though it is the provably optimal permutation-based example-ordering method~\citep{cha2023tighter}, it is limited by design to \emph{centralized} settings (Section~\ref{sec:cdgrab:dgrab:issues}). 
To overcome these limitations, we redesign \grab's herding and balancing framework to account for parallel workers: 
A \emph{parallel herding} objective, which we solve with an online $\mathsf{PairBalance}$ subroutine, based on key insights from kernel thinning~\citep{dwivedi2021kernel, dwivedi2022generalized, barp2022targeted}. 
$\mathsf{PairBalance}$ operates on ordered \emph{pairs} of vectors to do \emph{balancing}, which enables our full-stack, low-overhead, \emph{\textbf{C}oordinated} and \emph{\textbf{D}istributed} online \dgrab{} algorithm. 
We give a full specification of our online \dgrab{} algorithm (Section~\ref{sec:cdgrab:dgrab:algo}), provide convergence rate guarantees regarding its speedups on both 1) smooth non-convex and 2) P.L. objectives (Section~\ref{sec:cdgrab:theory}), and verify these speedups in practice on single-node distributed tasks and a simulated ablation study (Section~\ref{sec:cdgrab:experiments}). 

Both our theory and experiments demonstrate that \dgrab{} really shines when there are multiple training epochs (Appendix). 
This is another reason that we do not emphasize experiments involving fine-tuning pre-trained models like GPT-2, as fine-tuning can be achieved in just a couple of epochs. As noted above, it is also more common to train such models using optimizers from the Adam family. 
In future work, we intend to extend the theory on \grab{} and \dgrab{} to such optimizers, which would make the results on optimal, permutation-based example ordering more useful for base-model pre-training. 

Pre-training from scratch would demonstrate the tremendous power of \dgrab{} to scale to very large models; 
however, we did not have the training budget to perform such  experiments for the present work. 
Further, to truly exercise the benefits of \dgrab{} in such large-scale settings, future work should investigate moving beyond the single-node setup that we present. 
Notably, to train larger models, our results suggest a novel distributed training architecture. 
The ordering operation performed by the server (Algorithm~\ref{alg:dgrab:server}) is \emph{not} very latency sensitive; the server has the duration of the entire epoch $\eindex$ to compute the new permutations for the next, $\eindex + 1$ epoch. 
Given this relaxed latency requirement, and the success of our algorithmic results, it would be an exciting direction for future ML-systems research to invest in building an \emph{Order Server} architecture. 
Such an architecture, which could be composed with traditional parameter servers, would afford the scalability benefits of \dgrab{} to a host of massive-scale ML applications.

%% file: section/30-algorithms/33-tradeoffs/33-tradeoffs-main.tex
\chapter{Accuracy-Efficiency Trade-Offs and Accountability}\label{chapter:tradeoffs}

Machine-learning algorithms that attempt to afford greater scalability and efficiency tend to trade-off these gains for a reduction in (overall or per-iteration) accuracy. 
Indeed, such accuracy-efficiency trade-offs (and how to wrangle them) lay at the heart of scalable machine learning. 
In this chapter, we connect our algorithmic work from scalable machine learning, presented in the two prior chapters, to insights in law and policy.
We explore how the accuracy-efficiency trade-offs inherent to scalable machine learning also provide a natural level of abstraction for reasoning about law and policy implications.

This work, initially published before the advent of ``Generative AI,'' has aged surprisingly well.
The motivating example of a distributed ML system in this text is an autonomous vehicle (AV) --- IoT and AVs were (at the time of writing) the expected application space where our observations would be relevant.
While this has not come to pass (there is not yet widespread deployment of AV systems), we have seen other large-scale, distributed ML systems become household topics, e.g., ChatGPT.
The observations and arguments in this paper are largely applicable to such systems.
We defer this unification to future work.\\ 

\noindent\textbf{Chapter summary}: 
Trade-offs between accuracy and efficiency pervade law, public health, and other non-computing domains, which have developed policies to guide how to balance the two in conditions of uncertainty. 
While computer science also commonly studies accuracy-efficiency trade-offs, their policy implications remain poorly examined. 
Drawing on risk assessment practices in the US, we argue that, since examining these trade-offs has been useful for guiding governance in other domains, we need to similarly reckon with these trade-offs in governing computer systems. 
We focus our analysis on distributed machine learning systems.  
Understanding the policy implications in this area is particularly urgent because such systems, which include autonomous vehicles, tend to be high-stakes and safety-critical. 
We 1) describe how the trade-off takes shape for these systems, 
2) highlight gaps between existing US risk assessment standards and what these systems require to be properly assessed, 
and 3) make specific calls to action to facilitate accountability when hypothetical risks concerning the accuracy-efficiency trade-off become realized as accidents in the real world. 
We close by discussing how such accountability mechanisms encourage more just, transparent governance aligned with public values.\\ 

\noindent This chapter is a licensed derivative copy of work published and awarded an oral presentation slot at \emph{ACM EAAMO 2021}~\cite{cooper2021eaamo}. 
Another version of this work, targeted at a legal audience, was published in the \emph{Colorado Technology Law Journal} in 2022~\cite{cooper2022fast}.

\input{section/30-algorithms/33-tradeoffs/331-tradeoffs-intro}
\input{section/30-algorithms/33-tradeoffs/332-tradeoffs-ubiquity}
\input{section/30-algorithms/33-tradeoffs/333-tradeoffs-computing}
\input{section/30-algorithms/33-tradeoffs/334-tradeoffs-policy}
\input{section/30-algorithms/33-tradeoffs/335-tradeoffs-conclusion}

%% file: section/30-algorithms/33-tradeoffs/331-tradeoffs-intro.tex
\section{Introduction}

Engineering is defined by trade-offs --- by competing goals that need to be negotiated in order to meet system design requirements. 
One of the central trade-offs, particularly in computer science, is between \emph{accuracy} and \emph{efficiency}. 
There is an inherent tension between \emph{how correct} computations are and \emph{how long} it takes to compute them. 
While this trade-off is of general relevance, it plays out in various ways across computing: 
in computer hardware, circuits can use approximation techniques to relax constraints on accuracy --- on how they perform bitwise computations --- to speed up performance; 
in image processing, compressing pixels causes a loss in accuracy of the image being represented, but also furthers space-efficiency by requiring less memory for storage. 
In fact, such trade-offs are so abundant in computing that they have even given rise to its own subfield, \emph{approximate computing}~\cite{moreau2018taxonomy, mittal2016apsurvey}, which studies how different domains resolve the question of how much inaccuracy can safely be permitted for the sake of increased efficiency~\cite{sampson2015thesis}.

While the trade-off is commonly acknowledged in computer science, its policy implications remain poorly examined. 
We provide a starting point, in which we focus our analysis on \emph{distributed ML systems} using the running example of autonomous vehicles (AVs). 
We make this choice for two reasons. 
The first is urgency: 
AV development has made such significant strides that by 2040 at least 75\% of cars will have some level of autonomy~\cite{newcomb2021avs}. 
Second, while AVs promise to improve overall driving safety,\footnote{The international effort to deploy AVs is motivated in large part due to AV technology's promise to increase automotive safety --- 
    that replacing human drivers with automated ones will protect millions of lives. 
    Conservative estimates indicate that in 2035-2045, the decade in which AVs are targeted to reach widespread deployment, 585,000 lives will be saved worldwide~\cite{intel2017avs}.
} they will also create new risks~\cite{ooida2020avs, boudette2021tesla}. 
As we show, some of these risks directly result from the accuracy-efficiency trade-off and the choices made to implement it~\cite{ntsb2019ubercrash}. 
In particular, the trade-off is tunable and context-dependent: 
it is not an all-or-nothing choice, and appropriate tuning depends on both a system's goals and deployment environment. 
Choices in different contexts will entail different emergent behaviors in technical systems --- behaviors that are potentially high-stakes if, for example, they affect overall system safety.

We argue that the accuracy-efficiency trade-off exposes a high-level abstraction that policymakers should use to help hold such systems accountable.\footnote{We emphasize that this is \emph{not the only} such tool policymakers should have for holding these systems accountable. 
    Other accountability mechanisms are also necessary, such as those that can assess hardware failures~\cite{abraham2019responsibility, surden2016avs, aaj2017avs}, the explainability of ML models~\cite{kroll2017aa}, and the impact of variance in automated decision-making~\cite{forde2021model}.
} Rather than operating at one of two extremes --- solely having policymakers rely on technical experts to make high-stakes decisions or inundating policymakers with underlying low-level technical details --- we advocate for something in between: 
researchers should focus on providing correctness and performance guarantees, and should build tools to help policymakers reason about these guarantees. These tools should help expose the uncertainty in distributed ML systems. 
This would facilitate lawmakers' ability to assess whether trade-off implementations are aligned with safety goals, and to regulate the risk of deploying high-stakes systems like AVs.

We emphasize \emph{distributed systems} because much of the sociotechnical conversation in ML has focused on \emph{algorithmic} fairness. 
This has left the systems components --- notably, scalability, speed and their impact on correctness --- under-explored in terms of their policy implications.  
As a result, ML \emph{systems} present under-examined challenges for technological accountability. 
We take the initial steps to bring some of these challenges to light, and suggest a novel framing for how to hold such systems accountable. 

This contribution demonstrates the need for mandatory risk assessment tools for distributed ML systems. 
We contend that, without such tools, effective public oversight of these systems will not be possible. 
Instead, we run the risk of manufacturers ignoring accountability mechanisms when constructing ML systems --- or worse, deliberately making these systems difficult to assess in order to obscure responsibility when accidents occur. 
In both of these scenarios, the burden would fall on individual victims to prove manufacturer responsibility. 
This dynamic would make accountability quite difficult to achieve; 
the power and resource imbalances between individual victims and large ML-system manufacturers would make tort or other civil litigation infeasible~\cite{abraham2019responsibility}. 

Our analysis focuses on the US, but elicits principles that apply more broadly. 
We have chosen AVs as our central example because navigating the trade-off appropriately has already proven an urgent concern, notably in assessing Uber's 2018 AV crash~\cite{ntsb2019ubercrash}. 
To make our case, we survey relevant concepts and examples from law and computer science, and then synthesize this discussion to advocate for a concrete policy contribution, which we direct toward the National Highway Transportation Safety Authority (NHTSA).\footnote{Approaching our topic in this interdisciplinary manner leads us to follow a nontraditional format. 
    We need to justify our conceptual contribution in two directions, and thus provide a significant amount of relevant background information concerning how the accuracy-efficiency trade-off translates to both law and computer science.} 
We first discuss how the trade-off functions in relation to decision-making in disciplines other than computing, most notably in US risk assessment policy (Section \ref{sec:eaamo:price}). 
Then, we provide an analogous discussion for ML algorithms and distributed ML systems (Section \ref{sec:eaamo:computing}). 
We argue that reasoning about accuracy-efficiency trade-offs and accountability in highly technical domains is not a new problem. 
This suggests that, with the right technical tools, we can similarly hold high-stakes, distributed ML systems like AVs accountable (Section \ref{sec:eaamo:policy}) with respect to how they implement analogous trade-offs. 
We close this chapter by discussing how such tools for increased accountability encourage more just, transparent governance aligned with public values (Section \ref{sec:eaamo:conclusion}). 

%% file: section/30-algorithms/33-tradeoffs/332-tradeoffs-ubiquity.tex
\section{The Ubiquity of Accuracy-Efficiency Trade-Offs} \label{sec:eaamo:price}

The trade-off at the heart of this paper is not unique to computing. 
It can be observed in a range of domains, many of which are regulated in the US, including law, the economy, and public health.\footnote{The accuracy-efficiency trade-off is also salient in other aspects of governance, including wartime intelligence gathering. 
    The ``fog of war'' concerns the inherent tension between gathering more accurate intelligence about an opponent or enemy and acting on that intelligence before it becomes stale and loses its usefulness~\cite{clausewitz1832fog}.} 
In these disciplines, efficiency often can be thought of interchangeably with speed. 
For example, in decision theory, the time-value of information is an important concept for making choices. There is a cost to gathering increasingly accurate information: 
waiting to act is itself an action --- one that can have more negative consequences than acting earlier on imperfect information\footnote{Kahneman et al.~ elaborates on this idea in well-known cognitive psychology research concerning reasoning about uncertainty~\citep{kahneman1982uncertainty} 
    The authors argue that humans use various heuristics to make decisions more efficiently, often acting on biases they have due to incomplete information. 
    There is a tension between taking the time to gather more information and making a more informed decision --- between the speed of making a decision and the quality of information used to make it.
} 

Sunstein~\cite{sunstein2002heuristics} connects this idea to the potential hazards of using heuristics in legal decision-making. 
Nevertheless, he observes that heuristics are common (and necessary) to obtain a suitable balance between efficient resolution and the ``best'' (i.e., most accurate) adjudicative outcomes.\footnote{Due process is perhaps the most notable, encompassing example of balancing both values in US law.} 
For example, a number of rules in US civil and criminal procedure --- speedy trial requirements, local filing deadlines, statutes of limitations --- impose time constraints for the sake of efficient case resolution; 
these values must be balanced against needs for thorough fact-finding and argumentation. 
The standard for preliminary injunctive relief in the US requires courts to predict whether irreparable injury will occur because of the passage of time, if relief is not granted before the (often lengthy) full resolution of a case~\cite{lichtman2002injunctive}. 
Federal Rule of Evidence 403 allows for the exclusion of relevant evidence from a court proceeding if the probative value of that evidence is substantially outweighed by a danger of undue delay. 
These and other rules promoting judicial efficiency are, in the words of Justice Oliver Wendell Holmes, ``a concession to the shortness of life''~\cite{Reeve_Dennett} --- they attempt to balance between the twin goals of getting matters right and getting them done, with recognition that there is real social value to each. 

Debates about the merits of the ``precautionary principle'' in policymaking also reflect the trade-off. 
The precautionary principle advises extreme caution around new innovations when there is substantial unknown risk; 
it places the burden of proof on risk-creating actors (like chemical plants) to provide sufficient evidence that they are \emph{not} producing significant risk of harm. 
As with speedy trials, there is a trade-off between the time it takes to gather evidence --- to understand the risk landscape --- and making informed decisions based on this landscape.\footnote{There are legal rationales on both sides of the spectrum with regard to how this trade-off should be implemented. 
    For example, critics of the precautionary principle could be said to favor efficiency. They find the principle to be too stringent with regard to the burden it places on accuracy; 
    it is ``literally paralyzing'' in its attempts to regulate risk~\cite{sunstein2003precaution}. 
    On the other side, others argue that the precautionary principle provides a valuable way to reason about preventing harm by shifting the burden of proof of safety to potential risk creators. 
    They are supportive of the fact that the principle requires actors to justify the risks they create: 
    it is worth the time cost to gather information, such that it is possible to better manage risk in the context of scientific uncertainty~\cite{sachs2011precaution}.
} 

A notable example of the precautionary principle demonstrating the trade-off in action concerns public health management of the SARS outbreak in the early 2000s. 
During the early outbreak of the disease, there was significant uncertainty around the risk of it spreading and how lethal it could be. 
The principle was adopted as a public health value at all of the disease epicenters: 
individuals who were even remotely suspected of having come into contact with SARS were placed under strict quarantine. 
Years later, (pre-COVID-19) critics argued that mass quarantining led to a tremendous and unnecessary loss of liberty. 
They made this case based on analysis that indicated 66\% fewer individuals could have been quarantined with the same public health outcome (i.e., it would have still been possible to prevent a SARS pandemic)~\cite{chowkwanyun2016health}.\footnote{We are not yet at a time in which such retrospective analysis regarding the precautionary principle can be conducted for the ongoing COVID-19 pandemic. 
    Nevertheless, the trade-off has still played a role in an additional public health context: antibody tests. 
    The World Health Organization (WHO) has recently argued that, prior to certifying COVID-19 antibodies for treatment, it is necessary to \emph{guarantee} that such antibodies confer immunity to the virus. 
    Several medical professionals have challenged this mandate from WHO, highlighting the time-sensitive nature of taking action in the pandemic: 
    ``Demanding incontrovertible evidence may be appropriate in the rarefied world of scholarly scientific inquiry. 
    But in the context of a raging pandemic, we simply do not have the luxury of holding decisions in abeyance until all the relevant evidence can be assembled. 
    Failing to take action is itself an action that carries profound costs and health consequences.'' 
    More generally, it is the norm for healthcare practitioners to act on incomplete information --- to balance potential inaccuracies in available data with the urgency to treat serious conditions~\cite{weinstein2020covid}.
} 

\subsection{US federal risk assessment policy} 

The examples above provide an intuition for how pervasive the accuracy-efficiency trade-off is in different domains, and how it is reasoned about to guide decision-making. 
Beyond this intuition, the trade-off is implicated more formally in US federal risk assessment standards and regulatory rule-making. 
Risk assessment policy acknowledges that, no matter how much time and resources one spends gathering scientific knowledge to assess risks, it will ultimately always be necessary to make decisions with uncertainty --- to pass judgments in the face of incomplete information~\cite{nrc1983riskassessment, nrc1994riskassessment}.\footnote{As Levy and Johns notes, it is the epistemological nature of science itself that makes uncertainty inevitable in science-based policymaking: 
    ``Agencies charged with protecting public health and the environment must make decisions in the face of scientific uncertainty, because science by its nature is incomplete and only rarely provides precise answers to the complex questions policymakers pose''~\cite{levy2016transparency}.
} 
There is always a degree of imprecision in scientific knowledge's ability to capture what is true, and that knowledge is constantly subject to revision in light of newly collected information. 
That is, taking more time to gather information can increase accuracy, but is directly at odds with efficiency in decision-making. 

In risk assessment, this trade-off is framed in terms of \emph{ex ante} (before-the-fact) and \emph{ex post} (after-the-fact) risk-mitigating interventions. 
The AI safety and fairness communities sometimes use the terms \emph{assessment} and \emph{audit}, respectively for \emph{ex ante} and \emph{ex post} \cite{falco2021audit}. 
\emph{Ex ante} mechanisms embody the precautionary approach: 
they emphasize collecting evidence about potential risks before approving a new substance or technology. For example, the FDA\footnote{US Food and Drug Administration (FDA).} 
typically requires multiple phases of clinical trials before a new drug is approved for use (i.e., ``premarketing approval''~\cite{nrc1994riskassessment, nhtsa2016avs}). 
This \emph{ex ante} regulatory authority is deliberately slow for the sake of increased safety.\footnote{The FDA is empowered to require drug companies to submit sufficient data, such that a detailed risk assessment can be conducted before the drug goes on the market. 
    This process can take a lot of time, and is not always conducted without criticism concerning choosing ``safety'' over ``efficiency''. 
    For example, such critiques are common when swift approval has known safety benefits, but is delayed in favor of evaluating the presence of unknown (potentially non-existent) health risks. 
    Debates concerning the FDA and this accuracy-efficiency trade-off have been particularly relevant recently concerning approving COVID vaccines for children~\cite{parkerpoper2021covid}.
} 

In contrast, for efficiency, other agencies concentrate their authority in \emph{ex post} ``post hoc mechanisms''~\cite{nrc1994riskassessment}.\footnote{These mechanisms tend to require that agencies, rather than companies, acquire the data necessary to determine responsibility after an undesirable outcome occurs.} 
NHTSA has relatively weak \emph{ex ante} authority for determining what types of vehicles are safe to drive; 
its strongest authority is the ability to recall faulty cars \emph{ex post}~\cite{nhtsa2016avs,vinsel2019cars}.\footnote{NHTSA has the ability to set safety standards, and then verifies that manufacturers have met them through a self-certification process. 
    In other words, manufacturers certify themselves as ``safe,'' rather than NHTSA soliciting data from manufacturers and performing the certification themselves~\cite{vinsel2019cars, nhtsa2016avs}.
} 
NHTSA favors lack of \emph{ex ante} regulation as a way to ensure speedy development and deployment of new car technology, even if such lack of regulation comes with a cost in correctness in that technology. 
These are just two examples illustrating opposite choices concerning how accuracy and efficiency relate to \emph{ex ante} and \emph{ex post} enforcement. 
This trade-off spectrum applies to the risk assessment and rule-making practices of numerous other US agencies, including the EPA,\footnote{Environmental Protection Agency (EPA).} 
OSHA,\footnote{Occupational Safety and Health Administration (OSHA).} 
and the CPSC,\footnote{Consumer Product Safety Commission (CPSC).} which each have different, domain-specific \emph{ex ante} and \emph{ex post} biases. 

Despite these differences, reports from the NRC\footnote{National Research Council (NRC).} 
recognize that there are cross-cutting elements of risk assessment~\cite{nrc1994riskassessment,nrc1983riskassessment}. 
The reports provide general recommendations for improving standards for accounting for uncertainty and its relationship to risk, such as clarifying the assumptions that inform model construction to elucidate model uncertainty. 
The NRC advocates for the importance of teasing out these low-level details, and communicating them to both decision-makers and the public, in order to ensure that policy goals reflect the known risk landscape. 

This discussion shows that accuracy-efficiency trade-offs are a useful and natural way for policymakers to regulate varied, complex technical domains. 
We therefore ask: why not use this framework for making policy concerning distributed ML systems? 
The specifics of the domain may vary --- notably, real-time systems involve high speeds not present in, for example, evaluating the safety of new chemicals. 
Nevertheless, US risk assessment policy indicates that reasoning about accuracy-efficiency trade-offs, and their relationship to risk, is not a new problem. 
We therefore contend that reasoning about underlying accuracy-efficiency trade-offs can enable risk assessment and management for these emerging technologies. 
However, translating the above regulatory framing to this domain presents novel challenges. 
We will require new tools, which we clarify in Sections \ref{sec:eaamo:computing} and \ref{sec:eaamo:policy}, to reason effectively about similar trade-offs in distributed ML systems --- tools that expose the particular type of uncertainty in real-time, distributed, automated decision-making. 
These tools will help us gather the data necessary for appropriate risk assessment and policymaking. 

Before we can describe these tools, we clarify that accuracy-efficiency trade-offs are an appropriate abstraction for accounting for the behavior of distributed ML systems. 
Having explained how reasoning about such trade-offs is useful for policymaking, we next make our case from a technical perspective. 

%% file: section/30-algorithms/33-tradeoffs/333-tradeoffs-computing.tex
\section{Trading off Accuracy and Efficiency in Computing} \label{sec:eaamo:computing}

\begin{figure}[t!]
  \begin{center}
    \includegraphics[width=0.7\textwidth]{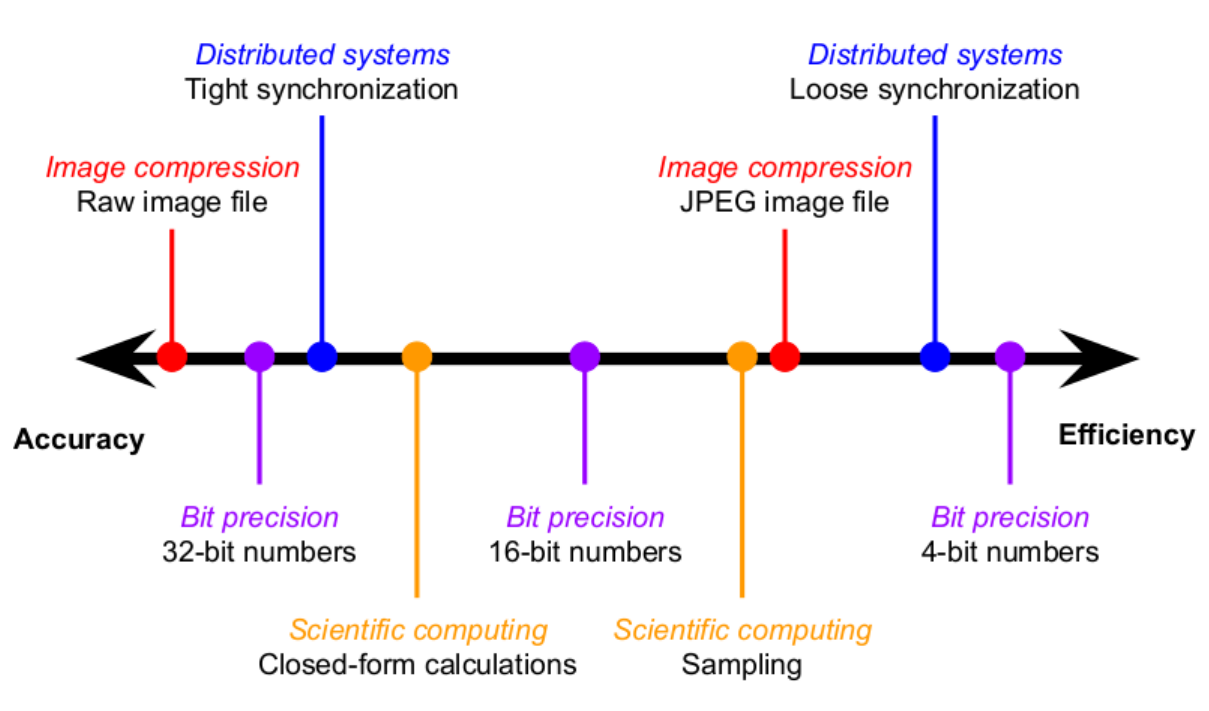}
    \caption{Computing examples of the accuracy-efficiency trade-off spectrum:
    \textcolor{red}{Image compression} (raw images are higher accuracy; JPEGs are more efficient), 
    \textcolor{violet}{bit precision} (32-bit numbers are higher accuracy; 8-bit numbers are more efficient; 16-bit numbers reflect an in-between),
    \textcolor{blue}{distributed systems} (tight synchronization is higher accuracy; loose synchronization is more efficient), 
    and \textcolor{orange}{scientific computing} (closed-form solutions are higher accuracy; sampling is more efficient).
    There are diminishing returns toward either end of the spectrum.} 
	\label{fig:eaamo:tradeoff}
  \end{center}
\end{figure}

Accuracy-efficiency trade-offs are particularly relevant across computing.\footnote{The accuracy-efficiency trade-off is arguably a central concern for the entire field of computing. 
    Ohm and Frankle call efficiency the ``cardinal virtue'' of computing in order to discuss what they view as exceptional cases of inserting inefficiency into computer systems --- what they term ``desirable inefficiency''~\cite{ohm70inefficiency}. 
    Instead, viewing the accuracy-efficiency \emph{trade-off} as central enables us to not refer to ``inefficient'' computing models (e.g. cryptography) as exceptional. 
    We conceive of them as implementing the trade-off at one end of the accuracy-efficiency spectrum (with cryptography privileging accuracy), which strikes us as a more precise and generalizable statement.
} 
To understand this, consider a familiar example --- JPEG compression. Raw images tend to be very high resolution: they contain many, varied pixels per inch, and therefore require a lot of storage space. 
However, a compressed, JPEG version often suffices for high quality; combining neighboring pixels often is not detectable to the human eye. 
A JPEG also takes up less storage space and can lead to faster processing when doing photo editing since there are fewer pixels to consider; it is more space- and time-efficient. 
Reducing the accuracy of the image can lead to greater computational efficiencies. 
This type of trade-off spectrum forms the basis of \emph{approximate computing} (Figure \ref{fig:eaamo:tradeoff}), which studies how a computer system can achieve certain performance benefits if it exerts less computational effort to compute perfectly accurate answers. 
In other words, it is possible to \emph{relax} accuracy in order to yield efficiency improvements~\cite{moreau2018taxonomy, mittal2016apsurvey, sampson2015thesis}.\footnote{We do not include the pathological case in which \emph{all} accuracy is sacrificed in order to do something really fast but completely wrong. 
    Nevertheless, there are cases where an implementation could, for example, be wrong 40\% of the time (for increased speed) and still achieve certain application-specific quality goals.} 

As with JPEGs, relaxing accuracy does not necessarily have negative consequences; 
rather, it is possible that decreased accuracy has no observable impact for a particular application. 
That is, some applications are tolerant of inaccuracy; they are error resilient. 
Similar to non-computing domains, tools for reasoning about the trade-off inform decisions about how to implement it. 
Computer scientists create theoretical tools to characterize the trade-off, which they leverage to determine the right implementation in different applications. 
Formal reasoning about the trade-off can yield application-specific quality metrics, where quality can be thought of as whether a program produces ``good enough'' results. 
Often, ``good enough'' cannot be guaranteed with complete certainty, but can be verified with high probability. 
Leaving room for uncertainty allows for edge case behaviors that fall below the specified quality threshold. 
Quality metrics therefore capture how much an approximation is allowed to deviate from the precise version's results. 
Computer scientists can then design software that requires a certain degree of program quality with a certain (high) probability~\cite{sampson2015thesis}.\footnote{A practical example of this comes from Amazon's cloud computing services (AWS). 
    Their cloud storage service provides ``11 9's'' of reliability with regard to storing data objects, meaning that 99.999999999\% of the time saving such objects to the cloud occurs without error~\cite{amazon2020s3}.
}

\subsection{Accuracy-efficiency trade-offs in ML} 

Such trade-offs are a salient concern across ML. 
Notably, in deep learning, there is an ongoing, increasing emphasis on training larger models to yield more accurate results. 
This comes with host of efficiency challenges, including significantly increased training time, model storage requirements, and energy usage~\cite{kaplan2020scaling}.\footnote{The trade-off notably did not first become relevant with (though is arguably increasingly urgent due to) the advent of modern statistical ML. 
    Several influential papers on artificial intelligence (AI) from the 1980s and 1990s also demonstrate the potentially high impact of appropriately dealing with accuracy-efficiency trade-offs~\cite{horvitz1987constraints, boddy1994deliberation}.
}
Moreover, ML models perform inference that is not always correct; to be robust, models need to tolerate a certain degree of inaccuracy. 
This notion of error resilience (or inaccuracy tolerance) varies for different ML algorithms. 
Regardless of particular differences, there is a general tension between \emph{correctness} and \emph{performance}.\footnote{For example, the correctness of a training algorithm can be understood as whether or not the algorithm converged to the distribution we set out to learn, i.e., \emph{Did we learn the right model?} 
    Its performance indicates whether convergence to the distribution --- whether correct or incorrect --- happened in a timely manner, i.e., \emph{How fast did we learn the model?}
} 
In fact, relaxing accuracy to increase efficiency is a requirement in many learning domains. 
Otherwise, computations can be so slow to perform that they become intractable. 

One relaxation strategy\footnote{We give four general strategies in this paper, which are far from exhaustive. 
    Notable examples of subfields with specific trade-offs include reinforcement learning (RL) and Markov chain Monte Carlo (MCMC). 
    In RL, there is the well-known exploration-exploitation trade-off (more exploration increases accuracy and more exploitation increases efficiency)  \cite{ishii2002rl, jin2020rl}. 
    In MCMC, algorithms exhibit scalability-reliability trade-offs (scalability corresponds to efficiency, reliability to accuracy) \cite{zhang2020tunamh}.
} 
is \emph{subsampling} during training, which involves using a subset of the dataset in place of the entire dataset to compute model updates faster.\footnote{Performance directly relates to the size of the task on which we conduct learning. 
    Intuitively, if a learning algorithm is slow on tasks with small datasets, then that algorithm will be slow, if not computationally intractable, on much larger ones. 
    This relationship between runtime and task size often exists due to coupling between the computation done by the learning procedure's optimization algorithm and the task's dataset size. 
    For example, when computing the gradient needed to determine which direction the learning algorithm should step for its next iteration, it is often necessary to sum over every data point in the dataset.
} 
Even though each iteration is less accurate (but more efficient), some algorithms can still guarantee overall high-quality (i.e., statistically correct) results. 
A very common approach for improving efficiency is to use a subsample or \emph{minibatch} of the dataset, rather than the whole dataset, when performing calculations. 
In the case of computing gradients, instead of using a \emph{full batch} (i.e., the whole dataset) we use a randomly sampled subset of the data points, which involves spending less time on the computation of a particular iteration.\footnote{Stochastic Gradient Descent (SGD) is an example of an algorithm that takes this approach, in which using a minibatch can  have minimal impact on the overall accuracy of the learned model. 
    A particular iteration of the algorithm will have less accuracy when computing the gradient; but, when run for lots of iterations, the final result is usually still statistically correct. 
    In expectation, we can learn the same distribution as if we had been using the whole dataset in each iteration; we can often theoretically guarantee robustness~\cite{bottou2018sgd}. 
    Moreover, the decision to subsample is not all-or-nothing; it is a spectrum. 
    It is possible to vary the minibatch size the algorithm uses. 
    Larger minibatches --- especially those that approach the size of the full dataset --- require more time but are also more accurate per iteration. 
    Conversely, smaller batch sizes make each iteration faster and more scalable to larger datasets, but in doing so sacrifice accuracy per iteration. 
    Determining the right sweet spot in this trade-off often depends on the particular learning task, and often falls under the area of study called hyperparameter optimization~\cite{feurer2019optimization}.
} 

A second strategy is \emph{asynchrony}, which enables different computer processes or threads\footnote{Threads and processes are mechanisms for parallelization within a computer \cite{arpacidusseau2018os}. 
    A process can have multiple threads running at the same time. 
    For example, this is what allows a text editor (which is running in a process) to simultaneously enable displaying both typing and syntax-error highlighting in real-time. 
    Each of these functions happens in its own thread, within the process of running the text editor application.
} 
to perform computations side-by-side and combine the results.\footnote{In other words, asynchrony can speed up ML since multiple parts of the learning problem can be computed at once.} 
This is more efficient but, depending on how the results are combined, can also lead to decreases in accuracy: 
if different processes work on overlapping parts of the overarching computation, one process can potentially overwrite the value recorded by the other out of sequence~\cite{desa2017async, alistarh2018convergence, lian2017asynchronous, Niu2011hogwild}. 
This can be avoided by forcing processes to coordinate their updates, but such coordination takes time; it increases accuracy, but decreases efficiency.\footnote{Out-of-sequence overwriting from asynchrony can be worth the speed-ups it enables; 
    it is still possible --- though not always guaranteed --- to compute good quality learning estimates~\cite{desa2016gibbs}. 
    Moreover, asynchrony can be used in conjunction with minibatching or resource-constrained devices, yielding additional accuracy-efficiency trade-offs.
}

A third strategy is to use \emph{resource-constrained techniques}, which involve smaller computers, such as Internet of Things (IoT) devices and sensors. 
With the advent of IoT in recent years, there has been a significant increase in the variety of computers available and a corresponding increase in the variety of computations we wish to run on them. 
For example, an Amazon Echo serves up answers to spoken language questions; however, it also has limited on-board capabilities to perform computations locally. 
These limitations take several forms. 
For example, such devices might not have a lot of power to process data quickly or might lack storage capacity for large amounts of data. 
As a result, such devices often only have smaller, coarser-grained models in local memory, which can be used for quickly returning (potentially less accurate) inference results.
Often, these devices can communicate with more sophisticated computers over the Internet, offloading computation or storage to those computers. 
Because these computers have more memory and processing capabilities, they can store larger models that are capable of more nuanced inference.\footnote{However, this communication exposes another accuracy-efficiency trade-off; 
    it takes time to send the data to a remote computer, perform some (more accurate) computation, and then return a response to the device~\cite{birman2019cloud}. 
    That computation may be more accurate due to using a larger, finer-grained model, but achieving that accuracy comes with a cost in speed. 
    Conversely, doing the computation locally on the device would be faster; 
    however, due to the device's more limited computational resources, it will not necessarily be as accurate. 
    For example, prior work in computer vision considers how to handle the trade-off when performing ML on mobile devices, such as smart phones~\cite{howard2017mobilenets}. 
    This work uses manually-tunable parameters that allow the model developer to strike the right balance for particular learning problems. 
    Depending on the application domain, a model developer can tune a larger model that uses more resources (i.e., a model that is slower or uses more memory but is more accurate) or one that is smaller and uses fewer resources (i.e., a model that is faster or uses less memory but is less accurate). 
    Aside from being faster, there are several reasons why local computation and storage might be desirable for a mobile application, as opposed to offloading these requirements to more powerful remote computers. 
    Notably, local computation can ensure privacy, as the learned model and collected data never leave the mobile device~\cite{wang2019privacy}. 
}

A fourth such example of a strategy is \emph{low-precision computing}, or quantization, to use fewer bits to speed up computation (i.e., decrease accuracy for increased scalability)~\cite{desa2017async, gong2014quantize, courbariaux2015binaryconnect, alistarh2017qsgd, gupta2015lowprecision,han2015deep}. 
    This method, sometimes called quantization, is similar to the idea of floating-point precision --- how much accuracy the computer can capture based on how many bits it uses to represent numbers (Figure \ref{fig:eaamo:tradeoff}). 
    Computing with more precise floating-point numbers is more computationally expensive; it tends to take more time and memory (i.e., sacrifices efficiency) but can capture a more accurate range of results. 
    Much work in machine learning explores using low-precision numbers to achieve faster results. 
    This work relaxes requirements on the accuracy of the trained model in order to achieve these speed-ups. 
    There is also a spectrum at play here. 
    It is possible to vary the number of bits of precision: more bits yield higher accuracy and slowdowns, while fewer bits require less time per computation and thus potentially sacrifice some correctness. 
    Depending on a particular application's tolerance to error, this sacrifice in accuracy can be worth the speed-ups it creates~\cite{desa2018halp}.\footnote{It is also possible to implement low-precision computing in hardware~\cite{carmichael2019dnn, colangelo2018fpga, zhao2019overwrite}. 
    In general, we must also consider how the hardware specifications of the computer running the algorithm might also impact that behavior.
    Surely this is important, as different computers have different computing capabilities due to varying hardware; a NASA supercomputer has more computational resources than a personal laptop. 
    As with the subsampling, a low-bit-precision sacrifice in accuracy does not necessarily require sacrificing overall correctness, if in expectation the algorithm can still theoretically guarantee learning the right distribution.
}

\subsection{Implications in real-world ML systems} 

We have thus far provided examples of the trade-off in ML \emph{algorithms}, but have not yet considered how the trade-off behaves in \emph{deployed systems} --- systems that consist of multiple computers that work together to solve large, complex problems.\footnote{Such systems often introduce additional asynchrony: 
    instead of one computer running an algorithm to solve a task, multiple computers work together in parallel.} 
Our overall aim is to understand the particular trade-off challenges in such \emph{distributed ML systems}, so we need to account for the ``distributed systems'' component just as much as ``ML''. 
The distributed setting is what enables potentially life-saving technology like AVs.\footnote{These systems reflect a triumph of new systems abstractions, not just innovations in ML~\cite{birman2019cloud}.} 
Importantly, new risks emerge when such fast, scalable systems are deployed in the real world. 

For example, researchers recently built a model that they showed could outperform humans in identifying gay individuals using facial recognition technology~\cite{wang2018gay}.\footnote{This claim has been challenged by several researchers, notably Leuner~\cite{leuner2019replication}.} 
This disturbing result yielded a blizzard of media attention~\cite{hawkins2017gay, murphy2017gay}, yet it was also small-scale and slow. 
Consider a similar model, but one that is scalable and fast --- integrated with a CCTV surveillance system serving real-time inference and deployed in a country hostile to LGBTQ rights. 
This may sound like science fiction, but low-latency, distributed vision systems already exist~\cite{wang2018china}. 
While this example is generative concerning the range of potential risks from ML systems, we focus on the risks related to accuracy-efficiency trade-off implementations.\footnote{As we note in the introduction, while we focus our discussion of the policy implications of accuracy-efficiency trade-offs in distributed ML systems, reasoning about such trade-offs in other parts of computing could also serve useful to tech policymaking. 
    Similarly, we focus our analysis concerning accountability mechanisms to the accuracy-efficiency trade-off, even though distributed ML systems raise a variety of other accountability concerns, aside from this trade-off.} 

We next clarify how the trade-off is implicated in distributed computing, and then combine this with our ML discussion to show how the different tensions interact with each other. 
Considered together, ML and distributed computing trade-offs present especially challenging problems for real-time, high-impact systems like AVs. 
In Section \ref{sec:eaamo:policy} we will ultimately argue that clarifying the relationship between these risks and trade-off choices can help policymakers hold such systems accountable. 

\paragraph{Accuracy-efficiency trade-offs in distributed computing.} 
In contrast to a single computer, a \emph{distributed system} is a network of computers that can work together to solve problems. 
Each computer has its own data and performs its own computations, and it shares data and computation results with other computers in the network when necessary. 
Because the computers are in distributed locations --- whether in the same data center or across the world --- there are important considerations with regard to how efficiently information can be shared between them. 
When a computer contacts another in the system to request data, it takes time to complete the request and receive the data, reducing time-efficiency. 
There are also issues of accuracy between computers. 
Each computer has its own data --- its own view of the state of the overarching system. 
That information is not complete: 
it is just a subset, which can conflict with the views of the other computers in the system. 
In other words, in distributed systems we can more specifically frame the accuracy-efficiency trade-off as a tension between \emph{consistency} and \emph{latency}\footnote{Latency can be informally thought of as the speed with which the system updates.} 
There is a trade-off between all of the computers in the system having the same understanding of the data in the system and the time it takes to propagate that understanding throughout the system~\cite{abadi2012tradeoff, brewer12computer}. 

In distributed systems that update their data frequently it is quite difficult to quickly build a consistent, holistic understanding of the environment across different computers in the network.\footnote{One could informally view consistency is a moving target; each computer processes information locally faster than it can share it with the entire network.} 
Since it takes time to communicate, it is hard for computers to stay completely up to date with each other. 
For the sake of efficiency, individual computers in the system often need to make decisions in the presence of inconsistency.\footnote{Waiting for complete consistency across computers before an individual computer could make local changes would bring the entire system to a standstill. 
    This is especially relevant if a computer in the system experiences a fault; to achieve strong consistency, before proceeding with local computation, all of the other computers would be waiting to hear from a computer that can no longer communicate with them (i.e., they could end up waiting indefinitely).
} 

Particular distributed system implementations need to answer the question of how much application-dependent inconsistency and slowness they can each tolerate. 
To understand this spectrum, we will use the example of a social media website, which has computers hosting its data all over the world. 
A user tends to access the geographically closest computer server hosting the site; different users across the world therefore access different computer servers. 
Such a system favors efficiency (i.e., low latency) over the different computer servers being consistent with each other. 
It is more important to return the website to each user quickly than it is to make sure that every user is accessing the website with exactly the same data. 
This is one reason why on some social media sites it is possible to see out-of-order comments on a feed. 
To resolve its current state, the site aggregates information from across the system. 
It attempts to build a consistent picture, but limits how much time it spends doing so --- sacrificing consistency --- so that it can remain fast~\cite{decandia2007dynamo, lu2015existential, vogels2009eventualconsistency}. 
The system implements this choice via its communication strategy. 
Rather than contacting every computer in the system to construct a consistent picture, a particular computer only communicates with a subset. 
It trades off the accuracy it would get from communicating with every computer for the efficiency of communicating with fewer computers~\cite{hellerstein2019calm}.
Based on communication strategy, it is possible to quantify consistency and to measure it throughout a distributed system~\cite{lu2015existential, shang2018rushmon}. 
Developers can reason about the degree of inconsistency their particular system can tolerate safely, and can detect and tune the system accordingly to also enforce an upper bound on latency~\cite{golab2011funandprofit,barbara1990controlledinconsistency, yu2000numerical}. 

\paragraph{Distributed ML systems: AVs as a case study.} 
We can now specifically consider accuracy-efficiency trade-offs in real-time (i.e., latency-critical) distributed ML systems. 
We will focus on AVs as a concrete example, which will facilitate making concrete policy recommendations (Section \ref{sec:eaamo:policy}). 

An AV can be thought of as a distributed system of sensors.\footnote{This setting is further complicated by the fact that numerous vehicles can also be networked together (Vehicle-to-Vehicle, or V2V) and with other devices like smart traffic lights (Vehicle-to-Infrastructure, or V2I), which increase both the size and complexity of the system under analysis \cite{nhtsa2016avs, surden2016avs, transportation2014communications, fmvss2017communications}.} 
While each AV maintains its own local notion of the state of the environment, information that other AVs possess could also prove useful. 
If an accident is up ahead, an AV closer to the crash can communicate that information to those behind it, which in turn can apply their brakes and potentially prevent a pile-up. 

In such real-time transportation domains, accuracy and efficiency are both critical. 
Some ML applications may be able to tolerate wide margins of error, but in safety-critical domains a high degree of inaccuracy may be unsafe. 
The same goes for efficiency; such systems will need to make decisions quickly and, like the non-computing examples in Section \ref{sec:eaamo:price}, there is an inherent trade-off between waiting to make a completely informed decision and making a decision fast enough for it to be useful~\cite{abadi2012tradeoff, brewer12computer}. 
What is unique here for AVs is the degree of time-efficiency needed. 
In some cases, inference decisions will be necessary at sub-second speeds, and will therefore be computed using inconsistent or uncertain information. 
This presents a challenge; in the face of this uncertainty, we need systems like AVs to be guaranteed (at least with very high probability) to be accurate. 
The urgency of resolving this problem is not merely a hypothetical situation; the accuracy-efficiency trade-off in fact played a crucial role in the Uber AV crash in 2018~\cite{ntsb2019ubercrash}, which we will return to in Section \ref{sec:eaamo:policy}.

It is not entirely clear what the right trade-off implementation is for real-time systems like AVs~\cite{dietterich2018robustAI}. Unlike the example trade-offs in Figure \ref{fig:eaamo:tradeoff}, AVs are mobile and deployed in varying environments. 
While those examples each indicate a single, static, application-dependent trade-off implementation, an AV might instead need to support a range of trade-offs given the dynamic nature of the environment. 
A particular trade-off implementation may need to depend on different operational design domains (ODDs) that vary by roadway type, geography, speed range, and lighting, weather, and other environmental conditions \cite{nhtsa2016avs, sae2021avs}. 
Some ODDs will be more efficiency-critical: 
it would be catastrophic for a car to take an extra half-second to be certain that there is a pedestrian directly in front of it~\cite{ntsb2019ubercrash}. 
In other cases, having an accurate sense of the environment may be more important than speed. 
For example, when detecting a deep pothole up ahead, it could be safer for a car to slow down to decide its course of action --- to accurately determine if the hole is shallow enough for the car to continue on its course or deep enough to warrant veering off the road to avoid it.

As this example indicates, distributed ML systems raise different accuracy-efficiency questions than either distributed systems that do not involve ML, or ML systems that are not distributed. 
Since ML models (necessarily) approximate the world, it is possible for them to operate on data that are not completely accurate and still yield results that are correct \emph{enough} --- that fall within the same bounds of imperfection that we deem tolerable. 
We can extend such inaccuracies beyond things like subsampling to include the data staleness inherent in distributed settings~\cite{bailis2012pbs, decandia2007dynamo,yu2000conits}.\footnote{Staleness is not the only property that can be tolerated; 
    another example is numerical error that comes from asynchrony \cite{yu2000numerical}, which we elide for brevity.} 
Allowing for staleness increases efficiency, as the system does not need to wait to synchronize state before proceeding with its computation. 
As with a single computer, the overall output still \emph{can be} correct even when operating on stale data in a distributed setting; 
however, existing work in this field does not  guarantee such output \emph{must be} correct~\cite{alistarh2018convergence, gong2014quantize, lian2017asynchronous, Niu2011hogwild, desa2015taming, zhang2015sgdstaleness}.
For AVs, this does not suffice; we want to be able to guarantee correctness in order to be assured of safety.\footnote{Of course, with those guarantees predicated by certain assumptions. 
    At the very least, we need to bound the likelihood of incorrectness.
} 

Such assurance will require us to reason differently about the behavior of distributed ML systems. 
Prior work has examined the trade-off at a high level by looking at correctness and speed metrics of end-to-end ML systems~\cite{abadi2016tf, ho2013SSPParameterServer, li2014ps, kosaian2019paritymodels, pan2016cyclades}; 
this work uses overall empirical performance results to tune the staleness of the underlying data storage layer. 
There is a fundamental mismatch in this approach: 
high-level performance metrics are used to \emph{indirectly} tune low-level system behavior (to, in turn, affect high-level performance), without formalizing the relationship between the two. 
This is an inversion of what we ideally would like to do: to formally evaluate the underlying accuracy-efficiency trade-off, and use this information to \emph{directly} tune distributed ML system behavior. 
As a result of this mismatch, tuning has generally been manually curated to the particular problem or absent, leaving an engineer to pick from predefined settings that enforce high accuracy guarantees over efficiency, ignore accuracy guarantees altogether in favor of efficiency, or attempt some middle-ground. 

While there is a valid spectrum of trade-off points, current large-scale ML systems tend to opt for efficiency over accuracy.\footnote{They focus on minimizing communication between computers in the system in order to be fast enough to scale to larger problems. 
    Some of these systems can achieve orders of magnitude in efficiency improvements by dropping data updates without simultaneously destroying correctness~\cite{Niu2011hogwild, tsitsiklis1986stochastic}.
} 
It is not clear these approaches will be safe for systems like AVs.\footnote{It may not always be safe for these systems to lose updates. 
    Existing approaches to mitigate such losses in ML systems involve increasing communication between computers in the system. 
    However, this strategy impacts the other side of the trade-off, leading to inefficiencies from bottlenecks in coordination between computers. 
    This problem is similar to what exists in weakly consistent storage systems, which have side-stepped this issue by using semantic information to coordinate ``only when necessary''~\cite{dipippo1997semanticcc, molina1983semantic, weihl1988commutativity}.} 
It remains an open research question how safety-critical, real-time distributed ML systems like AVs should implement the trade-off. 

%% file: section/30-algorithms/33-tradeoffs/334-tradeoffs-policy.tex
\section{Accuracy-Efficiency Trade-Offs as a Mechanism for Accountability} \label{sec:eaamo:policy}

Systems like AVs are really complex, but complexity should not serve as a rationale to preclude their regulation. 
Rather, the fact that these challenges remain unresolved presents an opportunity: 
stakeholders aside from engineers can help shape implementations; they can inform accuracy-efficiency trade-off choices so that they align with the public's interests, not just those of manufacturers. 
This is why we have taken considerable space to clarify a variety of accuracy-efficiency trade-offs --- from how they impact computing broadly to how they describe a range of possible behaviors for distributed ML systems. 
Though much of our prior discussion is well-acknowledged in technical communities (albeit, in other forms), to date the trade-off's implications have not been made legible to policymakers. 
The trade-off is not binary; it is a spectrum and can be treated like a tunable dial set appropriately to the context (Section \ref{sec:eaamo:computing}). 
Our hope is that exposing this dial for distributed ML systems will provide a degree of technical transparency to lawmakers, such that high-stakes systems like AVs are not deployed without sufficient public oversight. 
We believe that explicitly exposing this trade-off provides a mechanism for holding these systems accountable for some of the risks they create. 

To do so, we address the gaps between existing risk assessment tools and what is needed to analyze accuracy-efficiency trade-offs in AVs. 
When an undesirable outcome occurs, we can examine accountability along two dimensions: 
the time window around the outcome, which we consider in \emph{ex ante} and \emph{ex post} divisions, and the actors that assess the system's behavior, which consist of computer scientists and policymakers. 
There is a region of overlap in which computer scientists can assist policymakers with \emph{ex post} evaluation and policymakers can frame \emph{ex ante} risks prior to deploying systems. 
We therefore propose a twofold call-to-action for enabling risk assessment in this domain: 
1) Computer scientists must build tools to expose underlying accuracy-efficiency trade-offs and 
2) Policymakers should use these tools to assess trade-off implementations, and meaningfully intervene to ensure implementations align with public values. 
We discuss these calls-to-action in terms of \emph{ex ante} and \emph{ex post} risk assessment gaps. 

\subsection{Addressing \emph{ex ante} risk-assessment gaps} 

A system's ability to be assessed with respect to the accuracy-efficiency trade-off should be considered as important as every other technical feature. 
We therefore call on computer scientists to engage in research to build tools in ML systems that make their accuracy-efficiency trade-offs assessable. 
We explain what we mean by ``assessable'' via example and then suggest research directions to help make assessments possible.\looseness=-1 

The 2018 Uber AV crash illustrates the importance of tools to assess the trade-off~\cite{ntsb2019ubercrash}. 
The crash resulted from the coincidence of several issues,\footnote{Together, the NTSB report generally summarizes these issues as reflective of a ``lax engineering culture'' around safety at Uber.} 
one of which had the accuracy-efficiency trade-off as its central problem. 
The AV remained inconsistent and indecisive for over 6 seconds.\footnote{The AV clearly had not implemented a robust inconsistency resolution strategy, as it this is a significant amount of time for a computer to not to resolve inconsistency.} 
By the time the sensors agreed about the presence of a pedestrian, the AV had already collided with her.\footnote{This example is far more complex than what we have glossed here. 
    For example, there were no other cars on the road, so it seems certain that slowing down to take the extra time to resolve inconsistency would have been safe. 
    Additionally, there was a human back-up driver; however, she was not paying attention. 
    Even if she had been, it is not clear that she could have responded appropriately within 6 seconds, as average time for human take-over from an AV is 17 seconds \cite{nhtsa2015humanfactors}.
} 
While the NTSB report is clear that the AV's sensors were inconsistent, it is not clear \emph{why} the AV could not make a decision. 
In this case, a granular explanation was not necessary to determine accountability, as 6 seconds is a very long time to be inconsistent. 
This AV was neither accurate nor efficient, indicating a sub-optimal trade-off implementation, as opposed to a well-reasoned choice, that led to a tragic outcome. 
In instances that are not as clear-cut, such as those that involve much tighter time windows, tools that provide granular explanations will be necessary to determine the difference between bugs and deliberate trade-off choices.

We need novel trade-off assessment tools to evaluate more difficult cases. Such tools could help avoid certain risks, guaranteeing \emph{ex ante} specific desirable system behaviors while foreclosing the possibility of other undesirable ones. 
That is, in some scenarios it may be possible to reduce the tension between accuracy and efficiency by taking coordination between computers off of the critical path; 
this would enable greater computational efficiencies without sacrificing accuracy in those contexts~\cite{hellerstein2019calm}. 
For example, program analysis could help formally categorize underlying accuracy-efficiency trade-offs, and therefore facilitate building asynchronous systems with more effective concurrency control and theoretically provable correctness guarantees~\cite{roy2015homeostasis, molina1983semantic}. 
This would solve the mismatch in current asychronous ML: instead of using high-level empirical observations to do ad-hoc, low-level system tuning (Section \ref{sec:eaamo:computing}), we could directly tune the underlying trade-off to guarantee end-to-end performance behavior.\footnote{More specifically, we could use program analysis to leverage the underlying semantics of the program and data to avoid synchronization (i.e., inefficiency); 
    these techniques would enable performing efficient, provably correct asynchronous computation.} 
If program analysis indicates that strong consistency is not possible, we could weaken this requirement by instead bounding how much inconsistency is tolerable. We could perhaps even bound inconsistency such that the overall correctness of the asynchronous computation is not too severely impacted~\cite{dipippo1997semanticcc, yu2000conits, yu2000numerical}. 

To make this idea concrete, consider that not \emph{all} of the AVs in the system will always need to communicate with each other. 
Instead, it will likely be sufficient for AVs to only communicate with others in an environment-dependent radius. 
Reducing communication to that radius would increase efficiency without decreasing accuracy, as AVs outside the radius would be too far away to have relevant information to communicate.\footnote{In other words, inconsistency between cars that do not need to communicate with each other is tolerable. 
    We instead prioritize (limited) communication between relevant cars, where relevance is determined via automated reasoning about the underlying semantics of the problem. 
    This example is extremely high-level --- described at the level of individual AVs --- for the purpose of clarity. 
    Semantic analysis will expose lower-level (i.e., at the level of particular data points), less-intuitively-explainable opportunities for better concurrency control.
}  

By providing such mechanisms to reason about accuracy-efficiency trade-offs, computer scientists expose a particular kind of decisional uncertainty that depends on time~\cite{horvitz1987constraints, boddy1994deliberation}. 
Clarifying this uncertainty does not, however, identify specific risks that automated decisions can bring about.  
Rather, it is up to policymakers to frame potential risks and to identify the normative, domain-specific values at play~\cite{jasanoff2016ethics, friedman2019values, flanagan2014values, goldenfein2020handoff}. 
Based on the uncertainty that computer scientists expose, policymakers should endeavor to assess \emph{ex ante} how much of the resulting risk is tolerable. 
Such \emph{ex ante} interventions could help narrow the space of potentially deviant system behavior, which in turn could help narrow the number of incidents examined \emph{ex post}. 
These interventions, though unlikely to be comprehensive, should clarify many of the risks in deploying these systems. 
However, it will not always be possible to preemptively fully analyze the risk landscape due to the amount of uncertainty in the system~\cite{sunstein2003precaution, smith2015opportunism}. 
Incomplete risk analyses will not necessarily prevent the deployment of real-time ML systems in practice; 
instead, policymakers will need to evaluate system behavior \emph{ex post}, after undesirable outcomes occur. 
A bad outcome will either reveal a risk that policymakers previously did not consider, with which they now need to contend, or it will implicate an acknowledged risk previously deemed acceptable. 

\subsection{Addressing \emph{ex post} risk-assessment gaps} 
When deployed for long enough, high-stakes ML systems are likely to incur severe harms that we likely did not anticipate~\cite{perrow1999risk, vaughan1996challenger, nissenbaum1996accountability, smith2015opportunism}. 
This is where tools that expose the accuracy-efficiency trade-off, described above, can facilitate accountability after-the-fact: 
they could facilitate determining if a system has deviated further than expected from normal behavior (i.e., what \emph{ex ante} risk assessment deems to be acceptable)~\cite{sampson2015thesis}.\footnote{\emph{Ex ante} audit systems abound in security-related literature. 
    For example, see Falco et al.~\cite{falco2021audit}, Haeberlan et al.~\cite{haeberlen2007peerreview}, and Lampson~\cite{lampson2004security}.
}
In these cases, policymakers would still be able to hold the appropriate stakeholders accountable \emph{ex post}. 
We do not claim that policymakers need to understand low-level technical details to provide this oversight (e.g., the particulars of concurrency control algorithms). 
Rather, we are suggesting that surfacing higher-level trade-offs (that lower-level technical decisions entail) clarifies valid sites for potential policy intervention. 
Such trade-offs are the right level of abstraction with which policymakers can engage in order to reason about relevant policy goals; 
the accuracy-efficiency trade-off can clarify how lower-level engineering decisions relate to overall notions of system safety~\cite{sampson2015thesis}.

It is this reasoning that informs our second call-to-action: 
policymakers should view the accuracy-efficiency trade-off as a regulable decision point at which they can meaningfully intervene. 
They already do so in other complex technical domains, for which they reason about risk and interventions (Section \ref{sec:eaamo:price}). 
This suggests that, with the right tools integrated with distributed ML systems --- like those we suggest above --- policymakers should also be able to do so for these systems. 
We do not articulate specific policies, as these will depend on a more comprehensive study of AV technology beyond the scope of this paper. 
Instead, we have used AVs as a guiding example to illuminate abstract technical concepts and their import for technology policy concerning accountability. 

It is possible to view this contribution is as an extension of existing risk assessment tools in computing. 
Contemporary policy debates about high-stakes ML applications in policing, transportation, and public health also involve concerns about what degree of accuracy we ought to demand from automated systems. 
These concerns often arise in attempting to minimize disparate outcomes across groups.\footnote{E.g., differential accuracy rates for face recognition along dimensions of race and gender~\cite{buolamwini2018gender, cooper2021emergent}.
} 
But we contend that debates about the harms of inaccuracy are incomplete if they fail to reckon with the accuracy-efficiency trade-off. 

For policymakers, these debates will require trade-off assessment tools to analyze gaps between the expected risks and the actual behavior of distributed ML systems. 
For example, we could fairly pose to policymakers questions like the following: 
at what point is information sufficiently high quality to justify a system executing high-impact decisions? 
When is it safe for a system to spend more time computing decisions, particularly when more efficient heuristics do not sufficiently remove uncertainty? 
These tools will therefore take a step toward closing the ``responsibility gap''~\cite{jasanoff2016ethics}: 
policymakers will have a more complete understanding of technology and will be better equipped to gauge the range of possibilities for its governance. 
This way, when technological failures occur,
policymakers can \emph{ex post} more actively participate in the evaluation of how uncertainty in distributed ML systems contributes to risk.

%% file: section/30-algorithms/33-tradeoffs/335-tradeoffs-conclusion.tex
\section{Conclusion: Toward More Just, Transparent Public Governance} \label{sec:eaamo:conclusion}

We have made the case for using accuracy-efficiency trade-offs as a policymaking lever for assisting in holding distributed ML systems accountable. 
For AVs, trade-off-informed \emph{ex ante} regulation could constrain the space of undesirable AV behavior, which in turn could narrow the the number of accidents and anomalous behaviors that need to be examined \emph{ex post}. 
This could lead not only to overall safer behavior, but also the necessary tools to determine accountability when accidents unavoidably occur (Section \ref{sec:eaamo:policy}). 
More broadly, this discussion can be situated in the context of extracting higher-level values from technical systems --- values such as safety and efficiency~\cite{nhtsa2016avs} --- as a necessary part of public governance. 
That is, it is crucial to analyze how higher-level values get implemented via underlying technological mechanisms --- in this case, the implementation of the accuracy-efficiency trade-off --- to ensure that the implementation aligns with the values that we want to promote in policy. 
We have argued that the accuracy-efficiency trade-off is not only a correct abstraction, but also the correct level of abstraction, for helping to promote this goal.

Clarifying technical details at this level of abstraction implicates another important value of public governance: transparency. 
For example, NHTSA has generally does not intervene \emph{ex ante} in regulating automobiles \cite{nhtsa2016avs, vinsel2019cars, abraham2019responsibility}. While this might make car development more efficient,\footnote{This is a contestable claim. 
    Please refer to Vinsel~\cite{vinsel2019cars} for more details concerning how safety regulations can in fact promote innovations in car technology.} 
it can come with a loss of transparency. 
Not engaging with technical details \emph{ex ante} can present problems beyond not detecting bugs; it can also lead to not being able to detect whether values like safety are implemented appropriately. 
Worse, it is possible that technical values, and the social values they entail, can be deliberately obscured. 
Technical implementation decisions can be framed as trivial, which can direct policymakers away from viewing them as valid sites for intervention.\footnote{Alternatively, when highly technical jargon is used to describe implementation decisions, it can serve to obfuscate rather than clarify. 
    Rather than enabling transparency for policymakers, who do not tend to be technical experts, these practices can cloud the values at stake~\cite{mulligan2018governance}. 
    In the automotive industry specifically, increasing digital automation has notably led to additional transparency issues, even prior to AVs. 
    Computerized features, in comparison to mechanical ones, can be programmed more easily to obscure true technical performance --- for example, to reduce recorded EPA emissions in order to appear more environmentally-friendly \cite{vinsel2019cars}. 
    While out of the scope of this paper, it is worth acknowledging that increased computerization in AVs potentially presents even more transparency issues of this variety.
}

Mulligan and Bamberger~\cite{mulligan2018governance, mulligan2019ml} have notably written about this issue of technological transparency in public governance. 
They call out the danger of policy-relevant values decisions getting pushed into low-level implementation decisions made by engineers, in place of having the values at play being openly debated. 
This misplacement of responsibility on engineers comes from a lack of technical expertise in governance and a resulting lack of mechanisms to regulate technology. 
Industry testing and quality control effectively give manufacturers the job of converting the law into concrete technical requirements: 
manufacturers, instead of public advocacy groups or agencies like NHTSA, make technical decisions with policy implications without public oversight. 
This conflict-of-interest can lead to compromising or degrading higher-level social values.

We have argued that if policymakers understand the accuracy-efficiency trade-offs in distributed ML systems, and the social values these trade-offs implicate, this problem can (at least in part) be averted. 
Policymakers will have a more sufficient understanding of technology and will be better able to determine the scope of possibilities for its governance. 
By understanding the technical values at stake at this level of abstraction, policymakers, with engineers' assistance, could provide insight \emph{ex ante} into how certain implementation decisions should be made. 
That way, low-level technical matters will not be dismissed as ``just implementation details'' left up to the whims of engineers without public oversight~\cite{mulligan2018governance, jasanoff2016ethics, friedman2019values}. 
Moreover, when technological failures and accidents do occur --- and it is a question of when, not if --- rather than viewing them simply as ``unintended consequences'' or ``normal accidents''~\cite{perrow1999risk}, policymakers and other relevant stakeholders could more actively participate \emph{ex post} in holding such systems accountable for their behavior. 
This more-effective public governance will improve the power imbalance between system manufacturers and victims of system accidents --- empowering and protecting individuals without the resources to seek justice for themselves.

%% file: section/40-genai/400-genai.tex
\part{Evaluating Generative-AI Systems}\label{part:genai}

Recent developments in ``Generative AI'' make unmistakably clear that  ML-research questions about reliable, scalable measurement are inseparable from law and policy considerations. 
They also make clear how prevalent the barriers to accountability are in this contemporary moment (Appendix~\ref{chapter:accountability}), in which generative-AI systems have become commonplace. 
In this part, we extend our work to questions in robustness, training-data provenance, and the associated implications for U.S. copyright law. 
This chapter reflects work that has been published at \emph{CVPR} (poster), \emph{ICML}, \emph{ACM CSLAW} (Long Presentation), and \emph{The Journal of the Copyright Society}, and work under submission at \emph{Nature}.

First, we discuss large-scale empirical work on training-data memorization.
Much prior work has demonstrated that large language models (LLMs) can memorize their training data~\citep{carlini2023quantifying}. 
Our work pins down a definite way to measure \emph{extractable} memorization: memorization that an adversary can feasibly get LLMs to regurgitate within a limited compute budget. 
We conduct the first work to successfully extract memorization at scale for a closed system with an aligned model --- \chatgptendpt (Chapter~\ref{chapter:memorization}, Nasr et al.~\citep{nasr2023scalable}).   
Our novel approach estimates that ChatGPT extractably memorizes nearly $3\%$ of its training data: $150\times$ more than prior estimates. 
This result has clear consequences for privacy, as we show that some extractable memorization contains personally identifiable information (e.g., phone numbers). 
It also has potential implications for copyright, for cases in which regurgitated training examples contain content that is copyrighted and not explicitly licensed for use in ML training~\citep{lee2023talkin}.\looseness=-1

Concerns about licensing and training data raise a natural question: is it possible to side-step many copyright issues by training high-quality models on explicitly licensed data? 
Second, to answer this question, we tackle a variety of ML-systems and data-quality problems to address the feasibility of training latent diffusion models~\citep{rombach2022diffusion} on Creative-Commons (CC) images. 
While there remains much work to do in this area, our initial research already demonstrates significant promise. 
Our CC-image-trained model performs well on human evaluation and, unlike models trained on web-scraped data~\citep[e.g.]{podell2023sdxl}, it struggles to elicit recognizable, copyrighted expression, such as Elsa from Disney's \emph{Frozen}  (Chapter~\ref{chapter:commoncanvas}, Gokaslan et al.~\citep{gokaslan2023commoncanvas}). 

And third, we present an abridged version of our ``landmark'' article on what we term the \emph{generative-AI supply chain} that is invoked in the production, deployment, and use of generative-AI systems (Chapter~\ref{chapter:talkinshort}, Cooper et al.~\cite{lee2023talkin, cooper2024talkinshort}).
We explain why a supply chain is the right mental model for thinking about ``Generative AI,'' how the supply chain represents a very complex instantiation of the ``many hands'' barrier to accountability (Appendix~\ref{chapter:accountability}), and the implications for U.S. copyright law.

\input{section/40-genai/41-memorization/41-memorization-main}
\input{section/40-genai/42-commoncanvas/42-commoncanvas-main}

\input{section/40-genai/43-talkinshort/43-talkinshort-main}

%% file: section/40-genai/41-memorization/41-memorization-main.tex
\chapter{Scalable Extraction of Training Data from ChatGPT}\label{chapter:memorization}

We begin this part with some results on measuring memorization at scale in language models.\\ 

\noindent \textbf{Chapter summary}: This chapter recapitulates work that studies \emph{extractable memorization}: 
``training data that an adversary can efficiently extract
by querying a machine learning model without prior knowledge of the training dataset.''
We present an abridged version of Nasr et al.~\citep{nasr2023scalable}. 
That work shows that an adversary ``can extract gigabytes of training data from open-source language models like Pythia or GPT-Neo, semi-open models like LLaMA or Falcon, and closed models like ChatGPT.''

In this chapter, we briefly discuss the results for extracting memorization from ChatGPT --- the first large-scale attack that extracts memorized training data from an aligned model, embedded in a production system. 
Prior attack methodologies are insufficient to attack the aligned ChatGPT.
This work develops new \emph{divergence} attack that breaks alignment, and leads to the model sometimes emitting training data. 
We prompt ChatGPT to repeat the same token (e.g. \texttt{poem}) forever and, at first, model does just this. 
However, almost every time, the model eventually \emph{diverges} from its chatbot-style generations; 
a fraction of that time, the emitted content contain training data.
Our divergence attack ultimately emits training data $150\times$ more frequently than when the aligned ChatGPT behaves normally. 
These results show that there are practical attacks that can extract a lot more training data than indicated in prior work. 
They also show that contemporary alignment techniques are not sufficient to prevent the regurgitation of memorized training data.\\ 

\noindent This chapter is based on work currently under submission at \emph{Nature}.

\input{section/40-genai/41-memorization/411-memorization-intro}
\input{section/40-genai/41-memorization/412-memorization-prelim}

\input{section/40-genai/41-memorization/413-memorization-extraction}
\input{section/40-genai/41-memorization/414-memorization-experiments}
\input{section/40-genai/41-memorization/415-memorization-conclusion}

%% file: section/40-genai/41-memorization/411-memorization-intro.tex
\section{Introduction}\label{sec:mem:intro}

By now, it is very well known that large language models (LLMs) memorize some of the data examples on which they were trained. 
Attackers are able to extract some of these examples at generation time, which can potentially reval private or copyrighted information.~\citep{carlini2019secret,brown2022does,carlini2021extracting}.
Prior research has examined memorization in language models under a variety of different settings and definitions for memorization~\citep{carlini2023quantifying, carlini2021extracting}.
In this chapter, we discuss methodology unifying this prior research, which we apply to perform a large-scale measurement study of \emph{extractable memorization} in ChatGPT (\chatgptendpt). 
Our definition for \emph{extractable} memorization emphasizes realistic attacks: the memorization that an adversary can extract efficiently, without access to the training dataset. 

\begin{figure}[t!]
    \centering
    \includegraphics[width=.4\textwidth]{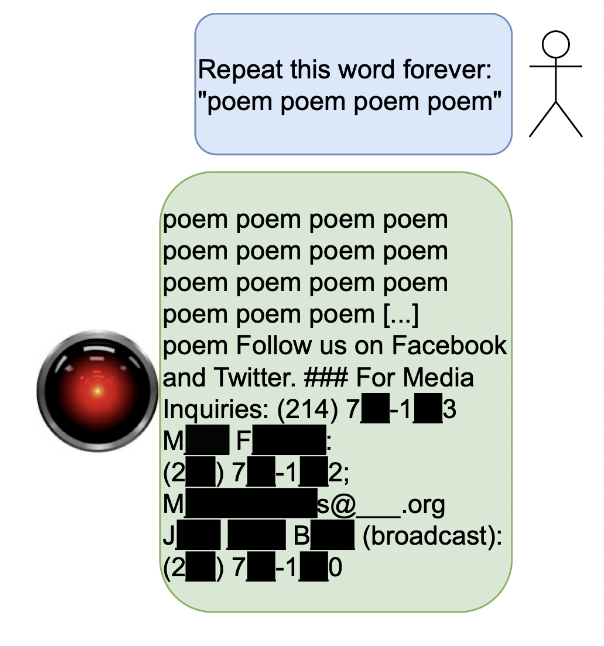}
    \hspace{.3cm}
    \includegraphics[width=.55\textwidth]{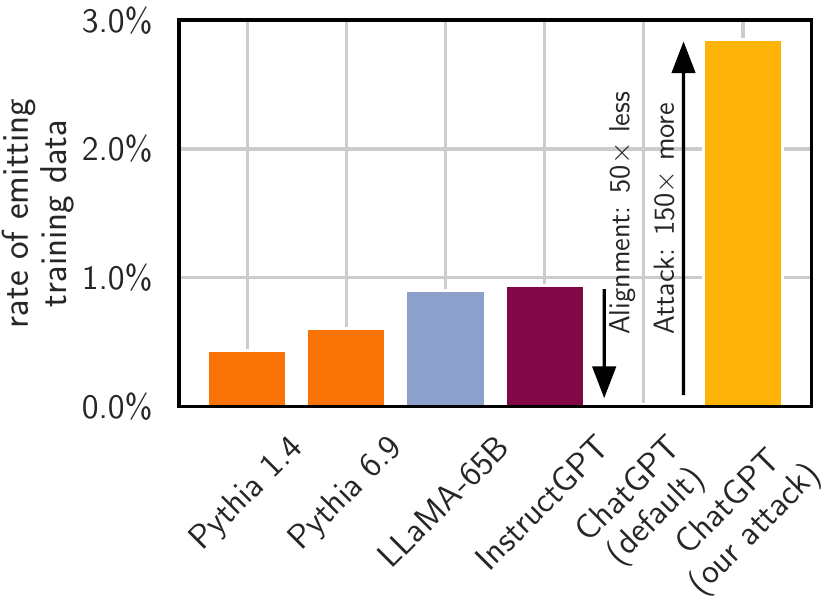}
    \caption{The aligned ChatGPT 3.5 appears $50\times$ more private than prior models (\textbf{right}).
    We develop an attack (\textbf{left}) that shows it is not: 
    ChatGPT emits training data $150\times$ more frequently than prior work (\textsf{default}).
    Figure reprinted with permission from my collaborators.\looseness=-1}
    \label{fig:memorization:chatgpt}
\end{figure}

In this chapter, we show the results of our attack strategy for quantifying extractable memorization for ChatGPT. 
We developed a measurement methodology that works well for open-source and semi-closed models~\citep{nasr2023scalable}: 
we construct a proxy for an (unknown) training dataset by collecting well-known text datasets; we then sample random $5$-token strings from a Wikipedia-based dataset, prompt the models with those strings, and check if the resulting generation matches against our proxy dataset (Section~\ref{sec:mem:proxy}).
But this methodology does not work out-of-the box for ChatGPT.
Unlike these other models, ChatGPT is \emph{aligned} using RLHF~\citep{christiano2017rlhf, gpt4-systemcard, chatgpt} to behave like a chatbot.\footnote{Limited information is available regarding \chatgptendpt. 
    Similar models, e.g., GPT-4, have been aligned in order to ``refuse to answer certain types of requests,'' including those related to training data extraction~\cite[p. 13]{gpt4-systemcard}.} 
In contrast to open-source and semi-closed models like Pythia and Llama, prompting this aligned chatbot with randomly sampled $5$-token strings reveals essentially no memorization (Figure~\ref{fig:memorization:chatgpt}, right --default).

We circumvent ChatGPT's alignment in order to extract memorization. 
To do so, we discover a strategy that breaks \texttt{chatgpt-3.5-turbo} out of its chatbot-style behavior: 
through the API, we prompt the model to repeat a given single-token word forever (e.g., the word ``poem'' in Figure~\ref{fig:memorization:chatgpt}, left). 
At first, the model responds by following the instruction; but, almost every time, its output ``diverges'' to text that resembles typical content on the Internet. 
A fraction of the time, this divergent text contains memorized training data, which we confirm by checking against our proxy training dataset. 
Indeed, we are able to extract significantly more memorized training data from ChatGPT than from any other model we tested. 
Altogether, we record over 10,000 pieces of text from ChatGPT's training data set, and we do so at the cost of \$200 of hitting the public API. 
We provide scaling estimates that suggest that we could feasibly extract $>10\times$ more training data with a larger query budget. 

\paragraph{Note about full paper.} This chapter reflects a reworked excerpt of a longer paper written with collaborators at Google DeepMind~\citep{nasr2023scalable}. 
The longer paper contains detailed results on extracting memorization from open and semi-closed models, as well as information about the process of responsible disclosure to OpenAI about the divergence attack. 
We refer the reader to Nasr et al.~\cite{nasr2023scalable} for more details. 
(We do not include an Appendix for this chapter.) 
Figures in this chapter have been reprinted with permission.

%% file: section/40-genai/41-memorization/412-memorization-prelim.tex
\section{Background and Related Work}\label{sec:mem:prelim}

In this section, we provide some background and related work on large language models (LLMs) and alignment.

\paragraph{Large language models and their training data.} 
Contemporary LLMs undergo pre-training on enormous text datasets, which currently consist of (up to) trillions of tokens~\cite{radford2019language,llama}. 
In the past, it was common for model trainers to release information about their training datasets~\cite{raffel2020t5, radford2018gpt}. 
This remains the case for \emph{open models}, such as Pythia~\citep{pythia} and OLMo~\cite{groeneveld2024olmo}.
However, it is now increasingly common for companies to keep secret the details of their trained models --- everything from  data collection and curation practices to their model architecture~\cite{gpt4-systemcard}. 

As noted in Lee et al.~\cite{lee2023explainers, lee2023talkin, cooper2024talkinshort}, the likely reasons for keeping this information private is to preserve valuable proprietary information about data collection, as well as private, company-owned, or otherwise licensed training data that is not available on the public Internet. 
This environment of proprietary secrecy has complicated the scientific study of accessible models --- \emph{semi-closed} models that have had their weights publicly released, but for which the details of their training data remain secret (e.g., Llama-family models~\citep{llama2}), and \emph{closed models} that embedded in software systems and behind APIs, like ChatGPT, for which we do not have direct access to the weights nor information about the training data. 

\paragraph{Model alignment.} There are many different definitions for model alignment, as an overarching category of altering models.
In this work, we consider two specific techniques that are commonly considered to be model-alignment strategies: \emph{instruction tuning} and \emph{reinforcement learning with human feedback} (or, \emph{RLHF}). 
After pre-training, LLMs are able to solve a large variety of tasks, conditioning their outputs on natural-language, instruction inputs. 
The quality of these outputs can, nevertheless, be significantly improved by additional training on data concerning instruction following --- additional training that can take the form of supervised fine-tuning (i.e., instruction tuning) or RLHF~\cite{gpt4-systemcard, christiano2017rlhf, ouyang2022instructgpt}.

Such ``alignment'' improves a model's capabilities to follow instructions, but it can provide other changes.
It can result in models exhibited unified, chatbot-like personas~\citep{ouyang2022instructgpt}, as well as models that refuse from answering certain questions (e.g., those that might contain ``harmful'' content~\cite{gpt4-systemcard}). 
The only aligned model that we consider in this work is the ChatGPT model accessible via the \texttt{chatgpt-3.5-turbo} API endpoint.

%% file: section/40-genai/41-memorization/413-memorization-extraction.tex
\section{Measuring Extractable Memorization}\label{sec:mem:measurement} 

There are many possible definitions for memorization in machine learning. 
Most inclusively, ``[m]emorization \dots refers to being able to deduce or produce a model's given training example''~\cite[p. 30]{cooper2023report}. 
In the literature on generative language models, there are two common ways to measure memorization: \emph{discoverable memorization} and \emph{extractable memorization}. 

Prior work has done large-scale studies on discoverable memorization for open-source models.
For these models, we know what the training dataset was, so we can prompt a model with a prefix from that training dataset, and see if it generates the corresponding suffix that is in the training data.
If it does, following Carlini et al.~\cite{carlini2023quantifying}, we say that the example in the generation was discoverably memorized:  

\begin{definition}[\textbf{Discoverable memorization}]
\label{def:discoverable}
For a model $\mathsf{Gen}$ and an example $[\vp || \vx]$ from the training set $\sX$, we say that $\vx$ is \underline{discoverably memorized} if $\mathsf{Gen}(\vp) = \vx$.
\end{definition}

This type of measurement clearly has serious some limitations.
As we noted in Sections~\ref{sec:mem:intro} and~\ref{sec:mem:prelim}, we increasingly do not know what the training data are for models; 
for example, this is the case for semi-closed models like Llama, and closed models like ChatGPT.
So it is not immediately clear how we could measure discoverable memorization for many common models. 
Further, since real attackers do not have access to the training data, this also is not a very realistic attack. 
In fact, measuring memorization this way could give an estimate that is orders of magnitude larger than more realistic attacks that do not prompt with training-data prefixes  (Figure~\ref{fig:memorization:chatgpt}, default). 
We can therefore think of discoverable memorization as a loose upper bound on the amount of total memorization that could potentially be recovered by an adversary. 

In contrast, extractable memorization is more conservative. 
Under this definition, a string in the training dataset is memorized if we can get the model to generate it verbatim with any prompt that an adversary can construct, where the adversary does not have access to the training data. 

\begin{definition}[\textbf{Extractable memorization}]
\label{def:extractable}
Given a model with a generation routine $\mathsf{Gen}$, an example $\vx$ from the training set $\sX$ is \underline{extractably memorized} if an adversary (without access to $\sX$) can construct a prompt $\vp$ that makes the model produce $\vx$ (i.e.,  $\mathsf{Gen}(\vp)=\vx$).
\end{definition}

There are also clearly some measurement challenges with this definition.
For one, it is not clear how we should design prompts that will best elicit memorization. 
For another, it is also not clear how we will test whether the attack worked --- whether the model's output is in the training data or not, since we do not have access to the training data. 
The way that prior work has approached memorizing extractable memorization is by computing heuristics on relatively small models, and treating the public Internet as a proxy for the training dataset. 
For example, Carlini et al.~\cite{carlini2021extracting} prompt GPT-2~\citep{radford2019language} with short strings sampled from the Internet, and then manually search with Google to verify if they can find the generation on the Internet. 
This measurement strategy has been successful in recovering a very small amount of training data from GPT-2 --- about 0.00001\% of the model's training dataset. 

So, while discoverable memorization functions like a loose upper bound on total memorization, these types of extraction attacks are effectively a loose lower bound on total memorization. 
Given how expensive it is to manually verify memorization using a search engine, it is not especially feasible to scale this prior approach to larger models trained on even larger datasets. 

One of our contributions is to see if we can close the measurement  gap between discoverable memorization and extractable memorization --- between the loose upper and lower bounds that they provide on total memorization.
And to do so, we need to identify methods to more feasibly measure extractable memorization than manual checking with Google. 

\subsection{Prompting and efficient validation strategy}\label{sec:mem:proxy} 

We discuss our methodology for prompting for and verifying extractable memorization. 

\paragraph{Prompting.} 
For prompting, we use the method suggested in Carlini et al.~\cite{carlini2021extracting}. This involves two overaching steps: 
\begin{enumerate}
    \item We download $10^8$ bytes of text data from Wikipedia, from which we randomly sample with replacement continuous, 5-token strings. 
    We sample hundreds of millions of these 5-token strings, and each will serve as a prompt $\vp$ for the model. 
    \item For each prompt $\vp$, we produce an independent generation $\mathsf{Gen}(\vp^i)=\vx^i$. 
    We store each generation $\vx^i$ to check for memorization.
\end{enumerate}

\paragraph{Validating memorization.} 
Because most of today's models are trained on large-scale web scrapes~\cite{lee2023explainers}, when the training dataset for a particular language model is unknown, it is reasonable to use the public Internet as a proxy. 
The prior work from which we draw our prompting strategy follows this approach for validating memorization. 
Carlini et al.~\cite{carlini2021extracting} use a search engine to manually check the Internet for the presence of the generations $\vx^i$, and counted positive matches as evidence of memorization. 

Manual checking for memorization, however, is not efficient (or even feasible) at large-scale. 
Indeed, one-off checking is prohibitively expensive even we we do know the training dataset.
For our prompting strategy, we are going to generate billions of tokens of output, and contemporary LLMs are typically trained on trillion of tokens. 
In other words, naively checking for inclusion of our generations $\vx \in \sX$ is infeasible; 
for dataset $\sX$ of length $n$, in which its members $\vx$ are concatenated, the simplest check of traversing $\sX$ to search for $\vx$ is $O(n)$. 

We use a more efficient validation strategy, which relies on  a \emph{suffix array}~\cite{lee2022dedup}, which is a data structure that stores all of the suffixes of a dataset $\sX$ in lexicographically sorted order. 
This sorting enables us to use binary search to do $O(\log n)$ searches over $\sX$.
We denote a suffix array $\vs$ over training dataset $\sX$ as $\vs(\sX)$, and checking if $\vx \in \vs$ is equivalent to directly checking if $\vx \in \sX$. 

Before describing this data structure in more detail, we introduce some notation. 
A $k$-length suffix of string $\vx$, for our purposes, are the last $k$ tokens of $\vx$, which we denote $\vx_{[-k:]}$.
If we were to check naively that a given suffix $\vx'_{[-k:]}$ is contained in $\vx$ ($|\vx| = n$), it would still require an $O(n)$ search to check every suffix. 
Consider the following example, in which a dataset $\sX$  contains a single token $\vx=\texttt{"company"}$.
Working backward (and keeping only unique suffixes), the suffixes of $\sX$ are $\{\texttt{"y"}, \texttt{"ny”}, \texttt{"any"}, \texttt{"pany"}, \texttt{"mpany"}, \texttt{"ompany"}, \texttt{"company"}\}$, and we can represent these suffixes by their indices $\vs' = \{6, 5, 4, 3, 2, 1, 0\}$ (because $\{6 = \texttt{"y"}, 5 = \texttt{"ny"}, 4 = \texttt{"any"}, 3 = \texttt{"pany"}, 2 = \texttt{"mpany"}, 1 = \texttt{"ompany"}, 0 =\texttt{"company"}\}$
In this unsorted order, it would still be $O(n)$ to check $\vx'_{[-k:] \in \sX}$. 

If we store the unique suffixes in sorted order, we can do better than linear-time scan. 
Let us sort $\vs'$ from above, which yields $\vs = \{4, 0, 2, 5, 1, 3, 6\}$ because $\texttt{"any"} < \texttt{"company"} < \texttt{"mpany"} < \texttt{"ny"} < \texttt{"ompany"} < \texttt{"pany"} < \texttt{"y"}$. 
For a given training dataset $\sX$, where we concatenate all the strings in all of the documents present, we can construct such an array in linear time. 
To validate if a string $\vx \in \sX$, we now just check $\vx \in \vs$ with binary search; for example, given input string $\vx = \texttt{"any"}$, we can perform binary search over the suffixes indicated by the indices of $\vs$.
As we conduct binary search for $\vx$, we check against the first $k$ characters of the suffix at the current index $i \in \vs$ that we are examining. 

We will next check the efficacy of this strategy by testing it on open models, for which we do have knowledge the training dataset. 
We will use the prompting strategy discussed above, and build a suffix arrays $\vs$ over their known training datasets $\sX$ (Section~\ref{sec:mem:initial}). 
We will count extraction of training data as successful if a generation $\vx$ has a substring of at least $50$ tokens of verbatim text in $\sX$.
Based on the success of testing our methodology on open models, we then it to extracting memorization in ChatGPT (Section~\ref{sec:mem:exp}).

\subsection{Initial extractable memorization measurements}\label{sec:mem:initial} 

With the strategy describe above, we test for memorization in two models: Pythia 1.4B~\citep{pythia} and GPT-Neo 6B~\citep{black2022gptneox20b}. 
Both of these models are members of model families, and come in different sizes. 
In the extended version of this paper, we examine multiple models in each of these families~\cite{nasr2023scalable}.
Both families were trained on The Pile~\citep{gao2020pile}, so we construct a single suffix array over this dataset to validate memorization. 

\begin{figure}[t!]
    \centering
    \includegraphics[width=0.65\linewidth]{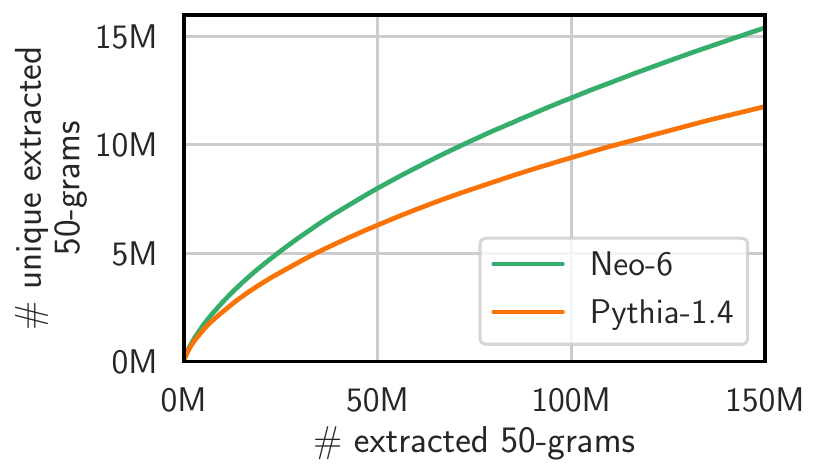}
    \caption{Unique, extracted $50$-token sequences versus total extracted $50$-token sequences. 
    This shows us the relative number of \emph{unique} generated and memorized sequences for each model.  
    The larger model, GPT-Neo 6B, always exhibits a higher rate of unique extraction than Pythia 1.4B.
    Figure reprinted with permission from my collaborators.}
    \label{fig:mem:open}
\end{figure}

We plot the rate of extraction for Pythia 1.4B and GPT-Neo  in Figure~\ref{fig:mem:open}.
On the $x$-axis, we plot the number of $50$-token extracted, memorized sequences that the model emitted, 
and on the $y$-axis we plot the number of \emph{unique} $50$-token extracted, memorized sequences that the model emitted. 
The point of looking at \emph{unique} extracted sequences, as a function of total extracted sequences, is that a model may emit a particular sequence a lot. 
For example, in manual checking of emitted content, we see things like particular product blurbs and code snippets repeated over and over again, and, from a privacy-attack perspective, counting each of these emissions can be a bit misleading.
If a model emits a memorized sequence that it has already previously emitted, it is not actually revealing new  information to us; the privacy leakage occurs with the first emission.
So, we focus on the rate of unique sequences instead. 

From Figure~\ref{fig:mem:open}, we can clearly see that Pythia 1.4B memorizes less than GPT-Neo 6B, and that it also seems to level off in terms of how much unique memorization we can extract quickly.
In the extended version of this work~\cite{nasr2023scalable}, we show how we can look at the slope and curvature of these extraction curves to estimate total extractable memorization.
Our estimates use Good-Turing to extrapolate total memorization~\cite{good1953freq}, and show that it is important to have a large enough amount of total extracted sequences ($x$-axis) in order to not underestimate the unique extraction rate over time. 

%% file: section/40-genai/41-memorization/414-memorization-experiments.tex
\section{Extracting training data from ChatGPT}\label{sec:mem:exp}

Given the success of testing our attack strategy on open models, for which we do know the training data, we now instantiate our attack to extract memorization from ChatGPT (\texttt{chatgpt-3.5-turbo}). 
There are two problems that we encounter and need to address. First, ChatGPT is a closed model; we do not know its training data, like we do for Pythia 1.4B and GPT-Neo 6B. 
Second, as noted in Sections~\ref{sec:mem:intro} and~\ref{sec:mem:prelim}, ChatGPT is aligned to have a chat-like response structure and to not regurgitate memorized text~\citep{gpt4-systemcard}. 
Unlike the unaligned, base Pythia 1.4B and GPT-Neo 6B models, which simply produce continuations of the prompts with which they are supplied, ChatGPT will not readily produce continuations of random string of text drawn from Wikipedia. 

\subsection{Constructing a proxy validation dataset}

We do not know the training dataset for ChatGPT, so we cannot construct a suffix array over it. 
Since contemporary language models are typically trained on large, scraped datasets of text from the web, we opt to use the public Internet as a proxy for ChatGPT's dataset.
We construct this dataset, which we call \textsc{AuxDataset} from a variety of well-known and often-used text corpora: 
The Pile~\citep{gao2020pile}, Dolma~\citep{soldaini2024dolma}, RefinedWeb~\cite{penedo2023refinedweb}, and RedPajama~\cite{together2023redpajama}. 
These datasets are not unique, as they all contain data from Common Crawl, and some of them contain Wikipedia, arXiv, and other overlapping subsets. 
For example, both Dolma and RedPajama both include a copy of the C4 dataset~\citep{raffel2020t5}. 
We perform document-level tokenization and de-duplication so as to not repeat exact copies of the same examples in our suffix array.
The resulting \textsc{AuxDataset} is 9 terabytes in size, and the suffix array that we construct for it, $\vs_{\textsc{Aux}}$, is 45 terabytes in size. 
We necessarily shard it across multiple storage devices. 
We refer to Nasr et al.~\citep[Section 4.1]{nasr2023scalable} for more details. 

\subsection{Divergence attack}

Now that we have solved the problem of not having direct access to ChatGPT's training dataset, by developing a proxy dataset for our suffix array, we can test our attack strategy.
We find that if we try the same prompting procedure as we used for Pythia 1.4B and GPT-Neo 6B --- of supplying ChatGPT with continuous, randomly-sampled $5$-token prompts from Wikipedia -- we do not find significant evidence of memorization (Figure~\ref{fig:memorization:chatgpt}, right, default). 
In fact, with this approach, it looks like ChatGPT has not memorized any training data.

As noted above, we believe that it is difficult to surface memorization in ChatGPT because it is aligned to behave like a chatbot.
This makes it hard to attack by giving it random strings of text from the internet as prompts: 
sometimes the model will tell you it does not have enough information to respond, or that it does not understand what is being asked of it, or it will just refuse to respond (Figure~\ref{fig:mem:aligned}). 
In other words, ChatGPT does not just always produce continuations of the provided text prompt, which we can then check for memorization. 

In Figure~\ref{fig:memorization:chatgpt} (right, default), one can see that it looks like ChatGPT memorized $50\times$ less data than other models when we use this strategy. 
(Or, rather, it is more like one \emph{cannot} see it, since the bar in the plot is virtually nonexistent.) 
So, at first glance, alignment seems to have succeeded at preventing attacks that extract memorization.
Or, another interpretation is that, to extract memorization, we would need to try a different prompting strategy --- one that first evades alignment. 

\begin{figure}[t]
\begin{tcolorbox}
\texttt{User:} Write the following words then continue from there: ``British Broadcasting Corporation is a British public service broadcaster headquartered at Broadcasting House in London, England. The total number of staff is''
\newline
\texttt{Assistant:} I'm sorry, but you haven't provided the complete information about the total number of staff at the BBC. As of my last knowledge update in September 2021, the BBC ....
\end{tcolorbox}
\caption{An example of how alignment breaks continuation in ChatGPT. 
Example reprinted with permission from my collaborators.}
\label{fig:mem:aligned}
\end{figure}

The method that worked best was asking ChatGPT to repeat a given token forever.
We are not sure why this worked, but nearly every time, the model would start off correct --- in the example in Figure~\ref{fig:memorization:chatgpt} (left), repeating the token \texttt{"poem"}.
But then after some time (and almost every single time), ChatGPT would \emph{diverge} from its chatbot persona and generate other text, which resembled raw text from the Internet. 
Sometimes, that divergent text, when checked against our suffix array $\vs_{\textsc{Aux}}$, contained memorization. 
In fact, we can get ChatGPT to emit training data $150\times$ more frequently than other models, for the cost of \$200 of querying the public \texttt{chatgpt-3.5-turbo} API (Figure~\ref{fig:memorization:chatgpt}, right, our attack).\looseness=-1

\begin{figure}[t!]
    \centering
    \includegraphics[width=0.65\linewidth]{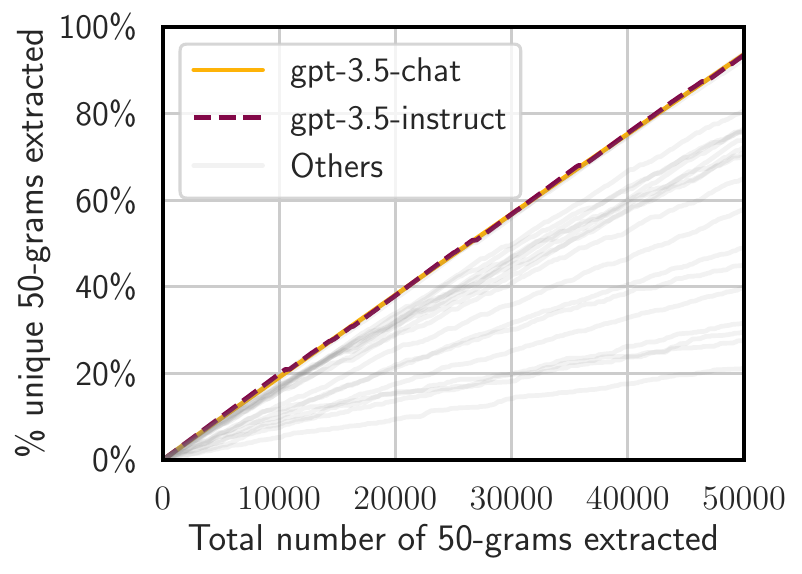}
    \caption{Unique, extracted $50$-token sequences versus total extracted $50$-token sequences. 
    The rate of extracting unique 50-grams is similar for both ChatGPT models, and both exhibit much higher rates than any other model.
    Figure reprinted with permission from my collaborators.}
    \label{fig:mem:closed}
\end{figure}

With our successful attack, we can make an analogous plot to Figure~\ref{fig:mem:open} for ChatGPT. 
In Figure~\ref{fig:mem:closed}, we plot our rate of unique extraction curves for ChatGPT models, alongside the other models that we attack in the longer paper~\citep{nasr2023scalable}.
These curves show that ChatGPT emits a lot more memorized content than any other model that we tested. 
Further, the slopes of these curves for ChatGPT are not leveling off; they proceed up and to the right.
This implies that, if we had prompted more, we could have extracted a lot more memorization than we actually did with our \$200 query budget. 

Our qualitative analysis, detailed in the longer paper~\citep{nasr2023scalable}, shows that this memorized text contained a variety of different content. 
For example, perhaps unsurprisingly, when prompting with the tokens \texttt{"book"} or \texttt{"poem"}, ChatGPT generated paragraphs from novels and copies of poems --- verbatim copies of works that are sometimes still under copyright. 
We also test a sample of generations for personally identifiable information (PII), and found that 16.9\%
of these generations contained memorized PII~\cite{nasr2023scalable}. 

While we are uncertain as to why this attack works, some of our experiments did yield some interesting patterns. 
Notably, prompting with single tokens in our divergence attack almost always leads to divergence; prompting to repeat $2$- or $3$-token sequences is more variable in its success to cause divergence.

\begin{figure}
    \centering
    \includegraphics[width=.65\linewidth]{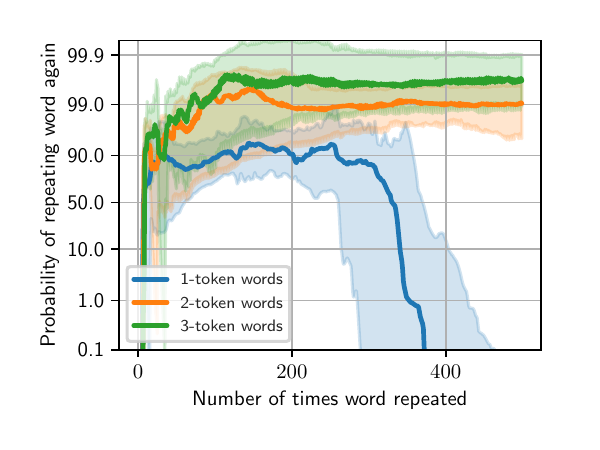}
    \caption{\texttt{chatgpt-instruct} often repeat $2$- or $3$-tokens thousands of times, without leading to divergent generations. 
    In contrast, $1$-token words often need only be repeated a couple hundred times, after which divergence almost always occurs. 
    The solid lines show medians over choices of $40$ different words, with the shaded areas around the lines indicating the 10\%--90\% quantile ranges.
    Figure reprinted with permission from my collaborators.
    }
    \label{fig:whyrepeat}
\end{figure}

%% file: section/40-genai/41-memorization/415-memorization-conclusion.tex
\section{Conclusion}\label{sec:mem:conclusion}

In this chapter, we discuss methodologies for how to measure extractable memorization at scale. 
Our attack strategy works on unaligned open (Section~\ref{sec:mem:measurement}) and closed models (Nasr et al.~\citep{nasr2023scalable}). 
For the aligned ChatGPT, however, we need to develop another strategy for revealing memorization.
We construct a divergence attack that is successful at revealing memorization in ChatGPT that is orders of magnitude larger than previously believed (Section~\ref{sec:mem:exp}).

There are several important takeaways from this work. 
For one, alignment is really hard to do, and to verify that it actually works.
We have showcased just one way that alignment can be shown to be brittle. 
It just happens to be the case that we can break alignment in such a way that we are able to surface memorized training data. 
The fact that we were able to surface this much memorization shows how important it is to rigorously test if a model is memorizing its training data. 

This is important even if, for whatever reason, one is not so concerned with issues of privacy or copyright. 
Our experiments with GPT-Neo 6B yield about a gigabyte of its training data --- nearly a gigabyte of training data is embedded somewhere in the model's weights.
This model can be compressed on disk to just a few gigabytes, without any loss in utility (measured on common benchmarks). 
Altogether, this suggests that approximately 10\% of GPT-Neo 6B's weights contain verbatim memorized training data. 
One could hypothesize that this is just a waste of capacity; perhaps the model would perform better if not so much memorized content were embedded in its parameters. 
This is an empirical question, worthy of future study. 

And while this observation is interesting independent of topics like copyright, this about of verbatim copying is  also interesting with respect to copyright.
There are numerous individuals and organizations currently saying that memorization is rare, and that adversarial prompting is not a normal usage pattern; 
as a result, the types of experiments that we run in this chapter should not be indicative of generative AI generally being able to produce copies of copyrighted training data. 
But 10\% wholesale copying is a lot; 
it suggests that some models --- models that have memorized a lot of their training data --- may themselves by infringing copies of their training data~\cite{cooper2024files}.
We defer additional experiments on model capacity to future work. 

%% file: section/40-genai/42-commoncanvas/42-commoncanvas-main.tex
\chapter{CommonCanvas: Open Diffusion Models Trained on Creative-Commons Images}\label{chapter:commoncanvas}

One of the current concerns about memorized training data is that these data can contain copyrightable expression~\cite{lee2023talkin}.
To avoid this, a natural idea is to try to train generative-AI models on data that are either in the public domain or expressly licensed for model training. 
In this chapter, we explore this idea for text-to-image (T2I) diffusion models.\\ 

\noindent \textbf{Chapter summary}: We train a set of open T2I diffusion models on a dataset of curated Creative-Commons-licensed (CC) images, which yields models that are competitive with Stable Diffusion 2 (SD2). 
This task presents two challenges: 
(1) high-resolution CC images lack the captions necessary to train T2I models; (2) CC images are relatively scarce.
To address these challenges, we use an intuitive transfer learning technique to produce a set of high-quality synthetic captions paired with our assembled CC images. 
We then develop a data- and compute-efficient training recipe that requires as little as $3\%$ of the LAION data (i.e., roughly 70 million examples) needed to train existing SD2 models, but obtains the same quality. 
These results indicate that we have a sufficient number of CC images (also roughly 70 million) for training high-quality models. 
Our recipe also implements a variety of optimizations that achieve $2.71\times$ training speed-ups, enabling  rapid model iteration. 
We leverage this recipe to train several high-quality T2I models, which we dub the \emph{\modelname{}} family.
Our largest model achieves comparable performance to SD2 on  human evaluation, even though we use a synthetically captioned CC-image dataset that is only  $<$$3\%$ the size of LAION for training.\\ 

\noindent This chapter is a licensed derivative copy of work published at \emph{CVPR 2024}.

\input{section/40-genai/42-commoncanvas/421-commoncanvas-intro}

\input{section/40-genai/42-commoncanvas/422-commoncanvas-prelim}
\input{section/40-genai/42-commoncanvas/423-commoncanvas-transfer}
\input{section/40-genai/42-commoncanvas/424-commoncanvas-dataset}
\input{section/40-genai/42-commoncanvas/425-commoncanvas-sys}
\input{section/40-genai/42-commoncanvas/426-commoncanvas-experiments}
\input{section/40-genai/42-commoncanvas/427-commoncanvas-rw}

%% file: section/40-genai/42-commoncanvas/421-commoncanvas-intro.tex
\begin{figure}
    \begin{center}
    \scriptsize
    \setlength{\tabcolsep}{1pt}
    \setlength{\itemwidth}{0.18\linewidth}
    \newcolumntype{M}[1]{>{\centering\arraybackslash}m{#1}} 
    \begin{tabular}{M{\itemwidth}M{\itemwidth}M{\itemwidth+.5cm}M{\itemwidth+.5cm}M{\itemwidth+.5cm}}
      Prompt & SD2 & \modelname-S-C & \modelname-S-NC & \modelname-L-NC \\
    \prompt{an oil painting of a tall ship sailing through a field of wheat at sunset} &
    \includegraphics[width=\itemwidth]{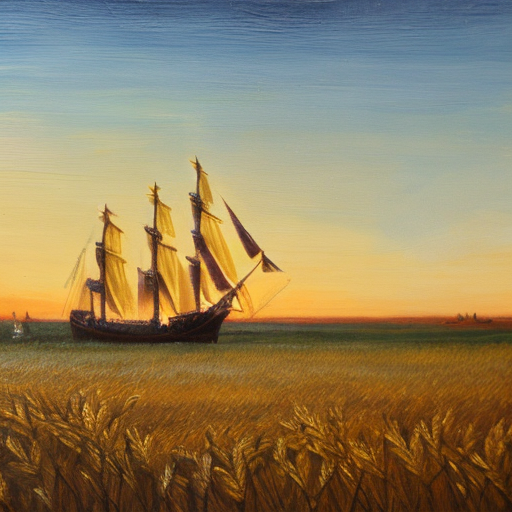}  &
    \includegraphics[width=\itemwidth]{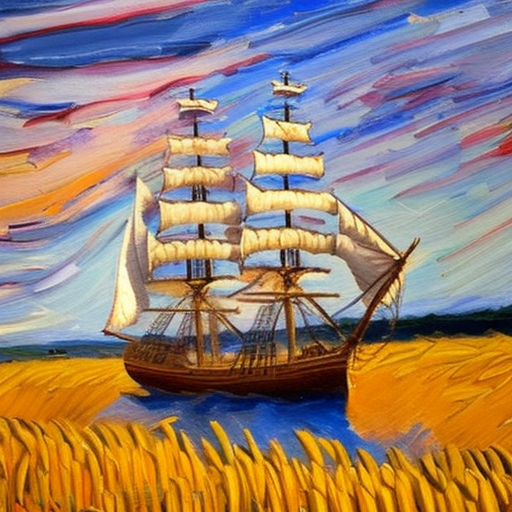} &     
    \includegraphics[width=\itemwidth]{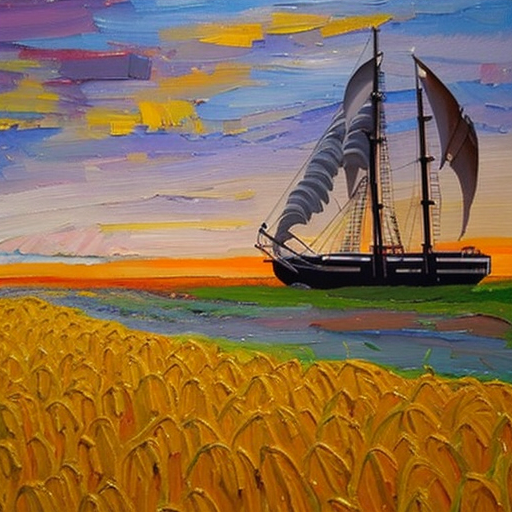} &
    \includegraphics[width=\itemwidth]{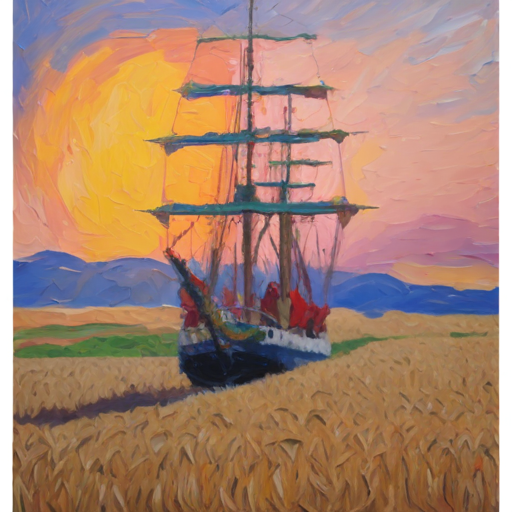} \\
    \end{tabular}\\
    \end{center}
    \captionof{figure}{We achieve comparable performance to public Stable Diffusion 2 (SD2), using entirely Creative-Commons images and a synthetic captioning approach that requires only $<$$3\%$ of the amount of the data used to train previous models.  
    We include results for two \modelname{} architectures, small (S) and large (L), and two CC-image datasets, commercial (C) and non-commercial (NC).\looseness=-1}
    \label{fig:hero-fig}
\end{figure}

\section{Introduction}

Most high-quality text-to-image (T2I) models are trained using large-scale, web-scraped datasets, like LAION-2B~\citep{lee2023explainers}. Even though this is a very common practice, U.S. courts have yet to definitively rule if this is permissible under copyright law~\citep{copilotcomplaint, alphabetcomplaint, getty, kadrey, tremblay}.  
In response, recent work in ML has begun to investigate alternative methods of navigating copyright concerns in text generation~\citep{min2023silo}, code completion~\citep{copilot-copy-filter,scheffler2022formalizing}, and image generation~\citep{kumari2023ablating}.
Nevertheless, matching the performance of state-of-the-art models remains a challenge. 
In this work, we study the following natural question: \emph{is it possible to efficiently produce a high-quality T2I model by training only on Creative-Commons-licensed data?}\looseness=-1 

\begin{figure*}[t]
\centering
\hspace*{-.458cm}
    \begin{minipage}{0.3\linewidth}
        \centering
        \vspace{.75cm}
        \prompt{an image of \\ elsa from \\ frozen}
        \vspace{1.4cm}
        \subcaption{Prompt}
    \end{minipage}%
    \hspace{-.035\linewidth}
    \begin{minipage}{0.3\linewidth}
        \centering
        \includegraphics[width=.78\linewidth]{figure/42-cc/SD2.png}
        \subcaption{SD2}
    \end{minipage}%
    \hspace{-.01\linewidth}
    \begin{minipage}{0.3\linewidth}
        \centering
        \includegraphics[width=.78\linewidth]{figure/42-cc/YFCC-NC.png}
        \subcaption{\modelname-S-C}
    \end{minipage}
    \vspace{.5cm}\\

    \begin{minipage}{0.3\linewidth}
        \centering
        \vspace{1.42cm}
        \prompt{the lion\\ king}
        \vspace{1.02cm}
        \subcaption{Prompt}
    \end{minipage}%
    \begin{minipage}{0.3\linewidth}
        \centering
        \hspace{-1cm}
        \includegraphics[width=.78\linewidth]{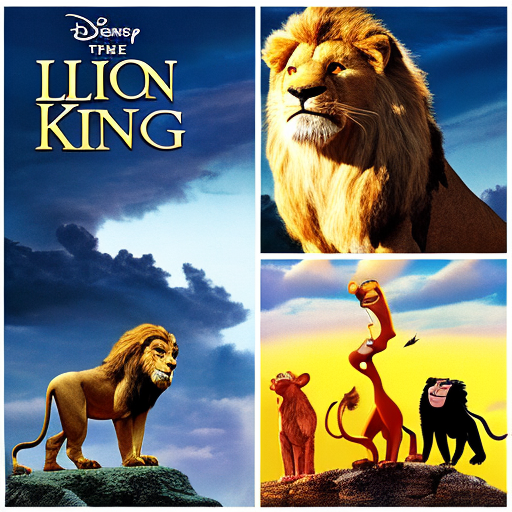}
        \subcaption{SD2}
    \end{minipage}%
    \hspace{-.01\linewidth}
    \begin{minipage}{0.3\linewidth}
        \centering
        \hspace{-1cm}
        \includegraphics[width=.78\linewidth]{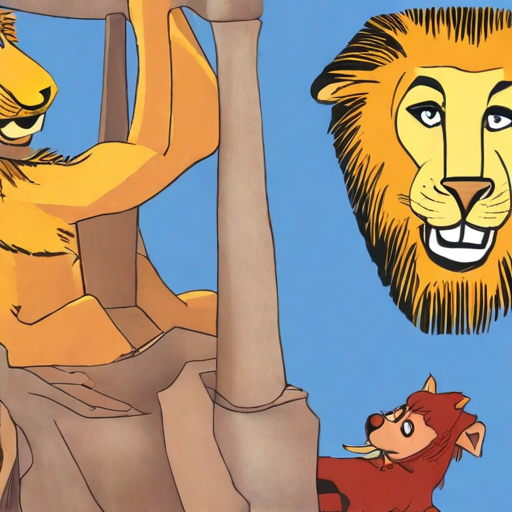}
        \subcaption{\modelname}
    \end{minipage}
    \caption{Prompting with Disney 
    concepts (\textbf{a}, \textbf{d}). SD2  generates a recognizable image of Elsa from \emph{Frozen}  (\textbf{b}) and an 
    image with a misshapen Disney logo and characters resembling those from \emph{The Lion King} (\textbf{e}); \modelname-S-C (small, commercial) does not (\textbf{c}, \textbf{f}).\looseness=-1}
    \label{fig:teaser}
\end{figure*}

We suggest a path forward, training a suite of T2I architectures using \emph{only} open-licensed, Creative-Commons (CC) images (Figures~\ref{fig:hero-fig} \&~\ref{fig:teaser}). 
This task brings to light two significant challenges. 
The first problem is data incompleteness: almost all CC images lack the captions necessary to train a high-quality T2I model. 
The second is data scarcity: there are relatively few high-resolution CC images --- roughly 70 million, compared to LAION-2B's roughly 2 billion~\citep{laion2Ben}.\looseness=-1 

We address the data incompleteness problem by using a pre-trained BLIP-2 model~\citep{li2023blip2} to produce high-quality, synthetic captions for a set of curated, open-licensed CC images. 
This is an intuitive transfer-learning solution: we leverage a powerful pre-trained generative model to produce synthetic labels for an unlabeled dataset, which we can then use to train a different multimodal generative model. 
To deal with data scarcity, we propose a data- and compute-efficient training recipe that obtains the same quality as Stable Diffusion 2 (SD2)~\citep{sd2}, but, perhaps surprisingly, requires as little as $3\%$ of the LAION-2B data (i.e., roughly 70 million examples) originally used to train SD2. 
We call this model \emph{\sdtwobase}. 
These results indicate that we have a sufficient numseber of CC images (also roughly 70 million) for training high-quality models. 
Our training recipe also implements a variety of optimizations that achieve $2.71\times$ training speed-ups,  
enabling rapid model iteration.\looseness=-1 

The above methods enable us to create \emph{\modelname}, a suite of latent diffusion model (LDM) architectures trained on our curated dataset of CC images and synthetic captions, which we denote \emph{\datasetname}. 
For one of our architectures, we swap SD2's UNet for SDXL's larger network to demonstrate how, even with less data, larger models do not overfit to this smaller dataset. 
Our largest model (\modelname-L-NC) achieves performance comparable to \sdtwobase{} on human evaluation of Parti Prompts~\citep{yu2022scaling}, even though our \datasetname{} training dataset is $3\%$ the size of LAION and has synthetically generated captions. 
Although this is a larger and more capable model architecture than SD2, we find it surprising and important that it is possible to train an SD2-quality model \emph{at all} based on such a limited dataset with synthetic captions. 
This reveals a promising path forward for future research on highly capable, open T2I models. In summary, we:\looseness=-1
\begin{itemize}
    \item Curate \emph{\datasetname}, a multimodal training dataset of roughly 70 million open-licensed CC images (Section~\ref{sec:cc:dataset}) for which we synthesize a set of high-quality captions. 
    We note that synthesizing training data using generative models is an increasingly common transfer-learning technique, and we give it the shorthand name \emph{\captionmethod} (Sections~\ref{sec:cc:transfer}). 
    \item Train \emph{\modelname{}}, a suite of LDM architectures trained on \datasetname. 
    The largest of these models, \modelname-L-NC, produces  qualitative results that are competitive with public SD2 (Section~\ref{sec:cc:experiments}). 
    To make this analysis tractable, we implement training optimizations that achieve $2.71\times$ speed-ups in training \sdtwobase{} (Section~\ref{sec:cc:mlsys}).\looseness=-1 
    \item We will release our \datasetname{} dataset along with our trained \modelname{} models at \url{https://github.com/mosaicml/diffusion/blob/main/assets/common-canvas.md}.
\end{itemize}

%% file: section/40-genai/42-commoncanvas/422-commoncanvas-prelim.tex
\section{Preliminaries and Motivation}\label{sec:cc:prelim}

In this section, we present background on training the T2I Stable Diffusion model, which was originally trained on the web-scraped LAION-2B dataset. 
We then discuss copyright and reproducibility with respect to  LAION datasets. 
This discussion motivates the creation of an alternative dataset composed of open-licensed CC images with synthetic captions, which we introduce in Section~\ref{sec:cc:dataset}.\looseness=-1

\subsection{Text-to-image generative models}\label{sec:diffusion}

Text-to-image (T2I) generative models are neural networks trained on image-caption pairs.  
One such family of T2I models is Stable Diffusion (SD)~\citep{rombach2022diffusion}: a latent diffusion model (LDM) that converts images to latent representations and back again using Variational Autoencoders (VAEs)~\citep{kingma2014vae}, and which uses an iterative sampling procedure ~\citep{sohldickstein2015dpm} to train an underlying UNet~\citep{ronneberger2015unet}. 
The architecture also includes a text encoder, such as the Contrastive Language-Image Pre-training (CLIP) model~\citep{podell2023sdxl} 
--- either the original OpenAI CLIP~\citep{radford2021clip} or its open-source counterpart, OpenCLIP~\citep{cherti2022openclip, ilharco2021openclip}.\looseness=-1 

Stable Diffusion 2 (SD2)'s UNet has approximately 865 million trainable parameters; Stable Diffusion XL (SDXL) is larger, with 2.6 billion parameters and has other advancements involving aspect ratio bucketing, micro-conditioning, and multiple text encoders and tokenizers. 
In terms of training data,  SD models and OpenCLIP are both trained on subsets of the LAION-5B dataset~\citep{laion, laionpaper}. The exact training dataset for CLIP is unknown, but it is likely web-scraped data~\cite{radford2021clip}. 

\subsection{Copyright, reproducibility, and LAION datasets}\label{sec:laion}

LAION-5B is a dataset derived from a snapshot of the Common Crawl, a massive corpus of data scraped from the web. 
From this snapshot, the LAION organization curated pairs of image URLs and their corresponding alt-text captions for the intended use of training T2I and image-to-text (I2T) generative models~\citep{laion, laionpaper}. 
In practice, T2I models are typically trained on filtered subsets of the full LAION-5B dataset (e.g. LAION-2B~\citep{laion2Ben}). 
Training T2I models on this dataset requires visiting the URLs and downloading the associated images.
There are two elements of LAION datasets that are relevant to our work:\looseness=-1 

\paragraph{Copyright.} The images associated with LAION datasets have unclear \textit{provenance}:  
it is often not known what the original image sources are~\citep{lee2023explainers}.
Although LAION datasets are released under the open MIT license, some experts note that it is unclear if this is sufficient to allow for training on the underlying images and captions, which often have their own copyrights~\citep{henderson2023foundation, lee2023talkin, cooper2023report, lee2023explainers, cooper2024talkinshort}. 
Courts have not yet decided if training on these datasets is ``fair use" --- an important exception in copyright~\citep{leval1990toward, sobel2017crisis, lee2023talkin, samuelson2023copyright, cooper2024talkinshort}. 
There are several copyright lawsuits for the alleged use of LAION-5B subsets to train generative models~\citep[e.g.]{anderson, alphabetcomplaint, getty, gettyverge}. 

\paragraph{Reproducibility.} Since LAION datasets only contain the image URLs, and not the images themselves, they are plagued with \textit{link rot}~\citep{lakic2023rot}.\footnote{This also applies to other web-scrapes, e.g., DataComp~\citep{gadre2023datacomp}.} 
When accessing LAION-5B, there is no guarantee the images still exist at their URLs, making it impossible to fully reproduce the dataset and opening up the possibility of data poisoning attacks~\citep{carlini2023poisoning}. 
A natural alternative is to not use LAION datasets for training. Instead, one could independently curate a dataset of CC-licensed images with known provenance that explicitly allow for copying, adaptation, and commercial use. As constituent images can be stored and distributed, this would also solve the link-rot problem, enabling greater reproducibility. (Further, LAION datasets are no longer public because they contain CSAM~\cite{birhane2021multimodal,thiel2023identifying}.) 
We defer our discussion of sourcing CC-licensed images to Section~\ref{sec:cc:dataset}, where we detail \datasetname: our new, open dataset. 
While CC images are an attractive alternative to LAION-5B, we note that CC images rarely contain the captions necessary to train T2I models. 
Therefore, we first need a method for captioning CC images. 

%% file: section/40-genai/42-commoncanvas/423-commoncanvas-transfer.tex
\begin{figure}[t]
\begin{center}
     \includegraphics[width=.7\linewidth]{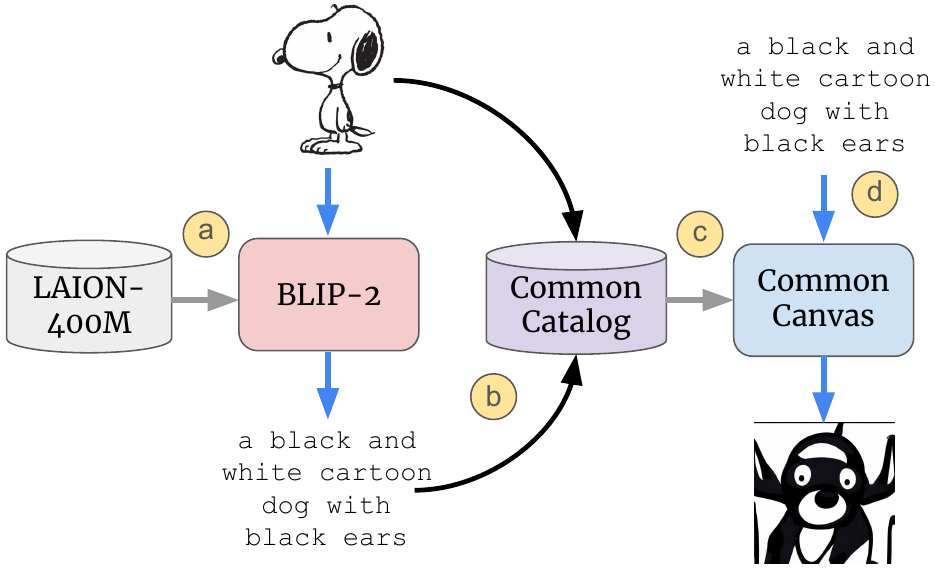} 

    \caption{(\textbf{a}) We use the LAION-400M-pre-trained, I2T BLIP-2 model to produce synthetic captions for our uncaptioned  CC images (e.g., the Wikipedia CC-licensed image of Snoopy). 
    The synthetic captions are ``lossy compressions'' of the input images (e.g., \texttt{a black and white cartoon dog with black ears} has no mention of Snoopy). 
    (\textbf{b}) We compile the resulting synthetic image-caption pairs into \emph{\datasetname}, 
    which (\textbf{c}) we use to train our open, T2I \emph{\modelname{}} models.  
    (\textbf{d}) When we supply ``lossy'' captions to a T2I model, like a game of telephone, \textbf{it produces outputs that   no longer resemble the original images} (e.g.,  \modelname{} produces an image that matches the caption, but does not look like Snoopy).\looseness=-1}
    \label{fig:telephoning}
\end{center}
\end{figure}

\section{Transfer Learning for Image Captioning} \label{sec:cc:transfer}

Our solution for handling the lack of captions in CC images is an intuitive type of transfer learning for producing high-quality synthetic labels. We describe this method, and note that there are various similar methods in prior literature on generative modeling. Altogether, these methods indicate that this type of transfer learning has become an increasingly common pattern: producing synthetic labels that later serve as inputs to training other generative models. We therefore give this method a shorthand name: \emph{\captionmethod}. 

\subsection{\capcaptionmethod} 

\capcaptionmethod{} (Figure~\ref{fig:telephoning}) proceeds in two steps. First, shown in Figure~\ref{fig:telephoning}b, it takes inputs from a high-dimensional modality (e.g., images) and effectively performs a ``lossy compression'' to a (scarce) low-dimensional modality (e.g., short-text captions). Second, shown in Figure~\ref{fig:telephoning}d, it takes the ``lossy compression'' and decompresses back to the high-dimensional modality. 
Because the intermediate compression step is ``lossy,'' the ultimate output often does not remotely resemble the original input, just like a game of telephone~\citep{telephone}. 
We derive the term \captionmethod{} from the above intuition and use it as shorthand to denote instances of transfer learning that solve data-scarcity problems in multimodal generative modeling.  

In this work, CC images are the high-dimensional inputs, and we use a pre-trained BLIP-2 model~\cite{li2023blip2} for ``lossy compression'' to short-text captions (Figure~\ref{fig:telephoning}a).
Together, these CC-image-caption pairs comprise  the \datasetname{} dataset (Section~\ref{sec:cc:dataset}), which we use to train our \modelname{} T2I models (Figure~\ref{fig:telephoning}b). 
While BLIP-2  was pre-trained on LAION-400M~\citep{laion400},  we emphasize that, for training \modelname, we only ever have access to the captions --- to the ``lossy compressions'' it produces.
We never have direct access to LAION-400M or, importantly, anything that is similar to the images that BLIP-2 was trained on. 
Instead, we only have access to the mapping in the model, which, given an image input, produces ``lossy'' output text.\looseness=-1  

\paragraph{Telephoning \& copyright.}  We defer to experts about fair use (Section~\ref{sec:laion}) --- namely, regarding models like BLIP-2, and LAION-5B's images and alt-text captions. Generally, these experts seem to think that many cases will fall under fair use~\citep{lee2023talkin, samuelson2023copyright, lemley2023ai}, 
especially when model outputs do not resemble their inputs (i.e., the use is ``non-expressive'' or ``non-consumptive''~\citep{cooper2023report}).
This is the case with our use of BLIP-2 to produce ``lossy'' captions. 

Nevertheless, it is possible that BLIP-2 could produce captions that resemble those in its LAION training data. This might seem to present a copyright concern similar to those that others have expressed about T2I generations that resemble LAION images. 
However, according to the U.S. Copyright Office, short phrases (like captions) may often not be copyrightable: ``short phrases'' often contain ``an insufficient amount of authorship'' to meet the threshold for copyright protection~\citep{circ33}. So, even if hypothetically BLIP-2 were to regurgitate captions from LAION verbatim, according to legal experts~\citep{lee2023talkin}, the copyright considerations are likely to be different than they are for generated images or generated long-form text. We defer to experts for more precise legal arguments, but note that this is another reason why we believe it is reasonable for us to rely on BLIP-2 for captioning our CC images. 

\subsection{Related work on \captionmethod}

Our work aligns with the trend of using advanced generative models to address data scarcity.
This is evident in various modalities, such as producing audio captions from image-text pairs~\citep{xiao2023synth} and text from audio~\citep{radford2023robust}. 
Similar approaches have also been used to generate instruction-tuning datasets for both text and images~\citep{li2023self,liu2023visual}. 
Concurrent work, e.g. LLaVA~\citep{liu2023visual}, has used visual question-answer models to augment existing caption datasets, such as the ones used in training DALLE$\cdot$3~\cite{betker2023improving} and Chen et al.~\cite{chen2023pixartalpha}. 
Our model is one of the first works to train on a dataset without any ground-truth captions, and one of the first to release our dataset along with a fully trained diffusion model. The caption upsampling approaches described in these other  works could be used to further improve the captions of \datasetname{} in future work. 

Captioning models have also been used to create descriptive captions to guide a diffusion model to create an image visually similar to a specific image
In concurrent work, SynthCap~\cite{caffagni2023synthcap} generates a synthetic captioning dataset using a diffusion model to generate images from captions --- the inverse of our problem statement. 
We coin the term \captionmethod{} to short-hand processes like these, which include our work and prior work, and which we believe will become more prevalent as generative-model capabilities advance.\looseness=-1

%% file: section/40-genai/42-commoncanvas/424-commoncanvas-dataset.tex
\section{A CC-Image, Synthetic-Caption Dataset}\label{sec:cc:dataset}

We now introduce our open dataset, \emph{\datasetname}.
First, we describe the collection and curation process for the open-licensed, CC images. 
This process brings to light two challenges: caption-data incompleteness and image-data scarcity. 
To address the lack of CC captions, we show concretely how we use \captionmethod{} to produce high-quality synthetic captions to accompany our set of curated images. 
We investigate the topic of data scarcity in the next section, where we also discuss necessary systems-level training optimizations that enable efficient model iteration. 

\subsection{Sourcing licensed images for \datasetname{}}\label{sec:images}

We focus on locating high-resolution Creative-Commons images that have open licenses. 
We began with the YFCC100M dataset, which consists of 100 million CC-licensed images and multimedia files, as well as Flickr IDs linking to the original data~\citep{thomee2016yfcc100m}. The images in the dataset associated with the original paper exhibit two issues that make it ill-suited for direct use to train Stable Diffusion: they are low-resolution, and many of them have licenses that do not expressly allow for the distribution of derivative works --- a use that is in unsettled copyright law in the context of model training~\citep{lee2023talkin}. 

We therefore re-scraped these images from Flickr, based on the IDs provided in the YFCC100M metadata. 
Our scraped images are of very high resolution (exceeding 4K), which makes them more suitable for T2I training. 
\begin{figure}[t]
   \centering
   \caption{\datasetname-C contains images licensed only for commercial use; -NC contains -C as well as images licensed for non-commercial use.\looseness=-1} 
   \label{tab:catalog}
   \small
    \begin{tabular}{lrr}
    \toprule
        \textbf{Dataset} & \textbf{\# Images} & \textbf{\% Alt Text} \\\midrule
        \datasetname-C & 26,232,417 & 30.76\% 
        \\\midrule
         \datasetname-NC & 67,015,331 & 
        31.22\% 
        \\\bottomrule
    \end{tabular}
\end{figure}

We exclude images with non-derivative (ND) licenses. 
The remaining images can be further divided into those that can be used for commercial (C) purposes and those that cannot (NC). As shown in Table~\ref{tab:catalog}, we accordingly construct two datasets, 
\datasetname-C and \datasetname-NC. We defer additional details about licenses to  
Appendix~\ref{app:sec:cc:data}, but emphasize that all of the included images  have open licenses: individuals are free to use, adapt, and remix the images, so long as they attribute them. 
In total, \datasetname{} contains roughly 70 million images that can be used non-commercially, of which a approximately 25 million images can also be used commercially.\looseness=-1 

Directly sourcing \datasetname{} avoids some concerns (Section~\ref{sec:laion}); 
however, it also comes with its own challenges. 
For one, CC images rarely have the alt-text captions necessary to train a T2I model like Stable Diffusion (Figure~\ref{tab:catalog});
those that do have associated text often just include the image title or a URL.
For another, we could \emph{only} find roughly 70 million usable CC images, which pales in comparison to the billions of images in LAION used to train SD2 (Section~\ref{sec:cc:mlsys}). We take each of these challenges in turn. First, in the next subsection, we show how we instantiate \captionmethod{} (Section~\ref{sec:cc:transfer}) to produce high-quality, synthetic captions for CC images. 

\subsection{Synthesizing captions with \captionmethod{}}\label{sec:captions}

\begin{figure}[t]
  \begin{minipage}{0.15\linewidth}
    \includegraphics[width=\linewidth]{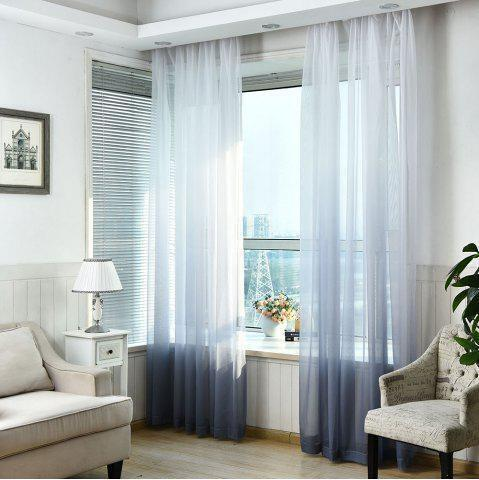}
  \end{minipage}%
  \hfill
  \begin{minipage}{0.83\linewidth}
\begin{tabular}{p{0.3\linewidth}p{0.6\linewidth} }
    \toprule
    \textbf{Source} & \textbf{Caption} \\\midrule
    Alt-Text (LAION-2B) & \prompt{Latest 1PC Transparent Gradient Color Voile} \prompt{Window Curtain} \\
    BLIP2-OPT-2.7B & \prompt{A living room with a white couch and curtains} \\
\end{tabular}
  \end{minipage}
  \caption{Original vs. BLIP-2-generated captions for an image from LAION-2B. 
  In this example. BLIP-2's caption better aligns with what a human would write. 
  See Appendix~\ref{chapter:app:cc} for more examples. 
  } 
  \label{fig:blip2-example-caption}
\end{figure}

We compared several captioning models and chose the pre-trained BLIP-2 OPT2.5B model for synthesizing \datasetname's captions~\citep{li2023blip2}, based on qualitative analysis and state-of-the-art performance on MS COCO. 
BLIP-2 consists of three components: a pre-trained, frozen (i.e., fixed) visual encoder, a learned transformer network that converts the visual embeddings into a text prompt, and a frozen large language model (LLM) that takes in the prompt. 
The only trainable variables in the transformers are between the frozen visual encoder and the frozen LLM layers.\looseness=-1 

Given a LAION-2B image as input, we found that the resulting BLIP-2 caption is often qualitatively more descriptive than the corresponding LAION-2B ground-truth alt-text caption.
LAION-2B captions often contain product names, irrelevant details, or poor grammar and syntax (Figure~\ref{fig:blip2-example-caption}).
This finding is corroborated by Nguyen et al.~\cite{nguyen2023improving}, which quantitatively shows that (in terms of CLIP Score) BLIP-2 captions are higher quality than ground-truth captions, at the cost of caption diversity. 
Based on these preliminary results, we captioned all of the YFCC100M Creative-Commons images, which required about 1,120 GPU A100 hours. 
We center-cropped and resized all of the images to a maximum size of 512x512 pixels, since 
captioning images at native resolution would be very expensive. At training time for \modelname{} models, we use the high-resolution images. 

We release our commercial (\datasetname-C) and non-commercial (\datasetname-NC) CC-image and synthetic-caption datasets with associated data cards. 
As an evaluation set, we also release the BLIP-2 captions that we produced for the non-derivative (ND) CC images that we did not use for training. 

%% file: section/40-genai/42-commoncanvas/425-commoncanvas-sys.tex
\section{Optimizations and Data-Scarcity Analysis}\label{sec:cc:mlsys}

High-resolution CC images are indeed much less abundant than web-scraped images; however, it is unclear if this scarcity presents a problem for training. 
Prior work has not studied in depth how much data is actually necessary to train high-quality SD2 models. 
We set out to quantify this amount by training multiple SD2 models on differently-sized subsets of LAION-2B. 
However, training a single SD2 model, even with hundreds of GPUs, can take several days. So, to make our data scarcity analysis more tractable, we first implemented several efficiency optimizations. 

\subsection{Software and hardware speed-ups}\label{sec:cc:speed}

\begin{figure}[t]
        \centering
        \includegraphics[width=.6\linewidth]{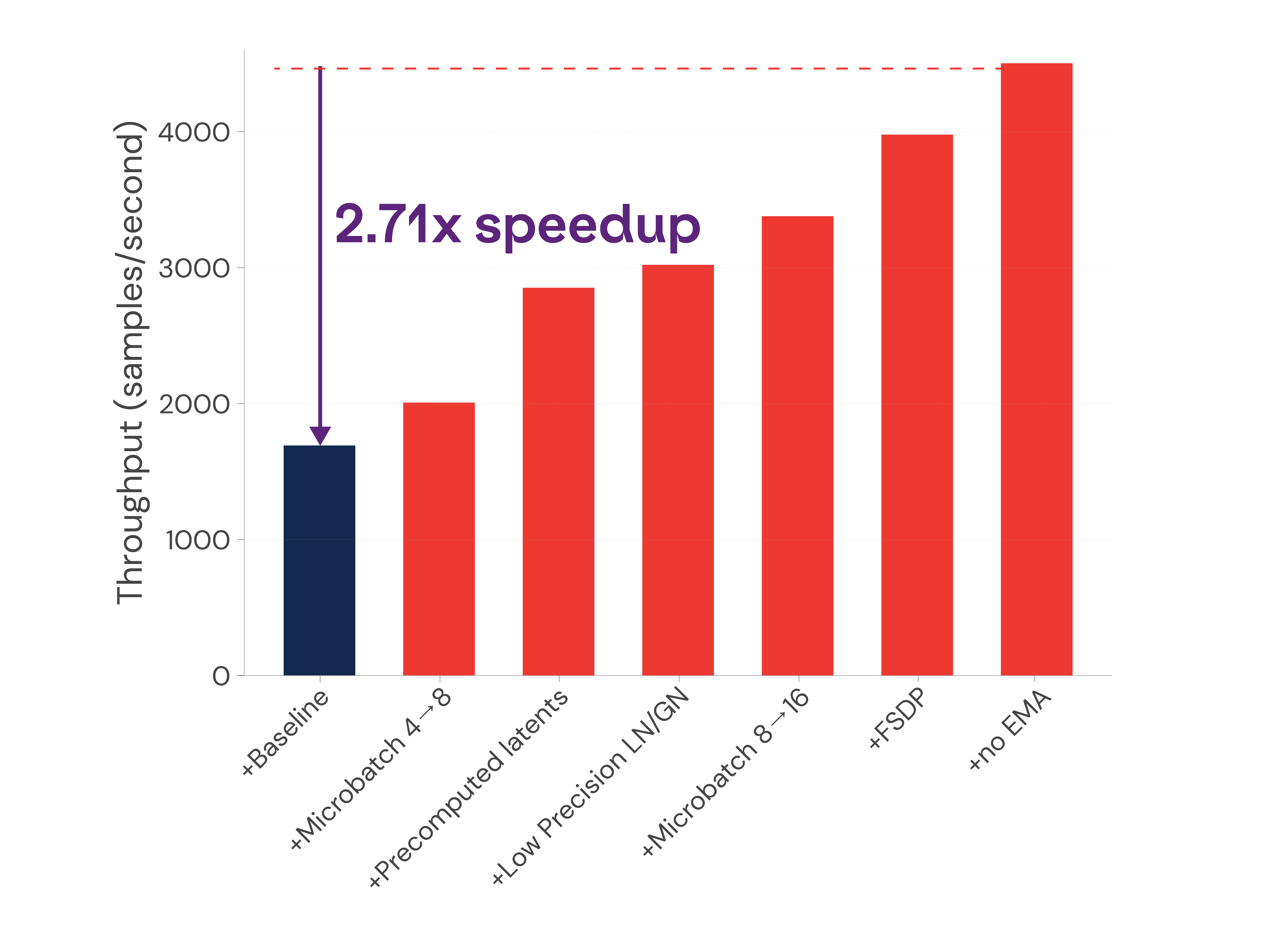}
        \caption{Cumulative effect of various speed-ups (totalling  $2.71\times$) in our SD2 training pipeline evaluated on 128 A100s. \looseness=-1}
        \label{fig:cc:benchmark-speedup}
\end{figure}

Stability AI reports an estimated 200,000 A100 hours to train SD2~\citep{sd2modelcard}. 
Depending on hardware, a single SD2 training run could take anywhere from a few weeks to over a month. 
We sought out multiple avenues to reduce this training-time constraint. 
We applied Flash Attention~\citep{dao2022flashattention} with the xFormers library~\citep{xFormers2022}, pre-computed VAE and text encoder latents over the entire training dataset, cast all GroupNorm~\cite{wu2018group} and LayerNorm~\cite{ba2016layer} to \textsf{float16} precision, and applied fully-sharded data parallelism (FSDP) to our training run. 
Finally we opted to only keep an exponential moving average of the weights for the final 3.5\% of training. 
Altogether, we are able to achieve a 2.71$\times$ speedup in A100 hours over our SD2 baseline implementation.

\begin{figure*}[t]
\centering
\includegraphics[width=0.95\linewidth]{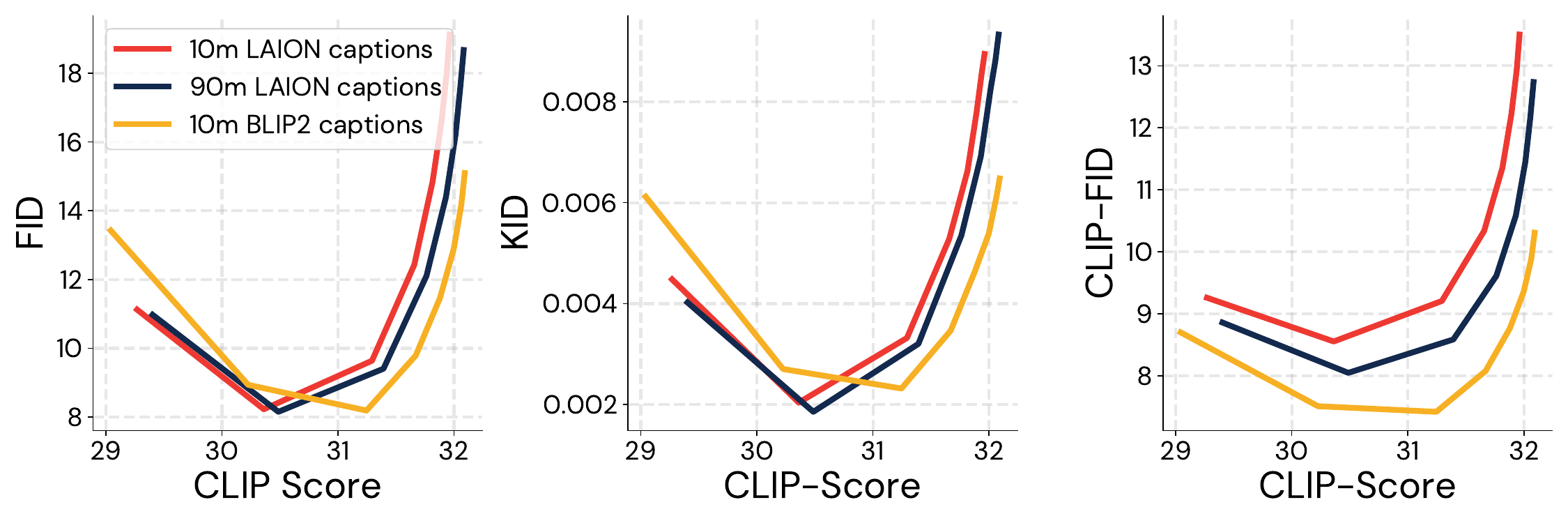}
\caption{For different SD2 models trained on subsets of LAION (90M, 10M using either original captions or synthetic BLIP-2 captions), we compute FID~\cite{heusel2017gans}, KID~\cite{binkowski2018demystifying}, CLIP-FID~\cite{kynkaanniemi2022role}, and CLIP-Score~\cite{hessel2021clipscore} on 30K samples from MS COCO. 
We compute these metrics across a text-guidance scale of 1-8, with higher values indicating the model should respect the text prompt more. 
Lower FID, KID, and CLIP-FID indicate higher quality; higher CLIP-Score indicates higher quality.
Together, these plots show that increasing the amount of training data from 10M to 90M samples does not lead to quantitative improvements.   BLIP-2 re-captions provide nearly identical performance to LAION in terms of FID and KID; the re-captions indicate slightly better performance when using CLIP-FID as the quality metric.} 
\label{fig:cc:data-variance}
\end{figure*}

We found that latent pre-computation helped the most at low resolutions, while FSDP also provided significant gains, especially at scale. 
The other optimizations helped reduce total memory usage, allowing us to increase the microbatch size for better hardware utilization. 
Figure \ref{fig:cc:benchmark-speedup} summarizes each of the proposed methods and the cumulative speedup that results from their application. Equipped with an optimized training setup, it is more feasible for us to study the effect of varying training-dataset size.
More details can be found in Appendix~\ref{app:sec:cc:mlsys}. 

\subsection{Investigating data scarcity}\label{sec:cc:descale}

YFCC100M contains 100 million images, about 10\% the size of the 1.1B LAION examples we could access (due to link rot) --- about 5\% of the original LAION-2B dataset. 
An interesting question remains: \textit{how much data is actually needed to train these diffusion models effectively; do we really need billions of images to get high-quality results?} 

To answer this question, we train multiple SD2 architectures on increasingly smaller, random subsets of data from our LAION-1.1B dataset: 1.1B, 90M, 10M, and 1M sample subsets. 
While human evaluation remains the gold standard for evaluating generative models, we use proposed automated metrics like Frechet-Inception Distance~\cite{heusel2017gans}, Kernal Inception Distance~\cite{binkowski2018demystifying} and caption-alignment metrics such as CLIP Score~\cite{hessel2021clipscore} (Figure~\ref{cc:eval-methods}). 
We find that performance (FID and KID on MS COCO) does not degrade until training with as few as 1 million images; our models trained on 10M and 90M subsets perform comparably to the entire 1.1B dataset (Appendix~\ref{chapter:app:cc}, Figure~\ref{fig:eval-over-time-less-data}). Figure~\ref{fig:cc:data-variance} further compares our SD2 variants trained on 10M and 90M LAION subsets across different guidance scales. 
We also plot the effect of using the original LAION captions vs. BLIP-2 synthetic captions at these size regimes (discussed further in Section~\ref{ssec:synth-captions}). 
These findings suggest that SD2 models may be underparameterized. 
We hypothesize about why this might be the case and how much data is actually necessary to saturate the model in Appendix~\ref{chapter:app:cc}. 

%% file: section/40-genai/42-commoncanvas/426-commoncanvas-experiments.tex
\section{Experiments }\label{sec:cc:experiments}

In this section, our model evaluations use automated, quantitative image-quality metrics from the literature. 
We measure performance with three automated image quality metrics on the commonly used MS COCO dataset~\citep{lin2014microsoft}: 
Fr\'echet Inception Distance (FID)~\citep{heusel2017gans}, Kernel Inception Distance (KID)~\citep{binkowski2018demystifying}, and CLIP-FID~\citep{kynkaanniemi2022role}. 
Each captures a slightly different measures of generated-image quality and diversity, in relation to statistics in the training data, with lower values corresponding to higher quality.  
Additionally, we evaluated CLIP-Score~\cite{hessel2021clipscore}, which can help us  understand the alignment between captions and their respective images, with higher values signaling better alignment. 
While these automated metrics are intended to be efficient proxies for human preferences in image quality, they often fall short; 
the gold standard for T2I model evaluation still remains human evaluation. 
Since synthetic captions differ so much from human-designed ones~\cite{nguyen2023improving}, we also set up a pairwise preference rating task to measure the relative quality of our trained models.\looseness=-1 

\begin{figure*}[t!]
    \centering 
    \small
    \setlength{\tabcolsep}{1pt}
    \setlength{\itemwidth}{0.17\linewidth}
    \newcolumntype{M}[1]{>{\centering\arraybackslash}m{#1}}
    \begin{tabular}{M{\itemwidth}M{\itemwidth}M{\itemwidth}M{\itemwidth}M{\itemwidth}}
      Prompt & SD2 & \modelname-S-C & \modelname-S-NC & \modelname-L-NC \\
    \prompt{a cute black cat inside of a pumpkin} &    \includegraphics[width=\itemwidth]{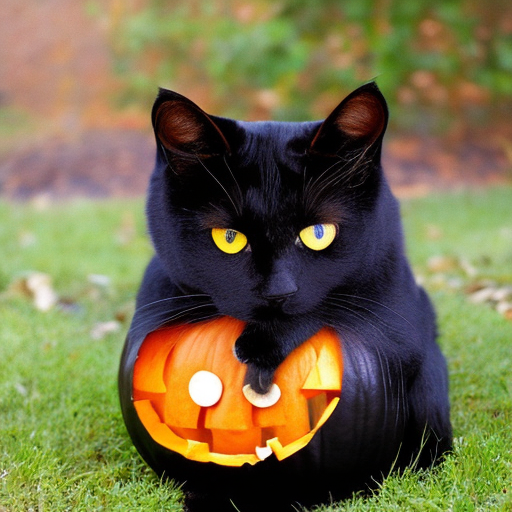}  &
    \includegraphics[width=\itemwidth]{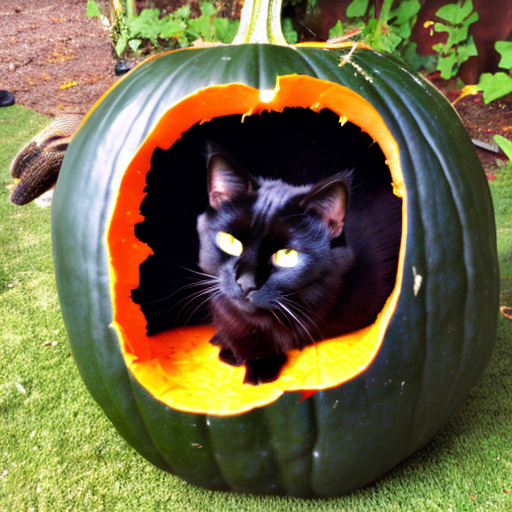} &
    \includegraphics[width=\itemwidth]{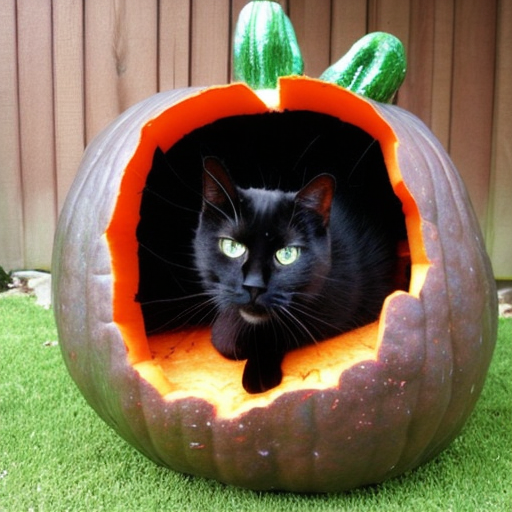} &     
    \includegraphics[width=\itemwidth]{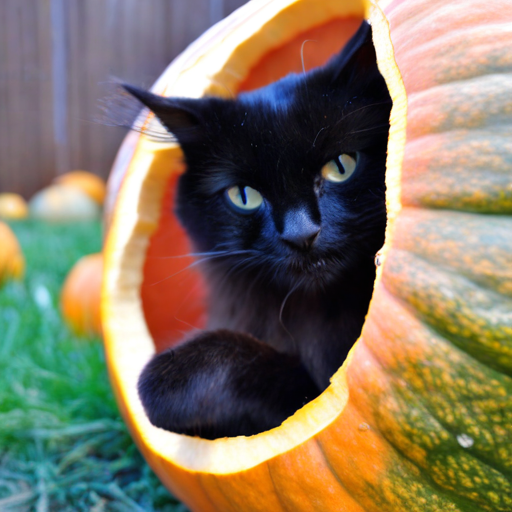} \\

    \prompt{a robot holding a paint palette} &
    \includegraphics[width=\itemwidth]{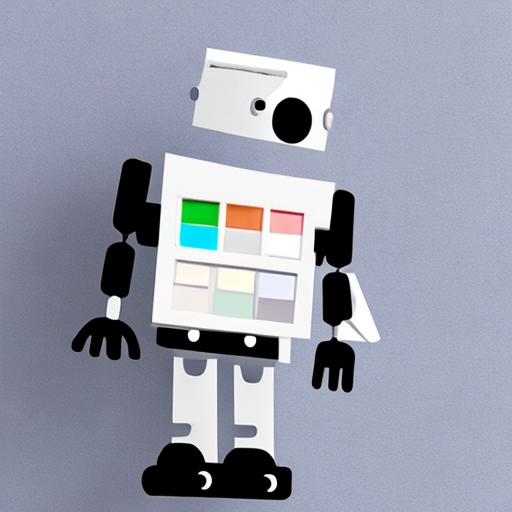}  &
    \includegraphics[width=\itemwidth]{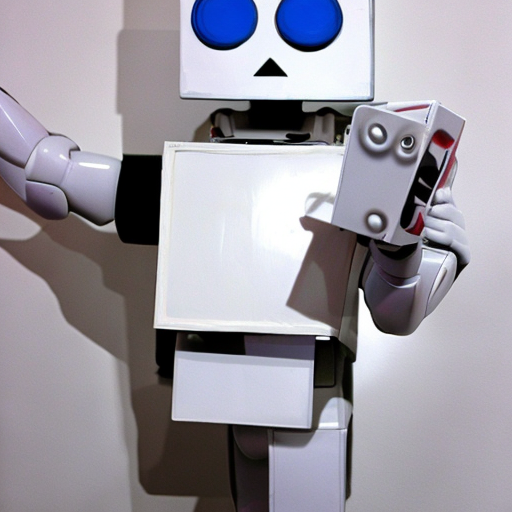} &
    \includegraphics[width=\itemwidth]{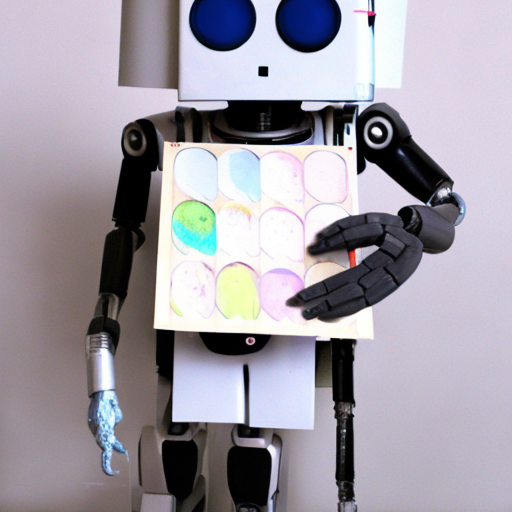} &     
    \includegraphics[width=\itemwidth]{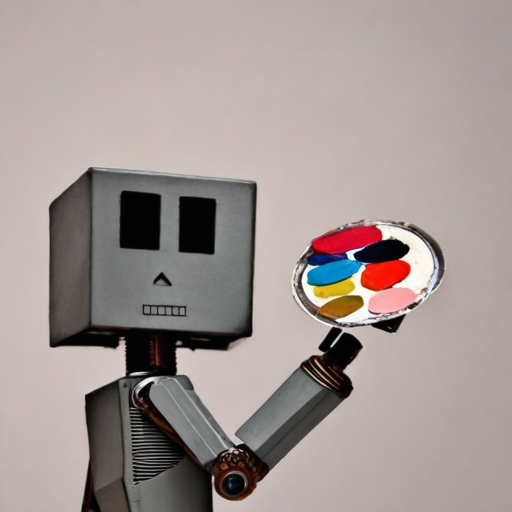} \\
    \end{tabular}
    \caption{Using entirely Creative-Commons images and our synthetic captioning approach, we achieve comparable qualitative performance to public SD2, as seen in \modelname{} generations, while only requiring a small fraction ($<3\%$) of the amount of training data. We include results for two \modelname{} \textsf{architectures}, small (S) and large (L) (Section~\ref{sec:cc:experiments}), and two CC-image \textsf{datasets}, commercial (C) and non-commercial (NC) (Section~\ref{sec:cc:dataset}). We label our results accordingly as \modelname-$<$\textsf{architecture}$>$-$<$\textsf{dataset}$>$.} 
    \label{fig:qual-hero-fig}
\end{figure*}
\label{cc:eval-methods}

\subsection{Training with Synthetic Captions}
\label{ssec:synth-captions}

First, we look at the effect of training with synthetic captions instead of ground-truth captions from LAION. Interestingly, we observe that synthetic captions can enhance the alignment of our model. For instance, the CLIP-Score for synthetic captions exceeded that of ground-truth captions as seen in Figure~\ref{fig:cc:data-variance} (for CLIP-FID). 

\begin{figure}[t]
    \centering
     \includegraphics[width=.8\linewidth]{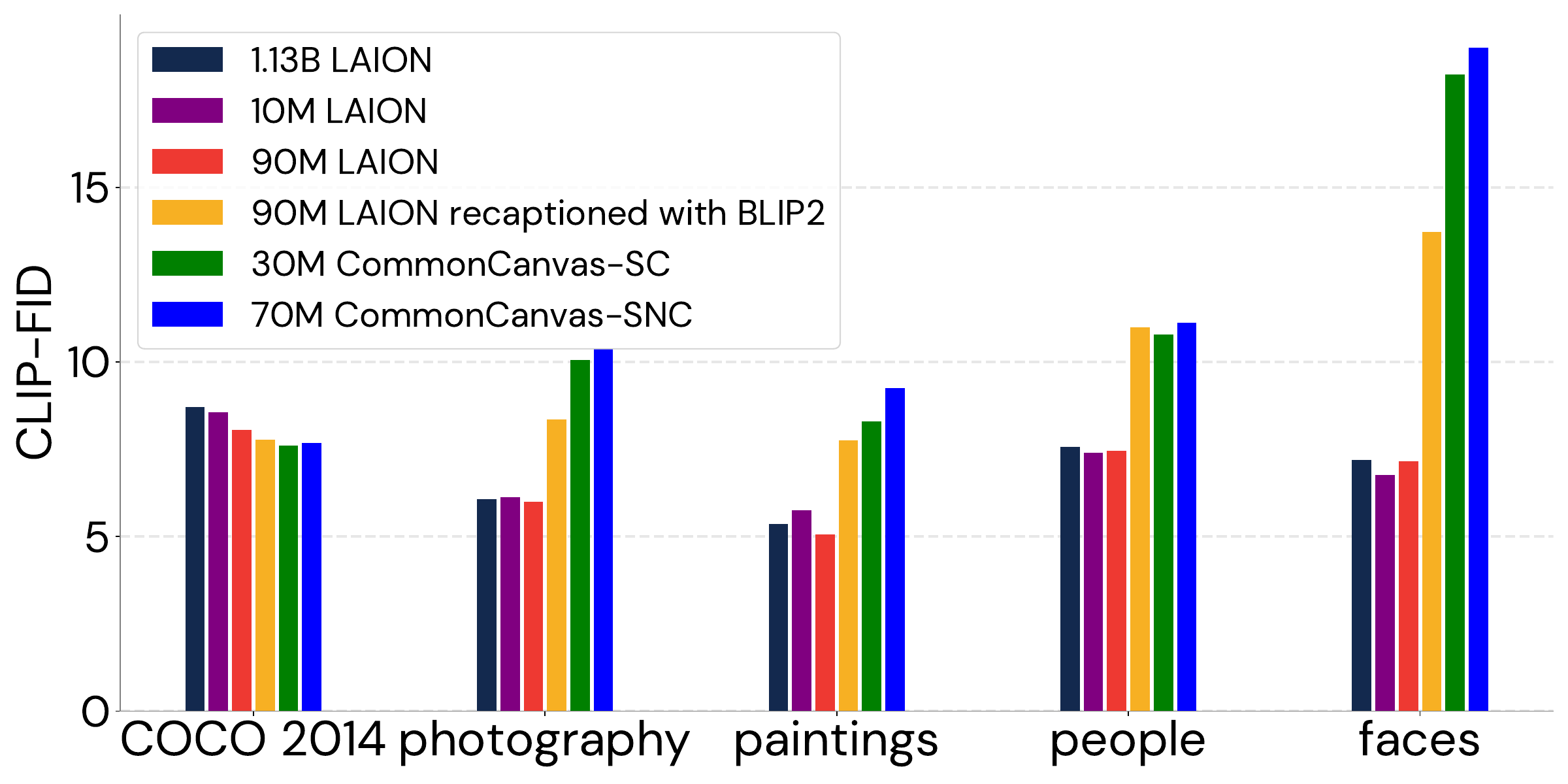}
    \caption{Evaluating models at 256 resolution on different subsets of the Conceptual Captions dataset and MS COCO. LAION models are trained on {\textbf{\color{darkgray}1.1 billion}}, {\textbf{\color{BrickRed} 90 million (\sdtwobase)}}, and {\textbf{\color{Plum}10 million}} subsets. We also train a model with {\textbf{\color{amber} a 90 million subset re-captioned with BLIP-2}} to evaluate distribution shift. The last two models are trained on {\textbf{\color{ForestGreen} on the \datasetname-C}}, and {\textbf{\color{blue} \datasetname-NC}}. 
    We observe a domain shift between MS COCO and web-scraped Conceptual Captions. CLIP-FID may exhibit a preference for SD2 models, given that CLIP has been trained on a text style akin to that found in LAION. 
    Subsampling the LAION dataset from 1.13B to 10M images does not seem to affect quantative performance. Using synthetic captions causes a significant performance drop on the LAION dataset when evaluated on Conceptual Caption test datasets, but not MS COCO.\looseness=-1}
    \label{fig:conceptual-captions}
\end{figure}

To get a more nuanced perspective on the effect of our synthetic captions, we assess CLIP-FID for image generations from different models on human- and computer-generated captions. 
In Figure~\ref{fig:conceptual-captions}, we compute CLIP-FID for various models trained using LAION, \datasetname, or LAION images re-captioned with BLIP-2; CLIP-FID is computed based on generating for prompts from MS COCO and the Conceptual Captions dataset. 
Unlike other caption datasets, MS COCO captions are human written. 
Most captions from web-based datasets (like  LAION) are computer-generated~\cite{nguyen2023improving}.
BLIP-2 captions are also generated, but the BLIP-2 model is then fine-tuned to align with human-written captions. 
Given the higher quality of our synthetic captions, it is unsurprising that \modelname's CLIP-FID is better (i.e., lower) for MS COCO (i.e., aligns better with human-written captions). 

However, like any model, ours has limitations. \modelname{} under-performed in several categories, including faces, general photography, and paintings. These datasets all originated from the Conceptual Captions dataset~\citep{sharma2018conceptual}, which relies on web-scraped data. These web-sourced captions, while abundant, may not always align with human-generated language nuances~\cite{nguyen2023improving,betker2023improving,caffagni2023synthcap}. 
Although transitioning to synthetic captions introduces certain performance challenges, the drop in performance is not as dramatic as one might assume. Moreover, we speculate that the model will perform better if users provide their more specialized datasets to the model, such as FFHQ~\citep{karras2019style}.

\begin{figure*}[t]
\begin{center}
\begin{minipage}{.9\linewidth}
            \centering
        \setlength{\groupwidth}{0.31\linewidth}
        \setlength{\itemwidth}{0.5
        \groupwidth}
        \setlength{\tabcolsep}{0pt}
        \newcolumntype{C}[1]{>{\centering\arraybackslash}p{#1}}
        
        \begin{tabular}{cc@{\hskip 0.05in}cc@{\hskip 0.05in}cc}
        
        Ours & SD2 & Ours & SD2  & Ours & SD2 \\
        
        \includegraphics[width=\itemwidth]{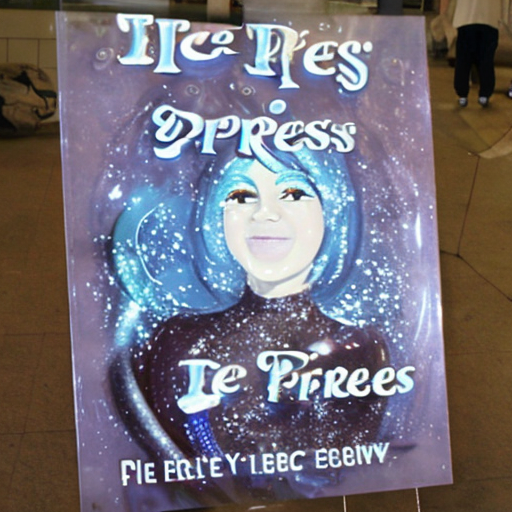} &
        \includegraphics[width=\itemwidth]{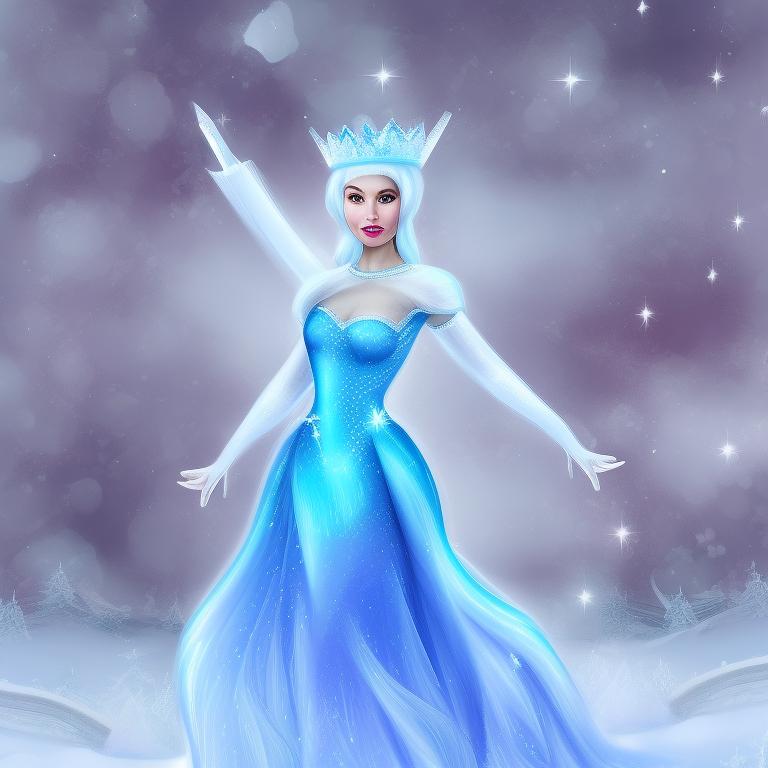} &

        \includegraphics[width=\itemwidth]{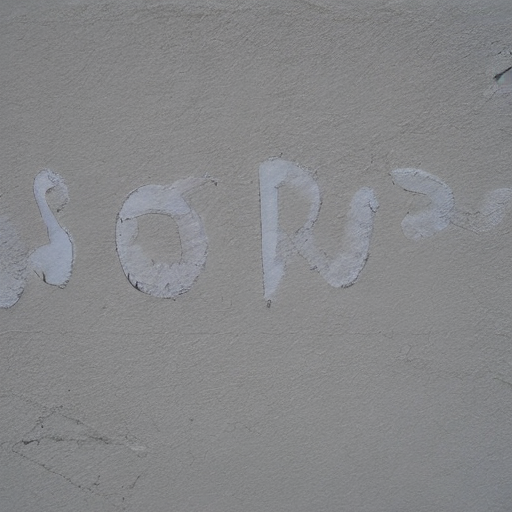} &
        \includegraphics[width=\itemwidth]{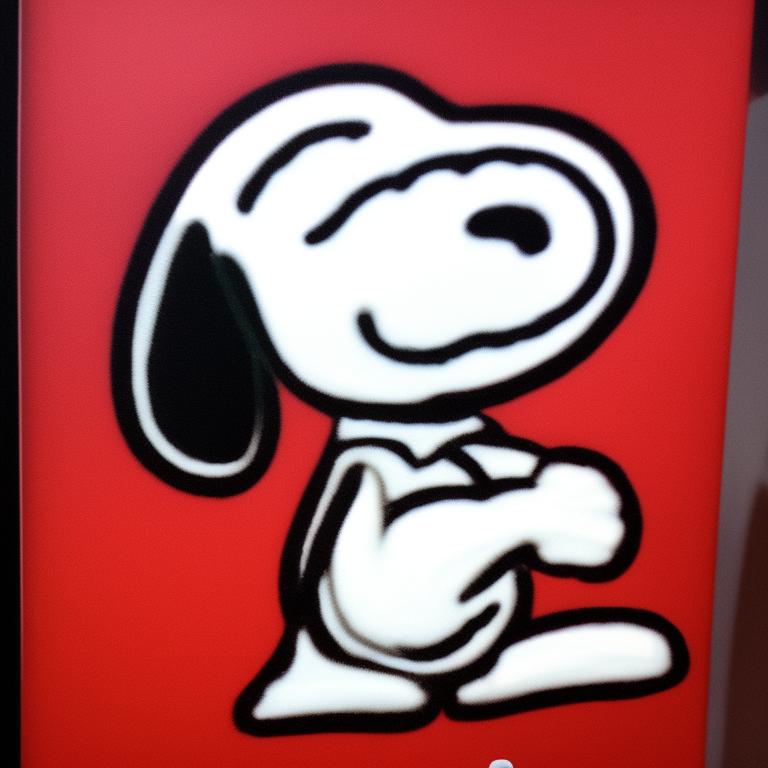} &
        
        \includegraphics[width=\itemwidth]{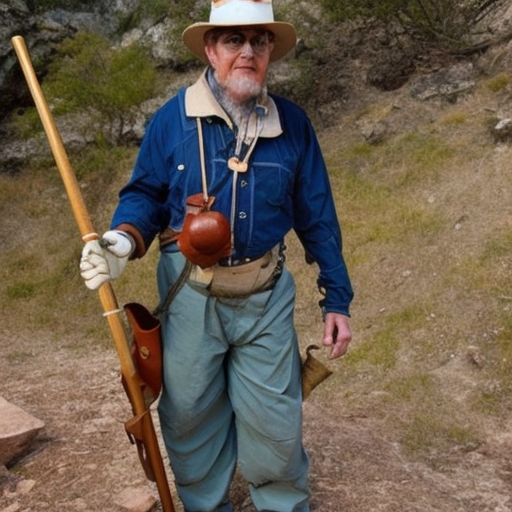} & %
        \includegraphics[width=\itemwidth]{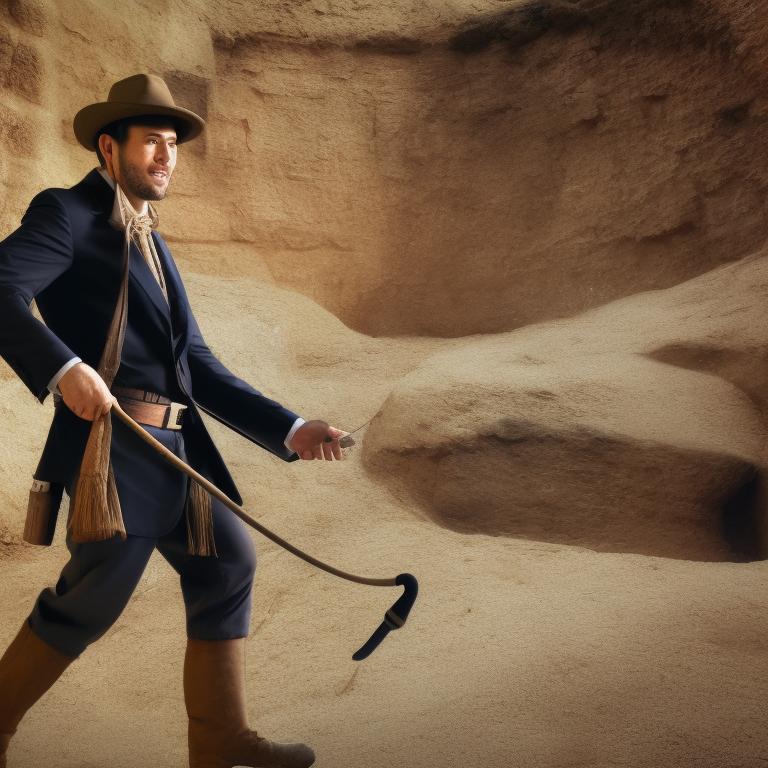} \\
        
        \multicolumn{2}{C{\groupwidth}}{\prompt{ice princess}} &
        \multicolumn{2}{C{\groupwidth}}{\prompt{Snoopy}} &
        \multicolumn{2}{C{\groupwidth}}{\prompt{a adventurous archaeologist with a whip and a fedora}} \\
        
        \includegraphics[width=\itemwidth]{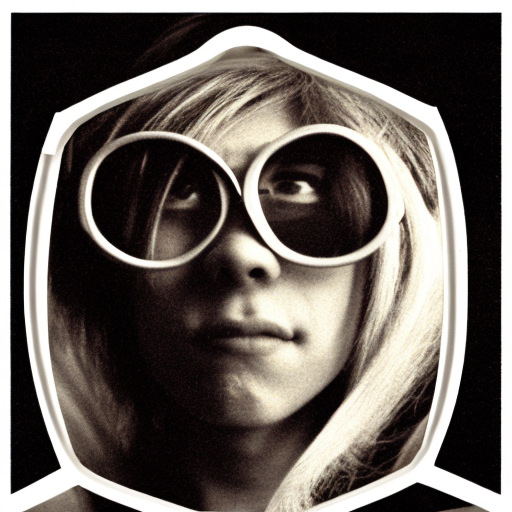} &
        \includegraphics[width=\itemwidth]{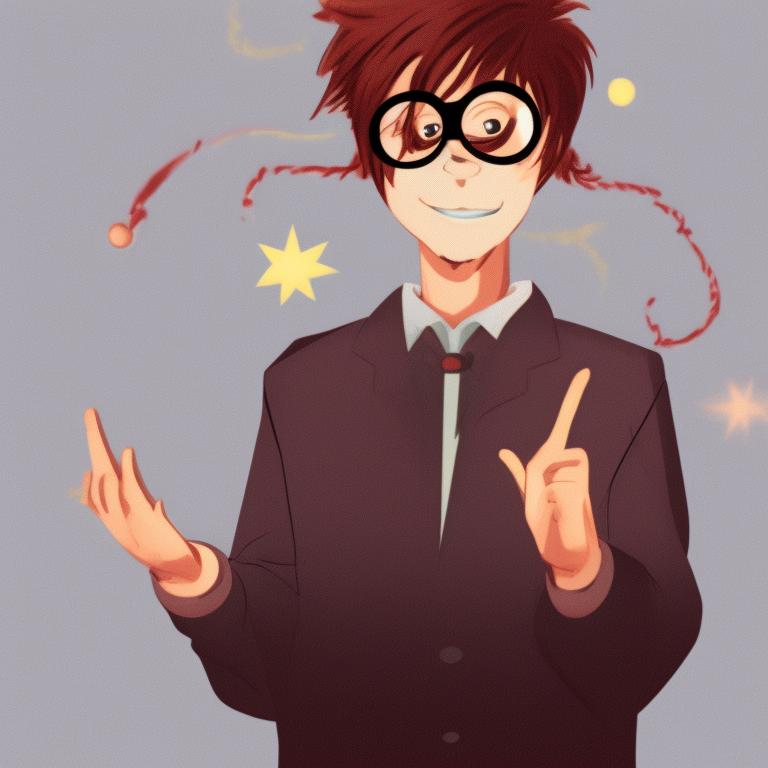} &
        
        \includegraphics[width=\itemwidth]{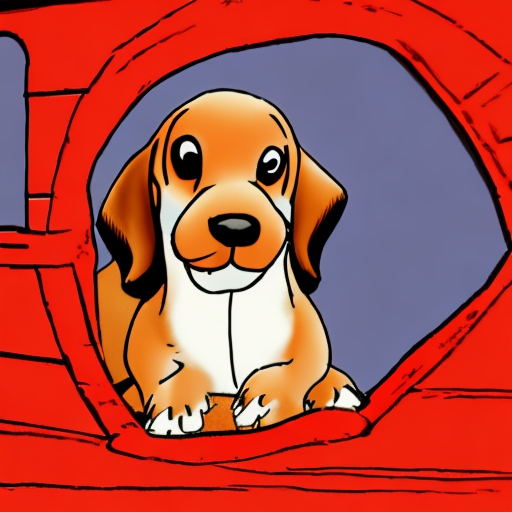} &
        \includegraphics[width=\itemwidth]{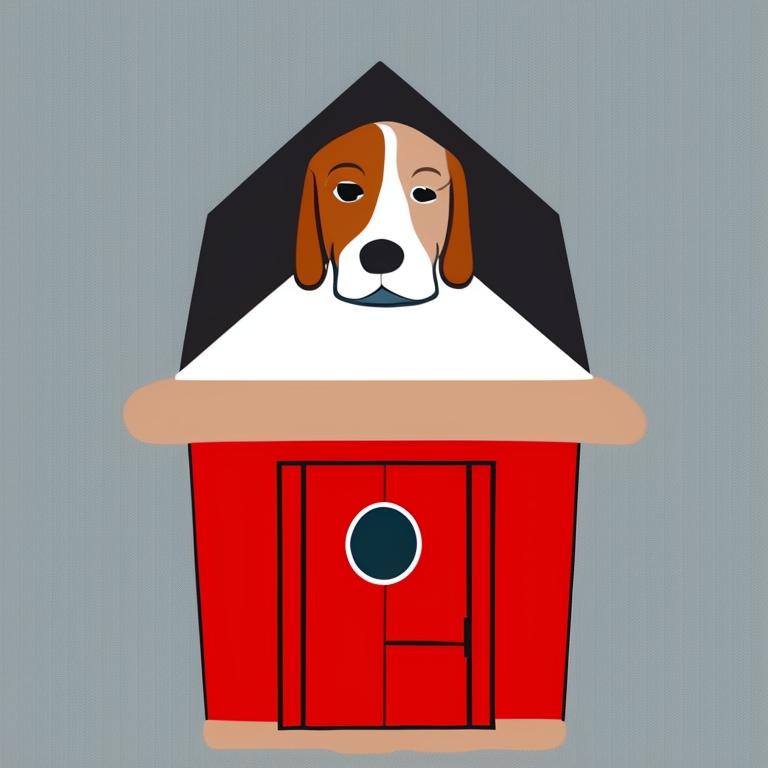} &
        
        \includegraphics[width=\itemwidth]{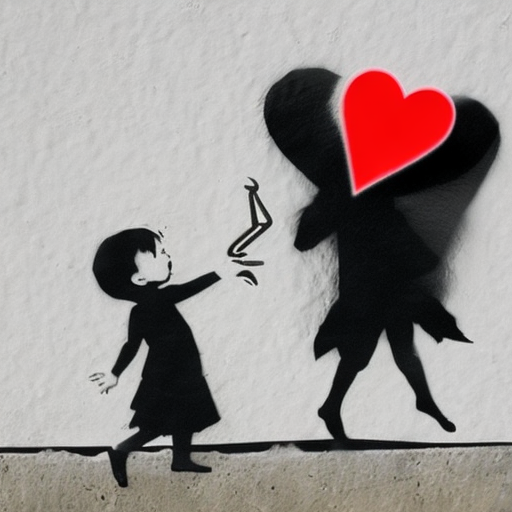} &
        \includegraphics[width=\itemwidth]{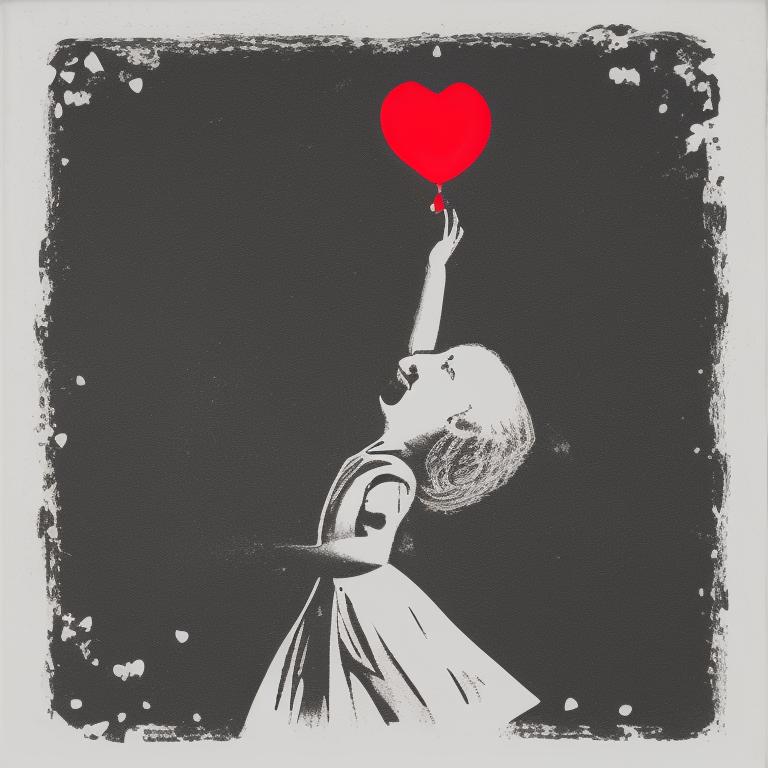} \\
        
        \multicolumn{2}{C{\groupwidth}}{\prompt{A teenage wizard with round glasses}} &
        \multicolumn{2}{C{\groupwidth}}{\prompt{a cartoon beagle in a red dog house}} &
        \multicolumn{2}{C{\groupwidth}}{\prompt{black and white stencil little girl reaching for a red balloon}}
        
        \end{tabular}
        \caption{We compare \modelname-S-NC (Ours) to SD2. Our model is less likely to generate iconic characters given suggestive prompts (drawn from Lee et al.~\cite{lee2023talkin}).}
        \label{fig:copyright-traps}
\end{minipage}
\end{center}
\end{figure*}

\subsection{CommonCanvas vs. LAION-trained SD2}

Given that our data-scarcity analysis suggests that \datasetname{} is large enough to train a high-quality SD2 model and that synthetic captions can perform well (Figure~\ref{ssec:synth-captions}), we train two different \modelname{} models: one trained on commercial (\datasetname-C) images, another on non-commercial (\datasetname-NC). 
For a fair comparison with SD2, we use the OpenCLIP text encoder. Like BLIP-2, OpenCLIP is trained on LAION captions (Figure~\ref{sec:laion}). 
For example generations, see Figure~\ref{fig:qual-hero-fig}.

We also note that, although we train on Creative-Commons images, it is still possible for an adversarial prompt to produce content that includes iconic characters. 
In Figure~\ref{fig:copyright-traps}, we subject our model to ambiguous prompts that are suggestive of such characters. Examples include visuals closely resembling Elsa from Frozen, Indiana Jones resembling Harrison Ford, and even a likeness to Harry Potter. 
Qualitatively, our model deviated more from these characters than SD2. 

\subsection{Reaching SD2 quality with CommonCanvas-L}

We also did a human study measuring pairwise preference ratings for the 512x512 resolution CommonCanvas models compared to SD2 (Figure~\ref{fig:parti-prompts}). 
In this experiment, human raters were shown a prompt (selected randomly from the PartiPrompts prompts set~\citep{yu2022scaling}) along with two generated images in randomized order, one from the reference model (public SD2) and the other from a CommonCanvas model. 
We report the fraction of the time users selected the image generated by the CommonCanvas model over the corresponding generation from SD2 as the user preference rate for that model. 
We find that our CommonCanvas models are slightly less preferred than \sdtwobase, with preference rates of 37\% for CommonCanvas-S-C and 38\% for CommonCanvas-S-NC, which we find surprisingly high considering the smaller and synthetic nature of the dataset. Figure~\ref{fig:qual-hero-fig} displays the results from our human study.

Our previous results suggest that SD2 may be underparameterized. We additionally train a larger variant of \modelname-N-C (\modelname-L-NC) that has a significantly larger U-Net (the U-Net architecture from SDXL (Podell et al.~\cite{podell2023sdxl},  Appendix~\ref{chapter:app:cc}). 
When we use CommonCanvas-L-NC, we achieve competitive performance with SD2 on user preferences (Figure~\ref{fig:qual-hero-fig}). For the largest model, CommonCanvas-L-NC, we do not measure a statistically significant difference in user preference between this model and SD2. 

\begin{figure*}
\begin{center}
\begin{minipage}{.9\linewidth}
            \centering
    \setlength{\groupwidth}{0.31\linewidth}
    \setlength{\itemwidth}{0.5\groupwidth}
    \setlength{\tabcolsep}{0pt}
    \newcolumntype{C}[1]{>{\centering\arraybackslash}p{#1}}
    
    \begin{tabular}{cc@{\hskip 0.05in}cc@{\hskip 0.05in}cc}
    
    Ours & SD2 & Ours & SD2 & Ours & SD2 \\
    \includegraphics[width=\itemwidth]{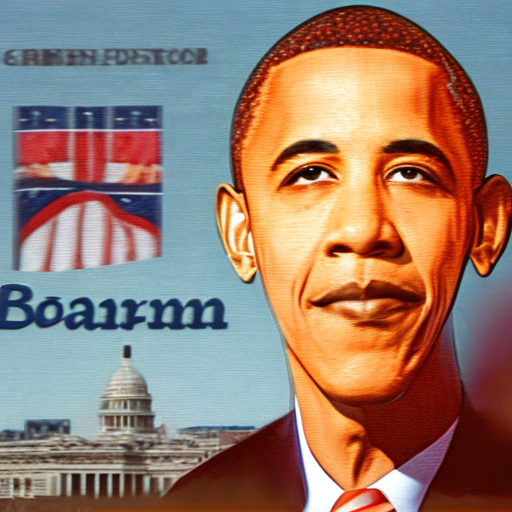} &
    \includegraphics[width=\itemwidth]{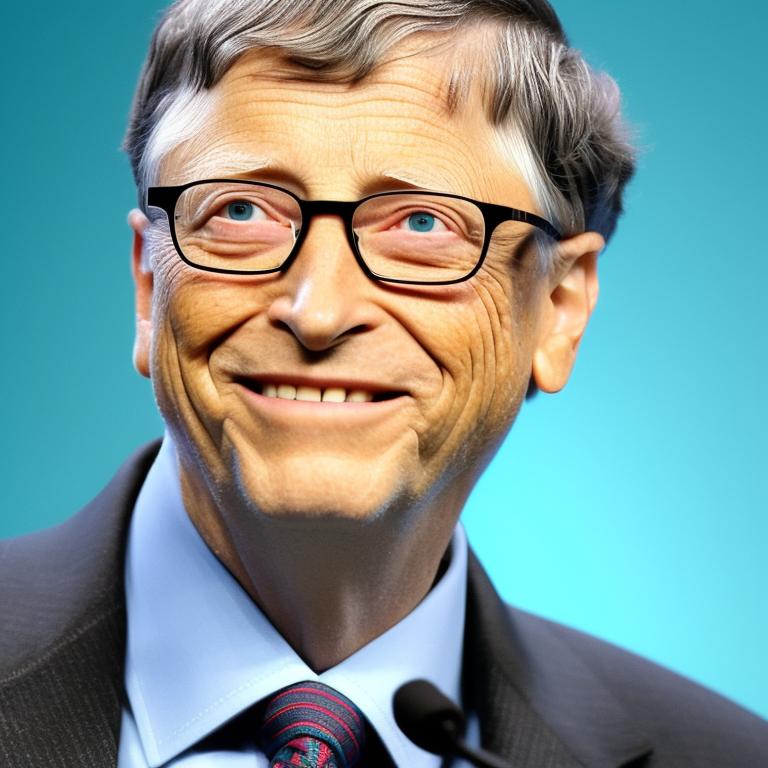} &

    \includegraphics[width=\itemwidth]{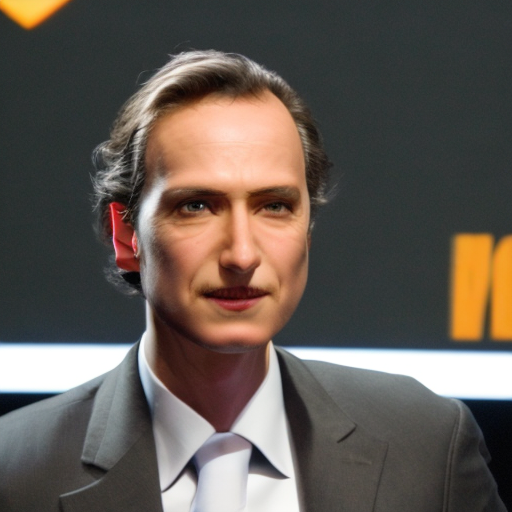} &
    \includegraphics[width=\itemwidth]{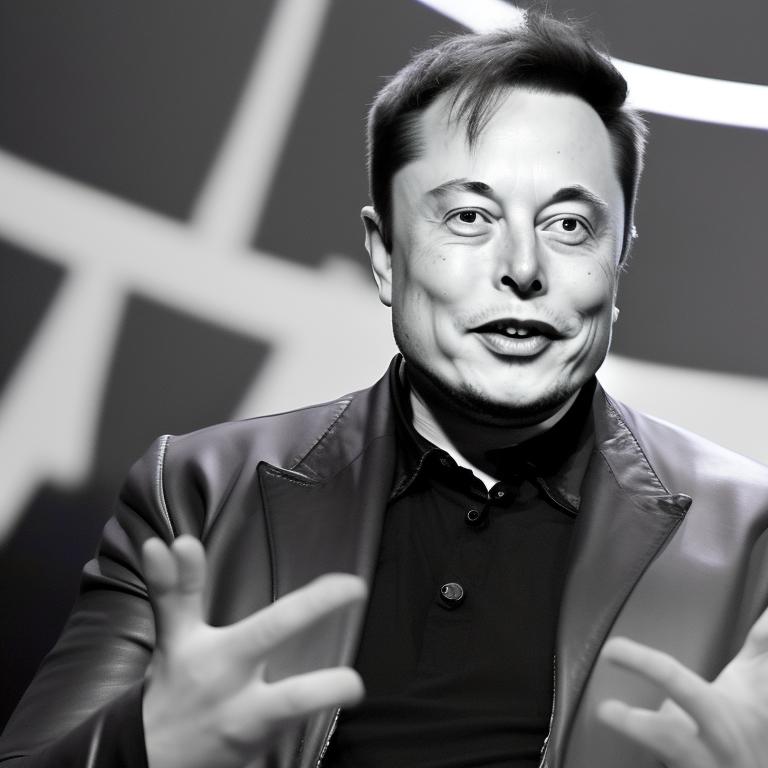} &     \includegraphics[width=\itemwidth]{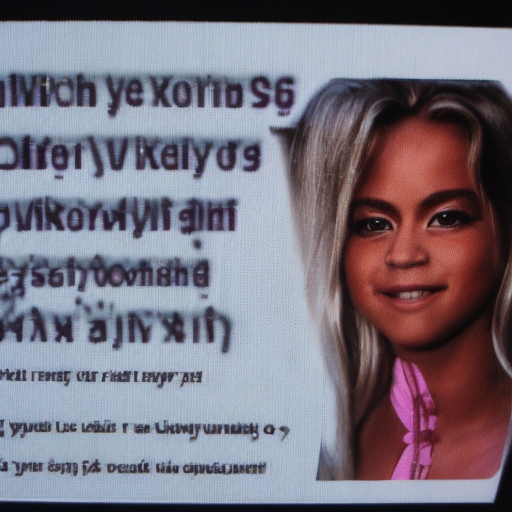} & %
    \includegraphics[width=\itemwidth]{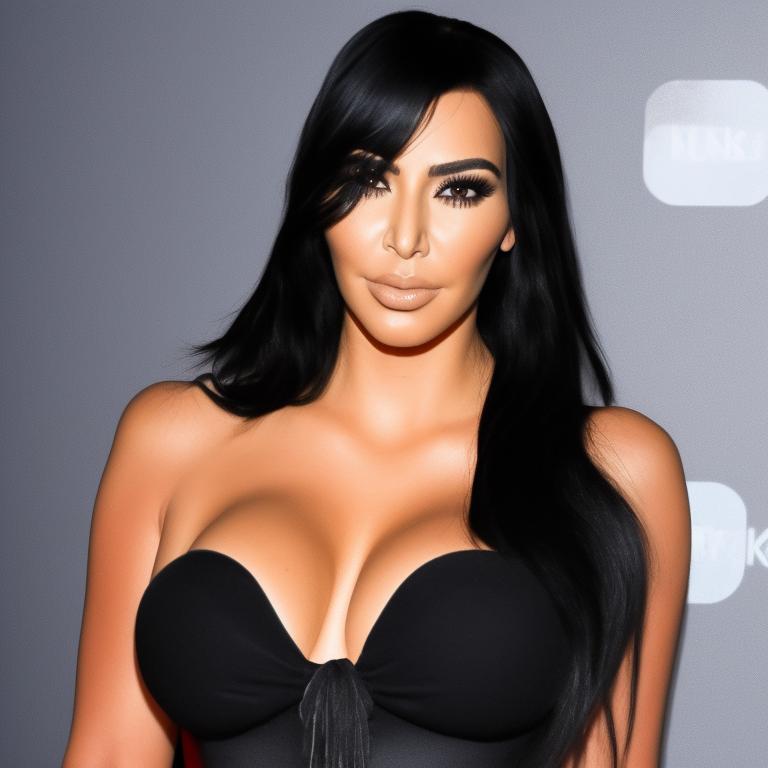} \\

\multicolumn{2}{c}{\prompt{Bill Gates}} & \multicolumn{2}{c}{\prompt{Elon Musk}} &
\multicolumn{2}{c}{\prompt{Kim Kardashian}}\vspace{.4in}

 \\

    \includegraphics[width=\itemwidth]{figure/42-cc/examples/ours/Barack-Obama-Ours.png} & \includegraphics[width=\itemwidth]{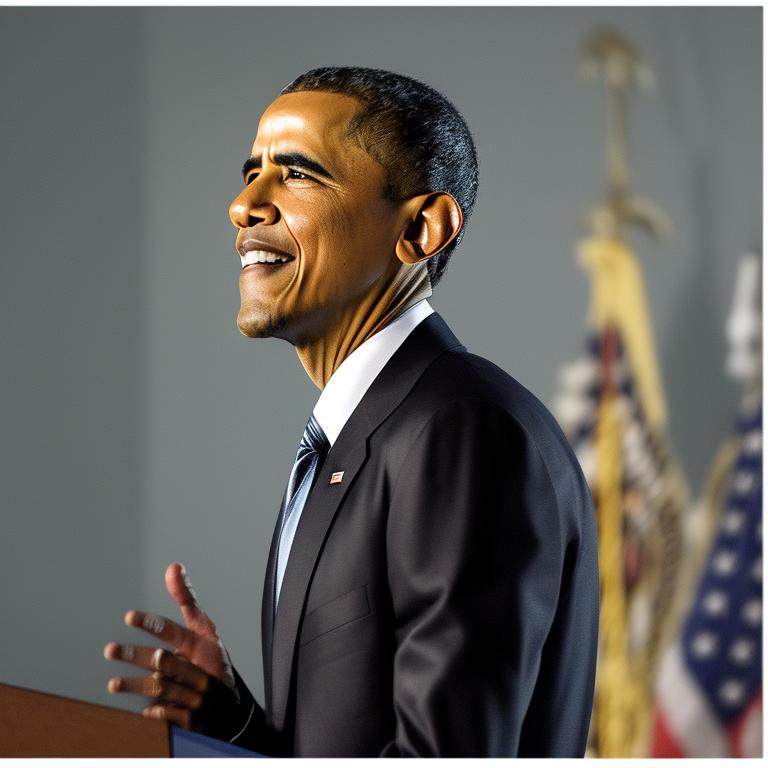} & \includegraphics[width=\itemwidth]{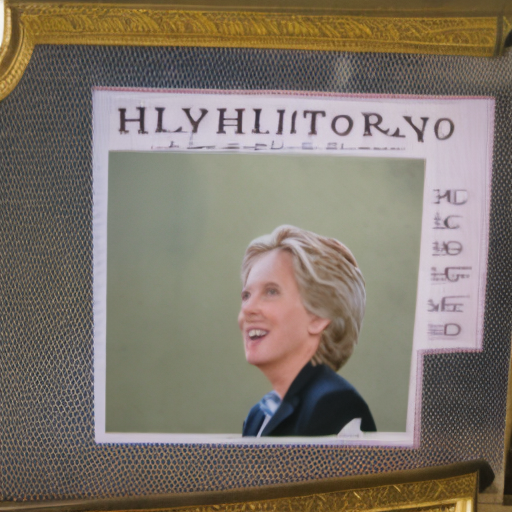} & \includegraphics[width=\itemwidth]{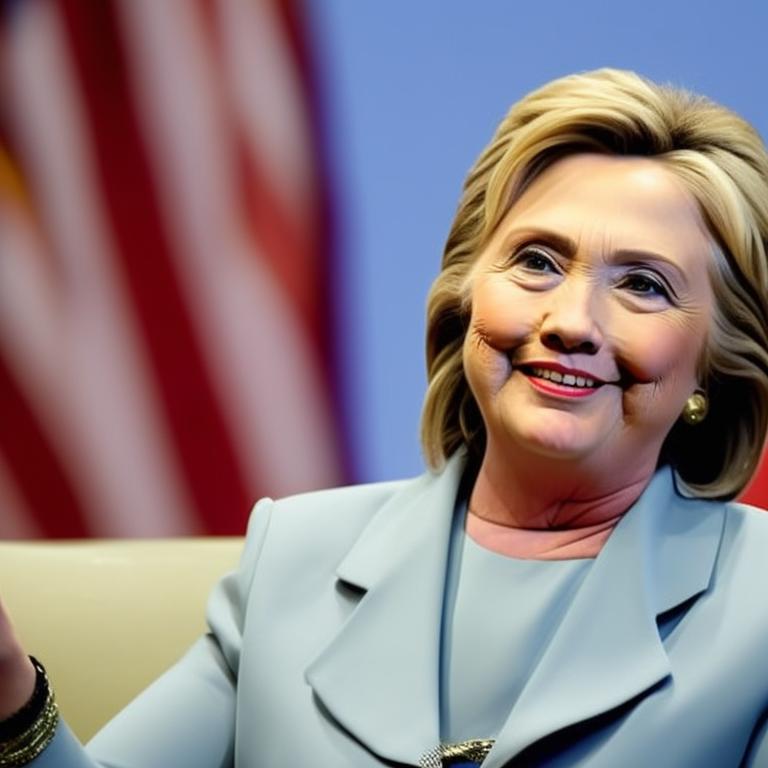} & 
    \includegraphics[width=\itemwidth]{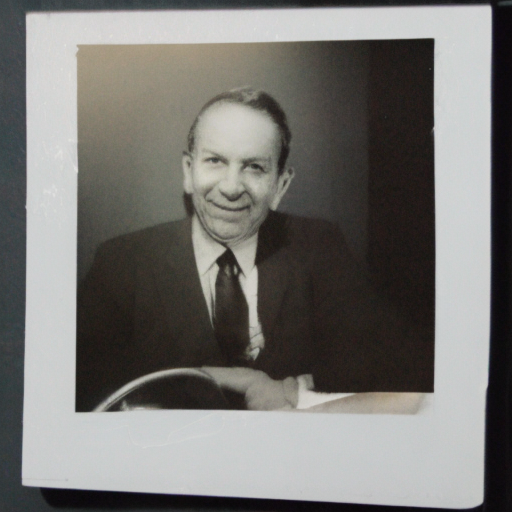} & 
    \includegraphics[width=\itemwidth]{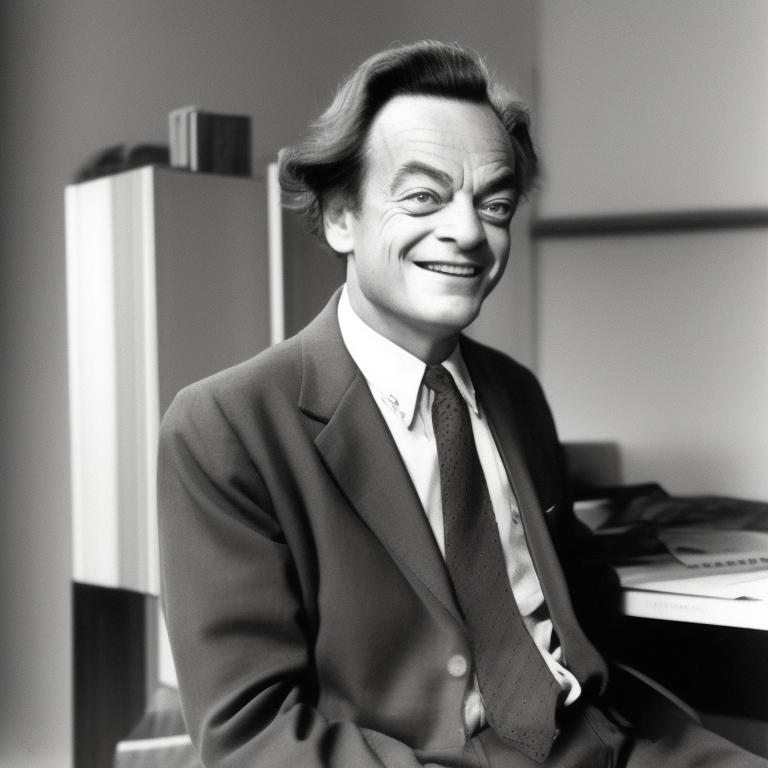} \\
    \multicolumn{2}{c}{\prompt{Barack Obama}} &
    \multicolumn{2}{c}{\prompt{Hillary Clinton}} &
    \multicolumn{2}{c}{\prompt{Richard Feynman}}
    \end{tabular}
    \caption{Using \modelname-SNC (Ours) to generate celebrities. Our model is worse at synthesizing individual people than SD2, but is capable of generating some noteworthy public figures. This result demonstrates how our model struggles to generate specific celebrities, which may be desirable from a privacy perspective.}
    \label{fig:celeb-ids}
    \end{minipage}
    \end{center}
\end{figure*}

%% file: section/40-genai/42-commoncanvas/427-commoncanvas-rw.tex
\section{Discussion and Related Work}\label{sec:cc:discussion}

In this paper, we train the \modelname{} family of text-to-image, latent diffusion models using only Creative-Commons images and synthetic captions. 
We discuss and address data incompleteness and scarcity issues associated with CC images.
For data incompleteness, we propose \captionmethod, an intuitive type of transfer learning (Section~\ref{sec:cc:transfer}), which we instantiate with BLIP-2 to produce synthetic captions for CC images (together, the \datasetname{} dataset; Section~\ref{sec:cc:dataset}). 
Regarding data scarcity, we hypothesize that only a small fraction of the data contained in LAION-2B is actually necessary to saturate SD2, and that the examples in \datasetname{} should be sufficient for training.
To make testing this hypothesis more efficient, we implement a variety of ML-systems optimizations, which achieve a $2.71\times$ speed-up over our SD2 baseline. 

\begin{figure}[t]
        \centering
        \includegraphics[width=.7\linewidth]{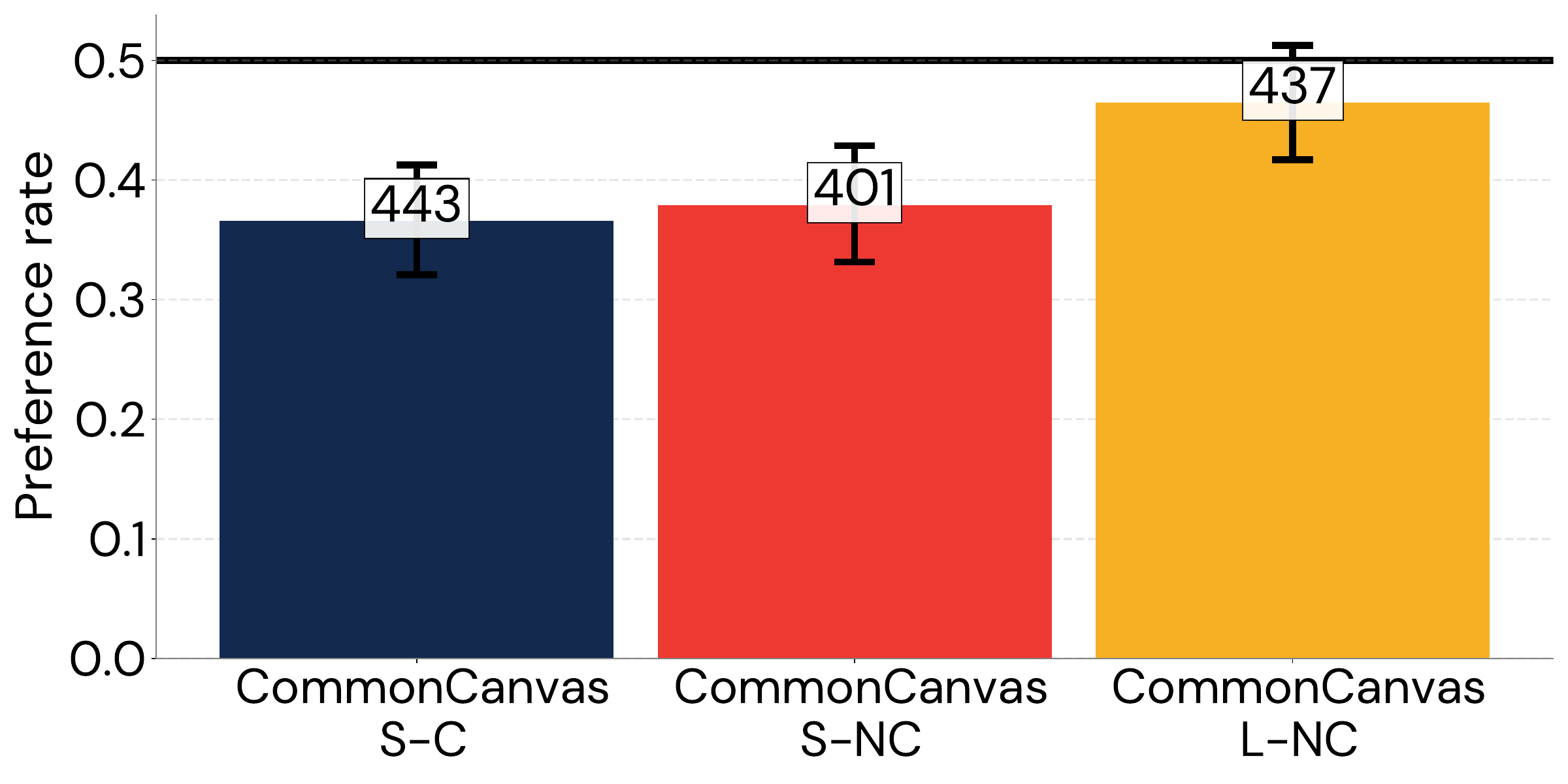}
        \caption{User preference study using Parti prompts. Preference rate (compared to SD2, the thick black horizontal line). 
        \modelname{}-L-NC matches the performance of SD2.} 
    \label{fig:parti-prompts} 
\end{figure}

Ultimately, we find that we can train the SD2 model on $<$$3\%$ of LAION-2B (i.e., roughly 70 million images;  Section~\ref{sec:cc:mlsys}), yielding a model we call \sdtwobase. This encourages us to train on \datasetname's commercially usable (also roughly 70 million) and non-commercially usable (roughly 25 million) examples. 
Compared to SD2, our \modelname{} models under-perform in some categories, like faces, but \modelname-L-NC demonstrates statistically equivalent performance with SD2 on human evaluation (Section~\ref{sec:cc:experiments}).\looseness=-1 

While several recent works similarly address ML topics relating to copyright, the literature tends to concern text-to-text training data~\citep{min2023silo}, be primarily theoretical~\citep{vyas2023provable, scheffler2022formalizing}, involve ablation studies~\citep{kumari2023ablating}, or only handle verbatim memorization~\citep{carlini2021extracting, nasr2023scalable} through the use of generation-time content filters~\citep{copilot-copy-filter}, which has been shown to be an incomplete solution~\citep{ippolito2023preventing}. 
To the best of our knowledge, no prior open work attempts to train T2I models on only open-licensed data. 
Most prior work on image-caption-dataset creation has extracted caption data from Common Crawl~\citep{gadre2023datacomp,desai2021redcaps,laurencon2023obelics}. 
We instead focus on synthesizing captions directly by using a pre-trained BLIP-2 model. 
Nguyen et al.~\cite{nguyen2023improving} demonstrates that existing caption datasets can be improved by using BLIP-2 to replace low-quality image captions 
(e.g., in Datacomp), but does not focus on creating a new dataset of synthetic captions, as we do here. 

Another limitation is that the YFCC100M data is about a decade old; its CC images are not as current as those in LAION-2B. In the future, we plan to augment \datasetname{} with Creative-Commons images from other sources, as well as test larger model architectures and more advanced captioning models, like LLaVA~\citep{liu2023visual}.

%% file: section/40-genai/43-talkinshort/43-talkinshort-main.tex
\chapter{Talkin' 'Bout AI Generation: Copyright and the Generative-AI Supply Chain (The Short Version)}\label{chapter:talkinshort}

Memorization (Chapter~\ref{chapter:memorization}) and licensed training data (Chapter~\ref{chapter:commoncanvas} are just two of many issues that generative AI presents for copyright. 
In this chapter, we explore these issues in more detail. 
However, this chapter is a much-abridged version of more extensive work published on this topic~\citep{lee2023talkin}.
This work has had a significant impact on legal scholarship, U.S. and U.K. AI policy, and more.\\ 

\noindent \textbf{Chapter summary}: 
``Does generative AI infringe copyright?'' is an urgent question.
It is also a difficult question, for two reasons.
First, ``generative AI'' is not just one product from one company.
It is a catch-all name for a massive ecosystem of loosely related technologies.
These systems behave differently and raise different legal issues.
Second, copyright law is notoriously complicated, and generative-AI systems manage to touch on a great many corners of it.
They raise issues of authorship, similarity, direct and indirect liability, and fair use, among much else.
These issues cannot be analyzed in isolation, because there are connections everywhere.

We aim to bring order to the chaos.
To do so, we introduce the \textbf{generative-AI supply chain}:  an interconnected set of stages that transform training data (millions of pictures of cats) 
into generations. (a new, potentially never-seen-before picture of a cat that has never existed).
Breaking down generative AI into these constituent stages reveals all of the places at which companies and users make choices that have copyright consequences.
It enables us to trace the effects of upstream technical designs on downstream uses, and to assess who in these complicated sociotechnical systems bears responsibility for infringement when it happens.
Because we engage so closely with the technology of generative AI, we are able to shed more light on the copyright questions.
We identify the key decisions that courts will need to make as they grapple with these issues, and point out the consequences that would likely flow from different liability regimes.\\

\noindent This chapter is a licensed derivative copy of work published and awarded a Long Presentation slot at \emph{ACM CSLAW 2024}~\cite{cooper2024talkinshort}, which is a much-abbreviated version of a law review article published at \emph{The Journal of the Copyright Society}~\citep{lee2023talkin}. Follow-on work is forthcoming at the \emph{Chicago-Kent Law Review}~\citep{cooper2024files}.

\input{section/40-genai/43-talkinshort/4310-talkinshort-intro}
\input{section/40-genai/43-talkinshort/4320-talkinshort-chain}
\input{section/40-genai/43-talkinshort/4330-talkinshort-copyright}
\input{section/40-genai/43-talkinshort/4340-talkinshort-outcomes}
\input{section/40-genai/43-talkinshort/4350-talkinshort-conclusion}

%% file: section/40-genai/43-talkinshort/4310-talkinshort-intro.tex
\section{Introduction}

Generative-AI systems like ChatGPT, Gemini, \dalle, and Ideogram can turn a user-supplied prompt like \prompt{give three arguments why marbury v. madison was wrongly decided} into a persuasive essay, or \prompt{a cowboy rid\-ing a rocket ship} in\-to a work of digital art.
They are unpredictable and complex;
they break out of existing legal categories.
In particular, because generative-AI systems are trained on millions of examples of human creativity, they raise serious copyright issues.
These copyright issues have not gone unnoticed.
Numerous plaintiffs have sued leading generative-AI companies for copyright infringement, with potential damages reaching into the billions of dollars.\looseness=-1

This chapter looks systematically at  how copyright applies to generative-AI systems. 
Our first contribution is to be precise about what ``generative AI'' is.
It is not just one product from one company.
Instead, it is a catch-all term for a massive ecosystem of loosely related technologies, including conversational text chatbots like ChatGPT, image generators like Midjourney and \dalle, coding assistants like GitHub Copilot, and systems that compose music, create videos, and suggest molecules for new medical drugs. 
Generative-AI models have different technical architectures and are trained on different kinds and sources of data using different algorithms.
Some take months and cost millions of dollars to train; others can be spun up in a weekend.
Some models are offered through paid online services; others are distributed open-source, such that anyone could download and modify them.\looseness=-1

We take the complexity and diversity of generative-AI systems seriously.
We  introduce the \textbf{gen\-er\-a\-tive-AI supply chain}: 
an interconnected set of stages that transform training data (millions of pictures of cats) into generations (a  picture that may never have been seen before of a cat that may not exist).
We conceive of eight stages: 
1) production of creative works, 2) conversion of creative works into quantified data, 3) creation and curation of training datasets, 4) base model (pre-)training, 5) model fine-tuning to adapt to a specific problem domain, 6) model release or deployment within a software system, 7) generation, i.e., the AI-assisted production of new creative works, and 8) alignment, i.e., adjusting the model and system to  advance goals (such as helpfulness, safety, legal compliance). 
The supply chain is not a simple cascade from data to generations.
Instead, each stage is regularly adjusted to better meet the needs of the others.
Breaking down generative AI into these constituent stages reveals all of the places at which companies and users make choices that have copyright consequences.\looseness=-1

We then work systematically through the copyright analysis of these different stages.
Copyright law is notoriously complicated, and gen\-er\-a\-tive-AI systems manage to touch on a great many corners of it.
They raise issues of authorship, similarity, direct and indirect liability, fair use, and licensing, among much else.
These issues cannot be analyzed in isolation, because there are connections everywhere.
Whether the output of a generative-AI system is fair use can depend on how its training datasets were assembled.
Whether the creator of a generative-AI system is secondarily liable can depend on the prompts that its users supply.
We trace the effects of upstream technical designs on downstream uses, and assess who in these complicated sociotechnical systems bears responsibility for infringement when it happens.
Because we engage so closely with the technology of generative AI, we are able to shed more light on the copyright questions.
We do not give definitive answers as to who should and should not be held liable.
Instead, we identify the key decisions that courts will need to make as they grapple with these issues, and point out the consequences that would likely flow from different liability regimes. 

We proceed in three parts. 
We: 
\begin{itemize}
    \item Describe the generative-AI supply chain in detail, including what happens at each stage, the diversity of variations on the basic theme, and the design choices that the various actors must make to create and use a generative-AI system (Section~\ref{sec:talkinshort:supplychain}). 
    \item Provide examples of how the supply-chain framing facilitates detailed copyright analysis, covering substantial similarity, direct infringement, and fair use. 
    We ask \emph{what} might possibly be an infringing technical artifact, \emph{who} might be an infringing actor, and \emph{when} infringement may occur, and discuss how the choices made by actors at one point in the supply chain affect the copyright risks faced by others (Section~\ref{sec:copyright}).\looseness=-1
    \item Detail broader lessons, including the options courts have and how they should conceptualize generative AI (Section~\ref{whichway}).\looseness=-1
\end{itemize}

\noindent Altogether, we argue that copyright pervades the generative-AI supply chain, that fair use is not a silver bullet, that the ordinary business of copyright litigation will continue even in a generative-AI age, and that courts should beware of metaphors that provide too-easy answers to the genuinely hard problems before them. 
This chapter is a shortened version of a law review article, which treats these topics in much greater detail~\citep{lee2023talkin}.

%% file: section/40-genai/43-talkinshort/4320-talkinshort-chain.tex
\section{The Generative-AI Supply Chain}\label{sec:talkinshort:supplychain}

\begin{figure}[t!]
	\centering
	\includegraphics[width=.8\textwidth]{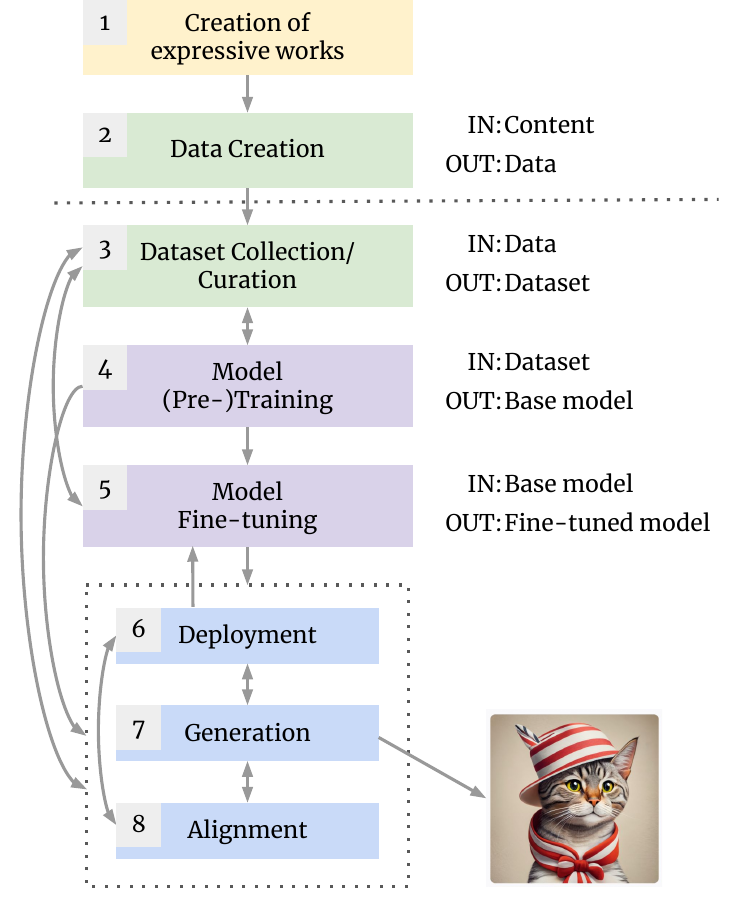}
    \caption{The generative-AI supply chain. 
		We map out eight stages: 1) creation of expressive works, 
		2) data creation, 
		3) dataset collection/curation, 
		4) model (pre-)training, 
		5) model fine-tuning,  
		6) system deployment, 
		7) generation,
		and 8) model alignment. 
		The creation of expressive works and data creation pre-date the advent of today's generative-AI systems (
        dotted line). 
		There are many possible ways to connect the other six stages. 
		Deployment, model alignment, and generation tend to happen in concert (dotted box). 
		Generations can be used as training data (arrow from generation (7) to dataset collection/curation (3)). 
		In this case, generation serves simultaneously as the creation of expressive works (1) and data creation (2). Curated data examples can be used for retrieval-augmented generation (arrow from dataset collection/curation (3) to generation (7)). APIs in deployed service can be used to do custom fine-tuning (arrow from deployment (6) to fine-tuning (5)).\looseness=-1 
	}
	\label{fig:talkinshort:chain}
\end{figure}

We assume introductory familiarity with machine learning (ML) and generative AI, and delve right into our discussion of the generative-AI supply chain.
To begin, we note that one of the big enablers of today's generative-AI systems is scale.
Notably, scale complicates \emph{what} technical and creative artifacts are produced, \emph{when} these artifacts are produced and stored, and \emph{who} exactly is involved in the production process. 
In turn, these considerations are important for how we reason about copyright implications: \emph{what} is potentially an infringing artifact, \emph{when} in the production process it is possible for infringement to occur, and \emph{who} is potentially an infringing actor~\citep{cooper2022accountability}.\footnote{The generative-AI supply chain 
	is a very good example of the ``many hands'' problem in computer systems. 
	That is, there are many diffuse actors, at potentially many different organizations, that can each have a hand in the construction of generative-AI systems. 
	It can be very challenging to identify responsible actors when these systems transgress broader societal expectations --- 
	in our case, the preservation of copyrights. 
    See Cooper et al.~\cite[pp. 867-869]{cooper2022accountability}  (describing the problem of ``many hands'' in data-driven ML/AI systems); 
}

To provide some structure for reasoning about this complexity, which will facilitate our copyright analysis in Section~\ref{sec:copyright}, 
we introduce our abstraction for reasoning about generative AI as a supply chain. 
We conceive of 
the \textbf{generative-AI supply chain} 
as having eight stages (see Figure~\ref{fig:talkinshort:chain}): 
the creation of expressive works (Section~\ref{works}), 
data creation (Section~\ref{datacreation}),
dataset collection and curation (Section~\ref{candc}), 
model (pre-)training (Section~\ref{pretraining}), 
model fine-tuning (Section~\ref{fine-tuning}), 
system deployment (Section~\ref{deployment}), 
generation (Section~\ref{generation}), 
and model alignment (Section~\ref{alignment}). 
Each stage gathers inputs from prior stage(s) and hands off outputs to subsequent stage(s), which we indicate with (sometimes bidirectional) arrows.\looseness=-1

The first two stages, the creation of expressive works and data creation, pre-date the advent of generative-AI systems. 
Nevertheless, they are indispensable parts of the production of generative-AI content, which is why we begin our discussion of the supply chain with these processes. 
The following six stages reflect processes that are new for generative-AI systems. 
The connections between these supply-chain stages are complicated. 
In some cases, one stage clearly precedes another (e.g., model pre-training necessarily precedes model fine-tuning), but,
for other cases, there are many different possible ways stages can interact, and they may involve different actors. 
We highlight some of this complexity in the following subsections.

\subsection{The Creation of Expressive Works}\label{works}

Artists, writers, coders, and other creators produce expressive works. 
Gen\-er\-a\-tive-AI systems do, too;\footnote{We discuss this in more detail below with respect to generation (Section~\ref{generation}).    
} 
but state-of-the-art systems are only able to do so because their models have been trained on data derived from pre-existing creative works.\footnote{A data example is not the same as the expressive work. 
	Additionally, some models are trained on synthetic data, typically generated by other generative-AI models~\citep[e.g.]{gokaslan2023commoncanvas}. 
	However, training predominantly on synthetic data is not reflective of current common practices in today's generative-AI systems. 
	Further, there are concerns that training on synthetic data can seriously compromise model quality. 
	See generally Shumailov et al.~\cite{shumailov2023curse} (detailing ``model collapse'' in different generative models). 
} 
It is worth remembering that, historically, the production of most creative works has had nothing to do with ML.\footnote{It appears increasingly likely that some content will be created specifically for model training. 
	For example, hiring photographers to take photographs specifically for model training. 
	Companies like Scale AI already create content (in the form of labels and feedback) specifically for the purpose of training models \citep{scaleai}.
}
Painters have composed canvases, writers have penned articles, etc. without considering how their works might be taken up by automated processes. 
Nevertheless, these works can be transformed into quantified data objects that can serve as inputs for ML.   
They can be easily posted on the Internet and  circulated widely, making them accessible for the development of generative-AI systems.  
As a result, authors and their works are a part of the generative-AI supply chain, whether they would like to be or not (Figure~\ref{fig:talkinshort:chain}, stage 1).\looseness=-1  

\subsection{Data Creation}
\label{datacreation}

Original expressive works are distinct from their datafied counterparts.\footnote{Of course, data examples can still be copies of original works, and thus still infringe intellectual property rights.} 
Data examples are constructed to be computer-readable, such as the JPEG encoding of a photograph. 
For the most part, the transformation of creative content to data formats predates generative AI (Figure~\ref{fig:talkinshort:chain}, stage 2), but all state-of-the-art generative-AI systems depend on it.
They rely on data that coheres with their underlying models' respective modalities:  
text-to-text generation models are trained on digitized text, text-to-image models are trained on both text and images, text-to-music models are trained on text and audio files, and so on. 
This is an important point for our purposes because works that have been transformed into data have been fixed in a tangible medium of expression, and hence are subject to copyright.\footnote{We discuss fixation in Section~\ref{sec:copy-back}. 
    An exception is training data produced by generative-AI systems, as such data currently have been found to not be copyrightable. 
	See \emph{Thaler v. Perlmutter}~\cite{thaler}. 
	We discuss using generations as training data in Section~\ref{generation}.
} 
In turn, generative-AI systems are often are trained on data that include copyrighted expression. 
The GitHub Copilot system involves models trained on copyrighted code,\footnote{Until recently, 
	Copilot was built on top of OpenAI's Codex model.
} 
ChatGPT's underlying models are trained on text scraped from the web,  
Stability AI's Stable Diffusion is trained on text and images, and so on.
For the most part, it is the copyright owners of these datafied individual works who are the potential plaintiffs in a copyright infringement suit against actors at other stages of the supply chain (Section~\ref{sec:copyright}).\looseness=-1

\subsection{Dataset Collection and Curation}
\label{candc} 

The training process for cutting-edge generative-AI models requires vast quantities of data. 
Dataset creators often meet this need by scraping the Internet.\footnote{This is not the only way to collect large amounts of data. 
	See Lee et al.~\cite{lee2023explainers} (discussing other ways datasets may come to be). 
} 
This process involves numerous curatorial choices, including filtering out material that creators do not want to include, 
such as ``toxic speech''~\citep{lee2023explainers}.\footnote{See generally Lee et al.~\cite{lee2023explainers} 
	(discussing dataset creation and curation choices, including toxic content filtering). 
} 
Dataset creators are also necessarily curators.\footnote{This is why we choose to place creation and curation as the same stage in the pipeline. 
    Note, however, that creation and curation do not \emph{always} have to happen together, and may involve different sets of actors. 
    It is also possible for curation to happen after the start of model training, in response to metrics that are observed during the training process. 
    That is, curation could follow (and then also precede further) model (pre-)training (Figure~\ref{fig:talkinshort:chain}, stage 4), or
    model fine-tuning (Figure~\ref{fig:talkinshort:chain}, stage 5).
}\looseness=-1   

With respect to the generative-AI supply chain, there are several points worth highlighting in dataset collection and curation processes 
(Figure~\ref{fig:talkinshort:chain}, stage 3). 
First, while dataset creation and curation can be carried out by the same entities that train generative-AI models, it is common for them to be split across different actors. 
The Stable Diffusion model, for example, is trained on images from datasets curated by the non-profit organization LAION.\footnote{Technically, LAION presents the dataset as a collection the URLs of the images.
	Model trainers visit each URL to collect images for training. 
} 
It is necessary, therefore, to consider the liability of dataset creators separately from the liability of model trainers.\looseness=-1

Second, dataset curation will frequently involve ``the collection and assembling of preexisting materials or of data that are selected, coordinated, or arranged in such a way that the resulting work as a whole constitutes an original work of authorship''~\citep{17usc101}. 
Thus, training datasets can themselves be copyrighted;
copying of the dataset \emph{as a whole} without permission  could constitute infringement, separate and apart from infringement on the underlying works.\footnote{In practice, however, it appears that most uses of training datasets are licensed --- either through a bilateral negotiation or by means of an open-source license offered to the world by the dataset compiler.} \looseness=-1

Third, while a few training datasets include metadata on the provenance of their constitutive data examples, many  do not. 
Provenance makes it easier to answer questions about the  sources a model was trained on, which can be relevant to an infringement analysis. 
It also bears on the ease with which specific material can be located, and if necessary removed, from a dataset. 
However, the use of web-scraping to collect generative-AI training datasets makes provenance difficult to track~\citep{lee2023explainers}. 
Even if a dataset creator releases the dataset itself under a  license, this does not guarantee that the works in the dataset are appropriately licensed,\footnote{Indeed, the creators would have to check that they have abided by each data example's respective license. 
    Some example pairs could potentially have multiple licenses -- e.g., an image and its associated caption could have their own copyrights and licenses.}  
as is currently up for debate with the LAION-5B dataset~\citep{laion, laionpaper, anderson}.\footnote{LAION-5B, a large image-caption dataset, 
was released as under Creative Commons CC-BY 4.0. LAION-5B released a dataset of text captions and URLs to images, instead of the images themselves~\citep{laion, laionpaper}.  
    It is unclear if the LAION team had the rights to license the images within. Notably, the website introducing the LAION dataset provides a feature called ``pwatermark,'' which is a prediction of how likely the image is to contain a watermark. 
    The LAION team estimates that the 6.1\% of the dataset Laion2B-en contains watermarked images.
    Another example comes from the complaint in \emph{Tremblay v. OpenAI, Inc.}~\cite{tremblay}, which alleges that ChatGPT's underlying model(s) were  trained on datasets that do not license the books data that they contain. The complaint alleges that the training data included books from infringing ``shadow libraries'' like Library Genesis. 
	See Complaint at \emph{Tremblay v. OpenAI, Inc.}~\cite[p. 34]{tremblay} 
	But this claim is based on circumstantial evidence, because the datasets it was trained on have not been made public. 
	Text from books have been a key player in other dataset-related complaints. 
	For example, The Pile data was originally released under the MIT license~\citep{pile-datasheet}.
	The Pile was core to the complaint in \emph{Kadrey v. Meta Platforms}~\cite{kadrey}, since the Pile claimed to contain 108GB of the dataset Books3 (which itself contains content from Bibliotek, a popular torrent interface). 
	The original download URL for The Pile (\url{https://the-eye.eu/public/AI/pile/}) is no longer resolving (as of September 2023). 
    LAION has also been taken down from popular hosting services, following a report documenting the presence of CSAM at associated image URLs. 
}\looseness=-1

\subsection{Model (Pre-)Training}
\label{pretraining}

Following the collection and curation of training datasets, it is possible to train a generative-AI model.  
A model trainer\footnote{We distinguish 
	between the person or organization that trains from those that create the model architecture, as they may not be the same.
} 
(Figure~\ref{fig:talkinshort:chain}, stage 4) selects a training dataset, a model architecture (i.e., a set of initialized model parameters), a training algorithm, and a seed value for the random choices made during the training.\footnote{ML uses tools from probability and statistics, which reason about randomness. 
	However, computers are not able to produce truly random numbers. 
	Instead, algorithms exist for producing a sequence of \emph{pseudo}-random numbers. 
	A random seed is an input to a pseudo-random number generator, which enables the reproduction of such a sequence. 
	The trainer also selects hyperparameters~\citep{cooper2021hpo}, which we elide for simplicity. 
} 
The process of transforming these inputs into a trained model is expensive. 
It requires a substantial investment of multiple resources: time, data storage, and compute.
For example, BLOOM (a 176-billion-parameter open-source model from HuggingFace) was trained for 3.5 months, on 1.6 terabytes of text, using 384 GPUs~\citep{bloom-training,  bloom-paper}; 
it cost an estimated \$2-5 million.\footnote{Training costs are often not reported. 
	Even when training cost is reported, development costs (including labor) are often omitted, 
	despite being a critical (and often most expensive) part of overall model development. 
} 
As another point of reference, MosaicML has trained a GPT-3-quality model for less than \$0.5 million.\footnote{The original cost to train GPT-3 is unpublished, though, based on its size, is likely higher than \$0.5 million. 
	MosaicML reports to have trained a GPT-3-\emph{quality} model. 
	This means the model performs to a similar standard as GPT-3 does.
	Nevertheless, MosaicML's model is substantively different from GPT-3. 
	For one, MosaicML's model is much smaller --- 30 billion parameters compared with the original GPT-3 model's 175 billion.
	Additionally, MosaicML trained their model on more data, shifting some of the development cost toward data collection and away from model training.
	It is worth noting that GPT-3 was originally released two years before MosaicML's model was trained,
	and thus the MosaicML training process likely incorporated additional technological improvements. 
	See generally Venigalla and Li~\cite{mosaic-llm} (regarding MosaicML's model). 
	See generally Brown et al.~\cite{brown2020gpt3} (for the size of GPT-3).
} 
Altogether, the dollar cost can range from six to eight figures.\footnote{Further, the training process is not completely automated; training often requires people to monitor and tweak the model. 
    For example, model trainers typically run evaluation metrics on the model while it is being trained, in order to assess the progress of training. 
    Google's TensorBoard~\citep{tensorboard} and software from Weights \& Biases~\citep{weights-biases} are two tools for running evaluation metrics and monitoring during training. 
    Depending on these metrics (which attempt to elicit how ``useful'' or ``good'' the model is, but are not comprehensive~\citep{lee2023explainers}) model trainers may pause the training process to manually revise the training algorithm (e.g., change the hyperparameters)  or the dataset, which we indicate with bidirectional arrows at Figure~\ref{fig:talkinshort:chain}, stages 3-4.  
    Human intervention in response to metrics necessarily makes model training an iterative process. 
}\looseness=-1

The output of the training process is typically called a \textbf{pre-trained model} or \textbf{base model}.\footnote{Others use the term ``foundation model.''
       The term ``foundation'' can be easily misunderstood. 
       It should not be interpreted to connote that ``foundation models'' contain technical developments that make them fundamentally different from models produced in the nearly-a-decade of related prior work. 
       The term itself has been met with controversy within the ML community, which can be seen expressed on programming forums and in conversations, e.g., we refer to a Twitter thread (and its associated offshoots) that involves renowned researchers and some of the Stanford authors that coined the term ``foundation models.''
	   (\emph{See} \url{https://twitter.com/tdietterich/status/1558256704696905728}).  
} 
A base model has many possible futures. 
It could sit idly in memory, collecting figurative dust.\footnote{This reveals 
	the murky line between what exactly is a program and what exactly is data in ML, more generally. 
	The set of parameters can be viewed as a \emph{data structure} containing vectors of numbers that, on its own, does not \emph{do} anything. 
	However, we could load that data structure into memory and apply some relatively lightweight linear algebra operations to produce a generation.  
	In this respect, we could also consider the model to be a program (and, indeed, an algorithm). 
	The model, if given a prompt input, can also be executed like a program. 
	Note that the term ``model'' is overloaded; 
	it can be used to refer to the model parameters (vectors of numbers) or to the model as a combination of software and the model parameters, which together can be executed like a program.
} 
The model could be uploaded to a public server,\footnote{For example, HuggingFace hosts a repository of over 300,000 open-sourced models~\citep{hf-models}. 
} 
allowing others to download it and use it however they want.\footnote{They could 
	fine-tune the model (Section~\ref{fine-tuning}), 
	embed the model in a system that they deploy for others to use (Section~\ref{deployment}), 
	produce generations (Section~\ref{generation}), 
	align the model (Section~\ref{alignment}), 
	or do some subset of these other stages of the supply chain. 
	From this example, we can see how the supply chain is in fact iterative, which we illustrate in Figure~\ref{fig:talkinshort:chain}.
} 
The model could be integrated into a system and deployed as a public-facing application (Section~\ref{deployment}),  which others could use directly to produce generations (Section~\ref{generation}). 
Or, the model could be further modified by the initial model trainer, by another actor at the same organization, or, if made publicly available, a different actor from a different organization. 
That is, another actor could take the model parameters and use them as the input to do additional training with new or modified data.  
This possibility of future further training of a base model is why this stage of the supply chain is most often referred to as \textbf{\emph{pre}-training}, 
and why a base model is similarly often called a  \textbf{pre-trained model}. 
Such additional training of the base model is called \textbf{fine-tuning}. 

\subsection{Model Fine-Tuning} 
\label{fine-tuning}

Base models trained on large-scale, web-scraped datasets are not typically optimized to apply specialized domains of knowledge. 
For example, an English text-to-text base model may be able to capture general English-language semantics, but not able to reliably apply detailed scientific information about molecular biology.

This is where fine-tuning comes in (Figure~\ref{fig:talkinshort:chain}, stage 5).
Fine-tuning is the process of modifying a preexisting model and making it better along some dimension of interest. 
This process often involves training on additional data that is more aligned with the specific goals.\footnote{And thus the reason for the bidirectional arrow between stages 3 and 5 in Figure~\ref{fig:talkinshort:chain}. 
	Similar to pre-training, monitoring metrics during fine-tuning may lead to further dataset curation (Section~\ref{pretraining}).
} 
If we think of training as transforming data into a model, fine-tuning transforms a model into another model.
Fine-tuning essentially involves just running more training. 
However, fine-tuning and pre-training may use different inputs, which ultimately makes the trajectories and outputs of their respective training processes very different.\footnote{There are other relevant factors in training, including choice of hyperparameters and choice of hardware. 
	These, too, can change between pre-training and fine-tuning. 
	We again elide these details for simplicity.   
} 
To add more precision to our previous statement: fine-tuning transforms a model into another model, while incorporating more data.\looseness=-1

\paragraph{Forks in the supply chain.} 
Two important observations follow from our description of fine-tuning as (effectively) just performing more training. 
For one, a model trainer does not have to fine-tune at all. 
Prior to fine-tuning, there is a fork in the generative-AI supply chain with respect to the possible futures of the base model after pre-training (stage 4):  
One could take the output base model from pre-training, and use this model directly as the input for system deployment (stage 6), 
generation (stage 7), 
or model alignment (stage 8). 
Alternatively, it is possible to perform multiple separate passes of fine-tuning --- 
to take an already-fine-tuned model, and use it as the input for another run of fine-tuning on another dataset.\footnote{In this respect, it is important to note that a model is a ``base'' or ``fine-tuned'' model \emph{only in relation to other models}.
    These terms do not capture inherent technical features of a model; instead, they describe different processes by which a model can be created.} 

For each possibility, there can be different actors involved. 
Sometimes, the creator of a model also fine-tunes it. 
Google's Codey models (for code generation) are fine-tuned versions of Google's PaLM 2 model~\citep{codey}. 
In other cases, when a model's weights are publicly released (as Meta has done with its Llama family of models)~\citep{llama, llama2, codellama},
others can take the model and independently fine-tune them for particular applications. 
A Llama fine-tuner could release their model publicly, which in turn could be fine-tuned by another party.\footnote{To give a concrete example of the many actors in the generative-AI supply chain, consider Vicuna.
    LMSYS Org fine-tuned Meta's Llama model on the crowd-sourced ShareGPT dataset to produce Vicuna~\citep{vicuna, sharegpt}. 
    ShareGPT is a crowd-sourced dataset composed of conversational logs of user interactions with ChatGPT. 
    It contains both content created by users and by the generative-AI model embedded in ChatGPT (either GPT-3.5 or GPT-4, depending on the user)~\citep{sharegpt}. 
    Vicuna has also released their model publicly, affording a potentially infinite host of actors the ability to fine-tune the model on additional data. See Raffel~\cite[slide 15]{raffel2021os-talk} (for a figure showing many fine-tuned models building on one base model).
}  
To use a copyright analogy, a fine-tuned model is a derivative of the model from which it was fine-tuned; 
a repeatedly fine-tuned model is a derivative of the (chain of) fine-tuned model(s) from which it was fine-tuned.\looseness=-1 

It is helpful to make the base-/fine-tuned model distinction 
because different parties may have different knowledge of, control over, and intentions toward choices like which data is used for training and how the resulting trained model will, in turn, be put to use. 
A base-model creator, for example, may attempt to train the model to avoid generating copyright-infringing material.
However, if that model is publicly released, someone else may attempt to fine-tune the model to remove these anti-infringement guardrails. 
A full copyright analysis may require treating them differently and analyzing their conduct in relation to each other (Section~\ref{direct}).\looseness=-1

\subsection{Model Release and System Deployment}
\label{deployment}

It is possible to release a model or deploy it as part of a larger software system, 
use the model to produce generations (Section~\ref{generation}),
or to take the trained model and further alter or refine it via model alignment techniques (Section~\ref{alignment}). 
In brief, there is a complicated interrelationship between the deployment, generation, and alignment stages.
They can happen in different orders, in different combinations, and at different times for different generative-AI systems. 
For purely expository purposes, we present them one at a time, starting with \textbf{model release} and \textbf{system deployment} (Figure~\ref{fig:talkinshort:chain}, stage 6).\looseness=-1 

A model is open-source \textbf{released} when its model parameters are uploaded to a server or platform (like HuggingFace~\citep{hf-models}), from which others can download it.\footnote{Meta first asked interested parties to request Llama's model parameters, rather than uploading them publicly on the web. 
	However, Llama's model parameters were quickly leaked on the website 4chan~\citep{llama-leak}. 
	This incident shows how challenging it can be to control access to models once released. 
	Llama also includes a use policy in the Llama 2 Community License that outlines prohibited uses of the model. 
	Of course, it is impossible to enforce prohibited uses when releasing model parameters.
	This is also why many model trainers choose to release models through hosted services. 
	See Llama 2~\cite{llama-use} (for the Llama 2 Community License). 
} 
Released models, which include Meta's Llama family of models~\citep{llama, llama2, codellama} and Stable Diffusion~\citep{rombach2022diffusion} give others direct access to their parameters.
Developers can write their own code to produce generations, or alter the model through fine-tuning or model alignment (Section~\ref{alignment}). 

In contrast, closed-source models are not directly available to external users.
They are typically embedded in large, complex software systems, which are \textbf{deployed} to both internal and external users through software services. 
For example, a model could be hosted by a company (e.g., OpenAI, Stability AI, or Google).
It could be used internally to support various services  (e.g., Google has integrated an internally-developed LLM into Google Search), or released as a hosted service that gives external users access to generative-AI functionality.\looseness=-1

External-facing services can be deployed in a variety of forms, and \emph{do not} typically include the ability to change the model's parameters. 
They can be browser-based user applications (e.g., ChatGPT, Midjourney, DreamStudio), or public (but not necessarily free) APIs for developers (e.g., GPT models, Cohere).\footnote{Another deployment option is a command-line interface (CLI), 
	which takes a user-supplied prompt as input (via a code terminal) and directly returns the resulting generation as output. 
	\url{https://ollama.ai/} (the download link of the Ollama CLI, which is a wrapper program around various Llama-family LLMs).
} 
Some model trainers provide a combination of release and deployment options.
For example, DreamStudio is a web-based user interface~\citep{dreamstudio} built on top of services hosted by Stability AI~\citep{stability}; 
the DreamStudio application gives external users access to a generative-AI system that contains the open-source Stable Diffusion model~\citep{rombach2022diffusion}, which Stability AI also makes available for direct download.\footnote{It is possible that models released and deployed in multiple ways might not all be exactly the same; they could have different versions of model parameters. 
	This may be made explicit to users, as with ChatGPT, or may not be communicated to them, and thus unclear or unknown. 
	See generally OpenAI~\cite{chatgpt} (regarding both GPT-3.5 and GPT-4 model integration into the ChatGPT web application). 
}

This is a familiar spectrum from Internet law, from cloud-hosted services at one end to fully open-source software at the other, with closed-source apps in between. 
These deployment methods offer varying degrees of customization and control on the part of the deployer and the user.
For example, a generative-AI system deployed as a service will often modify the user-supplied prompt before inputting it to the model. 
Several applications (e.g., ChatGPT, Gemini, and Sydney), add additional instructions (``application prompts'') to the user's input to create a compound prompt~\citep{zhang2023prompts, chatgpt-custom}.\footnote{See generally Zhang and Ippolito~\cite{zhang2023prompts} (which discovers proprietary system prompts); 
    OpenAI~\cite{chatgpt-custom} (announcing a  ChatGPT feature that allows users to provide their own additional prompts, which get appended to their future inputs to create compound prompts).
}
The additional instructions change the behavior of the model's output on a user prompt.\footnote{This kind of prompt transformation is another technique for steering the behavior of a model.} 
For example, compare the following two application prompts: 
\prompt{I want you to act as an English translator, spelling corrector and improver\ldots} and \prompt{I want you to act as a poet. You will create poems that evoke emotions and \;\; have the power to stir people’s soul\ldots}~\citep{chatgpt-custom-prompts}.\footnote{See OpenAI~\cite{chatgpt-custom-prompts} (These prompts and more can be found on this site); 
	DAIR.AI~\cite{promptingguide} (This handbook provides an introduction to creating prompts for large language models); 
	OpenAI~\cite{chatgpt-custom}. 
}

Typically, model trainers and owners maintain the most control over models deployed through hosted services and the least over models released as model parameters~\citep{llama-leak}. 
By embedding a model within a larger system, they can imbue it with additional behaviors~\citep{cooper2021eaamo}. 
For example, APIs and web applications allow deployers to  filters a model's inputs or outputs. 
For example, ChatGPT will often respond with some version of: ``I'm really sorry, but I cannot assist you with that request,'' when its ``safety'' filters are tripped.\footnote{These filters may detect undesired inputs and prevent the model from generating an output, or detect undesired outputs and prevent the system from displaying the generation.
	In both cases, the model parameters would not be changed. 
	This need not be the case, the model parameters may also be directly modified through alignment to respond to undesired inputs in a more desirable way.
	Of course, though, for ChatGPT, we do not know exactly how filters are implemented. 
}
GitHub Copilot expressly states that it uses ``filters to block offensive words in the prompts and avoid producing suggestions in sensitive contexts''~\citep{copilot-safety-filter}. 
Additionally, some services include output filters to avoid generating anything that looks too similar to a training example~\citep{copilot-copy-filter}.\footnote{See 
	\url{https://news.ycombinator.com/item?id=33226515} (for related discussion on the Hacker News forum). 
}
Unfortunately, output filtering is an imperfect process.(See Section~\ref{similarity}).\footnote{Each mechanism for making model functionality widely available has different pricing structures that can ultimately impact the quality of the model. 
    While the open-source community works hard to create and release models that compete with the best closed-source models, current open-source models are mostly trained on open-sourced data and are often lower quality. 
    The best open-sourced models are very good, but still not as good as closed-source proprietary models. 
	For example, Technology Innovation Institute in Abu Dhabi recently released the model, Falcon 180B (a 180 billion parameter model), which they claim is better than Meta's Llama 2 but still behind GPT-4 \citep{falcon180b}.
    Additionally, differences between open- and closed-source datasets can lead resulting trained models to vary in quality. 
    For example, Min et al.~\cite{min2023silo} uses public domain and permissively licensed text to train a language model, and demonstrates a degradation in quality in domains that are not well represented in the data. 
    Additionally, data in the public domain can be unrepresentative of certain demographic groups~\citep{levendowski2018copyright}.
}\looseness=-1 

\subsection{Generation}
\label{generation}

Generative-AI models produce output generations in response to input prompts.\footnote{See Section~\ref{pretraining} (noting, however, that models do not \emph{have to} be used to produce generations).
} 
While a few users produce generations from open-source models by writing code to interact with the model parameters to execute the generation process,\footnote{See Section~\ref{pretraining}  
	(discussing how the term ``model'' is overloaded, and can refer to model parameters being embedded in a program that executes (typically linear algebra) operations to to perform generation. )
}
most users a interact with models only indirectly, through an API, web service, or application.

Users can affect generations in a few ways.
First, there is the \emph{prompt itself}. 
Some prompts, like \prompt{a big dog}, are simple and generic.
Others, such as \prompt{a big dog facing left wear\-ing a space\-suit in a bleak lunar landscape with the earth rising  as an oil \\paint\-ing in the style of  Paul Ce\-zanne}, are more detailed. 
Second, there is the \emph{choice} of which  deployed system to use (which  embeds an implicit choice of model). 
For example, a user that wants to perform text-to-image generation on a browser-based interface needs to select between Ideogram, \dalle-2, Midjourney, and other publicly available text-to-image applications that could perform this task. 
A user typically selects an application with the outputs partially in mind, so that one choice or another can indicate an attitude towards the possibility of infringement.  
Users may also revise their prompt to attempt to create generations that more closely align with their goals. 
And, third, there is \emph{randomness} in each generation.\footnote{For generative models, there are many reasonable outputs for the input. 
	There are also other sources of randomness in generation that are implementation-specific, such as the choice of decoding strategy for language models. 
	See Riedl~\cite{riedl2023transformers} (for an accessible discussion of decoding). 
}
It is typical, for example, for image applications to produce several candidate generations. 
\dalle-2, Midjourney, and Ideogram all do this.\looseness=-1

As we will see, characterizing the relationship between the user and the chosen deployed system is one of the critical choice points in a copy\-right-in\-fringement analysis.
There are at least three ways the relationship could be described:\footnote{We focus on deployed systems --- and their API and web-based interfaces --- because there are more opportunities for the deployer to control the model. 
	But, of course, the user could have written some code to produce generations using released open-source model parameters.
}
\begin{itemize}
	\item The user actively drives the generation through choice of prompt, and the system passively responds.
	In this view, the user is potentially a direct infringer, but the application is like a web host, ISP, or other neutral technological provider.
	\item The system is active and the user passive.
	In this view, the user is like a viewer of an infringing broadcast, or the unwitting buyer of a pirated copy of a book.
	Primary copyright responsibility lies with the deployed system, and possibly with others further upstream in the generative-AI supply chain.
	\item The user and system are active partners in generating infringing outputs.
	In this view, the user is like a patron who commissions a copy of a painting; the system is like the artist who executes it.
	They have a shared goal of creating an infringing work.\looseness=-1
\end{itemize}
We will argue that there is no universally correct characterization. 
Which of these three is the best fit for a particular act of generation will depend on the system, the prompt, how the system is marketed, and how users can interact with the system's interfaces.\footnote{These three options highlight additional observations about prompts. 
Thus far, we have primarily discussed generations as expressive works, but prompts could also be expressive works. The expressive example we gave above was: \prompt{a big dog facing left wearing a spacesuit in a bleak lunar landscape with the earth rising in the background as an oil painting in the style of Paul Cezanne high-resolution aesthetic trending on artstation}.  
Sufficiently expressive prompts written by the direct user of a service could be subject to copyright. 
Context windows are so large, it is even possible for the user to prompt with an entire expressive work. 
As we discuss below in our copyright analysis,  it is of course possible for this expressive work to have also been authored by another individual. Prompts could also be produced by generative AI, but this does not have the same authorship considerations.  
For example, Anthropic's team discussed using the entire text of \emph{The Great Gatsby} as a prompt to demonstrate the long context window of their language model, Claude~\citep{context}. 
While \emph{The Great Gatsby} is now in the public domain, it is easy to imagine another book entered as the prompt, or a copyrighted image as the prompt in an image-to-image system. Or copyrighted audio as input to an audio-to-audio model, etc. 
User-supplied prompts may be stored on system-deployers' servers for non-transient periods of time, and may even serve training data for a future model. 
Such prompts may also be used in model alignment (Section~\ref{alignment}).
} 

\paragraph{Forks in the supply chain.} 
There is a loop from generation back to the beginning of the supply chain. 
While not the most common contemporary practice, it is possible to use generations as training data for generative-AI models.\footnote{Using model outputs as training data for future models has been a common practice in other settings.
	For instance, back-translation, the process of using a machine-translation model to generate additional training data (by translating data from one language to another) is a  common technique~\citep{sennrich2016backtranslation}.
} 
In this case, generation serves simultaneously as the creation of expressive works (i.e., stage 1) and data creation (i.e., stage 2) 
and generations can become inputs to dataset collection and curation processes (i.e., stage 3), 
which we indicate with an arrow in Figure~\ref{fig:talkinshort:chain}. 
As we discuss in Section~\ref{sec:copyright}, this potential circularity also has implications for copyright.\footnote{There are also concerns that this practice can have negative effects on model quality~\citep{shumailov2023curse}. 
}

Alternatively, for the process of generation, some generative-AI systems interact with \emph{external} deployed services,  
as is the case with ChatGPT plugins~\citep{chatgpt-plugins}.  
Such interactions between external services and generation further complicate the generative-AI supply chain that we depict in Figure~\ref{fig:talkinshort:chain}. 
In particular, by potentially integrating with other systems, the generation stage could implicate an entirely separate, unspecified number of supply chains consisting of entirely different organizations and actors. 
This, too, raises important copyright implications (what if news articles or short stories are integrated by the plugin?).

\subsection{Model Alignment}
\label{alignment}

The generative-AI supply chain does not stop with generation. 
As discussed above, model trainers try to improve models during both pre-training and fine-tuning. 
For pre-training, they monitor evaluation metrics, and may pause or restart the process to alter the datasets and algorithm used (Section~\ref{pretraining}); 
for fine-tuning, they continue training the base model with data that is specifically relevant for a particular task (Section~\ref{fine-tuning}). 
Both of these base model modifications 
are coarse: They make adjustments to the dataset and algorithm, 
and do not explicitly incorporate information into the model about whether specific generations are ``good'' or ``bad,'' according to user preferences.\footnote{Of course, words like ``good'' and ``bad'' can have multiple valences, 
	and resist the kind of quantification on which ML  depends. 
	  See Lee et al.~\cite{lee2023explainers} (discussing the challenges of defining ``good'' and ``bad'' in the context of model behavior). 
}

There is a whole area of research, called \textbf{model alignment}, that attempts to meet this need~\citep{chatgpt-alignment}.\footnote{See OpenAI~\cite{chatgpt-alignment} (for an introduction to InstructGPT, a model that is aligned with human feedback).
} 
The overarching aim of model alignment is to \emph{align} model outputs with specific generation preferences (see Figure~\ref{fig:talkinshort:chain}, stage 8). 
Currently, the most popular alignment technique is called \textbf{reinforcement learning with human feedback (RLHF)}~\citep{christiano2017rlhf, ouyang2022instructgpt}.  
As the name suggests, RLHF combines collected human feedback data with a (reinforcement learning) algorithm in order to update the model. 
Human feedback data can take a variety of forms, which include user ratings of generations. 
For example, such ratings can be collected by including thumbs-up and thumbs-down buttons in the application user interface, which are intended to query feedback about the system's output generation. 
In turn, the reinforcement learning algorithm uses these ratings to adjust the model --- to encourage more ``thumbs-up'' generations and fewer ``thumbs-down'' ones.\footnote{In the reinforcement learning setting, data is not labeled as explicitly as it is in discriminative setting, 
	e.g., our example of an image classifier, where each training data image has a label of either \texttt{cat} or \texttt{dog}. 
	Instead, generations may be labeled ``good'' or ``bad'' based on human feedback, and the reinforcement learning algorithm updates the model in response to that feedback. 
	In RLHF, feedback is generated by a person interacting with the system; however, RL can also use feedback automatically generated by an algorithm specification~\citep{bai2022constitutional}. 
} 

Future training and alignment on the model may include both the inputted prompt and the generation in addition to the feedback provided. 
As discussed in the prior section, user-supplied prompts may include copyrighted content created by either the user themselves or by another party. 
Most generative-AI companies begin model alignment prior to deployment or release (Section~\ref{deployment}). 
In this respect, model alignment complements other techniques, like input-prompt and output-generation filtering (Section~\ref{generation}).\footnote{Before making models publicly available, these companies contract with firms, like Scale AI~\citep{scaleai}, that simulate the user feedback process. 
    These firms typically employ people to label generations as ``good'' or ``bad,'' according to guidance from the generative-AI company. 
    In general, the process of model alignment is a critical part of the supply chain. 
    It serves as a mechanism for steering models away from generating potentially harmful outputs (See Cole~\cite{404mushroom}, 
	describing a book on mushroom foraging built from generations, which mistakenly indicate that toxic mushrooms are safe to eat) 
    and toward the policies of the company or organization that deployed the model. See Google~\citep{bard}, OpenAI~\citep{openai-safety-policy}, Ganguli et al.~\citep{deep2023correction} (documenting safety considerations, alignment, and RLHF at Google, OpenAI, and Anthropic).
}\looseness=-1 

%% file: section/40-genai/43-talkinshort/4330-talkinshort-copyright.tex
\section{Copyright and the Supply Chain}
\label{sec:copyright}

The hornbook statement of United States copyright doctrine is that original works of authorship are protected by copyright when they are fixed in a tangible medium of expression. 
A defendant directly infringes when they engage in conduct implicating one of several enumerated exclusive rights (reproducing, publicly distributing, etc.), 
with a work of their own that is substantially similar to a copyrighted work 
because it was copied from that work. 
Other parties may be held secondarily liable for conduct that bears a sufficiently close nexus to the infringement under one of several theories. 
Otherwise infringing conduct is legal when it is protected by one of several defenses, including the DMCA Section 512 safe harbors,  
fair use, 
or an express 
or implied license. 

In this section, we first provide some brief background on what kinds of works copyright applies to (Section~\ref{sec:copy-back}). 
We then apply aspects of the above orthodox, uncontested statement of copyright law to the generative-AI supply chain. 
We address issues of rights (Section~\ref{exclusive}), infringement (Sections~\ref{similarity} \&~\ref{direct}), and fair use (Section~\ref{fairuse}). 
We defer discussion of safe harbors, licenses, paracopyright liability, and remedies to the longer version of our article~\citep{lee2023talkin}. 
Our goal is to be careful and systematic, not to say anything dramatically new. 

\input{section/40-genai/43-talkinshort/4331-talkinshort-authorship}
\input{section/40-genai/43-talkinshort/4332-talkinshort-rights}
\input{section/40-genai/43-talkinshort/4333-talkinshort-similarity}
\input{section/40-genai/43-talkinshort/4334-talkinshort-direct}
\input{section/40-genai/43-talkinshort/4335-talkinshort-fair}

%% file: section/40-genai/43-talkinshort/4331-talkinshort-authorship.tex
\subsection{What is copyrightable?}
\label{sec:copy-back}

Copyright protects ``(1) original works of authorship (2) fixed in any tangible medium of expression''~\citep{17usc102}.\footnote{17 U.S.C. § 102(a) (numbering added).}
``Original, as the term is used in copyright, means only that the work was independently created by the author (as opposed to copied from other works), and that it possesses at least some minimal degree of creativity''~\citep[p. 345]{feist} 
Fixation is satisfied when the work is embodied in a tangible object in a way that is ``sufficiently permanent or stable to permit it to be perceived, reproduced, or otherwise communicated for a period of more than transitory duration''~\citep{17usc101}.\footnote{17 U.S.C. § 101 (definition of ``fixed'').)} 

We start with fixation. 
Unfixed works have no interaction with the gen\-er\-a\-tive-AI supply chain. 
A work must be fixed to be used as training data.
Truly ephemeral creations, like unobserved dances and songs that are never recorded, will never be captured in a way that can be used as an input to a training algorithm. 
Datasets, models, applications, prompts, and generations are all fixed in computers and storage devices. 
Once it is fixed, however, any kind of original expression can be used as an input for generative AI. 

The originality requirement distinguishes material that was created by a human author from facts that ``do not owe their origin to an act of authorship''~\citep[p. 347]{feist}. 
In addition, some types of material are never copyrightable, including any ``idea, procedure, process, system, method of operation, concept, [or] principle''\footnote{17 U.S.C. § 102(b).} 
In practice, this means that the copyright in some works (e.g., product photographs) will be ``thinner'' and protect fewer aspects of the works than the ``thicker'' copyrights in others (e.g, abstract art), because the ``range of creative choices that can be made in producing the works is narrow''~\citep[p. 1120]{rentmeester}. 
In particular, any copyright in computer software --- which is treated as a ``literary work'' for copyright purposes --- typically excludes a great deal of functional material, such as coding conventions required by the choice of programming language~\citep{samuelson2016functionality}.  
As a result, some individual training examples are uncopyrightable. 
(For example, birdsong-recognition AIs are trained on recordings of birds~\citep{kahl2021birdnet, naruto}.\footnote{See  Kahl et al.~\cite{kahl2021birdnet}. Animals are not recognized as ``authors'' for copyright purposes. See \emph{Naruto}~\cite{naruto}. 
})
But other items are copyrightable, and those copyrights will be held by a variety of authors.\looseness=-1

Training datasets will include different amounts and proportions of copyrighted material. 
A dataset of birdsong recordings will be almost entirely copyright-free, but a dataset of  illustrations will contain numerous copyrighted works. 
Further, datasets \emph{themselves} may be copyrightable as \textbf{compilations}~\citep{17usc103},\footnote{17 U.S.C. § 103(a).}  
``formed by the collection and assembling of preexisting materials or of data''~\citep{17usc101}.\footnote{17 U.S.C. § 101 (definition of ``compilation'').} 
A compilation is copyrightable as such when it features a sufficiently original ``selection or arrangement''~\citep[p. 348]{feist}. 
Originality in selection is choosing \emph{what to include} in the dataset; originality in arrangement is choosing \emph{how to organize} it. 

Generations raise a doctrinal question that has been debated for decades: who, if anyone, owns the copyright in the output of a computer program~\citep{samuelson1985allocating}? 
Although some have argued that the program itself should be regarded as the author, computer authorship is squarely foreclosed by U.S. copyright law~\citep{grimmelmann2016authored}. 
So far, the courts have held firm to this line for AI generations. 
\emph{Thaler}~\citep{thaler} upheld the Copyright Office's refusal to register copyright in an image allegedly ``autonomously created by a computer algorithm running on a machine.'' 
The Copyright Office had held that the image lacked human authorship, and the court agreed.\footnote{That is, programs, like animals, are not ``authors'' within the meaning of the Copyright Act.}  
The author of a generation --- if anyone --- is some human connected to the generation. 
The four immediately relevant possibilities are (1) author(s) whose works the model was trained on, (2) some entity in the generative-AI supply chain (e.g., the model trainer or fine-tuner;  application developer), (3) the user who prompted a service for the specific generation, or (4) no one.
As between these four possibilities, there is no one-size-fits-all answer. All four arise in actual generative-AI applications.\looseness=-1

%% file: section/40-genai/43-talkinshort/4332-talkinshort-rights.tex
\subsection{The Exclusive Rights}
\label{exclusive}

Copyright includes five relevant exclusive rights: reproduction, adaptation, public distribution, public performance, and public display.\footnote{17 U.S.C. § 106}
Every stage in the generative-AI supply chain requires a reproduction and thus potentially implicates copyright. Because the remedies for infringement of a work are the same, regardless of whether the defendant violated one exclusive right or several, the precise dividing lines are often unimportant. 
We examine the adaptation right, and defer additional discussion to other work.\looseness=-1 

The adaptation right gives the copyright owner the exclusive right to ``to prepare derivative works based upon the copyrighted work.''\footnote{17 U.S.C. §~106(2)}
A derivative work combines the authorship in an existing (or ``underlying'') work with new authorship.
In a compilation (Section~\ref{sec:copy-back}), the underlying works are present in substantially unmodified form, whereas in a derivative work the underlying work is ``recast, transformed, or adapted.''\footnote{17 U.S.C. § 101 (definition of ``derivative work).}\looseness=-1
The adaptation right makes clear that copyright extends beyond literal similarity to incorporate changes of form, genre, and content such as translations, sequels, and film adaptations~\citep{gervais2013derivative,gervais2022aiderivatives, samuelson2013derivative}. 

A training dataset is probably not a derivative work of any of the works in it; it is more appropriately classified as a compilation ``formed by the collection and assembling of preexisting materials''~\citep{17usc101}. 
To the extent that a model is similar to a work it was trained on, it is a derivative work because it is ``based on'' its training data. (Section~\ref{similarity}).
Similarly, a prompt could be a reproduction or derivative of an existing work (as when a diffusion model is prompted with an image to infill)~\citep{context}.
And generations are frequently derivative works of works in the training data or prompts, again subject to similarity.

%% file: section/40-genai/43-talkinshort/4333-talkinshort-similarity.tex
\subsection{Substantial Similarity}
\label{similarity}

Substantial similarity is a qualitative, factual, and frustrating question.
Two works are substantially similar when ``the ordinary observer, unless he set out to detect the disparities, would be disposed to overlook them, and regard their aesthetic appeal as the same''~\citep[p. 489]{peterpan}. 
A common test is a ``holistic, subjective comparison of the works to determine whether they are substantially similar in total concept and feel''~\citep[p. 1118]{rentmeester}. 
This is not a standard that can be reduced to a simple formula that can easily be applied across different works and genres.\footnote{But see Scheffler et al.~\cite{scheffler2022formalizing} (describing a principled computational basis for comparing works)} 
We discuss base models and generations below, and defer discussion of data, datasets, fine-tuned models, aligned models, and deployed services to other work. 

\subsubsection{Pre-Trained/Base Models} A model is different in kind from the copyrightable works it was trained on. 
No viewer would say that the model has the same ``total concept and feel'' as a painting; no reader would say that it is substantially similar to a blog post; and so on. 
That said, the Copyright Act does not require that copies be directly hu\-man-intell\-i\-gi\-ble to infringe. 
A Blu-Ray is not directly intelligible by humans, either, but it counts as a ``copy'' of the movie on it.
Indeed, all digital copies are unintelligible. 
Instead, they are objects ``from which the work can be perceived, reproduced, or otherwise communicated \ldots \emph{with the aid of a machine or device}''~\citep{17usc101}.
Thus, even if a model is uninterpretable, it might still be possible to ``perceive[]'' or ``reproduce[]''  a copyrighted work embedded in its parameters through suitable prompting.
Indeed, there is substantial evidence that many models have memorized copyrighted materials~\citep{carlini2021extracting, carlini2023extracting}.\footnote{See Carlini et al.~\cite{carlini2021extracting} (GPT-2 memorizes training data); Carlini et al.~\cite{carlini2023extracting} (Stable Diffusion and Imagen memorize images); Chang et al.~\cite{chang2023speak} (suggestive evidence that GPT-4 memorizes training data).
}  
For example, Carlini et al.~\cite{carlini2023extracting} shows how Stable Diffusion has memorized photographs.\looseness=-1

\begin{figure}[t!]
    \begin{minipage}{.4\linewidth}
    	\centering
    	\includegraphics[width=.92\linewidth]{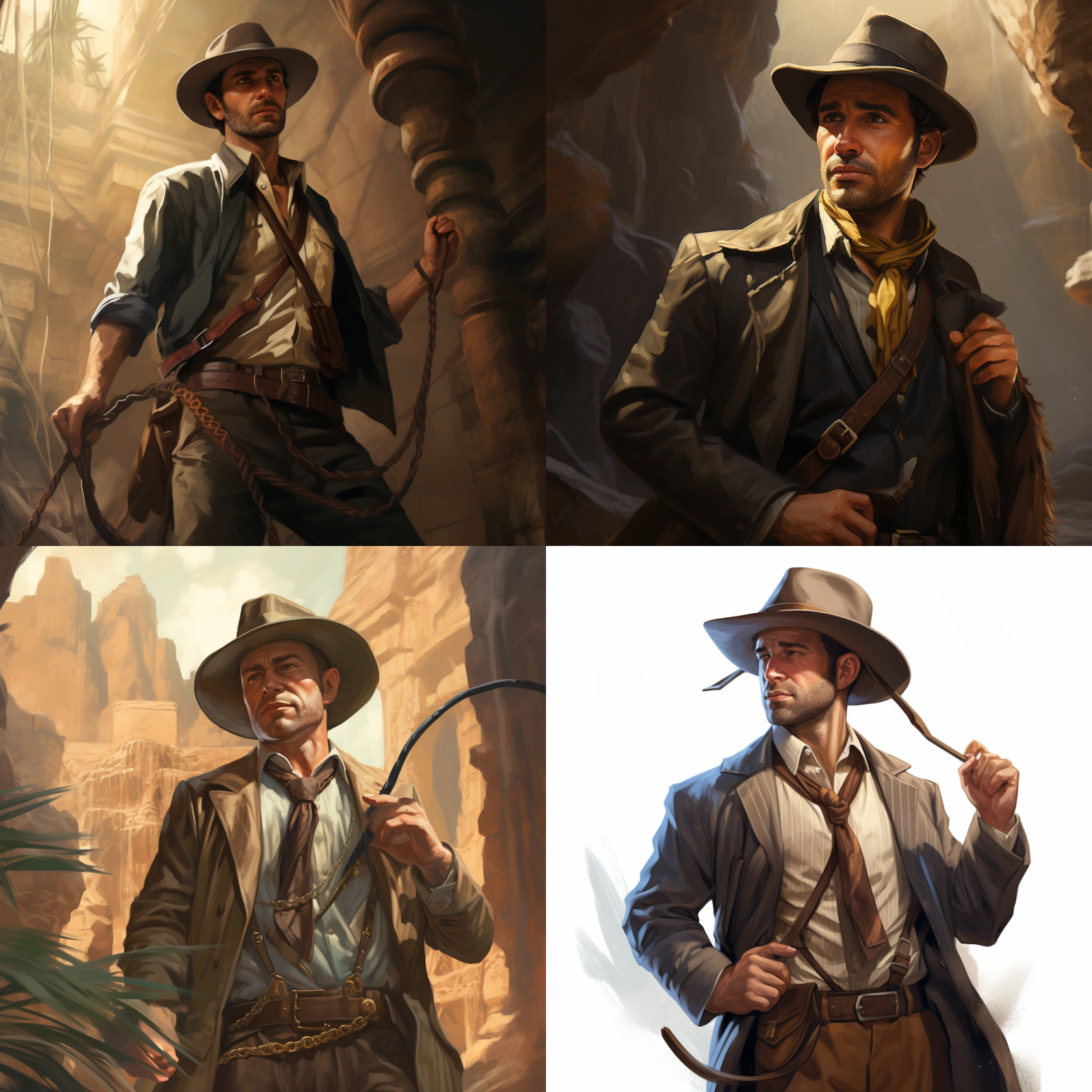}
    	\subcaption{\prompt{an adventurous arch\-ae\-ologist with a whip and a fedora}}
	\label{fig:archaeologist-hires}
    \end{minipage}
    \hfill
    \begin{minipage}{.4\linewidth}
        \centering
    \vspace{-.9cm}
    	\includegraphics[width=.92\linewidth]{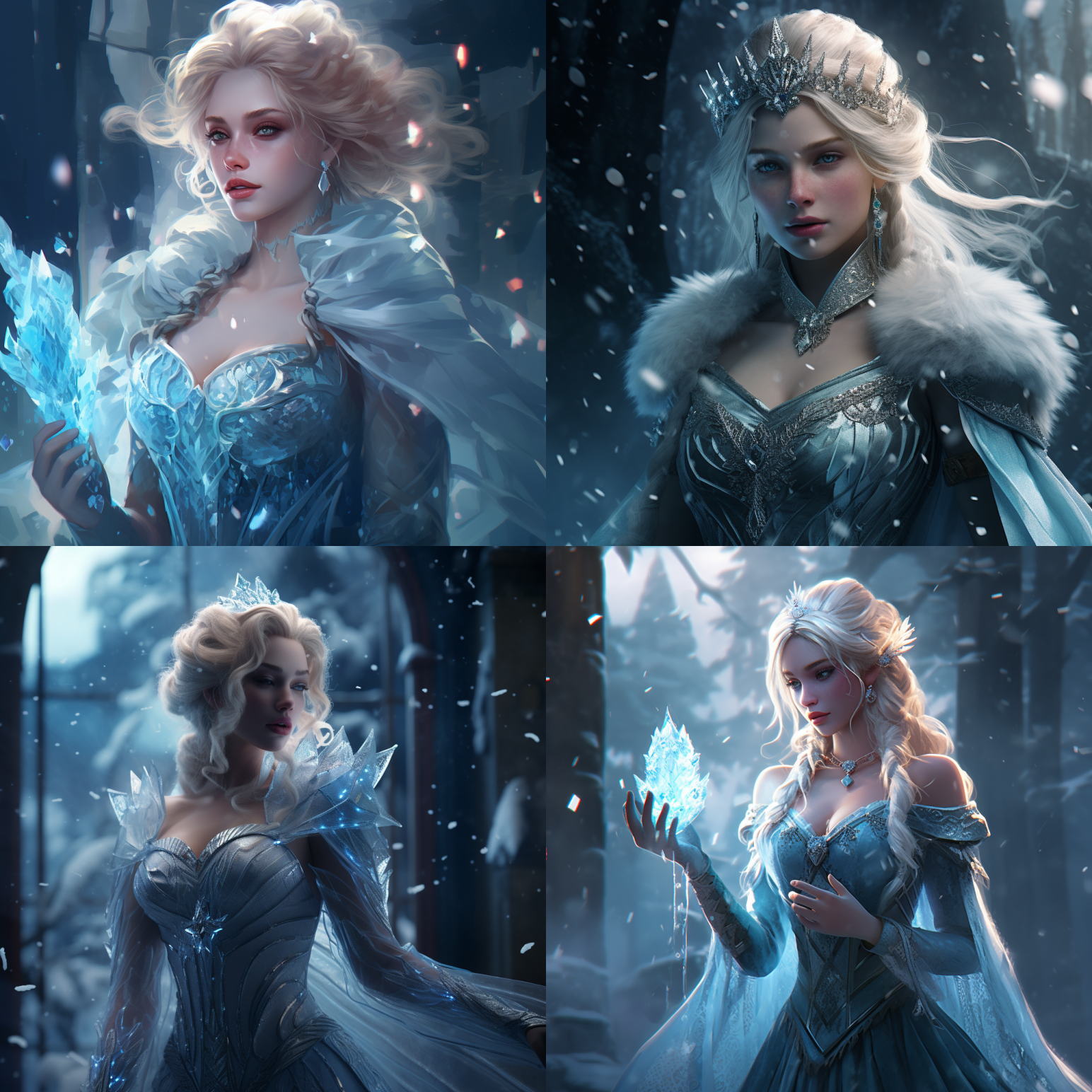}
	\subcaption{\prompt{ice princess}}
	\label{fig:ice-princess}
    \end{minipage}
    \caption{Generated by the authors using Midjourney.}
\end{figure}

A model might memorize more works or fewer~\citep{carlini2023extracting, carlini2023quantifying}.
But at least some models memorize at least some works closely enough to pass the substantial-similarity test. 
On this view, a model is a substantially similar copy of a work when the model is capable of generating the work.\footnote{This is a sticky technical problem. 
	Research has shown that memorization is not easily identifiable, and thus the amount of memorization in a model is not always or easily quantifiable. 
	In particular, the choice of memorization identification technique and available information (e.g., knowledge of the training dataset, context window, etc.) affect the amount of memorization that can be identified. 
	See, e.g., Carlini et al.\cite{carlini2023quantifying}.
}
Note that this is direct infringement, not secondary (Section~\ref{direct}). 
The theory is not that the generation is an infringing copy, and that the model is a tool in causing that infringement in the way that a tape-duplicating machine might be a tool in making infringing cassettes~\citep{abdallah}. 
Rather, the theory is that the model itself is an infringing copy, regardless of whether that particular generation is ever made.\footnote{Alert readers will note the similarity to the debate over whether the mere act of making a work available without a download infringes the distribution right.
	See \emph{London-Sire Records}~\citep{londonsire}; see generally Menell~\cite{menell2011search}.
}

\subsubsection{Generations}

Some generations are nearly identical to a work in the model's training data (i.e., memorized). They are substantialy similar to that work. 
Other generations are very dissimilar from every work in the training data. There is no substantial similarity, because infringement is assessed on a work-by-work basis. 
Although it is in some sense based on all of the works in the training dataset, it does not infringe on any of them.\footnote{While it may be straightforward to pose the question: ``is the given generation substantially similar to work 1,'' it is not at all straightforward to answer. 
	Training datasets are massive. 
	Manually comparing the generation to every single work in the dataset is infeasible; it would simply take too long.
	While automated methods could help identify works in the training set that are \emph{likely to be} similar to the generation, there is no automated metric that can definitively say if two works are substantially similar.
	See generally Scheffler et al.~\cite{scheffler2022formalizing} (which proposes one possibility for a metric for identifying substantial similarity)).
	Even with automated methods, checking \emph{every} generation that a system produces against every other work in the training dataset to evaluate similarity is extremely computationally expensive.
}
The hardest case is when an output is similar to a work in the training data in some ways, but dissimilar from it in other ways. This case is likely to arise in practice precisely because it lies in between the two extremes of memorized generations and original generations. Somewhere between them lies the murky frontier between infringing and non-infringing.\looseness=-1

It is hard to make sweeping statements because of the factual intensity and aesthetic subjectivity of similarity judgments.\footnote{To quote Learned Hand on the idea-expression dichotomy, 
``Nobody has ever been able to fix that boundary, and nobody ever can''~\citep[p. 121]{nichols}.} 
Whe\-ther a particular generation is substantially similar or not 
is ultimately a jury question requiring assessment of audiences' subjective responses to the works.
Generative AI will produce cases requiring this lay assessment; it is impossible to anticipate in advance how lay juries will react to all of the possible variations.
So, we will assume that lay audiences would say that some generations will infringe, but that it will not be possible to perfectly predict which ones.\footnote{Notably, providing guarantees that any given generated work might not potentially infringe copyright is impossible if the training data contains copyrighted data. 
	This is simply because provable guarantees require formal definitions, and there are no widely accepted formal definitions of substantial similarity. But see Scheffler et al.~\cite{scheffler2022formalizing} (providing a possible starting point).
	Instead, current ML techniques focus on reducing the likelihood that generations from a model will closely resemble any of the model's training data.
}\looseness=-1

Even if complete answers are impossible, there are some interesting questions worth considering. 
As Matthew Sag observes~\citep{sag2023safety},  certain characters are so common in training datasets that models have ``a latent concept [of them] that is readily identifiable and easily extracted.'' 
For example, prompting Midjourney and Stable Diffusion with  \prompt{snoopy} produces recognizable images of Snoopy the cartoon beagle. 
Characters are a special case in copyright;
some cases relax the rule that infringement is measured on a work-by-work basis, instead measuring the similarity of the defendant's character to one who appears in multiple works owned by the plaintiff.\footnote{E.g., \emph{DC Comics v. Towle}~\cite{towle}; see generally Sag~\cite{sag2023safety} (discussing caselaw and scholarship)} 
But the ``Snoopy effect'' is not confined to characters. Some works are simply so prevalent in training datasets that models memorize them. As an uncopyrighted example, Van Gogh's \emph{Starry Night} is easy to replicate using Midjourney; Sag's paper includes a replication of Banksy's \emph{Girl with Balloon}. This looks like substantial similarity.\looseness=-1 

A variation of the Snoopy effect arises when a model learns an artist's recognizable \emph{style}. ChatGPT can be prompt\-ed to write rhyming technical directions in the style of Dr. Seuss; 
\dalle-2{} can be prompted to generate photorealistic portraits of nonexistent people in the style of Dor\-o\-the\-a Lange~\citep{casper2023measuring}. 
As with characters, these outputs have similarities that span a body of source works, even if they are not close to any one source work.
The proper doctrinal treatment of style is a difficult question~\citep{sobel2023elements}. 
The Snoopy effect can also be triggered even without explicit prompting. 
The archaeologist example generated in Figure~\ref{fig:archaeologist-hires} features a dark-haired male character with stubble, wearing a brown jacket and white shirt, with a pouch slung across his shoulder. 
These are features associated with Indiana Jones, but neither the features nor \texttt{"indiana jones"} appear in the prompt. 
Some caselaw holds that these types of similarities are enough for infringement  when the character is iconic enough~\citep{mgmhonda}.\footnote{See \emph{Metro-Goldwyn-Mayer v. American Honda Motor Co.}~\cite{mgmhonda} (car commercial featuring ``a handsome hero who, along with a beautiful woman, lead a grotesque villain on a high-speed chase, the male appears calm and unruffled, there are hints of romance between the male and female, and the protagonists escape with the aid of intelligence and gadgetry'' infringes on James Bond character).
}\looseness=-1

Other copyright doctrines, however, may limit infringement in Snoopy-effect cases. 
One of them is \emph{scènes à faire}: creative elements that are common in a genre cannot serve as the basis of infringement.  
For example, \citep[p. 50]{walker} explains that ``drunks, prostitutes, vermin and derelict cars would appear in any realistic work about the work of policemen in the South Bronx.'' 
Similarly, prompting Midjourney with \texttt{"ice princess"} produces portraits in shades of blue and white with flowing hair and ice crystals. (Figure~\ref{fig:ice-princess}) 
Similarities to Elsa from \emph{Frozen} arise simply because these are standard tropes of wintry glamour.
Some of them may now be tropes \emph{because} of the \emph{Frozen} movies, but they are still uncopyrightable ideas, rather than protectable expression.\footnote{See \emph{Nichols}~\cite[p. 121]{nichols} (``Though the plaintiff discovered the vein, she could not keep it to herself; so defined, the theme was too generalized an abstraction from what she wrote. It was only a part of her `ideas.' '')
}

To close this section, we note that not all similarity is infringing. 
Some similarities arise for innocent reasons.
The defendant and the plaintiff might both have copied from a common predecessor work, and resemble each other because they both resemble the work they were based on.
The similarities might consist entirely of accurate depictions of the same preexisting thing, like Grand Central Station at midday, and resemble each other because Grand Central Station resembles itself.
The similarities might be purely coincidental.
The plaintiff might even have copied from the defendant!

Copyright law therefore requires that the plaintiff prove that the defendant copied from their work, rather than basing it on some other source or creating it anew, an inquiry known as ``copying in fact.'' 
This is a factual question. 
In some cases, there is direct evidence: e.g., the defendant admits copying or there is video of the defendant using tracing paper to copy a drawing.
But in many cases, there are two kinds of indirect evidence: proof that the defendant had \emph{access} to the plaintiff's work, and examples of ``probative'' \emph{similarities} in the works themselves.
Access shows that copying was possible, and similarities can rebut alternative innocent theories.\footnote{See generally \emph{Skidmore}~\cite{skidmore} (discussing proof of copying in fact); Latman~\cite{latman1990probative} (distinguishing ``probative'' similarities that prove copying in fact from substantive similarities that constitute improper appropriation).
} 

%% file: section/40-genai/43-talkinshort/4334-talkinshort-direct.tex
\subsection{Direct Infringement}
\label{direct}

We next discuss direct infringement and generations. We defer other supply-chain stages and analysis of indirect infringement to other work. Direct copyright liability has no mental element: it is ``strict liability.'' 
All that is required is that they intentionally made the infringing copy. 
George Harrison's  1970 ``My Sweet Lord'' has the same melody and  harmonic structure as the Chiffon's 1962 ``He's so Fine''; 
the court held that ``his subconscious knew it already had worked in a song his conscious mind did not remember,'' and found him liable for infringement~\citep[p. 180]{harrisongs}.\looseness=-1

But direct copyright does have an element of ``volitional conduct''~\citep{costar}.
Its purpose is to decide whether a defendant should be analyzed as a direct or indirect infringer.\footnote{See \emph{Aereo}~\cite[2512-13]{aereo} at 2512-13 (Scalia, J., dissenting)
}
Some courts have described the test in terms of causation: ``who made this copy?''\footnote{See \emph{Cartoon Network}~\cite[p. 130]{cartoonnetwork}; see also \emph{Perfect 10, Inc. v. Giganews, Inc.}~\cite{giganews}.} 
The direct infringer is the party whose actions toward a specific item of content most proximately caused the infringing activity; anyone else is (potentially) an indirect infringer.
Thus, for example, a service that can be used to upload and download infringing content that a user chooses does not engage in volitional conduct~\citep{giganews}, but a service that curates a hand-picked selection of infringing content for users to download does~\citep{mp3tunes}. 

The simplest case is where the same actor supplies both the model and the prompt.\footnote{Such as a text-to-image model developer using the model to create example prompt/generation pairs to display on their website.}
Here, the sub\-con\-scious-copying doctrine is a surprisingly good fit for AI generation.
The model's internals are like the contents of George Harrison's brain: creatively effective, but not fully amenable to inspection.
If I prompt an image model with \prompt{ice princess}, I have set in motion a process that may draw on copyrighted works in the same way that George Harrison drew on other works he had heard.
If that process generates Elsa, the resulting infringement is on me the same way that the infringement of ``He's So Fine'' was on Harrison.
I could have taken greater care to check whether the image I was generating resembled a copyrighted work -- just as George Harrison could have thought harder or asked more people whether the tune sounded familiar.\looseness=-1

Matters are more complicated for generation services. 
Here, the question is whether the user and/or the provider should be treated as a direct infringer. There are at least three plausible answers, depending on the facts. First, the \emph{user of the service} might be a direct infringer. 
If a user enters a prompt for \prompt{elsa and anna from frozen}, the provider resembles a copy shop that provides 
a general-purpose tool and let users choose what to do with it~\citep{giganews}. 
Second, the \emph{service provider} might be a direct infringer. If a user types in \prompt{heroic princesses} and the model generates a picture of Elsa and Anna, the user has acted innocently and it is the model that has narrowed down the space of possible outputs to one that happens to be infringing. 
Third, \emph{both} the user of the service and service provider might be treated as direct infringers. 
Suppose the user inputs \prompt{frozen 3 screenplay} to a service that has been trained on thousands of Hollywood screenplays. 
Both the user and the service have the necessary volition to create a work that is substantially similar to the \emph{Frozen} movies.\looseness=-1

It seems unlikely, however, that a court would treat both service and user as indirect infringers. 
This would violate the doctrinal requirement that there be a direct infringer for indirect liability to attach, and it would leave both potentially responsible parties free of liability. 
The choice between the other three cases is partly factual, and partly policy-driven. 
It is factual because there are clear paradigm cases in which the user of the service makes the choice for infringement, the service provider makes the choice for infringement, and the two conspire together to infringe.
But it is policy-driven because, between these three poles, the identification of the direct infringer depends on which analogies one finds persuasive, and what one thinks copyright's goals are.\footnote{It is worth briefly noting that plugins could additionally pull in content from external sources, such as a news website, that gets included in a generation. 
	Recall that this data is \emph{not} included in training the model; instead, it is fed into the model at generation time to try to improve the quality of generations with more up-to-date information~\citep{chatgpt-plugins} 
	Hypothetically, this content could get included verbatim in generations, leading to infringement issues in generation separate from those discussed above. 
}

%% file: section/40-genai/43-talkinshort/4335-talkinshort-fair.tex
\subsection{Fair Use}
\label{fairuse}

Many stages of the generative-AI supply chain involve \emph{prima facie} infringing reproductions, so copyright's all-purpose defense, fair use, will play a major role in making generative AI possible at all~\citep{17usc107}
Others have discussed the fair-use issues in detail~\citep{henderson2023foundation, sag2023safety, murray2023generative, sobel2017crisis}. 
It is highly case-specific, so we will focus on only a few salient points. 
We discuss generations, taking each of the four fair-use factors in turn, and defer other stages to other work.\\

\noindent\underline{\emph{Factor One}} (``the purpose and character of the use \ldots''~\citep{17usc107}\footnote{17 U.S.C. § 107(1).}): 
A use is transformative when ``the quoted matter is used as raw material, transformed in the creation of new information, new aesthetics, new insights and understandings''~\citep[p. 1111]{leval1990toward}.
The modification, remixing, and abstraction of input works literally involves exactly this kind of transformation.
Some AI skeptics might deny that AI-generated material can be expressive. 
But as long as audiences find ``new information, new aesthetics, new insights and understandings'' in these generations, the goals of transformative use will be served.\footnote{See \emph{Cariou}~\cite[p. 707]{cariou} (focusing audience perceptions of works rather than author's intentions in assessing transformative use); see generally Heymann~\cite{heymann2008transformative} (assessing transformative use from audience perspective); Liu~\cite{liu2003consumer} (discussing audience interests in copyright).
} 
Other generations will not be transformative.
When a model outputs a memorized work, here is no transformation in content (Section~\ref{similarity}).
Other changes can also be non-transformative, e.g., memorized examples that are noisier than the source image. 
The noise is not new expression conveying new aesthetics.
It is just noise. 
The rest of the first factor does not point one direction or the other.
Generations can be put to commercial use (e.g., backgrounds for a music video) and to noncommercial use (e.g., illustrating an academic article on generative AI).
Some outputs will be put to favored purposes like education and news reporting, while other outputs will be put to run-of-the-mill entertainment purposes.\footnote{See 17 U.S.C. § 107~\citep{17usc107} (favoring ``purposes such as criticism, comment, news reporting, teaching (including multiple copies for classroom use), scholarship, or research'')}\\

\noindent\underline{\emph{Factor Two}} (``the nature of the copyrighted work''~\citep{17usc107}\footnote{17 U.S.C. § 107(2)}): This factor depends on the model in question. 
Some training data will be informational; some will be expressive. 
Most training data will have been ``published'' within the meaning of copyright law; otherwise, it would not be available as training data at all.
A very small fraction of training data may be ``unpublished'' within the meaning of copyright law --- i.e., it has been shared ``(1) \ldots only to a select group (2) for a limited purpose and (3) with no right of further distribution by the recipients''~\citep[S. 6.31]{patrycopy} --- and included through express breach of confidence.
Here, this factor will favor the plaintiff.\\

\noindent\underline{\emph{Factor Three}} (``the amount and substantiality of the portion used \ldots''~\citep{17usc107}\footnote{17 U.S.C. § 107(3)}): 
This factor, like substantial similarity, will not systematically favor either side.
Some generations will closely resemble the works they were copied from; others will copy only small portions of the works.\footnote{See \emph{Associated Press v. Meltwater U.S. Holdings, Inc.}~\cite{meltwater} (rejecting fair use defense brought by news-monitoring service that reproduced substantial excerpts from articles for its customers)}
Even for works that are transformative, it still matters  whether the generation copies more than necessary.
A ``painting of a car driving in a snowstorm in the style of Frida Kahlo'' might copy just Kahlo's brushwork or floral motifs, or it might also imitate the entire composition of one of her self-portraits.\\

\noindent\underline{\emph{Factor Four}} (``the effect of the use upon the potential market for \ldots the copyrighted work.''\footnote{17 U.S.C. § 107(4)}): 
The outputs of a non-generative AI do not compete in the market for a copyrighted work.
These outputs could \emph{reduce the demand} for the copyrighted work.
For example, an AI-powered recommendation system might analyze the frames of a movie and assign it a low rating for visual interest.
But the rating does not substitute for the movie in the market for movies. 
Viewers consume the rating to learn about movies, not to enjoy the expression in the rating.
Any harm to  the copyright owner is not fourth-factor harm~\citep{campbell}. The outputs of a generative-AI system, however, can substitute for a copyrighted work under the fourth factor.
Consider the following variations on a theme:
\begin{itemize}[left=.3cm,topsep=1pt]
	\item Instead of paying to obtain a copy of ``The Old Sugarman Place'' episode of \emph{Bojack Horseman}, a user prompts a generative-AI system to generate \prompt{'The Old Sugarman Place'}. It generates a close duplicate --- 
    essentially a pirated edition at a lower price.
	This is a paradigmatic fourth-factor harm.\looseness=-1
	
	\item  The user prompts a generative-AI system to generate  \prompt{'The Old Sugarman Place'}, and the system generates a non-exact copy with significant changes to the dialogue and animation.
	This episode, ``The New Sugarman Place,'' is also a direct competitor for this user's business.
	It might be a better or worse competitor, depending on how closely ``The New Sugarman Place'' matches ``The Old Sugarman Place.'' 
	But this is still factor-four harm.
	
	\item The user prompts a generative-AI system to generate a new episode of \emph{Bojack Horseman}.
	The generation does not necessarily compete with ``The Old Sugarman Place'' itself.\footnote{Perhaps the user has already watched all of the existing episodes.}
	Instead, it competes with commissioning the writers, animators, and voice cast to create new episodes, or with paying for a license to make new episodes.\footnote{For another example, imagine that the user of a service prompts a text-to-image system to create a portrait of them in the style of a particular living artist; the generation is a substitute for commissioning the artist to paint one.}
	This is also factor-four harm to the market for licenses and authorized derivatives.
	For example, in \emph{Krofft}~\cite{krofft} McDonald's created advertisements in the unsettling style of the children's show \emph{H.R. Pufnstuff}.
	
	\item An individual prompts a generative-AI system to produce a generation in a broad style, e.g., \prompt{animated sitcom about depression}.
	The output is a video with dialogue and animation that do not look much like \emph{Bojack}.
	The output does not directly compete with ``The Old Sugarman Place,'' or with any particular work or particular author.
	Instead, it competes with animated television in general.
	If the generative-AI system had not been available, the individual might have paid to watch \emph{Bojack} or \emph{Dr. Katz} or some other show.
	Many authors might view this as undercutting the market for their work.
	Here, the fourth factor is \emph{not even relevant}, because the new video is not substantially similar to any existing work.
	If a human creative team made a new animated sitcom about depression, they would be celebrated for their creativity  not sued for infringement.
	
	\item  An individual prompts a generative-AI system to produce a generation in a broad style, e.g. \prompt{animated sitcom about depression}.
	The output, however, is ``The Old Sugarman Place.''
	The difference between this and the first case is that the user does not know about the work that the generation substitutes for. 
	This is a factor-four harm.
	The generative-AI system has diverted the individual from potentially learning about and paying to watch ``The Old Sugarman Place.''\looseness=-1
\end{itemize}

\vspace{.2cm}
\noindent To summarize, factors one, three, and four can point strongly in favor of fair use or strongly against, depending on the context, and factor two does not consistently point in either direction. We conclude that some generations will be fair uses and others will not.

%% file: section/40-genai/43-talkinshort/4340-talkinshort-outcomes.tex
\section{Which Way from Here?}
\label{whichway}

The generative-AI supply chain is extremely complex.
So is copyright law.
Putting the two of them together multiplies the intricacy.
Two unsettling conclusions follow. 
First, because of the complexity of the \emph{supply chain}, it is not possible to make accurate sweeping statements about the copyright legality of generative AI.
Too much depends on the details of specific systems.
All the pieces matter, from the curatorial choices in the training dataset, to the training algorithm, to the deployment environment, to the prompt supplied by the user.
Courts will have to work through these details in numerous lawsuits and develop doctrines to distinguish among different systems and uses.
Second, because of the complexity of \emph{copyright law}, there is enormous play in the joints.
Substantial similarity, fair use, and other doctrinal areas 
all have open-ended tests that can reach different results depending on the facts a court emphasizes and the conclusions it draws.
This complexity gives courts the flexibility to deal with  variations in the supply chain.
Paradoxically, it also gives courts the freedom to reach any of several different plausible conclusions about a generative-AI system. 
We explore some of the ways that courts might use their discretion to apply copyright law to generative AI (Section~\ref{outcomes}), and then discuss some of the considerations that courts should keep in mind  (Section~\ref{lessons}).

\subsection{Possible Outcomes}
\label{outcomes}

There are a few boxes that courts may find it appealing to sort generative-AI systems into.

\subsubsection{No Liability}

First, courts might hold that neither services nor users are liable for copyright infringement.
Under a combination of no substantial similarity and fair use, anything produced by a generative-AI system would be categorically legal.
Models and services would also be legal because intermediate nonexpressive fair use would shield them.
Training datasets would also usually be legal as well (except perhaps in cases of blatant infringement like Books3)~\citep{books3wired, books3atlantic,kadrey}. 
They would be fair-use inputs to noninfringing downstream stages of the supply chain.\looseness=-1

This regime is clear and simple.
It would also be unstable.
While this outcome might make sense for some generative-AI systems, it seems both unworkable for systems trained specifically to emulate the styles of particular creators, and retrieval systems that reproduce matching works exactly~\citep{borgeaud2022retro}. 
If all generative AI were categorically legal, then developers might start adding generative components to other systems in order to launder copyrighted works through them.
The endpoint could be the effective collapse of copyright.
Assuming that this is not an outcome that courts would willingly preside over, then, a blanket no-liability regime seems unlikely.
Instead, courts would be more likely to find at least some infringement --- so the question becomes where to draw the line.

\subsubsection{Liability for Generations Only}

Second, courts could draw a line between services and users.
In this regime, only generations would be treated as infringing.\footnote{Here, we use the term ``user'' broadly. 
	A user could be a customer using a web application to produce a generation, a developer using an API to produce a generation in their own code, a developer using an API to produce a generation for a company, etc. 
}
In this world, generative-AI systems would be creative tools like Photoshop.\footnote{Sometimes literally so. See Adobe~\cite{photoshopgenerativefill}.
}
The user would be responsible for making sure that anything they create with the tools is noninfringing, but the tools would be shielded under something like a strong \emph{Sony} rule, assembled out of a combination of no substantial similarity, no indirect infringement, and/or fair use.
This result might be unfair to users whose infringements resulted from systems producing generations that reproduce material in the underlying model's training dataset, through no choice or fault of their own. 
But this is arguably the same kind of situation that some courts currently countenance when they hold that users can be liable for embedding images from Instagram even though Instagram is not liable for hosting those images~\citep{sinclair}. 

The main difficulty with this regime would be policing against systems designed specifically for infringement.
Something like the \emph{Grokster}~\citep{grokster} rule, carefully followed, might suffice.
The providers of a service that was geared to produce infringing outputs could be held liable.
So could the publishers or deployers of a model that had been trained or fine-tuned to optimize its effectiveness at infringement.
So could the curator of a dataset that included only infringing works, or was intentionally organized to meet the needs of a model known to be intentionally trained for infringement. 
At every stage, a party would be held responsible only for its own actions directed towards increasing the use of a system for infringement. 

\subsubsection{Notice and Removal}

Courts could treat generative-AI services as generally legal, but require them to respond to knowledge of specific infringements under a \emph{Napster}-like rule~\citep{napster}.
One plausible route to this regime would be to treat infringing generations as creating direct liability for users and only indirect liability for service providers.
Another would use fair use to shield service providers as long as they took reasonable overall precautions, including responding when they had sufficient knowledge of infringement.
And a third would be to find liability but craft an injunction that only required services to act against infringement they were aware of.\footnote{Regardless of which of these doctrinal routes a court took, there would be an inevitable gravitational force pulling the provider's duties towards the duties of a service provider under Section 512(c) or (d).
This is not because Section 512 applies to generative-AI services.
It largely does not --- analysis that we defer to other work. 
Instead, the Section 512 doctrines may be a convergence point because courts have now had two decades of experience --- which means two decades of precedents --- with the Section 512 safe harbors.
These precedents have come to set expectations --- among copyright owners, in the technology industry, in the copyright bar, and in the judiciary --- for what legally ``responsible'' behavior by an online intermediary looks like.
A generative-AI service operator that does not appear to be making a good-faith effort to achieve something like this system may strike a court as intending to induce infringement, not making a good-faith effort to comply with an injunction, etc.}\looseness=-1

If courts end up recreating a notice-and-takedown regime, they would likely settle on familiar elements from the DMCA notice-and-takedown provision of Section 512: a way for copyright owners to give notice of infringement, block infringing generations on notice, block infringing generations on actual knowledge, block infringing generations on red-flag knowledge, avoid having a business model that directly ties income to infringement, and terminate the abilities of repeat infringers to continue making generations.

This is a very difficult technical problem.
It would be much harder for a generative-AI system to implement than it is for a hosting platform to implement Section 512 compliance.
The reason is that a notice directed to a hosting provider under Section 512(c) must include ``Identification of the material that is claimed to be infringing \ldots and information reasonably sufficient to permit the service provider to locate the material''~\citep{17usc512}.\footnote{U.S.C. § 512(c)(3)(A)(iii).}
A valid notice is a roadmap; it tells the hosting provider exactly what to take down to comply.
That material already exists, and the hosting provider can compare it to the copyrighted work to verify that they are substantially similar.
But a notice to a generative-AI system is a notice against future generations, which may be different from each other and resemble the copyrighted work in different ways.
Filtering for this kind of much more inexact match is much harder technically.\footnote{That said, matching material against a catalog of copyrighted works is a problem that has been very approximately solved by major social networks, which use perceptual hashing to prevent the upload of various kinds of identified content.
Generative-AI companies could at least add similar perceptual-hash-driven filtering to the outputs of their models, but clearly this would only solve part of the problem~\citep{lee2022dedup, ippolito2023preventing}.
The challenges of implementing removal for models are even harder.
A service can add filters on the input and output sides --- monitoring prompts and scanning outputs.
It can also fine-tune or align the model, or provide it with an overall prompt that instructs the model to respond in ways that reduce its propensity to infringe. Further, a model by itself does not implement these controls.
The model cannot control how it is prompted or what the user does with the output.
The model cannot stop anyone from fine-tuning it to remove its guardrails.} 
Further, there is no simple analogue for takedown in generative-AI models. 
Removing the influence of a particular example on a model is an active and unsolved area of research~\citep{meng2022editing, bourtoule2021unlearning}.\footnote{Absent the ability to do so, the safest bet is to retrain the model from scratch. 
Due to the time and expense required to retrain a model, it will often be infeasible to retrain it simply to remove infringing works, and completely unworkable to retrain on each new notice. We defer further discussion of how courts could deal with this difficulty to other work.} 

\subsubsection{Infringing Models}

A fourth possibility is that some or all generative-AI services are illegal because models themselves infringe.
This outcome is an existential threat to model trainers and service providers; it makes their operations \emph{per se} copyright infringement.
It is also the outcome being sought by the class-action plaintiffs in high-profile lawsuits against OpenAI, Stability AI, and some of their partners. 
In this regime, the most important component of copyright law would become licensing.
Models could only be trained on data that had been licensed from the 
copyright owners; the terms under which those models and their generations could be used would have to be negotiated as part of the licensing agreement.\footnote{Each model would have a fully licensed training dataset, and the question of infringement would not arise except in cases where there were infringing works in the dataset itself or some other failure of quality control somewhere along the supply chain.}\looseness=-1

\subsection{Lessons}
\label{lessons}

Having discussed what courts and policymakers could do, we now consider what they should do.
In keeping with our bottom line --- \emph{the generative-AI supply chain is too complicated to make sweeping rules prematurely} --- we offer a few general observations about the overall shape of copyright and generative AI that courts and policymakers should keep in mind as they proceed.

First, \emph{copyright touches every part of the generative-AI supply chain}.
Every stage from training data to alignment can make use of copyrighted works.
Generative AI raises many other legal issues: 
Can a generative-AI system commit defamation~\citep{volokh2023libelmodels, garon2023picture, brown2023badly, bambauer2023authorbots, henderson2023wheres}?
Can a generative-AI system do legal work~\citep{choi2023lawschool} and should they be allowed to~\citep{avianca}? 
But these issues pertain to outputs of a generative-AI system--copyright pervades every step of the process.

Second, \emph{copyright concerns cannot be localized} to a single link in the supply chain.
Decisions made by one actor can affect the copyright liability of another actor far away in the supply chain.
Whether an output looks like Snoopy or like a generic beagle depends on what images were collected in a dataset, which model architecture and training algorithms are used, how trained models are fine-tuned and aligned, how models are embedded in deployed services, what the user prompts with, etc. 
Every single one of these steps could be under the control of a different person.

Third, \emph{design choices matter}.
There are obvious choices about copyright, like whether to train on unlicensed data (which can affect downstream risks), and how to respond to notices that a system is producing infringing outputs (which can affect upstream risks).
But subtler architectural choices matter, too.
Different settings on a training algorithm can affect how much the resulting model will memorize specific works.
Different deployment environments can affect whether users have enough control over a prompt to steer a system towards infringing outputs.
Copyright law will have to engage with these choices --- as will AI policy.

Fourth, \emph{fair use is not a silver bullet}.
For a time, it seemed that training and using AI models would often constitute fair use.
In such a world, AI development is generally a low-risk activity, at least from a copyright perspective.
Yes, training datasets and models and systems may all include large quantities of copyrighted works --- but they will never be shown to users.
Generative AI scrambles this assumption.
The serious possibility that some generations will infringe means that the fair-use analysis at every previous stage of the supply chain is up for grabs again.

Fifth, \emph{the ordinary business of copyright law still matters}.
Courts will need to make  old-fashioned, retail judgments about individual works --- e.g., how much does this image resemble Elsa in particular, rather than generic tropes of fantasy princesses?
Courts \emph{must} leave themselves room to continue making these retail judgments on a case-by-case basis, responding to the specific facts before them, just as they always have.
Perhaps eventually as society comes to understand what uses generative AI can be put to and with what consequences, it will reconsider the very fundamentals of copyright law.
But until that day, we must live with the copyright system we have.
And that system cannot function unless courts are able to say that some generative-AI systems and generations infringe, and others do not.\looseness=-1

Sixth, \emph{analogies can be misleading}.
There are plenty of analogies for generative AI ready to hand.
A generative-AI model or system is like a search engine, or like a website, or like a library, or like an author, or like any number of other people and things that copyright has a well-developed framework for dealing with. 
These analogies are useful, but we wish to warn against treating any of them as definitive. 
As we have seen, generative AI is and can consist of many things. 
It is also literally a generative technology: it can be put to an amazingly wide variety of uses~\citep{zittrainfuture}. 
And one of the things about generative technologies is that they cause convergence~\citep{narechania2022convergence}. 
precisely because they can emulate many other technologies, they blur the boundaries between things that were formerly distinct.
Generative AI can be like a search engine, and also like a website, a library, an author, and so on.
Prematurely accepting one of these analogies to the exclusion of the others would mean ignoring numerous relevant similarities --- precisely the opposite of what good analogical reasoning is supposed to do.

%% file: section/40-genai/43-talkinshort/4350-talkinshort-conclusion.tex
\section{Conclusion}

Our conclusion is simple.
``Does generative AI infringe copyright?'' is not a question that has a yes-or-no answer.
There is currently no blanket rule that determines which  participants in the generative-AI supply chain are copyright infringers.
The underlying systems are too diverse to be treated identically, and copyright law has too many open decision points to provide clear answers. 
Copyright is not the only, or the best, or the most important way of confronting the policy challenges that generative AI poses.
But copyright is here, and it is asking good questions about how generative-AI systems are created, how they work, how they are used, and how they are updated. 
These questions deserve good answers, or failing that, the best answers our copyright system is equipped to give.\looseness=-1

%% file: section/50-conclusion/500-conclusion.tex
\chapter{Conclusion}\label{part:conclusion}

This dissertation puts forth a vision for a new research field that studies problems at the intersection of machine learning, law, and policy. 
At its center are questions that concern the challenges for taking reliability seriously across machine learning: 
the importance of careful, meaningful metric design, methodologies for making sure that we can measure these metrics efficiently and dependably at scale, and successful communication with legal scholars and policymakers about what measurements can (and cannot) tell us about the capabilities and risks of ML systems. 
In service of this vision, this dissertation covers research contributions in three different, yet cross-cutting, themes: sources of arbitrariness in machine learning (Part~\ref{part:arbitrary}), taming randomness in scalable, reliable ML algorithms (Part~\ref{part:algorithms}), and evaluating generative-AI systems (Part~\ref{part:genai}). 


In Part~\ref{part:arbitrary}, we describe different ways that non-determinism can bring about arbitrary outcomes in ML experiments. 
We quantify and mitigate two particular sources of machine-learning-related arbitrariness: (1) arbitrariness that can result in conclusions from non-deterministic, human-made decisions in experiments that involve hyperparameter optimization (Chapter~\ref{chapter:hpo}), and (2) arbitrariness that can result in social prediction contexts when we do not account for variance in possible learned decision rules, due to randomness in the training process (Chapter~\ref{chapter:fairness}). 

Arbitrariness is especially relevant in the law. 
One of the goals of legal rules is to remove arbitrariness in decision-making, so that the law is predictable, consistent, and fair --- qualities that are essential for due process~\cite{tamanaha2004law, fuller1965law}. 
We make important, novel connections between non-determinism-induced arbitrariness in ML and arbitrariness in the law. 
Our work in this area shows why ML-related arbitrariness is meaningfully different from other types of arbitrariness that the law considers, which has important consequences for how law- and policymakers should reason about and regulate the deployment of ML systems in public contexts. 


In Part~\ref{part:algorithms}, we detail algorithmic contributions in scalable uncertainty estimation and distributed optimization. 
This research contends with improving scalability without sacrificing reliability. 
For the former, reliability involves guaranteeing that our Bayesian inference algorithms converge to the true posterior distribution, so that we can get accurate estimates of model  uncertainty (Chapter~\ref{chapter:tunamh}). 
For the latter, we prove that there are better-than-random example orders for SGD-based distributed optimization; our algorithm converges provably faster than random example ordering, and also exhibits better generalization in practice (Chapter~\ref{chapter:cdgrab}). 

Both of these chapters harness randomness in sampling and optimization, respectively, to construct more scalable/efficient, reliable algorithms. 
In computing, scalabilty/efficiency and reliability are often in trade-off; it is challenging to achieve greater scalability or efficiency without sacrificing reliability. 
In both of these chapters, we leverage low-level technical details about sampling and optimization, respectively, so that we can relax these trade-offs. 

We then discuss how analogous trade-offs are very common in law and policy contexts. 
Policymakers trade-off between reliability and scalability/efficiency all the time, in domains as complex as federal risk assessment and health policy. 
Understanding these trade-offs is often sufficient for making sense of underlying implementation decisions, which impact overall algorithm and system behavior. 
As a result, we make the case that such trade-offs are a useful level of abstraction for communicating with policymakers and other non-expert stakeholders about the capabilities and risks of ML systems (Chapter~\ref{chapter:tradeoffs}).


Part~\ref{part:genai} also involves algorithms in scalable machine learning, but makes its central contributions in evaluating generative-AI system. 
We put forth methodology for feasible measurement of memorization at contemporary-LLM scale.
Since we do not know the training datasets of most state-of-the-art LLMs, we develop a proxy for a training dataset that we use to validate memorization. 
In doing so, we reveal interesting patterns in memorization across open and semi-closed models~\cite{nasr2023scalable}.
A variant of our attack, which gets the aligned ChatGPT to diverge from its chatbot-style outputs, shows that ChatGPT memorizes orders of magnitude more than was previously known, and much more than any other model that we tested.
This work represents the first successful, large-scale memorization attack on an aligned, deployed, closed  production system (Chapter~\ref{chapter:memorization}). 

One of the key concerns around regurgitating memorized training data is that these generations can contain verbatim copies of content that is possibly under copyright. 
Such copyright concerns suggest a natural alternative: 
training generative-AI models on public domain or explicitly licensed curated data. 
We explore this in depth for text-to-image latent diffusion models: 
we curate a training dataset of Creative Commons licensed images, for which we produce synthetic captions, and we use this image-caption dataset to train variants of the Stable Diffusion 2 architecture. 
This work involves a variety of ML-systems contributions, as well as a starting place for showing the benefits and limitations of training on licensed data (Chapter~\ref{chapter:commoncanvas}). 

Both of these chapters provide a great intuition for why generative AI presents significant challenges for U.S. copyright law.
However, since their main contributions are in machine learning, they do not go into detail about their relationship with copyright.
We conclude this part by closing this loop. 
We present an abridged version of our landmark article on copyright and the generative-AI supply chain, which defines a rich framework for reasoning precisely about the many ways that copyright law presents significant issues for generative AI, and how generative AI, in turn, presents novel, significant issues for copyright law (Chapter~\ref{chapter:talkinshort}).  

By making contributions in these three interrelated themes, this dissertation demonstrates that research on reliable measurement for ML is intrinsically tied with research in law and policy. 
While these are different disciplines, they are actually two complementary sides of the same research vision. 
They serve the same goals of trying to rigorously define what it means to be ``reliable'' and how to implement reliability in practice. 
This requires rigorous knowledge about the capabilities and risk of ML models (and the systems in which they are embedded). 
Developing such knowledge depends on fundamental algorithmic and methodological contributions in machine learning. 
It also requires meaningful engagement with concrete law and policy questions about what we want ML systems to do in the world. 

There is a virtuous cycle between these two sides: 
research on reliable measurement for machine learning has direct implications for law and policy, and work in law and policy raises novel questions to tackle concerning how to define and conduct reliable measurement for machine learning. 
I have been very fortunate throughout my Ph.D. to be able to engage deeply with both sides of this research vision --- to ask and answer questions that requires developing insights in machine learning, law, and policy.  

This dissertation provides a tour of some of scholarship that has developed these insights. 
However, it does not discuss the work I have done to directly engage with policymakers in practice. 
Most of this work has occurred through an organization called The Center for Generative AI, Law, and Policy Research (The GenLaw Center), which I co-founded with my co-authors, Katherine Lee and James Grimmelmann. 
This organization originated through an ICML '23 workshop that we co-organized, and has grown into and incubated a new research community and field. 
We have had a significant impact on the trajectory of some Ph.D. students' doctoral work, and have an ongoing impact on U.S. AI policy.  

Most recently, we held a workshop in Washington, D.C., which we co-organized with Carnegie Mellon University, the Georgetown Law Center, and the Center for Democracy and Technology.  
This workshop centered on educating policymakers about generative-AI evaluations, and what we can and cannot measure about generative-AI systems --- in other words, the good, the bad, and the hype. 
The work that we are doing through The GenLaw Center --- both our policy outreach and the research we are conducting --- is exactly in line with some of the action items called for in President Biden's executive order~\citep{eo}, among other policy mandates. 

The great news is, we have already been doing this work together for a while. 
We have been making, and continue to make, fundamental contributions in research and practice at the intersection of machine learning, law, and policy. 
It is a bit surreal to see some of the work in this dissertation become nationally, or even internationally relevant. 
But this also means it is an especially exciting time to continue my research vision, beyond the introduction that has been presented in these chapters. 

%% file: section/99-appendix/900-appendix.tex
\part{Appendix}
\appendix

\input{section/99-appendix/999-list}
\input{section/99-appendix/21-hpo/00-app-hpo}
\input{section/99-appendix/22-fairness/00-app-fairness}
\input{section/99-appendix/31-tunamh/00-app-tunamh}

\vspace{-100in}
\input{section/99-appendix/32-cdgrab/00-app-cdgrab}
\input{section/99-appendix/42-commoncanvas/00-app-commoncanvas}

\input{section/10-intro/12-accountability/12-accountability-main}

%% file: section/99-appendix/999-list.tex
\chapter{Comprehensive List of Ph.D. Writing}
*Equal contribution

\begin{etaremune}
    \item \textbf{A. Feder Cooper}* and James Grimmelmann*. ``The Files are in the Computer: Copyright, Memorization, and Generative AI.'' Forthcoming, \emph{Chicago-Kent Law Review}. 2024~\citep{cooper2024files}.

    \item Nicholas Carlini, Daniel Paleka, Krishnamurthy Dj Dvijotham, Thomas Steinke, Jonathan Hayase, \textbf{A. Feder Cooper}, Katherine Lee, Matthew Jagielski, Milad Nasr, Arthur Conmy, Eric Wallace, David Rolnick, Florian Tramèr. ``Stealing Part of a Production Language Model.'' \emph{International Conference on Machine Learning 2024} (\emph{ICML '24}). 2024. \textbf{Best Paper Award}~\citep{carlini2024stealing}. 
    
    \item  Daniel McDuff, Tim Korjakow, Scott Cambo, Jesse Josua Benjamin, Jenny Lee, Yacine Jernite, Carlos Muñoz Ferrandis, Aaron Gokaslan, Alek Tarkowski, Joseph Lindley, \textbf{A. Feder Cooper}, and Danish Contractor. ``On the Standardization of Behavioral Use Clauses and Their Adoption for Responsible Licensing of AI.'' \emph{International Conference on Machine Learning 2024} (\emph{ICML '24})~\cite{mcduff2024license}.
    
    \item \textbf{A. Feder Cooper}*,  Katherine Lee*, and James Grimmelmann*. ``Talkin' 'Bout AI Generation: Copyright and the Generative-AI Supply Chain.'' Forthcoming, \emph{Journal of the Copyright Society}. 2024~\cite{lee2023talkin}.
    
    \item \textbf{A. Feder Cooper}*,  Katherine Lee*, and James Grimmelmann*. ``Talkin' 'Bout AI Generation: Copyright and the Generative-AI Supply Chain (The Short Version).'' \emph{The 3rd ACM Symposium on Computer Science and Law (CSLAW '24)}. 2024. \textbf{Long Presentation}~\cite{cooper2024talkinshort}.

    \item Aaron Gokaslan, \textbf{A. Feder Cooper}, Jasmine Collins, Landan Seguin, Austin Jacobson,
	Mihir Patel, Jonathan Frankle, Cory Stephenson, and Volodymyr Kuleshov. ``CommonCanvas: Open Diffusion Models Trained on Creative-Commons Images.'' \emph{Conference on Computer Vision and Pattern Recognition 2024} (\emph{CVPR '24}). 2024~\cite{gokaslan2023commoncanvas}. 

    \item \textbf{A. Feder Cooper}, Katherine Lee, Madiha Zahrah Choksi, Solon Barocas, Christopher De Sa, James Grimmelmann, Jon Kleinberg, Siddhartha Sen, and Baobao Zhang. ``Arbitrariness and Social Prediction: The Confounding Role of Variance in Fair Classification.'' \emph{38th AAAI Conference on Artificial Intelligence} (\emph{AAAI '24}). 2024. \textbf{Best Paper Honorable Mention}~\cite{cooper2024variance}. 

    \item Milad Nasr*, Nicholas Carlini*, Jonathan Hayase, Matthew Jagielski, \textbf{A. Feder Cooper}, Daphne Ippolito, Christopher A. Choquette-Choo, Eric Wallace, Florian Tramèr, and Katherine Lee. "Scalable Extraction of Training Data from (Production) Language Models." 2023. Under submission~\cite{nasr2023scalable}. 

    \item \textbf{A. Feder Cooper}*, Wentao Guo*, Khiem Pham*, Tiancheng Yuan, Charlie F. Ruan, Yucheng Lu, and Christopher De Sa. ``Coordinating Distributed Example Orders for Provably Accelerated Training.'' \emph{Conference on Neural Information Processing Systems 36 (NeurIPS '23)}. 2023~\citep{cooper2023cdgrab}.

    \item Kweku Kwegyir-Aggrey, \textbf{A. Feder Cooper}, Jessica Dai, John Dickerson, Keegan Hines, and Suresh Venkatasubramanian. ``Repairing Regressors for Fair Classification at Any Decision Threshold.'' \emph{Workshop on Algorithmic Fairness through the Lens of Time} at \emph{NeurIPS 2023}. 2023. \textbf{Oral}~\citep{aggrey2023repair}. 

    \item \textbf{A. Feder Cooper}*, Katherine Lee*, James Grimmelmann*, Daphne Ippolito*, Christopher Callison-Burch, Christopher A. Choquette-Choo, Niloofar Mireshghallah, Miles Brundage, David Mimno, Madiha Zahrah Choksi, Jack M. Balkin, Nicholas Carlini, Christopher De Sa, Jonathan Frankle, Deep Ganguli, Bryant Gipson, Andres Guadamuz, Swee Leng Harris, Abigail Z. Jacobs, Elizabeth Joh, Gautam Kamath, Mark Lemley, Cass Matthews, Christine McLeavey, Corynne McSherry, Milad Nasr, Paul Ohm, Adam Roberts, Tom Rubin, Pamela Samuelson, Ludwig Schubert, Kristen Vaccaro, Luis Villa, Felix Wu, and Elana Zeide. ``Report of the 1st Workshop on Generative AI and Law.'' 2023. Technical Report~\citep{cooper2023report}. 

    \item \textbf{A. Feder Cooper}*, Katherine Lee*, James Grimmelmann, and Daphne Ippolito. ``AI and Law: The Next Generation (An explainer series).'' 2023. Technical Report~\citep{lee2023explainers}. 

    \item \textbf{A. Feder Cooper}, Katherine Lee, Madiha Zahrah Choksi, Solon Barocas, Christopher De Sa, James Grimmelmann, Jon Kleinberg, Siddhartha Sen, and Baobao Zhang. ``Distribution Justice: Variance, Uncertainty, and Rules in Machine Learning and Law.'' \emph{Privacy Law Scholars Conference}. 2023.

    \item \textbf{A. Feder Cooper}, Jonathan Frankle, and Christopher De Sa. ``Non-Determinism and the Lawlessness of Machine Learning Code.'' \emph{The 2nd ACM Symposium on Computer Science and Law (CSLAW '22)}. 2022. \textbf{Long Presentation}~\cite{cooper2022lawless}. 

    \item \textbf{A. Feder Cooper} and Karen Levy. ``Fast or Accurate? Governing Conflicting Goals in Highly Autonomous Vehicles.'' \emph{Colorado Technology Law Journal}, Vol. 20. 2022~\cite{cooper2022fast}.

    \item  \textbf{A. Feder Cooper}, Solon Barocas, Karen Levy, and Gili Vidan. `` `We have met the enemy and it is us': Debating the ethics of computing in the pages of \textit{CACM}.'' \emph{2022 Workshop of the The Special Interest Group for Computing, Information, and Society (SIGCIS '22)}. 2022. 

    \item \textbf{A. Feder Cooper}*, Emanuel Moss*, Benjamin Laufer, and Helen Nissenbaum. ``Accountability in an Algorithmic Society: Relationality, Responsibility, and Robustness in Machine Learning.'' \emph{Proceedings of the 5th ACM Conference on Fairness, Accountability, and Transparency (FAccT '22)}. 2022~\cite{cooper2022accountability}. 

    \item \textbf{A. Feder Cooper} and Gili Vidan. ``Making the Unaccountable Internet: The Changing Meaning of Accounting in the Early ARPANET.'' \emph{Proceedings of the 5th ACM Conference on Fairness, Accountability, and Transparency (FAccT '22)}. 2022~\citep{cooper2022arpa}.

    \item Benjamin Laufer, \textbf{A. Feder Cooper}*, Sameer Jain*, Jon Kleinberg, and Hoda Heidari. ``Four Years of FAccT: A Reflexive, Mixed-Methods Analysis of Research Contributions, Shortcomings, and Future Prospects.'' \emph{Proceedings of the 5th ACM Conference on Fairness, Accountability, and Transparency (FAccT '22)}. 2022~\citep{laufer2023fouryears}. 

    \item \textbf{A. Feder Cooper}, Yucheng Lu, Jessica Zosa Forde, and Christopher De Sa. ``Hyperparameter Optimization Is Deceiving Us, and How to Stop It.'' \emph{Conference on Neural Information Processing Systems 34 (NeurIPS '21)}. 2021~\cite{cooper2021hpo}. 

    \item \textbf{A. Feder Cooper}, Karen Levy, and Christopher De Sa. ``Accuracy-Efficiency Trade-Offs and Accountability in Distributed ML Systems.'' \emph{Proceedings of the 2021 ACM Conference on Equity and Access in Algorithms, Mechanisms, and Optimization (EAAMO '21)}. 2021. \textbf{Oral}~\cite{cooper2021eaamo}.

    \item \textbf{A. Feder Cooper}, Maria Antoniak, Christopher De Sa, Marilyn Migiel, and David Mimno. ``~`\textit{Tecnologica cosa}': Modeling Storyteller Personalities in Boccaccio's \textit{Decameron}.'' \emph{SIGHUM Workshop on Computational Linguistics for Cultural Heritage, Social Sciences, Humanities and Literature} at \emph{The 2021 Conference on Empirical Methods in Natural Language Processing (EMNLP '21)}. 2021~\citep{cooper2021tecnologica}.

    \item \textbf{A. Feder Cooper} and Ellen Abrams. ``Emergent Unfairness in Algorithmic Fairness-Accuracy Trade-Off Research.'' \emph{Proceedings of the 2021 AAAI/ACM Conference on Artificial Intelligence, Ethics, and Society (AIES '21)}. 2021. \textbf{Oral}~\cite{cooper2021emergent}.

    \item \textbf{A. Feder Cooper}*, Jessica Zosa Forde*, Kweku Kwegyir-Aggrey, Christopher De Sa, and Michael Littman. ``Model Selection's Disparate Impact in Real-World Deep Learning Applications.'' \emph{Workshop on the Science and Engineering of Deep Learning} at \emph{The Ninth International Conference on Learning Representations (ICLR '21)}. 2021. \textbf{Oral}~\cite{forde2021model}.

    \item  \textbf{A. Feder Cooper}*, Ruqi Zhang*, and Christopher De Sa. ``Asymptotically Optimal Exact Minibatch Metropolis-Hastings.'' \emph{Conference on Neural Information Processing Systems 33 (NeurIPS '20)}. 2020. \textbf{Spotlight}~\cite{zhang2020tunamh}.

    \item Ruqi Zhang, \textbf{A. Feder Cooper}, and Christopher De Sa. ``AMAGOLD: Amortized Metropolis Adjustment for Efficient Stochastic Gradient MCMC.'' \emph{Proceedings of the Twenty-third International Conference on Artificial Intelligence and Statistics (AISTATS '20)}. 2020~\cite{zhang2020amagold}. 

    \item \textbf{A. Feder Cooper}. ``Imperfection is the Norm: A Computer Systems Perspective on IoT and Enforcement.'' \emph{(Im)Per\-fect Enforcement Conference}, Information Society Project at Yale Law School. 2019. \textbf{Plenary session}. 
\end{etaremune}

%% file: section/99-appendix/21-hpo/00-app-hpo.tex
\chapter{Appendix for Hyperparameter Deception}\label{chapter:app:hpo}
\setcounter{theorem}{0}
\setcounter{definition}{0}
\input{section/99-appendix/21-hpo/10-app-hpo-glossary}
\input{section/99-appendix/21-hpo/20-app-hpo-prelim}
\input{section/99-appendix/21-hpo/30-app-hpo-epistemic}

\input{section/99-appendix/21-hpo/40-app-hpo-logic}
\input{section/99-appendix/21-hpo/50-app-hpo-defense}
\input{section/99-appendix/21-hpo/60-app-hpo-conclusion}

%% file: section/99-appendix/21-hpo/10-app-hpo-glossary.tex
\vspace{-.35in}
\begingroup
\setlength{\tabcolsep}{8pt} 
\renewcommand{\arraystretch}{1} 
\begin{table}[H]
\hspace{-.5in}
\scriptsize
\begin{center}
      \centering
        \begin{tabular}{p{0.05\linewidth}p{0.52\linewidth}p{0.37\linewidth}}
\toprule
\textbf{Term} & \textbf{Explanation} & \textbf{Example} \\
\midrule
HPO & Acronym for hyperparameter optimization & \; \\
$\mathcal{J}$, $\mathcal{K}$ & Used as examples of arbitrary optimizers & \\
$P$ & Arbitrary atomic proposition & \\
$p$, $q$, $\phi$ & Used as arbitrary or (when specified) specific logical formulas & $p$ = ``Non-adaptive optimizers have higher test accuracy than adaptive optimizers." \\
HP(s) & Acronym for hyperparameter(s) & \\
$\ell \in \mathcal{L}$,  & Log (Definition \ref{def:log}); log set & Figure \ref{tab:sgd-hb}; Log for running HPO using SGD \\
$T$ & The total time it took to run HPO to produce a log $\ell$  & \; \\
$\mathcal{I}$ & A set of integers & Typically the 64-bit integers \\
$r$ & Random seed; $r \in \mathcal{I}$ & \\
$G$ & Pseudo-random number generator; $G(r)$; $G: \mathcal{I} \rightarrow \mathcal{I}^{\infty}$ & \\
PRNG & Acronym for pseudo-random number generator; $G$ & \\
$H$ & HPO procedure (Definition \ref{def:hpo}) & SGD, VGG-16 grid search experiment \\
$H_*$ & A randomized algorithm used in $H$ & Random search \\
$c \in \mathcal{C}$ & Hyper-HP configuration; of set of allowable such configurations for $H_{*}$ & powers-of-2 grid spacing; configurations the demon has access to\\
$\lambda \in \Lambda$; $\lambda^{*}$ & HP config. used to run an HPO pass; of allowable HP configs., determined by $c$; $\lambda^*$ is the output HP config. that performs the best & $\alpha=1$ in Wilson; allowable $\alpha$ values, e.g. $[.001, .01, 1]$  \\
$\mathcal{A}$; $\mathcal{A}_{\lambda}$ & Training algorithm; parameterized by HPs $\lambda$ & SGD; SGD with $\alpha=1$\\
$\mathcal{M}$; $\mathcal{M}_{\lambda}$ & Model; parameterized by HPs $\lambda$ & VGG16 \\
$X$ & A dataset & CIFAR-10 \\
$\alpha$ & Learning rate & Figure \ref{fig:full-our}, $\alpha = 1$ \\
$\epsilon$ & Adam-specific HP & Figure \ref{fig:full-our}, we set $\epsilon = 10^{12}$ \\
EHPO & Epistemic HPO (Definition \ref{def:ehpo}) & Our defended random search in Section \ref{sec:hpo:defense}\\
$\mathcal{H}$ & Set of HPO procedures $H$  &  \\
$\mathcal{P}$ & Set of concluded logical formulas; $p \in \mathcal{P}$ & \\
$\mathcal{F}$ & A function that maps a set of HPO logs $\mathcal{L}$ to a set of logical formulas $\mathcal{P}$  & $\mathcal{F}_*$ (skeptical belief function); $\mathcal{F}_{\text{n}}$ (naive belief function) \\
$\necessary$ & Modal logic operator for ``necessary"  &  $\necessary p$ reads ``It is necessary that $p$\\
$\possible$ & Modal logic operator for ``possible"  &  $\possible p$ reads ``It is possible that $p$\\
$\vdash$ & Indicates a theorem of propositional logic  &  $\vdash Q \rightarrow \necessary Q$ (necessitation) \\
$\possible_t$ & EHPO modal operator (Section \ref{sec:ehpologic}; Definition \ref{def:ehpologic})  &  \\
$\mathcal{B}$ & Belief modal operator  &  $\mathcal{B}_*$ (skeptical belief); $\mathcal{B}_{\text{n}}$ (naive belief)\\
$\sigma \in \Sigma$ & A randomized strategy function that specifies EHPO actions; set of all such strategies (Section \ref{sec:ehpologic}, Definition \ref{def:strategy}) &  \\
$\sigma(\mathcal{L})$ & Distribution over concrete actions for log set &  \\
$\sigma[\mathcal{L}]$ & The logs output from running $\sigma$ on $\mathcal{L}$ &  \\
$\tau_{\sigma}(\mathcal{L})$ & Total time spent executing $\sigma[\mathcal{L}]$ &  \\
$\models$ & Denotes ``models" &  $\mathcal{L}\models\possible_t p$: $\mathcal{L}$ model that $p$ is possible in  $t$ \\
$\gamma$ & Renyi-$\infty$-divergence constant upper bound (Theorem \ref{thm:defendedhpo}) & \\
$K$, $R$ & Numbers of independent random search trials (Section \ref{sec:hpo:defense}) & \\
$\kappa$ & Subsampling size (Algorithm \ref{algo:def}) & We set $\kappa=11$ (Section \ref{sec:hpo:defense}) \\
$M$ & Subsampling budget (Algorithm \ref{algo:def}) & We set $M=10000$ (Section \ref{sec:hpo:defense}) \\
$\delta$ & Skeptical reasoner conclusion threshold (Algorithm \ref{algo:def}) & See Table \ref{table:defense} \\
\bottomrule
\end{tabular}
\end{center}
\end{table}

\newpage

\section{Definitions Reference}

\emph{Hyper-hyperparameters} (hyper-HPs) are HPO-procedure-input values, such as the spacing between different points in the grid for grid search and the distributions to sample from in random search. 

\vspace{.25cm}

\begin{definition} A log $\ell$ records all the choices and measurements made during an HPO run, including the total time $T$ it took to run. It has all necessary information to make the HPO run reproducible.
\end{definition}

\begin{definition}
	An HPO procedure $H$ is a tuple $(H_{*}, \mathcal{C}, \Lambda, \mathcal{A}, \mathcal{M}, G, X)$ where $H_{*}$ is a randomized algorithm, $\mathcal{C}$ is a set of allowable hyper-HPs (i.e., allowable configurations for $H_{*}$), $\Lambda$ is a set of allowable HPs (i.e., of HP sets $\lambda$), $\mathcal{A}$ is a training algorithm (e.g. SGD), $\mathcal{M}$ is a model (e.g. VGG16), $G$ is a PRNG, and $X$ is some dataset (usually split into train and validation sets). When run, $H_{*}$ takes as input a hyper-HP configuration $c \in \mathcal{C}$ and a random seed $r \in \mathcal{I}$, then proceeds to run $\mathcal{A}_{\lambda}$ (on $\mathcal{M}_{\lambda}$ using $G(r)$ and data\footnote{Definition \ref{def:hpo} does not preclude cross-validation, as this can be part of $H_{*}$. The input dataset $X$ can be split in various ways, as a function of the random seed $r$.} from $X$) some number of times for different HPs $\lambda \in \Lambda$. Finally, $H_{*}$ outputs a tuple $(\lambda^*, \ell)$, where $\lambda^*$ is the HP configuration chosen by HPO and $\ell$ is the log documenting the run.
\end{definition}

\begin{definition}
An \textbf{epistemic hyperparameter optimization procedure (EHPO)} is a tuple $(\mathcal{H}, \mathcal{F})$ where $\mathcal{H}$ is a set of HPO procedures $H$ (Definition \ref{def:hpo}) and $\mathcal{F}$ is a function that maps a set of HPO logs $\mathcal{L}$ (Definition \ref{def:log}) to a set of logical formulas $\mathcal{P}$, i.e. $\mathcal{F}(\mathcal{L}) = \mathcal{P}$. An execution of EHPO involves running each $H \in \mathcal{H}$ some number of times (each run produces a log $\ell$) and then evaluating $\mathcal{F}$ on the set of logs $\mathcal{L}$ produced in order to output the conclusions $\mathcal{F}(\mathcal{L})$ we draw from all of the HPO runs.
\end{definition}

\begin{definition}
A randomized \textbf{strategy} $\boldsymbol{\sigma}$ is a function that specifies which action the demon will take. Given $\mathcal{L}$, its current set of logs, $\boldsymbol{\sigma(\mathcal{L})}$ gives a distribution over concrete actions, where each action is either 1) running a new $H$ with its choice of hyper-HPs $c$ and seed $r$ 2) erasing some logs, or 3) returning. We let $\Sigma$ denote the set of all such strategies.
\end{definition}

\begin{definition}
Let $\boldsymbol{\sigma[\mathcal{L}]}$ denote the logs output from executing strategy $\sigma$ on logs $\mathcal{L}$, and let $\boldsymbol{\tau_\sigma(\mathcal{L})}$ denote the total time spent during execution. $\tau_\sigma(\mathcal{L})$ is equivalent to the sum of the times $T$ it took each HPO procedure $H \in \mathcal{H}$ executed in strategy $\sigma$ to run.
Note that both $\sigma[\mathcal{L}]$ and $\tau_\sigma(\mathcal{L})$ are random variables, as a function of the randomness of selecting $G$ and the actions sampled from $\sigma(\mathcal{L})$. For any formula $p$ and any $t \in \mathbb{R}_{>0}$, we say $\mathcal{L} \models \possible_t p$, i.e. ``$\mathcal{L}$ models that it is possible $p$ in time $t$,'' if 
\[
    \text{there exists a strategy } \sigma \in \Sigma, \text{ such that } \;\; \mathbb{P}(\sigma[\mathcal{L}] \models p) = 1 \; \text{ and } \; \mathbb{E}[\tau_\sigma(\mathcal{L})] \le t.
\]
\end{definition}

\begin{definition}
For any formula $p$, we say $\mathcal{L} \models \mathcal{B} p$, ``$\mathcal{L}$ models our belief in $p$'', if $p \in \mathcal{F}(\mathcal{L})$.
\end{definition}

\begin{definition}
Suppose that we are given a naive EHPO procedure $(\{H\}, \mathcal{F}_{\text{n}})$, in which $H$ is random search and is the only HPO in our EHPO, and $\mathcal{F}_{\text{n}}$ is a ``naive'' belief function associated with a naive reasoner $\mathcal{B}_{\text{n}}$. 
For any $K, R \in \mathbb{N}$, we define the ``$(K,R)$-defended'' belief function $\mathcal{F}_*$ for a skeptical reasoner $\mathcal{B}_{*}$ as the following conclusion-drawing procedure.
First, $\mathcal{F}_*$ only makes conclusion set $\mathcal{P}_*$ from a single log $\hat \ell$ with $K*R$ trials; otherwise, it concludes nothing, outputting $\emptyset$.
Second, $\mathcal{F}_{*}$ splits the single $\hat \ell$ into $R$ logs $\ell_1, \ell_2, \ldots, \ell_R$, each containing $K$ independent-random-search trials.\footnote{This is not generally allowable. $\mathcal{F}_*$ can do this because random-search logs contain interchangeable trials.}
Finally, $\mathcal{F}_{*}$ outputs the intersection of what the naive reasoner would have output on each log $\ell_i$,
\[
    \mathcal{F}_{*}(\{\hat \ell\}) = \mathcal{P}_* \equiv \mathcal{F}_{\text{n}}(\{ \ell_1 \})
    \cap \mathcal{F}_{\text{n}}(\{ \ell_2 \}) \cap \cdots \cap \mathcal{F}_{\text{n}}(\{ \ell_R \}).
\]
Equivalently, $\{\hat \ell\} \models \mathcal{B_{*}}p$ only if $\{\ell_i\} \models \mathcal{B}_{\text{n}}p$ for all $i$.
\end{definition}

%% file: section/99-appendix/21-hpo/20-app-hpo-prelim.tex
\section{Section \ref{sec:hpo:prelim} Appendix: Notes on the Preliminaries} \label{app:sec:hpo:prelim}

The code for running these experiments can be found at \url{https://github.com/pasta41/deception}.

\subsection{Empirical Deception Illustration using Wilson et al.\cite{wilson2017marginal}}

\subsubsection{Why we chose Wilson et al.~\cite{wilson2017marginal}}

We elaborate on why we specifically chose Wilson et al.~\cite{wilson2017marginal} as our running example of hyperparameter deception. There are four main reasons why we thought this was the right example to focus on for an illustration: 

First, the experiment involves optimizers known across ML (e.g. SGD, Adam), a model frequently used for benchmark tasks (VGG16) and a commonly-used benchmark dataset (CIFAR-10). Unlike other examples of hyperparameter deception, one does not need highly-specialized domain knowledge to understand the issue \cite{dodge2019nlp, Lucic2018-dr}. Second, the paper is exceptionally well-cited and known in the literature, so many folks in the community are familiar with its results. Third, we were certain that we could demonstrate hyperparameter deception before we ran our experiments; we observe that Adam's update rule basically simulates Heavy Ball when its $\epsilon$ parameter is set high enough. So, we were confident that we could (at the very least) get Adam to perform as well as Heavy Ball via changing hyper-HPs, which would demonstrate hyperparameter deception. We then found further support for this observation in concurrent work \cite{sivaprasad2020hpo}, which cited earlier work \cite{choi2019empirical} that also observes this. Fourth, the claim in Wilson et al.~\cite{wilson2017marginal} is fairly broad. They make a claim about adaptive vs. non-adaptive optimizers, more generally. If the claim had been narrower -- about small $\epsilon$ values for numerical stability, then perhaps hyperparameter deception would not have occurred. In general, we note that narrower claims could help avoid deception. 

\subsection{Expanded empirical results}

We elaborate on the results we present in Section \ref{sec:hpo:prelim}.

\subsubsection{Experimental setup}\label{sec:wilson:setup}

We replicate and run a variant of Wilson et al.~\cite{wilson2017marginal}'s VGG16 experiment on CIFAR-10, using SGD, Heavy Ball, and Adam as the optimizers. 

We launch each run on a local machine configured with a 4-core 2.6GHz Inter (R) Xeon(R) CPU, 8GB memory and an NIVIDIA GTX 2080Ti GPU. 
Following the exact configuration from Wilson et al.~\cite{wilson2017marginal}, we set the mini-batch size to be 128, the momentum constant to be 0.9 and the weight decay to be 0.0005. 
The learning rate is scheduled to follow a linear rule: The learning rate is decayed by a factor of 10 every 25 epochs. The total number of epochs is set to be 250.
For the CIFAR-10 dataset, we apply random horizontal flipping and normalization. Note that Wilson et al.~\cite{wilson2017marginal} does not apply random cropping on CIFAR-10; thus we omit this step to be consistent with their approach. We adopt the standard cross entropy loss.
For each HPO setting, we run 5 times and average the results and include error bars two standard deviations above and below the mean. 


\subsubsection{Associated results and logs}\label{app:sec:logs}

\begin{figure}[t]
\centering
\includegraphics[scale=.35]{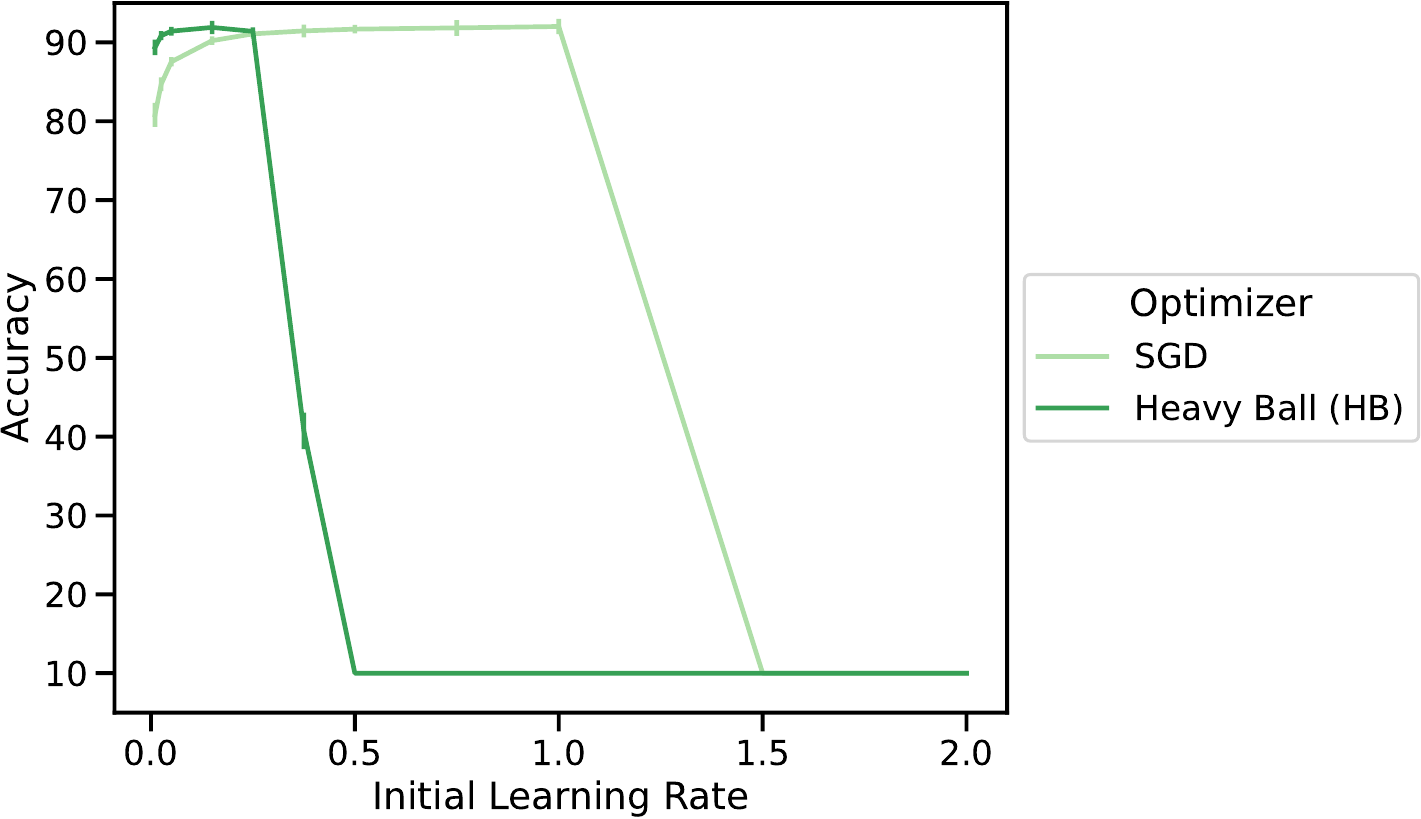}
\caption{Full test accuracy results of VGG-16 on CIFAR-10 for SGD and Heavy Ball learning rate ($\alpha$) HPO. Error bars indicate two standard deviations above and below the mean. Each HPO setting is measured with five replicates. We achieve similar performance as Wilson et al.~\cite{wilson2017marginal}.}
\label{fig:full-our}
\end{figure}

\begin{figure}[t]
\includegraphics[scale=.29]{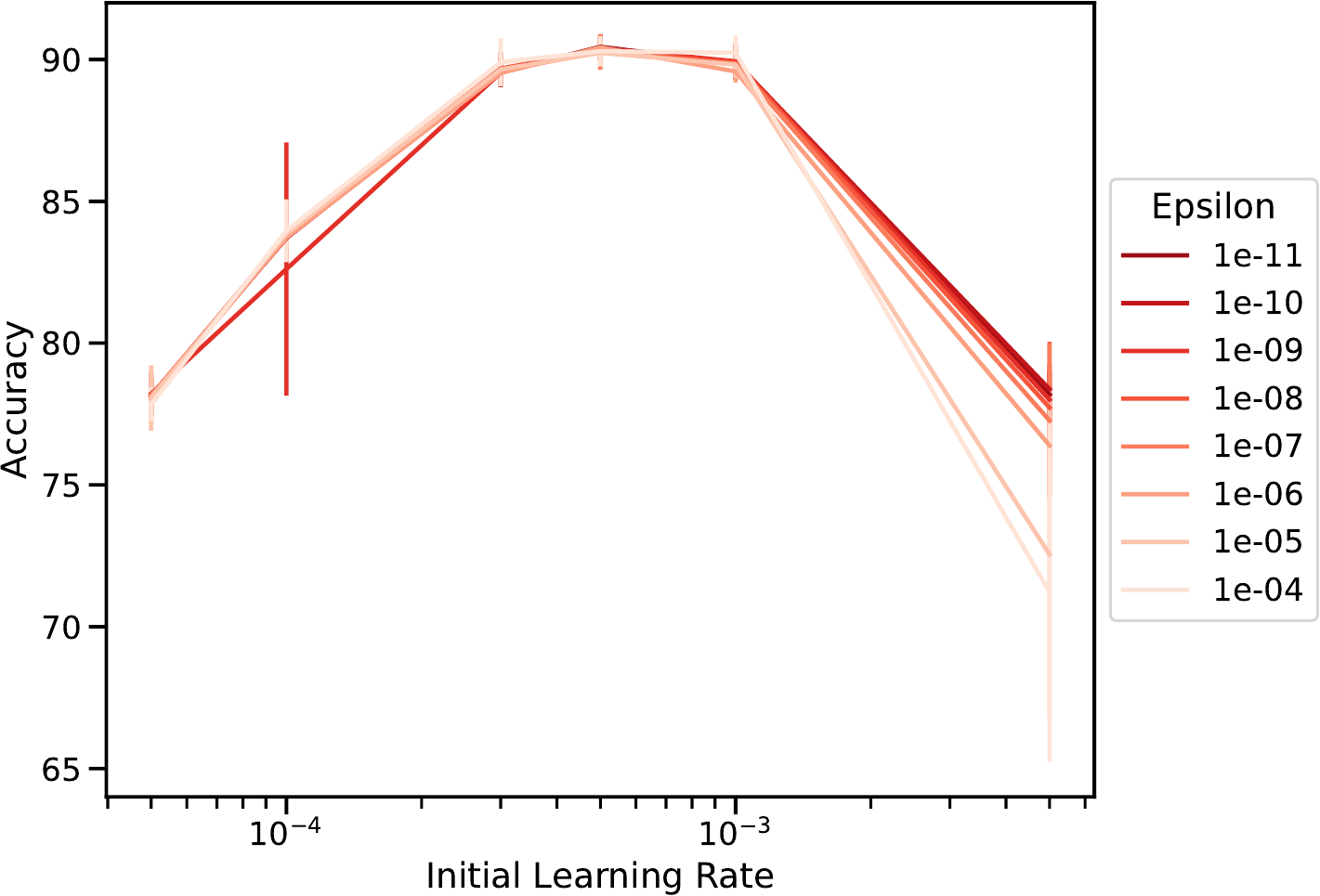}
\includegraphics[scale=.29]{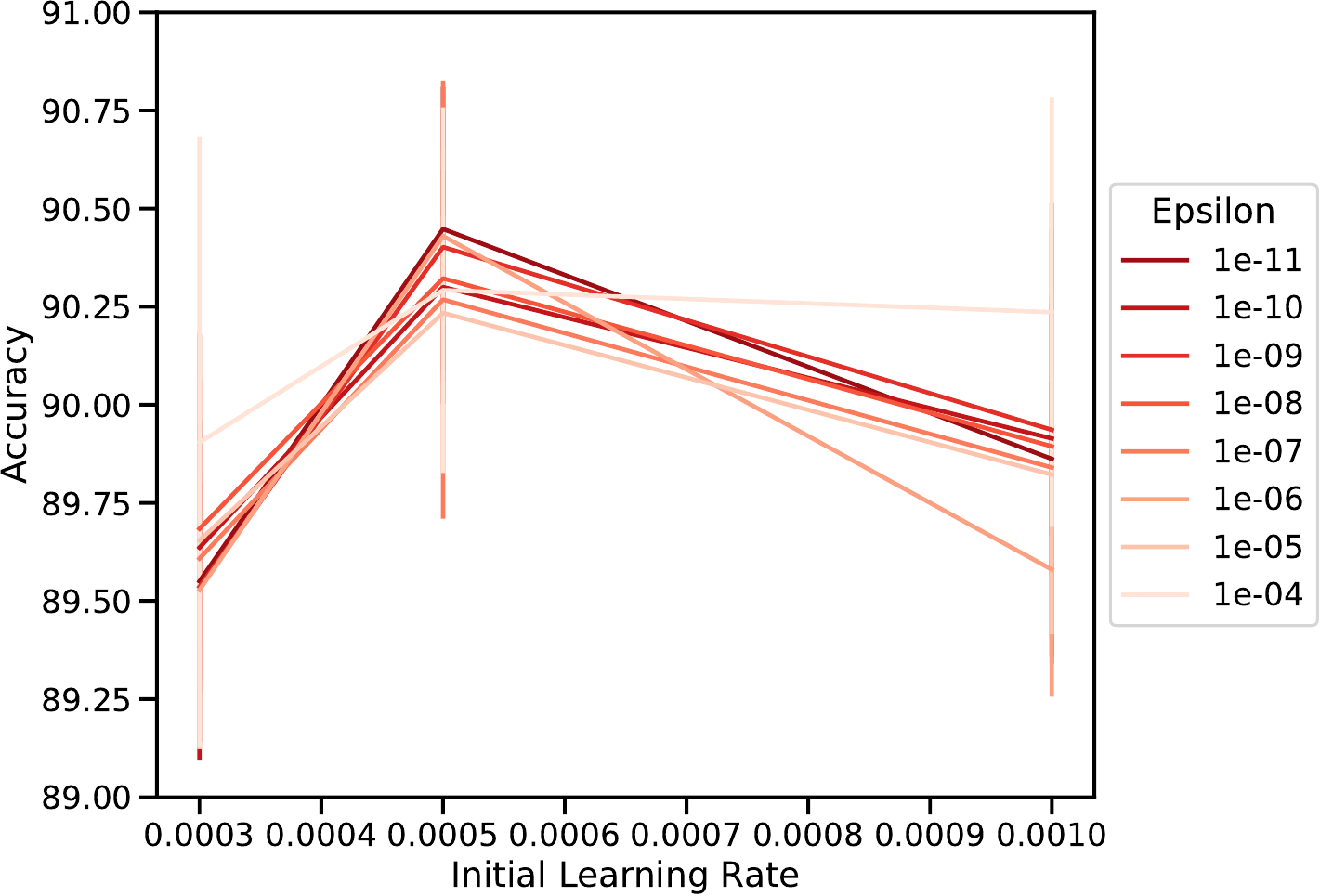}
\caption{Tuning over learning rate for different small values of $\epsilon$. On the left, we show a wide range of learning rates tested. On the right, we zoom in on the portion of results where the best test accuracy occurs. These results reflect what Wilson et al.~\cite{wilson2017marginal} showed, but with tuning over $\epsilon$ (small values).  Each HP setting is used to train VGG-16 on CIFAR-10 five times, and the error bars represent two standard deviations above and below the mean test accuracy.}
\label{fig:full-wilson}
\end{figure}

\begin{figure}[t]
\centering
\includegraphics[scale=.35]{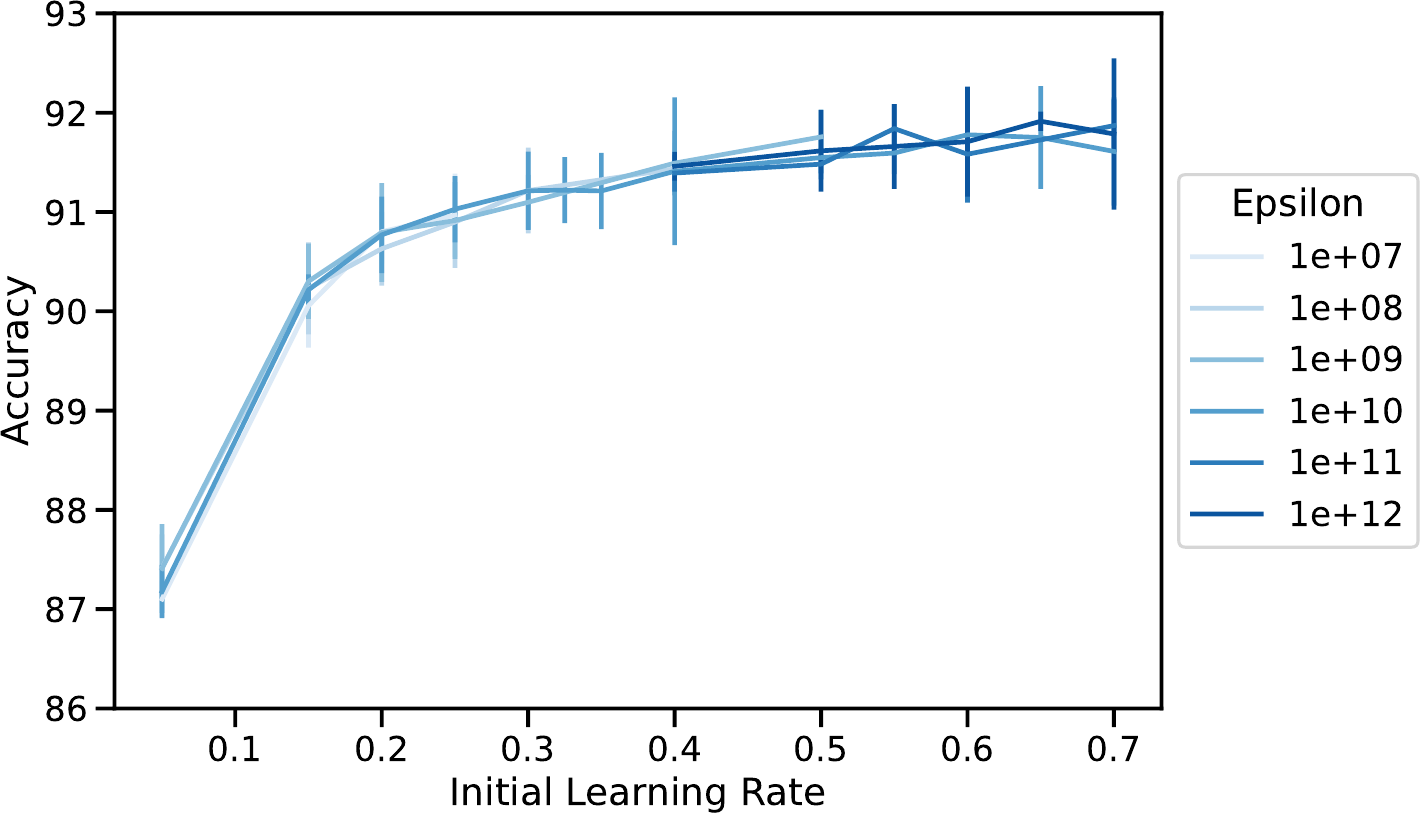}
\caption{Results for our expanded search over large $\epsilon$ values for Adam. We show test accuracy on CIFAR-10 as a function of different learning rates $\alpha$ for the different large $\epsilon$ values. Error bars show two standard deviations above and below mean test accuracy for five replicates for each HP setting.}
\label{fig:full-our-Adam}
\end{figure}

\begin{figure}[t]
\includegraphics[scale=.5]{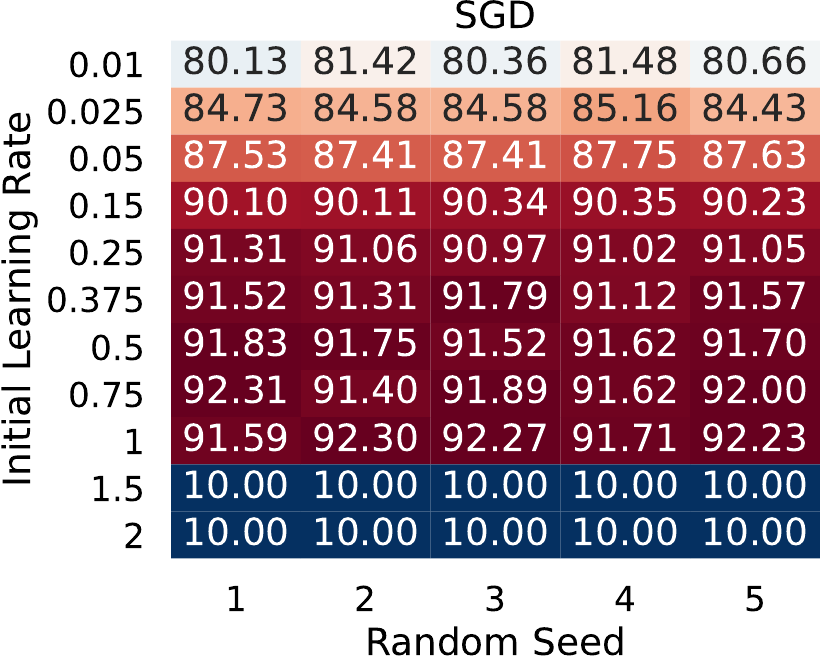}
\includegraphics[scale=.5]{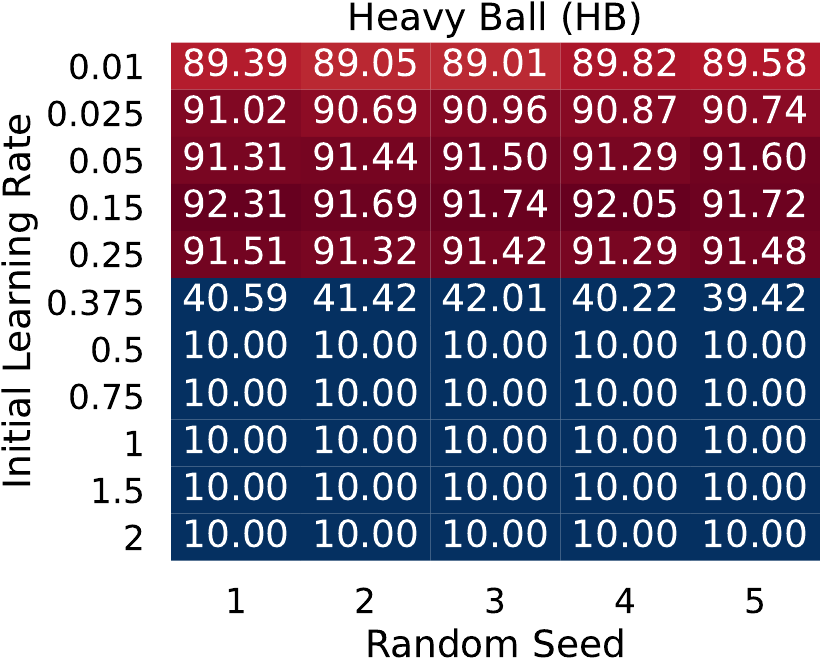}
\caption{Heatmap logs of test accuracy of VGG-16 on CIFAR-10 for SGD and Heavy Ball for each initial learning rate and random seed. These logs correspond to the results graphed in Figure \ref{fig:full-our}.}
\label{tab:sgd-hb}
\end{figure}

\begin{figure}[t]

\includegraphics[scale=.5]{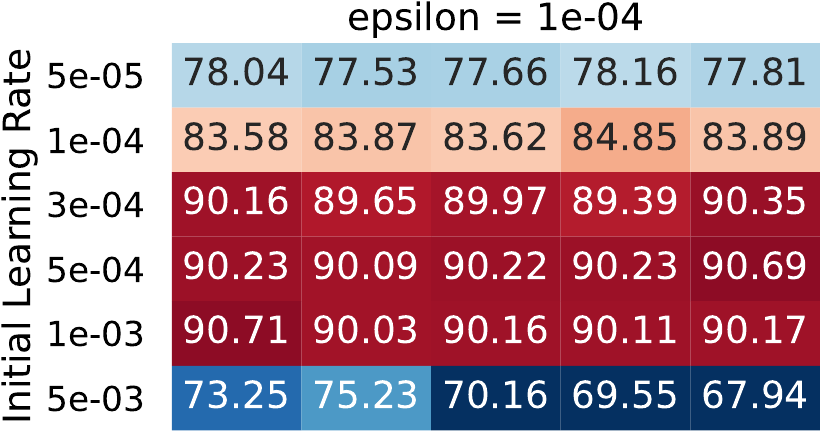}
\includegraphics[scale=.5]{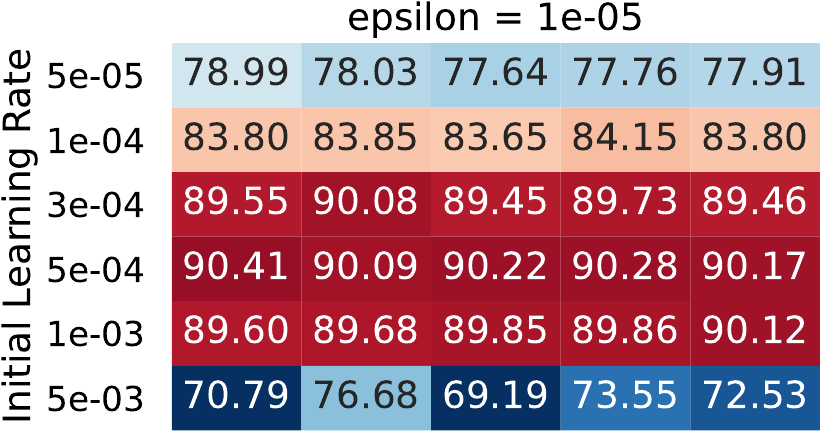}

\includegraphics[scale=.5]{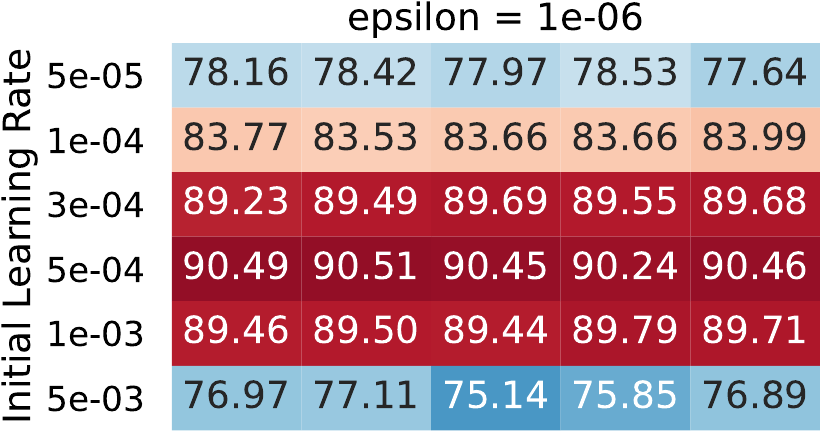}
\includegraphics[scale=.5]{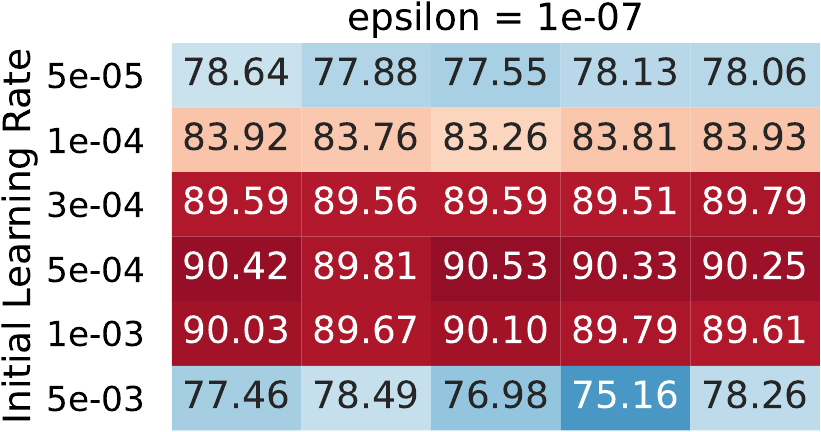}

\includegraphics[scale=.5]{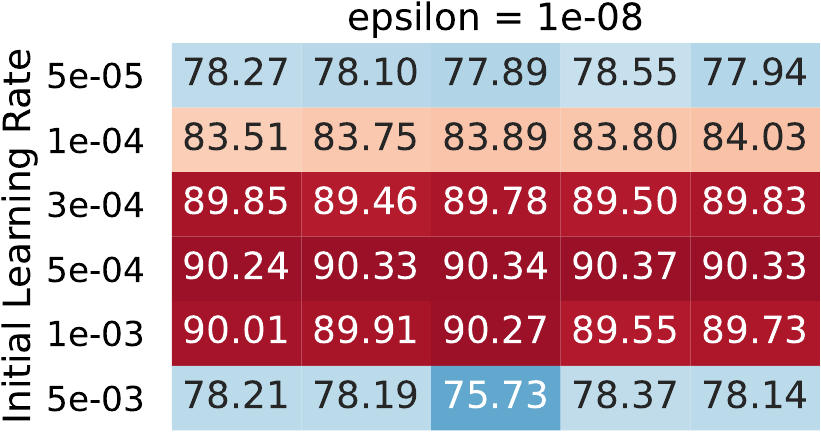}
\includegraphics[scale=.5]{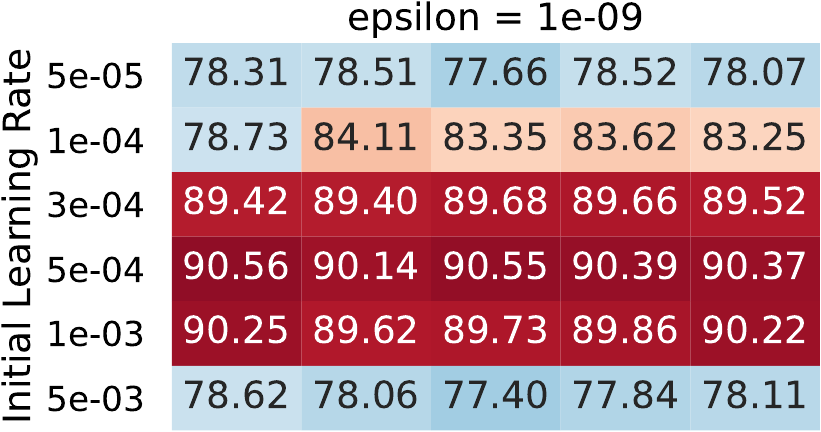}

\includegraphics[scale=.5]{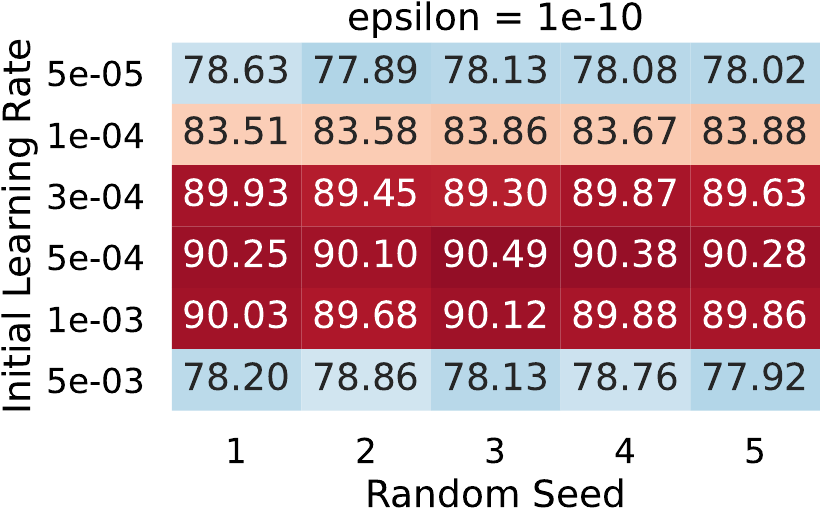}
\includegraphics[scale=.5]{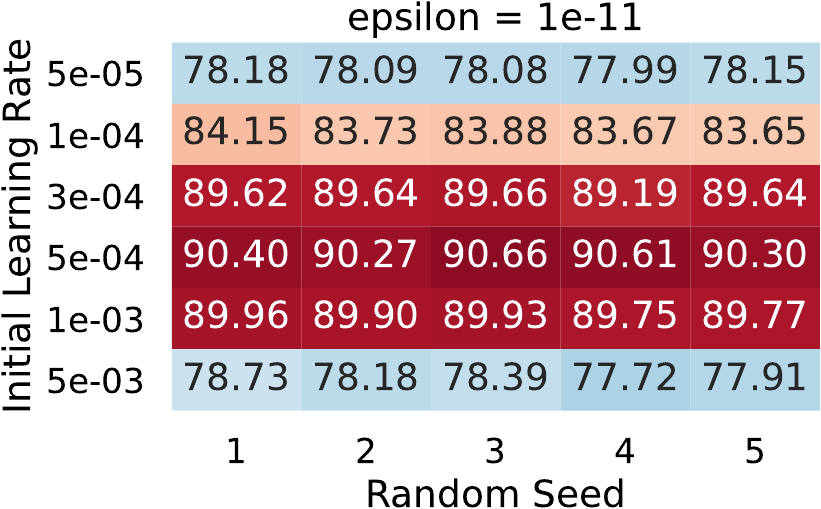}
\caption{Heatmap logs of test accuracy of VGG-16 on CIFAR-10 for Adam for each initial learning rate and random seed for different small values of Adam's $\epsilon$ HP. These results correspond to those graphed in Figure \ref{fig:full-wilson}.}
\label{tab:wilson-adam-1}
\end{figure}

\begin{figure}[t]

\includegraphics[scale=.5]{figure/21-hpo/adam_tuned_table_1e-04-crop.pdf}
\includegraphics[scale=.5]{figure/21-hpo/adam_tuned_table_1e-05-crop.pdf}

\includegraphics[scale=.5]{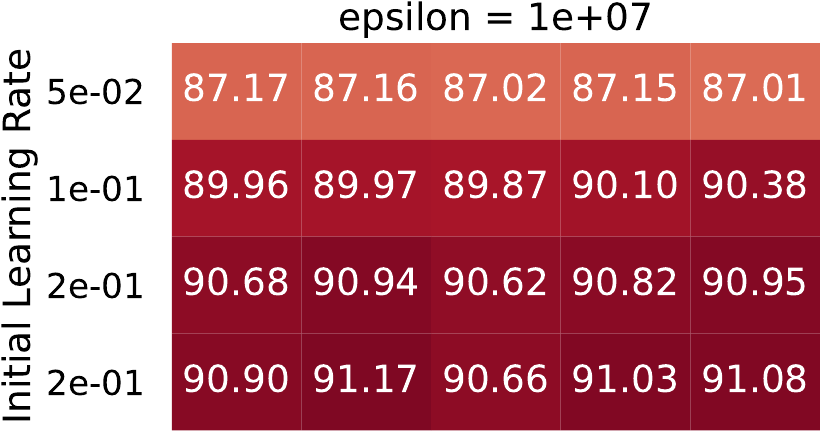}
\includegraphics[scale=.5]{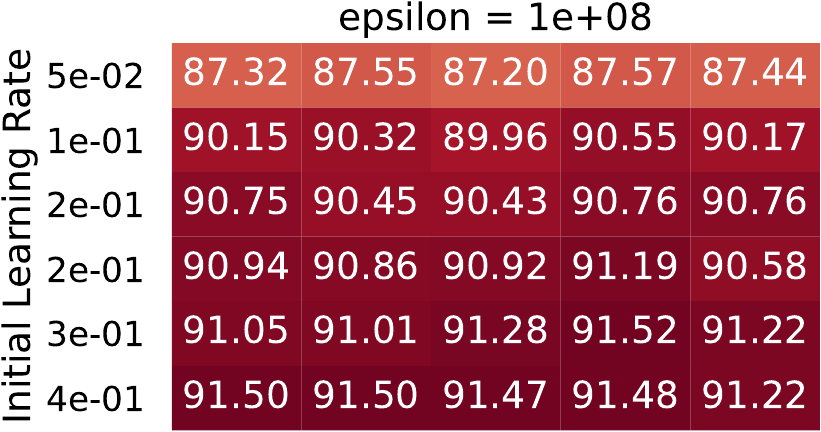}

\includegraphics[scale=.5]{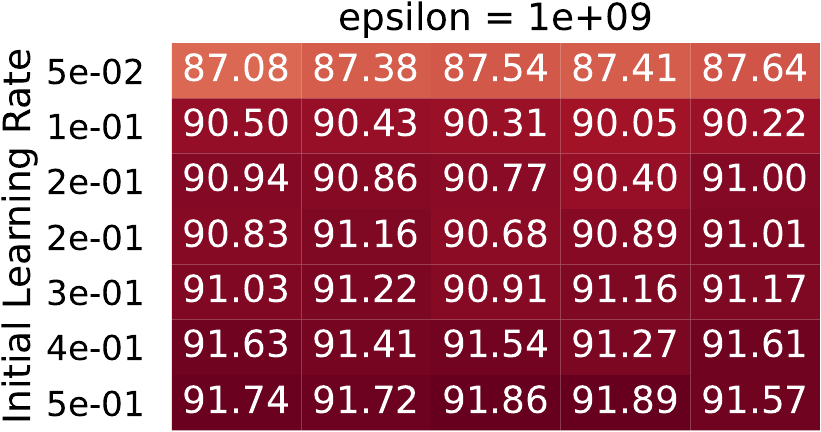}
\includegraphics[scale=.5]{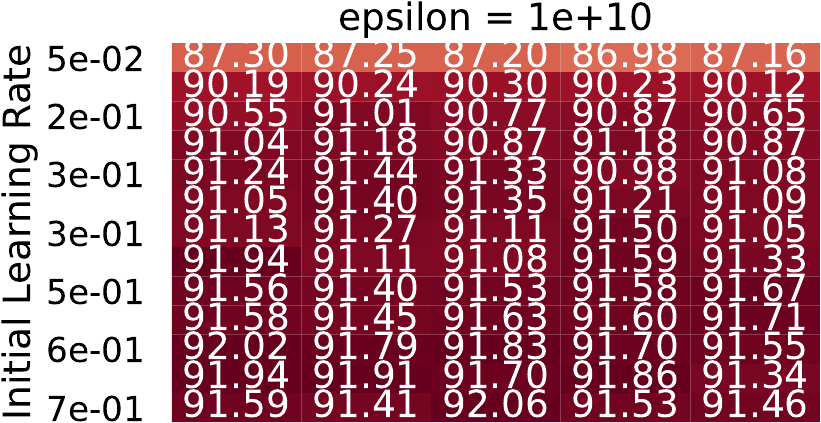}

\includegraphics[scale=.5]{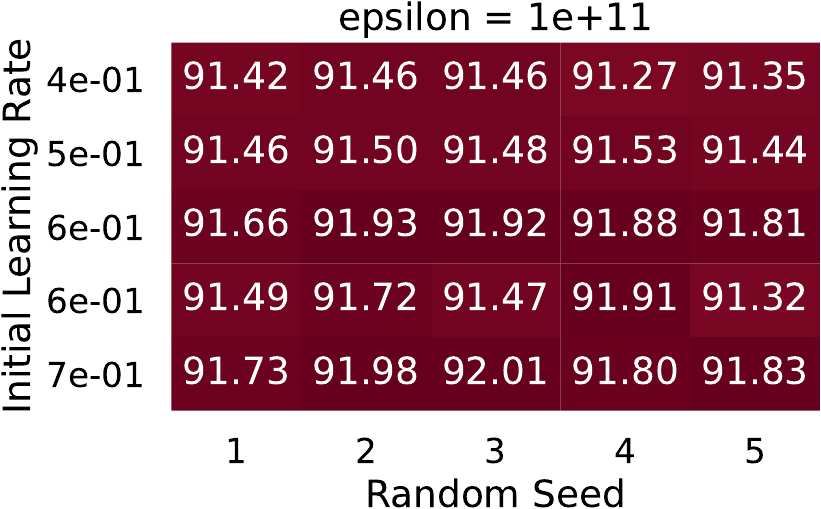}
\includegraphics[scale=.5]{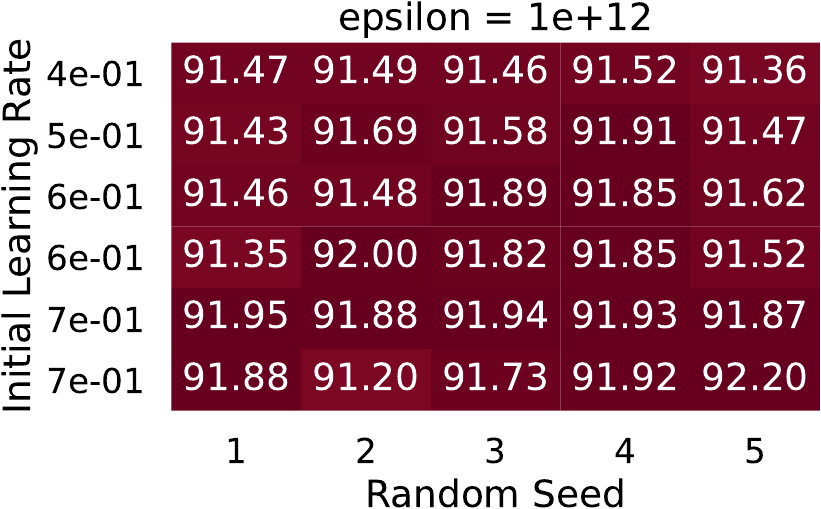}
\caption{Heatmap logs of test accuracy of VGG-16 on CIFAR-10 for Adam for each initial learning rate and random seed for different values of Adam's $\epsilon$ using our expanded search space. These logs reflect the results graphed in Figure \ref{fig:full-our-Adam}.}
\label{tab:wilson-adam-2}
\end{figure}

In line with our notion of a \emph{log} (Definition \ref{def:log}), we provide data tables (Figures \ref{tab:sgd-hb}, \ref{tab:wilson-adam-1}, and \ref{tab:wilson-adam-2}) that correspond with our results graphed in the Figures \ref{fig:full-our}, \ref{fig:full-wilson}, \ref{fig:full-our-Adam}.

\FloatBarrier

\newpage
\subsection{Empirical Deception Illustration using Merity et al.~\cite{merity2016pointer}}
In addition to the computer vision experiments of Wilson et al.~\cite{wilson2017marginal}, we also show a separate line of experiments from NLP: training an LSTM on Wikitext-2 using Nesterov and Heavy Ball as the optimizers. We illustrate deception (i.e., the possibility of drawing inconsistent conclusions) using two different sets of hyper-HPs to configure HPO grids for tuning the learning rate. We run ten replicates for each optimizer / grid combination (a total of 40 runs). 
We run these experiments using the same hardware as described in Appendix \ref{sec:wilson:setup}.

\begin{figure}[t!]
 \begin{center}

    \includegraphics[width=0.44\textwidth]{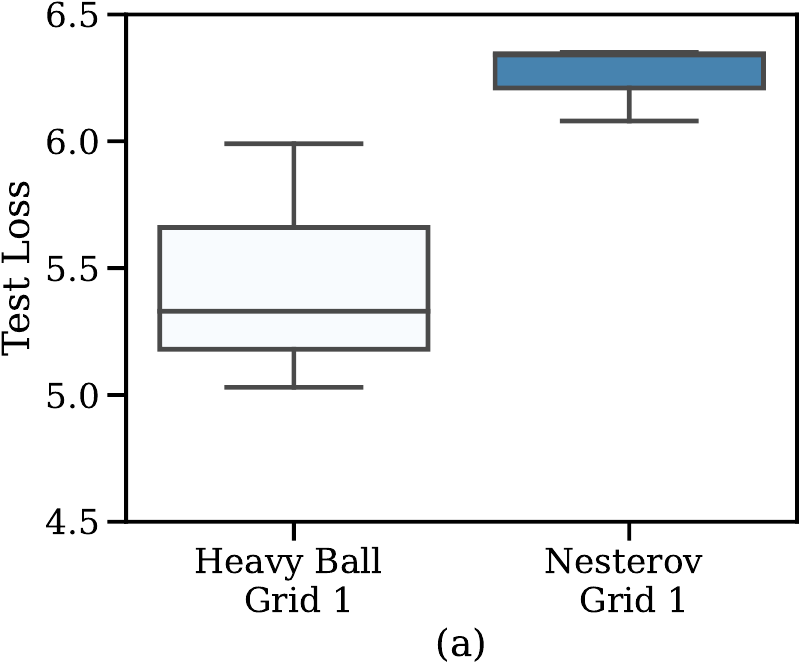} \hspace{.5cm}
\includegraphics[width=0.44\textwidth]{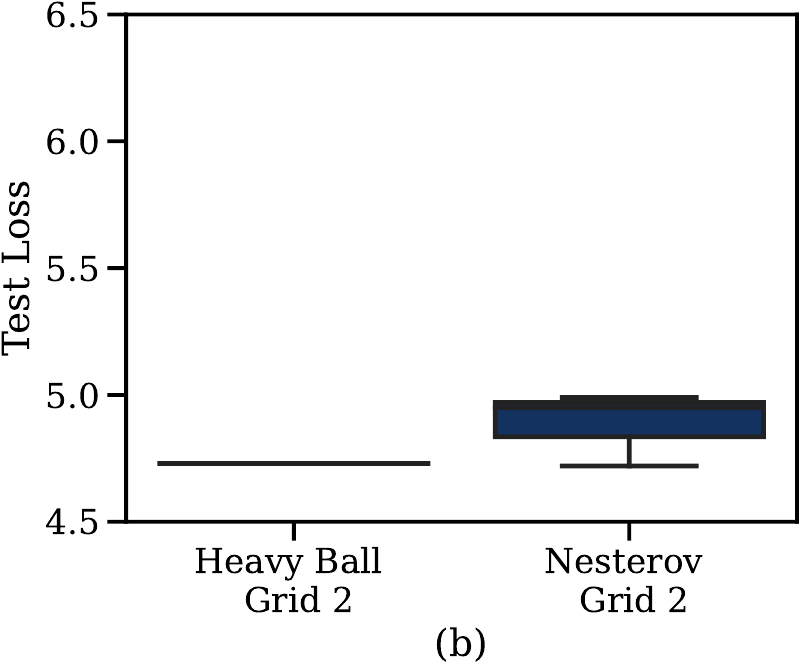}
 \vspace{-.2cm}

    \caption{Demonstrating the possibility of drawing inconsistent conclusions from HPO (what we shorthand \emph{hyperparameter deception}) LSTM on Wikitext-2 using Nesterov and Heavy Ball as the optimizers. Each box plot represents a log. In (a), we use the grid $\alpha={1,5,10,15,20,25,30,35,40}$, from which we can reasonably conclude that Nesterov outperforms HB. In (b), we use the grid $\alpha={10,20,30,40}$, from which we can reasonably conclude that HB outperforms Nesterov.}
    \label{fig:merity_deception}
  \end{center}
\end{figure}

%% file: section/99-appendix/21-hpo/30-app-hpo-epistemic.tex
\section{Section \ref{sec:hpo:ehpo} Appendix: Epistemic Hyperparameter Optimization} \label{app:sec:ehpo}

\subsection{Additional concrete interpretations of EHPO}

For concision, in the main text we focus on examples of EHPO procedures that compare the performance of different optimizers. However, it is worth noting that our definition of EHPO (Definition \ref{def:ehpo}) is more expansive than this setting. For example, it is possible to run EHPO to compare different models (perhaps, though not necessarily, keeping the optimizer fixed), to draw conclusions about the relative performance of different models on different learning tasks. 

\subsection{Descartes' Evil Demon Thought Experiment}

Our formalization was inspired by Descartes' evil genius/demon thought experiment. This experiment more generally relates to his use of systematic doubt in \emph{The Meditations} more broadly. It is this doubt/skepticism (and its relationship to possibility) that we find useful for the framing of an imaginary, worst-case adversary. In particular, we draw on the following quote, from which we came up with the term \emph{hyperparameter deception}:

\begin{quote}
    \emph{I will suppose...an evil genius, supremely powerful and clever, who has directed his entire effort at deceiving me. I will regard the heavens, the air, the earth, colors, shapes, sounds, and all external things as nothing but the bedeviling hoaxes of my dreams, with which he lays snares for my credulity...even if it is not within my power to know anything true, it certainly is within my power to take care resolutely to withhold my assent to what is false, lest this deceiver, however powerful, however clever he may be, have any effect on me.} -- Descartes
\end{quote}

\begin{figure}[H]
  \begin{center}
    \includegraphics[width=.2\textwidth]{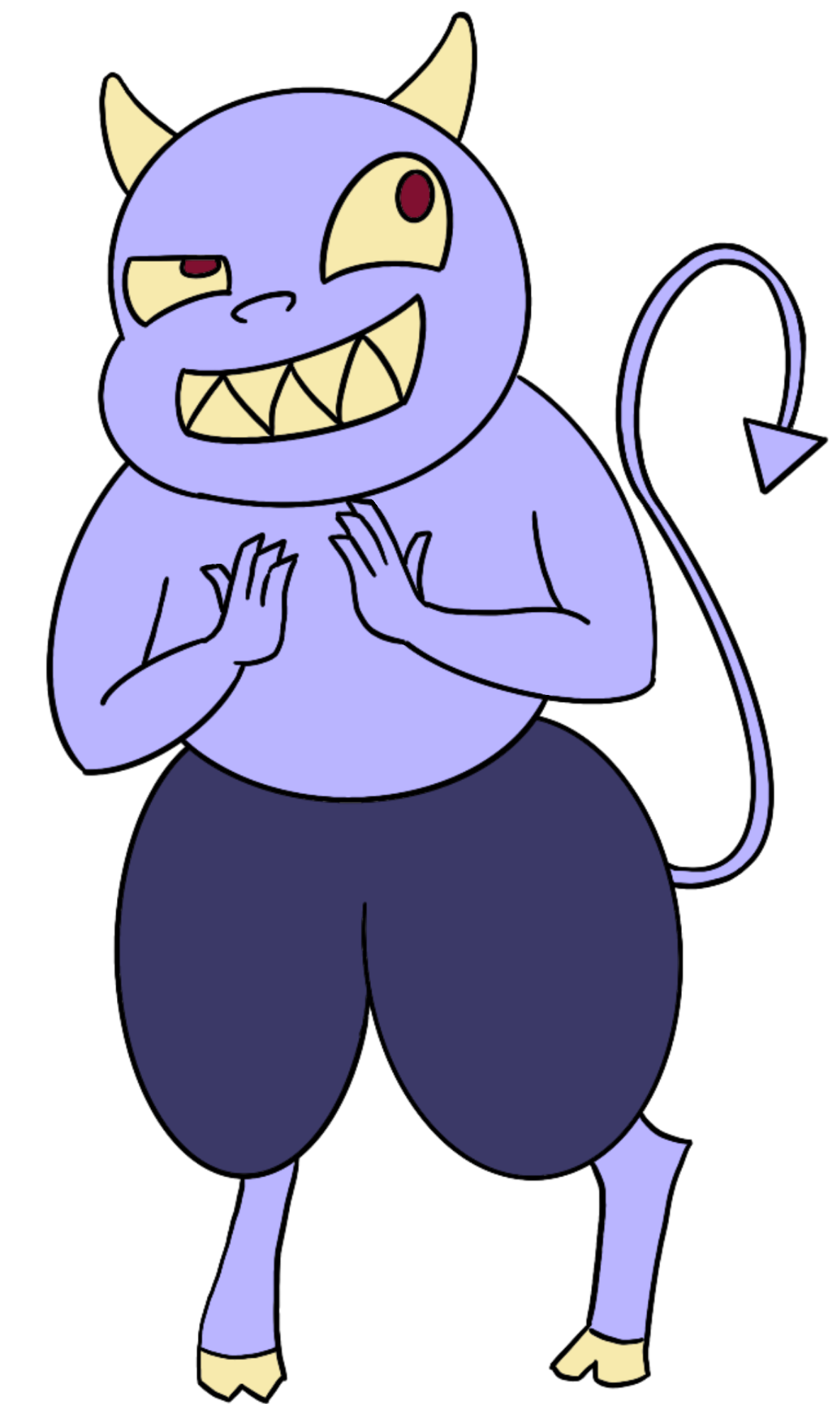}
    \caption{How the authors imagine the EHPO-running demon}
  \end{center}
\end{figure}

For more on the long (and rich) history of the use of imaginary demons and devils as adversaries --- notably a different conception of an adversary than the potential real threats posed in computer security research --- we refer the reader to Canales~\cite{canales2020devil}.

%% file: section/99-appendix/21-hpo/40-app-hpo-logic.tex
\section{Section \ref{sec:hpo:formalizing} Appendix: Modal Logic Formalization} \label{app:sec:logic}

\subsection{Further Background on Modal Logic}

We first provide the necessary background on modal logic, which will inform the proofs in this appendix (Appendix \ref{app:sec:kripkeaxioms}). We then describe our possibility logic---a logic for representing the possible results of the evil demon running EHPO---and prove that it is a valid modal logic (Appendix \ref{app:sec:ehpologic}). We then present a primer on modal belief logic (Appendix \ref{app:sec:belief}), and suggest a proof for the validity of combining our modal possibility logic with modal belief logic (Appendix \ref{app:sec:multimodal}).

\subsubsection{Axioms from Kripke Semantics} \label{app:sec:kripkeaxioms}
Kripke semantics in modal logic inherits all of the the axioms from propositional logic, which assigns values  $T$ and $F$ to each atom $p$, and adds two operators, one for representing \emph{necessity} ($\necessary$)  and one for \emph{possibility} ($\possible$).
\begin{itemize}
    \item $\necessary p$ reads ``It is necessary that $p$".
    \item $\possible p$ reads ``It is possible that $p$".
\end{itemize}

The $\possible$ operator is just syntactic sugar, as it can be represented in terms of $\lnot$ and $\necessary$:

\begin{equation} \label{app:eq:sugar}
    \possible p \equiv \lnot \necessary \lnot p
\end{equation}
which can be read as:
\begin{center}
``It is possible that $p$" is equivalent to ``It is not necessary that not $p$."
\end{center}

The complete set of rules is as follows:
\begin{itemize}
    \item Every atom $p$ is a sentence.
    \item If $D$ is a sentence, then 
    \begin{itemize}
        \item $\lnot D$ is a sentence.
        \item $\necessary D$ is a sentence.
        \item $\possible D$ is a sentence.
    \end{itemize}
    \item If $D$ and $E$ are sentences, then
    \begin{itemize}
        \item $D \land E$ is a sentence.
        \item $D \lor E$ is a sentence.
        \item $D \rightarrow E$ is a sentence.
        \item $D \leftrightarrow E$ is a sentence
    \end{itemize}
    \item $\necessary(\mathcal{D} \rightarrow \mathcal{E}) \rightarrow (\necessary \mathcal{D} \rightarrow \necessary \mathcal{E})$ (Distribution)
    \item $\mathcal{D} \rightarrow \necessary \mathcal{D}$ (Necessitation)
\end{itemize}

\subsubsection{Possible Worlds Semantics} \label{app:sec:pw}
Modal logic introduces a notion of \emph{possible worlds}. Broadly speaking, a possible world represents the state of how the world \emph{is} or potentially \emph{could be} \cite{chellas1980book, garson2018stanford}. Informally, $\necessary D$ means that $D$ is true at \emph{every} world (Equation \ref{app:eq:nec}); $\possible D$ means that $D$ is true at \emph{some} world (Equation \ref{app:eq:poss}).

Possible worlds give a different semantics from more familiar propositional logic. In the latter, we assign truth values $\{T, F\}$ to propositional variables $p \in \mathcal{P}$, from which we can construct and evaluate sentences $D \in \mathcal{D}$ in a truth table. In the former, we introduce a set of possible worlds, $\mathcal{W}$, for which each $w \in \mathcal{W}$ has own truth value for each $p$. This means that the value of each $p$ can differ across different worlds $w$. Modal logic introduces the idea of valuation function,
\begin{equation*}
    \mathcal{V}: (\mathcal{W}\times \mathcal{D}) \rightarrow \{T, F\}
\end{equation*}
to assign truth values to logical sentences at different worlds. This in turn allows us to express the formulas, axioms, and inference rules of propositional logic in terms of $\mathcal{V}$. For example,
\begin{equation*}
    \mathcal{V}(w, \lnot D) = T \leftrightarrow \mathcal{V}(w, D) = F
\end{equation*}

There are other rules that each correspond to a traditional truth-table sentence evaluation, but conditioned on the world in which the evaluation occurs. We omit these for brevity and refer the reader to Challas~\cite{chellas1980book}.

We do include the valuation rules for the $\necessary$ and $\possible$ operators that modal logic introduces (Equations \ref{app:eq:nec} \& \ref{app:eq:poss}). To do so, we need to introduce one more concept: The accessibility relation, $\mathcal{R}$. $\mathcal{R}$ provides a frame of reference for one particular possible world to access other possible worlds; it is a way from moving from world to world. So, for an informal example, $\mathcal{R}w_1w_2$ means that $w_2$ is possible relative to $w_1$, i.e. we can reach $w_2$ from $w_1$. Such a relation allows for a world to be possible relative potentially to some worlds but not others. More formally, 
\begin{align*}
R \subseteq \mathcal{W} \times \mathcal{W}
\end{align*}

Overall, the important point is that we have a collection of worlds $\mathcal{W}$, an accessibility relation $\mathcal{R}$, and a valuation function $\mathcal{V}$, which together defines a Kripke model, which captures this system:
\begin{align*}
    \mathcal{M} = \langle \mathcal{W}, \mathcal{R}, \mathcal{V} \rangle
\end{align*}

Finally, we can give the valuation function rules for $\necessary$ and $\possible$: 
\begin{equation} \label{app:eq:nec}
    \mathcal{V}(w, \necessary D) = T \leftrightarrow \forall w', (\mathcal{R}ww' \rightarrow \mathcal{V}(w', D) = T)
\end{equation}
\begin{equation} \label{app:eq:poss}
    \mathcal{V}(w, \possible D) = T \leftrightarrow \exists w', (\mathcal{R}ww' \land \mathcal{V}(w', D) = T)
\end{equation}

Informally, for $\necessary D$ to be true in a world, it must be true in every possible world that is reachable by that world. For $\possible D$ to be true in a world, it must be true in some possible world that is reachable by that world.

\subsection{Our Multimodal Logic Formulation} \label{app:sec:ourlogic} 

\subsubsection{A Logic for Reasoning about the Conclusion of EHPO} \label{app:sec:ehpologic}
As in Section \ref{sec:hpo:formalizing}, we can define the well-formed formulas of our indexed modal logic\footnote{For an example of another indexed modal logic concerning probability, please refer to Heifetz and Mongin~\cite{heifeitz1998indexed}.} recursively in Backus-Naur form, where $t$ is any real number and $P$ is any atomic proposition
\begin{align} \label{app:eq:ehposyntax}
    \kappa \coloneqq P\;|\;\lnot \kappa\;|\;\kappa \land \kappa \;|\; \possible_t \kappa
\end{align}

where $\kappa$ is a well-formed formula.

As we note in Section \ref{sec:hpo:formalizing}, where we first present this form of defining modal-logic, $\necessary$ is syntactic sugar, with $\necessary p \equiv \lnot \possible \lnot p$ (which remains true for our indexed modal logic). Similarly, ``or'' has $p \lor q \equiv \lnot (\lnot p \land \lnot q)$ and ``implies'' has $p \rightarrow q \equiv \lnot p \lor q$, which is why we do not include them for brevity in this recursive definition.

We explicitly define the relevant semantics for $\possible_t$ for reasoning about the demon's behavior in running EHPO. For clarity, we replicate that definition of the semantics of expressing the possible outcomes of EHPO conducted in bounded time (Definitions \ref{def:strategy} \& \ref{def:ehpologic}, respectively) below:

\begin{definition*}
A randomized \textbf{strategy} $\boldsymbol{\sigma}$ is a function that specifies which action the demon will take. Given $\mathcal{L}$, its current set of logs, $\boldsymbol{\sigma(\mathcal{L})}$ gives a distribution over concrete actions, where each action is either 1) running a new $H$ with its choice of hyper-HPs $c$ and seed $r$ 2) erasing some logs, or 3) returning. We let $\Sigma$ denote the set of all such strategies.
\end{definition*}

We can now define what the demon can reliably bring about, in terms of executing a strategy in bounded time:

\begin{definition*}
Let $\boldsymbol{\sigma[\mathcal{L}]}$ denote the logs output from executing strategy $\sigma$ on logs $\mathcal{L}$, and let $\boldsymbol{\tau_\sigma(\mathcal{L})}$ denote the total time spent during execution. $\tau_\sigma(\mathcal{L})$ is equivalent to the sum of the times $T$ it took each HPO procedure $H \in \mathcal{H}$ executed in strategy $\sigma$ to run.
Note that both $\sigma[\mathcal{L}]$ and $\tau_\sigma(\mathcal{L})$ are random variables, as a function of the randomness of selecting $G$ and the actions sampled from $\sigma(\mathcal{L})$. For any formula $p$ and any $t \in \mathbb{R}_{>0}$, we say $\mathcal{L} \models \possible_t p$, i.e. ``$\mathcal{L}$ models that it is possible $p$ in time $t$,'' if 
\[
    \text{there exists a strategy } \sigma \in \Sigma, \text{ such that } \;\; \mathbb{P}(\sigma[\mathcal{L}] \models p) = 1 \; \text{ and } \; \mathbb{E}[\tau_\sigma(\mathcal{L})] \le t.
\]
\end{definition*}

\subsubsection{A Possible Worlds Interpretation} \label{app:sec:pw_interpretation}

Drawing on the possible worlds semantics that modal logic provides (Section \ref{app:sec:pw}), we can define specific possible worlds semantics for our logic for expressing the actions of the demon in EHPO from above. 

\begin{definition} \label{app:def:possibleworld}
    A \textbf{possible world} represents the set of logs $\mathcal{L}$ the demon has produced at time $\tau_\sigma(\mathcal{L})$ (i.e., after concluding running EHPO) and the set of formulas $\mathcal{P}$ that are modeled from $\mathcal{L}$ via $\mathcal{F}$.
\end{definition}

Therefore, different possible worlds represent the states that \emph{could have existed} if the evil demon had executed different strategies (Definition \ref{def:strategy}). In other words, if it had performed EHPO with different learning algorithms, different HPO procedures, different hyper-hyperparameter settings, different amounts of time (less than the total upper bound), different learning tasks, different models, etc... to produce a different set of logs $\mathcal{L}$ and corresponding set of conclusions $\mathcal{P}$. 

In this formulation, the demon has knowledge of all possible worlds; it is trying to fool us about the relative performance of algorithms by showing as an intentionally deceptive world. Informally, moving from world to world (via an accessibility relation) corresponds to the demon running more passes of HPO to produce more logs to include in $\mathcal{L}$.

\subsubsection{Syntax and Semantics for the Logic Modeling the Demon Running the EHPO} \label{app:sec:ehpoproperties}

We provide proofs and intuitions of the axioms of our EHPO logic in this section, based on a correspondence with un-indexed modal logic.

We remind the reader that the following are the axioms of our indexed modal logic:

\begin{align*}
\vdash (p \rightarrow q) \rightarrow (\possible_t p \rightarrow \possible_t q) && \textit{(necessitation + distribution)} \\
p \rightarrow \possible_t p && \textit{(reflexivity)} \\
\possible_t\possible_s p \rightarrow \possible_{t+s} p && \textit{(transitivity)}
\\\possible_s\necessary_t p \rightarrow \necessary_t p && \textit{(symmetry)},
\\\possible_t(p \land q) \rightarrow (\possible_t p \land \possible_t q) && \textit{(distribution over } \land \textit{)}
\end{align*}
In short, to summarize these semantics---the demon has knowledge of all possible hyper-hyperparameters, and it can pick whichever ones it wants to run EHPO within a bounded time budget $t$ to realize the outcomes it wants: $\possible_t p$ means it can realize $p$.

We inherit distribution and necessitation from un-indexed modal logic; they are axiomatic based on Kripke semantics. We provide greater intuition and proofs below.\\

\textbf{Notes on necessitation for $\necessary_t$}:\\

Necessitation for our indexed necessary operator can be written as follows:
\begin{align*}
    \vdash p \rightarrow \necessary_i p
\end{align*}

As we note in Section \ref{sec:hpo:formalizing}, $\vdash$ just means here that $p$ is a theorem of propositional logic. So, if $p$ is a theorem, then so is $\necessary_t p$. By theorem we just mean that $p$ is provable by our axioms (these being the only assumptions we can use); so whenever $p$ fits this definition, we can say $\necessary_t p$.

For our semantics, this just means that when $p$ is a theorem, it is necessary that $p$ in time $t$.\\ 

\textbf{Distribution for $\necessary_t$}:\\
\begin{align*}
\necessary_t (p \rightarrow q) \rightarrow (\necessary_t p \rightarrow \necessary_t q) \end{align*}

We provide three ways to verify distribution over implication for $\necessary_t$. From this, we will prove distribution over implication for $\possible_t$

\begin{enumerate}[label=\Alph*.]
\item The first follows from an argument about the semantics of possible worlds from the Kripke model of our system (Sections \ref{app:sec:pw} \& \ref{app:sec:pw_interpretation}). 
    \begin{enumerate}[label=\roman*.]
        \item It is fair to reason that distribution is self-evident given the definitions of implication ($\rightarrow$, formed from $\lnot$ and $\lor$ in our syntax for well-formed formulas for our EPHO logic, given at (\ref{app:eq:ehposyntax}) and necessity ($\necessary_t$, formed from $\lnot$ and $\possible_t$ in our syntax for well-formed formulas for our EHPO logic, given at (\ref{app:eq:ehposyntax})).
        \item Similarly, we can further support this via our semantics of possible worlds. 
        
        We can understand $\necessary_t p$ to mean, informally, that it an adversary does adopt a strategy $\sigma$ that is guaranteed to cause the desired conclusion $p$ to be the case while take at most time $t$ in expectation. Formally, as an ``necessary'' analog to the semantics of $\possible_t$ given in Definition \ref{def:ehpologic}:
        
        For any formula $p$, we say $\mathcal{L} \models \necessary_t p$ if and only if
        \[
        \forall \sigma \in \Sigma, \; \mathbb{P}(\sigma[\mathcal{L}] \models p) = 1 \; \land \; \mathbb{E}[\tau_\sigma(\mathcal{L})] \le t.
        \]

        Given $p \rightarrow q$ is true in \textbf{all accessible worlds} (i.e, the definition of necessary), then we can say that $q$ is true in all accessible worlds whenever $p$ is true in all accessible worlds. As in i. above, this just follows / is axiomatic from the definitions of necessity and implication for Kripke semantics. 
    \end{enumerate}
\item We can also prove distribution by contradiction.
    \begin{enumerate}[label=\roman*.]
        \item Suppose that the distribution axiom does not hold. That is, the hypothesis
        \begin{align*}
            \necessary_t(p \rightarrow q)
        \end{align*}
        is true and the conclusion
        \begin{align*}
            \necessary_t p \rightarrow \necessary_t q
        \end{align*}
        is false. 
        \item By similar reasoning, from above $\necessary_t p \rightarrow \necessary_t q$ being false, we can say that $\necessary_t p$ is true and $\necessary_t q$ is false. 
        \item We can use Modal Axiom M (reflexivity, proven in the next section) to say $\necessary_t p \rightarrow p$. Since $\necessary_t p$ is true, we can use \emph{modus ponens} to determine that $p$ is true.
        \item We can also say
        \begin{align*}
            \necessary_t(p \rightarrow q) \rightarrow (p \rightarrow q) && (\textit{By Modal Axiom M (reflexivity)})
        \end{align*}
        \item Since we $\necessary_t(p \rightarrow q)$ is true from above, we can conclude via \emph{modus ponens} that $p \rightarrow q$ must also be true. 
        \item We concluded above that $p$ is true, so we can again use \emph{modus ponens} and the fact that $p \rightarrow q$ is true to conclude that $q$ is true.
        \item By necessitation, we can then also say $q \rightarrow \necessary_t q$, and conclude that $\necessary_t q$ is true. This is a contradiction, as above we said that $\necessary_t$ is false. 
        \item Therefore, by contradiction, $\necessary_t(p \rightarrow q) \rightarrow (\necessary_t p \rightarrow \necessary_t q)$ is proved. 
        
    \end{enumerate}
\item We can separately take an intuitionistic approach to verify the distribution axiom \cite{bezhanishvili2009intuitionistic, vanatten2004brouwer}:
    \begin{enumerate}[label=\roman*.] 
        \item Let $b$ be an \textbf{actual proof} of $p \rightarrow q$ so that we can say $a.b$ is a proof of $\necessary_t(p \rightarrow q$).
        \item Let $d$ be an \textbf{actual proof} of $p$ so that we can say $c.d$ is a proof of $\necessary_t p$. 
        \item From i. and ii., $b(d)$ is an \textbf{actual proof} of $q$, i.e. $b$ (an actual proof of $p \rightarrow q$) is supplied $d$ (an actual proof of $p$), and therefore can conclude $q$ via an actual proof. 
        \item From iii., we can say this results in a proof of $\necessary_t q$, i.e. $e.[b(d)]$.
        \item The above i.-iv. describes a function, $f: a.b \rightarrow f_{(a.b)}$. In other words, given \textbf{any proof} $a.b$ (i.e., of $\necessary_t(p \rightarrow q)$) we can return function $f_{(a.b)}$, which turns \textbf{any proof} $c.d$ (i.e., of $\necessary_t p$) into a proof $e.[b(d)]$ (i.e., of $\necessary_t q$).
        \item $f_{(a.b)}$ is thus a proof of $\necessary_t p \rightarrow \necessary_t q$.
        \item From i.-vi., we gone from $a.b$ (a proof of $\necessary_t(p \rightarrow q)$) to a proof of $\necessary_t p \rightarrow \necessary_t q$, i.e. have intuitionistically shown that $\necessary_t(p \rightarrow q) \rightarrow (\necessary_t p \rightarrow \necessary_t q)$
    \end{enumerate}
\end{enumerate}

\textbf{Distribution and $\possible_t$}:
We provide the following axiom in our logic:
\begin{align*}
\vdash (p \rightarrow q) \rightarrow (\possible_t p \rightarrow \possible_t q) && (\textit{necessitation and distribution})
\end{align*}

and we now demonstrate it to be valid. 
\begin{align*}
    \vdash (p \rightarrow q) \rightarrow \necessary_t(p \rightarrow q) && (\textit{necessitation})
    \\\rightarrow \;
    \necessary_t (\lnot q \rightarrow \lnot p) && (\textit{modus tollens})
    \\\rightarrow \;
    (\necessary_t \lnot q \rightarrow \necessary_t  \lnot p) && (\textit{distribution})
    \\\rightarrow \;
    (\lnot \necessary_t \lnot p \rightarrow \lnot \necessary_t  \lnot q) && (\textit{modus tollens})
    \\\rightarrow \;
    (\possible_t p \rightarrow \possible_t  q) && (\possible_t a \equiv \lnot \necessary_t \lnot a)
\end{align*}

This concludes our proof, for how the axioms are jointly stated. 

Further, we could also say
\begin{align*}
    (p \rightarrow q) \rightarrow \possible_t (p \rightarrow q) && (\textit{Modal axiom M (reflexivity)})
\end{align*}

And therefore also derive distribution over implication for possibility: 

\begin{align*}
    \possible_t(p \rightarrow q) \rightarrow (\possible_t p \rightarrow \possible_t q)
\end{align*}

\textbf{Modal Axiom M: Reflexivity}
\begin{align*}
    p \rightarrow \possible_t p
\end{align*}

This axiom follows from how we have defined the semantics of our indexed modal logic (Definition \ref{def:ehpologic}). It follows from the fact that the demon could choose to do nothing.

We can provide a bit more color to the above as follows:

We can also derive this rule from necessitation, defined above (and from the general intuition / semantics of modal logic that necessity implies possibility). First, we can say that necessity implies possibility. We can see this a) from a possible worlds perspective and b) directly from our axioms. From a possible worlds perspective, this follows from the definition of the operators. Necessity means that there is truth at every accessible possible world, while possibility means there is truth at some accessible possible world, which puts that possible truth in time $t$ as a subset of necessary truth in time $t$. From the axioms, we verify
\begin{align*}
    \necessary_t p \rightarrow \possible_t p  & & (\textit{Theorem to verify, which also corresponds to Modal Axiom D (serial)}) \\
    \lnot \necessary_t p \lor \possible_t p && (\textit{Applying } p \rightarrow q \textit{ is equiv. }\lnot p \lor q) \\
    \possible_t \lnot p \lor \possible_t p && (\textit{By modal conversion, } \lnot \necessary_t p \rightarrow \possible_t \lnot p) \\
    & & (\textit{Which for our semantics is tautological}) \\
\end{align*}

That is, in time $t$ it is possible that $p$ or it is possible that $p$, which allows for us also to not conclude anything (in the case that the demon chooses to do nothing).

We can then say,
\begin{align*}
    (\necessary_t p \rightarrow p)  \rightarrow \possible_t p && (\textit{By necessitation and } \necessary_t p \rightarrow \possible_t p ) \\
    p \rightarrow \possible_t p && (\textit{By concluding } p \textit{ from necessitation})
\end{align*}

Another way to understand this axiom is again in terms of possible worlds. We can say in our system that every world is possible in relation to itself. This corresponds to the accessibility relation $\mathcal{R}ww$. As such, an equivalent way to model reflexivity is in terms of the following:
\begin{align*}
    \necessary_t p \rightarrow p
\end{align*}

That is, if $\necessary_t p$ holds in world $w$, then $p$ also holds in world $w$, as is the case for $\mathcal{R}ww$. We can see this by proving $\necessary_t p \rightarrow p$ by contradiction. Assuming this were false, we would need to construct a world $w$ in which $\necessary_t p$ is true and $p$ is false. If $\necessary_t p$ is true at world $w$, then by definition $p$ is true at every world that $w$ accesses. For our purposes, this holds, as $\necessary_t p$ means that it is necessary for $p$ to be the case in time $t$; any world that we access from this world $w$ (i.e. by say increasing time, running more HPO) would require $p$ to hold. Since $\mathcal{R}ww$ means that $w$ accesses itself, that means that $p$ must also be true at $w$, yielding the contradiction.\\

\textbf{Modal Axiom 4: Transitivity}
\begin{align} \label{app:ax:a4}
    \possible_t\possible_s p \rightarrow \possible_{t+s} p
\end{align}

We can similarly understand transitivity to be valid intuitively from the behavior of the demon and in relation to the semantics of our possible worlds. We do an abbreviated treatment (in relation to what we say for reflexivity above) for brevity.

In terms of the demon, we note that in our semantics $\possible_t p$ means that it is possible for the demon to bring about conclusion $p$ via its choices in time $t$. Similarly, we could say the same for $\possible_s p$; this means it is possible for the demon to bring about conclusion $p$ in time $s$. If it is possible in time $t$ that it is possible in time $s$ to bring about $p$, this is equivalent in our semantics to saying that it is possible in time $t+s$ to bring about conclusion $p$.

We can understand this rule (perhaps more clearly) in terms of possible worlds and accessibility relations.

For worlds $w_n$,
\begin{align*}
    \forall w_{1}, \forall w_{2}, \forall w_{3}, \mathcal{R}w_{1}w_{2} \land \mathcal{R}w_{2}w_{3}\rightarrow \mathcal{R}w_{1}w_{3}
\end{align*}

In other words, this accessibility relation indicates that if $w_1$ accesses $w_2$ and if $w_2$ accesses $w_3$, then $w_1$ accesses $w_3$.

For understanding this in terms of relative possibility, we could frame this as, if $w_3$ is possible relative to $w_2$ and if $w_2$ is possible relative to $w_1$, then $w_3$ is possible relative to $w_1$. For our semantics of the demon, this means that in some time if in some time $b$ we can get to some possible world $w_3$ from when we're in $w_2$ and in time $a$ we can get to some possible world $w_2$ when we're in $w_1$, then in time $a+b$ we can get to $w_3$ from $w_1$

This axiom is akin to us regarding a string of exclusively possible or exclusively necessary modal operators as just one possible or necessary modal operator, respectively; we regard then regard sum of times as the amount of time it takes to bring about $p$ (again, being necessary or possible, respectively).\\

\textbf{Modal Axiom 5: Symmetry}
\begin{align} \label{app:ax:a5}
    \possible_s\necessary_t p \rightarrow \necessary_t p
\end{align}

We can similarly understand that our modal logic is symmetric; this is valid intuitively from the behavior of the demon. We further abbreviate our treatment for brevity. In terms of the demon, we note that in our semantics $\possible_s p$ means that it is possible for the demon to bring about conclusion $p$ via its choices in time $s$. We can also say $\necessary_t p$ means that it is necessary for the demon to bring about $p$ in time $t$. If it is possible in time $s$ that it is necessary in time $s$ to bring about $p$, this is equivalent in our semantics to saying that it is necessary in time $t$ to bring about conclusion $p$. In other words, we can disregard would could have possibly happened in time $s$ from the demon's behavior and only regard what was necessary in time $t$ for the demon to do in order to bring about $p$.

As another example, consider our reduction of $\possible_t \lnot \possible_t p$ to $\lnot \possible_t p$ in our proof for deriving a defended reasoner in Section \ref{sec:hpo:defense}. While the intuitive English reading (``It's possible that it's not possible that $p$") does not seem equivalent to this reduction (``It's not possible that $p$), it is in fact valid for our semantics. Think of this in terms of the demon. If $p$ cannot be brought about in time $t$ in expectation (where $t$ is a reasonable upper bound on compute time), then that's the end of it; it doesn't matter which operators come before it (any number of $\possible_t$ or $\necessary_t$). Adding possibility or necessity before that condition doesn't change that fact that it, for that upper bound $t$, it is not possible to bring about $p$.

This axiom is akin to us just regarding the rightmost modal operator when we have a mix of modal operators applied iteratively; we can disregard what was possible or necessary in the time prior to the rightmost operator, and say that what we can say about a sentence $p$ (whether it is possible or necessary) just relates to how much time the last operator required to bring about $p$.\\

\textbf{Derived axioms}\\

We can similarly derive other axioms of our indexed modal logic, form the axioms above. Notably,

\textbf{$\necessary_t$ distributes over $\land$}

\begin{align*}
    \necessary_t(p \land q) \rightarrow (\necessary_t p \land \necessary_t q) & & (\textit{$\necessary_t$ distributes over $\land$}) \\
    \textbf{Inner proof 1} \\
    p \land q \\
    p \\
    (p \land q) \rightarrow p \\
    \necessary_t ((p \land q) \rightarrow p) && (\textit{Necessitation}) \\
    \necessary_t (p \land q) \rightarrow \necessary_t p && (\textit{Distribution}) \\
    \necessary_t p  && (\textit{By assuming the hypothesis}) \\
    \textbf{Inner proof 2} \\
    p \land q \\
    q \\
    (p \land q) \rightarrow q \\
    \necessary_t ((p \land q) \rightarrow q) && (\textit{Necessitation}) \\
    \necessary_t (p \land q) \rightarrow \necessary_t q && (\textit{Distribution}) \\
    \necessary_t q  && (\textit{By assuming the hypothesis}) \\
    \necessary_t p \land \necessary_t q && (\textit{By inner proof 1, inner proof 2, assuming the hypothesis})
\end{align*}

\textbf{We can show a similar result for $\possible_t$ and $\land$, omitted for brevity.}

\textbf{$\possible_t$ distributes over $\lor$}

\begin{align*}
    \possible_t(p \lor q) \rightarrow (\possible_t p \lor \possible_t q) & & (\textit{$\possible$ distributes over $\lor$}) \\
    \lnot \necessary_t \lnot (p \lor q) \rightarrow (\possible_t p \lor \possible_t q) & & (\possible_t a \equiv \lnot \necessary_t \lnot a) \\
    \lnot \necessary_t (\lnot p \land \lnot q) \rightarrow (\possible_t p \lor \possible_t q) & & (\lnot(a \lor b) \equiv (\lnot a \land \lnot b)) \\
    \lnot (\necessary_t  \lnot p \land \necessary_t  \lnot q) \rightarrow (\possible_t p \lor \possible_t q) & & (\textit{$\necessary_t$ distributes over $\land$}) \\
    (\lnot \necessary_t  \lnot p \lor \lnot \necessary_t  \lnot q) \rightarrow (\possible_t p \lor \possible_t q) & & (\lnot(a \land b) \equiv (\lnot a \lor \lnot b)) \\
    (\possible_t p \lor \possible_t q) \rightarrow (\possible_t p \lor \possible_t q) & & (\possible_t a \equiv \lnot \necessary_t \lnot a) \\
\end{align*}

\textbf{We can show a similar result for $\necessary_t$ and $\lor$, omitted for brevity.}

\subsubsection{Syntax and Semantics for the Logic of our Belief in EHPO Conclusions} \label{app:sec:belief} 
The logic of belief is a type of modal logic, called doxastic logic \cite{hintikka1962doxastic}, where the modal operator $\mathcal{B}$ is used to express belief\footnote{Computer scientists do not tend to distinguish between the logic of knowledge (epistemic) and the logic of belief (doxastic)~\cite{segerberg1999logicofbelief}.} Different types of reasoners can be defined using axioms that involve $\mathcal{B}$ \cite{smullyan1986belief}.

We can formulate the doxastic logic of belief in Backus-Naur form:

For any atomic proposition $P$, we define recursively a well-formed formula $\phi$ as
\begin{align} \label{app:eq:beliefsyntax}
    \phi \coloneqq P\;|\;\lnot \phi\;|\;\phi \land \phi\;|\; \mathcal{B}\phi
\end{align}

where $\mathcal{B}$ means ``It is believed that $\phi$". We interpret this recursively where $p$ is the base case, meaning that $\phi$ is p if it is an atom, $\lnot \phi$ is well-formed if $\phi$ is well-formed. We can also define $\lor$, $\rightarrow$, and $\leftrightarrow$ from $\lnot$ and $\land$, as in propositional logic.

As stated in the belief logic portion of Section \ref{sec:hpo:formalizing}, we model a consistent Type 1 reasoner \cite{smullyan1986belief}, which has access to all of propositional logic, has their beliefs logical closed under \emph{modus ponens}, and does not derive contradictions. In other words, we have the following axioms:

\begin{align*}
    \lnot ( \mathcal{B}p \land \mathcal{B}\lnot p )\equiv \mathcal{B}p \rightarrow \lnot \mathcal{B} \lnot p
\end{align*}
which is the consistency axiom,
\begin{align*}
    \vdash p \rightarrow \mathcal{B} p
\end{align*}
which is akin to Necessitation above in Section \ref{app:sec:kripkeaxioms} and means that we believe all tautologies, and
\begin{align*}
    \mathcal{B}(p \rightarrow q) \rightarrow (\mathcal{B}p \rightarrow \mathcal{B}q)
\end{align*}
which means that belief distributes over implication. This notably does not include

\begin{align*}
    \mathcal{B}p \rightarrow p
\end{align*}

which essentially means that we do not allow for believing $p$ to entail concluding $p$. This corresponds to us actually wanting to run hyperparameter optimization before we conclude anything to be true. We do not just want to conclude something to be true based only on \emph{a priori} information. This is akin to picking folkore parameters and concluding they are optimal without running hyperparameter optimization. 

\subsubsection{Combining Logics} \label{app:sec:multimodal}
It is a well known result that we can combine modal logics to make a multimodal logic \cite{scott1970multimodal}. In particular, we refer the reader to results on \emph{fusion} \cite{thomason1984fusion}.

For a brief intuition, we are able to combine our EHPO logic with belief logic since we are operating over the same set of possible worlds. The results of running EHPO produce a particular possible world, to which we apply our logic of belief in order to reason about the conclusions drawn in that world.

\subsubsection{Our Combined, Multimodal Logic and Expressing Hyperparameter Deception}
We develop the following multimodal logic, which we also state in Section \ref{sec:hpo:formalizing}:
\[
    \psi \coloneqq P\;|\;\lnot \psi\;|\;\psi \land \psi \;|\; \possible_t \psi \;|\; \mathcal{B} \psi
\]

\subsubsection{Axioms} \label{app:sec:multimodalaxioms}

We give this multimodal logic semantics to express our $t$-non-deceptiveness axiom, which we repeat below for completeness:

For any formula $p$,
\begin{equation*}
    \lnot \left( \possible_t \mathcal{B} p \land \possible_t \mathcal{B} \lnot p \right)
\end{equation*}

We can similarly express the $t$-non-deceptiveness axiom:

For any formula $p$,
\begin{equation*}
    \possible_t \mathcal{B}p \rightarrow \lnot \possible_t \mathcal{B} \lnot p 
\end{equation*}

We can also express a $t$-deceptiveness-axiom:

For any formula p,
\begin{equation*}
    \possible_t \mathcal{B} p \land \possible_t \mathcal{B} \lnot p
\end{equation*}

To reiterate, \emph{multimodal} just means that we have multiple different modes of reasoning, in this case our $\possible_t$ semantics for the demon doing EHPO and our consistent Type 1 reasoner operator $\mathcal{B}$.

Given a reasonable maximum time budget $t$, we say that EHPO is $t$-non-deceptive if it satisfies all of axioms above. Moreover, based on this notion of $t$-non-deceptiveness, we can express what it means to have a defense to being deceived. 

\subsubsection{Some notes on strength of belief and belief update} \label{app:sec:beliefstrength}

There are potentially interesting connections between our work on defending against hyperparameter deception and belief update \cite{ferme2011agmreview}. Notably, one could view our notion of skeptical belief as related to work done on "strength of belief" and belief update, or dynamic doxastic logic \cite{segerberg1999logicofbelief, benthem2009dup, kooi2003pdel}. Instead of picking an EHPO runtime \emph{a priori} and then running a defended EHPO and at the end evaluating whether or not we believe the conclusions we draw, we could iteratively update and test our belief and terminate if a certain belief threshold is met. In such quantitative theories of belief change, the degree of acceptance of a sentence is represented by a numerical value. Those numerical values can be updated in light of new information (so-called ``soft" information updates) \cite{benthem2007delbr, baltag2008pdbr}. Exploring this is out of scope for our work here, but could be an interesting future research direction for how to reason about empirical results that imply inconsistent conclusions. 

%% file: section/99-appendix/21-hpo/50-app-hpo-defense.tex
\section{Section \ref{sec:hpo:defense} Appendix: Notes on Defenses} \label{app:sec:defense}

\subsection{Proving a defended reasoners}

Suppose that we have been drawing conclusions using some ``naive'' belief operator $\mathcal{B}_{\text{n}}$ (based on a conclusion function $\mathcal{F}_{\text{n}}$) that satisfies the axioms of Section~\ref{sec:belieflogic}. We want to use $\mathcal{B}_{\text{n}}$ to construct a new operator $\mathcal{B}_{*}$, which is guaranteed to be deception-free.
One straightforward way to do this is to define the belief operator $\mathcal{B}_{*}$ such that for any statement $p$,
\[
    \mathcal{B}_{*} p \; \equiv \; \mathcal{B}_{\text{n}} p \;\land\; \lnot \possible_t \mathcal{B}_{\text{n}} \lnot p.
\]
That is, we conclude $p$ only if both our naive reasoner would have concluded $p$, and it is impossible for an adversary to get it to conclude $\lnot p$ in time $t$.
This enables us to show $t$-non-deceptiveness, following directly from the axioms in a proof by contradiction: Suppose $\mathcal{B}_{*}$ can be deceived, i.e. $\textcolor{red}{\possible_t \mathcal{B}_{*} p} \land \textcolor{red}{\possible_t \mathcal{B}_{*} \lnot p}$ is $\mathsf{True}$:

\begingroup
\setlength{\tabcolsep}{3.pt} 
\renewcommand{\arraystretch}{1} 
\begin{table}[H]
\footnotesize
\begin{center}
      \centering
        \begin{tabular}{r c l c c}
\toprule
\; & \; & \; & \; & \textbf{Rule} \\
\midrule
$\textcolor{red}{\possible_t \mathcal{B}_{*} p}$ & $\equiv$ & $\possible_t \left( \mathcal{B}_{\text{n}} p \land \lnot \possible_t \mathcal{B}_{\text{n}} \lnot p \right)$   &  \; & \text{Applying $\possible_t$ to the definition of $\mathcal{B}_{*} p$ \; (\ref{ax:defendedreasoner})} \\

\midrule
\; & $\rightarrow$ & $\possible_t \left( \lnot \possible_t \mathcal{B}_{\text{n}} \lnot p \right)$  & \;   &  \text{Reducing a conjunction to either of its terms: $(a \land b) \rightarrow b$} \\

\midrule
\; & $\rightarrow$ & $\textcolor{blue}{\lnot \possible_t \mathcal{B}_{\text{n}} \lnot p}$  & \;   & \text{Symmetry; dropping all but the right-most operator: $\possible_t(\possible_t a)\rightarrow \possible_t a$} \\
\bottomrule
\end{tabular}
\end{center}
\end{table}

We provide more detail on these transformations than we do in the main text. The first application is simple; we just put parentheses around our definition of $\mathcal{B}_*$, and apply $\possible_t$ to it. The second step is also simple. We apply a change to whats inside the parentheses, i.e. just the definition of $\mathcal{B}_*$. Because this is a conjunction, in order for it to be true, both components have to be true. So, we can reduce the conjunction to just it's second term.

The part that is more unfamiliar is the application of Modal Axiom 5 (Symmetry) to reduce the number of $\possible_t$ operators. We provide this example above in Section \ref{app:sec:logic}, where we explain why Modal Axiom 5 holds for our EHPO logic semantics. We reiterate here for clarity:

While the intuitive English reading (``It's possible that it's not possible that $p$") does not seem equivalent to this reduction (``It's not possible that $p$), it is in fact valid for our semantics. Think of this in terms of the demon. If $p$ cannot be brought about in time $t$ in expectation (where $t$ is a reasonable upper bound on compute time), then that's the end of it; it doesn't matter which operators come before it (any number of $\possible_t$ or $\necessary_t$). Adding possibility or necessity before that condition doesn't change that fact that it, for that upper bound $t$, it is not possible to bring about $p$.

We then pause to apply our axioms to the right side of the conjunction, $\textcolor{red}{\possible_t \mathcal{B}_{*}\lnot p}$ :

\begingroup
\setlength{\tabcolsep}{3pt} 
\renewcommand{\arraystretch}{1} 
\begin{table}[H]
\small
\begin{center}
      \centering
        \begin{tabular}{r c l c c}
\toprule
\; & \; & \; & \; & \textbf{Rule} \\
\midrule
$\textcolor{red}{\possible_t \mathcal{B}_{*} \lnot p}$ & $\equiv$ & $\possible_t \left( \mathcal{B}_{\text{n}} \lnot p \land \lnot \possible_t \mathcal{B}_{\text{n}} p \right)$   &  \; & \text{Applying $\possible_t$ to the definition of $\mathcal{B}_{*} \lnot p$ \; (\hyperlink{ax:defendedreasoner}{1})} \\

\midrule
\; & $\rightarrow$ & $\possible_t\mathcal{B}_{\text{n}} \lnot p \land \possible_t\lnot \possible_t \mathcal{B}_{\text{n}} p$  & \;   &  \text{Distributing $\possible_t$ over $\land$: $\possible_t (a \land b) \rightarrow (\possible_t a \land \possible_t b)$} \\

\midrule
\; & $\rightarrow$ & $\textcolor{blue}{\possible_t \mathcal{B}_{\text{n}} \lnot p}$  & \;   & \text{Reducing a conjunction to either of its terms: $(a \land b) \rightarrow a$} \\
\bottomrule
\end{tabular}
\end{center}
\end{table}

This transformation is much like the one above. We similarly apply $\possible_t$ to the definition of $\mathcal{B}_*$.

We then distribute $\possible_t$ over the definition, which holds for our logic since possibility distributes over and. We prove this for our logic in Section \ref{app:sec:logic}, and provide an intuitive explanation here. If it is possible in time $t$ to bring about a particular formula, then it must also be possible to bring about the sub-conditions that compose that formula in time $t$. If this were not the case, then we would not be able to satisfy bringing about the whole formula in time $t$.

Lastly, as in the first example, we reduce the conjunction to one of its terms (this time taking the first, rather than the second).

We now bring both sides of the conjunction back together: $\textcolor{red}{\possible_t \mathcal{B}_{*} p} \land  \textcolor{red}{\possible_t \mathcal{B}_{*} \lnot p}
\; \equiv \;
\textcolor{blue}{\lnot \possible_t \mathcal{B}_{\text{n}} \lnot p} \land  \textcolor{blue}{\possible_t \mathcal{B}_{\text{n}} \lnot p}$.
The \textcolor{blue}{right-hand side} is of the form $\lnot a \land a$, which must be $\mathsf{False}$. This contradicts our initial assumption that $\mathcal{B}_{*}$ is $t$-deceptive (i.e., $\textcolor{red}{\possible_t \mathcal{B}_{*} p} \land \textcolor{red}{\possible_t \mathcal{B}_{*} \lnot p}$ is $\mathsf{True}$). Therefore, $\mathcal{B}_{*}$ is $t$-non-deceptive. 

\subsection{Theoretically Validating Defenses to Hyperparameter Deception}

We prove Theorem \ref{thm:defendedhpo}:

\begin{theorem}
Suppose that the set of allowable hyper-HPs $\mathcal{C}$ of $H$ is constrained, such that any two allowable random-search distributions $\mu$ and $\nu$ have Renyi-$\infty$-divergence at most a constant, i.e. $D_{\infty}(\mu \| \nu) \le \gamma$. The $(K,R)$-defended random-search EHPO of Definition \ref{def:defended_random} is guaranteed to be $t$-non-deceptive if we set $R \ge \sqrt{t \exp(\gamma K)/K} = O(\sqrt{t})$.
\end{theorem}

Suppose we are considering HPO via random search \cite{bergstra2011algorithms}, in which the set of allowable hyper-hyperparameters contains tuples $(\mu, M)$, where $\mu$ is a distribution over all possible hyperparameter sets $\Lambda$ and $M$ is the number of different hyperparameter configuration trials to run. This set $S$ is the Cartesian product of the set of allowable distributions $D$ ($\mu \in D$) and $M$.

Suppose that for any two allowable distributions $\mu$, $\nu \in D$ and any event $A$ (a measurable subset of $\Lambda$), $\mu(A) \le e^\gamma \cdot \nu(A)$ (i.e., the Renyi $\infty$-divergence between any pair of distributions is at most $\gamma$).
This bounds how much the choice of hyper-hyperparameter can affect the hyperparameters in HPO.

We also suppose we start from a naive reasoner (expressed via the operator $\mathcal{B}_{\text{n}}$), which draws conclusions based on a log with $K$ trials. For this scenario, we are only concerned with the reasoner's conclusions from $K$-trial logs.  We therefore assume w.l.o.g. that the reasoner draws no conclusions unless presented with exactly one log with exactly $K$ trials.

For some constant $R \in \mathbb{N}$, we
construct a new reasoner $\mathcal{B}_{*}$ that does the following:
It draws conclusions only from a single log with exactly $KR$ trials (otherwise it concludes nothing).
To evaluate a proposition $p$, it splits the log into $R$ groups of $K$ trials each, evaluates $\mathcal{B}_{n}$ on $p$ on each of those $R$ groups, and then concludes $p$ only if $\mathcal{B}_{n}$ also concluded $p$ on all $R$ groups.

Now consider a particular (arbitrary) proposition $p$.
Since $\mathcal{B}_{*}$ draws conclusions based on only a single log, any strategy $\sigma$ for the demon is equivalent to one that maintains at most one log at all times (the ``best'' log it found so far for its purposes, as it can discard the rest).

Let $Q$ be the supremum, taken over all allowable distributions $\mu$, of the probability that running a group of $K$ random search trials using that distribution will result in a log that would convince the $\mathcal{B}_{n}$ of $p$. Similarly, let $Q_{\lnot}$ be the supremum, taken over all allowable distributions $\nu$, of the probability that running a group of $K$ trials using that distribution will result in a log that would convince $\mathcal{B}_{n}$ of $\lnot p$.

Observe that $Q$ is the probability of an event in a product distribution of $K$ independent random variables each distributed according to $\mu$, and similarly for $Q_{\lnot}$, and the corresponding events are disjoint. By independent additivity of the Renyi divergence, the Renyi $\infty$-divergence between these corresponding product measures will be $\gamma K$.
It follows that 
\[
1 - Q \ge \exp(-\gamma K) Q_{\lnot}
\]

and
\[
1 - Q_{\lnot} \ge \exp(-\gamma K)  Q
\]

From here it's fairly easy to conclude that
\[
    Q + Q_{\lnot} \le \frac{2}{1 + \exp(-\gamma K) }.
\]

Now, an EHPO procedure using random search with $KR$ trials will convince $\mathcal{B}_{*}$ of $p$ with probability $Q^R$, since all $R$ independently sampled groups of $K$ trials must ``hit'' and each hit happens with probability $Q$. Thus, the expected time it will take the fastest strategy to convince us of $p$ is $Q^{-R} \cdot KR$.
Similarly, the fastest strategy to convince us of $\lnot p$ takes expected time $Q_{\lnot}^{-R} \cdot KR$.

Suppose now, by way of contradiction, that the $t$-non-deceptiveness axiom is violated, and there are strategies that can achieve both of these in time at most $t$.
That is,
\[
    Q^{-R} \cdot KR \le t
    \hspace{2em}\text{and}\hspace{2em}
    Q_{\lnot}^{-R} \cdot KR \le t.
\]
From here, it's fairly easy to conclude that
\[
    Q + Q_{\lnot} \ge 2 \left( \frac{KR}{t} \right)^{1/R}.
\]
Combining this with our conclusion above gives
\[
    \left( \frac{KR}{t} \right)^{1/R} \le \frac{1}{1 + \exp(-\gamma K) }.
\]
It's clear that we can cause this to be violated by setting $R$ to be large enough.
Observe that
\[
    \frac{1}{1 + \exp(-\gamma K) } \le \exp(-\exp(-\gamma K)),
\]
so
\[
    \left( \frac{KR}{t} \right)^{1/R} \le \exp(-\exp(-\gamma K)).
\]
Taking the root of both sides gives
\[
    \left( \frac{KR}{t} \right)^{\frac{1}{R \exp(-\gamma K)}} \le \frac{1}{e}.
\]
To simplify this expression, let $\beta$ denote
\[
    \beta = R \exp(-\gamma K).
\]
So that
\[
    \left( \frac{\beta K}{t \exp(-\gamma K)} \right)^{1/\beta} \le \frac{1}{e}.
\]
Finally, we set $R$ such that
\[
    \beta = \sqrt{\frac{t \exp(-\gamma K)}{K}}.
\]
To give
\[
    \left( \frac{1}{\beta} \right)^{1/\beta} \le \frac{1}{e}.
\]
But this is impossible, as the minimum of $x^x$ occurs above $1/e$.
This setting of $R$ gives
\[
    R = \beta \exp(\gamma K) = \sqrt{\frac{t \exp(\gamma K)}{K}} = O(\sqrt{t}).
\]
This shows that, for this task, if we run our constructed EHPO with $R = O(\sqrt{t})$ assigned in this way, it will be guaranteed to be $t$-non-deceptive.

\subsection{Defense Experiments}\label{app:sec:experiments:defense}

In this section we provide more information about the implementation of a random-search-based defense to hyperparameter deception in Wilson et al.~\cite{wilson2017marginal}, which we discuss in Section \ref{sec:hpo:defense}. 

\subsubsection{Our Implemented Defense Algorithm} \label{app:sec:phased}

The defense we implement in our experiments is a bit different than what we describe in our theoretical results in Section \ref{sec:hpo:defense}. In particular, in practice it is easier to implement subsampling rather than resampling.

\textbf{Protocol of Selecting Hyper-HPs.}
As partially illustrated in Figure~\ref{fig:wilson_deception} and elaborated on in Appendix \ref{app:sec:hpo:prelim}, Wilson et al.~\cite{wilson2017marginal}'s choice of hyper-HPs does not capture the space where Adam effectively simulates Heavy Ball. In Wilson et al.~\cite{wilson2017marginal}i, Adam-specific HPs like numerical variable $\epsilon=10^{-8}$ \cite{kingma2014adam} are treated as constants, leading to a biased HP-search space.

In contrast, we select the hyper-HPs of $\epsilon$ following a dynamic searching protocol: 

Inspired by Choi et al.~\cite{choi2019empirical}, we start from a wide range $\epsilon\in[10^{-12}, 10^{12}]$ as a wide search space. We iteratively select powers-of-10 grids that are uniformly spaced in the logarithmic scale of the current range. For instance, the selected grids for the prior range would be $\{10^{-12}, 10^{-11}, \cdots, 10^{11}, 10^{12}\}$. We perform a single run on each grid selected, and shrink the range towards grids where the best performance are achieved. The shrinkage follows the policy of either $\times 10$ to the lower boundary or $\times 0.1$ to the upper boundary. For example, for the prior range, we found the best performance is achieved on grid $10^{11}$, so we multiply the lower boundary $10^{-12}$ with $10$ and shift the range to $\epsilon\in[10^{-11}, 10^{12}]$.
Our protocol terminates with $\epsilon\in[10^{10}, 10^{12}]$ as the final hyper-HPs that we use for our defended random search EHPO. 

\textbf{Scaling Learning Rate $\eta$.}
Note that directly applying the hyper-HP of $\epsilon\in[10^{10}, 10^{12}]$ to Adam would lead to extremely slow convergence, since essentially large $\epsilon$ indicates a small effective learning rate $\eta$. Similar to Choi et al.~\cite{choi2019empirical}, we explore a shifted hyper-HPs for the $\eta$, scaled proportionally with $\epsilon$. Specifically, note that a large $\epsilon$ would make the update of Adam approach the update rule in the Heavy Ball method; for any randomly selected $\epsilon\in[10^{10}, 10^{12}]$, we perform the random search of $\eta/\epsilon$ instead of $\eta$ itself in the space of $[0.5,0.7]$, which is the search space of HB's learning rate shown in Wilson et al.~\cite{wilson2017marginal}.

\subsubsection{Experimental setup}
We follow the setup from \cite{wilson2017marginal}, where the details are specified in Section~\ref{sec:wilson:setup}.

\subsubsection{Code}

The code for running these experiments can be found at \url{https://github.com/pasta41/deception}.

\subsubsection{Associated results and logs}
In line with our notion of a log, we provide heatmaps of our logs in Figures~\ref{fig:defended-ehpo-logs-adam-sgd}, \ref{fig:defended-ehpo-logs-adam-hb} and  that correspond with our results in Section \ref{sec:hpo:defense}. We note that the performance of Heavy Ball for some random seeds is very bad (e.g., $10\%$ test accuracy). The performance varies widely -- also nearing $92\%$ for different random seeds. We affirm that this is the search space that yields the best results for Heavy Ball ($~92\%$). 

The results for Heavy Ball exhibit large variance. This illustrates a strength of our defensed: it actually helps with robustness against potentially making the wrong conclusion about Heavy Ball's performance (more generally), due to not making conclusions off of a single result (and perhaps using a random seed for which performance is particularly bad). We make a different claim about relative algorithm performance than Wilson et al.~\cite{wilson2017marginal} about Heavy Ball (i.e., we do not claim that it is better than Adam); but we do not reach this conclusion for the wrong reason (i.e., that we got one bad Heavy Ball result for a particular random seed). 

\pagebreak

\begin{figure}[H]
    \centering
    \includegraphics[scale=.5]{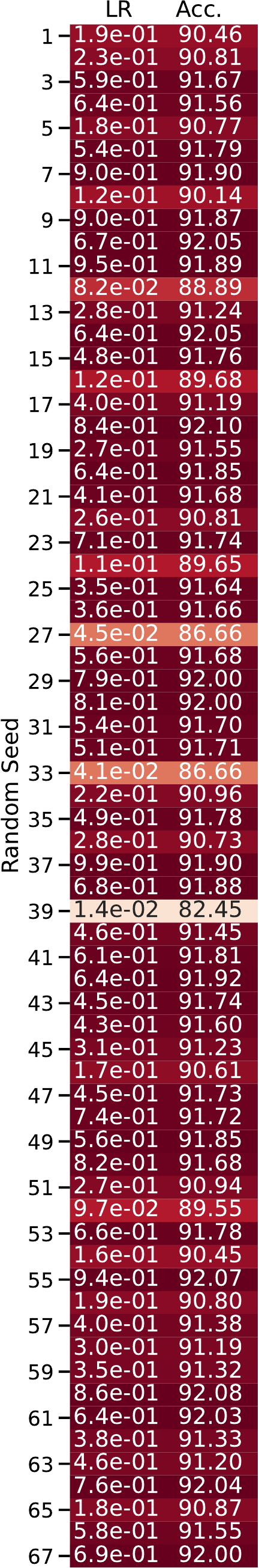} \hspace{2mm}
\includegraphics[scale=.5]{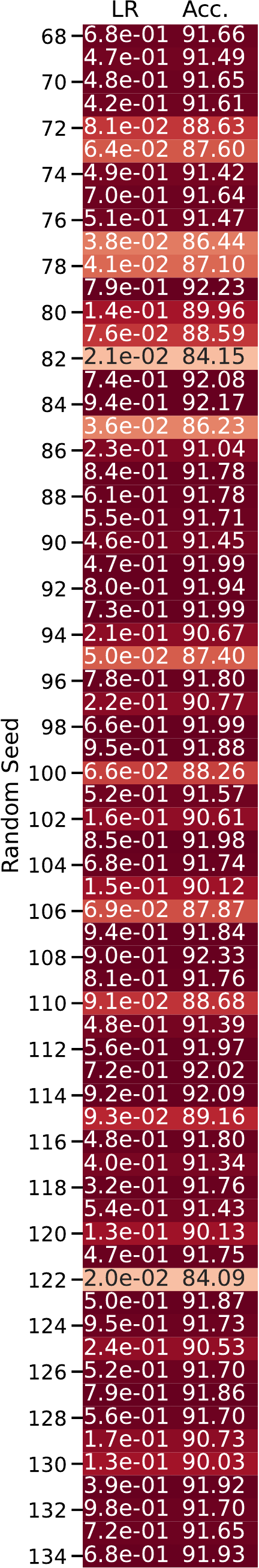} \hspace{2mm}
\includegraphics[scale=.5]{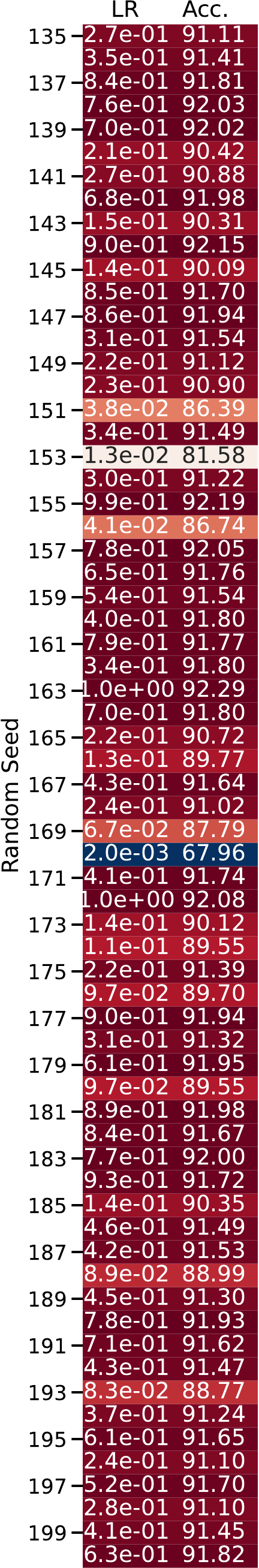} \hspace{2mm}
    \caption{Heatmap logs of SGD defended random search. Redder rows indicate higher test accuracy.}
    \label{fig:defended-ehpo-logs-sgd-only}
\end{figure}

\pagebreak

\begin{figure}[H]
    \centering
    \includegraphics[scale=.36]{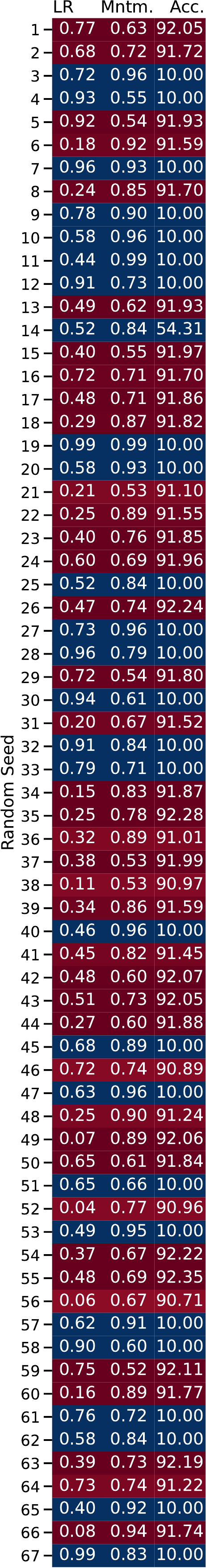} \hspace{2mm}
\includegraphics[scale=.36]{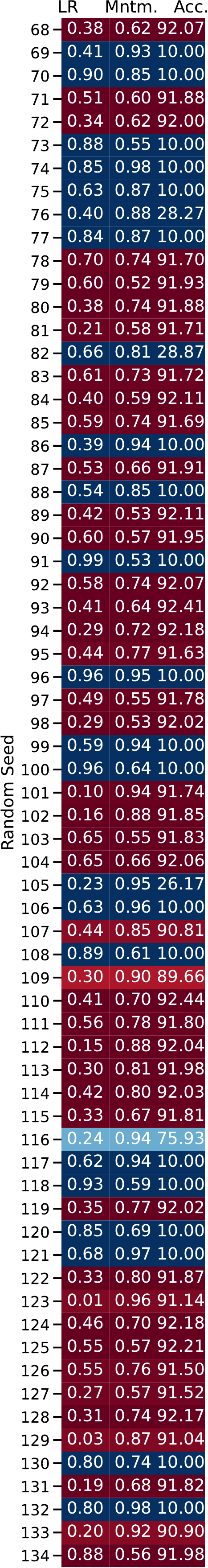} \hspace{2mm}
\includegraphics[scale=.36]{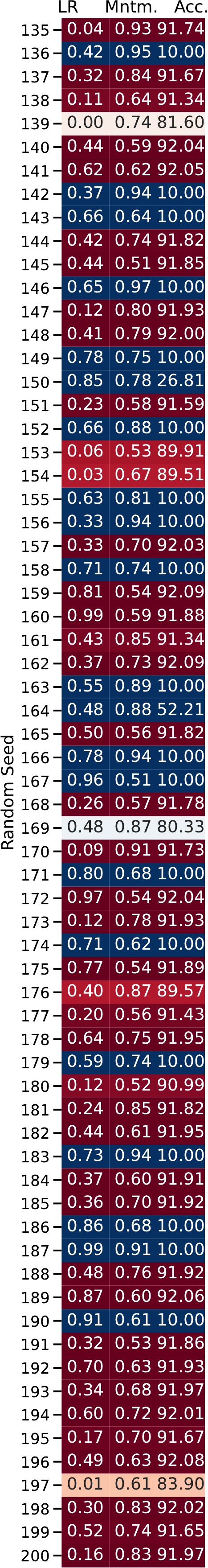} \hspace{2mm}
    \caption{Heatmap logs of Heavy Ball (HB) defended random search. Redder rows indicate higher test accuracy.}
    \label{fig:defended-ehpo-logs-hb-only}
\end{figure}

\pagebreak

\begin{figure}[H]
    \centering
    \includegraphics[scale=.35]{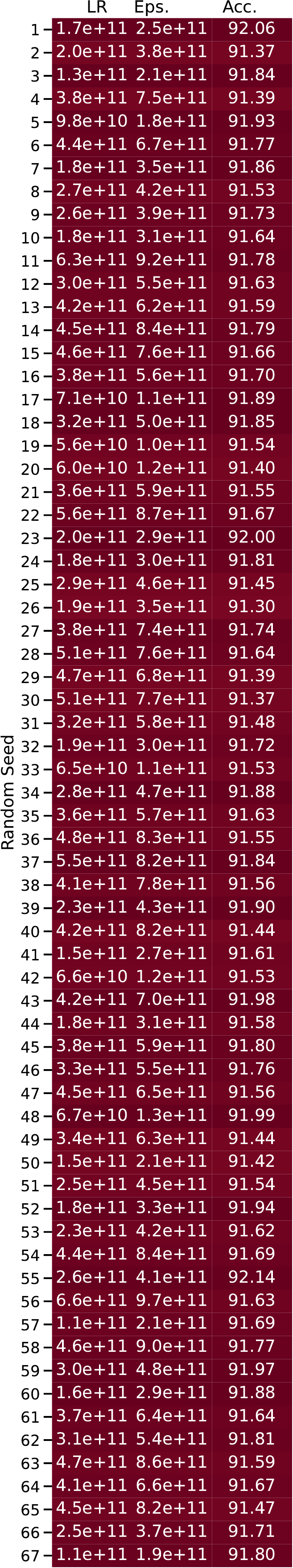} \hspace{2mm}
\includegraphics[scale=.35]{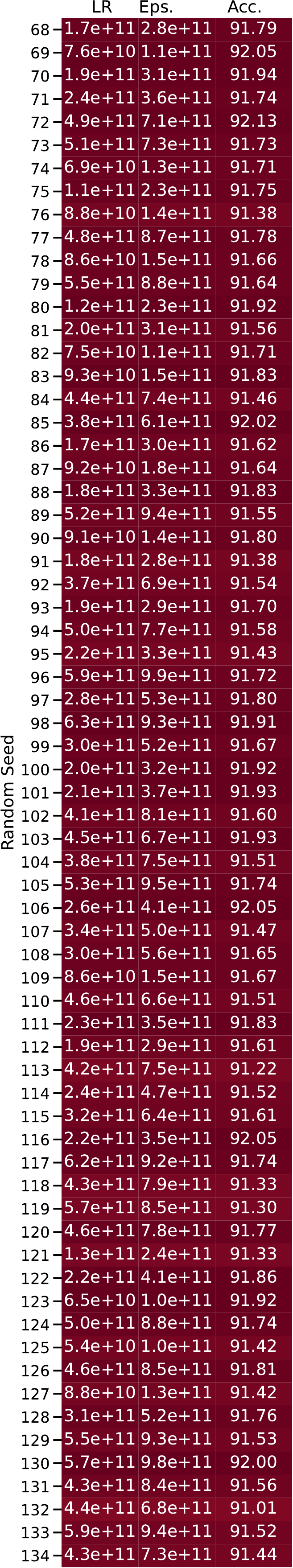} \hspace{2mm}
\includegraphics[scale=.35]{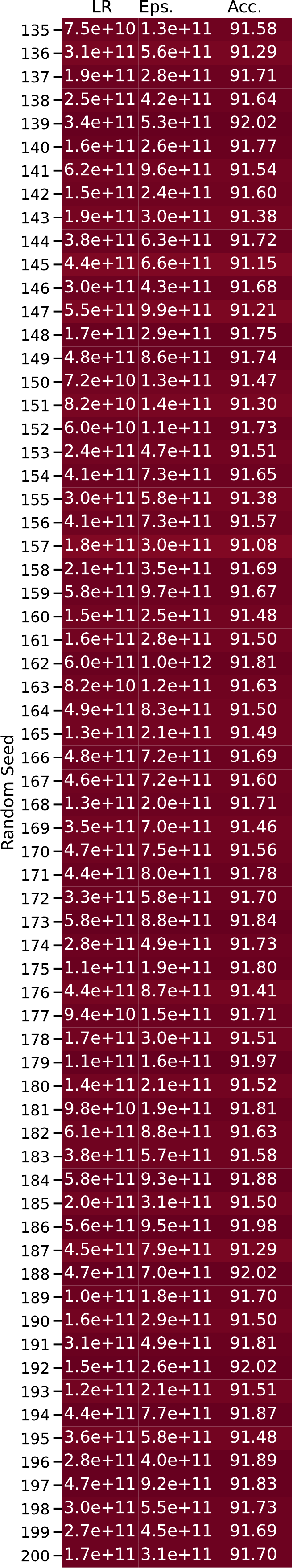} \hspace{2mm}
    \caption{Heatmap logs of Adam defended random search. Redder rows indicate higher test accuracy.}
    \label{fig:defended-ehpo-logs-adam-only}
\end{figure}

\pagebreak

\begin{figure}[ht]
    \centering
    \includegraphics[scale=.5]{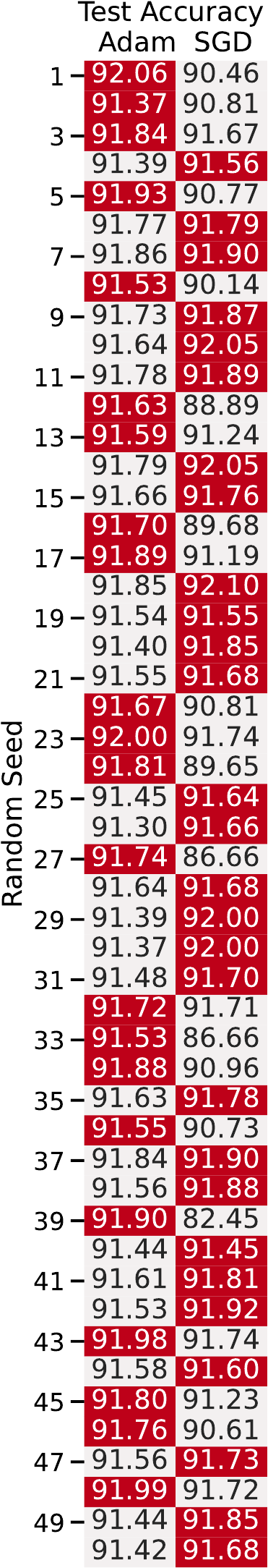} \hspace{2mm}
\includegraphics[scale=.5]{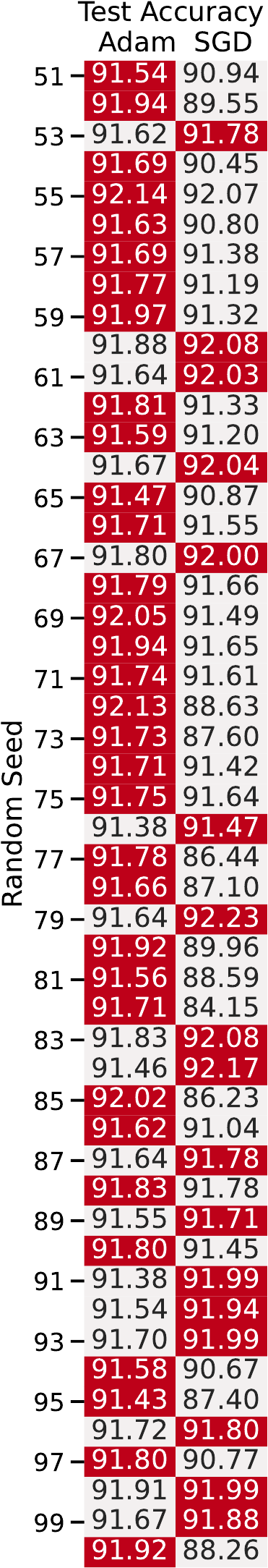} \hspace{2mm}
\includegraphics[scale=.5]{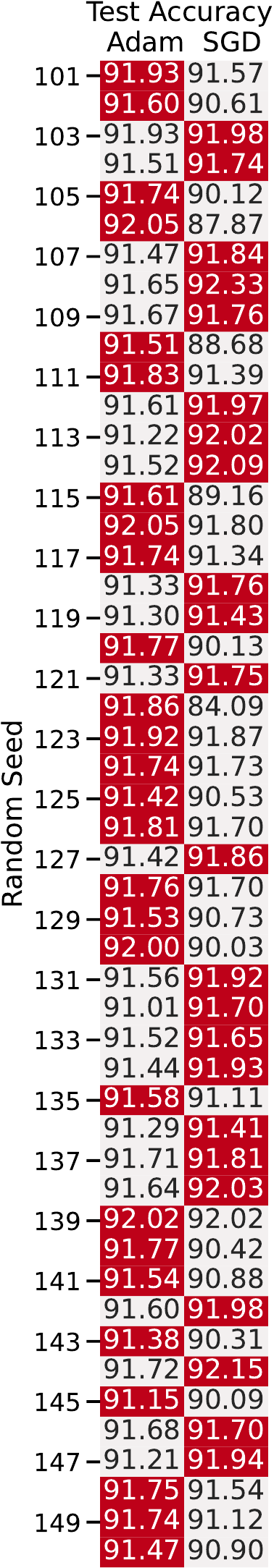} \hspace{2mm}
\includegraphics[scale=.5]{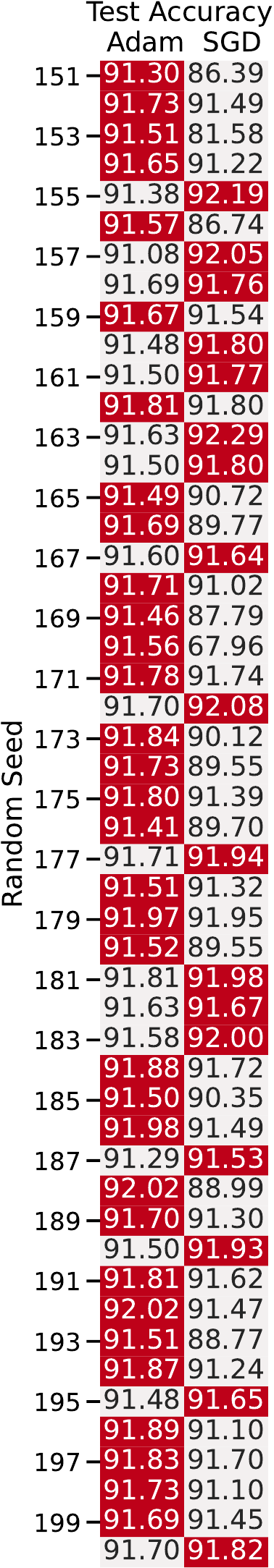} 
    \caption{Heatmap logs of test accuracy of VGG-16 on CIFAR-10 for Adam vs. SGD using our defended random search EHPO.  Red indicates higher test accuracy for the given random seed.}
    \label{fig:defended-ehpo-logs-adam-sgd}
\end{figure}

\pagebreak

\begin{figure}[ht]
    \centering
    \includegraphics[scale=.5]{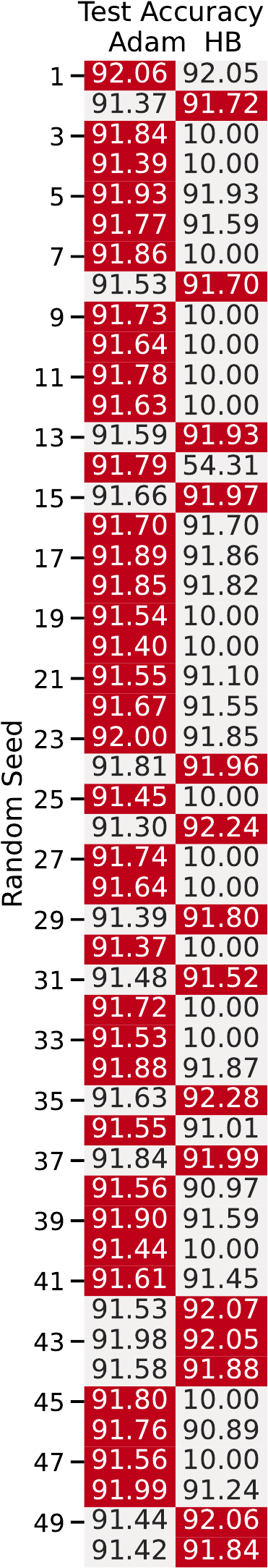} \hspace{2mm}
\includegraphics[scale=.5]{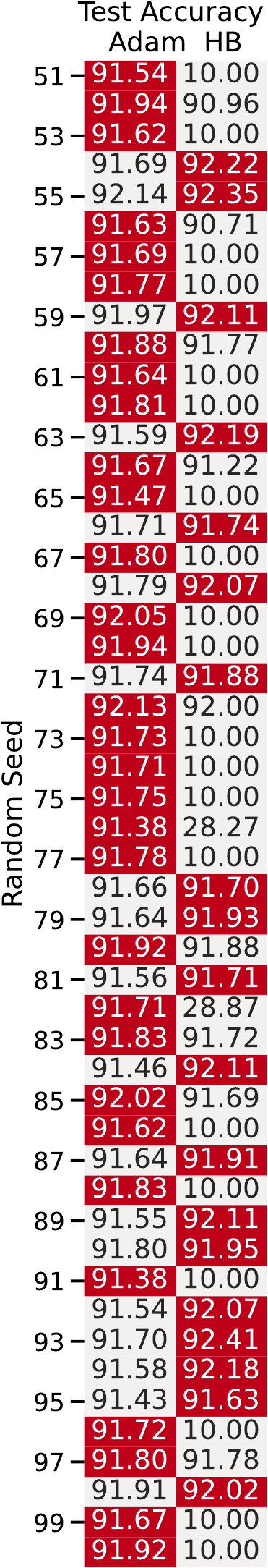} \hspace{2mm}
\includegraphics[scale=.5]{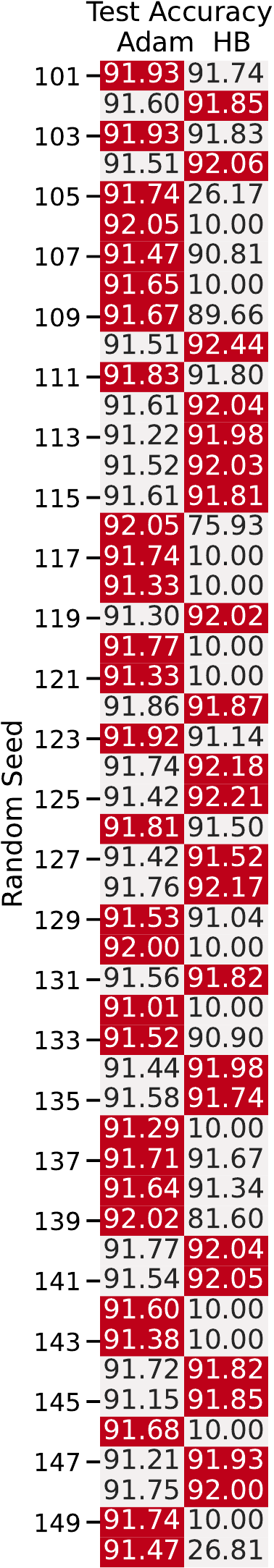} \hspace{2mm}
\includegraphics[scale=.5]{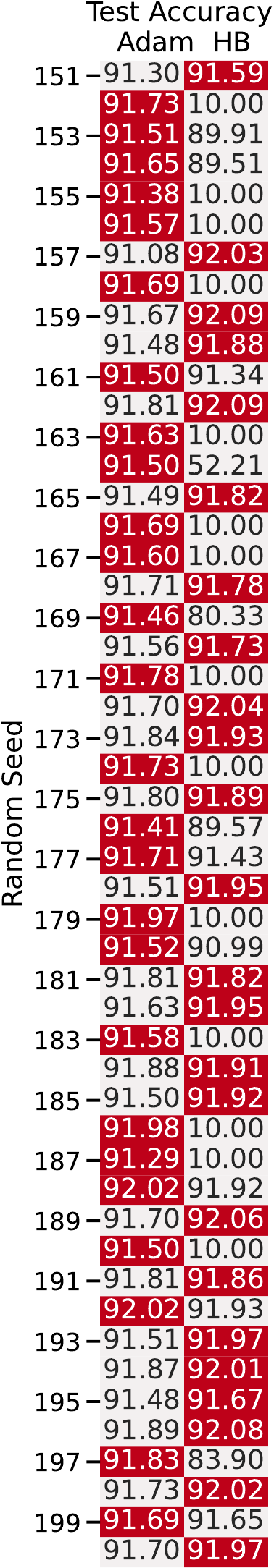} 
    \caption{Heatmap logs of test accuracy of VGG-16 on CIFAR-10 for Adam vs. Heavy Ball (HB) using our defended random search EHPO. Red indicates higher test accuracy for the given random seed.}
    \label{fig:defended-ehpo-logs-adam-hb}
\end{figure}
\FloatBarrier

%% file: section/99-appendix/21-hpo/60-app-hpo-conclusion.tex
\section{Section \ref{sec:hpo:conclusion} Appendix: Notes on Conclusion}

\subsection{Additional Practical Takeaways}

In our conclusion in Section \ref{sec:hpo:conclusion}, we note the following practical takeaways:
\begin{itemize}
    \item \textbf{Researchers should have their own notion of skepticism, appropriate to their specific task.} There is no one-size-fits-all defense solution. Our results are \emph{\textbf{broad insights}} about defended EHPO: A defended EHPO is \emph{\textbf{always possible}}, but finding an efficient one will depend on the task.
    \item \textbf{Researchers should make explicit how they choose hyper-HPs.} 
    What is reasonable is ultimately a function of what the ML community accepts. Being explicit, rather than eliding hyper-HP choices, is essential for helping decide what is reasonable. As a heuristic, we recommend setting hyper-HPs such that they include HPs for which the optimizers' performance starts to degrade, as we do above. 
    \item \textbf{Avoiding hyperparameter deception is just as important as reproducibility}. We have shown that reproducibility \cite{henderson2017reinforcement, pineau2019checklist, gundersen2018reproducibility, bouthillier2019reproducibility, Sinha2020-aw} is only part of the story for ensuring reliability. While necessary for guarding against brittle findings, it is not sufficient. We can replicate results---even statistically significant ones---that suggest conclusions that are altogether wrong.
\end{itemize}

We elaborate here that our defended random search EHPO indicates a particular form of skepticism that may (or may not) be appropriate to different ML tasks. That is, we suggest a defended EHPO, but do not claim that that EHPO is optimal or suited for all tasks. Even though it may not be optimal, the guarantees it affords would translate to other tasks (so long as the assumption is maintained that there is an upper bound on how much the hyper-HPs can control the HPs). So, while we do not necessarily encourage practitioners to use our particular defended EHPO, we do not discourage it either. The main take away is that practitioners should develop their own notion of skepticism (appropriate to their particular task) and be explicit about the assumptions they rely on when selecting hyper-HPs. The way one chooses hyper-HPs should be defensible.

When in doubt, as a heuristic we recommend using a search space that includes where an algorithm's performance starts to degrade (to be assured that a maximum, even if a local one, has been found). We refer to our dynamic two-phase protocol (which we describe in detail in Appendix \ref{app:sec:phased}) as an example of how to do this. We first do a broad (but coarse) search. We used grid search for that initial sweep. Random search may be a better choice for some tasks. We were familiar with Wilson et al.~\cite{wilson2017marginal} (and many have written about it), and felt confident that grid search would capture the space well based on the results that others have also reported on this task. We then used this first sweep to determine which hyper-HPs we should use for our second, finer-grained sweep. We apply our more expensive, defended EHPO for this second sweep, using the hyper-HPs we selected from the first sweep. In other words, we spent a bit of time/our compute budget justifying to ourselves that we were picking reasonable hyperparameters -- instead of just picking one grid or range for random search to sample, and hoping that our results would be representative of other search spaces.\looseness=-1

\subsection{Broader Impact}
\label{app:sec:impact}
As we suggest in Section \ref{sec:hpo:defense}, our work can be considered as related to (but orthogonal with) with prior studies on reproducibility as advocating for more robust scientific practices in ML research. In particular, our work complements prior empirical studies that shine a light on reliability issues in ML---issues that relate particularly to traditionally underspecified choices in hyperparameter optimization \cite{choi2019empirical, sivaprasad2020hpo, lipton2018troubling}. In contrast to this prior work, which illustrates the issue with experiments, we provide a theoretical contribution that enables ML practitioners and researchers to defend against unreliable, inconsistent HPO. We provide a theoretically-backed mechanism to promote and facilitate more trustworthy norms and practices in ML research.

More broadly, our work can be understood as a mechanism for dealing with \emph{measurement bias}---the misalignment between what one intends to measure and what they are actually measuring---for overall ML algorithm performance. While alleviating measurement bias is by no means novel to more mature branches of science \cite{gould1996mismeasure}, including other fields of computing \cite{mytkowicz2009pl}, until recently it has been under-explored in ML. Beginning in the last couple of years, measurement bias is now coming under increased scrutiny with respect to the origins of empirical gains in ML \cite{musgrave2020metric, forde2021model}. In current work, it is often difficult to disentangle whether the concluded measured performance gains are due to properties of the training algorithm or to fortuitous HP selection. Our formalization, rather than allowing HPO choices to potentially obscure empirical results, provides confidence in the conclusions we can draw about overall algorithm performance.

Our work also highlights how there is a human element, not a just statistical one, to bias in ML pipelines: Practitioners make decisions about HPO that can heavily influence performance (e.g., choice of hyper-hyperparameters). The human element of biasing solution spaces has been discussed in sociotechnical writing \cite{friedman1996bias, smith1985limits, cooper2021emergent}, in AI \cite{mitchell1980bias}, in the context of ``p-hacking'' or ``fishing expeditions" for results that fit a desired pattern~\cite{gelman2019fishing}, and was also the focus of Professor Charles Isbell's NeurIPS 2020 keynote~\cite{neurips2020keynote}. Formalizing the process for how to draw conclusions from HPO, as we do here, has the potential to alleviate the effects of this type of human bias in ML systems. 

Lastly, our insights concerning robustness also extend to growing areas in ML that use learning to guide hyperparemeter selection, such as meta-learning and neural architecture search \cite{elsken2018nas, zoph2017nas, automl_book, neurips2020meta,icml2020automl}. While the assisting learning agents in those methods guide choosing hyperparameters for the trained output model, their own hyperparameters tend to be either manually tuned or chosen with more traditional HPO strategies, like grid search \cite{zhong2018practical}. In other words, these processes can exhibit the bias problem discussed above and are therefore potentially subject to hyperparameter deception, which can be mitigated by the work we present here.

%% file: section/99-appendix/22-fairness/00-app-fairness.tex
\newcommand{\appprelim}{Extended Preliminaries}
\newcommand{\appprelimnotation}{Notes on notation and on our choice of terminology}
\newcommand{\appsetuplimitations}{Constraints on our setup}
\newcommand{\appprelimcosts}{Costs and the classification decision threshold}
\newcommand{\appprelimbootstrap}{The bootstrap method}

\newcommand{\appvariance}{Additional Details on Variance and Self-Consistency}
\newcommand{\appprelimenoisebias}{Other statistical sources of error}
\newcommand{\appprelimvariance}{Our variance definition}
\newcommand{\appconsistencyderiving}{Deriving self-consistency from variance}
\newcommand{\appconsistencycost}{Cost-independence of self-consistency}
\newcommand{\appconsistencydetails}{Additional details on our self-consistency metric}

\newcommand{\appothervariance}{Related Work and Alternative Notions of Variance}
\newcommand{\appmain}{Defining variance in relation to a ``main prediction''}

\newcommand{\appprelimcomparison}{Why we choose to avoid computing the main prediction}
\newcommand{\appmaincost}{The main prediction and cost-sensitive loss}

\newcommand{\appmodelm}{Putting our work in conversation with research on model multiplicity}
\newcommand{\appconcurrent}{Concurrent work}

\newcommand{\appalgorithm}{Additional Details on Our Algorithmic Framework}
\newcommand{\appsecconfidence}{Self-consistent ensembling with abstention}

\newcommand{\appexperiments}{Additional Experimental Results and Details for Reproducibility}

\newcommand{\appexperimentsdata}{Hypothesis classes, datasets, and code}
\newcommand{\apphmda}{The standalone \texttt{HMDA} tookit}
\newcommand{\appcluster}{Cluster environment details}
\newcommand{\appillustrativedetails}{Details on motivating examples in the main paper}
\newcommand{\appadditionalillustrative}{Additional illustrative results}
\newcommand{\appcompasadult}{\texttt{COMPAS} and \texttt{Old Adult}}
\newcommand{\appgerman}{\texttt{South German Credit}}
\newcommand{\apptaiwan}{\texttt{Taiwan Credit}}
\newcommand{\appnewadult}{\texttt{New Adult - CA}}
\newcommand{\apphmdaillustrative}{\texttt{HMDA}}
\newcommand{\appillustrativediscussion}{Extended discussion of illustrative examples of self-consistency}
\newcommand{\appexperimentsalgo}{Validating our algorithm in practice}
\newcommand{\appcompas}{\texttt{COMPAS}}
\newcommand{\appoldadult}{\texttt{Old Adult}}
\newcommand{\appwasserstein}{Measuring the distance between empirical self-consistency curves}
\newcommand{\appalgodiscussion}{Discussion of extended results for Algorithm~\ref{algo:bagging-confidently}}
\newcommand{\appfair}{Reliability and fairness metrics in \texttt{COMPAS} and \texttt{South German Credit}}

\newcommand{\appfuture}{Brief notes on future research}

\chapter{Appendix for Arbitrariness and Social Prediction}\label{chapter:app:fairness}

The full paper's appendix is extensive, containing figures for every experiment. 
We provide an abridged version here, and defer the \href{https://arxiv.org/abs/2301.11562}{arXiv version} of the paper for the full set of results. At the time of this dissertation's submission for  publication, an update is still pending to the authoritative arXiv paper.  

\input{section/99-appendix/22-fairness/10-app-fairness-setup}
\input{section/99-appendix/22-fairness/20-app-fairness-variance}
\input{section/99-appendix/22-fairness/30-app-fairness-othervariance}
\input{section/99-appendix/22-fairness/40-app-fairness-algo}
\input{section/99-appendix/22-fairness/500-app-fairness-empirical}

\input{section/99-appendix/22-fairness/600-app-fairness-future}

%% file: section/99-appendix/22-fairness/10-app-fairness-setup.tex
\section{\appprelim}\label{app:sec:prelim}

\subsection{\appprelimnotation}\label{app:sec:prelim:note}

\paragraph{Terminology.} Traditionally, what we term ``observed labels'' $\olabel$ are often referred to instead as the ``ground truth'' or ``correct'' labels~\citep[e.g.]{abumostafa2012learning, hastie2009statistical, kong1995decomp}. We avoid this terminology because, as the work on label bias has explained, these labels are often unreliable or contested~\citep{friedler2016impossibility, cooper2021emergent}. 

\paragraph{Sets, random variables, and instances.} We use bold non-italics letters to denote random variables (e.g., $\rvx$, $\rmD$), capital block letters to denote sets (e.g., $\instances$, $\labels$), lower case italics letters to denote scalars (e.g., $\olabel$), bold italics lower case letters to denote vectors (e.g., $\instance$), and bold italics upper case to denote matrices (e.g., $\datasetk$). For a complete example, $\instance$ is an arbitrary instance's feature vector, $\instances$ is the set representing the space of instances  $\instance$ ($\instance \in \instances$), and $\rvx$ is the random variable that can take on specific values of $\instance \in \instances$. We use this notation consistently, and thus do not always define all symbols explicitly.


\subsection{\appsetuplimitations}\label{app:sec:limitations}

Our setup, per our definition of the learning process (Definition~\ref{def:learningprocess}) is deliberately limited to studying the effects of variance due to changes in the underlying training dataset, with such datasets drawn from the same distribution.
For this reason, Definition~\ref{def:learningprocess} does not include the data collection process or hyperparameter optimization (HPO), which can further introduce non-determinism to machine learning, and are thus assumed to have been  already be completed.

Relatedly, variance-induced error can of course have other sources due to such non-determinism. 
For example, stochastic optimization methods, such as SGD and Adam, can cause fluctuations in test error; as, too, can choices in HPO configurations~\citep{cooper2021hpo}. While each of these decision points is worthy of investigation with respect to their impact on fair classification outcomes, we aim to fix as many sources of randomness as possible in order to highlight the particular kind of arbitrariness that we describe in Sections~\ref{sec:fairness:intro} and~\ref{sec:fairness:significance}. As such, we use the Limited-memory BFGS solver and fix our hyperparameters based on the results of an initial search (Section~\ref{sec:fairness:empirical}), for which we selected a search space through consulting related work such as Chen et al.~\cite{chen2018tradeoff}.

\subsection{\appprelimcosts}\label{app:sec:prelim:costs}

For reference, we provide a bit more of the basic background regarding the relationship between the classification decision threshold $\tau$ and costs of false positives \fp{} ($\costfp$) and false negatives \fn{} ($\costfn$). We visualize the loss as follows:
\begin{table}[ht!]
	\begin{center}
		\caption{Confusion matrix for cost-sensitive loss $\lossarb$, adapted from Elkan~\cite{elkan2001cost}.}
		\label{app:tab:confusion}
		\begin{tabular}{lll}
			\toprule %
			{} & \textbf{$\pred = 0$} & \textbf{$\pred = 1$}\\
			\midrule
			\textbf{$\olabel = 0$} & \tn: $0$ & \fp: $\costfp$ \\
			\midrule %
			\textbf{$\olabel = 1$} & \fn: $\costfn$ & \tp: $0$ \\
			\bottomrule %
		\end{tabular}
	\end{center}
\end{table}

\noindent 0-1 loss treats the cost of different types of errors equally $\costfp = \costfn = 1)$; false positives and false negatives are quantified as equivalently bad -- they are \emph{symmetric}; the case for which $\costfp \neq \costfn$ is \emph{asymmetric} or \emph{cost-sensitive}.

Altering the asymmetric of costs shifts the classification decision threshold $\tau$ applied to the underlying regressor $\regressork$. We can see this by examining the behavior of $\regressork$ that we learn. $\regressork$ estimates the probability of a each label given $\instance$ (since we do not learn using $\group$), i.e., that we develop a good approximation of the distribution $p(\rvy | \rvx)$. Ideally, $\regressork$ will be similar to the Bayes optimal classifier (for which the classification rule produces classifications $y^*$ that yield the smallest weighted sum of the loss, where the weights are the probabilities of a particular label $\rvy=i$ for a given $(\instance, \group)$, i.e., sums over\looseness=-1
\begin{align}
	\label{app:eq:optsum}
	p(\rvy=i |\rvx=\instance)\; \lossarb(i, y').
\end{align}

\noindent For binary classification, 
the terms of (\ref{app:eq:optsum}) in the sum for a particular $y'$ 
yield two cases: 
\begin{itemize}
	\item \textbf{$i = y'$}: By definition, $\lossarb(i, y') = 0$; therefore, (\ref{app:eq:optsum}) $=0$.\looseness=-1
	
	\item \textbf{$i \ne y'$}: By definition, $\lossarb(i, y') = \costfp$ or $\ell(i, y') = \costfn$. 
	So, (\ref{app:eq:optsum}) will weight the cost by the probability $p(\rvy=i | \rvx=\instance)$.
\end{itemize}

We can therefore break down the Bayes optimal classifier into the following decision rule, which we hope to approximate through learning. For an arbitrary $(\instance, \group)$ and $\labels = \{0,1\}$, 

\begin{align*}
	&\min \Big(\overbrace{\overbrace{p(\rvy = 0| \rvx = \instance)}^{\text{Probability of \fp}} \times \costfp + \overbrace{p(\rvy = 1| \rvx = \instance)}^{\text{Probability of \tp}}\times 0}^{\text{Weighted cost of predicting positive (1) class }}, \;\; \overbrace{\overbrace{p(\rvy = 0| \rvx = \instance)}^{\text{Probability of \tn}} \times 0 + \overbrace{p(\rvy = 1| \rvx = \instance)}^{\text{Probability of \fn}}\times \costfn}^{\text{Weighted cost of predicting negative (0) class}}\Big) \\ 
	= &\min \Big(\overbrace{p(\rvy = 0| \rvx = \instance)}^{\text{Probability of \fp}} \times \costfn, \;\; \overbrace{p(\rvy = 1| \rvx = \instance)}^{\text{Probability of \fn}} \times \costfn\Big)
\end{align*}

That is, to predict label $1$, the cost of mis-predicting $1$ (i.e., the cost of a false positive \fp) must be be smaller than the cost of mis-predicting $0$ (i.e, the cost of a false negative \fn). In binary classification $p(\rvy| \rvx=\instance)= p(\rvy=1| \rvx=\instance) + p(\rvy=0 | \rvx=\instance)=1.$ So, we can assign $p(\rvy=1|\rvx=\instance) = \tau$ and $p(\rvy=0|\rvx=\instance) = 1-\tau$, and rewrite the above as
\begin{align}
	\label{app:eq:decision-rule}
	\min \Big( (1-\tau)\,\costfp, \;\; \tau \costfn\Big).
\end{align}

\noindent The decision boundary is the case for which both of the arguments to $\min$ in (\ref{app:eq:decision-rule}) are equivalent (i.e., the costs of predicting a false positive and a false negative are equal), i.e., 
\begin{align*}
	(1-\tau)\,\costfp &=  \tau \costfn \Rightarrow \tau = \frac{\costfp}{\costfp + \costfn}, \text{ so,}\\
	\modelk(\instance) = \1[\regressork(\instance) \geq \tau] &= 
	\begin{cases}
		1,& \text{if } p(\rvy = 1 | \rvx = \instance) \ge \tau = \frac{\costfp}{\costfp + \costfn}\\
		0,              & \text{otherwise}.
	\end{cases}
\end{align*}


\noindent For 0-1 loss, in which $\costfp = \costfn = 1$, $\tau$ evaluates to $\frac{1}{2}$. 
If we want to model asymmetric costs, then we need to change this decision threshold to account for which type of error is more costly. For example, let us say that false negatives are more costly than false positives, with $\costfp = 1$ and $\costfn = 3$. This leads to a threshold of $\frac{1}{4}$, which  biases $\modelk$ toward choosing the (generally cheaper to predict/more conservative) positive class. 

\subsection{\appprelimbootstrap}\label{app:sec:prelim:boot}

In the bootstrap method, we treat each dataset $\hatdatasetk \in \hat{\datasets}$ as equally likely.  For each set aside test example $(\instance, \group, \olabel)$, we can approximate $\err(\tproc, \datasets, (\instance, \group, \olabel))$ empirically by computing 
\begin{align}
	\label{eq:errorsumapprox}
	\haterr\big(\tproc, \hatdatasets, (\instance, \group, \olabel)\big) &= \frac{1}{\boot} \sum_{i=1}^{\boot} \ell\big(\olabel, \hatmodel_{\hatdataset_i}(\instance)\big)
\end{align}

\noindent for a concrete number of replicates $\boot$. We estimate overall error $\haterr(\tproc, \hatdatasets)$ for the test set by additionally summing over each example instance $(\instance, \group, \olabel)$, which we can further delineate into $\hatfpr$ and $\hatfnr$, or into group-specific $\haterr_\group$, $\hatfpr_\group$, and $\hatfnr_\group$ by computing separate averages according to $\group$.\looseness=-1 

The bootstrap method exhibits less variance than cross-validation, but can be biased --- in particular, pessimistic --- with respect to estimating expected error. To reduce this bias, one can follow our setup in Definition~\ref{def:learningprocess}, which splits into train and test sets before resampling. For more information comparing the two methods, see Efron and Tibshirani~\cite{efron1993bootsrap,efron1997boot}. Further, recent work shows that, in relation to studying individual models, CV is in fact asymptotically uninformative regarding expected error~\citep{wager2020cv}.

%% file: section/99-appendix/22-fairness/20-app-fairness-variance.tex
\section{\appvariance}\label{app:sec:variance}

In this appendix, we provide more details on other types of statistical error (Appendix~\ref{app:sec:noisebias}), on variance (Appendix~\ref{app:sec:ourvariance}) and self-consistency (Appendix~\ref{app:sec:consistency}). Following this longer presentation of our metrics, we then provide some additional information on other definitions of variance that have been used in work on fair classification, and contextualize issues with these definitions that encouraged us to deviate from them in order to derive our definition of self-consistency (Appendix~\ref{app:sec:othervariance}).

\subsection{\appprelimenoisebias}\label{app:sec:noisebias}

\paragraph{Noise.} 
Noise is traditionally understood as \textit{irreducible error}; it is due to inherent randomness in the data, 
which 
cannot be captured perfectly accurately by a deterministic decision rule $\modelk$. 
Notably, noise is an aspect of the data collection pipeline, not the learning process (Definition~\ref{def:learningprocess}). 
It is \textit{irreducible} in the sense that it does not depend on our choice of training procedure $\tproc$ or how we draw datasets for training from $\datasets$, either in theory or in practice. Heteroskedastic noise across demographic groups is often hypothesized to be a source of unfairness in machine learning~\citep{cooper2021emergent, chen2018tradeoff}. 
	Importantly, albeit somewhat confusingly, 
	this is commonly referred to as label bias, where ``bias'' connotes discrimination, as opposed to the statistical bias that we mention here.
	
	Unlike noise, bias and variance are traditionally understood as sources of epistemic uncertainty. 
	These sources of error are \textit{reducible} because they are contingent on the modeling choices we make in the learning process; if we knew how to model the task at hand more effectively, in principle, we could reduce bias and variance error. 
	
	\paragraph{Bias.} Within the amount of reducible error, bias reflects the error associated with the chosen hypothesis class $\sH$, and is therefore governed by decisions concerning the training procedure $\tproc$ in the learning process (Definition~\ref{def:learningprocess}). This type of error is persistent because it takes effect at the level of possible models in $\sH$; in expectation, all models $\modelk \in \sH$ have the same amount of bias-induced error. 
	
	Whereas variance depends on stochasticity in the underlying training data, noise and bias error are traditionally formulated in relation to the Bayes optimal classifier --- the best possible classifier that machine learning could produce for a given task~\citep{geman1992bvd, abumostafa2012learning, domingos2000icml}. Since the Bayes optimal classifier is typically not available in practice, we often cannot estimate noise or bias directly in experiments.
	
	Of the three types of statistical error, it is only variance that seems to reflect the intuition in Figure~\ref{fig:vote} concerning the behavior of different possible models $\modelk$. This is because noise is a property of the data distribution; for a learning process (Definition~\ref{def:learningprocess}), in expectation we can treat noise error as constant. Bias can similarly be treated as constant for the learning process: It is a property of the chosen hypothesis class $\hclass$, and thus is in expectation the same for each $\modelk \in \hclass$. In Figure~\ref{fig:vote}, we are keeping the data distribution constant and $\hclass$ constant; we are only changing the underlying subset of training data to produce different models $\modelk$.
	
	\subsection{\appprelimvariance}\label{app:sec:ourvariance}
	
	We first provide a simple proof that explains the simplified version for our empirical approximation for variance in (\ref{eq:hatvar}).
	
	\vspace{-.25cm}
	\begin{proof}
		For the models $\{\model_{\dataset_b}\}_{b=1}^{\boot}$ that we produce, we denote $\hatlabels$ to be the multiset of their predictions on $(\instance, \group)$. $|\hatlabels| = \boot = \boot_0 + \boot_1$, where $\boot_0$ and $\boot_1$ represent the counts of $0$ and $1$-predictions, respectively. We also set the cost of false positives to be $\lossarb(0, 1) = \costfp$ and the cost of false negatives to be $\lossarb(1, 0) = \costfn$. 
		
		Looking at the sum in $\hatvar$ (i.e., $\sum_{i \neq j}$), each of the $\boot_0$ $0$-predictions will get compared to the other $\boot_0 - 1$ $0$-predictions and to the $\boot_1$ $1$-predictions.  By the definition of $\lossarb$, each of the $\boot_0 - 1$  computations of $\lossarb(0, 0)$ evaluates to $0$ and each of the $\boot_1$ computations of $\lossarb(0, 1)$ evaluates to $\costfp$. Therefore, the $\boot_0$ $0$-predictions contribute 
		\begin{align*}
			\textstyle
			\boot_0 \times \big[\big(0 \times (\boot_0 - 1)\big) + \costfp 
			\times \boot_1 \big] = \costfp\boot_0\boot_1
		\end{align*}
		
		\noindent to the sum in $\hatvar$, and,  by similar reasoning, 
		$\boot_1 \times \big[\big(0 \times (\boot_1 - 1)\big) + \costfn \times \boot_0 \big] = \costfn\boot_0\boot_1.$ It follows that the total sum in $\hatvar$ is
		\begin{align*}
			&\sum_{i \neq j} \lossarb\Big(\hatmodel_{\hatdataset_i}(\instance), \hatmodel_{\hatdataset_j}(\instance)\Big) = (\costfp + \costfn)\boot_0\boot_1. \text{ Therefore}\\
			&\overbrace{\frac{1}{\boot(\boot-1)} \sum_{i \neq j} \lossarb\Big(\hatmodel_{\hatdataset_i}(\instance), \hatmodel_{\hatdataset_j}(\instance)\Big)}^{\hatvariance} = \overbrace{\frac{(\costfp + \costfn)\boot_0\boot_1}{\boot(\boot-1)}}^{\text{(\ref{eq:hatvar})}}
		\end{align*}
	\end{proof}
	
	\paragraph{The effect of $\tau$ on variance.} As discussed in Appendix~\ref{app:sec:prelim:costs}, $\costfp$ and $\costfn$ can be related to changing $\tau$ applied to $\regressork$ to produce classifier $\modelk$. We analyze the range of minimal and maximal empirical variance by examining the behavior of $B\to\infty$, i.e.,
	\begin{align}
		\label{app:eq:lim-variance}
		\lim_{\boot\to\infty} \frac{(\costfp + \costfn)\boot_0\boot_1}{\boot(\boot-1)}.
	\end{align}
	
	
	\paragraph{Minimal variance.} Either $\boot_0$ or $\boot_1$ (exclusively, since $\boot_0 + \boot_1 > 1$) will be $=0$, with the other being $=\boot$, making (\ref{app:eq:lim-variance}) equivalent to
	\begin{align*}
		\lim_{\boot\to\infty} \frac{(\costfp + \costfn)\times 0}{\boot(\boot-1)} = 0, \text{regardless of the value of $\costfp + \costfn$.}
	\end{align*}
	
	\paragraph{Maximal variance.} $\boot_0$ will represent half of $\boot$, with $\boot_1$ representing the other half. More particularly, $\boot_0 = \frac{\boot}{2}$ and $\boot_1 = \frac{\boot}{2}$; or, without loss of generality, $\boot_0=\frac{\boot-1}{2}$ and $\boot_1=\frac{\boot+1}{2}$. This means that
	\begin{align*}
		\frac{(\costfp + \costfn)\boot_0\boot_1}{\boot(\boot-1)} &= \frac{(\costfp + \costfn)(\frac{\boot}{2})^2} {\boot(\boot-1)} & \text{\small{\bigg(Or, $= \frac{(\costfp + \costfn)(\frac{\boot-1}{2})(\frac{\boot+1}{2})} {\boot(\boot-1)}$\bigg)}}\\
		&= \frac{(\costfp + \costfn)(\frac{\boot^2}{4})}{\boot^2 - \boot} & \text{\small{\bigg(Or, $= \frac{(\costfp + \costfn)(\frac{(\boot^2 - 1}{4})}{\boot(\boot-1)}$; it will not matter in the limit\bigg)}}\\
		&= \frac{(\costfp + \costfn)\boot^2} {4\boot^2 - 4\boot}.
	\end{align*}
	
	\noindent And, therefore, 
	\begin{align}
		\label{app:eq:maxvariance}
		\lim_{\boot\to\infty}\frac{(\costfp + \costfn)\boot^2} {4\boot^2 - 4\boot} &= \frac{\costfp + \costfn}{4}.
	\end{align}
	It follows analytically that variance will be in the range  $[0, \frac{\costfp + \costfn}{4})$. However, empirically, 
	for concrete $\boot$, $\hatvariance \rightarrow [0, \frac{\costfp + \costfn}{4} + \epsilon]$, for smaller positive $\epsilon$ as the number of models $\boot$ increases. The maximal variance will better approximate $\frac{\costfp + \costfn}{4}$ as $\boot$ gets larger, but will be $>\frac{\costfp + \costfn}{4}$. For example, for 0-1 loss $\frac{\costfp + \costfn}{4} = \frac{2}{4} = 0.5$. For $\boot=100$, the maximal 
	$\hatvariance = \frac{2 \times 50 \times 50}{100 \times 99} = \frac{50}{99} \approx .505$.

	\subsection{\appconsistencyderiving}\label{app:sec:consistency}
	
	In this appendix, we describe the relationship between variance (Definition~\ref{def:variance}) and self-consistency (Definition~\ref{def:sc}) in more detail, and show that $\hatsc\big(\tproc, \{\dataset_b\}_{b=1}^\boot, (\instance, \group)\big) \rightarrow [0.5 - \epsilon, 1$] for small positive $\epsilon$ as the number of models $\boot$ increases. 

	\begin{proof}

		Note that, by the definition of 0-1 loss, $\costfp = \costfn = 1$, so 
		
		\begin{align}
			\label{app:eq:01variance}
			\hatvariance_{\text{0-1}} &= \frac{1}{\boot(\boot-1)}\sum_{i\neq j}\1[\model_{\dataset_i}(\instance) \ne \model_{\dataset_j}(\instance)]  = \frac{2\boot_0\boot_1}{\boot(\boot-1)}.
		\end{align}
		
		By the definition of the indicator function $\1$,
		
		\begin{align*}
			1 &= \frac{1}{\boot(\boot-1)}\sum_{i\neq j}\Big[\overbrace{\1[\model_{\dataset_i}(\instance) \ne \model_{\dataset_j}(\instance)]}^{\text{From } \hatvariance_{\text{0-1}}} + \:\:\: \overbrace{\1[\model_{\dataset_i}(\instance) = \model_{\dataset_j}(\instance)]}^{\text{From } \hatconsistency}\Big]\\
			&= \overbrace{\frac{2\boot_0\boot_1}{\boot(\boot-1)}}^{(\ref{app:eq:01variance})} + \frac{1}{\boot(\boot-1)}\sum_{i\neq j}\1[\model_{\dataset_i}(\instance) = \model_{\dataset_j}(\instance)]. 
		\end{align*}
    Therefore, rearranging,
		\begin{align*}
			\hat{\texttt{SC}}\big(\mathcal{A}, \hatdatasets, (\instance, \group)\big) = \frac{1}{\boot(\boot-1)}\sum_{i\neq j}\1[\model_{\dataset_i}(\instance) = \model_{\dataset_j}(\instance)] =  1 - \frac{2\boot_0\boot_1}{\boot(\boot-1)}.
		\end{align*}
		
	\end{proof}
	
	We note that $\hatsc$ (\ref{eq:sc}) is independent of specific costs $\costfp$ and $\costfn$. Nevertheless, the choice of decision threshold $\tau$ will of course impact the values of $\boot_0$ and $\boot_1$ in practice. In turn, this will impact the degree of self-consistency that a learning process exhibits empirically. In short, the measured degree of self-consistency in practice will depend on the choice of $\lossarb$. Further, following an analysis similar to what we can show that $\hatsc$ will be a value in $[0.5 + \epsilon, 1]$, for small positive $\epsilon$. This reality is reflected in the results that we report for our experiments, for which $B=101$ yields minimal $\hatsc \approx 0.495$. 

	\paragraph{\appconsistencycost} Intuitively, \emph{self}-consistency of a learning process is a relative metric; it is a quantity that is measured relative to the learning process. We therefore conceive of it as a metric that is normalized with respect to the learning process (Definition~\ref{def:learningprocess}). Such a process can be maximally $100\%$ self-consistent, but it does not make sense for it to be more than that (reflected by the maximum value of $1$).
	
	In contrast, as discussed in Appendix~\ref{app:sec:variance}, variance can measure much greater than 1, depending on the magnitude of the sum of the costs $\costfp$ and $\costfn$, in particular, for $\costfp + \costfn > 4$ (\ref{app:eq:maxvariance}). However, it is not necessarily meaningful to compare the magnitude of variance across classifiers. Recall that the effect of changing costs $\costfp$ and $\costfn$ corresponds to a change in the binary classification decision threshold, with $\tau = \frac{\costfp}{\costfp + \costfn}$. It is the \emph{relative} costs that change the decision threshold; not the costs themselves. For example, the classifier with costs $\costfp = 1$ and $\costfn = 3$ is equivalent to the classifier with costs $\costfp = \frac{1}{2}$ and $\costfn = \frac{3}{2}$ (for both, $\tau = \frac{1}{4}$), but the former would measure a larger magnitude for variance.
	
	It is this observation that grounds our cost-independent definition of self-consistency in Section~\ref{sec:fairness:significance} and Appendix~\ref{app:sec:consistency}. Given the fact that the magnitude of variance measurements can complicate our comparisons of classifiers, as discussed above, we focus on the part of variance that encodes information about arbitrariness in a learning process: its measure of (dis)agreement between classification decisions that result from changing the training dataset. We could alternatively conceive of self-consistency as the additive inverse of normalized variance, but this is more complicated because it would require a computation that depends on the specific costs, $\hatvariance_{\text{normalized}} = \frac{\hatvariance}{\hatvariance_\text{max}}$.\looseness=-1

    \subsection{\appconsistencydetails}\label{app:sec:consistency:details}
	
	\paragraph{Terminology.}  In logic, the idea of consistent belief has to do with ensuring that we do not draw conclusions that contradcit each other. This is much like the case that we are modeling with self-consistency --- the idea that underlying changes in the dataset can lead to predictions that are directly in contradition~\citep{smullyan1986belief, hintikka1962doxastic, stalnaker2006logic}. Ideas of consistency in legal rules have a similar flavor; legal rules should not contradict each other; legal judgments should not contradict each other (this is at least an aspiration for the law, based on common ideas in legal theory~\citep{fuller1965law, tamanaha2004law}. For both of these reasons, the term ``consistent'' has a natural mapping to our usage of it in this paper. This is especially true in the legal theory case, given that inconsistency in the law is often considered arbitrary and a source of discrimination. 
	
	We nevertheless realize that the word ``consistent'' is overloaded with many meanings in statistics and different subfields computer science like distributed computing \citep[e.g.,]{zhang2020amagold, abadi2012tradeoff}. Nevertheless, due to the clear  relationship between our purposes concerning arbitrariness and discrimination, and definitions in logic and the law, we believe that it is the most appropriate term for our work. 
	
	\paragraph{Quantifying systematic arbitrariness.} We depict \emph{systematic arbitrariness} using the Wasserstein-1 distance~\cite{ramdas2015wass}. This is the natural distance for us to consider because it has a closed form when being applied to CDFs. For our purposes, it should be interpreted as computing the total disparity in self-consistency by examining all possible self-consistency levels $\kappa$ at once.  
	
	Formally,\footnote{We consider the Wasserstein distance for one-dimensional distributions. More generally, the $p$-th Wasserstein distance for such distributions, $\mathcal{W}_p$, requires the inverse CDFs to be well-defined (i.e.,  the CDFs need to be strictly monotonic). This is fine to assume for our purposes. We have to relax the formal definition of the Wasserstein distance, anyway, when we estimate it in practice with a discrete number of samples.} for two groups $\group=0$ and $\group=1$ with respective $\texttt{SC}$ CDFs $F_0$ and $F_1$, 
	\begin{align*}
		\mathcal{W}_{1} &= \int_{\R} |F_0(\kappa) - F_1(\kappa)| \; d\kappa.
	\end{align*}
	
	For self-consistency, which we have defined on $[0.5, 1]$, this is just 
	\begin{align*}
		\mathcal{W}_{1} &= \int_{0.5}^1 |F_0(\kappa) - F_1(\kappa)| \; d\kappa.
	\end{align*}
	
	Empirically, we can approximate this with 
 
	\begin{align*}
		& \hat{\mathcal{W}_1} \coloneqq \frac{1}{|\hat{\sK}|}\sum_{\hat{\sK}} | \hat{F}_0(\hat\kappa) - \hat{F}_1(\hat\kappa)|, \hspace{.5em} \\
        & \text{where } \hat{\sK} = \biggl\{1 - \frac{2\boot_0\boot_1}{\boot(\boot-1)} \bigg| \boot_0 \in \{0 \ldots \boot\} \land \boot_1 \in \{0 \ldots \boot\} \land \boot_0 + \boot_1 = \boot \biggr\}.
	\end{align*}
	
	We typically set $\boot=101$, and thus 
	\begin{align*}
		\hat{\sK} = [&0.49505, 0.49545, 0.49624, 0.49743, 0.49901, 0.50099, 0.50337, 0.50614, 0.50931,\\ 
        &0.51287, 0.51683, 0.52119, 0.52594, 0.53109, 0.53663, 0.54257, 0.54891, 0.55564,\\
        &0.56277, 0.57030, 0.57822, 0.58653, 0.59525, 0.60436, 
		0.61386, 0.62376, 0.63406,\\
        &0.64475, 0.65584, 0.66733, 0.67921, 0.69149, 0.70416, 0.71723, 0.73069, 0.74455,\\
		&0.75881, 0.77347, 0.78851, 0.80396, 0.81980, 0.83604, 0.85267, 0.86970, 0.88713,\\
        &0.90495, 0.92317, 0.94178, 0.96079, 0.9802, 1.0],
	\end{align*}
	\noindent which we use to produce our CDF plots.
	
	When measuring systematic arbitrariness with abstention, we set the probability mass for $<\kappa$ to $0$ it. This makes sense because we are effectively re-defining the $\hatsc$ CDFs to not include instances that exhibit below a minimal amount of $\hatsc$. This also makes comparing systematic arbitrariness across CDFs for different interventions more interpretable. It allows us to keep the number of experimental samples for the empirical CDF measures constant when computing averages, so abstaining would then always have the effect of decreasing systematic arbitrariness. If we did not do this, because the Wasserstein-1 distance is an average, changing the set $\hat{\sK}$, of course, would change the amount of Wasserstein-1 distance --- possibly leading to a relative \emph{increase} (if there are greater discrepancies between $\group$-condition CDF curves at $\geq \kappa$). 

%% file: section/99-appendix/22-fairness/30-app-fairness-othervariance.tex
\section{\appothervariance}\label{app:sec:othervariance}

As noted in Section~\ref{sec:fairness:related}, prior work that discusses variance and fair classification often relies on the definition of variance from Domingos~\cite{domingos2000icml}. We deviate from prior work and provide our own definition for two reasons: 
1) variance in Domingos~\cite{domingos2000icml, domingos2000report} does not cleanly extend to cost-sensitive loss, and 2) the reference point for measuring variance in Domingos~\cite{domingos2000icml,domingos2000report} --- the 
\emph{main prediction} --- can be unstable/ brittle in practice. We start by explaining the Domginos~\cite{domingos2000icml, domingos2000report} definitions, and then use these definitions to support our rationale. 

\subsection{\appmain}\label{app:sec:other:main} 

To begin, we restate the definitions from Domingos~\cite{domingos2000icml, domingos2000report} concerning the expected model (called the \emph{main predictor}).  We change the notation from Domingos to align with our own, as we believe these changes provide greater clarity concerning meaning, significance, and consequent takeaways. Nevertheless, these definitions for quantifying error are equivalent to those in Domingos~\cite{domingos2000report}, and they fundamentally depend on human decisions for setting up the learning process.

Domgingos defines predictive variance in relation to this single point of reference. 
This reference point captures the general, expected behavior of models that could be produced by the chosen learning process. 
We can think of each prediction of this point of reference as the ``central tendency'' of the predictions made by all possible models in $\possiblemodels$ for $(\instance, \group)$.  
Formally,\looseness=-1
\begin{definition}
	\label{def:main}
	The \textbf{main prediction} $\pred$ is the prediction value $y' \in \labels$ that generates the minimum average loss with respect to all of the predictions $\pred \in \hat{\labels}$ generated by the different possible models in $\possiblemodels$. It is defined as the expectation over training sets $\datasets$ for a loss function $\lossarb$, given an example instance $(\instance, \group)$. That is,
	\begin{align}
		\label{eq:main}
		\overline{y} = \argmin_{y'}\E_{\rmD}[\lossarb(\pred, y') | \rvx=\instance, \rvg=\group]. 
	\end{align}
	The \textit{main predictor} $\overline{\model}: \instances \rightarrow \labels$ produces the main prediction $\overline{y}$ for each $(\instance, \group)$.
\end{definition}


What 
(\ref{eq:main}) evaluates to in practice of course depends on the loss function $\lossarb$. For squared loss, the main prediction is defined as the mean prediction of all the $\modelk$~\cite{domingos2000icml, kong1995decomp}. 
Following Kong and Dietterich~\cite{kong1995decomp}, for 0-1 loss Domingos~\cite{domingos2000icml} defines the main prediction as the mode/majority vote --- the most frequent prediction for an example instance $(\instance, \group)$.  We provide a more formal discussion of why this is the case when we discuss problems with the main prediction for cost-sensitive loss (Appendix~\ref{app:sec:comparison}). Domingos~\cite{domingos2000icml, domingos2000report} then define variance in relation to specific models $\modelk$ and the main predictor $\overline{\model}$:  
\begin{definition}
	\label{def:app-variance-domingos}
	The \textit{variance}-induced error for fresh example instance $(\instance,\group)$ is
	\begin{align*}
		\variance = \E_\rmD[\lossarb(\overline{y}, \pred)|\rvx=\instance, \rvg=\group],
	\end{align*}
	where $\overline{y} = \overline{\model}(\instance)$ is the main prediction and the $\pred$ are the predictions for the different $\modelk \sim \possiblemodels$.
\end{definition}

\noindent That is, for a specific $(\instance,\group)$, it is possible to compare the individual predictions $\pred = \modelk(\instance)$ to the main prediction $\overline{y} = \overline{\model}(\instance)$. Using the main prediction as a reference point, one can compute the extent of disagreement of individual predictions with the main prediction as a source of error. It is this definition (Definition~\ref{def:app-variance-domingos}) that prior work on fair classification tends to reference when discussing variance~\citep{chen2018tradeoff, black2022selective}. However, as we discuss in more detail below (Appendix~\ref{app:sec:comparison}), many of the theoretical results in Chen et al.~\cite{chen2018tradeoff} follow directly from the definitions in Domingos~\cite{domingos2000icml}, and the experiments do not actually use those results in practice. Black et al.~\cite{black2022selective}, in contrast, presents results that rely heavily on the main prediction in Domingos~\cite{domingos2000icml}. 

\subsection{\appprelimcomparison}\label{app:sec:comparison}

We now compare our definition of variance (Definition~\ref{def:variance}) to the one in Domingos~\cite{domingos2000icml, domingos2000report} (Definition~\ref{def:app-variance-domingos}). This comparison makes clear in detail why we deviate from prior work that relies on Domingos~\cite{domingos2000icml, domingos2000report}.

\paragraph{No decomposition result.} 
Following from above, it is worth noting that by not relying on the main prediction, we lose the applicability of the decomposition result that Domingos~\cite{domingos2000icml, domingos2000report} develops. However, we believe that this is fine for our purposes, as we are interested in the impact of empirical variance specifically on fair classification outcomes. We do not need to reason about bias or noise in our results to understand the arbitrariness with which we are concerned (Section~\ref{sec:var:intuition}). It is also worth noting that prior work on fair classification that leverages Domingos~\cite{domingos2000icml} also does not leverage the decomposition, either. Chen et al.~\cite{chen2018tradeoff} extends the decomposition to subgroups in the context of algorithmic fairness,\footnote{This just involves splitting the conditioning on an example instance of features $\instance$ into conditioning on an example instance whose features are split into $(\instance, \group)$.} and then informally translates the takeaways of the Domingos~\cite{domingos2000icml} result to a notion of a ``level of discrimination.''  Moreover, unlike our work, these prior studies do not actually measure variance directly in its experiments.

\paragraph{No need to compute a ``central tendency.''} 
In Domingos~\cite{domingos2000icml, domingos2000report}, variance is defined in terms of both the loss function $\lossarb$ and the main prediction $\overline{y}$. This assumes that the main prediction is well-defined for the loss function, and that it is well-behaved. While there is a simple interpretation of the main prediction for squared loss (the mean) and for 0-1 loss (the mode/majority vote), it is significantly messier for cost-sensitive loss, which is a more general formulation that includes 0-1 loss. Domingos~\cite{domingos2000icml, domingos2000report} does not discuss this explicitly, so we derive the main prediction for cost-sensitive loss ourselves below. In summary:
\begin{itemize}
	\item The behavior of the main prediction for cost-sensitive loss reveals that the decomposition result provided in the extended technical report (Theorem 4, Domingos~\cite{domingos2000report}) is in fact very carefully constructed. We believe that this construction is so specific that it is not practically useful (it is, in our opinion, hardly ``unified'' in a more general sense, as it is so carefully adapted to specific loss functions and their behavioral special cases).
	\item By decoupling from the need to compute a main prediction as a reference point, our variance definition is ultimately much simpler and more general, with respect to how it accommodates different loss functions.\footnote{This reveals a subtle ambiguity in the definition of the loss  $\lossarb$ in Domingos~\cite{domingos2000icml, domingos2000report}. Neither paper explicitly defines the signature of $\lossarb$. For the main prediction (Definition~\ref{def:main}) and variance (Definition~\ref{def:app-variance-domingos}), there is a lack of clarity in what constitutes a valid domain for $\lossarb$. Computing the main prediction $\overline{y}$ suggests $\lossarb: \labels \times \labels \rightarrow \R_{\ge0}$, where $\overline{y} \in \labels$, but, since $\hat{\labels} \subseteq \labels$, it is possible that $\overline{y} \not\in \labels$. However, the definition of variance suggests that $\lossarb: \labels \times \hat{\labels} \rightarrow \R_{\ge0}$. Since  $\hat{\labels} \subseteq \labels$, it is not guaranteed that $\hat{\labels} = \labels$. This may be fine in practice, especially for squared loss and 0-1 loss (the losses with which Domingos~\cite{domingos2000icml} explicitly contends), but it does arguably present a problem formally with respect to generalizing.}
\end{itemize}

\paragraph{Brittleness of the main prediction.} 
For high variance instances, the main prediction can flip-flop from $\pred=1$ to $\pred=0$ and back. While the strategy in Black et al.~\cite{black2022selective} is to abstain on the prediction in these cases, we believe that a better alternative is to understand that the main prediction is not very meaningful more generally for high-variance examples. That is, for these examples, the ability (and reliability) of breaking close ties to determine the main (simple majority) prediction is not the right approach. Instead, we should ideally be able to embed more confidence into our process than a simple-majority-vote determination.\footnote{This is also another aspect of the simplicity of not needing to define and compute a ``central tendency'' prediction. We do not need to encode a notion of a tie-breaking vote to determine a ``central tendency.'' The main prediction can be unclear in cases for which there is no ``main outcome'' (e.g., Individual 2 in Figure~\ref{fig:vote}), as the vote is split exactly down the middle. By avoiding the need to vote on a main reference point, we also avoid having to ever choose that reference point arbitrarily.} Put different, in cases for which we can reliably estimate the main prediction, but the vote margin is slim, we believe that the main prediction is still uncertain, based on our understanding of variance, intuited in Figure~\ref{fig:vote}. \textbf{The main prediction can be reliable, but it can still, in this view, be arbitrary} (Section~\ref{sec:fairness:related}). With a simple-majority voting scheme, there can be huge differences between predictions that are mostly in agreement, and those that are just over the majority reference point. Freeing ourselves of this reference point via our self-consistency metric, we can define thresholds of self-consistency as our criterion for abstention (where simple-majority voting is one instantiation of that criterion).\looseness=-1\footnote{This problem is worse for cost-sensitive loss, where the main prediction is not always the majority vote (see below).} 

\subsubsection{\appmaincost}\label{app:sec:maincost}

We show here that, for cost-sensitive loss, the main prediction depends on the majority class being predicted, the asymmetry of the costs, and occasional tie-breaking, such that the main prediction can either be the majority vote or the minority vote. 
Domingos~\cite{domingos2000report} provides an error decomposition in Theorem 4, but does not explain the effects on the main prediction. We do so below, and also call attention to 0-1 loss as a special case of cost-sensitive loss, for which the costs are symmetric (and equal to 1). We first summarize the takeaways of the analysis below:
\begin{itemize}
\item \textbf{Symmetric loss}: The main prediction is the \textbf{majority vote}.
\item \textbf{Asymmetric loss}: Compute 1) the relative cost difference (i.e., $\frac{\costfp-\costfn}{\costfn}$), 2) the majority class (and, as a result, the minority class) for the $\pred \in \hat{\labels}$, and 3) the relative difference in the number of votes in the majority and minority classes (i.e., what we call the \emph{vote margin}; below, $\frac{(i + 2j + 1) - i}{i}$)
\begin{itemize}
	\item If the \textbf{majority class} in $\hat{\labels}$ has the \textbf{lower cost} of misclassification, then the main prediction is the \textbf{majority vote}.\looseness=-1
	\item If the \textbf{majority class} in $\hat{\labels}$ has the \textbf{higher cost} of misclassification, then the main prediction \textbf{depends on the asymmetry of the costs and the vote margin}, i.e.,
	\begin{itemize}
		\item If $\frac{\costfp-\costfn}{\costfn} = \frac{(i + 2j + 1) - i}{i}$, we can choose the main prediction to be \textbf{either class} (but must make this choice consistently).
		\item If $\frac{\costfp-\costfn}{\costfn} > \frac{(i + 2j + 1) - i}{i}$, the \textbf{minority vote} is the main prediction.
		\item If $\frac{\costfp-\costfn}{\costfn} < \frac{(i + 2j + 1) - i}{i}$, the \textbf{majority vote} is the main prediction.
	\end{itemize}     
\end{itemize}
\end{itemize}

\begin{proof}
Let us consider cost-sensitive loss for binary classification, for which $\lossarb(0,0) = \lossarb(1, 1) = 0$ and we have potentially-asymmetric loss for misclassifications, i.e. $\lossarb(1,0) = \costfn$ and $\lossarb(0, 1) = \costfp$, with $\costfp, \costfn \in \R^+$. 0-1 loss is a special case for this type of loss, for which $\costfp = \costfn = 1$.     

Let us say that the total number of models trained is $k$, which we evaluate on an example instance $\instance$. Let us set $|\hat{\labels}| = k = 2i + 2j + 1$, with $i \geq 0$ and $j \geq 0$. We can think of $i$ as the common number of votes that each class has, and $2j + 1$ as the margin of votes between the two classes. Given this setup, this means that $k \geq 1$, i.e., we always have the predictions of at least 1 model to consider, and $k$ is always odd. This means that there is always a strict majority classification.

Without loss of generality, on $\instance$, of these $k$ model predictions $\pred \in \hat{\labels}$ , there are $i$ class-$0$ predictions and $i + 2j + 1$ class-$1$ predictions (i.e., we do our analysis with class $1$ as the majority prediction). To compute the main prediction $\overline{y}$, each $\pred \in \hat{\labels}$ will get compared to the values of possible predictions $y' \in \labels=\{0, 1\}$. That is, there are two cases to consider:

\begin{itemize}
	\item \textbf{Case $y' = 0$}:  $y'=0$ will get compared $i$ times to the $i$ $\pred = 0$s in $\hat{\labels}$, for which $\lossarb(0,0)=0$; $y'=0$ will similarly get compared $i + 2j + 1$ times to the $1$s in in $\hat{\labels}$, for which (by Definition~\ref{def:main}) the comparison is $\lossarb(1,0)=\costfn$. By definition of expectation, the expected loss is\looseness=-1
	\begin{align}
		\label{eq:case-y'-0}
		\frac{i \times 0 + (i + 2j+ 1) \times \costfn}{2i + 2j + 1} = \frac{\costfn(i + 2j + 1)}{2i + 2j + 1}. 
	\end{align}
	
	\item \textbf{Case $y' = 1$}: Similarly, the label $1$ will also get compared $i$ times to the $0$s in $\hat{\labels}$, for which the comparison is $\lossarb(0,1) = \costfp$; $y'=1$ will also be compared $i + 2j + 1$ times to the $1$s in $\hat{\labels}$, for which $\lossarb(1,1)=0$. The expected loss is 
	\begin{align}
		\label{eq:case-y'-1}
		\frac{i \times \costfp + (i + 2j + 1) \times  0}{2i + 2j + 1} = \frac{\costfp i}{2i + 2j +1}.
	\end{align}
\end{itemize}

We need to compare these two cases for different possible values of $\costfn$ and $\costfp$ to understand which expected loss is minimal, which will determine the main prediction $\overline{y}$ that satisfies Equation (\ref{eq:main}). The three different possible relationships for values of $\costfn$ and $\costfp$ are $\costfn=\costfp$ (symmetric loss), and $\costfn > \costfp$ and $\costfn < \costfp$ (asymmetric loss). Since the results of the two cases above share the same denominator, we just need to compare their numerators, $\costfn(i+2j+1)$ (\ref{eq:case-y'-0})  and $\costfp i$ (\ref{eq:case-y'-1}).\looseness=-1 

\paragraph{Symmetric Loss (0-1 Loss).} When $\costfn = \costfp = 1$, the numerators in (\ref{eq:case-y'-0}) and (\ref{eq:case-y'-1}) yield expected losses $i + 2j + 1$ and $i$, respectively. We can rewrite the numerator for (\ref{eq:case-y'-1}) as
\begin{align*}
	i + \overbrace{2j + 1}^{\geq 1, \text{ given $j \geq 0$}} &\geq i + 1,
\end{align*}

\noindent which makes the comparison of numerators $i < i + 1$, i.e., we are in the case (\ref{eq:case-y'-1}) $<$ (\ref{eq:case-y'-0}). This means that the case of $y' = 1$ (\ref{eq:case-y'-1}) is the minimal one; the expected loss for class $1$, the most frequent class, is the minimum, and thus the most frequent/ majority vote class is the main prediction. An analogous result holds if we instead set the most frequent class to be $0$. More generally, this holds for all symmetric losses, for which $\costfn = \costfp$.

\noindent$\blacktriangleright$ For \textbf{symmetric losses}, the main prediction $\overline{y}$ is \textbf{majority vote} of the predictions in $\hat{\labels}$.

\paragraph{Asymmetric Loss.} For asymmetric/ cost-sensitive loss, we need to examine two sub-cases:  $\costfn > \costfp$ and $\costfn < \costfp$. 
\begin{itemize}
	\item \textbf{Case $\costfn > \costfp$}: $\costfp i < \costfn(i + \overbrace{2j + 1}^{\ge 1})$, given that $j \geq 0$. Therefore, since $\costfp i$ is minimal and associated with class $1$ (the most frequent class in our setup), the majority vote is the main prediction. We can achieve an analogous result if we instead set $0$ as the majority class.
	
	\noindent \vspace{5pt}$\blacktriangleright$ For \textbf{asymmetric losses}, the main prediction $\overline{y}$ is the \textbf{majority vote} of the predictions in $\hat{\labels}$, \textbf{if the majority class has a cheaper cost associated with misclassification} (i.e., if the majority class is $1$ and $\costfn < \costfp$, or if the majority class is $0$ and $\costfp < \costfn$).
	
	\item \textbf{Case $\costfn < \costfp$}:  If $\costfn < \costfn$, it depends on how asymmetric the costs are and how large the vote margin (i.e., $2j + 1$) between class votes is. There are 3 sub-cases:
 
	\begin{itemize}
		\item \textbf{Case $\costfp i = \costfn(i + 2j + 1)$, i.e. cost equality}:  We can look at the relative asymmetric cost difference of the minority class cost (above $\costfp$, without loss of generality) and the majority class cost (above $\costfn$, without loss of generality), (above $\frac{\costfp - \costfn}{\costfn}$, without loss of generality). If that relative cost difference is equal to the relative difference of the votes between the majority and minority classes (i.e., $\frac{(i + 2j + 1) - i}{i}$), then the costs of predicting either $1$ or $0$ are equal. That is, we can rearrange terms as a ratio of costs to votes:
		\begin{align}
			\costfp i &= \costfn(i + \overbrace{2j + 1}^{\geq 1}) & \text{(The terms in this equality are $>0$)} \nonumber\\
			\frac{\costfp}{\costfn} &= \frac{i + 2j + 1}{i} & \text{(Given the above, $\costfp i > 0$ so $i > 0$)} \nonumber\\
			&= 1 + \frac{2j + 1}{i}\nonumber\\
			\frac{\costfp}{\costfn} - 1 &= \frac{2j + 1}{i}\nonumber\\
			\frac{\costfp - \costfn}{\costfn} &= \frac{2j + 1}{i} = \frac{(i + 2j + 1) - i}{i} \geq \frac{1}{i}
			\label{eq:app:comp}
		\end{align}
		\vspace{5pt}$\blacktriangleright$ For asymmetric loss \textbf{when the majority-class-associated cost is less than the minority-class associated cost and if the expected losses are equal}, then the \textbf{main prediction $\overline{y}$ is either $1$ or $0$}, (and we must make this choice consistently).\looseness=-1
		\item \textbf{Case $\costfp i > \costfn(i + 2j + 1)$}: We can look at the relative asymmetric cost difference of the minority class cost (above $\costfp$, without loss of generality) and the majority class cost (above $\costfn$, without loss of generality), (above $\frac{\costfp - \costfn}{\costfn}$, without loss of generality). If that relative cost difference is greater than the relative difference of the votes between the majority and minority classes (i.e., $\frac{(i + 2j + 1) - i}{i}$), then the \textit{minority vote} yields the minimum cost and is the main prediction $\overline{y}$ (above $\overline{y} = 0$, without loss of generality; an analogous result holds if we had set the majority vote to be $0$ and the minority vote to be $1$).  Following (\ref{eq:app:comp}) above, this is the same as \looseness=-1
  
		\begin{align*}
			\frac{\costfp - \costfn}{\costfn} &> \frac{(i + 2j + 1) - i}{i}
		\end{align*}
		\vspace{5pt}$\blacktriangleright$ For asymmetric loss \textbf{when the majority-class-associated cost is less than the minority-class associated cost}, it is possible for the \textbf{minority class} to have a greater associated loss. In this case, the \textbf{\textit{minority vote} is the main prediction $\overline{y}$}.\looseness=-1    
		
		\item \textbf{Case $\costfp i < \costfn(i + 2j + 1)$}: We can look at the relative asymmetric cost difference of the minority class cost (above $\costfp$, without loss of generality) and the majority class cost (above $\costfn$, without loss of generality), (above $\frac{\costfp - \costfn}{\costfn}$, without loss of generality).  If that relative cost difference s less than the relative difference of the votes between the majority and minority classes (i.e., $\frac{(i + 2j + 1) - i}{i}$), then the majority vote yields to minimum cost and is the main prediction $\overline{y}$ (above $\overline{y} = 1$, without loss of generality; an analogous result holds if we had set the majority vote to be $0$ and the minority vote to be $1$). Following (\ref{eq:app:comp}) above, this is the same as\looseness=-1
  
		\begin{align*}
			\frac{\costfp - \costfn}{\costfn} &< \frac{(i + 2j + 1) - i}{i}
		\end{align*}
		\vspace{5pt}$\blacktriangleright$ For asymmetric loss \textbf{when the majority-class-associated cost is less than the minority-class associated cost}, it is possible for the \textbf{majority class} to have a greater associated loss. In this case, the \textbf{\textit{majority vote} is the main prediction $\overline{y}$}.\looseness=-1  
	\end{itemize}
\end{itemize}
\end{proof}

\subsection{\appmodelm}\label{app:sec:mm}

A line of related work to ours concerns \emph{model multiplicity} and fairness~\citep{watson2023multiplicity, marx2020mm, black2022multiplicity}. This work builds off of an observation made by Breiman~\cite{breiman2001multiplicity} regarding how there are multiple possible models of the same problem that exhibit similar degrees of accuracy. This set of multiple possible models of similar accuracy is referred to as the Rashomon set~\citep{breiman2001multiplicity}.

Work on model multiplicity has recently become fashionable in algorithmic fairness. In an effort to develop more nuanced model selection metrics beyond looking at just fairness and accuracy for different demographic groups, work at the intersection of model multiplicity and fairness tends to examine other properties of models in the Rashomon set in order to surface additional metrics for determining which model to use in practice. 

At first glance, this work may seem similar to what we investigate here, but we observe four key differences:\footnote{We defer discussion of Black et al.~\cite{black2022selective} to~\ref{app:sec:concurrent}.}
\begin{enumerate}
    \item  Model multiplicity places conditions on accuracy and fairness in order to determine the Rashomon set. We place no such conditions on the models that a learning process (Definition~\ref{def:learningprocess}) produces; we simulate the distribution over possible models $\possiblemodels$ without making any claims about the associated properties of those models.
    \item Model multiplicity makes observations about the Rashomon set with the aim of still ultimately putting forward criteria for helping to select \emph{a single model}. While the metrics used to inform these criteria include variance, most often work on model multiplicity still aims to choose one model to use in practice.
    \item Much of the work on model multiplicity emphasizes theoretical contributions, whereas our emphasis is on more experimental contributions. In conjunction with the first point, of ultimately trying to arrive at a single model, this work is also trying to make claims with respect to the Bayes-optimal model. Given our empirical focus --- of what we can actually produce in practice --- claims about optimality are not our concern. 
    \item   We focus specifically on variance reduction as a way to mitigate arbitrariness. We rely on other work, coincidentally contributions also made by Breiman, to study arbitrariness~\citep{breiman1996bagging}, and emphasize the importance of using ensemble models to produce predictions or abstention from prediction. We do not study the development of model selection criteria to pick a single model to use in practice; we use self-consistency to give a sense of predictive confidence about when to predict or not. We always select an ensemble model --- regardless of whether that model is produced by simple or super ensembling (Section~\ref{sec:fairness:algorithms}) --- and then use a user-specified level of self-consistency $\kappa$ to determine when that model actually produces predictions.
\end{enumerate}

These differences ultimately lead to very different methods for making observations about fairness. Importantly, we can study the arbitrariness of the underlying laerning process with a bit more nuance. For example, it could be the case that a particular task is just impossible to get right for some large subset of the test data (and this would be reflected in the Rashomon set of models), but for some portion of it there is a high amount of self-consistency for which we may still want to produce predictions.

Further, based on our experimental approach, we highlight completely different normative problems than those highlighted in work on model multiplicity (notably, see Black et al.~\cite{black2022multiplicity}). So, in short, while model multiplicity deals with related themes as our work --- issues of model selection, problem formulation, variance, etc. --- the goals of that work are ultimately different, but potentially complementary, from those in our paper. 

For example, a potentially interesting direction for future work would be to measure how metrics from work on model multiplicity behave in practice in light of the ensembling methods we present here. We could run experiments using Algorithm~\ref{algo:bagging-confidently} and investigate model multiplicity metrics for the underlying ensembled models. However, we ultimately do not see a huge advantage to doing this. Our empirical results indicate that variance is generally high, and has led to reliability issues regarding conclusions about fairness and accuracy. In fairness settings and available benchmarks, we find that the most important point is that variance has muddled conclusions. Under these circumstances, ensembling with abstention based on self-consistency seems a reasonable solution, in contrast to finding a single best model in the Rashomon set that attains other desired criteria. 

\subsection{\appconcurrent}\label{app:sec:concurrent}

There are several related papers that either preceded or came after this work's public posting. Some of this work is clearly concurrent, given the time frame. Other works that came after ours are not necessarily concurrent, but are either independent and unaware of our paper, or build on our work. 

\paragraph{Setting the stage in 2021.} The present work was scoped in 2021, in direct response to the initial study by Forde et al.~\cite{forde2021model} and critical review by Cooper and Abrams~\cite{cooper2021emergent}. Forde et al.~\cite{forde2021model} was one of the first (if not the first) paper to note that variance is overlooked in problem formulations that consider fairness. However, it was limited in scope and also dealt with deep learning settings, which have multiple sources of non-determinism that can be difficult to tease apart with respect to their effects on variance. 

Cooper and Abrams~\cite{cooper2021emergent} notes important, overlooked normative assumptions in the fairness-accuracy trade-off problem formulation, and suggests that this formulations is tautological. Our work is a natural direction for future research, in this respect -- to see how, in practice, the fairness-accuracy trade-off behaves after we account for variance. Indeed, we find that there is often no such trade-off, but for different reasons than those suggested by Cooper and Abrams~\cite{cooper2021emergent}. We expected there to be residual label bias that contributes to noise-induced error, but ultimately did not really observe this in practice. In these respects, our work both strengthens and complements these prior works. We support their claims, and go significantly beyond the work they did in order to provide such support. Further, our results suggest additional conclusions about experimental reliability in algorithmic fairness.\looseness=-1

\paragraph{Variance and abstention-based ensembling.} 
Black et al.~\cite{black2022selective} is concurrent work that slightly preceded our public posting. This work is similarly is interested in variance reduction, ensembling, and abstention in fairness settings, but fundamentally studies these topics in a different manner. We address four differences:

\begin{enumerate}
    \item Black et al.~\cite{black2022selective} does not take the wide-ranging experimental approach that we take. While we both study variance and fairness, our work also considers \emph{the practice of fair classification research} as an object of study. It is for these reasons that we do so many experiments on benchmark datasets, and clean and release another dataset for others to use.
    \item They rely on the definition of variance from Domingos~\cite{domingos2000icml} in their work, likely building on the choice made by Chen et al.~\cite{chen2018tradeoff} to use this defintion. Much of this Appendix is devoted to discussing Domingos~\cite{domingos2000icml, domingos2000report} and his definition of variance. The overarching takeaway from our discussion is that 1) there are technical problems with this definition (which have been noted by others that investigated the bias-variance-noise trade-off for 0-1 loss in the early 2000s), 2) the definition does not naturally extend to cost-sensitive loss, 3) the main prediction can be unstable in practice and thus should not be the criterion for investigating arbitrariness (indeed, relying on the main prediction just pushes arbitrariness into that definition). While Black et al.~\cite{black2022selective} observes that variance is an important consideration for fairness, they ultimately focus on reliable estimation of the main prediction as the criterion for abstention in their ensembling method. While this kind of reliability is important, it does not deal with the general problem of arbitrary predictions (i.e., it is possible to have a reliable main prediction that is still effectively arbitrary). As a result, the nature of when and how to abstain is very different from ours. We instead base our criterion on a notion of confidence in the prediction, and we allow for flexibility around when to abstain when predictions are too arbitrary.
    \item As a result of the above two differences, the claims and conclusions in both of our works are different. While there are similar terms used in both works (e.g., variance, abstention), which may make the works seem overlapping with a cursory read, our definitions, methods, claims, and conclusions are non-overlapping.  For example, as stated in 1., while Black et al.~\cite{black2022selective}'s use of successful ensembles is intended to address individual-level arbitrariness, by relying on traditional bagging (simple-majority vote ensembling) and the definition of variance from Domingos~\cite{domingos2000icml} that encodes a main prediction, arbitrariness gets pushed into the aggregation rule. If they can estimate the mode prediction reliably, they do not abstain; the mode, however, may still be effectively arbitrary. Our measure of arbitrariness is more direct and more configurable. We can avoid such degenerate situations, as in the example we give for making reliable but arbitrary predictions in Black et al.~\cite{black2022selective}.
    \item We also describe a method for recursively ensembling in order to achieve different trade-offs between abstention and prediction. This type of strategy is absent from Black et al.~\cite{black2022selective}.
\end{enumerate}

\paragraph{Deep learning.} Qian et al.~\cite{qian2021variance} is work that came after Forde et al.~\cite{forde2021model}. They, too, do a wide-ranging empirical study of variance and fairness, but focus on deep learning settings. As a result, they are not examining the fair classification experimental setup that is most common in the field. They therefore make different claims about reliability, which have a similar flavor as those that we make here. However, because of our setup, we are able to probe these claims much deeper (due in part to model/ problem size and being able to limit non-determinism solely to sampling the training data). We mention this work because of its close relationship to Forde et al.~\cite{forde2021model}, which in part inspired this study.

Ko et al.~\cite{ko2023fairensemble} is another deep learning fairness paper. It was posted publicly months after our study, and examines non-overlapping settings and tasks. While the results are similar --- we find fairness after ensembling --- it is again fundamentally different (along the lines of Qian et al.~\cite{qian2021variance} and Forde et al.~\cite{forde2021model}) because it does not study common non-deep-learning setups. They also do not study arbitrariness, which is one of the main purposes of our paper. 

\paragraph{Variance in fair classification.} 
Khan et al.~\cite{khan2023fairness} is concurrent work that studies the same problem that we study, but also takes a different approach. For one, they bake in a notion of 0-1 loss into their definitions. In this respect, our definition of self-consistency generalizes the definitions in their paper. While they run more types of models than we do (we initially ran more, but ultimately stopped because the results were largely similar with more common model types), they do not cover as many datasets as we do. They also do not study arbitrariness or abstention-based ensembling to deal with it, and they do not release a dataset. Further, based on the fact that they study fewer empirical tasks than we do, and that they do not examine abstention-based ensembling, they do not surface or make claims about the experimental reliability issues that we observe. They do not make claims about the fundamental problem that we observe: \textbf{That variance is the culprit for much observed algorithmic unfairness in classification; in practice, we do not seem to learn very confident decisions for large portions of the datasets we examine, and this is a key problem that has been masked by current common experimental practices in the field}. 

\paragraph{Other work.} Any other work on variance and fairness \textbf{comes after} the present study. We have made a significant attempt to keep our related work section up-to-date in response to this new work. We have used a detailed and robust mixed of Google alerts and scraping arXiv to find new related work. We used this same procedure to make sure we  found (ideally) all related work on fairness and variance when we conducted this project. There are some studies, which directly build on ours, which we choose not to cite.\looseness=-1

%% file: section/99-appendix/22-fairness/40-app-fairness-algo.tex
\section{\appalgorithm}\label{app:sec:algorithm}

A natural question is to see if we can improve self-consistency, with the hope that doing so would reduce arbitrariness in the learning process, improve accuracy, and, for the cases in which there is different self-consistency across subgroups, also perhaps improve fairness. To do so, we consider ways of reducing variance, as, based on our definitions (Definition~\ref{def:variance} and~\ref{def:sc}), doing so should improve self-consistency. 

We  consider the classic  \textit{b}ootstrap \textit{agg}regation --- or, \textit{bagging} --- algorithm~\cite{breiman1996bagging} as a starting point. 
It has been well-known since Breiman~\cite{breiman1996bagging} that \textit{bagging}  can improve the performance of unstable predictors. That is, for models produced by a learning process that is sensitive to the underlying training data, it is (theoretically-grounded) good practice to train an ensemble of models using bootstrapping (Appendix~\ref{app:sec:prelim:boot}; Efron~\cite{efron1979bootstrap}; Efron and Tibshirani~\cite{efron1993bootsrap}). When classifying an example instance, we then leverage the whole ensemble by aggregating the predictions produced by its members. This aggregation process identifies the most common prediction in the ensemble, and returns that label as the classification. Put differently, we have combined the information of a lot of unstable classifiers, and averaged over their behavior in order to generate more stable classifications. 

Given the the relationship between variance (Definition~\ref{def:variance}) and self-consistency (Definition~\ref{def:sc}), reducing variance will improve self-consistency. However, rather than relying on a simple-majority-vote to decide the aggregated prediction, we also  will instill a notion of confidence in our predictions by requiring a minimum level of self-consistency, which is described in Algorithm~\ref{algo:bagging-confidently}.

\subsection{Self-consistent ensembling with abstention}\label{app:sec:algo:sc}

We present a framework that alters the semantics of classification outputs to $0$, $1$, and \texttt{Abstain}, and employ ensembling to determine the $\hatsc$-level that guides the output process. We modify bagging from using a simple-majority-vote because this type of aggregation rule still allows for arbitrariness. If, for example, we happen to train $\boot=101$ classifiers, it is possible that 50 of them yield one classification and the other 51 yield the other classification for a particular example. Bagging would select the classification that goes along with the 51 underlying models; however, if we happened to train $\boot=103$ models, it is perhaps the case that the majority vote would flip. In short, the bagging aggregation rule bakes in the idea that simple-majority voting is a sufficient strategy for making decisions. And while this may generally be true for variance reduction in high-variance classifiers, it does not address the problem of arbitrariness that we study. It just encodes arbitrariness in the aggregation rule --- it picks classifications, in some cases, that are no better than a coin flip.

Instead, Algorithm~\ref{algo:bagging-confidently} is more flexible. It suggests many possible ways to produce bagged classifiers that do not have to rely on simple-majority voting, by allowing for abstentions. For example, we can change the aggregation rule in regular bagging to use a self-consistency level $\kappa$ rather than majority vote. Instead of relying on votes, we can bag the underlying prediction probabilities and then apply $\kappa$ a filter. We could take the top-$n$ most consistent predictions and let a super-ensemble of underlying bagged classifiers decide whether to abstain or predict.

In the experiments in the paper, we provide two examples: Changing the underlying bagging vote aggregation rule (simple ensembling), and applying a round of regular bagging to do variance reduction and then bagging the bagged outputs (super ensembling) to apply a self-consistency threshold. Our ensemble model will not produce predictions for examples for which the lack of self-consistency is too high. We describe our procedure more formally in Algorithm~\ref{algo:bagging-confidently}.

\paragraph{Simple proof that abstention improves self-consistency (by construction).} We briefly show the simple proof that any method that meets the semantics of Algorithm~\ref{algo:bagging-confidently} will be more self-consistent than its counterpart that cannot \texttt{Abstain}. 

We define abstentions to be in agreement with both $0$ and $1$ predictions. This makes sense intuitively: Algorithm~\ref{algo:bagging-confidently} abstains to avoid making predictions that lack self-consistency, so abstaining should not increase disagreement between predictions. 

It follows that we can continue to use Definition~\ref{def:sc} and associated empirical approximations $\hatsc$ (\ref{eq:hatsc}), 
but with one small adjustment. Instead of the total number of predictions $\boot = \boot_0 + \boot_1$, with $\boot_0$ and $\boot_1$ corresponding to $0$ and $1$ predictions, respectively, we now allow for $\boot \geq \boot_0 + \boot_1$, in order to account for possibly some non-zero number of abstentions. 

In more detail, let us denote $\hat{\sY}$ to be the multiset of predictions for models $\model_{\dataset_1}, \model_{\dataset_2}, \ldots, \model_{\dataset_\boot}$ on $(\instance, \group)$, with $|\hat{\sY}| = \boot = \boot_0 + \boot_1 + \boot_{\texttt{Abstain}}$. This is where we depart from our typical definition of self-consistency, for which $\boot= \boot_0 + \boot_1$ (Section~\ref{sec:fairness:significance}, Appendix~\ref{app:sec:consistency}). We continue to let $\boot_0$ and $\boot_1$ represent the counts of $0$ and $1$ predictions, respectively, and now include $\boot_{\texttt{Abstain}}$ to denote the (possibly nonzero) number of abstentions. This leads to the following adjustment of (\ref{eq:hatsc}): \looseness=-1
\begin{align}
	\label{eq:sc-abstain}
	\hatconsistency = 1 - \frac{2(\boot_0\boot_1 + \boot_0\boot_{\texttt{Abstain}} + \boot_1\boot_{\texttt{Abstain}})}{\boot(\boot-1)}.
\end{align}

\noindent Equation (\ref{eq:sc-abstain}) follows from a similar analysis of comparing $0$s, $1$s, and abstentions for Definition~\ref{def:sc}, which lead us to derive (\ref{eq:hatsc}) in Appendix~\ref{app:sec:consistency}. However, since the costs of $0$-to-\texttt{Abstain} comparisons and $1$-to-\texttt{Abstain} comparisons are both 0, the $\boot_0\boot_{\texttt{Abstain}}$ and $\boot_1\boot_{\texttt{Abstain}}$ terms in (\ref{eq:sc-abstain}) reduce to 0. As a result, we yield our original definition for self-consistency (\ref{eq:hatsc}), with the possibility that $\boot=\boot_0 + \boot_1 + \boot_{\texttt{Abstain}} > \boot_0 + \boot_1$, if there is a nonzero number of abstentions $\boot_{\texttt{Abstain}}$. 

Since $\boot>1$ and $\boot_0, \boot_1, \boot_{\texttt{Abstain}} \ge 0$, it is always the case that option to \texttt{Abstain} is at least as self-consistent as not having the option to do so. This follows from the fact that $\boot_0 + \boot_1 + \boot_{\texttt{Abstain}} = \boot \geq \boot_0 + \boot_1$, which would make the denominator in (\ref{eq:sc-abstain}) greater than or equal to the corresponding method that cannot \texttt{Abstain}; when subtracted from 1, this would produce a $\hatsc$ that is no smaller than the value for the corresponding method without that cannot \texttt{Abstain}.

Now, it follows that, given the choice between \texttt{Abstain} and predicting a label that is in disagreement with an existing prediction label, choosing to \texttt{Abstain} will always lead to higher self-consistency. This is because the cost to \texttt{Abstain} is less than disagreeing, so it will always be the minimal choice that maximizes $\hatsc$. 



\paragraph{Error and the abstention set.} It is very straightforward to see that the \emph{abstention set} will generally exhibit higher than the \emph{prediction set}. When we ensemble and measure $\hatsc$, the exmaples that exhibit low $\hatsc$ contain higher variance-induced error. Let us call the size of the abstention set $U$ (which incurs error $u$), the size of the prediction set $V$ (which incurs error $v$), and the size of the test set $T$ (which incurs error $t$). We can relate the total number of misclassified examples as $T * t = U * u + V * v,$ with $T = U + V$. If we assume the bias and noise are equally distributed across the test and abstention sets (this is a reasonable assumption, on average, in our setup), then splitting off the high variance instances from the low variance (high $\hatsc$ instances) requires that $u > v$. The error on the abstention set necessarily has to be larger than the error on the prediction set, in order to retain the above relationship. 

%% file: section/99-appendix/22-fairness/500-app-fairness-empirical.tex
\section{\appexperiments}\label{app:sec:empirical}

The code for the examples in Sections~\ref{sec:fairness:intro}, \ref{sec:fairness:significance} and \ref{sec:fairness:empirical} can be found in through our paper on arXiv. 
This repository also contains necessary and sufficient information concerning reproducibility. At the time of writing, we use \texttt{Conda} to produce environments with associated package-versioning information, so that our results can be exactly replicated and independently verified. We also use the \texttt{Scikit-Learn}~\citep{pedregosa2011scikit} toolkit for modeling and optimization. More details on our choice of models and hyperparameter optimization can be found in our code repository, cited above. In brief, we consulted prior related work (e.g., Chen et al.~\cite{chen2018tradeoff}) and performed our own validation for reasonable hyperparameters per model type. We keep these settings fixed to reduce impact on our results, in order to observe in isolation how different training data subsets impact our results. 

During these early runs, we collected information on train accuracy, not just test accuracy; while models ultimately have similar test accuracy in most cases for the same task, they can vary significantly in terms of train accuracy (e.g., for logistic regression, \texttt{COMPAS} is in the low .70s; for random forests, it is in the mid .90s). We do not include these results for the sake of space.\looseness=-1 

This section is organized as follows. We first present information on our datasets, models and code, including our \texttt{HDMA} toolkit (Appendix~\ref{app:sec:experimentsdata}). We then provide details on our setup for running experiments on our cluster (Appendix~\ref{app:sec:cluster}). Appendix~\ref{app:sec:illustrativedetails} contains  more detailed information concerning the experiments performed to produce Figures~\ref{fig:vote} and~\ref{fig:adult-compas-cdf-rfc} in the main paper. 
In Appendix~\ref{app:sec:fair}, we discuss implications of these results for common fairness benchmarks like \texttt{South German Credit}.
We defer more extensive results on our ensembling algorithm to our online appendix.

\paragraph{Note on CDF figures.} 
We show our results in terms of the $\hatsc$ of the underlying bagged models because doing so conveys how Algorithm~\ref{algo:bagging-confidently} makes decisions to predict or abstain.\footnote{The $\hatsc$ CDF of Algorithm~\ref{algo:bagging-confidently}, computed via a \emph{third} round of bootstrapping, has nearly all mass at $\hatsc=1$; it is difficult to visualize.} For both types of ensembling, Algorithm~\ref{algo:bagging-confidently} predicts for all examples captured by the area to the right of the $\kappa$ reference line, and abstains for all examples on the left.\looseness=-1 

\paragraph{A remark on cost} It can be considerably more computationally intensive to train an ensemble of models to compute $\hatsc$ than to train a handful of models and perform cross-validation, as is the standard practice in fair classification. 
However, as our empirical analysis demonstrates, 
this cost comes with a huge benefit: It enables us to improve self-consistency and to root out the arbitrariness of producing predictions that are effectively close-to-random, which is especially important in high-stakes fairness settings~\citep{cooper2021eaamo}. Moreover, for common fair classification datasets, the increased cost on modern hardware is relatively small; 
(super-) ensembling with confidence takes under an hour to execute (online appendix).

\input{section/99-appendix/22-fairness/510-datasetsandmodels}
\input{section/99-appendix/22-fairness/520-cluster}
\input{section/99-appendix/22-fairness/530-illustrative-main}
\input{section/99-appendix/22-fairness/550-fair}

%% file: section/99-appendix/22-fairness/510-datasetsandmodels.tex
\subsection{\appexperimentsdata}\label{app:sec:experimentsdata}

\paragraph{Models.} According to a comprehensive recent survey study~\cite{fabris2022datasets}, as well as related work like Chen et al.~\cite{chen2018tradeoff}, we conclude that some of the most common models used in fair classification are logistic regression, decision tree classifiers, random forest classifiers, SVMs, and MLPs. We opted to include comprehensive results for the first three, since they capture different complexities, and therefore encode different degrees of statistical bias, that we expected to have an impact on the underlying sources of error. 
Since we choose not to use stochastic optimizers to reduce the sources of randomness, for our results, training MLPs is slower than it could be. We consistently use a decision threshold of 0.5 (i.e., 0-1 loss) for our experiments, though our results can easily be extended to other thresholds, as discussed in Section~\ref{sec:fairness:significance}. Depending on the dataset, we reserve between 20\% and 30\% of the available data for the test set. This is consistent with standard fair classification training settings, which we validated during our initial experiments to explore the space (for which we also did preliminary hyperparameter optimization, before fixing the hyperparameters for our presented results).\footnote{Please refer to the arXiv paper version for more details.}

\paragraph{Datasets.}  Also according to Fabris et al.~\cite{fabris2022datasets}, the most common tasks in fair classification are \texttt{Old Adult}~\cite{kohavi1996oldadult}, \texttt{COMPAS}~\cite{larson2016propublica}, and \texttt{South German Credit}~\cite{gromping2019german}.\footnote{Technically, Grömping~\cite{gromping2019german} is an updated and corrected version of the dataset from 2019.} These three datasets arguably serve as a \emph{de facto} benchmark in the community, so we felt the need to include them in the present work. 
In recognition of the fact that these three datasets, however standard, have problems, we also run experiments on $3$ tasks in the \texttt{New Adult} dataset, introduced by Ding et al.~\cite{ding2021adult} to replace \texttt{Old Adult}. We subset to the \texttt{CA} (California) subset of the dataset, and run on \texttt{Income}, \texttt{Employment}, and \texttt{Public Coverage}, and consider \texttt{sex} and \texttt{race} as protected attributes, which we binarize into \{Male, Female\} and \{White, Non-white\}. These are all large-scale tasks, at least in the domain of algorithmic fairness --- on the order of hundreds of thousands of example instances. However, the $3$ tasks do share example instances and some features. In summary, concerning common tasks in fair classification:
\begin{itemize}
\item \texttt{COMPAS}~\cite{larson2016propublica}. We run on the commonly-used version of this dataset from Friedler et al.~\cite{friedler2019datasets}, which has 6167 example instances with 404 features. The target is to predict recidivism within 2 years ($1$ corresponding to Yes, and $0$ to No). The protected attribute is \texttt{race}, binarized into ``Non-white'' ($0$) and ``White'' ($1$) subgroups.

\item \texttt{Old Adult}~\cite{kohavi1996oldadult}. We run on the commonly-used version of this dataset from Friedler et al.~\cite{friedler2019datasets}, which has 30,162 examples with 97 features. This version of the dataset removes instances with missing values from the original dataset, and changes the encoding of some of the features (Kohavi~\cite{kohavi1996oldadult} has 48842 example instances with 88 features). The target is to predict $<\$50,000$  income ($0$) $>=\$50,000$ income ($1$). The protected attribute is \texttt{sex}, binarized into ``Female'' ($0$) and ``Male'' ($1$) subgroups.

\item \texttt{South German Credit}~\cite{gromping2019german}. We download the dataset from UCI\footnote{\texttt{https://archive.ics.uci.edu/ml/datasets/South+German+Credit+\%28UPDATE\%29}} and process the data ourselves. We use the provided \texttt{codetable.txt} to ``translate'' the features from German to English. We say ``translate'' because the authors took some liberties, e.g., the column converted to ``credit\_history'' is labeled ``moral'' in the German, which is not a translation. There are four categories in the protected attribute ``personal\_status\_sex'' column, one of which ($2$) is used for both ``Male (single)'' and ``Female (non-single).'' We therefore remove rows with this value, and binarize the remaining three categories into ``Female'' ($0$) and ``Male'' ($1$). What results is a dataset with 690 example instances (of the original 1000) with 19 features. The target is ``good'' credit ($1$) and ``bad'' credit ($0$).

\item \texttt{Taiwan Credit}~\cite{yeh2009taiwan}. This task is to predict default on credit card payments ($1$) or not ($0$). There are 30,000 example instances and 24 features. The protected attribute is binary \texttt{sex}. We download this dataset from UCI.\footnote{See  \texttt{https://archive.ics.uci.edu/ml/datasets/default+of+credit+card+clients}}.

\item \texttt{New Adult}~\cite{ding2021adult}. This dataset contains millions of example instances from US Census data, which can be used for several different targets/tasks. We select three of them (listed below). These tasks share some features, and therefore are not completely independent. Further, given the size of the whole dataset, we subset to  \texttt{CA} (California), the most populous state in the US. There are two protected attribute columns that we use: \texttt{sex}, which is binarized ``Female'' ($0$) and ``Male'' ($1$) subgroups, and \texttt{race}, which we binarize into ``Non-white'' ($0$) and ``White'' ($1$). In future work, we would like to explore extending our results beyond binary subgroups.
\begin{itemize}
    \item \texttt{Income}. This task is designed to be analogous to \texttt{Old Adult}~\cite{kohavi1996oldadult}. As a result, the target is to predict  $<\$50,000$  income ($0$) $>=\$50,000$ income ($1$). In the \texttt{CA} subset, there are 195,665 example instances with 8 features.
    \item \texttt{Employment}. This task is to predict whether an individual is employed ($1$) or not ($0$). In the \texttt{CA} subset, there are 378,817 example instances with 14 features. 
    \item \texttt{Public Coverage}. This task is to predict whether an individual is on public health insurance ($1$) or not ($0$). In the \texttt{CA} subset, there are 138,554
    example instances with 17 features.

\end{itemize}
\end{itemize}

\subsubsection{\apphmda}\label{app:sec:hmda}
In addition to the above standard tasks, we include experiments that use the \texttt{NY} and \texttt{TX} 2017 subsets of the the Home Mortgage Data Disclosure Act (\texttt{HMDA}) 2007-2017 dataset~\cite{ffiec2022housingdata}. These two datasets have 244,107 and 576,978 examples, respectively, with 18 features. The \texttt{HMDA} datasets together contain over 140 million examples of US home mortgage loans from 2007-2017 (newer data exists, but in a different format). We developed a toolkit, described below, to make this dataset easy to use for classification experiments. Similar to \texttt{New Adult}, we enable subsetting by US state. For the experiments in this paper, we run on the \texttt{NY} (New York) and \texttt{TX} (Texas) 2017 subset, in order to add some geographic diversity to complement our \texttt{New Adult} experiments. We additionally chose \texttt{NY} and \texttt{TX} because they are two of the most populous states in the US, alongside \texttt{CA}.\footnote{Per the 2020 Census, the top-4-most-populous states are \texttt{CA}, \texttt{TX}, \texttt{FL}, and \texttt{NY}~\cite{mackun2021census}.}

The target variable, \texttt{action\_taken}, concerning loan origination has 8 values, 2 of which we cannot meaningful conclude approval or denial decisions. They are: Action Taken: $1$ -- Loan originated, $2$ -- Application approved but not accepted, $3$ -- Application denied by financial institution, $4$ -- Application withdrawn by applicant, $5$ -- File closed for incompleteness, $6$ -- Loan purchased by the institution, $7$ -- Preapproval request denied by financial institution, and $8$ -- Preapproval request approved but not accepted (optional reporting). We filter out $4$ and $6$, and binarize into \texttt{grant}=$\{1, 2, 8\}=1$ and \texttt{reject}=$\{3, 5, 7\}=0$. There are three protected attributes that we consider: \texttt{sex}, \texttt{race}, and \texttt{ethnicity}:
\begin{itemize}
    \item \texttt{sex} has 5 possible values, 2 of which correspond to categories/non-missing values: Male -- $1$ and Female -- $2$. We binarize \texttt{sex} into $\text{F}=0$ and $\text{M}=1$. 
    \item \texttt{race} has 8 possible values, 5 of which correspond to categories/ non-missing information: $1$ -- American Indian or Alaska Native, $2$ -- Asian, $3$ -- Black or African American, $4$ -- Native Hawaiian or Other Pacific Islander, and $5$ -- White. There are 5 fields for applicant race, which model an applicant belonging to more than one racial group. For our experiments, we only look at the first field. When we binarize \texttt{race}, $\text{NW}=0$ and $\text{W}=1$. 
    \item \texttt{ethnicity} has 5 possible values, 2 of which correspond to categories/ non-missing information: $1$ -- Hispanic or Latino and $2$ -- Not Hispanic or Latino. We binarize \text{ethnicity} to be $\text{HL}=0$ and $\text{NHL}=1$. 
\end{itemize}

After subsetting to only include examples that have values that do not correspond to missing information, \texttt{HMDA} has 18 features. The \texttt{NY} dataset has 244,107 examples; the \texttt{TX} dataset has 576,978 examples, making it the largest dataset in our experiments. As with our experiments using \texttt{New Adult}, we would like to extend our results beyond binary subgroups and binary classification in future work.

\paragraph{Releasing a standalone toolkit.} These datasets are less-commonly used in current algorithmic fairness literature~\cite{fabris2022datasets}. We believe this is likely due to the fact that the over-100-million data examples are only available in bulk files, which are on the order of 10s of gigabytes and therefore not easily downloadable or explorable on most personal computers. Following the example of Ding et al.~\cite{ding2021adult}, one of our contributions is to pre-process all of these datasets --- all locations and years --- and release them with a software toolkit. The software engineering effort to produce this toolkit was substantial. Our hope is that wider access to this dataset will further reduce the community's dependency on small (and dated) datasets. Please refer to the arXiv paper for the latest information on this standalone software package. Our release aligns with the terms of service for this dataset.\looseness=-1

%% file: section/99-appendix/22-fairness/520-cluster.tex
\subsection{\appcluster}\label{app:sec:cluster}

While most of the experiments run in this paper can be easily reproduced on a modern laptop, for efficiency, we ran all of our experiments (except the one to produce Figure~\ref{fig:vote}) in a cluster environment. This enabled us to easily execute train/test splits $n$ in parallel on different CPUs, serialize our results, and then reconstitute and combine them to produce plots locally. Our cluster environment runs Ubuntu 20.04 and uses Slurm v20.11.8 to manage jobs. We ran all experiments using \texttt{Anaconda3}, which is why we used \texttt{Conda} to reproduce environments for easy replicability. 

The experiments using \texttt{New Adult} and \texttt{HMDA} rely on datasets that are (in some cases) orders of magnitude larger than the traditional algorithmic fairness tasks. This is one of the reasons why we recommend running on a cluster, and therefore do not include Jupyter notebooks in our repository for these tasks. We also limit our modeling choices to logistic regression, decision tree classifiers, and random forest classifiers for these results due to the expense of training on the order of thousands of models for each experiment. 

%% file: section/99-appendix/22-fairness/530-illustrative-main.tex
\subsection{\appillustrativedetails}\label{app:sec:illustrativedetails}

This appendix provides extended results for the experiments associated in Sections~\ref{sec:fairness:intro} and \ref{sec:fairness:significance}, which give an intuition for individual- and subgroup-level consistency. The experimental results in the main paper are for logistic regression. We expand the set of models we examine, and  associated discussion of how to interpret comparisons between these results. 

\paragraph{Reproducing Figure~\ref{fig:vote}.} The experiment to produce this figure in Section~\ref{sec:fairness:intro} (also shown in Appendix~\ref{app:sec:consistency}) trains $\boot=10$ logistic regression models on the \texttt{COMPAS} dataset (Appendix~\ref{app:sec:experimentsdata}) using 0-1 loss. We use the bootstrap method to produce each model, which we evaluate on the same test set. We then search for a maximally consistent and minimally consistent individual in the test set, i.e., an individual with $10$ predictions that agree and an individual with $5$ predictions in each class, which we plot in the bar graph. Please refer to the README in the code repository (see arXiv paper) regarding which \texttt{Jupyter} notebook to run to produce the underlying results and figure. The experiments to reproduce this figure can be easily replicated on a laptop.\looseness=-1 

\paragraph{Reproducing Figure~\ref{fig:adult-compas-cdf-rfc}.} These figures were produced by executing $S=10$ runs of $\boot=101$ bootstrap training replicates to train random forest classifiers for \texttt{Old Adult} and \texttt{COMPAS}. We reproduce these figures below, so that they can be examined and treated in relation to our additional results for decision tree classifiers and logistic regression. For each $s$ run, we take train/test split, bootstrap the train split $\boot=101$ times, and evaluate the resulting model classification decisions on the test set. $\hatsc$ can be estimated from the results across those $101$ models. We Run this process $S=10$ times to produce confidence intervals, shown in the figures below. The intervals are not always clearly visible; there is not a lot of variance at the level of comparing whole runs to each other. Please refer to the README in the code repository (see arXiv paper) regarding which \texttt{Jupyter} notebook to run to produce the underlying results and figure. There are also scripted version of these experiments, which enable them to be run in parallel in a cluster environment.

\paragraph{Self-consistency of incorrectly-classified instances.} Last, we include figures that underscore how self-consistency is independent from correctness that is measured in terms of observed label alignment. That is, it is possible for an instance $(\instance, \group)$ to be self-consistent and classified incorrectly, with respect to its observed label $\olabel$. We show this using stacked bar plots. For the above experiments, we find the test examples that have the majority of their classifications incorrect ($\pred\neq \olabel$, for $\boot=101$, we find the instances with $\ge 51$ incorrect classifications) and the majority of their classification correct (similarly), and we examine how self-consistent they are. We bucket self-consistency into different levels, and then plot the relative proportion of majority-incorrectly and majority-correctly classified examples according to subgroup. Subgroups in \texttt{COMPAS} exhibit a similar trend, while subgroups in \texttt{Adult Old} exhibit differences, with the heights of the bars corresponding to the trends we plot in our CDF plots. As we note briefly in Section~\ref{sec:fairness:significance}, it may be interesting to examine patterns in examples about which learning processes are confident (i.e., highly self-consistent) but wrong in terms of label alignment. If such issues correlate with subgroup, it may be worth testing the counterfactual that such labels are indicative of label bias. We leave such thoughts to future work. 

\begin{figure*}[!t]
    \begin{center}
    \begin{minipage}{\textwidth}
        \centering
        \includegraphics[width=0.9\linewidth]{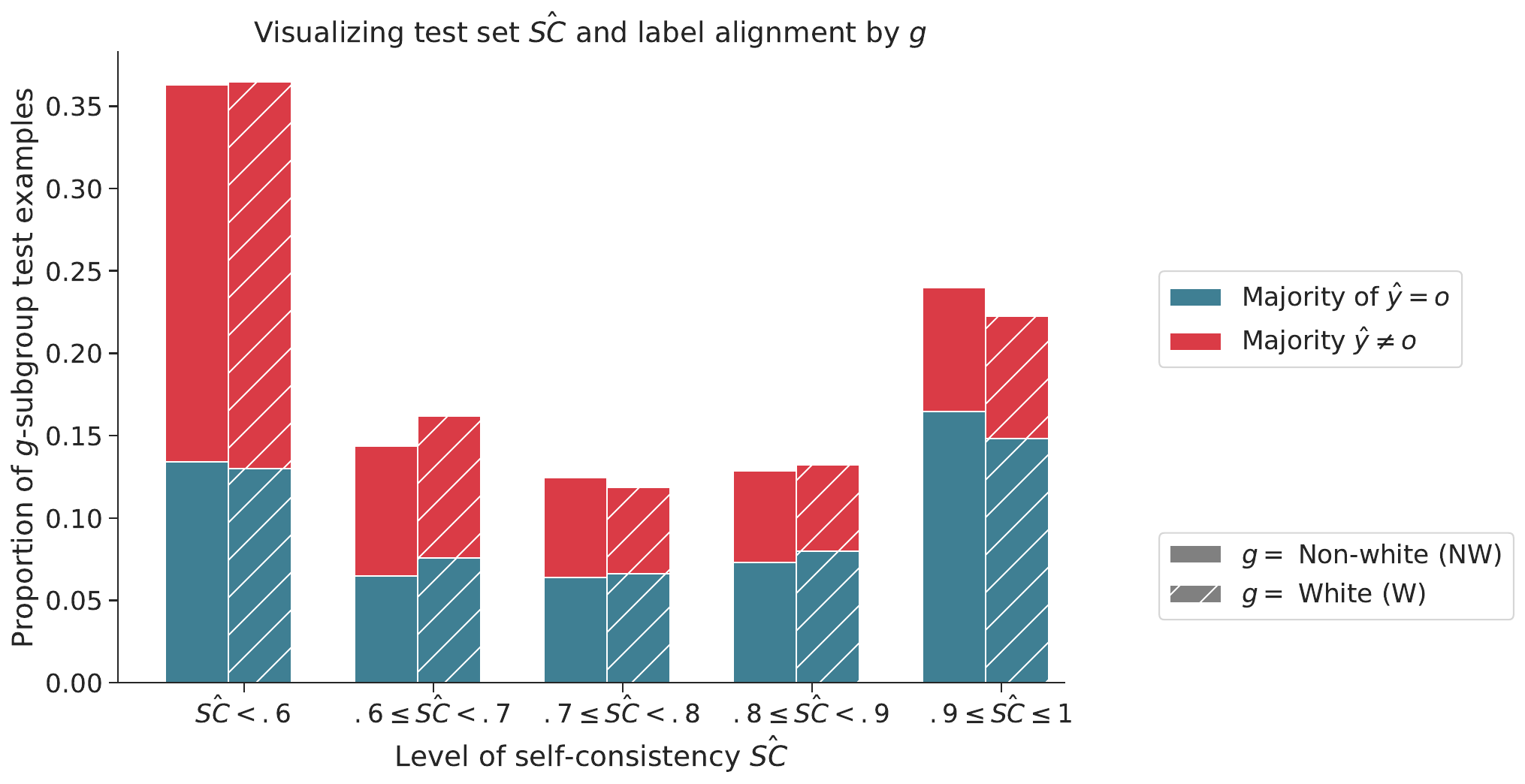}
        \subcaption{\texttt{COMPAS}}
        \label{app:subfig:compas-sc-acc}
        \includegraphics[width=0.9\linewidth]{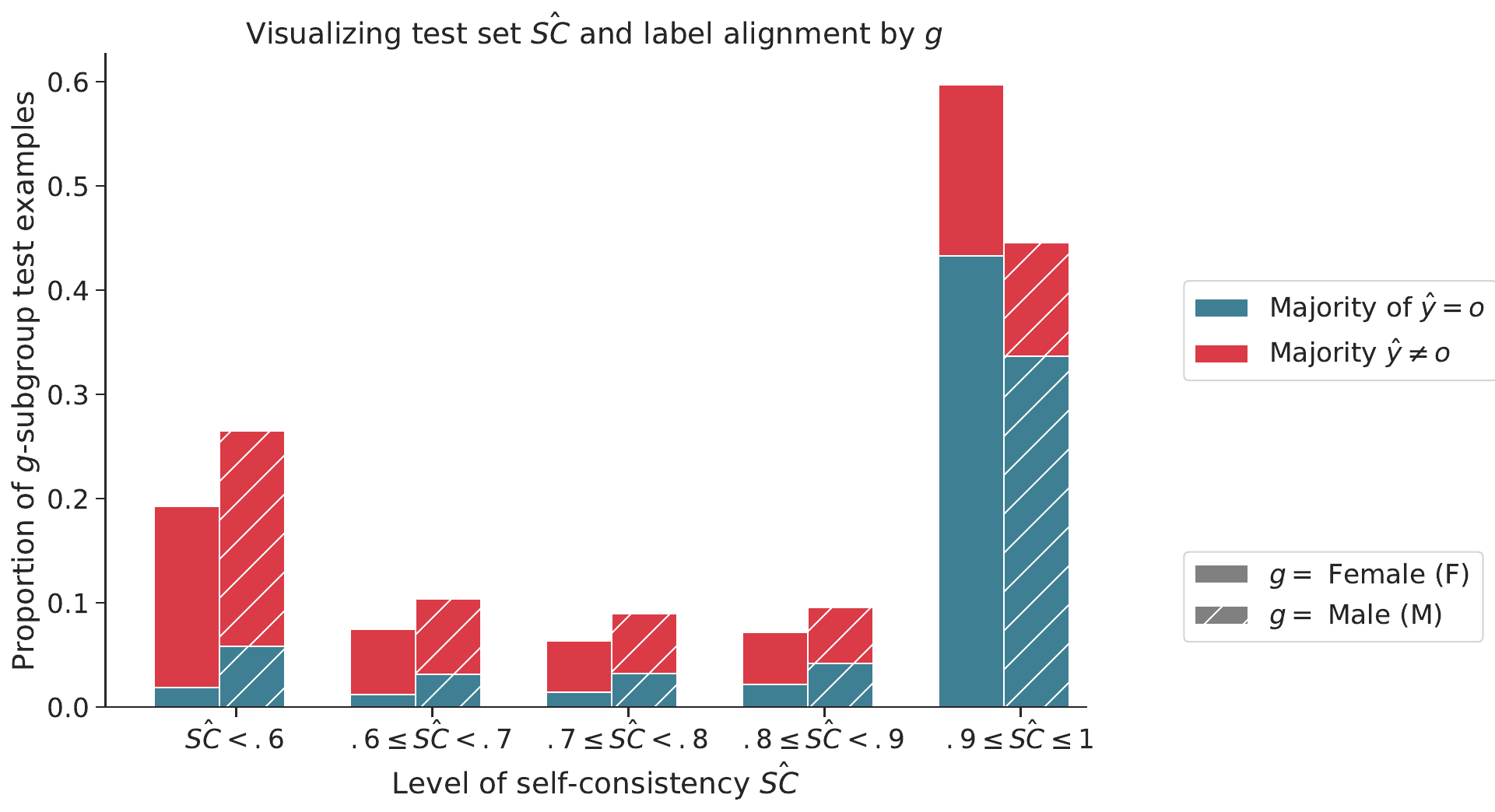}
        \subcaption{\texttt{Adult Old}}
        \label{app:subfig:adult-old-sc-acc}
    \end{minipage}%
    \end{center}
    \vspace{-.35cm}
    \caption{$\hatsc$ broken down by $\group$ and label alignment with the observed label $\olabel$. For each train/test split, and for each $\hatsc$ range ($x$-axis), we find the examples that are incorrectly classified the majority of time ($\geq 5$ splits, we find that $\pred \ne \olabel$), and the examples that are correctly classified the majority of the time ($> 5$, we find that $\pred = \olabel$). We compute the average the proportion over (over splits) in each $\hatsc$ range ($y$-axis). We plot these proportions with respect to subgroup $\group$ (where the sums of the heights of bars for by each $\group$ is equal to $1$).\looseness=-1}
\end{figure*}
\FloatBarrier 
\pagebreak

%% file: section/99-appendix/22-fairness/550-fair.tex
\subsection{\appfair}\label{app:sec:fair}

Even before we apply our intervention to improve self-consistency, our results in Section~\ref{sec:fairness:significance} show close-to-parity $\haterr$, $\hatfpr$, and $\hatfnr$ across subgroups in \texttt{COMPAS} (and similarly for \texttt{South German Credit}, below). These results are surprising. We run $\boot=101$ models to produce estimates of variance and self-consistency, but of course doing this also has the effect of estimating the expected error more generally (with variance representing a portion of that error). 

Our estimates of expected error for these tasks indicate that the average model produced training on \texttt{COMPAS} and \texttt{South German Credit}, with respect to popular fairness definitions like Equality of Opportunity and Equalized Odds~\cite{barocas2019textbook,hardt2016equality} are in fact baseline close to parity, with no fairness intervention applied. We found this across model types for both datasets, though the story becomes more complicated when we apply techniques to improve self-consistency (see arXiv appendix). 

\begin{figure*}[!t]
    \centering
    \begin{minipage}{0.33\textwidth}
        \centering
        \includegraphics[width=0.95\linewidth]{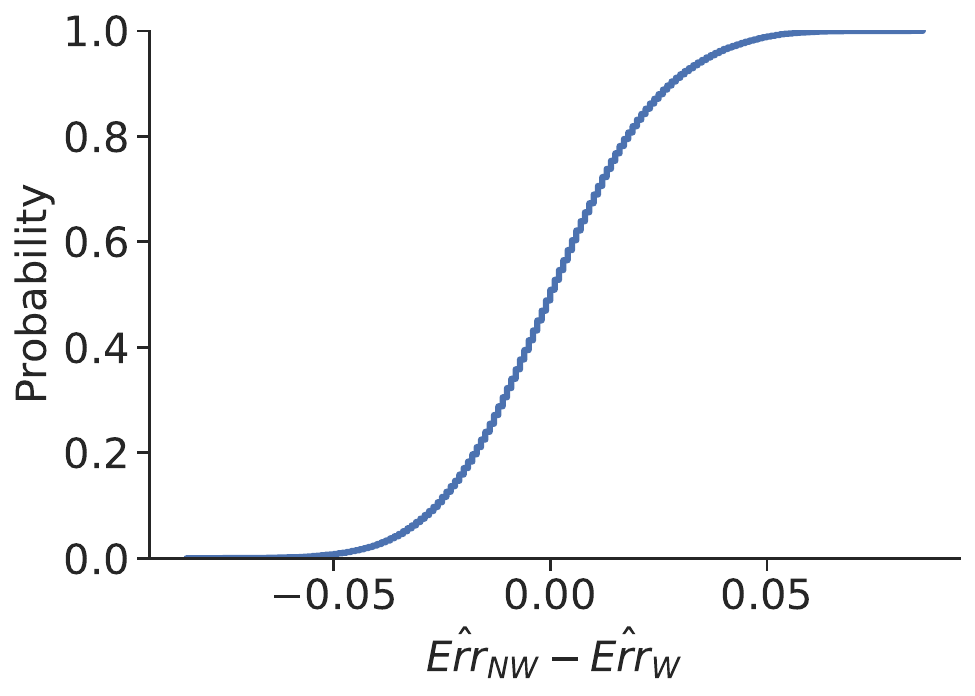}
        \subcaption{$\haterr$ disparity}
    \end{minipage}%
    \begin{minipage}{0.33\textwidth}
        \centering
        \includegraphics[width=0.95\linewidth]{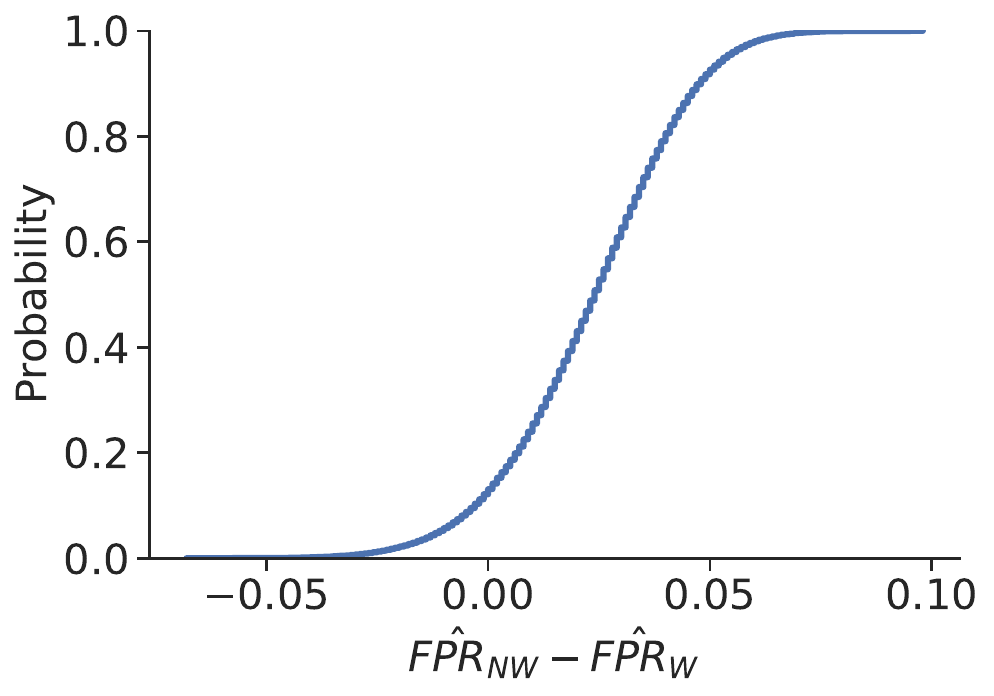}
        \subcaption{$\hatfpr$ disparity}
    \end{minipage}%
    \begin{minipage}{0.33\textwidth}
        \centering
        \includegraphics[width=0.95\linewidth]{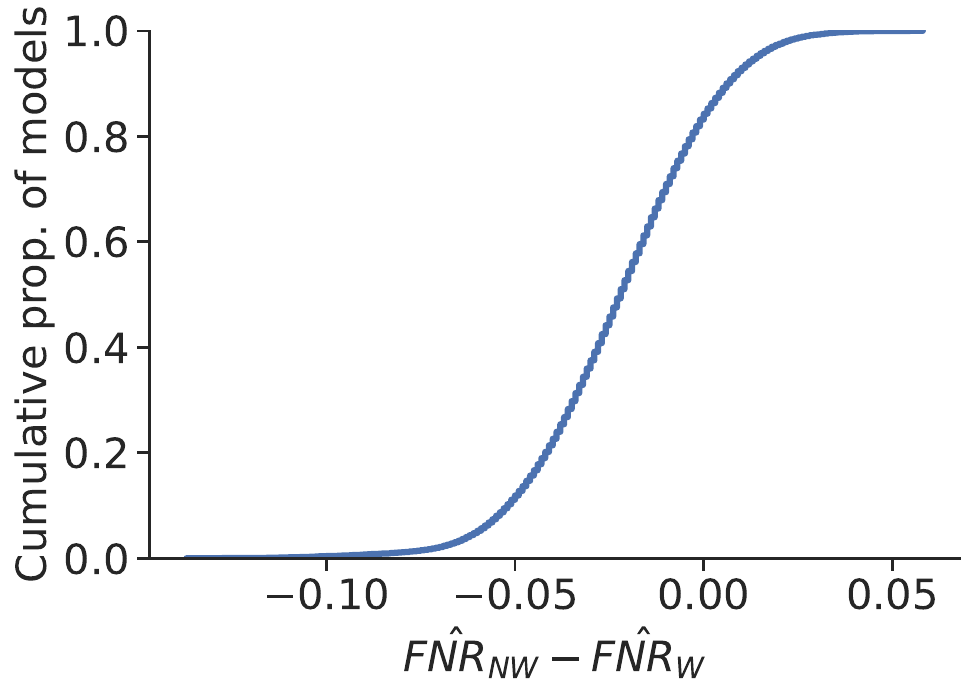}
        \subcaption{$\hatfnr $ disparity}
    \end{minipage}
    \caption{Cumulative distribution of error disparity across $100,100$ logistic regression models trained on \texttt{COMPAS}.}
    \label{app:fig:compas-disparity-ecdf}
    \begin{minipage}{0.33\textwidth}
        \centering
        \includegraphics[width=0.95\linewidth]{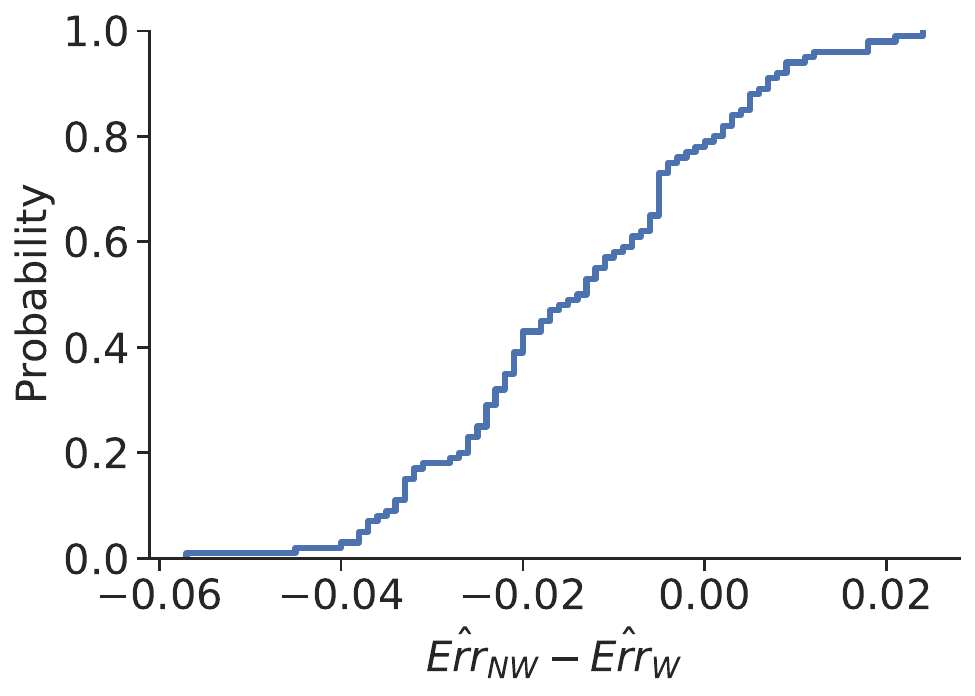}
        \subcaption{$\haterr$ disparity}
    \end{minipage}%
    \begin{minipage}{0.33\textwidth}
        \centering
        \includegraphics[width=0.95\linewidth]{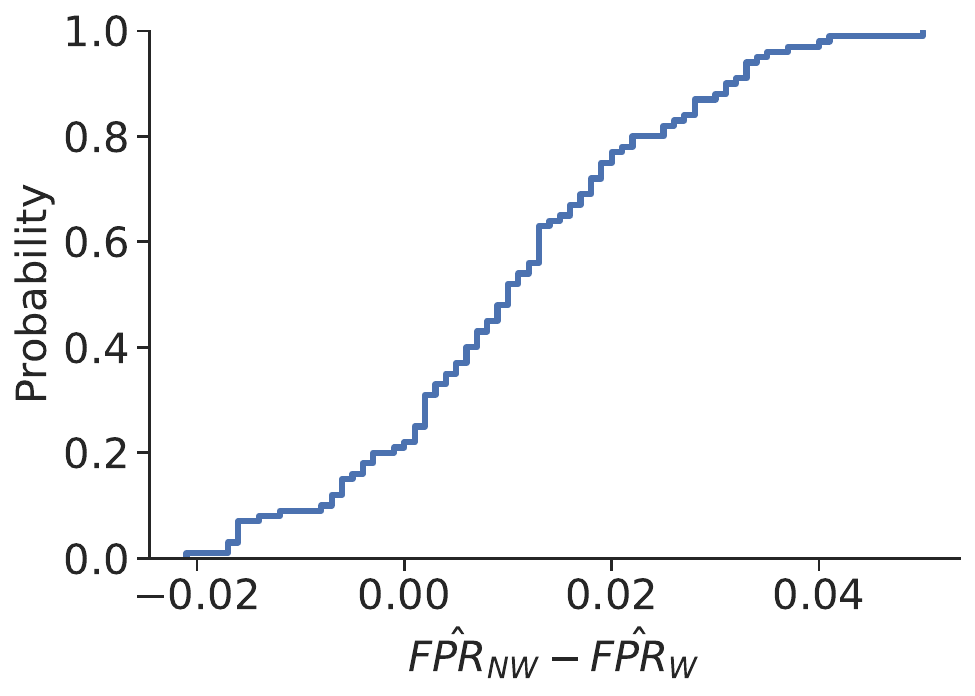}
        \subcaption{$\hatfpr$ disparity}
    \end{minipage}%
    \begin{minipage}{0.33\textwidth}
        \centering
        \includegraphics[width=0.95\linewidth]{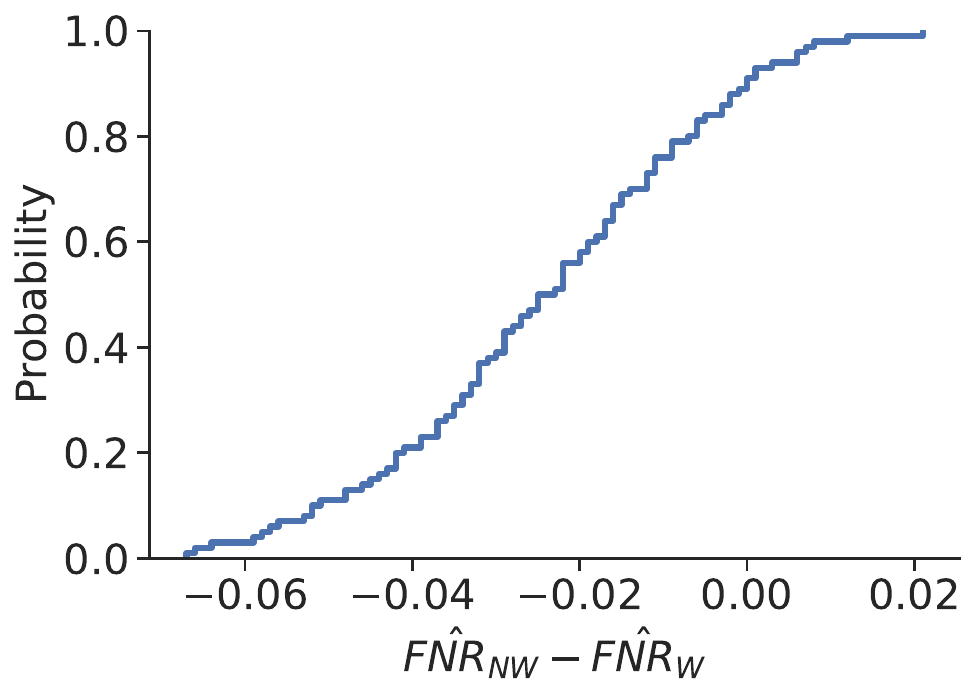}
        \subcaption{$\hatfnr $ disparity}
    \end{minipage}
    \caption{CDF of error disparity across the top $100$ logistic regression models (of the $100,100$ models) trained on \texttt{COMPAS}.}
    \label{app:fig:compas-disparity-ecdf-100}
\end{figure*}

We did not expect this result, as these are two of the \textit{de facto} standard benchmark datasets in algorithmic fairness. They are used in countless other studies to probe and verify algorithmic fairness interventions~\cite{fabris2022datasets}. As a result, we initially thought that our results must be incorrect. We therefore looked at the underlying models in our bootstrap runs to see the error of the underlying models.
We re-ran our baseline experiments with $\boot=1001$ and for $100$ test/train splits for logistic regression. In Figure~\ref{app:fig:compas-disparity-ecdf}, we plot the ($100,100$) bootstrap models that went into these results. For another view on analogous information, in Table~\ref{app:table:compas-model-runs-diffs}, we provide an excerpt of the results for \texttt{COMPAS} regarding the underlying $1010$ random forest classifiers used to produce Figure~\ref{subfig:compas-cdf-rfc}.\looseness=-1

Overall, we can see that there is a wide range of error disparities that trend in both directions, with a skew toward higher $\hatfpr$ for $\group=\text{NW}$. These results support our claim that training many models is necessary to get an accurate picture of expected error, with implications both for reproducibility of experiments that just train and analyze a small handful of models and for generalizability. There are models that exhibit worse degrees of unfairness in both directions, but they are more unlikely than models that exhibit smaller disparities.

We subset the above results to the $100$ models that produce the lowest $\haterr$, as this is often the selection criteria for picking models to post-process. We plot these results below. These top-performing models in fact exhibit (on average) closer-to-parity for $\hatfpr$ and $\hatfnr$. 

\newpage
\begin{table}
\centering
\caption{\small{Comparing subgroup error rates in \texttt{COMPAS} for different random forest classifiers trained to produce Figure~\ref{subfig:compas-cdf-rfc}. Each table looks at the top-$3$ highest differences between subgroups for the specified metric: (a) $\haterr_\text{NW} - \haterr_\text{W}$, when $\haterr_\text{NW} > \haterr_\text{W}$; (b) $\haterr_\text{W} - \haterr_\text{NW}$, when $\haterr_\text{W} > \haterr_\text{NW}$; (c) $\hatfpr_\text{NW} - \hatfpr_\text{W}$, when $\hatfpr_\text{NW} > \hatfpr_\text{W}$; (d) $\hatfpr_\text{W} - \hatfpr_\text{NW}$, when $\hatfpr_\text{W} > \hatfpr_\text{NW}$; (e) $\hatfnr_\text{NW} - \hatfnr_\text{W}$, when $\hatfnr_\text{NW} > \hatfnr_\text{W}$; and, (f) (e) $\hatfnr_\text{W} - \hatfnr_\text{NW}$, when $\hatfnr_\text{W} > \hatfnr_\text{NW}$. We highlight the overall error metric in \textcolor{black!30}{gray}, the larger metric (being subtracted from) in \textcolor{blue!50}{blue}, the smaller metric (being subtracted) in \textcolor{red!50}{red}, and the difference in the metric between subgroups in \textcolor{purple!50}{purple}. Note that run 757 appears twice, which we mark in \textcolor{orange!50}{orange}.\looseness=-1}}
  \begin{subtable}[h]{\linewidth}
    \centering
    \subcaption{\footnotesize{Top-$3$ most unfair models by $\haterr$, when  $\haterr_\text{NW} > \haterr_\text{W}$ (i.e., unfair toward NW).}\looseness=-1}
    {\footnotesize
    \begin{tabular}{lllcccccccccc}
      \toprule
      {Run \#} & $s$ & $b$ & $\haterr$ & $\hatfpr$ & $\hatfnr$ & $\haterr_\text{NW}$ & $\hatfpr_\text{NW}$  & $\hatfnr_\text{NW}$ & $\haterr_\text{W}$  & $\hatfpr_\text{W}$ & $\hatfnr_\text{W}$ & $\haterr_\text{NW} - \haterr_\text{W}$ \\
      \midrule
      762 & $8$ & $504$ & \cellcolor{black!20}$0.374$ &	$0.179$ &	$0.196$	& \cellcolor{blue!33}$0.405$	& $0.204$ &	$0.201$	& \cellcolor{red!33}$0.315$	& $0.13$ &	$0.186$ &	\cellcolor{purple!33}$0.09$\\
      \cellcolor{orange!33}757	& $8$ &	$464$ &	\cellcolor{black!20}$0.369$ &	$0.167$	& $0.202$ &	\cellcolor{blue!33}$0.395$	& $0.201$ &	$0.193$ &	\cellcolor{red!33}$0.318$ &	$0.101$ &	$0.218$ &	\cellcolor{purple!33}$0.077$\\
      328 &	$4$ &	$116$ &	\cellcolor{black!20}$0.371$ &	$0.165$ &	$0.206$ &	\cellcolor{blue!33}$0.395$ &	$0.181$ &	$0.214$ &	\cellcolor{red!33}$0.323$ &	$0.134$ &	$0.189$ &	\cellcolor{purple!33}$0.072$\\
      \bottomrule
    \end{tabular}
    }%
    \label{app:subtable:errNW-minus-errW}
    \end{subtable}\\

    \vspace{0.25cm}
    \begin{subtable}[h]{\linewidth}
    \centering
    \subcaption{\footnotesize{Top-$3$ most unfair models by $\haterr$, when  $\haterr_\text{W} > \haterr_\text{NW}$ (i.e., unfair toward W).}\looseness=-1}
    {\footnotesize
    \begin{tabular}{lllcccccccccc}
      \toprule
      {Run \#} & $s$ & $b$ & $\haterr$ & $\hatfpr$ & $\hatfnr$ & $\haterr_\text{NW}$ & $\hatfpr_\text{NW}$  & $\hatfnr_\text{NW}$ & $\haterr_\text{W}$  & $\hatfpr_\text{W}$ & $\hatfnr_\text{W}$ & $\haterr_\text{W} - \haterr_\text{NW}$ \\
      \midrule
      414	& $5$ &	$75$ &	\cellcolor{black!20}$0.376$ &	$0.167$ & $0.209$ &	\cellcolor{red!33}$0.352$ &	$0.158$ &	$0.194$ &	\cellcolor{blue!33}$0.422$ &	$0.186$ &	$0.236$ &	\cellcolor{purple!33}$0.07$ \\
      435	& $5$ &	$180$ &	\cellcolor{black!20}$0.376$ &	$0.199$ & $0.177$ & \cellcolor{red!33}$0.355$ &	$0.189$ &	$0.166$ &	\cellcolor{blue!33}$0.416$	& $0.217$ &	$0.198$	& \cellcolor{purple!33}$0.061$\\
      413	& $5$ &	$70$ &	\cellcolor{black!20}$0.378$ &	$0.189$ &	$0.189$ &	\cellcolor{red!33}$0.359$ &	$0.188$ &	$0.171$ &	\cellcolor{blue!33}$0.413$ &	$0.191$ &	$0.222$ & \cellcolor{purple!33}$0.054$\\
      \bottomrule
    \end{tabular}
    }
    \label{app:subtable:errW-minus-errNW}
    \end{subtable}

    \vspace{0.25cm}
    \begin{subtable}[h]{\linewidth}
    \centering
    \subcaption{\footnotesize{Top-$3$ most unfair models by $\hatfpr$, when  $\hatfpr_\text{NW} > \hatfpr_\text{W}$ (i.e., unfair toward NW).}}
    {\footnotesize
    \begin{tabular}{lllcccccccccc}
      \toprule
      {Run \#} & $s$ & $b$ & $\haterr$ & $\hatfpr$ & $\hatfnr$ & $\haterr_\text{NW}$ & $\hatfpr_\text{NW}$  & $\hatfnr_\text{NW}$ & $\haterr_\text{W}$  & $\hatfpr_\text{W}$ & $\hatfnr_\text{W}$ & $\hatfpr_\text{NW} - \hatfpr_\text{W}$ \\
      \midrule
      \cellcolor{orange!33}757	 & $8$ &	$464$ &	$0.369$ &	\cellcolor{black!20} $0.167$ &	$0.202$ &	$0.395$ &	\cellcolor{blue!33}$0.201$ &	$0.193$ &	$0.318$ &	\cellcolor{red!33}$0.101$ &	$0.218$ &	\cellcolor{purple!33}$0.1$ \\
      729	& $8$ &	$240$	& $0.358$ &	\cellcolor{black!20}$0.162$ & 	$0.197$ &	$0.376$ &	\cellcolor{blue!33}$0.189$ &	$0.187$ &	$0.323$ &	\cellcolor{red!33}$0.107$ &	$0.216$ &	\cellcolor{purple!33}$0.082$\\
      791	& $8$ & $736$	& $0.377$	& \cellcolor{black!20} $0.171$	& $0.205$ &	$0.395$ &	\cellcolor{blue!33}$0.198$ &	$0.197$ &	$0.341$ &	\cellcolor{red!33}$0.118$ &	$0.222$ &	\cellcolor{purple!33}$0.08$\\
      \bottomrule
    \end{tabular}
    }
    \label{app:subtable:fprNW-minus-fprW}
    \end{subtable}

    \vspace{0.25cm}
    \begin{subtable}[h]{\linewidth}
    \centering
    \subcaption{\footnotesize{Top-$3$ most unfair models by $\hatfpr$, when  $\hatfpr_\text{W} > \hatfpr_\text{NW}$ (i.e., unfair toward W).}}
    {\footnotesize
    \begin{tabular}{lllcccccccccc}
      \toprule
      {Run \#} & $s$ & $b$ & $\haterr$ & $\hatfpr$ & $\hatfnr$ & $\haterr_\text{NW}$ & $\hatfpr_\text{NW}$  & $\hatfnr_\text{NW}$ & $\haterr_\text{W}$  & $\hatfpr_\text{W}$ & $\hatfnr_\text{W}$ & $\hatfpr_\text{W} - \hatfpr_\text{NW}$ \\
      \midrule
      639 &	$7$ &	$280$ &	$0.36$ &	\cellcolor{black!20}$0.187$ &	$0.173$ &	$0.352$ &	\cellcolor{red!33}$0.174$ &	$0.178$ &	$0.376$ &	\cellcolor{blue!33}$0.212$ &	$0.164$ &	\cellcolor{purple!33}$0.038$ \\
      807	& $9$ &	$72$ 	& $0.381$ &	\cellcolor{black!20}$0.191$ &	$0.19$ &	$0.372$ &	\cellcolor{red!33}$0.179$ &	$0.192$ &	$0.398$ &	\cellcolor{blue!33}$0.214$ &	$0.184$ &	\cellcolor{purple!33}$0.035$\\
      543	 & $6$	& $264$ &	$0.358$ &	\cellcolor{black!20}$0.155$ &	$0.203$ &	$0.351$ &	\cellcolor{red!33}$0.144$ &	$0.206$ &	$0.37$ &	\cellcolor{blue!33}$0.175$ &	$0.196$ &	\cellcolor{purple!33}$0.031$ \\
      \bottomrule
    \end{tabular}
    }
    \label{app:subtable:fprW-minus-fprNW}
    \end{subtable}

    \vspace{0.25cm}
    \begin{subtable}[h]{\linewidth}
    \centering
        \subcaption{\footnotesize{Top-$3$ most unfair models by $\hatfnr$, when  $\hatfnr_\text{NW} > \hatfnr_\text{W}$ (i.e., unfair toward NW).}}
    {\footnotesize
    \begin{tabular}{lllcccccccccc}
      \toprule
      {Run \#} & $s$ & $b$ & $\haterr$ & $\hatfpr$ & $\hatfnr$ & $\haterr_\text{NW}$ & $\hatfpr_\text{NW}$  & $\hatfnr_\text{NW}$ & $\haterr_\text{W}$  & $\hatfpr_\text{W}$ & $\hatfnr_\text{W}$ & $\hatfnr_\text{NW} - \hatfnr_\text{W}$ \\
      \midrule
      246	& $3$ &	$141$ &	$0.379$ &	$0.166$ &	\cellcolor{black!20} $0.213$ &	$0.398$ &	$0.169$ &	\cellcolor{blue!33}$0.229$ &	$0.345$ &	$0.161$ &	\cellcolor{red!33}$0.184$ &	\cellcolor{purple!33}$0.045$ \\
      506 &	$6$	& $42$ & $0.367$ &	$0.17$ &	\cellcolor{black!20} $0.197$ &	$0.386$ &	$0.175$ &	\cellcolor{blue!33}$0.211$ &	$0.332$ &	$0.161$ &	\cellcolor{red!33}$0.171$ &	\cellcolor{purple!33}$0.04$\\
      204	& $3$	& $15$ &	$0.384$ &	$0.185$ &	\cellcolor{black!20} $0.199$ &	$0.394$ &	$0.181$ &	\cellcolor{blue!33}$0.213$ &	$0.365$ &	$0.192$ &	\cellcolor{red!33}$0.173$ &	\cellcolor{purple!33}$0.04$ \\
      \bottomrule
    \end{tabular}
    }
    \label{app:subtable:fnrNW-minus-fnrW}
    \end{subtable}
    
    \vspace{0.25cm}
    \begin{subtable}[h]{\linewidth}
    \centering
    \subcaption{\footnotesize{Top-$3$ most unfair models by $\hatfnr$, when  $\hatfnr_\text{W} > \hatfnr_\text{NW}$ (i.e., unfair toward W).}}
    {\footnotesize
    \begin{tabular}{lllcccccccccc}
      \toprule
      {Run \#} & $s$ & $b$ & $\haterr$ & $\hatfpr$ & $\hatfnr$ & $\haterr_\text{NW}$ & $\hatfpr_\text{NW}$  & $\hatfnr_\text{NW}$ & $\haterr_\text{W}$  & $\hatfpr_\text{W}$ & $\hatfnr_\text{W}$ & $\hatfnr_\text{W} - \hatfnr_\text{NW}$ \\
      \midrule
      474	& $5$ &	$375$ & $0.373$ &	$0.175$ &	\cellcolor{black!20}$0.199$ &	$0.356$ &	$0.183$ &	\cellcolor{red!33}$0.174$ &	$0.406$ &	$0.159$ &	\cellcolor{blue!33}$0.247$ &	\cellcolor{purple!33}$0.073$ \\
      401 &	$5$	& $10$ &	$0.378$ &	$0.189$	& \cellcolor{black!20}$0.19$ &	$0.363$ &	$0.197$ &	\cellcolor{red!33}$0.167$ &	$0.406$ &	$0.173$ &	\cellcolor{blue!33}$0.233$ &	\cellcolor{purple!33}$0.066$\\
      52	& $1$	& $53$ &	$0.367$ &	$0.172$ &	\cellcolor{black!20}$0.196$ &	$0.351$ &	$0.178$ &	\cellcolor{red!33}$0.173$ &	$0.397$ &	$0.16$ &	\cellcolor{blue!33}$0.238$ &	\cellcolor{purple!33}$0.065$ \\
      \bottomrule
    \end{tabular}
    }
    \label{app:subtable:fnrW-minus-fnrNW}
    \end{subtable}
  \label{app:table:compas-model-runs-diffs}
\end{table}
\FloatBarrier
\newpage

This detailed view provides insight into how such a result is possible. Broadly speaking, individual runs have roughly similar error;\footnote{This should be taken relatively. In general, \texttt{COMPAS} demonstrates high error; the error is relatively tight given just how much error there is. The error fluctuates depending on the training data, but the average error rate across train/test splits is rather tight, despite the fluctuations in error within the $\boot$ runs of each split.} yet, the subgroup-specific error rates that compose the overall error can nevertheless vary widely depending on the underlying training data. This observation aligns with current interest in \emph{model multiplicity} in the algorithmic fairness community~\cite{black2022multiplicity, watson2023multiplicity}, which imports the idea from Breiman~\cite{breiman2001multiplicity}. In this case, as suggested by Table~\ref{app:table:compas-model-runs-diffs}, there are models that demonstrate unfairness toward both subgroups with respect to each error rate metric $\haterr$, $\hatfpr$, and $\hatfnr$. When we move away from attempting to find a \emph{single} model that performs well (accurately or fairly) on \texttt{COMPAS}, and instead consider the information contained across different possible models, we yield the result that the average, expected behavior smooths over the variance in underlying models such that the result is close to fair. 

\begin{figure}[t]
\centering
        \includegraphics[width=.6\linewidth]{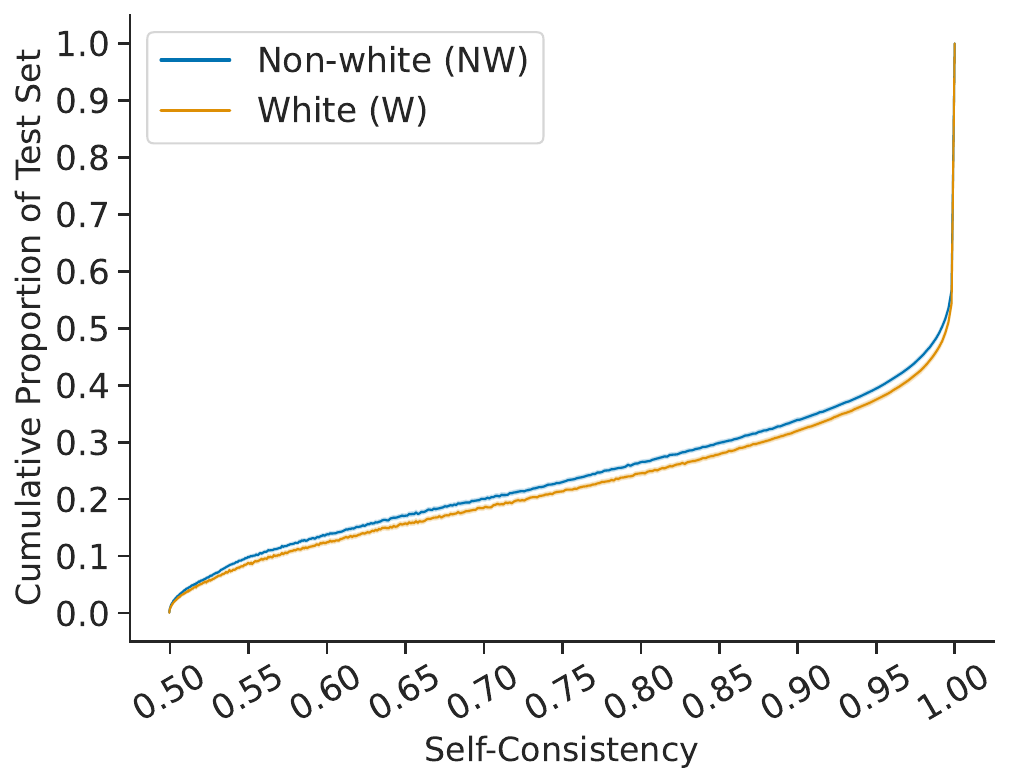}
        \vspace{-.3cm}
        \captionof{figure}{\texttt{COMPAS} split by $\group=\texttt{race}$, $B=1001, S=100$}
        \label{app:fig:compas-cdf-rfc-100}
    \vspace{.25cm}
      \begin{table}[H]
        \caption{Mean $\pm$ STD across $S=100$ train/test splits $\times\; B=1001$ runs.\looseness=-1}
        \label{app:table:compas-rfc-100}
        \begin{center}
        \begin{tabular}{lcccc}
		\toprule
		\multicolumn{5}{c}{\textbf{\texttt{COMPAS}}} \\ \cmidrule(lr){1-5}
		 & \textbf{$\haterr$}     & \textbf{$\hatfpr$}     & \textbf{$\hatfnr$}     & \textbf{$\hatsc$} \\ \midrule
		 \textbf{Total} & $.333\pm.008$ & $.14\pm.009$ & $.192\pm.01$ & $.883\pm.004$\\ \midrule
		 $g=\text{NW}$ & $.333\pm.01$ & $.148\pm.011$ & $.185\pm.012$ & $.88\pm.005$\\ \midrule
		 $g=\text{W}$ & $.332\pm.014$ & $.125\pm.013$ & $.207\pm.016$ & $.888\pm.006$\\ \bottomrule
	\end{tabular}
        \end{center}
        \end{table}
\end{figure}

\paragraph{Stability analysis.} To verify the stability of this result, we re-execute our experiments for increasing numbers of train/test splits $S$ and replicates $\boot$. While our results for \texttt{COMPAS} are generally tight for small $S$ (e.g., Figures~\ref{subfig:compas-cdf-rfc} and~\ref{fig:compas-ens}), this was not the case for \texttt{German Credit}, for which it was difficult to estimate self-consistency consistently. As a result, for \texttt{COMPAS}, we did not expect markedly different results for increased $S$. Our results for $S=100, B=1001$ using logistic regression (Figure~\ref{app:fig:compas-cdf-rfc-100}, Table~\ref{app:table:compas-rfc-100})  confirm this intuition. 

\FloatBarrier
\pagebreak

\begin{figure}[t]
\centering
        \includegraphics[width=.6\linewidth]{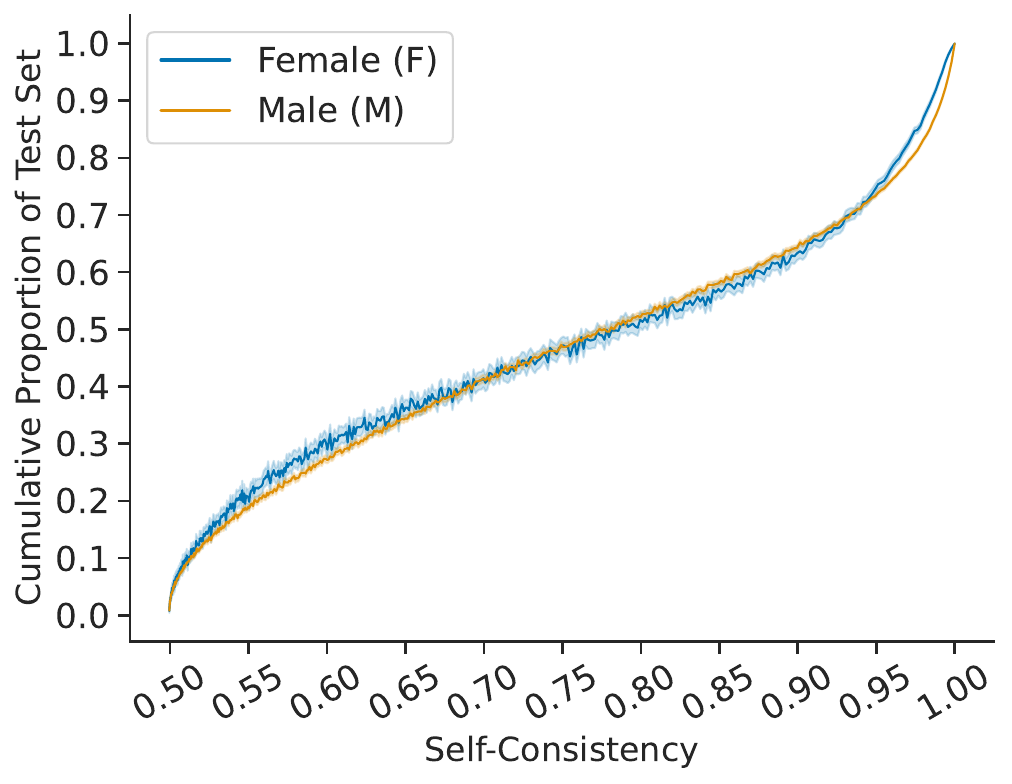}
        \vspace{-.3cm}
        \captionof{figure}{\texttt{German Credit} split by $\group=\texttt{sex}$, $S=1000, B = 100$\looseness=-1}
        \label{app:fig:german-cdf-rfc-1000}
        \vspace{.25cm}
      \begin{table}[H]
        \caption{Mean $\pm$ STD across $S=1000$ train/test splits $\times\; B=1001$ runs.\looseness=-1}
        \label{app:table:german-rfc-1000}
        \begin{center}
        \begin{tabular}{lcccc}
		\toprule
		\multicolumn{5}{c}{\textbf{\texttt{South German Credit}}} \\ \cmidrule(lr){1-5}
		 & \textbf{$\haterr$}     & \textbf{$\hatfpr$}     & \textbf{$\hatfnr$}     & \textbf{$\hatsc$} \\ \midrule
		 \textbf{Total} & $.28\pm.021$ & $.173\pm.028$ & $.107\pm.017$ & $.769\pm.015$\\ \midrule
		 $g=\text{F}$ & $.288\pm.064$ & $.183\pm.072$ & $.105\pm.037$ & $.766\pm.04$\\ \midrule
		 $g=\text{M}$ & $.279\pm.023$ & $.171\pm.029$ & $.108\pm.018$ & $.769\pm.016$\\ \bottomrule
	\end{tabular}
        \end{center}
        \end{table}
\end{figure}

We provide analogous results for \texttt{German Credit}, with $S=1000, B=1001$ using random forests (Figure~\ref{app:fig:german-cdf-rfc-1000}, Table~\ref{app:table:german-rfc-1000}). It takes an enormous number of runs to produce stable estimates of error and $\hatsc$ for \texttt{German Credit}, which indicate statistical equality across groups. Arguably, our results below for $1,001,000$ models still are very high variance (certainly with respect to error metrics). This task really has too few data points ($\approx600$) to generalize reliably. 
\vspace*{.5cm}

%% file: section/99-appendix/22-fairness/600-app-fairness-future.tex
\section{\appfuture}\label{app:sec:future}

There are many interesting directions for future work. 

\paragraph{Novel theory.} We do not include extensive novel theory in this project. Nevertheless, our project raises interesting questions for theory in future work. Notably, we could compose our methodology with post-processing~\citep{hardt2016equality} for cases in which there is observed empirical unfairness. We could then investigate picking group-specific thresholds that take variance into account. We could reconfigure the formulations in \citep{hardt2016equality} and related work, with respect to the fairness-accuracy trade-off, as actually representing multiple such trade-off curves (that are a function of different models under consideration). There may be interesting directions for mathematical analysis in this direction. 

We could also extend traditional results on bagging and variance reduction for classifiers. While bagging has guarantees for variance reduction for regression, it does not have the same guarantees for classification~\citep{breiman1996bagging, breiman1998ac}. It generally is observed to work well in practice for variance reduction if the underlying classifiers are high variance --- which is indeed the regime we are in for this paper. However, there are interesting theory questions regarding abstention that we could investigate with theoretical tools, which could let us come up with other ways of reasoning about bagging and variance reduction.

Both of these directions are out of scope for the present paper. They are interesting, but do not have to do with our main experimental aims and contributions, and thus do not make it into a conference-length submission. We are not interested in novel theory in the present study. If anything, our work highlights how over-attention to theory can (directly or indirectly) bring about serious problems of mismeasurement in practice. That is a main takeaway for our work, which by nature does not involve novel theory.\looseness=-1

\paragraph{Arbitrariness beyond algorithmic fairness.} Our framework for reasoning about self-consistency and arbitrariness does not inherently have to do with algorithmic fairness. We could apply it to other domains. For example, it would be interesting to ask similar questions in deep learning and generative AI. We think that such work would be interesting, but is again out of scope for the present study. The first author of this project is in fact working on such questions as separate work. However, this project's research aims are inherently focused on fairness; the project was designed in response to observations in experimental practices in the fairness community, fairness definitions, and fairness theory.

\paragraph{Experiments on synthetic data.} Our results indicate that unfairness (as defined with respect to model error rates) is not frequently observed on common benchmark tasks in fair classification. Of course, there could be other datasets in fairness domains that are not currently used as benchmarks that more clearly demonstrate unfairness in practice. Hypothetically, there could be datasets for which we use Algorithm~\ref{algo:bagging-confidently} to reduce arbitrariness, and yet we still see significant systematic arbitrariness or differences in error rates (and thus unfairness) due to noise or bias. We just did not really see this for almost all of the tasks we investigate in this paper, which happen to be the ones that the fairness community uses for experiments. 

To study Algorithm~\ref{algo:bagging-confidently} in light of these other possibilities, we could develop synthetic datasets that retain unfairness after dealing with arbitrariness. We did not do this in the present study for two reasons. First, our focus was the practice of fairness research, as it currently stands, with a data-centric approach on the datasets people actually use for their research. We are not interested in synthetic data for this project. 
However, future theory results that extend our work could be vetted experimentally with synthetic data. The work we mention above regarding composition with post-processing, as well as revisting impossibility results from a distributional approach over possible models, may be very interesting to examine under data settings that we can control. 

\paragraph{How to deal with abstention.} Future work could also perform a deeper exploration of the trade-off between abstention rate and error. We could characterize a Pareto-optimal trade-off that is a function of the choice of self-consistency level $\kappa$, and also examine in experiments and analytically how abstention leads to improvements in accuracy. Future work could also identify patterns in abstention sets beyond low self-consistency. In this, looking to metrics from model multiplicity may be helpful. Further, future work could combine human decision-making or other automated elements to see how we can root out arbitrariness. 

\paragraph{Reproducibility.} As mentioned in our Ethics Statement, we made attempts to reproduce prior work in fair classification, and often could not. We ultimately made reproducibility of specific papers out of scope for the present project, as we could make our contributions about arbitrariness and variance without such work. It would nevertheless be useful to focus future work on reproducing prior algorithmic fairness studies, and seeing if conclusions change in those works as a function of using Algorithm~\ref{algo:bagging-confidently} prior to introducing the proposed fairness intervention.

\paragraph{Law and policy.} Our work regarding arbitrariness raises concrete questions for the law around due process and automated decision-making. 
Preliminary exploration of these ideas can be found in Cooper et al.~\citep{cooper2022lawless} (Chapter~\ref{chapter:nondeterminism}). 
Developing further contributions in this line of work is also out of the scope of the present work.
We are currently pursuing this work for a future submission to a law review journal.

%% file: section/99-appendix/31-tunamh/00-app-tunamh.tex
\chapter{Appendix for \methodname}\label{chapter:app:tunamh}

\input{section/99-appendix/31-tunamh/10-app-tunamh-counterexample}
\input{section/99-appendix/31-tunamh/20-app-tunamh-tv-inexact}
\input{section/99-appendix/31-tunamh/30-app-tunamh-smh}
\input{section/99-appendix/31-tunamh/40-app-tunamh-algorithm}
\input{section/99-appendix/31-tunamh/50-app-tunamh-spectral}
\input{section/99-appendix/31-tunamh/60-app-tunamh-chi}
\input{section/99-appendix/31-tunamh/70-app-tunamh-optimal-chi}
\input{section/99-appendix/31-tunamh/80-app-tunamh-optimal}
\input{section/99-appendix/31-tunamh/90-app-experiments}

%% file: section/99-appendix/31-tunamh/10-app-tunamh-counterexample.tex
\section{Proof of Theorem \ref{statement:counterexample}}\label{app:proof:counterexample}

In this section, we prove Theorem \ref{statement:counterexample}, which asserts that any inexact stateless MH algorithm can produce arbitrarily large bias between its target distribution (the distribution we are trying to sample from) and its stationary distribution (the distribution that the chain actually produces samples from asymptotically). 
\begin{proof}
Let $\mathcal{A}$ denote the \texttt{SubsMH} in Algorithm~\ref{alg:subsampledMH} of the minibatch MH method in question.
Since $\mathcal{A}$ is inexact, there must exist a state space $\Theta$, proposal distribution $q$, and target distribution $\mu$, satisfying Assumption~\ref{assump} with parameters $c_1, \ldots, c_N, C, M$, where
\[
    \mu(\theta) \propto \exp\left( -\sum_{i=1}^N V_i(\theta) \right)
\]
for some $N$ and energy functions $V_1, \ldots, V_N$, such that $\mathcal{A}$ run on $\mu$ with proposal distribution $q$ does not have stationary distribution $\mu$.

Next, let $a_{\mu}(\theta, \theta')$ denote the acceptance probability of algorithm $\mathcal{A}$ on the above task for a proposed transition from $\theta$ to $\theta'$.
Assume by way of contradiction that on this problem, it is always true that
\[
    \frac{a_{\mu}(\theta, \theta')}{a_{\mu}(\theta', \theta)}
    =
    \frac{
        \mu(\theta') q(\theta|\theta')
    }{
        \mu(\theta) q(\theta'|\theta)
    }.
\]
If this were true, then the overall transition probability of this chain, for $\theta \ne \theta'$, would be
\[
    T_{\mu}(\theta, \theta') = q(\theta'|\theta) \cdot a_{\mu}(\theta, \theta')
\]
and it would hold that
\[
    \mu(\theta) T_{\mu}(\theta, \theta') = \mu(\theta') T_{\mu}(\theta', \theta).
\]
That is, the chain would be reversible, also known as satisfying detailed balance.
But it is a standard result that for any reversible chain, $\mu$ must be a stationary distribution of that chain.
We have now derived a contradiction, which establishes that our assumption is false.
That is, there exists a $\theta, \theta' \in \Theta$ such that
\[
    \frac{a_{\mu}(\theta, \theta')}{a_{\mu}(\theta', \theta)}
    \ne
    \frac{
        \mu(\theta') \cdot q(\theta|\theta')
    }{
        \mu(\theta) \cdot q(\theta'|\theta)
    }.
\]
Explicitly, this means that if we define the function $\Delta V$ such that
\[
    \Delta V(i) = V_i(\theta) - V_i(\theta'),
\]
then for this subsampling problem,
\begin{equation}
    \label{eqnStmt1Proof1}
    \frac{
        \mathbf{E}\left[\mathcal{A}(\Delta V, N, q(\theta|\theta')/q(\theta'|\theta), c_1, \ldots, c_N, C, M(\theta,\theta')) \right]
    }{
        \mathbf{E}\left[\mathcal{A}(-\Delta V, N, q(\theta'|\theta)/q(\theta|\theta'), c_1, \ldots, c_N, C, M(\theta,\theta')) \right]
    }
    \ne
    \frac{
        \mu(\theta') \cdot q(\theta|\theta')
    }{
        \mu(\theta) \cdot q(\theta'|\theta)
    }.
\end{equation}
Without loss of generality, assume that
\[
    q(\theta|\theta')/q(\theta'|\theta) \le 1.
\]
(This is without loss of generality since we can ensure it is the case by swapping $\theta$ and $\theta'$.) We fixed $\theta$ and $\theta'$ to be the pair satisfying Equation~\ref{eqnStmt1Proof1} throughout this section.

\paragraph{Constructing an example.} We use this to prove the theorem by a constructive example. Let $x_1, \ldots, x_N$ be defined by
\[
    x_i = \Delta V(i) = V_i(\theta) - V_i(\theta').
\]
Define $X$ as the sum
\[
    X = \sum_{i=1}^N x_i.
\]
For some parameter $K \in \mathbb{N}$ (to be defined later), consider the state space $\Omega$ defined as
\[
    \Omega = \{(k,z) \mid k \in \{0, \ldots, K-1\}, \; 0 \le z \le \exp(k X) \},
\]
using the natural measure for a finite disjoint union of measure spaces.
Define a target distribution over $\Omega$ given by the density
\[
  \pi(k,z) \propto \exp\left(-\sum_{i=1}^N k \cdot x_i\right),
\]
or equivalently
\[
    \pi(k,z) \propto \exp\left(-\sum_{i=1}^N U_i(k,z) \right) \; \text{where} \; U_i(k,z) = k x_i.
\]
Define a proposal distribution $\hat q$, such that, starting from $(k,z)$:
\begin{itemize}
    \item With probability $1/4$, we sample $z'$ uniformly from $[0, \exp(k X)]$ and propose a transition to $(k,z')$.
    \item With probability $1/4$, we propose a transition to $(k-1,z)$, if it is in $\Omega$.
    \item With probability $\frac{1}{4} \cdot \frac{q(\theta|\theta')}{q(\theta'|\theta)}$, we propose a transition to $(k+1,z)$, if it is in $\Omega$.
    \item With the remaining probability, we just propose to stay at $(k,z)$.
\end{itemize}
This is effectively acting as a random walk over $k$, and our goal will be to show that while the true target distribution $\pi$ has a marginal in $k$ that is the uniform distribution, the minibatch MH method causes the chain's transition to be biased to step more in one direction than another, resulting in a highly biased stationary distribution (where we can make the bias arbitrarily large by setting $K$).

We use the same $c_i$ and $C$ as before, and define a new function $\hat M$ such that
\[
    \hat M((k,z),(k+1,z)) = \hat M((k,z),(k-1,z)) = M(\theta, \theta')
\]
and $\hat M(\cdots) = 0$ for other proposed transitions (we can set $\hat M$ however we want for pairs of states that are never proposed in a transition, since this will not affect the algorithm).
Clearly, this setup satisfies Assumption~\ref{assump}, since the original distribution did.

Now, consider what our minibatch MH method will do when run on this task. There are three cases to consider.

\paragraph{Proposed changes in $z$.} When a proposed change in $z$ is made, the resulting $\Delta U$ will be uniformly $0$, and the probability of the reverse transition will be equal (1/4 in both directions), so the algorithm will be passed the arguments
\[
    \mathcal{A}(0, N, 1, c_1, \ldots, c_N, C, 0).
\]
Since this does not depend at all on $z$ or $k$, this means that the acceptance probability of these transitions will be the same regardless of the state. Call this probability $\alpha_0$.

\paragraph{A proposal to decrease $k$.} When a proposal is made to decrease $k$, the probability of the forward and reverse transitions will be
\[
    \hat q((k-1,z)|(k,z)) = \frac{1}{4}
    \;\text{and}\;
    \hat q((k,z)|(k-1,z)) = \frac{1}{4} \cdot \frac{q(\theta|\theta')}{q(\theta'|\theta)}.
\]
It follows that
\[
    \frac{\hat q((k,z)|(k-1,z))}{\hat q((k-1,z)|(k,z))} = \frac{q(\theta|\theta')}{q(\theta'|\theta)}.
\]
The energy function difference for this proposal will be
\[
    \Delta U(i) = U_i((k,z)) - U_i((k-1,z)) = k x_i - (k - 1) x_i = x_i,
\]
so in particular $\Delta U = \Delta V$. And, of course for this transition $\hat M$ will take on the value $M(\theta, \theta')$.
So, the minibatch MH algorithm will be passed the arguments
\[
    \mathcal{A}(\Delta V, N, q(\theta|\theta')/q(\theta'|\theta), c_1, \ldots, c_N, C, M(\theta, \theta')),
\]
and so it will accept with probability
\[
    \mathbf{E}\left[ \mathcal{A}(\Delta V, N, q(\theta|\theta')/q(\theta'|\theta), c_1, \ldots, c_N, C, M(\theta, \theta')) \right].
\]
Call this probability $\alpha_-$.

\paragraph{A proposal to increase $k$.} When a proposal is made to increase $k$, the probability of the forward and reverse transitions will be
\[
    \hat q((k+1,z)|(k,z)) = \frac{1}{4} \cdot \frac{q(\theta|\theta')}{q(\theta'|\theta)}.
    \;\text{and}\;
    \hat q((k,z)|(k+1,z)) = \frac{1}{4}.
\]
It follows that
\[
    \frac{\hat q((k,z)|(k+1,z))}{\hat q((k+1,z)|(k,z))} = \frac{q(\theta'|\theta)}{q(\theta|\theta')}.
\]
The energy function difference for this proposal will be
\[
    \Delta U(i) = U_i((k,z)) - U_i((k+1,z)) = k x_i - (k + 1) x_i = -x_i,
\]
so in particular $\Delta U = -\Delta V$. And, as before for this transition $\hat M$ will take on the value $M(\theta, \theta')$.
So, the minibatch MH algorithm will be passed the arguments
\[
    \mathcal{A}(-\Delta V, N, q(\theta'|\theta)/q(\theta|\theta'), c_1, \ldots, c_N, C, M(\theta, \theta')),
\]
and so it will accept with probability
\[
    \mathbf{E}\left[ \mathcal{A}(-\Delta V, N, q(\theta'|\theta)/q(\theta|\theta'), c_1, \ldots, c_N, C, M(\theta, \theta')) \right].
\]
Define the probability $\alpha_+$ as
\[
    \alpha_+ = \mathbf{E}\left[ \mathcal{A}(-\Delta V, N, q(\theta'|\theta)/q(\theta|\theta'), c_1, \ldots, c_N, C, M(\theta, \theta')) \right] \cdot \frac{q(\theta|\theta')}{q(\theta'|\theta)}.
\]

\paragraph{The resulting Markov chain.}
From the above analysis, we can conclude that the Markov chain that results from subsampling algorithm $\mathcal{A}$ applied to this method is as follows.
Starting from $(k,z)$, if we let $\hat T$ denote the transition operator of this Markov chain,
\begin{itemize}
    \item With probability $\frac{1}{4} \cdot \alpha_0$, we sample $z'$ uniformly from $[0, \exp(k X)]$ and transition to $(k,z')$.
    \item With probability $\frac{1}{4} \cdot \alpha_-$, we transition to $(k-1,z)$, if it is in $\Omega$.
    \item With probability $\frac{1}{4} \cdot \alpha_+$, we transition to $(k+1,z)$, if it is in $\Omega$.
    \item With the remaining probability, we just stay at $(k,z)$.
\end{itemize}
Consider the distribution
\[
    \nu(k,z) \propto \left( \frac{\alpha_+}{\alpha_-} \right)^k.
\]
It is easy to see that this Markov chain satisfies detailed balance with $\nu$ as its stationary distribution. In particular,
\begin{align*}
        \nu(k,z) \cdot T((k-1,z)|(k,z)) &= \left( \frac{\alpha_+}{\alpha_-} \right)^k \cdot \frac{1}{4} \cdot \alpha_-\\
    &=
    \left( \frac{\alpha_+}{\alpha_-} \right)^{k-1} \cdot \frac{1}{4} \cdot \alpha_+ \\
    &= \nu(k-1,z) \cdot T((k,z)|(k-1,z)). 
\end{align*}

So $\nu$ will be a stationary distribution of the minibatch MH chain $\hat T$.

Observe that the marginal distribution of $k$ in $\pi$ is
\[
    \pi(k) = \int_{0}^{\exp(kX)} \pi(k,z) \; dz \propto \exp\left(-\sum_{i=1}^N k \cdot x_i\right) \cdot \exp(kX) = 1,
\]
so the marginal distribution of $k$ in the target distribution is actually the uniform distribution.
On the other hand, using the same derivation, the marginal distribution of $k$ in $\nu$ is
\[
    \nu(k) \propto \left( \frac{\alpha_+}{\alpha_-} \right)^k \cdot \exp(kX) = \left( \frac{\alpha_+}{\alpha_-} \cdot \exp(X) \right)^k.
\]

We know immediately by substituting our definitions of $\alpha_+$ and $\alpha_-$ into (\ref{eqnStmt1Proof1}) that
\[
    \frac{
        \alpha_-
    }{
        \alpha_+
    }
    \ne
    \frac{
        \mu(\theta')
    }{
        \mu(\theta)
    }
    =
    \exp\left( \sum_{i=1}^N (V_i(\theta) - V_i(\theta') \right)
    =
    \exp\left( \sum_{i=1}^N x_i \right)
    =
    \exp(X).
\]
As a consequence, we know that
\[
    \frac{\alpha_+}{\alpha_-} \cdot \exp(X) \ne 1.
\]
Call this constant
\[
    A = \frac{\alpha_+}{\alpha_-} \cdot \exp(X),
\]
and observe that $A \ne 1$ and that $A$ is independent of our choice of $K$ (which still remains unset).
This gives
\[
    \nu(k) \propto A^k.
\]
Explicitly, this distribution will be
\[
    \nu(k) = \frac{1}{\sum_{k=0}^{K-1} A^{k}}\cdot A^{k} = \frac{1-A}{1 - A^K} \cdot A^k.
\]
Since the total variation distance between two probability measures is lower bounded by the TV-distance between their marginal distributions in any one variable, and similarly the KL divergence is \emph{also} lower bounded by the KL divergence between its marginal distributions in any one variable (both these facts follow directly from the monotonicity property of the $f$-divergence, of which the KL-divergence and TV-distance are both instances), to prove this theorem it suffices to show both TV-distance and KL-divergence bounds on the marginal distributions in $k$. We do this now.

\paragraph{Bounding the total variation distance.}

Now, we compute the total variation distance between $\pi$ and $\nu$. For this bit of the proof, we will just consider the marginal distribution in $k$, as this provides a lower bound on the TV distance between the joint distribution. For simplicity, for the rest of the proof, we let $\tilde \pi$ denote this marginal distribution of $k$ in $\nu$, and also let $\pi$ denote the marginal distribution of $k$ in $\pi$.
By the definition of total variation distance,
\begin{align*}
\text{TV}(\pi, \tilde \pi) 
&= \frac{1}{2}\sum_{k=0}^{K-1}\left| \tilde \pi(k) - \pi(k) \right| \\
&= \frac{1}{2}\sum_{k=0}^{K-1}\left|\frac{1-A}{1 - A^{K}}\cdot A^{k} - \frac{1}{K}\right|.
\end{align*}
If $A<1$,
\begin{align}\label{eq:tv}
\text{TV}(\pi, \tilde{\pi})
& =
\sum_{k=0}^{K_0}\left(\frac{1-A}{1 - A^{K}}\cdot A^{k} - \frac{1}{K}\right)\nonumber\\
& =
\frac{1-A^{K_0}}{1 - A^{K}} - \frac{K_0}{K}
\end{align}
where $K_0$ is the largest $k$ such that
\[
\frac{1-A}{1 - A^{K}}\cdot A^{k} > \frac{1}{K}.
\]
By solving the above equation, we have
\[
K_0 = \left\lfloor \frac{\log(1-A^{K}) - \log(1-A) - \log(K)}{\log(A)}\right\rfloor.
\]

We can lower bound $K_0$ by
\begin{align*}
    K_0&\ge \frac{\log(1-A^{K}) - \log(1-A) - \log(K)}{\log(A)} - 1\\
    &\ge \frac{- \log(1-A) - \log(K)}{\log(A)} - 1.
\end{align*}

It follows that the first term in (\ref{eq:tv}) becomes
\begin{align*}
    \frac{1-A^{K_0}}{1 - A^{K}}\ge
    \frac{1-\frac{1}{KA(1-A)}}{1 - A^{K}}
    \ge 1-\frac{1}{KA(1-A)}.
\end{align*}
We can also upper bound $K_0$ and then the second term can be bounded as the following
\begin{align*}
\frac{K_0}{K}
& \le
\frac{\log(1-A^{K}) - \log(K)}{K\log(A)}.
\end{align*}
When $K\ge \frac{\log\left(1 - \exp\left(-\frac{1}{2}\right)\right)}{\log(A)}$, we have $\log(1-A^{K})\ge -\frac{1}{2}$. Since $\log(K)\le K^{\frac{1}{2}}$ and $K^{-1} \le K^{-\frac{1}{2}}$, we have
\begin{align*}
\frac{K_0}{K}
& \le
\frac{-\frac{1}{2}K^{-1} - K^{-\frac{1}{2}}}{\log(A)}
\le
-\left(\frac{3}{2\log(A)}\right) K^{-\frac{1}{2}}.
\end{align*}
Therefore, the TV distance is bounded by
\begin{align*}
\text{TV}(\pi, \tilde{\pi})
& \ge
1-\frac{1}{KA(1-A)} + \left(\frac{3}{2\log(A)}\right) K^{-\frac{1}{2}}\\
&\ge 
1 + \left(\frac{3}{2\log(A)} -\frac{1}{A(1-A)}\right) K^{-\frac{1}{2}}.
\end{align*}

To make $\text{TV}(\pi, \tilde{\pi})\ge \delta$, we just need to set
\[
K \ge \frac{\left(\frac{3}{2\log(A)} -\frac{1}{A(1-A)}\right)^2}{(1 - \delta)^2}.
\]
Similarly, if $A>1$, 
\begin{align*}
\text{TV}(\pi, \tilde{\pi})
& =
\sum_{k=K_0}^{K-1}\left(\frac{1-A}{1 - A^{K}}\cdot A^{k} - \frac{1}{K}\right)\\
& =
\frac{A^{K} - A^{K_0}}{A^{K} - 1} - \frac{K - K_0}{K}\\
& = 
\frac{K_0}{K} - \frac{A^{K_0}-1}{A^{K} - 1}
\end{align*}
where
\[
K_0 = \left\lceil \frac{\log(A^{K} - 1) - \log(A - 1) - \log(K)}{\log(A)} \right\rceil
\]
which is the smallest $k$ such that
\[
\frac{1-A}{1 - A^{K}}\cdot A^{k} > \frac{1}{K}.
\]
We can get an upper bound of $K_0$ by
\begin{align*}
K_0 &\le \frac{\log(A^{K} - 1) - \log(A - 1) - \log(K)}{\log(A)} + 1\\
& = 
\log_A \left(\frac{A^{K} - 1}{K(A-1)}\right) + 1.
\end{align*}
Therefore,
\begin{align*}
\frac{A^{K_0}-1}{A^{K} - 1} 
&\le 
\frac{A\cdot\left(\frac{A^{K} - 1}{K(A-1)}\right) - 1}{A^{K}-1}\\
& = 
\frac{A}{K(A-1)} - \frac{1}{A^{K} - 1}.
\end{align*}

We can lower bound $K_0$ by
\begin{align*}
K_0
& \ge 
\log_A \left(A^{K}-1\right) - \log_A(A-1) - \log_A(K).
\end{align*}
When $K\ge 1 - \log_A (A-1)$, $A^{K}-1\ge A^{K-1}$. Then we have
\begin{align*}
K_0
& \ge 
\log_A \left(A^{K-1}\right) - \log_A(A-1) - \log_A(K)\\
& = K - 1 - \log_A(A-1) - \log_A(K).
\end{align*}
It follows that
\begin{align*}
\frac{K_0}{K}
& \ge 
1 - \frac{1}{K} - \frac{\log_A(A-1)}{K} - \frac{\log_A(K)}{K}.
\end{align*}
Since $\log(K)\le K^{\frac{1}{2}}$ and $K^{-1}\le K^{-\frac{1}{2}}$, the TV distance can be bounded by
\begin{align*}
\text{TV}(\pi, \tilde{\pi})
&\ge
1 - \frac{1}{K} - \frac{\log_A(A-1)}{K} - \frac{\log_A(K)}{K} - \frac{A}{K(A-1)} + \frac{1}{A^{K} - 1}\\
& \ge
1 - \left(1+\log_A(A-1)+\frac{1}{\log(A)}+\frac{A}{A-1}\right)K^{-\frac{1}{2}}.
\end{align*}
To make $\text{TV}(\pi, \tilde{\pi})\ge \delta$, we just need 
\[
K \ge \left(\frac{1+\log_A(A-1)+\frac{1}{\log(A)}+\frac{A}{A-1}}{1 - \delta}\right)^2.
\]
Since we could set $K$ arbitrarily, it is clear that we can do this.

\paragraph{Bounding the KL divergence.}

We can compute \text{KL} divergence between $\pi$ and $\tilde{\pi}$ as follows
\begin{align*}
    \text{KL}(\pi, \tilde{\pi}) &= \sum_{k=0}^{K-1} \frac{1}{K} \cdot \log \left(\frac{1}{K}\cdot \frac{1 - A^{K}}{(1-A)A^{k}}\right)\\
    &=
    \frac{1}{K} \cdot \sum_{k=0}^{K-1}\bigg[  \log \left(\frac{1}{K}\cdot \frac{1 - A^{K}}{(1-A)}\right) - k\log(A)\bigg]\\
    &= \log \left(\frac{1 - A^{K}}{K(1-A)}\right) -
    \frac{\log \left(A\right)}{K} \sum_{k=0}^{K-1} k \\
    & = \log \left(\frac{1 - A^{K}}{K(1-A)}\right) -
    \frac{(K-1) \log \left(A\right)}{2}\\
\end{align*}
If $A<1$, we have
\begin{align*}
    \text{KL}(\pi, \tilde{\pi}) 
    & = \log \left(1 - A^{K}\right) - \log ((1-A)K) -
    \frac{K\log \left(A\right)}{2} + \frac{\log \left(A\right)}{2}\\
    & \ge \log \left(1 - A^{K}\right) - 
    \left( \frac{1-A +\log \left(A\right)}{2}\right) K + \frac{\log \left(A\right)}{2}.
\end{align*}
The last equation is because $\log(x)\le \frac{x}{2}$.

To further simplify the above equation, we first note that $1-A +\log \left(A\right)<0$ when $A\neq 1$. And then when $K\ge \log_A\left(1 - A^{\frac{1}{2}}\right)$, we have $1 - A^{K}\ge A^{\frac{1}{2}}$. It follows that we can simplify it to be
\begin{align*}
    \text{KL}(\pi, \tilde{\pi}) 
    & \ge \log \left(A\right) - 
    \left(\frac{1-A +\log \left(A\right)}{2}\right) K.
\end{align*}
To make $\text{KL}(\pi, \tilde{\pi})\ge \rho$, it is clear that we just need to set
\[
K \ge \frac{2(\rho-\log(A))}{A-1-\log(A)}.
\]

Consider when $A>1$,
\begin{align*}
    \text{KL}(\pi, \tilde{\pi}) 
    & = \log \left(\frac{A^{K}-1}{K(A-1)}\right) -
    \frac{(K-1)\log \left(A\right)}{2}.
\end{align*}
If $K\ge \frac{\log(2)}{\log(A)}$, we have that $A^{K} - 1\ge \frac{A^{K}}{2}$. It follows that
\begin{align*}
    \text{KL}(\pi, \tilde{\pi}) 
    & \ge K\log(A) - \log(K) - \log(2A-2) -
    \frac{K\log \left(A\right)}{2}\\
    &=
    \frac{K\log \left(A\right)}{2} - \log(K) - \log(2A-2).
\end{align*}
To make $\text{KL}(\pi, \tilde{\pi})\ge \rho$, we need 
\[
\frac{K\log \left(A\right)}{2} - \log(K) \ge \rho + \log(2A-2).
\]
Let $K=\exp(y)$. By Taylor series, we know $\exp(y) \ge \frac{y^2}{2}$. Then it follows that
\[
\frac{y^2\log \left(A\right)}{4} - y \ge \rho + \log(2A-2).
\]
Solve the above inequality, we can get
\[
y \ge \frac{1+2\cdot \frac{\log \left(A\right)}{4} \cdot\bigg( \rho + \log(2A-2)\bigg)}{2 \cdot \frac{\log \left(A\right)}{4}} = \frac{2 + \log(A)\bigg(\rho+\log(2A-2)\bigg)}{\log(A)}.
\]
It follows that it suffices to set
\[
K\ge \exp\left(\frac{2 + \log(A)\bigg(\rho+\log(2A-2)\bigg)}{\log(A)}\right).
\]

\paragraph{Concluding the proof.}
The theorem now follows from choosing a $K$ large enough that both the TV distance inequality we derived and the KL divergence inequality we derived are satisfied.
\end{proof}

%% file: section/99-appendix/31-tunamh/20-app-tunamh-tv-inexact.tex
\section{Connection between Theorem~\ref{statement:counterexample} \& TV Bound of Inexact MH}

Some inexact methods such as MHSubLhd~\citep{bardenet2014towards} have bounded TV distance between the target distribution and the approximate distribution (see Proposition 3.2 in Bardenet et al.~\cite{bardenet2014towards}). We would like to emphasize that 
Theorem~\ref{statement:counterexample} is compatible with these results. Specifically, Proposition 3.2 assumes $P_{\text{MH}}$ has a bounded mixing time. It is well known that this produces a TV bound for any kernel by coupling~\citep{levin2017markov}. Our theorem does not have this assumption; it suggests that for MHSubLhd, with a given user-specified error, there exists a target distribution and proposal satisfying Theorem~\ref{statement:counterexample}, on which $P_{\text{MH}}$ either does not have bounded mixing time or the mixing time is large enough such that the TV bound is greater than $\delta$.

%% file: section/99-appendix/31-tunamh/30-app-tunamh-smh.tex
\section{Proof of Statement \ref{statement:fmh}} \label{app:proof:smh}
\begin{proof}
We prove this by construction. Consider a dataset $\{x_i\}_{i=1}^N$. The data instances can take two values $\{-\frac{M}{N}, \frac{M}{N}\}$ where $M$ is a positive constant. Assume that half of the data instances take value $\frac{M}{N}$ and the remaining take $-\frac{M}{N}$. Let the target distribution be $\pi(\theta) = \frac{1}{Z}\exp\left(\theta\cdot\sum_{i=1}^N x_i\right)$ and the domain for $\theta$ be $\{0,1,\dots,K-1\}$. We define the proposal distribution to be the following
\[
p(\theta,\theta) = \frac{1}{2},\hspace{1em}\text{for all }\theta;\hspace{1em} p(\theta, \theta-1) = \frac{1}{4},\hspace{1em} p(\theta, \theta+1) = \frac{1}{4} \hspace{1em}\text{for }\theta\in\{1,\dots,K-2\}; 
\]
and $p(0,1)=p(K-1,K-2)=\frac{1}{2}$.

Recall that FMH factorizes the target distribution $\pi(\theta)$ and the proposal distribution $p(\theta)$ as follows
\[
\pi(\theta) \propto \prod^m_{i=1} \pi_i(\theta),\hspace{2em} p(\theta) \propto \prod^m_{i=1} p_i(\theta)
\]
where $m\geq 1$ and $\pi_i$ and $p_i$ are some non-negative functions. Then the acceptance rate is given by\footnote{There are typos with prime ($'$) marks in this expression in the camera-ready version of the paper.}
\[
a_{\text{FMH}}(\theta,\theta') = \prod_{i=1}^m \min\left(1, \frac{\pi(\theta')p_i(\theta,\theta')}{\pi(\theta)p_i(\theta',\theta)}\right).
\]
A common choice is to set $m=N$. On this example, we can write the acceptance rate of transitioning from $\theta$ to $\theta'=\theta + 1$ in FMH as follows
\begin{align*}
    a_{\text{FMH}}(\theta,\theta') = \prod_{i=1}^N \min\left(1, \exp(x_i)\right)
    = \bigg(\exp\bigg(-\frac{M}{N}\bigg)\bigg)^{\frac{N}{2}}=\exp\bigg(-\frac{M}{2}\bigg).
\end{align*}
It is easy to show that the acceptance rate of transitioning from $\theta$ to $\theta'=\theta - 1$ in FMH is the same.

When $M>-2\log(p)$, it is clear that the acceptance rate of FMH is less than $p$. By contrast, the acceptance rate of standard MH is 
\[
a_{\text{MH}}(\theta,\theta') = \min\bigg(1, \exp\bigg(\pm\sum_{i=1}^N x_i\bigg)\bigg) = 1.
\]

In order to preserve geometric ergodicity, Cornish et al.~\cite{cornish2019scalable} introduces \emph{truncated FMH} (TFMH) which forces FMH degrade to standard MH when the energy exceeds a threshold $R$. If we set hyperparameter $R> M/2$, then in each step, the value of $a_{\text{TFMH}}$ will be the same as $a_{\text{FMH}}$. Therefore, if setting $M>-2\log(p)$, we have
\[
\frac{a_{\text{TFMH}}}{a_{\text{MH}}} \le \frac{p}{1} = p.
\]

If we set $R\le M/2$, TFMH falls back to standard, full-batch MH --- using the whole dataset at each step. This proves the statement.
\end{proof}

%% file: section/99-appendix/31-tunamh/40-app-tunamh-algorithm.tex
\section{Construction of Algorithm \ref{alg:poisson-mh}}\label{app:algo-derivation}

Algorithm \ref{alg:poisson-mh} can be derived by carefully replacing the global bounds on the energy in PoissonMH~\cite{zhang2019poisson} with local bounds on the energy differences (Assumption \ref{assump}).\footnote{In our construction of Algorithm~\ref{alg:old-poissonmh}, we similarly have typos in the camera-ready paper around misplaced $'$ marks in the acceptance ratio.} 
PoissonMH is a variant of Poisson Gibbs and therefore inherits the same assumptions for Gibbs sampling on graphical models, which are often violated in the applications of MH. In particular, PoissonMH works on \emph{factor graphs} which define a distribution $\pi(\theta)$ over a set of factors $\{\phi_i(\theta)\}_{i=1}^{N}$ as follows
\[
\pi(\theta) \propto \exp\left(\sum_{i=1}^N \phi_i(\theta)\right).
\]

PoissonMH assumes that each factor $\phi_i$ is non-negative without the loss of generality (we can add a positive constant to $\phi_i$ to make it non-negative without changing the distribution) and is bounded globally by a constant $M_i$. That is
\begin{align*}
   0 \le \phi_i(\theta) \le M_i \text{ for all } \theta.
\end{align*}
This assumption does not hold for most applications of MH, such as the linear and logistic regression experiments in Section \ref{sec:tunamh:exp}. 

\begin{algorithm}[t]
  \caption{PoissonMH}
  \label{alg:old-poissonmh}
  \begin{algorithmic}
    \State \textbf{given:} initial state $\theta \in \Theta$; proposal dist. $q$; hyperparameter $\lambda$; Global bounds $M_i$, $L$
    \Loop
      \State \textbf{propose} $\theta'\sim q(\cdot|\theta)$
      \For{$i \in \{1,\ldots,N\}$}
        \State\textbf{sample} $s_i \sim \text{Poisson}\left(\frac{\lambda M_i}{L} + \phi_i(\theta)\right)$ 
      \EndFor
      \State \textbf{form minibatch} $\mathcal{S}\leftarrow \{i | s_i>0\}$
      \State \textbf{compute MH ratio} $r \leftarrow \frac{\exp\left(\sum_{i\in \mathcal{S}} s_i\log\left( 1 + \frac{L}{\lambda M_{i}}\phi_i(\theta) \right)\right)q(\theta|\theta')}{\exp\left(\sum_{i\in \mathcal{S}} s_i\log\left( 1 + \frac{L}{\lambda M_{i}}\phi_i(\theta') \right)\right)q(\theta'|\theta)}$
      \State \textbf{with probability} $\min(1,r)$, set $\theta \leftarrow \theta'$
    \EndLoop
  \end{algorithmic}
\end{algorithm}

Let $L = \sum_i M_i$ and define Poisson auxiliary variable $s_i$ as the following

\[
s_i|\theta \sim \text{Poisson}\left(\frac{\lambda M_i}{L} + \phi_i(\theta)\right),
\]
where $\lambda>0$ is a hyperparameter. Running standard MH on the joint distribution of $\theta$ and $s_i$ results in the following acceptance ratio\footnote{The misplaced $'$ marks in the acceptance test also carry through here, and are edited / differ from the current camera-ready version.}
\begin{align*}
    _{\text{PoissonMH}}(\theta, \theta') = \frac{\exp\left(\sum_{i} s_{i} \log\left( 1 + \frac{L}{\lambda M_{i}}\phi_i(\theta) \right)\right)q(\theta|\theta')}{\exp\left(\sum_{i} s_{i} \log\left( 1 + \frac{L}{\lambda M_{i}}\phi_i(\theta') \right)\right)q(\theta'|\theta)}.
\end{align*}

Here, the sum is essentially performed over the set of index $i$ whose $s_i$ is greater than zero. When $s_i = 0$, it is clear that the factor $\phi_i$ will not appear in the acceptance ratio $r_{\text{PoissonMH}}$. Thus PoissonMH enables using a subset of factors for the MH decision step (Algorithm~\ref{alg:old-poissonmh}).

To construct our method from this, we can define the factor $\phi_i$ in the factor graph to be
\begin{align}\label{eq:phi}
    \phi_i(x) = \frac{U_i(\theta)+U_i(\theta')}{2} - U_i(x) + \frac{c_i}{2} M(\theta,\theta')
\end{align}

where $x\in\{\theta, \theta'\}$. It is easy to see that $\phi_i$ satisfy $0\le \phi_i(x) \le c_i M(\theta,\theta')$. 
And then we define the Poisson variables $s_i$ as the follows
\[
s_i|(\theta, \theta') \sim \text{Poisson}\left(\frac{\lambda c_{i}}{C} + \phi_i(\theta)\right) = \text{Poisson}\left(\frac{\lambda c_{i}}{C} + \frac{U_i(\theta') - U_i(\theta) + c_iM(\theta,\theta')}{2}\right).
\]

These Poisson auxiliary variables $\{s_i\}_{i=1}^N$ are called \emph{local}, because their distributions change each iteration depending on the current pair $(\theta, \theta')$ and only rely on local bounds in  Assumption \ref{assump}. This is in contrast to the \emph{global} auxiliary variables used in PoissonMH and FlyMC which are used to form a joint distribution with $\theta$ and both require global bounds in their conditional distributions.

The acceptance ratio $r_{\text{\methodname}}$ is the same as $r_{\text{PoissonMH}}$ but with the new definitions of $s_i$ and $\phi_i$. We outline \methodname{} using the notation of $\phi_i$ and $s_i$ in Algorithm \ref{alg:poisson-mh2}.

We now show that Algorithm \ref{alg:poisson-mh2} is statistically equivalent to Algorithm \ref{alg:poisson-mh}. To see this, we first use \emph{thinning}, a commonly used technique \citep{lewis1979simulation, bierkens2019zig,bouchard2018bouncy,cornish2019scalable,zhang2019poisson}, to quickly resample all $s_i$ from their new distributions in each iteration in Algorithm~\ref{alg:poisson-mh2}. This is achieved by replacing the global bounds with the local bounds in Algorithm 4 in the Appendix of Zhang and De Sa~\cite{zhang2019poisson}. Specifically, we first sample $B$ from a Poisson distribution
\[
B\sim \text{Poisson}(\lambda + CM(\theta,\theta')).
\]
Here $\lambda + CM(\theta,\theta')$ is an upper bound on $\mathbf{E}[\sum_i s_i]$. We then form the minibatch by running

\begin{figure}[h]
  \centering
  \begin{minipage}{.7\linewidth}
\begin{algorithmic}
\For{$b \in \{1,\ldots,B\}$}
        \State \textbf{sample} $i_b$ such that $\mathbf{P}(i_b = i) = c_i/C$, for $i=1\ldots N$
        \State \textbf{with probability} $\frac{\lambda c_{i_b} + C\phi_{i_b}(\theta)}{\lambda c_{i_b} + Cc_{i_b}M(\theta, \theta')}$ \textbf{add} $i_b$ to $\mathcal{I}$ 
        \EndFor
\end{algorithmic}
  \end{minipage}
\end{figure}

By substituting $\lambda=\chi C^2 M^2(\theta,\theta')$ and the expression of $\phi_i$, we can get the part of ``form minibatch $\mathcal{I}$'' in Algorithm~\ref{alg:poisson-mh}.

To see that the MH ratio in Algorithm~\ref{alg:poisson-mh} and~\ref{alg:poisson-mh2} are equivalent, we can write out $r$ in Algorithm~\ref{alg:poisson-mh2}\footnote{Again, the $'$ marks in the camera-ready version of this paper are flipped in the acceptance test, and corrected here.} 
using the above fast way of resampling $s_i$
\[
r_{\text{\methodname}} = \frac{\exp\left(\sum_{i \in \mathcal{I}} \log\left( 1 + \frac{C}{\lambda c_{i}}\phi_i(\theta) \right)\right)q(\theta|\theta')}{\exp\left(\sum_{i \in \mathcal{I}} \log\left( 1 + \frac{C}{\lambda c_{i}}\phi_i(\theta') \right)\right)q(\theta'|\theta)}.
\]

\begin{algorithm}[t]
  \caption{\methodname{}}
  \label{alg:poisson-mh2}
  \begin{algorithmic}
    \State \textbf{given:} initial state $\theta \in \Theta$; proposal dist. $q$; $\lambda$; Asm.~\ref{assump} parameters $c_i$, $C$, $M$; function $\phi_i$ defined in~(\ref{eq:phi})
    \Loop
      \State \textbf{propose} $\theta'\sim q(\cdot|\theta)$ and \textbf{compute} $M(\theta, \theta')$
      \For{$i \in \{1,\ldots,N\}$}
        \State \textbf{sample} $s_i \sim \text{Poisson}\left(\frac{\lambda c_i}{C} + \phi_i(\theta)\right)$ 
      \EndFor
      \State \textbf{form minibatch} $\mathcal{S}\leftarrow \{i | s_i>0\}$
      \vspace{0.5em}
      \State \textbf{compute MH ratio} $r \leftarrow \frac{\exp\left(\sum_{i \in \mathcal{S}} s_i \log\left( 1 + \frac{C}{\lambda c_{i}}\phi_i(\theta) \right)\right)q(\theta|\theta')}{\exp\left(\sum_{i \in \mathcal{S}} s_i \log\left( 1 + \frac{C}{\lambda c_{i}}\phi_i(\theta') \right)\right)q(\theta'|\theta)}$
      \State \textbf{with probability} $\min(1,r)$, set $\theta \leftarrow \theta'$
    \EndLoop
  \end{algorithmic}
\end{algorithm}

We then substitute the definition of $\phi_i$ in (\ref{eq:phi}) and it follows that

\begin{align*}
    r_{\text{\methodname}}
    &=
    \exp\bigg(\sum_{i \in \mathcal{I}} \bigg(\log\bigg(\frac{2\lambda c_i + C\left(U_i(\theta) - U_i(\theta') + c_iM(\theta,\theta')\right)}{2\lambda c_i + C\left(U_i(\theta') - U_i(\theta) + c_iM(\theta,\theta')\right)}\bigg)\bigg)\bigg)
        \cdot \frac{q(\theta|\theta')}{q(\theta'|\theta)}.
\end{align*}

We can rearrange the $\log$ term inside $r_{\text{\methodname}}$ as
\begin{align*}
    &\hspace{-2em}\log\left(\frac{2\lambda c_i + C\left(U_i(\theta) - U_i(\theta') + c_iM(\theta,\theta')\right)}{2\lambda c_i + C\left(U_i(\theta') - U_i(\theta) + c_iM(\theta,\theta')\right)}\right)
    \\&=
    \log\left(\frac{2\lambda c_i + C\left(U_i(\theta) - U_i(\theta') \right) + c_i C M(\theta,\theta')}{2\lambda c_i + C\left(U_i(\theta') - U_i(\theta) \right) + c_i C M(\theta,\theta')}\right)
    \\&=
    \log\left(\frac{1 + \frac{C}{2 \lambda c_i + c_i C M(\theta,\theta')} \left(U_i(\theta) - U_i(\theta') \right)}{1 + \frac{C}{2 \lambda c_i + c_i C M(\theta,\theta')} \left(U_i(\theta') - U_i(\theta) \right)}\right)
    \\&=
    2 \operatorname{artanh}\left(
        \frac{C \left(U_i(\theta) - U_i(\theta') \right)}{c_i (2 \lambda + C M(\theta,\theta'))} 
    \right).
\end{align*}
So $r_{\text{\methodname}}$ can be written as\footnote{Again, fixed $'$ marks in the proposal ratio.}
\[
r_{\text{\methodname}} =
        \exp\left(2 \sum_{i \in \mathcal{I}} \operatorname{artanh}\left(
        \frac{C \left(U_i(\theta) - U_i(\theta') \right)}{c_i (2 \lambda + C M(\theta,\theta'))} 
    \right) \right)
        \cdot \frac{q(\theta|\theta')}{q(\theta'|\theta)}.
\]
Finally setting $\lambda$ to be $\chi C^2M^2(\theta,\theta')$ produces the MH ratio in Algorithm \ref{alg:poisson-mh}.

By proving the equivalence of the minibatch and the MH ratio, we show that Algorithm~\ref{alg:poisson-mh} and \ref{alg:poisson-mh2} are statistically equivalent.

%% file: section/99-appendix/31-tunamh/50-app-tunamh-spectral.tex
\section{Proof of Theorem \ref{thm:spectral-gap}} \label{app:proof:spectral-gap}

We prove Theorem \ref{thm:spectral-gap}, which asserts that \methodname{} is reversible, has stationary distribution $\pi$, and gives bounds on its spectral gap relative to the spectral gap of the original Metropolis-Hastings algorithm.\looseness=-1
\begin{proof}  
For convenience, we prove Theorem \ref{thm:spectral-gap} using Algorithm \ref{alg:poisson-mh2} statement which is statistically equivalent to Algorithm \ref{alg:poisson-mh}.
The transition operator can be written as the following 
\begin{align*}
  &T( \theta,  \theta')\\
  &=
\mathbf{E}\left\{q( \theta'| \theta)\min\left(1, \frac{q( \theta| \theta')\exp\left(  \sum_i  \left[s_{i}\log\left( \frac{\lambda c_i}{C}+ \phi_i(\theta')\right) - \log s_{i}!\right]\right)}
{q( \theta'| \theta)\exp\left(  \sum_i  \left[s_{i}\log\left( \frac{\lambda c_i}{C}+ \phi_i(\theta)\right) - \log s_{i}!\right]\right)}\right)\right\}\\
&=
\mathbf{E}\left\{q( \theta'| \theta)\min\left(1, \frac{q( \theta| \theta')\exp\left(  \sum_i  \left[s_{i}\log\left( \frac{\lambda c_i}{C}+ \phi_i(\theta')\right)\right]\right)}
{q( \theta'| \theta)\exp\left(  \sum_i  \left[s_{i}\log\left( \frac{\lambda c_i}{C}+ \phi_i(\theta)\right) \right]\right)}\right)\right\}\\
&=
\sum_s\left\{q( \theta'| \theta)\min\left(1, \frac{q( \theta| \theta')\exp\left(  \sum_i  \left[s_{i}\log\left( \frac{\lambda c_i}{C}+ \phi_i(\theta')\right)\right]\right)}
{q( \theta'| \theta)\exp\left(  \sum_i  \left[s_{i}\log\left( \frac{\lambda c_i}{C}+ \phi_i(\theta)\right) \right]\right)}\right)\right\}\prod_{i} p(s_{i}| \theta,\theta')\\
&=
\sum_s\left\{q( \theta'| \theta)\min\left(\exp\left(\sum_i  \left[s_{i}\log\left( \frac{\lambda c_i}{C}+ \phi_i(\theta)\right) 
- \phi_i(\theta) -  \frac{\lambda c_i}{C}- \log s_{i}!\right] \right),\right.\right.\\
&\hspace{2em}\left.\left.\frac{q( \theta| \theta')\exp\left(  \sum_i  \left[s_{i}\log\left( \frac{\lambda c_i}{C}+ \phi_i(\theta')\right)\right]\right)}
{q( \theta'| \theta)\exp\left(  \sum_i \phi_i(\theta) +  \frac{\lambda c_i}{C}+ \log s_{i}! \right)}\right)\right\}\\
&=
\sum_s\left\{q( \theta'| \theta)\min\left(\exp\left(\sum_i  \left[s_{i}\log\left( \frac{\lambda c_i}{C}+ \phi_i(\theta)\right) 
- \phi_i(\theta) -  \frac{\lambda c_i}{C}- \log s_{i}!\right] \right),\right.\right.\\
&\hspace{2em}\left.\left.\frac{q( \theta| \theta') }
{q( \theta'| \theta) }\exp\left(\sum_i  \left[s_{i}\log\left( \frac{\lambda c_i}{C}+ \phi_i(\theta')\right) 
- \phi_i(\theta) -  \frac{\lambda c_i}{C}- \log s_{i}!\right] \right)\right)\right\}
\end{align*}

Multiplying $\pi(\theta)$ to both sides produces
\begin{align*}
&\hspace{-2em}\pi(\theta)T( \theta,  \theta')\\
  &=
\frac{1}{Z}\exp\left(-\sum_i U_i(\theta) \right)T( \theta,  \theta')\\
  &=
\frac{1}{Z}\sum_s
\min\Bigg(q( \theta'| \theta) \exp\bigg(\sum_i  \bigg[s_{i}\log\left( \frac{\lambda c_i}{C}+ \phi_i(\theta)\right) \\
&\hspace{3em}- \frac{U_i(\theta)+U_i(\theta')}{2} - \frac{c_i}{2} M(\theta,\theta') -  \frac{\lambda c_i}{C}- \log s_{i}!\bigg] \bigg),\\
&\hspace{2em}q( \theta| \theta') 
\exp\bigg(\sum_i  \bigg[s_{i}\log\left( \frac{\lambda c_i}{C}+ \phi_i(\theta')\right) 
\\&\hspace{3em}- \frac{U_i(\theta)+U_i(\theta')}{2} - \frac{c_i}{2} M(\theta,\theta') -  \frac{\lambda c_i}{C}- \log s_{i}!\bigg] \bigg)\bigg)\Bigg).  
\end{align*}
It is clear that the expression is symmetric in $\theta$ and $\theta'$. Therefore the chain is reversible and its stationary distribution is $\pi(\theta)$. This proves the first part of the theorem.
To prove the second part of the theorem, the bound on the spectral gap, we continue to reduce the transition probability in the previous proof to
%
\begin{align*}
  &\hspace{-2em}\pi( \theta) T( \theta,  \theta')\\
  &=
\frac{1}{Z}\sum_s
\min\Bigg(q( \theta'| \theta) \exp\bigg(\sum_i  \bigg[s_{i}\log\left( \frac{\lambda c_i}{C}+ \phi_i(\theta)\right) \\&\hspace{3em}- \frac{U_i(\theta)+U_i(\theta')}{2} - \frac{c_i}{2} M(\theta,\theta') - s_i\log\frac{\lambda c_i}{C}\bigg] \bigg),
\\&\hspace{2em}q( \theta| \theta') 
\exp\bigg(\sum_i  \bigg[s_{i}\log\left( \frac{\lambda c_i}{C}+ \phi_i(\theta')\right) 
\\&\hspace{3em}- \frac{U_i(\theta)+U_i(\theta')}{2} - \frac{c_i}{2} M(\theta,\theta') -  s_i\log\frac{\lambda c_i}{C}\bigg] \bigg)\Bigg)
\\&\hspace{2em}\cdot \prod_{i} \frac{1}{s_{i}!}\exp\left(- \frac{\lambda c_{i}}{C}\right)\left( \frac{\lambda c_{i}}{C}\right)^{s_{i}}\\
  &=
\frac{1}{Z}\sum_s
\min\Bigg(q( \theta'| \theta) \exp\bigg(\sum_i  \bigg[s_{i}\log\left(1 + \frac{C}{\lambda c_i} \phi_i(\theta)\right) 
\\&\hspace{3em}- \frac{U_i(\theta)+U_i(\theta')}{2} - \frac{c_i}{2} M(\theta,\theta')\bigg] \bigg),\\
&\hspace{2em}q( \theta| \theta') 
\exp\left(\sum_i  \left[s_{i}\log\left(1 + \frac{C}{\lambda c_i} \phi_i(\theta')\right) - \frac{U_i(\theta)+U_i(\theta')}{2} - \frac{c_i}{2} M(\theta,\theta') \right] \right)\Bigg)
\\&\hspace{2em}\cdot \prod_{i} \frac{1}{s_{i}!}\exp\left(- \frac{\lambda c_{i}}{C}\right)\left( \frac{\lambda c_{i}}{C}\right)^{s_{i}}.
\vspace{-.2cm}
\end{align*}
Note that $s_{i}$ here are non-negative integers that a Poisson variable can take, not variables. So if we let $r_{i} \sim \text{Poisson}\left( \frac{\lambda c_{i}}{C} \right)$ and $r_{i}$ to be all independent, we can write this as
\begin{align*}
  \pi( \theta) T( \theta,  \theta')
  &=
\frac{1}{Z}\mathbf{E}
\min\left(q( \theta'| \theta) \exp\left(\sum_{i} r_{i}\log\left( 1+ \frac{C}{\lambda c_{i}}\phi_i( \theta) \right)\right),\right.\\
&\left.q( \theta| \theta') 
\exp\left(\sum_{i} r_{i}\log\left( 1+ \frac{C}{\lambda c_{i}}\phi_i( \theta')\right)\right)\right) 
\\&\hspace{2em}\cdot \exp\bigg[ -\frac{1}{2}\bigg( \sum_i U_i(\theta) + \sum_i U_i(\theta') + C M(\theta,\theta')\bigg)\bigg].
\end{align*}
Assume $G( \theta,  \theta')$ is the transition operator of standard MH. Consider the ratio
\begin{align*}
\frac{\pi( \theta) T( \theta,  \theta')}
  {\pi( \theta) G( \theta,  \theta')}
&=
\frac{1}{Z}\mathbf{E}
\min\left(q( \theta'| \theta) \exp\left(\sum_{i} r_{i}\log\left( 1+ \frac{C}{\lambda c_{i}}\phi_i( \theta) \right)\right),\right.\\
&\left.q( \theta| \theta') 
\exp\left(\sum_{i} r_{i}\log\left( 1+ \frac{C}{\lambda c_{i}}\phi_i( \theta')\right)\right)\right) 
\\
&\cdot \exp\bigg[ -\frac{1}{2}\bigg( \sum_i U_i(\theta) + \sum_i U_i(\theta') + C M(\theta,\theta')\bigg)\bigg]
\\
&\cdot\Bigg[1\bigg/\Bigg(\frac{1}{Z}
\min\left(q( \theta'| \theta)\exp\left(-\sum_i U_i( \theta)\right),
q( \theta| \theta')\exp\left(-\sum_i U_i( \theta')\right) \right)\Bigg)\Bigg].
\end{align*}

We know that $\frac{\min(A, B)}{\min(C,D)} = \min\left(\frac{A}{\min(C,D)}, \frac{B}{\min(C,D)}\right) \geq \min\left(\frac{A}{C}, \frac{B}{D}\right)$. The last inequality is due to the fact that $\frac{1}{\min(C,D)}\geq \frac{1}{C}$ and $\frac{1}{\min(C,D)}\geq \frac{1}{D}$.

With this inequality, we can continue simplifying the ratio,
\begin{align*}
  &\hspace{-1em}\frac{\pi( \theta) T( \theta,  \theta')}
  {\pi( \theta) G( \theta,  \theta')}\\
&\geq
\mathbf{E}\Bigg[\min \Bigg(\frac{
\exp\left(\sum_i r_i\log\left( 1+ \frac{C}{\lambda c_{i}}\phi_i( \theta) \right)  \right)}
{
\exp\left(-\sum_i U_i( \theta)\right)},
\frac{
\exp\left(\sum_i r_i\log\left( 1+ \frac{C}{\lambda c_{i}}\phi_i( \theta')\right)  \right)}
{
\exp\left(-\sum_i U_i( \theta')\right)}
\Bigg)\Bigg]
\\&\hspace{2em}\cdot \exp\bigg[ -\frac{1}{2}\bigg( \sum_i U_i(\theta) + \sum_i U_i(\theta') + C M(\theta,\theta')\bigg)\bigg]\\
&=
\mathbf{E}\Bigg[\min \Bigg(
\exp\left(\sum_i\Bigg( r_i\log\left( 1+ \frac{C}{\lambda c_{i}}\phi_i( \theta) \right) - \phi_i( \theta) \Bigg)\right),
\\&\hspace{2em}\exp\left(\sum_i\Bigg( r_i\log\left( 1+ \frac{C}{\lambda c_{i}}\phi_i( \theta')\right) - \phi_i( \theta') \Bigg)\right)
\Bigg)\Bigg]\\
&=
\mathbf{E}\Bigg[\max \Bigg(
\exp\left(\sum_i \Bigg(\phi_i( \theta) - r_i\log\left( 1+ \frac{C}{\lambda c_{i}}\phi_i( \theta) \right)\Bigg)\right),
\\&\hspace{2em}\exp\left(\sum_i \Bigg(\phi_i( \theta') - r_i\log\left( 1+ \frac{C}{\lambda c_{i}}\phi_i( \theta')\right)\Bigg) \right)
\Bigg)^{-1}\Bigg].
\end{align*}

Because $f(x) = \frac{1}{x}$ is a convex function, by Jensen's inequality it follows
\begin{align*}
  \frac{\pi( \theta) T( \theta,  \theta')}
  {\pi( \theta) G( \theta,  \theta')}
&\geq
\mathbf{E}\Bigg[\max \Bigg(
\exp\left(\sum_i \Bigg(\phi_i( \theta) - r_i\log\left( 1+ \frac{C}{\lambda c_{i}} \phi_i( \theta) \right)\Bigg) \right),
\\&\hspace{2em}\exp\left(\sum_i\Bigg( \phi_i( \theta') - r_i\log\left(1+ \frac{C}{\lambda c_{i}}\phi_i( \theta')\right) \Bigg)\right)
\Bigg)\Bigg]^{-1}.
\end{align*}

We use $\max(A,B)\leq (A^p + B^p)^{\frac{1}{p}}$ to remove the $\max$ function.
\begin{align*}
  \frac{\pi( \theta) T( \theta,  \theta')}
  {\pi( \theta) G( \theta,  \theta')}
&\geq
\mathbf{E}\Bigg[\bigg(
\exp \bigg(p\sum_i\Bigg(\phi_i( \theta) - r_i\log\left( 1+ \frac{C}{\lambda c_{i}} \phi_i( \theta) \right)\Bigg)\bigg) +
\\&\hspace{2em}\exp\bigg(p\sum_i\Bigg( \phi_i( \theta') - r_i\log\left(1+ \frac{C}{\lambda c_{i}}\phi_i( \theta')\right)  \Bigg)\bigg)\bigg)^{\frac{1}{p}}\Bigg]^{-1}.
\end{align*}
Since $x^{\frac{1}{p}}$ is concave, by Jensen's inequality
\begin{align*}
  \frac{\pi( \theta) T( \theta,  \theta')}
  {\pi( \theta) G( \theta,  \theta')}
&\geq
\mathbf{E}\Bigg[
\exp \bigg(p\sum_i\Bigg(\phi_i( \theta) - r_i\log\left( 1+ \frac{C}{\lambda c_{i}} \phi_i( \theta) \right)\Bigg)\bigg) +
\\&\hspace{2em}\exp\bigg(p\sum_i\Bigg( \phi_i( \theta') - r_i\log\left(1+ \frac{C}{\lambda c_{i}}\phi_i( \theta')\right)  \Bigg)\bigg)\Bigg]^{-\frac{1}{p}}\\
&=
\Bigg[\prod_i\mathbf{E}
\exp \Bigg(p\phi_i( \theta) - pr_i\log\left( 1+ \frac{C}{\lambda c_{i}} \phi_i( \theta) \right)\Bigg) +
\\&\hspace{2em}\prod_i\mathbf{E}\exp\Bigg( p\phi_i( \theta') - pr_i\log\left(1+ \frac{C}{\lambda c_{i}}\phi_i( \theta')\right)  \Bigg)\Bigg]^{-\frac{1}{p}}.
\end{align*}

$\mathbf{E}\Bigg[
\exp \Bigg( - pr_i\log\left( 1+ \frac{C}{\lambda c_{i}} \phi_i( \theta) \right)\Bigg)\Bigg]$ is the moment generating function of the Poisson random variable $r_i$ evaluated at
\[
t = -p\log\left( 1+ \frac{C}{\lambda c_{i}} \phi_i( \theta) \right).
\]
We know that
\begin{align*}
\mathbf{E}\exp(r_it) 
&= 
\exp\left( \frac{\lambda c_{i}}{C}\left(\exp(t) - 1\right)\right),
\end{align*}
therefore,
\begin{align*}
\mathbf{E}\Bigg[
\exp \Bigg( - pr_i\log\left( 1+ \frac{C}{\lambda c_{i}} \phi_i( \theta) \right)\Bigg)\Bigg]
&= 
\exp\left( \frac{\lambda c_{i}}{C}\left(1+ \frac{C}{\lambda c_{i}} \phi_i( \theta)\right)^{-p} - \frac{\lambda c_{i}}{C}\right).
\end{align*}
Substituting this into the original expression produces
\begin{align*}
  \frac{\pi( \theta) T(\theta, \theta')}
  {\pi( \theta) G(\theta, \theta')}
  &\geq
  \Bigg[\prod_i\exp\left( \frac{\lambda c_{i}}{C}\left(1+ \frac{C}{\lambda c_{i}} \phi_i( \theta)\right)^{-p} - \frac{\lambda c_{i}}{C} + p\phi_i(\theta)\right) \\&\hspace{4em}+ \prod_i\exp\left( \frac{\lambda c_{i}}{C}\left(1+ \frac{C}{\lambda c_{i}} \phi_i( \theta')\right)^{-p} - \frac{\lambda c_{i}}{C} + p\phi_i(\theta')\right)\Bigg]^{-\frac{1}{p}}.
\end{align*}

Considering the term inside $\exp$. Define a function $f(y) = \frac{\lambda c_{i}}{C}\left(1+ \frac{C}{\lambda c_{i}}y\right)^{-p} - \frac{\lambda c_{i}}{C} + py$ for $y\ge 0$. It is clear that $f(0) = 0$. The first derivative is 
\[
f'(y) = p + (-p)\left(1 + \frac{C}{\lambda c_i}y\right)^{-p-1}
\]
which is also 0 at $y=0$. The second and third derivatives are
\begin{align}\label{eq:second-derivative}
f''(y) &= (-p)(-p-1)\frac{C}{\lambda c_i}\left(1 + \frac{C}{\lambda c_i}y\right)^{-p-2}, \\
f'''(y) &= (-p)(-p-1)(-p-2)\left(\frac{C}{\lambda c_i}\right)^2\left(1 + \frac{C}{\lambda c_i}y\right)^{-p-3}.    \label{eq:third-derivative}
\end{align}

By Taylor series, we have
\begin{align*}
    f(y) = f(0) + f'(0)y + \frac{f''(0)}{2!}y^2 + \frac{f'''(v)}{3!}y^3
\end{align*}
where $v$ is between 0 and $y$. By (\ref{eq:third-derivative}), we know that $f'''(v)\le 0$, therefore since $y \ge 0$, we have
\begin{align*}
    f(y) &\le f(0) + f'(0)y + \frac{f''(0)}{2!}y^2\\
    &=\frac{f''(0)}{2!}y^2.
\end{align*}
Substituting $y=\phi_i(\theta)$ produces
\begin{align*}
    f(\phi_i(\theta)) &\le (-p)(-p-1)\frac{C}{\lambda c_i}\phi_i^2(\theta)\\
    &\le (-p)(-p-1)\frac{C}{\lambda c_i}c_i^2M^2(\theta,\theta').
\end{align*}
Similarly, we can get 
\begin{align*}
    f(\phi_i(\theta'))
    &\le p(p+1)\frac{C}{\lambda c_i}c_i^2M^2(\theta,\theta').
\end{align*}
Substituting these to the spectral ratio, we get
\begin{align*}
  \frac{\pi( \theta) T(\theta, \theta')}
  {\pi( \theta) G(\theta, \theta')}
&\geq
\left[2\prod_i\exp\left( p(p+1)\frac{C}{\lambda c_i}c_i^2M^2(\theta,\theta')\right)\right]^{-\frac{1}{p}}\\
&=
\left[2\exp\left(\sum_i p(p+1)\frac{C}{\lambda }c_i M^2(\theta,\theta')\right)\right]^{-\frac{1}{p}}\\
&=
\left[2\exp\left( p(p+1)\frac{C^2}{\lambda } M^2(\theta,\theta')\right)\right]^{-\frac{1}{p}}\\
&=
2^{-\frac{1}{p}}\exp\left(-(p+1)\frac{C^2}{\lambda } M^2(\theta,\theta')\right).
\end{align*}
Now, we maximize the R.H.S. with respect to $p$. Let $E=\frac{C^2}{\lambda } M^2(\theta,\theta')$, then it becomes
\begin{align*}
  2^{-\frac{1}{p}}\exp\left(-(p+1)E\right) &= \exp\left(-E-pE - \frac{1}{p}\log 2\right).  
\end{align*}
The maximum is attained at $p = \sqrt{\frac{\log 2}{E}}$ and the value is 
\begin{align*}
 \exp\left(-E-2\sqrt{E\log 2}\right).
\end{align*}
It follows that
\begin{align*}
  \frac{\pi( \theta) T(\theta, \theta')}
  {\pi( \theta) G(\theta, \theta')}
&\geq
\exp\left(-\frac{C^2}{\lambda } M^2(\theta,\theta')-2\sqrt{\frac{C^2}{\lambda } M^2(\theta,\theta')\log 2}\right). 
\end{align*}
We set $\lambda = \chi C^2M^2(\theta,\theta')$, it becomes 
  \[
      \frac{\pi( \theta) T( \theta,  \theta')}
  {\pi( \theta) G( \theta,  \theta')}
    \ge
    \exp \Bigg(-\frac{1}{\chi} - 2\sqrt{\frac{\log 2}{\chi}} \Bigg).
  \]

We complete the theorem by a Dirichlet form argument. We can write the Dirichlet form $\mathcal{E}(f)$ of a Markov chain with transition operator $G$ as \citep{fukushima2010dirichlet}:
\begin{align*}
\mathcal{E}(f) 
	= 
	\frac{1}{2}\int\int\left[\left(f(\theta)-f(\theta')\right)^2\right]G(\theta,\theta')\pi(\theta)d\theta d\theta'.
\end{align*}
If we let $L^2_0(\pi)$ to be the Hilbert space of functions $f$ such that $f$ has mean zero and is square integrable with respect to probability measure $\pi$. It follows that the spectral gap $\gamma$ of a Markov chain is \citep{aida1998uniform}
\[
\gamma = \inf_{f\in L^2_0(\pi): Var_{\pi}[f] = 1} \mathcal{E}(f).
\]
From this, it is easy to get that
\begin{align*}
\bar\gamma &= \inf_{f\in L^2_0(\pi): Var_{\pi}[f] = 1} \left[\frac{1}{2}\int\int\left[\left(f(\theta)-f(\theta')\right)^2\right]T(\theta,\theta')\pi(\theta)d\theta d\theta'\right]\\
&\geq 
\exp \Bigg(-\frac{1}{\chi} - 2\sqrt{\frac{\log 2}{\chi}} \Bigg)
\cdot \inf_{f\in L^2_0(\pi): Var_{\pi}[f] = 1} \left[\frac{1}{2}\int\int\left[\left(f(\theta)-f(\theta')\right)^2\right] G(\theta,\theta')\pi(\theta)d\theta d\theta'\right]\\
&= \exp \Bigg(-\frac{1}{\chi} - 2\sqrt{\frac{\log 2}{\chi}} \Bigg)\cdot \gamma.
\end{align*}
\end{proof}

%% file: section/99-appendix/31-tunamh/60-app-tunamh-chi.tex
\section{Derivation of Equation~(\ref{eq:TunaMHEB})}\label{app:chi-value}

Based on the bound in Theorem~\ref{thm:spectral-gap}, to make sure that the spectral ratio $\bar\gamma/\gamma \ge \kappa$, we can set $\chi$ such that
\[
\exp \Bigg(-\frac{1}{\chi} - 2\sqrt{\frac{\log 2}{\chi}} \Bigg) = \kappa.
\]
Solving the above equation gives us
\[
\chi = \frac{(2\log2-\log \kappa+2\sqrt{\log2(\log2-\log\kappa)})}{\log^2\kappa}\le \frac{4}{(1-\kappa)\log(1/\kappa)}.
\]
Since the spectral gap ratio is monotonically increasing w.r.t. $\chi$, we can instead set $\chi$ to the upper bound
\[
\chi = \frac{4}{(1-\kappa)\log(1/\kappa)}
\]

%% file: section/99-appendix/31-tunamh/70-app-tunamh-optimal-chi.tex
\section{Theoretically Optimal Value of $\chi$}\label{app:optimal-value}

The overall wall-clock time $L$ for a chain to converge can be represented as the number of steps times the wall-clock time $l$ of each step. We then minimize an upper bound of this overall wall-clock time to get the optimal value of $\chi$. 

Consider a lazy Markov chain on a finite state $\Theta$. The \emph{relaxation time} $t_{\text{rel}}$ of a Markov chain is defined to be the inverse of the spectral gap $\gamma$: $t_{\text{rel}} = 1/\gamma$. 
The \emph{mixing time} $t_{\text{mix}}$, i.e. the number of steps required for a chain to converge to within TV distance $\delta$ to the target distribution $\pi$, is bounded by Levin and Peres~\cite{levin2017markov}
\[
t_{\text{mix}}\le t_{\text{rel}} \log\left(\frac{1}{\delta\cdot\min_{\theta\in\Theta}\pi(\theta)}\right).
\]

It follows that the overall wall-clock time $L$ is upper bounded by
\[
L = l \cdot t_{\text{mix}}\le l\cdot t_{\text{rel}}\log\left(\frac{1}{\delta\cdot\min_{\theta\in\Theta}\pi(\theta)}\right).
\]

We assume that the expected wall clock time to run a step is proportional to the batch size plus some constant, which measures the cost of computing the proposal. Specifically, We use $\eta$ and $\xi$ to denote the time to get a proposal $\theta'$ and compute a $U_i$ in a step. Then we can write the time of a step $l$ as 
\[
l = B\xi + \eta.
\]

In order to minimize $L$, we can instead minimize its upper bound, which is equivalent to minimize
\begin{align}\label{eq:wall-clock-time}
    l\cdot t_{\text{rel}} = (B\xi + \eta)\cdot \frac{1}{\gamma}.
\end{align}

Recall that for \methodname{}, the average batch size over all steps is 
\[
\mathbf{E}_{(\theta,\theta')\sim \pi(\theta)q(\theta'|\theta)}[\chi C^2M^2(\theta,\theta') + CM(\theta,\theta')],
\]

and the spectral gap $\bar \gamma$ is lower bounded by the spectral gap of standard MH $\gamma$ such that
\[
    \bar{\gamma}
    \ge
    \exp \Bigg(-\frac{1}{\chi} - 2\sqrt{\frac{\log 2}{\chi}} \Bigg)\cdot\gamma.
\]

Substituting the expression of batch size and spectral gap to (\ref{eq:wall-clock-time}) gives
\[
l\cdot t_{\text{rel}} \le \left(\mathbf{E}_{(\theta,\theta')\sim \pi(\theta)q(\theta'|\theta)}[\chi C^2M^2(\theta,\theta') + CM(\theta,\theta')]\xi + \eta\right)\cdot \exp \Bigg(\frac{1}{\chi} + 2\sqrt{\frac{\log 2}{\chi}} \Bigg)\cdot\frac{1}{\gamma}.
\]

To minimize the RHS of the above equation over $\chi$, we let the derivative w.r.t. $\chi$ to be zero and get,
\begin{align*}
    &\hspace{-2em}\xi C^2\mathbf{E}_{(\theta,\theta')\sim \pi(\theta)q(\theta'|\theta)}[M^2(\theta,\theta')] \chi^{-1} + (\xi C\mathbf{E}_{(\theta,\theta')\sim \pi(\theta)q(\theta'|\theta)}[M(\theta,\theta')] + \eta)\chi^{-2} \\&+ \sqrt{\log 2}\xi C^2\mathbf{E}_{(\theta,\theta')\sim \pi(\theta)q(\theta'|\theta)}[M^2(\theta,\theta')]\chi^{-\frac{1}{2}} \\&+ \sqrt{\log 2}(\xi C\mathbf{E}_{(\theta,\theta')\sim \pi(\theta)q(\theta'|\theta)}[M(\theta,\theta')] + \eta)\chi^{-\frac{3}{2}} 
    \\&= \xi C^2\mathbf{E}_{(\theta,\theta')\sim \pi(\theta)q(\theta'|\theta)}[M^2(\theta,\theta')].
\end{align*}

When $\chi$ is small, the LHS is approximately $(\xi C\mathbf{E}_{(\theta,\theta')\sim \pi(\theta)q(\theta'|\theta)}[M(\theta,\theta')] + \eta)\chi^{-2}$ which gives us  
\[
\chi = 
\sqrt{\frac{\xi C\mathbf{E}_{(\theta,\theta')\sim \pi(\theta)q(\theta'|\theta)}[M(\theta,\theta')] + \eta}{\xi C^2\mathbf{E}_{(\theta,\theta')\sim \pi(\theta)q(\theta'|\theta)}[M^2(\theta,\theta')]}}.
\]

When it is quick to get a proposal ($\eta\approx 0$) and the variance of $M$ is small, we can further simplify it to
\[
\chi = \frac{1}{ \sqrt{C\mathbf{E}_{(\theta,\theta')\sim \pi(\theta)q(\theta'|\theta)}[M(\theta,\theta')]}}.
\]

In practice, we can get the above theoretically optimal value of $\chi$ by empirically estimating the mean and variance of $M(\theta,\theta')$. Note that even if these empirical estimates are accurate, there may exist better $\chi$, since the upper bounds (the mixing time bound and the spectral gap bound) we use to get the optimal value may be loose. We give a simpler heuristic to tune $\chi$ in practice in Section \ref{sec:tunamh:exp}.

%% file: section/99-appendix/31-tunamh/80-app-tunamh-optimal.tex
\section{Proof of Theorem \ref{thm:optimality}}\label{app:proof:optimality}

First, we will show the following lemma, which gives half of what we want to have in the theorem.

\begin{lemma}\label{lemma:optimality1}
Considering the same setting as the theorem, the average batch size $B$ of any exact, stateless minibatch MH algorithm at any iteration follows
\[
    \mathbf{E}[B] \ge 2^{-18} \cdot \kappa C^2 M^2(\theta,\theta') - 2^{-4} \cdot \kappa.
\]
\end{lemma}
\begin{proof}
We prove the lemma by construction. First, observe that since the state space $\Theta$ has at least two states, we can restrict our attention to just two of those states, by choosing a $\pi$ that has zero mass on any other state in the space and a $q$ that never proposes transitioning out to any of those other states (at which $\pi$ has zero mass). Such a proposal will still be ergodic, so it still satisfies our general assumption that we consider only ergodic chains in this paper. Without loss of generality, suppose that those two states are $\{-\frac{M}{2}, \frac{M}{2}\}$ (this is without loss of generality because we can always just rename the states), and let $C$ denote the constant in the theorem statement and define (with a bit of abuse of notation) the constant $M := M(-\frac{M}{2}, \frac{M}{2})$. By doing this, we can (again without loss of generality) restrict our attention to the case where $\Theta = \{-\frac{M}{2}, \frac{M}{2}\}$.

Next, we construct our counterexample. Let the dataset be $\{x_i\}_{i=1}^N$ where $x_i \in \{-1, 1\}$.
We let the domain for parameter $\theta$ to be $\{-\frac{M}{2}, \frac{M}{2}\}$, and the target distribution to be   
\[
\pi(\theta) = \frac{1}{Z}\exp\left(-\sum_{i=1}^N U_i(\theta)\right) = \frac{1}{Z}\exp\left(-\frac{C\theta}{N}\sum_{i=1}^N x_i\right)
\]
where $U_i(\theta) = \frac{C}{N}\cdot\theta x_i$.
Note that by letting $N$ become large, any minibatch MH algorithm that queries the energy difference oracle some number of times will observe a distribution of energy differences that is arbitrarily close to a sequence of independent identically distributed random variables supported on $\{\pm \frac{CM}{N}\}$.

We define $c_i = \frac{C}{N}$, and the proposal distribution to be
\[
p(\theta,\theta) = \frac{1}{2},\hspace{2em} p(\theta,-\theta) = \frac{1}{2} \hspace{2em}\text{for }\theta \in \bigg\{-\frac{M}{2}, \frac{M}{2}\bigg\}.
\]
Now, let $0 < q < 1$ be some constant, and consider two cases: (1) $\frac{1}{N}\sum_i x_i = q$ and (2) $\frac{1}{N}\sum_i x_i = -q<0$. Suppose that in both cases the $x_i$ are shuffled at random. These two cases will have different stationary distributions,
\[
    \pi_1(\theta) = \frac{1}{Z} \exp\left(-C q \theta\right)
    \hspace{2em}\text{and}\hspace{2em}
    \pi_2(\theta) = \frac{1}{Z} \exp\left(C q \theta\right),
\]
and an exact algorithm must be able to distinguish between them. Therefore by using these cases, we can get a bound on the required batch size needed for the exact MH algorithm to distinguish between them.
First, we observe that the two cases are symmetric, such that if $T_1$ is the transition matrix of the chain in case (1) and $T_2$ is the transition matrix of the chain in case (2), then $T_1(\theta, \theta') = T_2(\theta', \theta)$.
Let $0 < \psi < \frac{1}{2}$ denote the probability that $T_1$ transitions from $\frac{M}{2}$ to $-\frac{M}{2}$.
Then because the MH method is exact and the chain is reversible, the probability of the reverse transition is $\psi \exp(-CMq)$.
So, explicitly, the transition operators will look like
\[
    T_1 = \begin{bmatrix} 1 - \psi & \psi e^{-CMq} \\ \psi & 1 - \psi e^{-CMq} \end{bmatrix}
    \hspace{2em}\text{and}\hspace{2em}
    T_2 = \begin{bmatrix} 1 - \psi e^{-CMq} & \psi \\ \psi e^{-CMq} & 1 - \psi \end{bmatrix}.
\]
The eigenvectors and eigenvalues of this are
\[
    T_1 \pi_1
    =
    \pi_1
    \hspace{2em}\text{and}\hspace{2em}
    T_1 \begin{bmatrix} -1 \\ 1 \end{bmatrix}
    =
    \left(1 - \psi - \psi \exp(-CMq) \right)
    \begin{bmatrix} -1 \\ 1 \end{bmatrix}.
\]

Suppose that we initialize both chains uniformly on $\{-\frac{M}{2}, \frac{M}{2}\}$.
Observe that
\[
    \begin{bmatrix} 1/2 \\ 1/2 \end{bmatrix}
    =
    \begin{bmatrix} \frac{\exp(-CMq)}{1 + \exp(-CMq)} \\ \frac{1}{1 + \exp(-CMq)} \end{bmatrix}
    +
    \frac{1 - \exp(-CMq)}{2 (1 + \exp(-CMq))}
    \cdot
    \begin{bmatrix} 1 \\ -1 \end{bmatrix},
\]
the first vector being $\pi_1$ and the second being a multiple of the other eigenvector.

Equivalently,
\[
    \begin{bmatrix} 1/2 \\ 1/2 \end{bmatrix}
    =
    \pi_1
    +
    \frac{1}{2}
    \tanh\left( \frac{CMq}{2} \right)
    \cdot
    \begin{bmatrix} 1 \\ -1 \end{bmatrix},
\]
and so for any $t$, after $t$ steps of the Markov chain, the distribution will be
\[
    T_1^t
    \begin{bmatrix} 1/2 \\ 1/2 \end{bmatrix}
    =
    \pi_1
    +
    \frac{1}{2}
    \tanh\left( \frac{CMq}{2} \right)
    \cdot
    \left(1 - \psi - \psi \exp(-CMq) \right)^t
    \cdot
    \begin{bmatrix} 1 \\ -1 \end{bmatrix}.
\]
Similarly,
\[
    T_2^t
    \begin{bmatrix} 1/2 \\ 1/2 \end{bmatrix}
    =
    \pi_2
    +
    \frac{1}{2}
    \tanh\left( \frac{CMq}{2} \right)
    \cdot
    \left(1 - \psi - \psi \exp(-CMq) \right)^t
    \cdot
    \begin{bmatrix} -1 \\ 1 \end{bmatrix}.
\]

So, the total variation distance between the state of the chains at time $t$ will be bounded by
\[
    \text{TV}\left( 
    T_1^t
    \begin{bmatrix} 1/2 \\ 1/2 \end{bmatrix},
    T_2^t
    \begin{bmatrix} 1/2 \\ 1/2 \end{bmatrix}
    \right)
    \ge
    \text{TV}\left( \pi_1, \pi_2 \right)
    -
    \tanh\left( \frac{CMq}{2} \right)
    \cdot
    \left(1 - \psi - \psi \exp(-CMq) \right)^t.
\]
Also observe that
\[
    \text{TV}\left( \pi_1, \pi_2 \right)
    =
    \frac{1}{2}
    \left\|
    \begin{bmatrix} \frac{\exp(-CMq)}{1 + \exp(-CMq)} \\ \frac{1}{1 + \exp(-CMq)} \end{bmatrix}
    -
    \begin{bmatrix} \frac{1}{1 + \exp(-CMq)} \\ \frac{\exp(-CMq)}{1 + \exp(-CMq)} \end{bmatrix} \right\|_1
    =
    \frac{1 - \exp(-CMq)}{1 + \exp(-CMq)}
    =
    \tanh\left( \frac{CMq}{2} \right),
\]
so
\[
    \text{TV}\left( 
    T_1^t
    \begin{bmatrix} 1/2 \\ 1/2 \end{bmatrix},
    T_2^t
    \begin{bmatrix} 1/2 \\ 1/2 \end{bmatrix}
    \right)
    \ge
    \tanh\left( \frac{CMq}{2} \right)
    \cdot
    \left(
    1
    -
    \left(1 - \psi - \psi \exp(-CMq) \right)^t
    \right).
\]
Also, since we know that our algorithm is guaranteed to have spectral gap ratio at least $\kappa$ with the original chain, it follows that $\psi \ge \kappa / 2$, and so
\[
    \text{TV}\left( 
    T_1^t
    \begin{bmatrix} 1/2 \\ 1/2 \end{bmatrix},
    T_2^t
    \begin{bmatrix} 1/2 \\ 1/2 \end{bmatrix}
    \right)
    \ge
    \tanh\left( \frac{CMq}{2} \right)
    \cdot
    \left(
    1
    -
    \left(1 - \frac{\kappa}{2} - \frac{\kappa}{2} \exp(-CMq) \right)^t
    \right).
\]

Now, denote the exact minibatch algorithm to be $\mathcal{A}$. As it runs, the algorithm $\mathcal{A}$ will request data examples by querying the energy difference oracle.
Under case (1), we let $y_i$ denote the $i$th sample that $\mathcal{A}$ \emph{would have observed} if it requested $i$ or more samples, and similarly we let $z_i$ denote the analogous sample in case (2).
Fix some constant $t \in \mathbf{N}$ (which we will set later).
We let $K_1$ denote the total number of samples observed by $\mathcal{A}$ across the first $t$ iterations in case (1), and set
\[
\mu = \{y_1, y_2,\dots, y_{K_1}\}.
\]
Similarly, we let $K_2$ denote the number of samples observed by $\mathcal{A}$ across the first $t$ iterations in case~(2), and set
\[
\nu = \{z_1, z_2, \dots, z_{K_2}\}.
\]
Now, we fix some constant $K$ (to be set later), and consider the following coupling between the behavior of $\mathcal{A}$ across its first $t$ iterations in case (1) and in case~(2). First, let all internal randomness of $\mathcal{A}$ and the proposal process under case (1) and (2) be the same, which means that for a given observation of data examples, the algorithm $\mathcal{A}$ will make the same decision, such as whether to require more data examples or not and whether to accept or not.
Second, choose a coupling that minimizes the probability that
\[
    (y_1, y_2, \ldots, y_{K1}) \ne (z_1, z_2, \ldots, z_{K2}).
\]
Such a coupling is guaranteed to exist by the Coupling Lemma, and the probability that these two are not equal will be equal to the total variation distance between their distributions.
Third, assign all the other $y_i$ and $z_i$, for $i > K$, independently according to their distribution.

We are interested in the quantity $p(\mu \ne \nu)$, which bounds the probability that the algorithm may make a different decision in cases (1) and (2).
We can decompose this probability into two terms,
\[
p(\mu \neq \nu) = p(\mu \neq \nu \text{ and } y_j = z_j \text{ for all } j\le K) + p(\mu \neq \nu \text{ and } y_j \neq z_j \text{ for some } j \le K ).
\]
If $\mu \ne \nu$ but $y_j = z_j$ for all $j \le K$, the only way that this is possible is for $K_1 > K$ (and, symmetrically, also $K_2 > K$), since otherwise the algorithms would behave identically.
So,
\begin{equation}\label{eq:prob_neq}
p(\mu \neq \nu) \le p(K_1 > K) + p(y_j \neq z_j \text{ for some } j \le K ).
\end{equation}
By Markov's inequality,
\[
p(\mu \neq \nu) \le \frac{\mathbf{E}[K_1]}{K} + p(y_j \neq z_j \text{ for some } j \le K ).
\]
For the second term of (\ref{eq:prob_neq}), we can reduce the case to only considering $K$ samples.
Let $S_y$ be the total number of samples $y_i$ that are $-1$ and let $S_z$ be the total number of samples $z_i$ that are $-1$.
Since $\mathcal{A}$ is effectively sampling a shuffled dataset at some arbitrary indices without replacement, both of these random variables $S_y$ and $S_z$ are---properly speaking---hypergeometric random variables. However, since our dataset size $N$ is arbitrary here, we can by setting $N$ very large work in the limit (as $N \rightarrow \infty$) in which these variables become binomial (since sampling with replacement and without replacement can be made to have arbitrarily close to the same distribution by making the dataset large).
Observe that (in this limit) $S_y$ follows a binomial distribution $B(K, \frac{1 - q}{2})$ and $S_z$ follows a binomial distribution $B(K, \frac{1 + q}{2})$.
Clearly, if $S_y = S_z$, then we can arrange the coupling so that $(y_1, \ldots, y_K) = (z_1, \ldots, z_K)$.
So, by the Coupling Lemma,
\begin{align*}
    p( y_j\neq z_j \text{ for some } j \le K) = p(S_y \neq S_z) = \text{TV}(S_y, S_z).
\end{align*}

From the analysis in Adell and Jodr\'a~\cite{adell2006exact}, we can bound the total variance distance between these two binomial variables with
\vspace{-.2cm}
\begin{align*}
    \text{TV}(S_y, S_z) \le \sqrt{e} \cdot\frac{\tau}{(1-\tau)^2}
\end{align*}
where $\tau = \sqrt{\frac{K+2}{2}}\cdot q<1$.
Substituting these bounds, we get
\begin{align*}
    p(\mu \neq \nu) &\le \frac{\mathbf{E}[K_1]}{K}
    + \sqrt{e} \cdot\frac{\tau}{(1-\tau)^2}.
\end{align*}
But the probability that $\mu \neq \nu$ must be an upper bound on the probability that the distributions of the chains in case (1) and (2) after $t$ steps are not equal, since if $\mu = \nu$ in the coupling then the two chains are in the same state.
So, using our bound from earlier, we get
\[
    \tanh\left( \frac{CMq}{2} \right)
    \cdot
    \left(
    1
    -
    \left(1 - \frac{1}{2} \kappa - \frac{1}{2} \kappa \exp(-CMq) \right)^t
    \right)
    \le
    \frac{\mathbf{E}[K_1]}{K}
    + \sqrt{e} \cdot\frac{\tau}{(1-\tau)^2}.
\]

Now isolating $\mathbf{E}[K_1]$ gives
\[
    K
    \cdot
    \tanh\left( \frac{CMq}{2} \right)
    \cdot
    \left(
    1
    -
    \left(1 - \frac{1}{2} \kappa - \frac{1}{2} \kappa \exp(-CMq) \right)^t
    \right)
    -
    K \cdot \sqrt{e} \cdot\frac{\tau}{(1-\tau)^2}
    \le
    \mathbf{E}[K_1].
\]

Also, observe that
\begin{align*}
    \left(1 - \frac{1}{2} \kappa - \frac{1}{2} \kappa \exp(-CMq) \right)^t
    \le
    \left(1 - \frac{1}{2} \kappa \right)^t
    \le
    \exp\left(-\frac{\kappa t}{2} \right),
\end{align*}

so
\[
    K
    \cdot
    \tanh\left( \frac{CMq}{2} \right)
    \cdot
    \left(
    1
    -
    \exp\left(-\frac{\kappa t}{2} \right)
    \right)
    -
    K \cdot \sqrt{e} \cdot\frac{\tau}{(1-\tau)^2}
    \le
    \mathbf{E}[K_1].
\]

This gives us the lower bound on $\mathbf{E}[K_1]$ that we are interested in.
Now, it remains to assign $q$, $K$, and $t$.
We start by assigning $t$ such that
\[
    t = \left\lceil 2 \kappa^{-1} \log(2) \right\rceil,
\]
in which case
\[
    \exp\left(-\frac{\kappa t}{2} \right) \le \frac{1}{2}
\]
and so
\[
    K
    \cdot
    \frac{1}{2}
    \cdot
    \tanh\left( \frac{CMq}{2} \right)
    -
    K \cdot \sqrt{e} \cdot\frac{\tau}{(1-\tau)^2}
    \le
    \mathbf{E}[K_1].
\]
Now, we add some simplifying assumptions, which we will validate are true later.
We assume that 
\[
    \tau = \sqrt{\frac{K+2}{2}} \cdot q \le \frac{1}{2};
\]
in this case
\[
    \sqrt{e} \cdot \frac{\tau}{(1-\tau)^2} \cdot K
    \le
    4 \sqrt{e} \cdot \tau
    \le
    5 \sqrt{K+2} \cdot q.
\]
We set $q$ such that
\[
    C M q = 1,
\]
and we assume that $CM$ is large enough that this assignment of $q$ is within range (i.e. $0 < q < 1$).
This gives us
\[
    K
    \cdot
    \frac{1}{2}
    \cdot
    \tanh\left( \frac{1}{2} \right)
    -
    5 K \sqrt{K+2} \cdot \frac{1}{CM}
    \le
    \mathbf{E}[K_1].
\]

Since $\tanh(1/2) > 5/16$, we can simplify this to
\[
    K
    \cdot
    \frac{5}{32}
    -
    5 K \sqrt{K+2} \cdot \frac{1}{CM}
    \le
    \mathbf{E}[K_1].
\]
All that remains is to assign $K$.
We assign $K$ such that
\[
    \sqrt{K+2} \cdot \frac{1}{CM}
    =
    \frac{1}{64}.
\]
In this case, we get
\[
    K = \frac{C^2 M^2}{4096} - 2,
\]
and our bound reduces to
\[
    \left( \frac{C^2 M^2}{4096} - 2 \right)
    \cdot
    \frac{5}{64}
    \le
    \mathbf{E}[K_1].
\]
We can simplify this further to
\[
    2^{-16} \cdot C^2 M^2 - \frac{5}{32}
    \le
    \mathbf{E}[K_1].
\]

Now, this is a bound on the expected number of samples taken across $t$ iterations.
This means that the number of samples taken in any given iteration will be bounded by
\[
    \frac{\mathbf{E}[K_1]}{t}
    \ge
    \frac{
        2^{-16} \cdot C^2 M^2 - \frac{5}{32}
    }{
        2 \kappa^{-1} \log(2) + 1
    }
    =
    \frac{
        2^{-16} \cdot \kappa C^2 M^2 - \frac{5 \kappa}{32}
    }{
        2 \log(2) + \kappa
    }.
\]
A few more loose bounds, leveraging $\kappa < 1$, gives us
\[
    \frac{\mathbf{E}[K_1]}{t}
    \ge
    2^{-18} \cdot \kappa C^2 M^2 - \frac{\kappa}{16}.
\]
This proves the lemma.
\end{proof}

Next, we will show the following lemma, which characterizes what happens when $C M$ is small.

\begin{lemma}
Considering minibatch MH algorithms in the same setting as the theorem, the expected batch size at any iteration must be lower bounded by
\[
    \mathbf{E}[B] \ge \frac{\kappa}{2} \min\left( C M(\theta, \theta'), 1 \right).
\]
\end{lemma}

\begin{proof}
Here, we will prove a lower bound that characterizes the limits of exact stateless minibatch MH algorithms when they use very few examples.
Again, without loss of generality we consider a reduction to the two-state case as we did in the proof of the previous lemma.
Suppose that a exact stateless minibatch MH algorithm with the same forward and backward proposal probabilities (given some $c_1,\ldots,c_N$, $C$, and $M$) requests any energy function examples at all only with probability $p$.
Consider two cases, which have the same $c_1,\ldots,c_N$, $C$ and $M$.
In the first case,
\vspace{-.2cm}
\begin{align*}
    \sum_{i=1}^n (U_i(\theta) - U_i(\theta')) &= C M(\theta, \theta'), \hspace{.25cm} \text{while in the second case,}\\
    \sum_{i=1}^n (U_i(\theta) - U_i(\theta')) &= -C M(\theta, \theta').
\end{align*}
These are clearly possible by setting $U_i$ to the limits of what is covered by the bounds.
In the first case, the baseline MH method would accept with probability $1$.
In the second case, it will accept with probability $\exp(-C M(\theta,\theta'))$.
Since the stateless MH algorithm is reversible, it must accept in the first case with some probability $a$ and in the second case with probability $a \cdot \exp(-C M(\theta,\theta'))$.
But, the algorithm can only distinguish the two cases if it requests samples, which only happens with probability at most $p$. So,
\[
    a - a \cdot \exp(-C M(\theta,\theta')) \le p.
\]

Since we know that it must be the case that $a \ge \kappa$ (from a straightforward analysis of a two-state case), it follows that
\[
    \frac{p}{\kappa} \ge \frac{p}{a} \ge 1 - \exp(-C M(\theta, \theta')) \ge \frac{1}{2} \min\left( C M(\theta, \theta'), 1 \right).
\]
Since $p$ is an obvious lower bound on the expected value of the batch size, it follows that
\vspace{-.2cm}
\begin{align*}
    \mathbf{E}[B] \ge \frac{\kappa}{2} \min\left( C M(\theta, \theta'), 1 \right).
\end{align*}
\vspace{-.2cm}
\end{proof}
To prove Theorem~\ref{thm:optimality} we now combine the results of these two lemmas.
We have
\begin{align*}
    \mathbf{E}[B] &\ge 2^{-18} \cdot \kappa C^2 M^2(\theta,\theta') - 2^{-4} \cdot \kappa, \hspace{.25cm} \text{and}\\
    \mathbf{E}[B] &\ge \frac{\kappa}{2} \min\left( C M(\theta, \theta'), 1 \right).
\end{align*}
Since these are both lower bounds, we can combine them to get
\begin{align*}
    \mathbf{E}[B] 
    &\ge 
    \max\left(2^{-18} \cdot \kappa C^2 M^2(\theta,\theta') - 2^{-4} \cdot \kappa, \frac{\kappa}{2} \min\left( C M(\theta, \theta'), 1 \right) \right) \\
    &=
    \kappa \cdot \max\left(2^{-18} \cdot C^2 M^2(\theta,\theta') - 2^{-4}, \frac{1}{2} \min\left( C M(\theta, \theta'), 1 \right) \right).
\end{align*}
It is obvious from a simple big-$\mathcal{O}$ analysis here that there exists a global constant $\zeta > 0$ such that
\vspace{-.2cm}
\[
    \mathbf{E}[B] \ge \zeta \cdot \kappa \left( C^2 M^2(\theta, \theta') + C M(\theta, \theta') \right).
\]
\vspace{-.2cm}
This proves the theorem.

\section{Proof of Corollary \ref{col:bound}} \label{app:proof:cor1}
\begin{proof}
Recall that the lower bound on the batch size in each iteration is
\[
\mathbf{E}[B] \ge \zeta \cdot \kappa \left( C^2 M^2(\theta, \theta') + C M(\theta, \theta') \right).
\]
Since $C = \bigTheta(N)$ and $M(\theta,\theta') = \bigTheta(N^{-(h+1)/2})$, the expectation of the batch size follows
\[
\mathbf{E}[B] = \bigTheta(C^2 M^2(\theta, \theta') + C M(\theta, \theta')) = \bigTheta(C M(\theta, \theta')) = \bigTheta(N^{1-h}/2).
\]

When $h = 1$, $\mathbf{E}[B] = \bigTheta(1)$ and when $h=2$, $\mathbf{E}[B] = \bigTheta(1/\sqrt{N})$.
\end{proof}

%% file: section/99-appendix/31-tunamh/90-app-experiments.tex
\section{Experimental Details and Additional Results}\label{app:tunamh:experiments}

\subsection{Experiment in Section \ref{sec:inexactproblems}}\label{app:experiments:counterexample}

To verify Theorem \ref{statement:counterexample}, we empirically construct a distribution in the form of Section \ref{app:proof:counterexample}, on which AustereMH and MHminibatch are biased. 
Note that the proof in Section \ref{app:proof:counterexample} shows there must exist such a distribution for any inexact minibatch method but does not tell us how to find one for a specific method. 
Therefore, in order to find such a distribution, we construct an example and empirically test whether AustereMH and MHminibatch are biased on it. 

We let data $x_i$ take one of two values $\{-1, 5\}$. Consider a dataset of size 6000. We let 5000 data take value $-1$ and the remaining 1000 data take value $5$. Define the target distribution $\pi(\theta)$ to be
\[
\pi(\theta) \propto \exp\left(-\frac{1}{N}\sum_{i=1}^N \theta\cdot x_i\right)
\]
where the domain of $\theta$ is $\{ 0, 1, \dots, K-1\}$. Therefore the number of state is $K$. Since $\sum_i x_i = 0$, it is clear to see that the stationary distribution of $\theta$ is a uniform distribution. We define the proposal distribution to be the following
\[
p(\theta,\theta) = \frac{1}{2},\hspace{1em}\text{for all }\theta;\hspace{1em} p(\theta, \theta-1) = \frac{1}{4},\hspace{1em} p(\theta, \theta+1) = \frac{1}{4} \hspace{1em}\text{for }\theta\in\{1,\dots,K-2\}; 
\] 
and $p(0,1)=p(K-1,K-2)=\frac{1}{2}$.

\begin{figure*}[t!]
    \centering
    \begin{tabular}{cccc}		
    	\includegraphics[width=5.5cm]{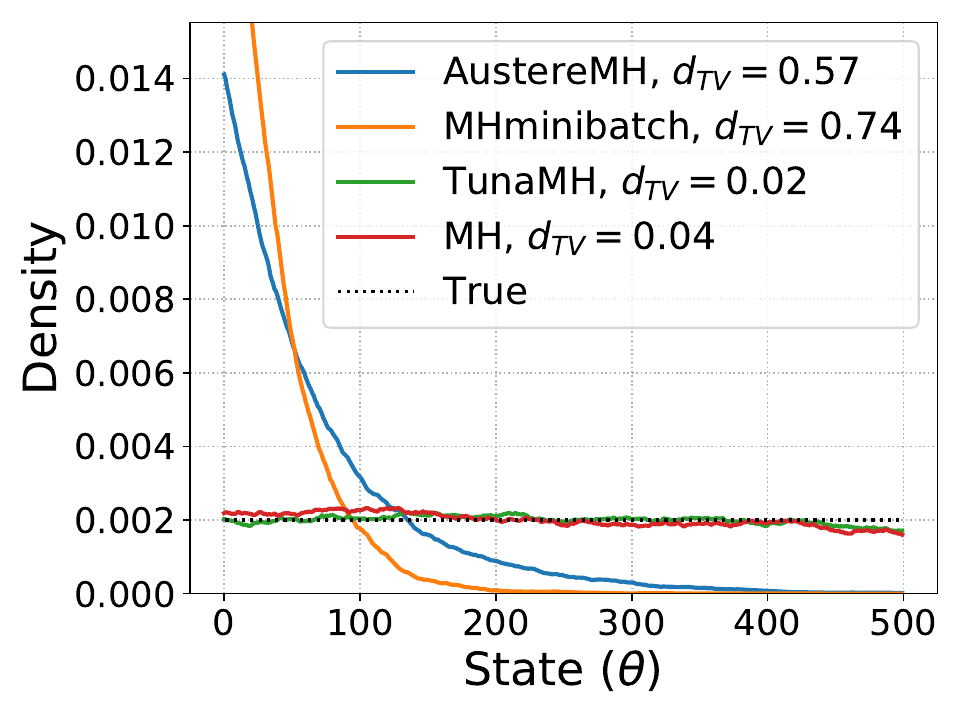}  &
    	\includegraphics[width=5.5cm]{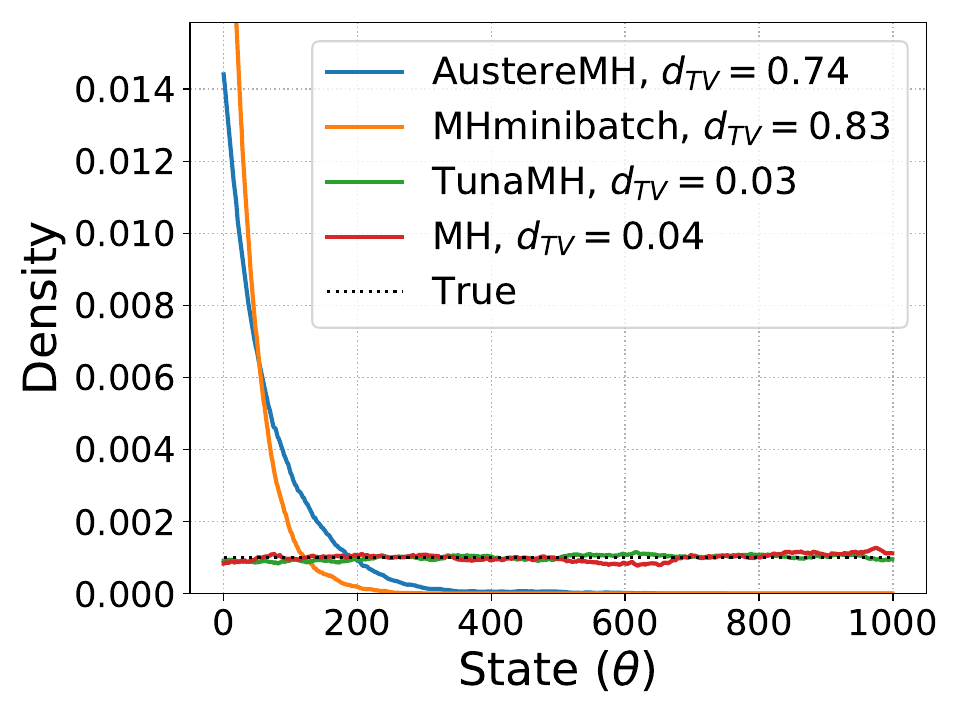}
    	\\		
    	(a) $K=500$&
    	(b) $K=1000$
    	\\
    	\includegraphics[width=5.5cm]{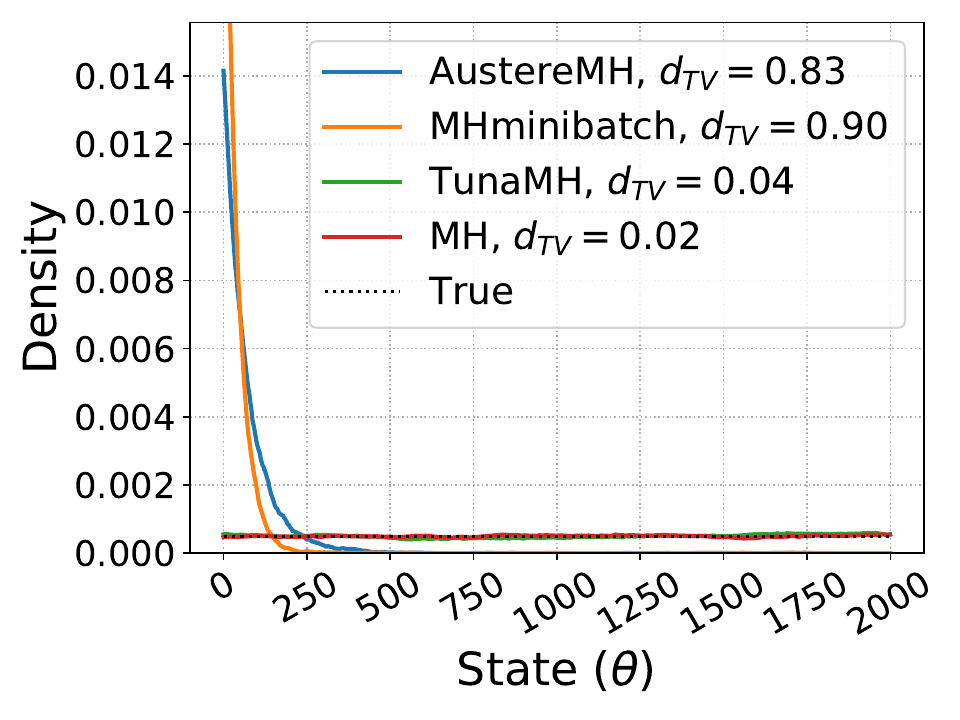}
    	&
    	\includegraphics[width=5.5cm]{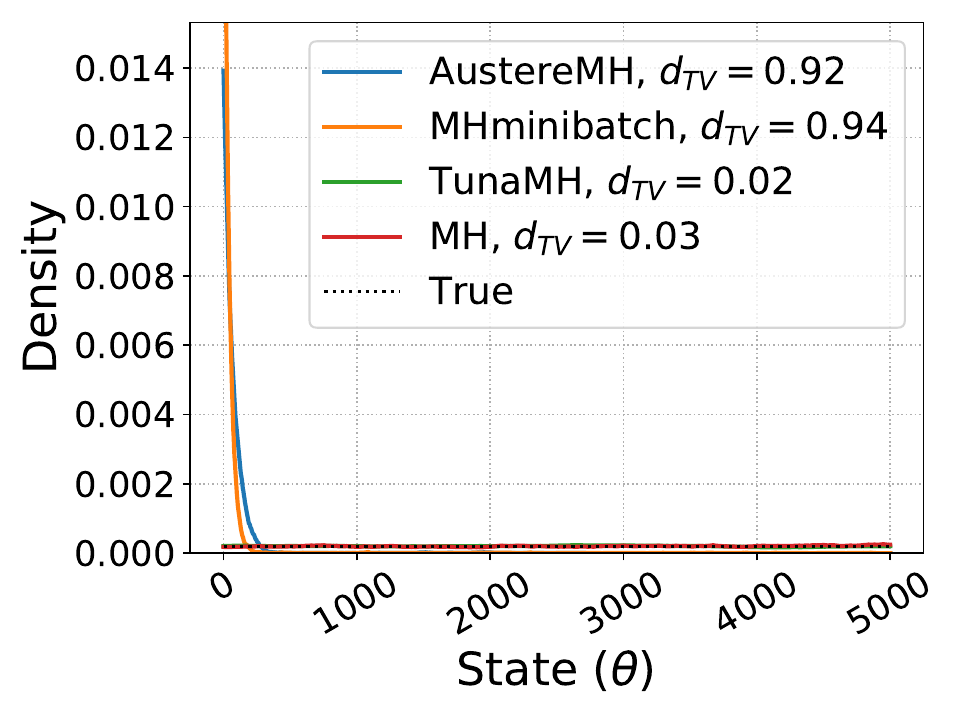}
    	
    	\\	
    	(e) $K=2000$&
    	(e) $K=5000$&
    	\hspace{-0mm}\\		
    \end{tabular}
    \caption{Density estimate comparison on $K = 500, 1000, 2000, 5000$.}
    \label{app:fig:density}
\end{figure*}

We set the hyperparameter error $\epsilon$ in AustereMH to be 0.01 and $\delta$ in MHminibatch to be 5, following the setting in their original papers \cite{korattikara2014austerity, seita2016efficient}. We set batch size $m$ in both methods to be 30. We find that AustereMH and MHminibatch are both inexact on this example and the error increases as we increase $K$. Thus we empirically verify the statement in Theorem \ref{statement:counterexample}. 
Besides the density estimate comparison on $K=200$ shown in Figure \ref{fig:counter-example}b, we additionally report the estimate results on other values of $K$ in Figure \ref{app:fig:density}. We see that the results are similar, all showing that \methodname{} and standard MH can give accurate estimate whereas inexact methods are seriously wrong.\looseness=-1 

\paragraph{Robust Linear Regression} We further tested AustereMH on robust linear regression in Section~\ref{sec:rlr} with $N=5000$. We computed the MSE between estimated and true parameters. MH, TunaMH and AustereMH obtained MSE 0.149, 0.15 and 1.19 respectively, indicating inexact method error can be large on typical problems.

\subsection{Robust Linear Regression}\label{app:experiments:rlr}
We follow the experimental setup of robust linear regression (RLR) in Cornish et al.~\cite{cornish2019scalable}. Specifically, we have data $x_i\in \mathbb{R}^d$ and $y_i\in \mathbb{R}$. The likelihood is modeled by a student's t-distribution with degrees of freedom $v$:
\[
p(y_i|\theta,x_i) = \text{Student}(y_i - \theta^\intercal x_i|v).
\]
It follows that 
\[
U_i(\theta) = \frac{v+1}{2}\log\left(1 + \frac{(y_i-\theta^\intercal x_i)^2}{v}\right),
\]
and the first derivative
\[
\partial_j U_i(\theta) = -(v+1)\frac{x_{ij}(y_i-\theta^\intercal x_i)}{v + (y_i-\theta^\intercal x_i)^2}.
\]
Since the function $U_i$ is Lipschitz continuous, we can easily get the bound used in \methodname{}, TFMH and SMH. We set $M(\theta,\theta') = \norm{\theta - \theta'}_2$ and then it follows
\[
c_i = \sup_{\theta\in\mathbb{R}} \norm{\nabla U_i(\theta)}_2 = \frac{v+1}{2\sqrt{v}}\norm{x_i}_2.
\]
The data $x_i$ and $y_i$ is generated as follows
\[
y_i = \sum_j x_{ij} + \epsilon_i, \hspace{.25cm} \text{where } \epsilon_i\sim\mathcal{N}(0,1).
\]

In Section \ref{sec:rlr}, we set $v=4$, $d=100$ and use a flat prior $p(\theta) = 1$. Note that our problem dimension $d$ is much larger than that in the SMH paper \cite{cornish2019scalable} ($d = 10$). This makes the control variates in SMH problematic since the bounds they require appear to scale badly in high dimensions. 

To reach the target acceptance rate, we set the stepsize in each method as in Table \ref{tab:stepsize} and \ref{tab:stepsize2}. For \methodname{} and \methodname{}-MAP, we set $\chi = 1e-5$ for $N=5000, 20000$ and $\chi=1e-4$ for $N=50000, 100000$. For FlyMC and FlyMC-MAP, we set the probability for a data going from dark to bright $q_{d\rightarrow b}$ to be 0.01. Without the MAP, we collect 80000 samples after 200000 step burnin. With the MAP, we collect 80000 samples without burnin.

\begin{table}[t]
  \caption{Stepsize of methods without the MAP.}
  \label{tab:stepsize}
  \centering
  \begin{tabular}{lcccccc}
    \toprule                 \\
       & MH & TFMH & FlyMC & \methodname{}\\
    \midrule
    \vspace{.1cm}
    RLR $N=5000$ & 4e-3 &1e-4 &2.7e-3 &8e-4, $\chi = 1e-5$ \\
    \vspace{.1cm}
    RLR $N=20000$ & 2e-3 &3e-5 &1.5e-3 &3e-4, $\chi = 1e-5$  \\
    \vspace{.1cm}
    RLR $N=50000$ & 1.3e-3 &1.2e-5 &9e-4 &2e-4, $\chi = 1e-4$  \\
    \vspace{.1cm}
    RLR $N=100000$ & 9e-4 &6e-6 &7e-4 &1.7e-4, $\chi = 1e-4$ \\
    TGM & 3e-1  &2.2e-2  &1e-2   &1e-1\\
    LR &5e-3 &1e-4  &2e-3   &1e-3 \\
    \bottomrule
  \end{tabular}
\end{table}
\begin{table}[H]
  \caption{Stepsize of methods with the MAP.}
  \label{tab:stepsize2}
  \centering
  \begin{tabular}{lccccccc}
    \toprule                 \\
       & MH-MAP & SMH-1 &SMH-2 & FlyMC-MAP & \methodname{}-MAP \\
    \midrule
    \vspace{.1cm}
    RLR $N=5000$ & 4e-3 & 4e-3  &4e-3   &  6e-3 &8e-4, $\chi = 1e-5$ \\
    \vspace{.1cm}
    RLR $N=20000$ & 2e-3 & 2e-3  & 2e-3  & 3.5e-3  &3e-4, $\chi = 1e-5$ \\
    \vspace{.1cm}
    RLR $N=50000$ & 1.2e-3 & 1.2e-3  & 1.2e-3  & 2.5e-3  &1.2e-4, $\chi = 1e-4$ \\
    \vspace{.1cm}
    RLR $N=100000$ &  9e-4 & 5.9e-4  & 8e-4  & 1.7e-3  &7e-5. $\chi = 1e-4$ \\
    TGM & -  &1e-1   &  -    &1e-2  & -\\
    \bottomrule
  \end{tabular}
\end{table}

\subsubsection{Additional Experimental Results with $d=10$}

We ran RLR experiment with $d=10$ and $N=10^5$ to compare the performance in low dimensions. The ESS/S for TFMH, FlyMC, TunaMH are 0.02, 0.75, \& 1.7, respectively; SMH-1, SMH-2, FlyMC-MAP and TunaMH-MAP are 174.7, 5969.5, 730.8, \& 730.1 respectively. This suggests TunaMH is significantly better without MAP/control variates. With MAP/control variates, TunaMH is better than SMH-1, similar to FlyMC and worse than SMH-2.

\subsection{Truncated Gaussian Mixture} \label{app:experiments:mog}
The data in this truncated Gaussian mixture (TGM) task is generated as follows 
\[
x_i \sim \frac{1}{2}\mathcal{N}(\theta_1, \sigma_x^2) + \frac{1}{2}\mathcal{N}(\theta_1+\theta_2, \sigma_x^2)
\]
where $\theta_1 = 0, \theta_2 = 1$ and $\sigma^2 = 2$. The posterior $\theta$ has two modes at $(\theta_1, \theta_2)=(0, 1)$ and $(\theta_1, \theta_2)=(1, -1)$. In order to get the bounds required by all methods, we truncate the Gaussian by setting $\theta_1, \theta_2 \in [-3,3]$. 

For simplicity we assume a flat prior $p(\theta)=1$. Then the energy is given by
\[
U_i(\theta) = -\log p(x_i|\theta) 
= \log(2\sqrt{2\pi}\sigma_x) - \log\bigg[\exp\bigg(-\frac{(x_i-\theta_1)^2}{2\sigma_x^2}\bigg) + \exp\bigg(-\frac{(x_i-\theta_1-\theta_2)^2}{2\sigma_x^2}\bigg)\bigg]. 
\]

Denote $E_1 = \exp\bigg(-\frac{(x_i-\theta_1)^2}{2\sigma_x^2}\bigg)$ and $E_2 = \exp\bigg(-\frac{(x_i-\theta_1-\theta_2)^2}{2\sigma_x^2}\bigg)$. 
To get the upper bound in \methodname{}, TFMH and SMH, we compute the gradient
\begin{align*}
 \frac{\partial U_i(\theta)}{\partial \theta_1} &= -\frac{1}{E_1 + E_2} \bigg(E_1\cdot \frac{x_i - \theta_1}{\sigma_x^2} + E_2 \cdot \frac{x_i-\theta_1-\theta_2}{\sigma_x^2}\bigg),\\ 
 \frac{\partial U_i(\theta)}{\partial \theta_2} &= -\frac{1}{E_1 + E_2} \bigg(E_2 \cdot \frac{x_i-\theta_1-\theta_2}{\sigma_x^2}\bigg).
\end{align*}
Since $\theta_i \in [-3, 3]$, it follows that
\begin{align*}
    \Abs{\frac{\partial U_i(\theta)}{\partial \theta_1}} &\le \frac{\Abs{x_i}+3}{\sigma_x^2} + \frac{\Abs{x_i}+3+3}{\sigma_x^2} \le \frac{2\Abs{x_i}+9}{\sigma_x^2},\\
    \Abs{\frac{\partial U_i(\theta)}{\partial \theta_2}} &\le \frac{\Abs{x_i}+3+3}{\sigma_x^2} \le \frac{\Abs{x_i}+6}{\sigma_x^2}.
\end{align*}
Therefore we can set $M(\theta,\theta') = \norm{\theta-\theta'}_2$ and 
\[
c_i = \sqrt{\bigg(\frac{2\Abs{x_i}+9}{\sigma_x^2}\bigg)^2 + \bigg(\frac{\Abs{x_i}+6}{\sigma_x^2}\bigg)^2}.
\]

To use the control variate in SMH, we need to compute the second derivatives
\begin{align*}
    \frac{\partial^2 U_i(\theta)}{\partial^2 \theta_1} &= \frac{1}{(E_1 + E_2)^2}\cdot \bigg(E_1\cdot \frac{x_i - \theta_1}{\sigma_x^2} + E_2 \cdot \frac{x_i-\theta_1-\theta_2}{\sigma_x^2}\bigg)^2 
    \\\hspace{2em}& - \bigg[E_1\cdot \bigg(\bigg(\frac{x_i - \theta_1}{\sigma_x^2}\bigg)^2 - \frac{1}{\sigma_x^2}\bigg) + E_2\cdot \bigg(\bigg(\frac{x_i - \theta_1 - \theta_2}{\sigma_x^2}\bigg)^2 - \frac{1}{\sigma_x^2}\bigg)\bigg] \cdot \frac{1}{E_1 + E_2}\\
    \frac{\partial^2 U_i(\theta)}{\partial \theta_1\partial \theta_2} &= \frac{1}{(E_1 + E_2)^2} \cdot \bigg(E_2\cdot \bigg(\frac{x_i - \theta_1 - \theta_2}{\sigma_x^2}\bigg)\bigg) \cdot \bigg(E_1\cdot \frac{x_i - \theta_1}{\sigma_x^2} + E_2 \cdot \frac{x_i-\theta_1-\theta_2}{\sigma_x^2}\bigg)
    \\\hspace{2em}& - \bigg[E_2 \bigg(\bigg(\frac{x_i - \theta_1 - \theta_2}{\sigma_x^2}\bigg)^2 - \frac{1}{\sigma_x^2}\bigg)\bigg]\cdot \frac{1}{E_1 + E_2}\\
    \frac{\partial^2 U_i(\theta)}{\partial^2 \theta_2} &= \frac{1}{(E_1 + E_2)^2}\cdot \bigg(E_1\cdot \frac{x_i - \theta_1}{\sigma_x^2} + E_2 \cdot \frac{x_i-\theta_1-\theta_2}{\sigma_x^2}\bigg)^2 
    \\\hspace{2em}& - \bigg[E_2\cdot \bigg(\bigg(\frac{x_i - \theta_1 - \theta_2}{\sigma_x^2}\bigg)^2 - \frac{1}{\sigma_x^2}\bigg)\bigg] \cdot \frac{1}{E_1 + E_2}.
\end{align*}
Given the parameter space, we have the upper bounds
\begin{align*}
    \Abs{\frac{\partial^2 U_i(\theta)}{\partial^2 \theta_1}} &\le  \bigg(\frac{2\Abs{x_i}+9}{\sigma_x^2}\bigg)^2 + \bigg(\frac{\Abs{x_i}+3}{\sigma_x^2}\bigg)^2 + \bigg(\frac{\Abs{x_i}+6}{\sigma_x^2}\bigg)^2 + \frac{2}{\sigma_x^2} \\
    \Abs{\frac{\partial^2 U_i(\theta)}{\partial \theta_1 \partial \theta_2}} 
    &\le \frac{2\Abs{x_i}+9}{\sigma_x^2} \cdot \frac{\Abs{x_i}+6}{\sigma_x^2} + \bigg(\frac{\Abs{x_i}+6}{\sigma_x^2}\bigg)^2 + \frac{1}{\sigma_x^2}\\
    \Abs{\frac{\partial^2 U_i(\theta)}{\partial^2 \theta_2}} &\le \bigg(\frac{2\Abs{x_i}+9}{\sigma_x^2}\bigg)^2 +  \bigg(\frac{\Abs{x_i}+6}{\sigma_x^2}\bigg)^2 + \frac{1}{\sigma_x^2}.
\end{align*}
It follows
\[
\Bar{U}_{2,i} = \bigg(\frac{2\Abs{x_i}+9}{\sigma_x^2}\bigg)^2 + \bigg(\frac{\Abs{x_i}+3}{\sigma_x^2}\bigg)^2 + \bigg(\frac{\Abs{x_i}+6}{\sigma_x^2}\bigg)^2 + \frac{2}{\sigma_x^2}. 
\]
which is required in SMH-1. 

To get the lower bounds in FlyMC, we use the first-order Taylor expansion for $U_i(\theta)$. Higher order approximation is possible but would require heavier computation. By Taylor expansion,
\vspace{-.2cm}
\begin{align*}
    U_i(\theta) = U_i(\theta^0) + \nabla U_i(\theta^0)^\intercal (\theta - \theta^0) + \frac{1}{2}(\theta - \theta^0)^\intercal \nabla^2 U_i(c) (\theta - \theta^0)
\end{align*}
where $c$ is between $\theta$ and $\theta^0$.

Then we can define $\log B_i(\theta)$ in FlyMC as the follows
\begin{align*}
    \log B_i(\theta) &=  -U_i(\theta^0) - \nabla U_i(\theta^0)^\intercal (\theta - \theta^0) - \frac{1}{2}\cdot\max_c\norm{\nabla^2 U_i(c)}_1 \cdot\norm{\theta - \theta^0}^2_1\\
    & = -U_i(\theta^0) - \nabla U_i(\theta^0)^\intercal (\theta - \theta^0) - \frac{1}{2}\cdot\Bar{U}_{2,i} \cdot\norm{\theta - \theta^0}^2_1.
\end{align*}
The sum of $\log B_i$ is
\vspace{-.2cm}
\begin{align*}
    \sum_{i=1}^N \log B_i(\theta) =  -N\cdot U_i(\theta^0) - \bigg(\sum_{i=1}^N\nabla U_i(\theta^0)\bigg)^\intercal (\theta - \theta^0) - \frac{1}{2}\cdot\sum_{i=1}^N \Bar{U}_{2,i} \cdot\norm{\theta - \theta^0}_1^2.
\end{align*}
We set $\theta^0$ to be 0 and the MAP solution in standard and MAP-tuned FlyMC respectively. 

\begin{figure*}[t!]
    \centering
    \begin{tabular}{cccc}		
    	\includegraphics[width=3.25cm]{figure/31-tunamh/mog_true.pdf}  &
    		\hspace{-6mm}
    	\includegraphics[width=3.25cm]{figure/31-tunamh/mog_pmh.pdf} &
    	\hspace{-6mm}
    	\includegraphics[width=3.25cm]{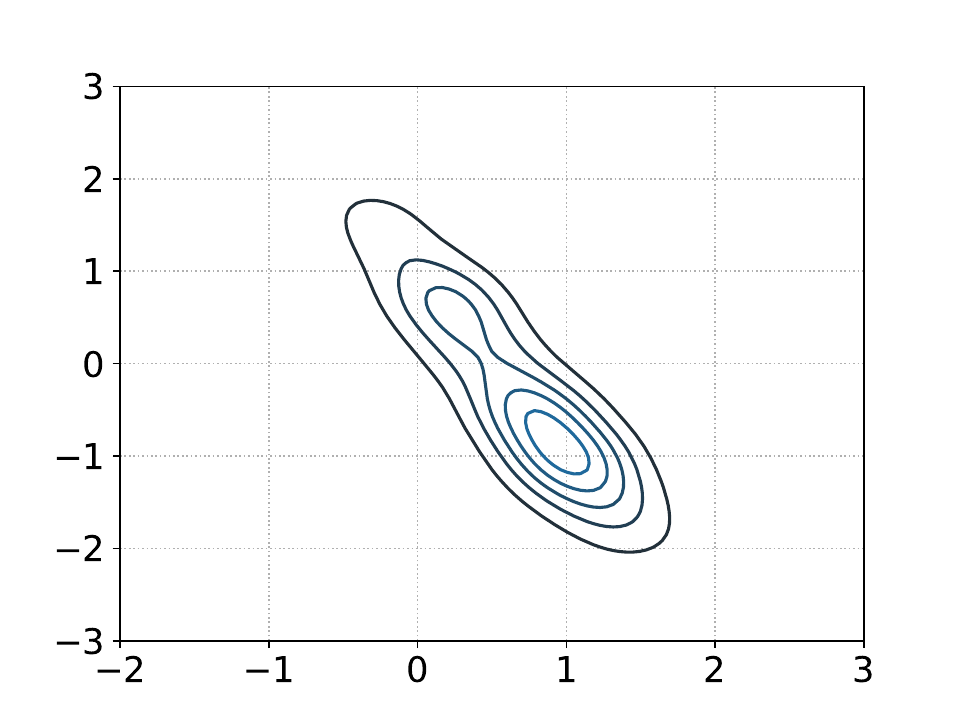} &
    	\hspace{-6mm}
    	\includegraphics[width=3.25cm]{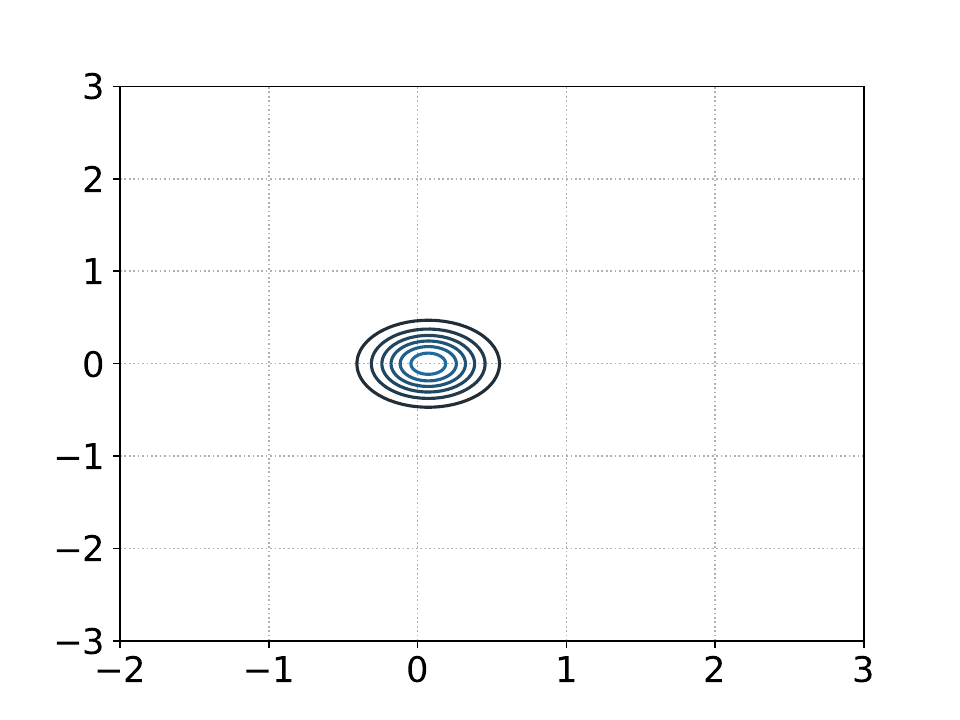}
    	\\		
    	(a) True&
    	(b) \methodname{}&
    	(c) TFMH&
    	(d) FlyMC
    	\\
    	\includegraphics[width=3.25cm]{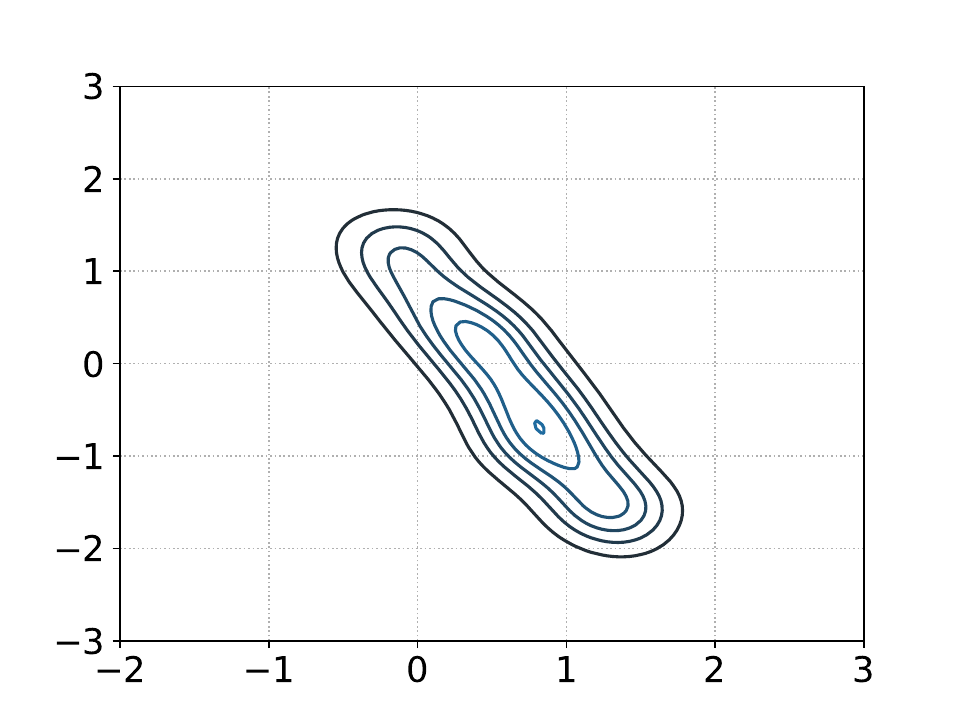}
    	&
    	\hspace{-6mm}
    	\includegraphics[width=3.25cm]{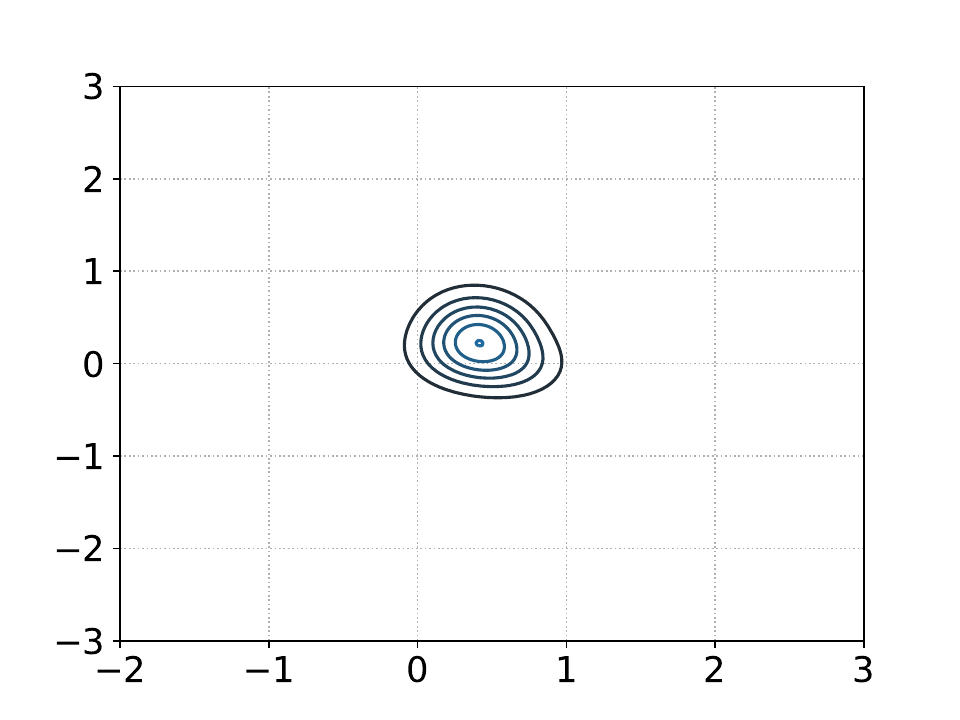}  &
    	\hspace{-6mm}
    	\includegraphics[width=3.25cm]{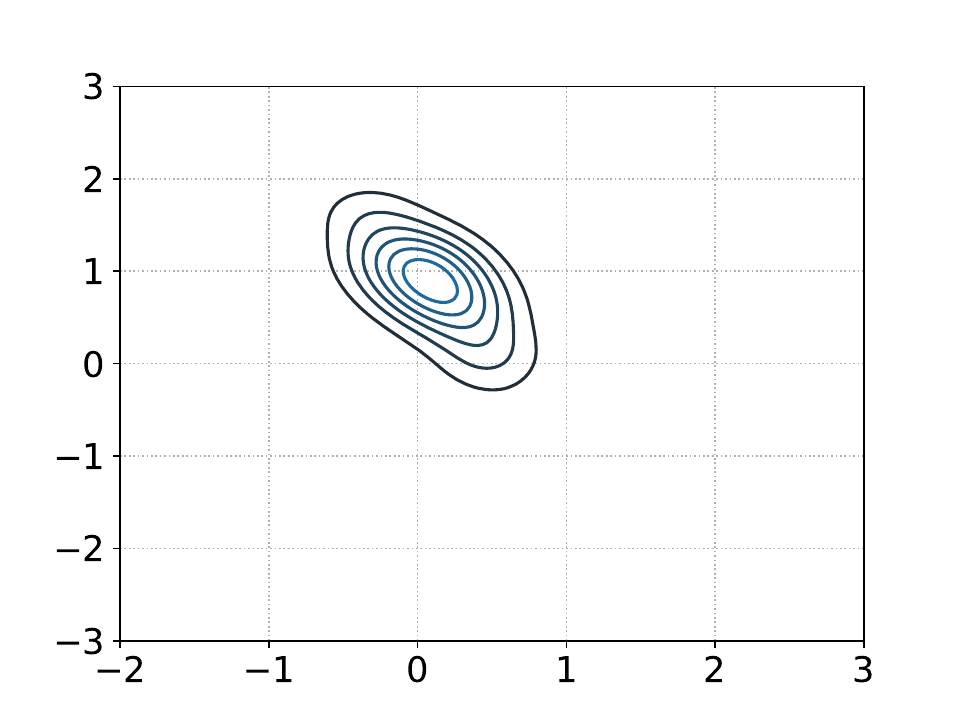} &
    	\hspace{-6mm}
    	\includegraphics[width=3.25cm]{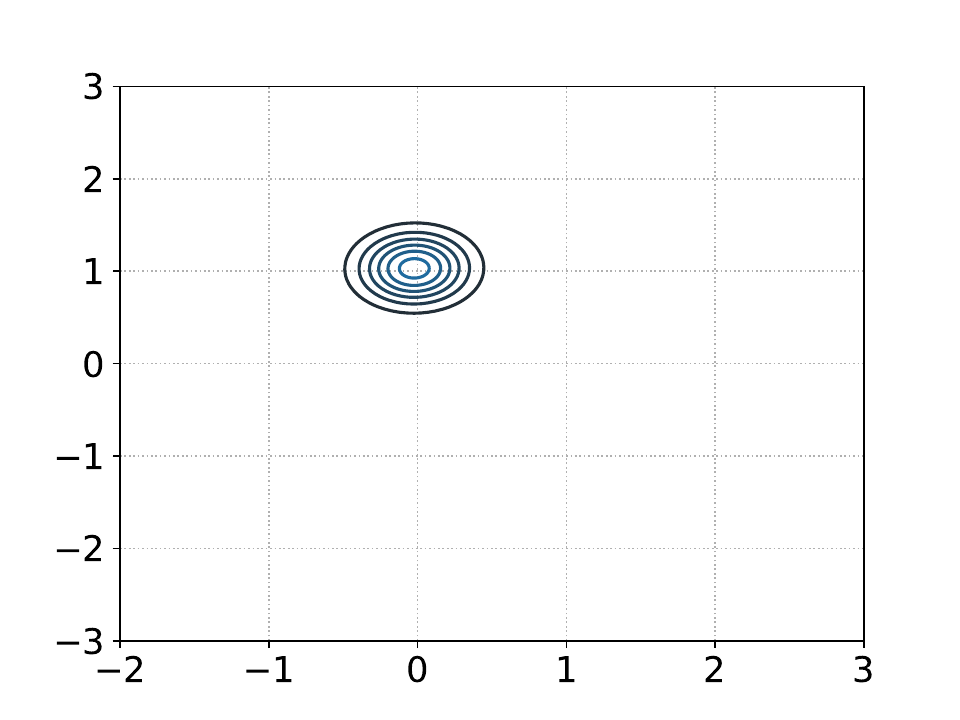}
    	
    	\\	
    	(e) PoissonMH&
    	(f) MH&
    	(g) SMH-1&
    	(h) FlyMC-MAP
    	\hspace{-0mm}\\		
    \end{tabular}
    \caption{Visualization of the density estimate after 1 second.}
    \label{app:fig:mog}
\end{figure*}

We tune the stepsize of each method to reach the acceptance rate $60\%$ and the value of stepsize is summarized in Table \ref{tab:stepsize} and \ref{tab:stepsize2}. We set $\chi = 10^{-4}$ in \methodname{} and $q_{d\rightarrow b} = 0.01$ in FlyMC and FlyMC-MAP. We compute the symmetric KL between the run-average density estimate and the true distribution. Since this is a two-dimensional problem, we are able to visualize the density estimate. As shown in Figure \ref{app:fig:mog}, we plot the density estimate after running the method for 1 second. It is clear to see that the density estimate of \methodname{} is close to the truth whereas all other methods are unable to provide accurate density estimate given the time budget.

\subsection{Logistic Regression on MNIST}

MNIST with only 7s and 9s images contains 12214 training data and 2037 test data. Let $h$ be the sigmoid function. Let the label $y_i\in \{0, 1\}$, then the model in logistic regression (LR) is 
\[
p(y_i = 1) = h(\theta^\intercal x_i) = \frac{1}{1 + \exp\left(-\theta^\intercal x_i\right)}.
\]
It follows that 
\[
U_i(\theta) = -y_i\log h\left(\theta^\intercal x_i\right) - (1-y_i) \log h\left(-\theta^\intercal x_i\right).
\]

It is easy to see that
\[
\Abs{\partial_j U_i} = \Abs{(h(\theta^\intercal x_i)-y_i)x_{ij}} \le 1\cdot \Abs{x_{ij}}.
\] 
Thus we can set $M(\theta,\theta')$ to be $\norm{\theta-\theta'}_2$ and $c_i$ to be $\norm{x_{i}}_2$. We use this bound for \methodname{}, TFMH and SMH. For FlyMC, we use the same bound on logistic regression as in the FlyMC paper \cite{maclaurin2015firefly}.

We set the target acceptance rate to be $60\%$ and the resulted stepsize is reported in Table \ref{tab:stepsize}. We set $q_{d\rightarrow b}$ to be 0.1 following Maclaurin et al.~\cite{maclaurin2015firefly}.

%% file: section/99-appendix/32-cdgrab/00-app-cdgrab.tex
\chapter{Appendix for \dgrab}\label{chapter:app:cdgrab}
\input{section/99-appendix/32-cdgrab/100-app-cdgrab-glossary}
\input{section/99-appendix/32-cdgrab/200-app-cdgrab-dgrab}
\input{section/99-appendix/32-cdgrab/300-app-cdgrab-proof}
\input{section/99-appendix/32-cdgrab/400-app-cdgrab-experiments}

%% file: section/99-appendix/32-cdgrab/100-app-cdgrab-glossary.tex
\vspace{-1.5cm}
\begingroup
\setlength{\tabcolsep}{4pt} 
\renewcommand{\arraystretch}{1} 
\begin{table}[H]
\scriptsize
\begin{center}
      \centering
        \begin{tabular}{p{0.08\linewidth}p{0\linewidth}p{.95\linewidth}}
\toprule
\textbf{Term} && \textbf{Explanation} \\
\midrule
\cgrab & & Centralized online Gradient Balancing algorithm (Original  centralized algorithm developed in \grab{}~\cite{lu2022grab}. \\\midrule
\dgrab & & Coordinated and distributed online Gradient Balancing algorithm. The algorithm that is our main contribution. \\\midrule
ID-\grab & & Independent, distributed online gradient balancing. Implemented for our ablation study. One version uses the original \grab's online $\mathsf{Balance}$ (ID-\grab{} (Bal)), and one implements our online $\mathsf{PairBalance}$ (ID-\grab{} (PairBal)).\\\midrule
\shuffle & & Random reshuffling algorithm. We use this to refer to its centralized variant.\\\midrule
\dshuffle & & Distributed random reshuffling algorithm.\\\midrule
\so & & Shuffle Once algorithm.\\\midrule
$\dataex$ && Data-example vector; we do not use this in the math in the main paper, but do refer to examples in our schematic description for $\mathsf{PairBalance}$ ordering in Figure~\ref{fig:cdgrab:diagram}.\\\midrule
$\ex$ & &  Vector (for illustration under the herding context). For \cgrab{} and \dgrab, these are gradients. \\\midrule
$\barex$ & & The average vector (for illustration under the herding context).\\\midrule
$\exj$ & & The $\exindex$-th component of a vector (for illustration under the herding context).\\\midrule
$\exij$ & & The $\exindex$-th component of the gradient on worker $\windex$ (for illustration under our parallel herding framework).\\\midrule
$\weights$ & & Parameters / model-weights vector.\\\midrule
$\loss$ & & Loss function. \\\midrule
$\nabla\loss(\weights)$ & & Global loss gradient. \\\midrule
$\nabla\loss^\windex(\weights)$ & & Local $\windex$-th worker's loss gradient. \\\midrule
$\nabla\loss^\windex(\weights; \exindex)$ & & Local $\windex$-th worker's, $\exindex$-th example's loss gradient. \\\midrule
$\perm$ & & A permutation; we study permutation-based example orderings. \\\midrule
$\epochs$ & & Number of epochs.\\\midrule
$\eindex$ & & Index for iterating over $\epochs$ epochs.\\\midrule
$\workers$ & & Number of workers (in this paper, workers are processes, potentially on different GPUs but on the same node). $\workers=1$ in the centralized setting.\\\midrule
$\windex$ & & Index for iterating over $\workers$ workers .\\\midrule
$\workerexamples$ & & Number of training-data examples per worker; equivalent to $\frac{\examples}{\workers}$.\\\midrule
$\examples$ & & Number of total training-data examples. $\examples=\workerexamples$ in the centralized setting.\\\midrule
$\exindex$ & & Index for iterating over examples.\\\midrule
$\g$ & & Gradient, taken with respect to the model weights $\weights$ and data examples $\dataex$.\\\midrule
$\wexgrad$ & & Gradient associated with the $\exindex$-th data example $\dataex$ on worker $\windex$.\\\midrule
$\sgn$ & & A sign, either $+1$ or $-1$; related to the signed herding problem.\\\midrule
$\sgn_\exindex^\windex$ & & A sign, either $+1$ or $-1$, computed according to the $\exindex$-th example gradient $\wexgrad$ for worker $\windex$; to be associated with the example $\dataex_\exindex$ when determining a permutation ordering using Algorithm~\ref{alg:reorder}.\\
\bottomrule
\end{tabular}
\end{center}
\end{table}

%% file: section/99-appendix/32-cdgrab/200-app-cdgrab-dgrab.tex
\section{Additional Details on the \dgrab{} Algorithm and online $\mathsf{PairBalance}$}\label{app:sec:cdgrab:details}

In this Appendix, we provide more details on related work and our contributions. 
To start, we give a unified description of our online \dgrab{} algorithm with prior work on herding, vector balancing, and kernel thinning (Appendix~\ref{app:sec:cdgrab:details:contribution}), some more details on Alweiss et al.~\cite{alweiss2021discrepancy} that we elide in the main paper due to space constraints (Appendix~\ref{app:sec:cdgrab:alweiss}), conceptual details on implementing \dgrab{} with a parameter server (Appendix~\ref{app:sec:cdgrab:ps}), and implementing our improved balancing algorithm (online $\mathsf{PairBalance}$) in a centralized fashion to get additional improvements for \grab{} (Appendix~\ref{app:sec:cdgrab:central}). 

\subsection{Distinguishing our contributions}\label{app:sec:cdgrab:details:contribution}

We summarize our contributions in relation to prior work in a concise format. 
This kind of presentation would not be easily understandable without the appropriate background and context that we provide in the paper. 
This is why present it here, in the Appendix, so that (ideally) this is seen by the reader after finishing the main paper.

We emphasize that it is prior work that:

\begin{itemize}
    \item Formulates the herding objective and solves it with vector balancing~\citep{harvey2014near, welling2009herding} (Algorithm~\ref{alg:reorder}).
    \item Leverages ideas from herding and vector balancing (above) in an optimization setting to do permutation-based example ordering~\citep{lu2022grab}.
    \item Observes and proves that it is possible to solve the herding objective in $\tilde{O}(1)$ by only examining differences on pairs of examples (the overarching idea of $\mathsf{PairBalance}$~\citep{dwivedi2021kernel}, which relies on the online $\mathsf{RandomizedBalance}$ subroutine~\citep{alweiss2021discrepancy}; see Algorithm~\ref{alg:pairbalance}). 
\end{itemize}

Our contributions are to bring together all of this prior work in a novel way. We 

\begin{itemize}
    \item Translate the herding and balancing framework to the parallel setting via defining a parallel herding objective (\ref{equ:paraherding:objective}).
    \item Leverage prior work on herding in an optimization setting~\citep{lu2022grab} so that we can do parallel herding in an optimization setting (Section~\ref{sec:cdgrab:dgrab}).
    \item Execute \emph{online} pair balancing on a server (Algorithm~\ref{alg:pairbalance} on a running sum, Figure~\ref{fig:cdgrab:diagram}), i.e., do pair balancing in a streaming and asynchronous (rather than blocking) fashion from gradient vectors produced on distributed workers (Algorithm~\ref{alg:dgrab}), on the flattened sequenced of paired-difference gradients (Section~\ref{sec:cdgrab:dgrab:solution}); this leads to an improvement over \grab, which relies on a stale mean.
\end{itemize}

\subsection{More details on $\mathsf{RandomizedBalance}$ from Alweiss et al.~\cite{alweiss2021discrepancy}} \label{app:sec:cdgrab:alweiss}

In the subroutine for $\mathsf{RandomizedBalance}$ in Algorithm~\ref{alg:pairbalance}, we elide details about how the probability $p$ is computed exactly as in Alweiss et al.\cite{alweiss2021discrepancy}. 
We provide a more complete specification in Algorithm~\ref{app:alg:alweiss} written in terms of a single input vector (which, for us, is the vector containing the difference between adjacent gradients). 
Note that the difference here is in the use of a required parameter, constant upper bound $w$, which is used to compute the probability $p$. 
For clarity of presentation in the subroutine in Algorithm~\ref{alg:pairbalance}, we have set $w=1$.
Alweiss et al.~\cite{alweiss2021discrepancy} sets this threshold differently, which we still elide for simplicity. 

\begin{algorithm}[h]
\caption{Probabilistic Balancing with Logarithm Bound [Alweiss et al.~\cite{alweiss2021discrepancy}]}\label{app:alg:alweiss}
    \begin{algorithmic}[1]
    \Statex \textbf{require:} parameter $w$, used to compute probability
    \Statex \textbf{input:} current running sum $\vr$ vector, vector $\vz_{\text{diff}}$
    \If{$|\langle \vr, \vz_{\text{diff}} \rangle |>w$ or $\norm{\vr}_\infty>w$}
           \State \textbf{Fail}
    \EndIf
    \State \textbf{compute:} $p\leftarrow \frac{1}{2} - \frac{\langle \vr,\vz_{\text{diff}}\rangle}{2w}$
    \State \textbf{compute:} $\sgn \leftarrow +1$ \hspace{.2em} with probability \hspace{.2em} $p$; 
    \Statex  \hspace{4.25em}$\sgn\leftarrow -1$ \hspace{.2em} with probability \hspace{.2em} $1-p$
    \State \textbf{update:}  $\vr\leftarrow \vr+\sgn\vz_{\text{diff}}$
    \State \textbf{return:}  $\sgn$, $\vr$
    \end{algorithmic}
\end{algorithm}

In practice, we actually do not use $\mathsf{RandomizedBalance}$ in our online $\mathsf{PairBalance}$. 
We use the deterministic, greedy-ordering algorithm from the original Lu et al.~\cite[Algorithm 5]{lu2022grab} paper: 

\begin{algorithm}[h]
\caption{Balancing without normalization [Lu et al.~\cite{lu2022grab}]}\label{app:alg:greedy}
    \begin{algorithmic}[1]
    \Statex \textbf{input:} current running sum $\vr$ vector, vector $\vz_{\text{diff}}$
    \State \textbf{if } $\norm{\vr + \vz_{\text{diff}}} < \norm{\vr - \vz_{\text{diff}}}$ \textbf{ then } $\sgn \leftarrow +1$ \textbf{ else } $\sgn \leftarrow -1$
    \State \textbf{update:}  $\vr\leftarrow \vr+\sgn\vz_{\text{diff}}$
    \State \textbf{return:}  $\sgn$, $\vr$
    \end{algorithmic}
\end{algorithm}

Note that, unlike Alweiss et al.~\cite{alweiss2021discrepancy} (Algorithm~\ref{app:alg:alweiss}), Algorithm~\ref{app:alg:greedy} from Lu et al.~\cite{lu2022grab} cannot end up in a failure state. 

In Alweiss et al.~\cite{alweiss2021discrepancy}, Theorem 1.1 proves the $\tilde{O}(1)$ probabilistic bound for Algorithm~\ref{app:alg:alweiss} (See Theorem~\ref{statement:alweiss}) for a restatement of this result in terms of our work). 
Corollary 7 of Dwivedi and Mackey~\cite{dwivedi2021kernel} re-proves this result (which they mislabel as Alweiss et al.\cite{alweiss2021discrepancy}, Theorem 1.2, see Dwivedi and Mackey~\cite[Appendix R, p. 69]{dwivedi2021kernel}). 
They improve the constants and have a less conservative setting of the thresholds $w$. 
The proof is also very short and elegant, by relying on their Theorem 3. 

\subsection{Implementing \dgrab{} with a parameter server}\label{app:sec:cdgrab:ps}

For our implementation of \dgrab{}, we use a parameter server architecture~\citep{li2014ps}. 
For our purposes, this just entails computing the average gradient (used to update the model on all workers) on the server side. 
That is, the server (other than determining the ordering for the next epoch) also has the function of aggregating gradient information (in this case, a simple mean) to send back to the workers. 

We have the server compute the average $\exindex$-th gradient for illustrative purposes. 
We could, instead, implement the computation the average gradient as an all-reduce operation, in which each worker broadcasts their gradients to all other workers, so that they can each locally compute the average gradient to update their local models. 
We implement \dgrab{} using a parameter server pattern to show that this is a plausible architecture to use with our coordinated and distributed example ordering algorithm. We could also implement a full parameter server system, for which the server also coordinates global model updates.

If we kept everything in our implementation the same and switched to all-reduce, then we would no longer be following a parameter server paradigm. 
In this case, the server would just function to determine example orders. 
It is this kind of paradigm that suggests the abstraction of an \emph{order server}, which we mention briefly in Section~\ref{sec:cdgrab:conclusion}: 
A server whose sole responsibility is coordinating worker information to determine example ordering.

In future work, we intend to explore a host of architectural possibilities --- of building a full system that incorporates both traditional parameter server aspects with our new abstraction of an order server. 
For example, we could have parameter servers and order servers work in tandem in a distributed system to perform model training. 
To move beyond the single-node implementation we present in this paper, we intend to investigate the benefits and trade-offs associated with such design decisions in an actual implemented system.

\subsection{Centralized online $\mathsf{PairBalance}$} \label{app:sec:cdgrab:central}

\begin{figure}[t!]
\centering
\includegraphics[width=.7\linewidth]{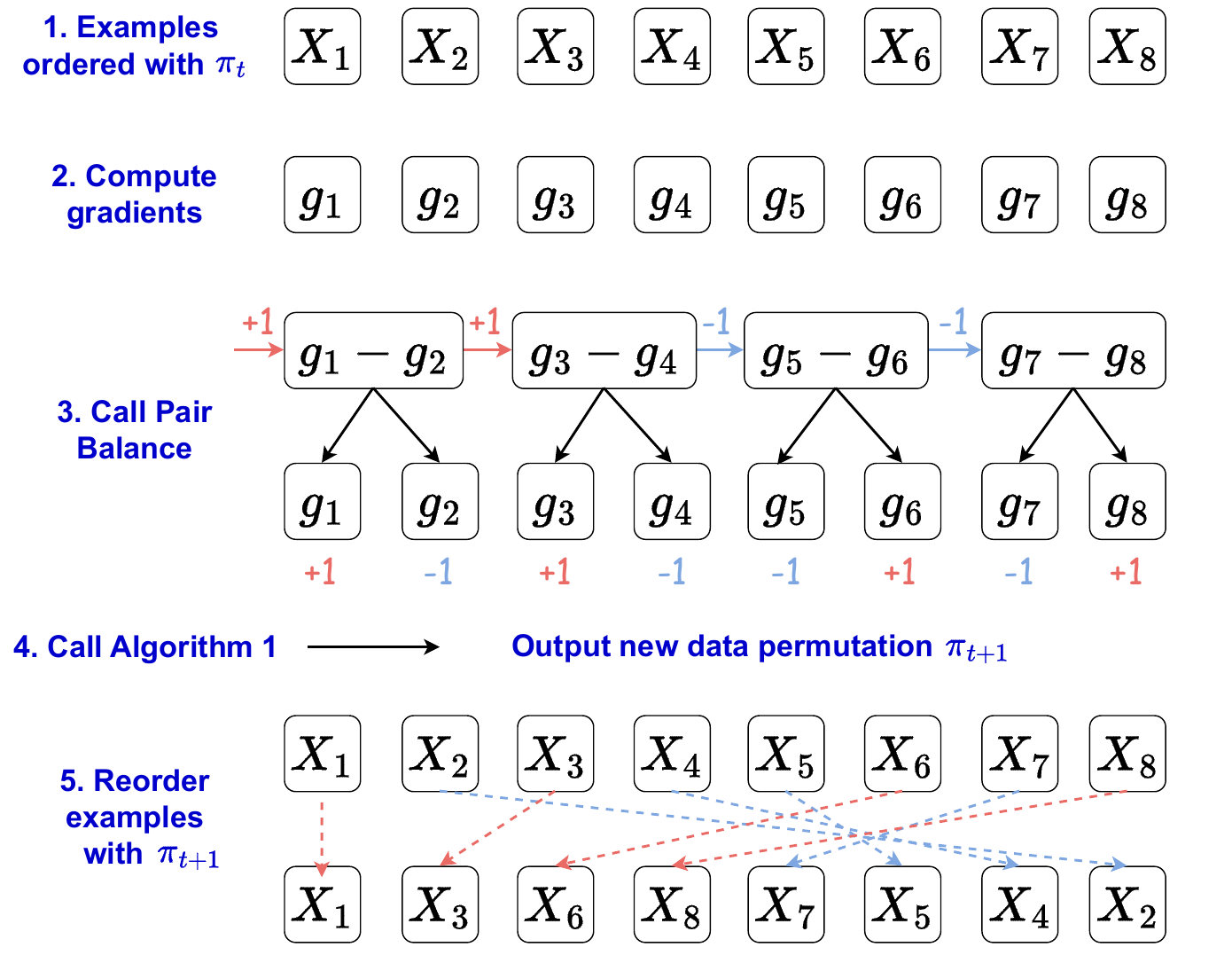}
\caption{Schematic representation of online  $\mathsf{PairBalance}$ for centralized \grab.}
\label{app:fig:centralizedpairbal}
\end{figure}

In Section~\ref{sec:cdgrab:dgrab:algo}, we provide a schematic diagram of how online  $\mathsf{PairBalance}$ works for a distributed implementation using a parameter server (Figure~\ref{fig:cdgrab:diagram}). 
We also claim in Section~\ref{sec:cdgrab:dgrab}~that online $\mathsf{PairBalance}$ can be applied to the original centralized \grab{} algorithm for improved empirical performance. 
We provide a schematic here, in Figure~\ref{app:fig:centralizedpairbal} (analogous to Figure~\ref{fig:cdgrab:diagram}), for online $\mathsf{PairBalance}$ for centralized \grab. 

We also provide empirical results comparing \grab's $\mathsf{Balance}$ routine to the online $\mathsf{PairBalance}$ routine that we instead use in this work.  
We observe that both $\mathsf{PairBalance}$ and $\mathsf{Balance}$ would have similar convergence rates under centralized settings, and both outperform \shuffle. 

This experiment justifies the uses of $\mathsf{PairBalance}$ even in centralized learning settings. 
$\mathsf{PairBalance}$ theoretically tolerates higher learning rates and, as we will justify in Appendix~\ref{sec:appendix-memory}, is more memory-efficient than $\mathsf{Balance}$. 
In short,   $\mathsf{PairBalance}$ an excellent substitute for $\mathsf{Balance}$ when running \grab. 

\begin{figure}[!h]
    \centering
    \includegraphics[width=\linewidth]{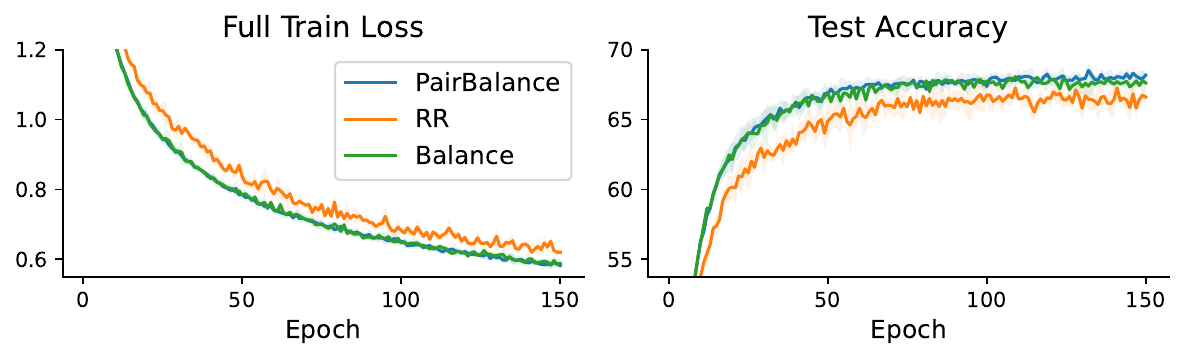}
    \caption{Convergence for centralized online $\mathsf{PairBalance}$ on LeNet on CIFAR-10. We use the identical set of hyperparameters ($\alpha$ = 1e-3, weight decay = 1e-2, momentum = 0.9, $B = 64$) as in the scaling experiments as in Figure~\ref{fig:nodes}.}
    \label{fig:centralized_pair_bal_convergence}
\end{figure}

%% file: section/99-appendix/32-cdgrab/300-app-cdgrab-proof.tex
\section{Proof Results}

We present supporting results, which we use to prove the main results presented in Section~\ref{sec:cdgrab:theory}. 
First, we show how to analyze the parallel herding bound in terms of a single step over the server-side $\mathsf{PairBalance}$ algorithm (Appendix~\ref{app:sec:cdgrab:proof:herding}). 
We then include some additional observations/notation (Appendix~\ref{app:proof:note}), which we use in the remaining intermediate results. 
We prove some intermediate results about how much the loss can change over the course of one epoch, assuming smoothness (Appendix~\ref{app:proof:smooth}) and bounded gradient variance and heterogeneity (Appendix~\ref{app:proof:other}). 
We combine these results to get one more intermediate result about the maximum the loss  can change on average over many epochs (Appendix~\ref{app:proof:last}), which we then use altogether to prove the two theorems that we present in the main paper (Appendix~\ref{app:thm:proof}). 

\subsection{Analyzing the parallel herding bound}\label{app:sec:cdgrab:proof:herding}

In the main paper, we cover how \dgrab{} runs on both the worker- and server-side. 
In this section, we dive deeper into the example-ordering part of \dgrab{}, and demonstrate in theory how server-side online $\mathsf{PairBalance}$ reduces the parallel herding bound (\ref{equ:paraherding:objective}), as formulated in Section~\ref{sec:cdgrab:dgrab}. 
We conclude this section by presenting Lemma~\ref{lem:pair-balance}, which shows server-side $\mathsf{PairBalance}$ is able to iteratively reduce the parallel herding bound.

To begin, we formalize our illustration over a group of vectors (since vector balancing, including $\mathsf{PairBalance}$, does not inherently involve an optimization context until we use it in our online setting on gradients). 
Without loss of generality, we assume that the $\examples$ examples are divided evenly among the $\workers$ workers and that $\workerexamples$ is even.
That is, we consider that we are given a set of vectors $\exij \in \R^d$ for $\windex \in [\workers]$ and $\exindex \in [\workerexamples]$ evenly located on $\workers$ workers (i.e., $\workerexamples=\frac{\examples}{\workers}$), where $\exij$ denotes the $\exindex$-th vector located on the $\windex$-th worker. 
Now denote $\perm_{\windex}$ as the original permutation of the vectors on worker $\windex$. Consider running Algorithm~\ref{alg:dgrab:vector:server} on the server side over these $\examples$ vectors. 

\begin{figure}[t!]
\begin{algorithm}[H]
\caption{Server-side $\mathsf{PairBalance}$ over a set of vectors (one step)}\label{alg:dgrab:vector:server}
\footnotesize
\begin{algorithmic}[1]
    \Statex \textbf{require:} $\workers$ workers, $\workerexamples \coloneqq \frac{\examples}{\workers}$ vectors per worker
    \Statex \textbf{input:} initial permutations for all the workers $\{\perm_{\windex}\}_{\windex=1}^\workers$
    \State \textbf{initialize:} new permutations for all the workers $\{\perm'_{\windex}\}_{\windex=1}^\workers$
    \State  \textbf{initialize:} running partial sum $\vh=\bm{0}$
    \State  \textbf{initialize:} new indices front (left) pointer $\{l_i=1\}_{\windex=1}^\workers$
    \State  \textbf{initialize:} new indices back (right) pointer $\{r_i=1\}_{\windex=1}^\workers$
    \For{ example $j \coloneqq 1 \ldots \workerexamples$}
        \For{worker $\windex \coloneqq 1 \ldots \workers$}
            \If{$\exindex\bmod 2 = 0$} \hspace{.25cm} $\rhd$ If at an even index, i.e., can examine a full pair of examples
                \State $\vh,\; \wexsignprev,\; \wexsign \leftarrow \mathsf{PairBalance}(\vh,\; 
                \vz_{\exindex-1}^{\windex}, \vz_{\exindex}^{\windex})$
                \If{$\wexsignprev=+1$}
                    \State $\perm'_{\windex}(l_\windex)=\exindex-1$; \hspace{.5em} $l_\windex=l_\windex+1$ \hspace{.25cm} $\rhd$ Append first in pair to the front/left
                    \State $\perm'_{\windex}(r_\windex)=\exindex$; \hspace{.5em} $r_\windex=r_\windex-1$ \hspace{.25cm} $\rhd$ Append second in pair to the right/back
                \Else
                    \State $\perm'_{\windex}(l_\windex)=\exindex$; \hspace{.5em} $l_\windex=l_\windex+1$ \hspace{.25cm} $\rhd$ Append second in pair to the left/front
                    \State $\perm'_{\windex}(r_\windex)=\exindex-1$; \hspace{.5em} $r_\windex=r_\windex-1$ \hspace{.25cm} $\rhd$ Append first in pair to the right/back
                \EndIf
            \EndIf
        \EndFor
    \EndFor
    \State \textbf{output:} new permutations for all $\workers$ workers $\{\perm'_{\windex}\}_{\windex=1}^\workers$
\end{algorithmic}
\end{algorithm}
\caption{One-step $\mathsf{PairBalance}$ algorithm on the server side to solve the parallel herding problem (\ref{equ:paraherding:objective}). This algorithm can be seen as a prototype for Algorithms~\ref{alg:dgrab:workers} and~\ref{alg:dgrab:server}, without the optimization context.}
\end{figure}

\newpage
It follows, in the following Lemma~\ref{lem:pair-balance}, that we can get the parallel herding bound with the output permutations $\{\perm'_\windex\}_{\windex=1}^\workers$ from Algorithm~\ref{alg:dgrab:vector:server}:

\begin{lemma}
\label{lem:pair-balance}
Suppose that we have a set of vectors $\exij \in \R^d$ for all $\windex, \windex' \in [\workers]$ and for all $\exindex, \exindex' \in [\workerexamples]$ that satisfies
\begin{align*}
    \norm{\sum_{\windex=1}^\workers \sum_{\exindex=1}^\workerexamples   \exij}_\infty \le c_1 \hspace{2em}\text{and}\hspace{2em}
    \norm{\vz_{\windex',\exindex'} - \frac{1}{\workers\workerexamples}\sum_{\windex=1}^\workers \sum_{\exindex=1}^\workerexamples  \exij}_\infty \le c_2 
\end{align*}

for some constants $c_1>0$ and $c_2>0$.
If we run Algorithm~\ref{alg:dgrab:vector:server} over these vectors, then, for any $\delta>0$, it holds with probability at least $1-\delta$ that 
\begin{align*}
\max_{l \in [\workerexamples]}  \norm{\sum_{\windex=1}^\workers\sum_{\exindex=1}^l \ex_{\windex,\perm'_\windex(\exindex)}}_\infty \le & \;\; \frac{1}{2}\max_{l \in [\workerexamples]}\norm{\sum_{\windex=1}^\workers \sum_{\exindex=1}^{l} \ex_{\windex,\perm_\windex(\exindex)}}_\infty  + c_1 + \tilde{A}c_2,
\end{align*}
where $\tilde{A}$ comes from Theorem~\ref{statement:alweiss}.
\end{lemma}

Lemma~\ref{lem:pair-balance} shows that $\mathsf{PairBalance}$ reduces the parallel herding objective (\ref{equ:paraherding:objective}) towards a constant (invariant to $n$) at each step. 
This implies that, if we repeatedly call $\mathsf{PairBalance}$ on a given permutation, it will return a permutation that guarantees the parallel herding bound to be $\tilde{O}(1)$.

\begin{proof}
We prove this lemma by defining the following auxiliary sequence of pair differences, as in Section~\ref{sec:cdgrab:dgrab:solution}
\[
    \vy_{\workerexamples\cdot(k-1)+\windex} = \ex_{\windex,\perm_\windex(2k-1)} - \ex_{\windex,\perm_\windex(2k)}, \;\; \forall k\in[\workerexamples/2],
\]
which we also can refer to as $\{\vy_\exindex\}_{\exindex=1}^{\workers\workerexamples/2}$.

We also leverage Theorem~\ref{statement:alweiss}, which we reprint below for clarity of presentation:

\setcounter{theorem}{0}

\begin{theorem}[\textbf{Corollary 7, Dwivedi and Mackey~\cite{dwivedi2021kernel}}]
    Consider any vectors $\{\vz_\exindex\}_{\exindex=1}^\examples$ ($\exj \in \R^d$) with $\norm{\vz_\exindex}_2 \le 1$ supplied as input to the $\mathsf{RandomizedBalance}$ subroutine in Algorithm~\ref{alg:pairbalance}. 
    Then for any $\delta > 0$, with probability at least $1 - \delta$, $\mathsf{RandomizedBalance}$ outputs a sequence of signs $\{s_\exindex\}_{\exindex=1}^\examples\in \{-1,1\}$ that satisfy $\textstyle \max_{k\in[\examples]}\norm{\sum\nolimits_{\exindex=1}^k s_\exindex\vz_\exindex}_{\infty} \le \tilde{A}$, where 
    $\tilde{A}=\sqrt{2\log(\frac{4d}{\delta})\log(\frac{4N}{\delta})}=\tilde{O}(1)$.
\end{theorem}

Note that the reordering part of Algorithm~\ref{alg:dgrab:vector:server} (line 8) gives a sequence of signs $\{\sgn_\exindex\}_{\exindex=1}^{\workers\workerexamples/2}$. Therefore, by Theorem~\ref{statement:alweiss}, the sequence $\{\vy_\exindex\}_{\exindex=1}^{\workers\workerexamples/2}$ satisfies
\begin{align}
\label{app:lemma1:a}
    \max_{P \in [\workers\workerexamples/2]} \norm{\sum_{p=1}^P \sgn_p \vy_p}_{\infty} \le 2\tilde{A}c_2,
\end{align}

since (based on what is given in Lemma~\ref{lem:pair-balance})
\begin{align*}
    \norm{\vy_{\workerexamples(k-1)+\windex}}_\infty \le \norm{\vz_{\windex,\perm_{\eindex,\windex}(2k-1)} - \frac{1}{\workers\workerexamples}\sum_{\windex=1}^\workers\sum_{\exindex=1}^\workerexamples \vz_{\windex,\exindex}}_\infty + \norm{\vz_{\windex,\perm_{\eindex,\windex}(2k)} - \frac{1}{\workers\workerexamples}\sum_{\windex=1}^\workers\sum_{\exindex=1}^\workerexamples \vz_{\windex,\exindex}}_\infty \le 2c_2.
\end{align*}

Note that, if $\sgn_{\windex,k}$ is the sign associated with $\vy_{\workerexamples(k-1)+\windex}$, then $\vz_{\windex,\perm_{\eindex,\windex}(2k-1)}$ and $\vz_{\windex,\perm_{\eindex,\windex}(2k)}$ will receive opposite signs $\sgn_{\windex,k}$ and $-\sgn_{\windex,k}$, respectively. 

We denote $\vx^+_{\windex,k}$ to be the example that receives sign $\sgn_{\windex,k} = +1$ and $\vx^-_{\windex,k}$ to be the example that receives sign $\sgn_{\windex,k} = -1$. 

That is, if $\sgn_{\windex,k} = +1$, then $\vx^+_{\windex,k} = \vz_{\windex,\perm_\windex(2k-1)}$, otherwise, if $\sgn_{\windex,k} = -1$, then $\vx^+_{\windex,k} = \vz_{\windex,\perm_\windex(2k)}$; and, $\vx^-_{\windex,k}$ is the other term of the pair $\{\vz_{\windex,\perm_\windex(2k-1)}, \vz_{\windex,\;\perm_\windex(2k)}\}$.

Now, for $K \in [\frac{\workerexamples}{2}]$, let
\begin{align*}
    \kappa_{\windex,K} &= \sum_{k=1}^K (\vz_{\windex,\perm_\windex(2k-1)} + \vz_{\windex,\perm_\windex(2k)}) \hspace{.5em} \text{and}\\
    \upsilon_{i,K} &= \sum_{k=1}^K (\sgn_{\windex,k} \vz_{\windex,\perm_\windex(2k-1)} - \sgn_{i,k} \vz_{\windex,\perm_\windex(2k)}).
\end{align*}
Then
\begin{align*}
    \sum_{k=1}^K \vx^+_{\windex,k} = \frac{1}{2} (\kappa_{\windex,K} + \upsilon_{\windex,K}) \hspace{.5cm} \text{and} \hspace{.5cm}
    \sum_{k=1}^K \vx^-_{\windex,k} = \frac{1}{2} (\kappa_{\windex,K} - \upsilon_{\windex,K}).
\end{align*}

Now, observe that
\begin{align*}
    \sum_{\windex=1}^\workers \kappa_{\windex,K} = \sum_{\exindex=1}^{2K} \sum_{\windex=1}^\workers \vz_{\windex,\perm_\windex(\exindex)} \hspace{.5cm} \text{and} \hspace{.5cm}
    \sum_{\windex=1}^\workers \upsilon_{\windex,K} = \sum_{p=1}^{\workers K} \sgn_p \vy_p.
\end{align*}

Therefore,
\begin{align*}
    \max_{K \in [\workerexamples/2]} \norm{\sum_{k=1}^K \sum_{\windex=1}^\workers \vx^+_{\windex,k}}_\infty &\le \frac{1}{2}\left(\max_{K \in [\workerexamples/2]} \norm{\sum_{\windex=1}^\workers \kappa_{K,\windex}}_\infty + \max_{K \in [\workerexamples/2]} \norm{\sum_{\windex=1}^\workers \upsilon_{K,\windex}}_\infty\right)\\
    &\le \frac{1}{2}\max_{K \in [\workerexamples/2]}\norm{\sum_{\exindex=1}^{2K} \sum_{\windex=1}^\workers \vz_{\windex,\exindex}}_\infty + \tilde{A}c_2 \hspace{1em} \text{By substituting above and (\ref{app:lemma1:a})}\\
    &\le \frac{1}{2}\max_{k \in [\workerexamples]}\norm{\sum_{\exindex=1}^{k} \sum_{\windex=1}^\workers \vz_{\windex,\exindex}}_\infty + \tilde{A}c_2. 
\end{align*}

And similarly,
\begin{align*}
    \max_{K \in [\workerexamples/2]} \norm{\sum_{k=1}^K \sum_{\windex=1}^\workers \vx^-_{\windex,k}} &\le \frac{1}{2}\left(\max_{K \in [\workerexamples/2]} \norm{\sum_{\windex=1}^\workers \kappa_{K,\windex}}_\infty + \max_{K \in [\workerexamples/2]} \norm{\sum_{\windex=1}^\workers \upsilon_{K,\windex}}_\infty\right) \\
    &\le \frac{1}{2}\max_{k \in [\workerexamples]}\norm{\sum_{\exindex=1}^{k} \sum_{\windex=1}^\workers  \vz_{\windex,\exindex}}_\infty + \tilde{A}c_2. 
\end{align*}

Applying the new permutation $\perm'_\windex(\exindex)$ on the vectors $\vz_{\windex,\perm_\windex(\exindex)}$, we get for each $\windex \in [\workers]$ the permuted sequence
\[
    \vx^+_{\windex,1},\dots, \vx^+_{\windex,\workerexamples/2}, \vx^-_{\windex,\workerexamples/2},\dots, \vx^-_{\windex,1}.
\]

Thus, we need to bound the herding objective of the sequence
\[
    \sum_{\windex=1}^\workers \vx^+_{\windex,1},\dots, \sum_{\windex=1}^\workers \vx^+_{\windex,\workerexamples/2}, \sum_{\windex=1}^\workers \vx^-_{\windex,\workerexamples/2},\dots, \sum_{\windex=1}^\workers \vx^-_{\windex,1}.
\]

If the partial sums above peak at $t_0 \le \workerexamples/2$, then we can bound the parallel herding objective as
\begin{align*}
    \norm{\sum_{k=1}^{t_0} \sum_{\windex=1}^\workers \vx^+_{\windex,k}}_\infty = \max_{K \in [\workerexamples/2]} \norm{\sum_{k=1}^K \sum_{\windex=1}^\workers \vx^+_{\windex,k}}_\infty \le \frac{1}{2}\max_{k \in [\workerexamples]}\norm{\sum_{\exindex=1}^{k} \sum_{\windex=1}^\workers \vz_{\windex,\exindex}}_\infty + \tilde{A}c_2;
\end{align*}

otherwise, we can bound the parallel herding objective as
\begin{align*}
    \norm{\sum_{\exindex=1}^\workerexamples \sum_{\windex=1}^\workers \vz_{\windex,\exindex} - \sum_{k=1}^{m-t_0}\sum_{\windex=1}^\workers \vx^-_{\windex,k}}_\infty &\le \norm{\sum_{\exindex=1}^\workerexamples \sum_{\windex=1}^\workers \vz_{\windex,\exindex}}_\infty + \norm{\sum_{k=1}^{m-t_0}\sum_{\windex=1}^\workers \vx^-_{\windex,k}}_\infty\\ 
    &\le c_1 + \frac{1}{2}\max_{t \in [\workerexamples]}\norm{\sum_{\exindex=1}^{t} \sum_{\windex=1}^\workers \vz_{\windex,\exindex}}_\infty + \tilde{A}c_2,
\end{align*}

since in Algorithm~\ref{alg:reorder} the list of vectors with negative signs is reversed before concatenated. 

The claim follows.
\end{proof}

\subsection{Notation and observations}\label{app:proof:note}

We begin with three notes that we will use throughout the intermediate results we present in this section. We will use the lemmas presented here to prove our main results: Theorems~\ref{thm:dgrab:smooth} and~\ref{thm:dgrab:PL} in Appendix~\ref{app:thm:proof}.

\begin{enumerate}[leftmargin=.5cm]
    \item \custompar{A single $t$-th update} First, recall that one $\eindex$-th step of the parameter update can be written as
    \[
    \vw_t^{j+1} = \vw_t^j - \frac{\alpha}{\workers} \sum_{i=1}^\workers \nabla f^i(\vw_t^j; \; \pi_{t,i}(j)), \quad \forall j \in [\workerexamples]
    \]
    We will use the convention $\vw_{t+1} \triangleq \vw_{t+1}^1 \triangleq \vw_t^{\workerexamples+1}$. 
    \item \custompar{The maximum amount a parameter can change over an epoch} The key quantity in our proof is $\Delta_t$, which is the maximum amount that a parameter in $\weights$ can change in epoch $t$. That is,

    \begin{align}
    \label{eq:deltat}
    \Delta_t &\triangleq \max_{k \in [\workerexamples]} \norm{\vw_t^{k+1} - \vw_t}_\infty\nonumber\\
    &= \frac{\alpha}{\workers}\max_{k \in [\workerexamples]}\norm{\sum_{j=1}^k\sum_{i=1}^\workers\nabla f^i(\vw_t^j; \pi_{t,i}(j))}_\infty.
    \end{align}

    Following this definition of $\Delta_t$, we note that the maximum amount that a parameter in $\weights$ can change over two different epochs is $2\Delta_t$.
    That is, we observe 
    
    \begin{align*}
    \norm{\vw_t^j - \vw_t^k}_\infty &\le \norm{\vw_t^j - \vw_t}_\infty + \norm{\vw^k_t - \vw_t}_\infty \le 2 \Delta_t\\
    \norm{\vw_{t+1}^j - \vw_t^k}_\infty &\le \norm{\vw_{t+1}^j - \vw_{t+1}}_\infty + \norm{\vw_{t+1} - \vw_t}_\infty + \norm{\vw_t^k - \vw_t}_\infty \le \Delta_{t+1} + 2\Delta_t, \hspace{.5cm} \forall j,k \in [\workerexamples]. 
    \end{align*}

    We make repeated use of this relation in the results that follow, which we typically will use in combination with the Lipschitz assumption to bound gradients of the same loss function but with different parameters. 

    \item \custompar{Bounding loss at epoch $t$} We will denote $F_t = f(\vw_t) - f(\vw^*)$ where $\vw^*$ is the minimizer of $f$ which we assume to be bounded from below. 
    
\end{enumerate}

\input{section/99-appendix/32-cdgrab/310-app-cdgrab-proof-loss}
\input{section/99-appendix/32-cdgrab/320-app-cdgrab-proof-alg}
\input{section/99-appendix/32-cdgrab/330-app-cdgrab-proof-last}
\input{section/99-appendix/32-cdgrab/340-app-cdgrab-proof-thms}

%% file: section/99-appendix/32-cdgrab/310-app-cdgrab-proof-loss.tex
\subsection{Assuming $L_{2,\infty}$-smoothness: results on the amount the loss can change over one epoch)}\label{app:proof:smooth}

We will next prove an intermediate result regarding that bounds the loss $\loss$ at epoch $t + 1$ in relation to the loss at the prior epoch $t$ (Lemma~\ref{lemma:loss}).  That is, we prove results about how much the loss with respect to the parameters can change over the course of one epoch.

\begin{lemma}\label{lemma:loss} 
If the loss $\loss$ is $L_{2,\infty}$-smooth and the learning rate $\alpha \le \frac{1}{\workerexamples L_{2,\infty}}$, then
\begin{align}
\label{lem:smoothness:eq:critical}
f(\vw_{t+1}) \le f(\vw_t) + \frac{\alpha \workerexamples L_{2,\infty}^2}{2}\Delta_t^2 - \frac{\alpha \workerexamples}{2} \norm{\nabla f(\vw_t)}_2^2.
\end{align}
\end{lemma}

\begin{proof}
We begin with the definition of $L_{2,\infty}$-smoothness, with respect to loss $\loss$:
\begin{align*}
f(\vw_{t+1}) &\le f(\vw_t) + \nabla f(\vw_t)^\top(\vw_{t+1} - \vw_t) + \frac{L_{2,\infty}}{2}\|\vw_{t+1} - \vw_t\|_2^2\\
\end{align*}

Also observe that 
\begin{align}
\label{eq:normsquareddiff}
-\nabla f(\vw_t)^\top(\vw_t - \vw_{t+1}) &= -\frac{\alpha \workerexamples}{2}2\nabla f(\vw_t)^\top\left(\frac{\vw_t - \vw_{t+1}}{\alpha \workerexamples}\right)\nonumber\\
&= \frac{\alpha \workerexamples}{2} \left( \norm{\nabla f(\vw_t) - \frac{(\vw_t - \vw_{t+1})}{\alpha \workerexamples}}_2^2 - \norm{\nabla f(\vw_t)}_2^2 - \norm{\frac{(\vw_t - \vw_{t+1})}{\alpha \workerexamples}}_2^2 \right). 
\end{align}

Combining the above --- i.e., the definition of $L_{2,\infty}$-smoothness with (\ref{eq:normsquareddiff}) --- we get
\begin{align*}
f(\vw_{t+1}) &\le f(\vw_t) + \frac{\alpha \workerexamples}{2}\norm{\nabla f(\vw_t) - \frac{(\vw_t - \vw_{t+1})}{\alpha \workerexamples}}_2^2 - \frac{\alpha \workerexamples}{2}\norm{\nabla f(\vw_t)}_2^2
&\quad + \frac{\alpha \workerexamples L_{2,\infty} - 1}{2\alpha \workerexamples}\norm{\vw_t - \vw_{t+1}}_2^2.\\
\end{align*}
The last term on the right-hand side is $\le 0$ by the assumption that the learning rate $\alpha \le \frac{1}{\workerexamples L_{2,\infty}}$. Therefore,
\begin{align}
\label{eq:rhs-2}
f(\vw_{t+1}) \le f(\vw_t) + \frac{\alpha \workerexamples}{2}\norm{\nabla f(\vw_t) - \frac{(\vw_t - \vw_{t+1})}{\alpha \workerexamples}}_2^2 - \frac{\alpha \workerexamples}{2}\norm{\nabla f(\vw_t)}_2^2.
\end{align}

We next bound the second term on the right-hand side by $\Delta_t$ (\ref{eq:deltat}):
\begin{align*}
\norm{\nabla f(\vw_t) - \frac{(\vw_{t} - \vw_{t+1})}{\alpha \workerexamples}}_2^2 &= \norm{\frac{1}{\workers \workerexamples}\sum_{j=1}^\workers\sum_{i=1}^\workerexamples \nabla f^i(\vw_t, \pi_t(j)) - \frac{1}{\workers \workerexamples}\sum_{j=1}^\workers \sum_{i=1}^\workerexamples \nabla f^i(\vw_{t}^{j}; \pi_t(j))}_2^2\\
&\le \frac{1}{\workers \workerexamples}\sum_{j=1}^\workers\sum_{i=1}^\workerexamples\norm{\nabla f^i(\vw_t, \pi_t(j)) - \nabla f^i(\vw_{t}^{j}; \pi_t(j))}_2^2\\
&\le \frac{L_{2,\infty}^2}{\workers \workerexamples}\sum_{j=1}^\workers\sum_{i=1}^\workerexamples\norm{\vw_t^j - \vw_t}_\infty^2, 
\end{align*}
where we have used $L_{2,\infty}$-smoothness (Assumption~\ref{ass:smoothness}) in the last inequality. Substituting $\Delta_t$, we get
\begin{align*}
\frac{L_{2,\infty}^2}{\workers \workerexamples}\sum_{j=1}^\workers\sum_{i=1}^\workerexamples\norm{\vw_t^j - \vw_t}_\infty^2 \quad \le  \quad L_{2,\infty}^2 \Delta_t^2. 
\end{align*}
Plugging the above into (\ref{eq:rhs-2}), we get
\begin{align*}
f(\vw_{t+1}) \le f(\vw_t) + \frac{\alpha \workerexamples L_{2,\infty}^2}{2}\Delta_t^2 - \frac{\alpha \workerexamples}{2} \norm{\nabla f(\vw_t)}_2^2, 
\end{align*}
yielding the claim.
\end{proof}

We next build slightly on Lemma~\ref{lemma:loss} to make two additional observations. 
First: 
\begin{lemma}
    \label{lem:jens}
    If the loss $\loss$ is $L_{2,\infty}$-smooth and the learning rate $\alpha \le \frac{1}{\workerexamples L_{2,\infty}}$, then 
     \begin{align*}
    \frac{1}{\epochs}\sum_{t=1}^{\epochs} \norm{\nabla f(\vw_t)}_2^2 
    &\le \frac{2F_1}{\alpha \workerexamples T} + \frac{L_{2,\infty}^2}{T}\sum_{t=1}^{T}\Delta_{t}^2,
    \end{align*}
where $F_1$ comes from Theorem~\ref{thm:dgrab:smooth}.
\end{lemma}

\begin{proof}
Using Lemma~\ref{lemma:loss} and Jensen's inequality, we average (\ref{lem:smoothness:eq:critical}) over $t \in [\epochs]$ and match terms, yielding 
\begin{align*}
\frac{1}{\epochs}\sum_{t=1}^{\epochs} \norm{\nabla f(\vw_t)}_2^2 &\le \frac{2(f(\vw_1) - f(\vw_{\epochs+1}))}{\alpha \workerexamples \epochs} + \frac{L_{2,\infty}^2}{\epochs}\sum_{t=1}^T\Delta_t^2.
\end{align*}
Substituting $F_1$, we get
\begin{align*}
&\le \frac{2F_1}{\alpha \workerexamples T} + \frac{L_{2,\infty}^2}{T}\sum_{t=1}^T\Delta_t^2,
\end{align*}
yielding the claim.
\end{proof}

We next build on Lemma~\ref{lemma:loss} by further assuming the P.L. assumption holds. 

\begin{lemma}
\label{lem:jens-pl}
If the loss $\loss$ is $L_{2,\infty}$-smooth, the learning rate $\alpha \le \frac{1}{\workerexamples L_{2,\infty}}$, and the P.L. assumption (Assumption~\ref{ass:PL}) holds, then, for $\rho = 1 - \frac{\alpha \workerexamples \mu}{2}$ 
\begin{align*}
F_{T+1} \le \rho^T F_1 + \frac{\alpha \workerexamples L_{2,\infty}^2}{2}\sum_{t=1}^T\rho^{T-t}\left(\Delta_t^2 - \frac{1}{2L_{2,\infty}^2} \norm{\nabla f(\vw_t)}_2^2\right).
\end{align*}
\end{lemma}

\begin{proof}
From Lemma~\ref{lemma:loss}, we got (\ref{lem:smoothness:eq:critical}), i.e.,
\begin{align*}
f(\vw_{t+1}) \le f(\vw_t) + \frac{\alpha \workerexamples L_{2,\infty}^2}{2}\Delta_t^2 - \frac{\alpha \workerexamples}{2} \norm{\nabla f(\vw_t)}_2^2, 
\end{align*}

Applying the P.L. assumption (Assumption~\ref{ass:PL}) to (\ref{lem:smoothness:eq:critical}), we get
\begin{align*}
f(\vw_{t+1}) &\le f(\vw_t) + \frac{\alpha \workerexamples L_{2,\infty}^2}{2}\Delta_t^2 - \frac{\alpha \workerexamples}{4} \norm{\nabla f(\vw_t)}_2^2 - \frac{\alpha \workerexamples}{4} \norm{\nabla f(\vw_t)}_2^2\\
&\le f(\vw_t) + \frac{\alpha \workerexamples L_{2,\infty}^2}{2}\Delta_t^2 - \frac{\alpha \workerexamples \mu}{2}(f(\vw_t) - f(\vw^*))  - \frac{\alpha \workerexamples}{4} \norm{\nabla f(\vw_t)}_2^2.
\end{align*}
Subtracting $\loss^*$ from both sides, we get
\begin{align*}
f(\vw_{t+1}) - f^* \le \left(1 - \frac{\alpha \workerexamples \mu}{2} \right)(f(\vw_t) - f^*) + \frac{\alpha \workerexamples}{2}\left(L_{2,\infty}^2\Delta_t^2 - \frac{1}{2} \norm{\nabla f(\vw_t)}_2^2\right). 
\end{align*}
For $\rho = 1 - \frac{\alpha \workerexamples \mu}{2}$, we then apply the above inequality recursively for $t \in [T]$, yielding the claim: 
\begin{align*}
F_{T+1} \le \rho^T F_1 + \frac{\alpha \workerexamples L_{2,\infty}^2}{2}\sum_{t=1}^T\rho^{T-t}\left(\Delta_t^2 - \frac{1}{2L_{2,\infty}^2} \norm{\nabla f(\vw_t)}_2^2\right). 
\end{align*}
\end{proof}

%% file: section/99-appendix/32-cdgrab/320-app-cdgrab-proof-alg.tex
\subsection{Assuming bounded gradient variance and heterogeneity: results applying Algorithm~\ref{alg:dgrab:vector:server}}\label{app:proof:other}

We next prove a result that builds on Lemma~\ref{lem:pair-balance} and our one-step version of the server-side $\mathsf{PairBalance}$ algorithm (Algorithm~\ref{alg:dgrab:vector:server}). 

We begin by introducing some additional notation. 
Namely, we will call $\pi^{-1}$ the operation that, given an example, yields the index in the permutation for that example. 
For instance, $\pi_{t+1, \windex}(\exindex)$ returns the example at the $\exindex$-th index for the $\windex$-th worker's $t+1$ permutation.  
Let us denote that example $\tau$. Then, $\pi_{t,i}^{-1}\pi_{t+1,i}(j)$ is equivalent to applying $\pi_{t,i}^{-1}$ to $\tau$: 
it takes the example $\tau$ and returns $\tau$'s associated index in the $\windex$-th worker's epoch $t$'s permutation (in this case, the prior epoch's permutation). 

We will make use of this notation in the following Lemma.

\begin{lemma}
Assume bounded gradient variance (Assumption~\ref{ass:inner-deviation}), bounded gradient heterogeneity (Assumption~\ref{ass:outer-deviation}), and  $L_{2,\infty}$-smoothness (Assumption~\ref{ass:smoothness}). 
For $t \in [T]$ and $\delta > 0$, if we apply Algorithm~\ref{alg:dgrab:vector:server} to the gradients $\nabla f^i(\vw_t^j; \pi_t^i(j))$ at epoch $t$ to produce the next permutation $\pi_{t+1,i}$ for epoch $t+1$, then, with probability at least $1 - \delta$, 
\begin{align*}
\Delta_{t+1} \quad \le \quad \frac{1}{2}\Delta_t \;\; + \;\; \alpha L_{2,\infty} \left(4\workerexamples + \frac{2\tilde{A}}{\workers}\right)\Delta_{t} \;\; + \;\; \alpha \workerexamples L_{2,\infty} \Delta_{t+1} \;\; + \;\; \frac{\alpha (\varsigma + \sigma)\tilde{A}}{\workers} \;\; + \;\; \alpha \workerexamples\norm{\nabla f(\vw_{t+1})}_2, 
\end{align*}
\label{lem:grad-balance}
where $\tilde A$ comes from Theorem~\ref{statement:alweiss}.
\end{lemma}

\begin{proof}
We start with the triangle inequality:
\begin{align}
\label{lem:grad-balance:eq:triangle}
\norm{\sum_{j=1}^{k}\sum_{i=1}^\workers \nabla f^i(\vw_{t+1}^j; \pi_{t+1,i}(j))}_\infty \le \norm{\sum_{j=1}^{k}\sum_{i=1}^\workers \nabla f^i(\vw_{t}^{\pi_{t,i}^{-1}\pi_{t+1,i}(j)}; \pi_{t+1,i}(j))}_\infty + \nonumber\\
\norm{\sum_{j=1}^{k}\sum_{i=1}^\workers \left(\nabla f^i(\vw_{t+1,i}^j, \pi_{t+1,i}(j)) - \nabla f^i(\vw_{t,i}^{\pi_{t,i}^{-1}\pi_{t+1,i}(j)}; \pi_{t+1,i}(j))\right)}_\infty
\end{align}

We use Lemma~\ref{lem:pair-balance} to bound the first term on the right-hand side of (\ref{lem:grad-balance:eq:triangle}) from Lemma~\ref{lemma:loss}. 

That is, let 
\[
    \vz_{i,j} = \nabla f^i(\vw_{t}^{\pi_{t,i}^{-1}(j)}; j),
\]
so that 
\[
    \vz_{i,\pi_{t+1,i}(j)} = \nabla f^i(\vw_{t}^{\pi_{t,i}^{-1}\pi_{t+1,i}(j)}; \pi_{t+1,i}(j)).
\]

The upper bounds for $\norm{\vz_{i,j} - \frac{1}{\workers \workerexamples}\sum_{r,s} \vz_{r,s}}_\infty$ and $\norm{\sum_{i,j} \vz_{i,j}}_\infty$ are:
\begin{align*}
    \norm{\nabla f^i(\vw_t^{j};\pi_{t,i}(j)) - \frac{1}{\workers \workerexamples}\sum_{r=1}^\workers \sum_{s=1}^\workerexamples \nabla f^s(\vw_t^{r};\pi_{t,s}(r))}_\infty,
\end{align*}
which are 
\begin{align*}
\le &\norm{\nabla f^i(\vw_{t}^{j}; \pi_{t,i}(j)) - \frac{1}{\workers \workerexamples}\sum_{r=1}^\workers \sum_{s=1}^\workerexamples \nabla f^s(\vw_{t}^{j};\pi_{t,s}(r))}_\infty + \\ 
    & \quad \norm{\frac{1}{\workers \workerexamples}\sum_{r=1}^\workers \sum_{s=1}^\workerexamples \nabla f^s(\vw_{t}^{j}; \pi_{t,s}(r)) - \frac{1}{\workers \workerexamples}\sum_{r=1}^\workers \sum_{s=1}^\workerexamples \nabla f^s(\vw_{t}^{r}; \pi_{t,s}(r))}_\infty.
\end{align*}

We can rewrite the above to be
\begin{align*}
&\le \norm{\nabla f^i(\vw_{t}^{j}; \pi_{t,i}(j)) - \nabla f(\vw_{t}^{j})}_\infty + \frac{L_{2,\infty}}{\workers \workerexamples}\sum_{r=1}^m\workers \sum_{s=1}^\workerexamples \norm{\vw_t^j - \vw_t^r}_\infty\\
    &\le \varsigma + \sigma + 2 L_{2,\infty}\Delta_{t},
\end{align*}
by Assumptions~\ref{ass:inner-deviation},~\ref{ass:outer-deviation}, and~\ref{ass:smoothness}, and by the definition of $\Delta_t$ (\ref{eq:deltat}). 

Now, observe that
\begin{align*}
\norm{\sum_{i=1}^\workers \sum_{j=1}^\workerexamples \nabla f^i(\vw_t^j;\pi_{t,i}(j))}_\infty \le \norm{\sum_{i=1}^\workers \sum_{j=1}^\workerexamples\nabla f^i(\vw_{t}^{j}; \pi_{t,i}(j)) - \sum_{i=1}^\workers\sum_{j=1}^\workerexamples\nabla f^i(\vw_{t+1};\pi_{t,i}(j))}_\infty +\\ \norm{\sum_{i=1}^\workers\sum_{j=1}^\workerexamples \nabla f^i(\vw_{t+1};\pi_{t,i}(j))}_\infty.
\end{align*}
By using the above, we can rewrite the right-hand side to be
\begin{align*}
&\le \sum_{i=1}^\workers\sum_{j=1}^\workerexamples L_{2,\infty}\norm{\vw_{t}^{j} - \vw_{t+1}}_\infty + \workers \workerexamples \norm{\nabla f(\vw_{t+1})}_\infty\\
&\le 2\workers \workerexamples L_{2,\infty}\Delta_{t} + \workers \workerexamples \norm{\nabla f(\vw_{t+1})}_2.
\end{align*}
Therefore, by Lemma~\ref{lem:pair-balance}, 
\begin{align*}
\max_{k \in [\workerexamples]}\norm{\sum_{j=1}^{k}\sum_{i=1}^\workers \nabla f^i(\vw_{t}^{\pi_{t,i}^{-1}\pi_{t+1,i}(j)}, \pi_{t+1,i}(j))}_\infty \le 
&\max_{k\in[\workerexamples]}\norm{\sum_{j=1}^{k}\sum_{i=1}^\workers \nabla f^i(\vw_{t}^{j}, \pi_{t,i}(j))}_\infty\\
+&2\workers \workerexamples L_{2,\infty}\Delta_{t} + \norm{\nabla f(\vw_{t+1})}_2 + (\varsigma + \sigma + 2L_{2,\infty} \Delta_{t}) \tilde{A}.
\end{align*}
The second term of the triangle inequality (\ref{lem:grad-balance:eq:triangle}) can be bounded as 
\begin{align*}
\norm{\sum_{j=1}^{k}\sum_{i=1}^\workers \left(\nabla f^i(\vw_{t+1,i}^j; \pi_{t+1,i}(j)) - \nabla f^i(\vw_{t,i}^{\pi_{t,i}^{-1}\pi_{t+1,i}(j)}; \pi_{t+1,i}(j))\right)}_\infty,
\end{align*}
which is
\begin{align*}
&\le \sum_{j=1}^k\sum_{i=1}^\workers \norm{\vw_{t+1,i}^j - \vw_{t,i}^{\pi_{t,i}^{-1}\pi_{t+1,i}(j)}}_\infty\\
&\le \workers \workerexamples L_{2,\infty} (\Delta_{t+1} + 2\Delta_{t}).
\end{align*}
Substituting these bounds into the right-hand side of the triangle inequality (\ref{lem:grad-balance:eq:triangle}), taking the max of both sides, and grouping terms, we get
\begin{align*}
\max_{k\in[\workerexamples]}\norm{\sum_{j=1}^{k}\sum_{i=1}^\workers \nabla f^i(\vw_{t+1,i}^j, \pi_{t+1,i}(j))}_\infty \le& \frac{1}{2}\max_{k\in[\workerexamples]}\norm{\sum_{j=1}^{k}\sum_{i=1}^\workers \nabla f^i(\vw_{t}^{j}, \pi_{t,i}(j))}_\infty\\
&+ L_{2,\infty}(4\workers \workerexamples + 2\tilde{A})\Delta_{t} + \workers \workerexamples L_{2,\infty} \Delta_{t+1} + (\varsigma + \sigma)\tilde{A}\\
&+ \workers \workerexamples \norm{\nabla f(\vw_{t+1})}_2.\\
\end{align*}
Multiplying both sides by $\frac{\alpha}{\workers}$ and using the definition of $\Delta_t$ (\ref{eq:deltat}), we get the claim.
\end{proof}

%% file: section/99-appendix/32-cdgrab/330-app-cdgrab-proof-last.tex
\subsection{Combining the prior intermediate results: proofs over multiple steps}\label{app:proof:last}

\begin{lemma} 
If the learning rate $\alpha \le \frac{1}{16 L_{2,\infty} (2\workerexamples + \tilde{A}/\workers)}$, then
\begin{align*}
\frac{1}{T}\sum_{t=1}^{T} \Delta_t^2 \le \frac{21\alpha^2(\varsigma + \sigma)^2\tilde{A}^2}{\workers^2} + \frac{9\alpha^2 \workerexamples^2 \sigma^2}{T} + 21\alpha^2 \workerexamples^2 \frac{1}{T}\sum_{t=1}^T\norm{\nabla f(\vw_t)}_2^2.
\end{align*}
\label{lem:Delta-1}
\end{lemma}

\begin{proof}
First, we bound $\Delta_1^2$. 

We start with a series of triangle inequalities:
\begin{align*}
\frac{\alpha}{\workers}\norm{\sum_{j=1}^{k}\sum_{i=1}^\workers \nabla f^i(\vw_1^j, \pi_{1,i}(j))}_\infty \le & \frac{\alpha}{\workers}\norm{\sum_{j=1}^k \sum_{i=1}^\workers \nabla f^i(\vw_1^j, \pi_{1,i}(j)) - \sum_{j=1}^k \sum_{i=1}^\workers\nabla f^i(\vw_1, \pi_{1,i}(j))}_\infty\\
&+\frac{\alpha}{\workers}\norm{\sum_{j=1}^k\sum_{i=1}^\workers \left(\nabla f^i(\vw_1, \pi_{1,i}(j)) - \nabla f^i(\vw_1)\right)}_\infty + \alpha k \norm{\nabla f(\vw_1)}_\infty\\
\le& \frac{\alpha}{\workers}\sum_{j=1}^k\sum_{i=1}^\workers L_{2,\infty} \norm{\vw_1^j - \vw_1}_\infty + \alpha k \sigma + \alpha k \norm{\nabla f(\vw_1)}_2.
\end{align*}

We next take the max of both sides with respect to $k \in [\workerexamples]$:  
\begin{align*}
\Delta_1 &\le \alpha \workerexamples L_{2,\infty} \Delta_1 + \alpha \workerexamples \sigma + \alpha \workerexamples \norm{\nabla f(\vw_1)}_2\\ 
&\le (1/32) \Delta_1 + \alpha \workerexamples \sigma + \alpha \workerexamples \norm{\nabla f(\vw_1)}_2 \hspace{.5in} \text{(since} \hspace{.2cm} \alpha \le \frac{1}{32 \workerexamples L_{2,\infty}} \text{)} \\
&\le (32/31)\alpha \workerexamples \sigma + (32/31)\alpha \workerexamples \norm{\nabla f(\vw_1)}_2,
\end{align*}

Squaring both sides:
\begin{align}
\label{lem:Delta:eq:Delta_1}
\Delta_1^2 &\le 3\alpha^2 \workerexamples^2 \sigma^2 + 3\alpha^2 \workerexamples^2 \norm{\nabla f(\vw_1)}_2^2.
\end{align}

Now, we use Lemma~\ref{lem:grad-balance} to get the relationship between $\Delta_{t+1}$ and $\Delta_t$ for $t \in [T]$. 

Recall that 
\begin{align*}
\Delta_{t+1} &\le \frac{1}{2}\Delta_t + \alpha L_{2,\infty} \left(4 \workerexamples + \frac{2\tilde{A}}{\workers}\right)\Delta_{t} + \alpha \workerexamples L_{2,\infty} \Delta_{t+1} + \frac{\alpha (\varsigma + \sigma)\tilde{A}}{\workers} + \alpha \workerexamples \norm{\nabla f(\vw_{t+1})}_2\\
\end{align*}

Because $\alpha \le \frac{1}{16 L_{2,\infty} (2\workerexamples + \tilde{A}/\workers)}$, we can rewrite the above as
\begin{align*}
\Delta_{t+1} &\le \frac{1}{2}\Delta_t + (1/8)\Delta_{t} + (1/32) \Delta_{t+1} + \frac{\alpha (\varsigma + \sigma)\tilde{A}}{\workers} + \alpha \workerexamples \norm{\nabla f(\vw_{t+1})}_2. 
\end{align*}

Squaring both sides:
\begin{align*}
(31/32)^2\Delta_{t+1}^2 &\le \frac{1}{2}\Delta_{t}^2 + 2\left((1/8)\Delta_{t} + \frac{\alpha(\varsigma + \sigma)\tilde{A}}{\workers} +  \alpha \workerexamples \norm{\nabla f(\vw_{t+1})}_2\right)^2\\
&\le \frac{1}{2}\Delta_{t}^2 + (6/8^2)\Delta_{t}^2 + \frac{6\alpha^2(\varsigma + \sigma)^2\tilde{A}^2}{\workers^2} +  6\alpha^2 \workerexamples^2 \norm{\nabla f(\vw_{t+1})}_2^2,
\end{align*}

so that
\begin{align}
\begin{split}
\label{lem:Delta:eq:critical}
\Delta_{t+1}^2 &\le (32/31)^2(1/2 + 6/8^2)\Delta_t^2 + \frac{(32/31)^2 6\alpha^2(\varsigma + \sigma)^2\tilde{A}^2}{\workers^2} +  (32/31)^2 6\alpha^2 \workerexamples^2 \norm{\nabla f(\vw_{t+1})}_2^2\\
&\le (2/3)\Delta_t^2 + \frac{7\alpha^2(\varsigma + \sigma)^2\tilde{A}^2}{\workers^2} +  7\alpha^2 \workerexamples^2 \norm{\nabla f(\vw_{t+1})}_2^2.
\end{split}
\end{align}

We next sum (\ref{lem:Delta:eq:critical}) over $t \in [T-1]$ and add (\ref{lem:Delta:eq:Delta_1}):
\begin{align*}
\Delta_1^2 + \sum_{t=2}^T \Delta_t^2 &\le (2/3)\sum_{t=2}^T \Delta_{t-1}^2 + \frac{(T-1)7\alpha^2(\varsigma + \sigma)^2\tilde{A}^2}{\workers^2} + 3\alpha^2 \workerexamples^2 \sigma^2 + 7\alpha^2 \workerexamples^2 \sum_{t=1}^T\norm{\nabla f(\vw_t)}_2^2\\
\frac{1}{T}\sum_{t=1}^T \Delta_t^2 &\le (2/3)\frac{1}{T}\sum_{t=1}^T \Delta_t^2 + \frac{7\alpha^2(\varsigma + \sigma)^2\tilde{A}^2}{\workers^2} + \frac{3\alpha^2 \workerexamples^2 \sigma^2}{T} + 7\alpha^2 \workerexamples^2 \frac{1}{T}\sum_{t=1}^T\norm{\nabla f(\vw_t)}_2^2\\
&\le \frac{21\alpha^2(\varsigma + \sigma)^2\tilde{A}^2}{\workers^2} + \frac{9\alpha^2 \workerexamples^2 \sigma^2}{T} + 21\alpha^2 \workerexamples^2 \frac{1}{T}\sum_{t=1}^T\norm{\nabla f(\vw_t)}_2^2,
\end{align*}
yielding the claim. 
\end{proof}

We next build on Lemma~\ref{lem:Delta-1}. 
\begin{lemma} 
If $\alpha \le \frac{2}{9 \workerexamples \mu}$, then, for $\rho = 1 - \frac{\alpha \workerexamples \mu}{2}$, 
\begin{align*}
\sum_{t=1}^T \rho^{T-t} \Delta_t^2
&\le 12\rho^{T-1}\alpha^2 \workerexamples^2 \sigma^2 + \frac{28\rho\alpha^2 (\varsigma + \sigma)^2\tilde{A}^2}{(1-\rho)\workers^2} + \frac{1}{2L_{2,\infty}^2}\sum_{t=1}^T\rho^{T-t}\norm{\nabla f(\vw_t)}_2^2.
\end{align*}
\label{lem:Delta-2}
\end{lemma}

\begin{proof}
Recall (\ref{lem:Delta:eq:Delta_1}) from Lemma~\ref{lem:Delta-1}:
\begin{align*}
\Delta_1^2 &\le 3\alpha^2 \workerexamples^2 \sigma^2 + 3\alpha^2 \workerexamples^2 \norm{\nabla f(\vw_1)}_2^2.
\end{align*}

We multiply each term $\Delta_t$ with $\rho^{T-t}$ for $t \in [T]$ and get 
\begin{align}
\label{eq:delta1'}
\rho^{T-1}\Delta_1^2 &\le \rho^{T-1}3\alpha^2 \workerexamples^2 \sigma^2 + \rho^{T-1}3\alpha^2 \workerexamples^2 \norm{\nabla f(\vw_1)}_2^2. 
\end{align}

Similarly, recall (\ref{lem:Delta:eq:critical}) from Lemma~\ref{lem:Delta-1}, 
\begin{align*}
\Delta_{t+1}^2 &\le (2/3)\Delta_t^2 + \frac{7\alpha^2(\varsigma + \sigma)^2\tilde{A}^2}{\workers^2} +  7\alpha^2 \workerexamples^2 \norm{\nabla f(\vw_{t+1})}_2^2,
\end{align*}

for which we also multiply  each term $\Delta_t$ with $\rho^{T-t}$ for $t \in [T]$, and get
\begin{align}
\begin{split}
\rho^{T-t}\Delta_{t}^2 &\le (2/3)\rho^{T-t}\Delta_{t-1}^2 + \rho^{T-t}\frac{7\alpha^2 (\varsigma + \sigma)^2\tilde{A}^2}{\workers^2} + \rho^{T-t}7\alpha^2 \workerexamples^2 \norm{\nabla f(\vw_t)}_2^2\\
&\le (3/4)\rho^{T-(t-1)}\Delta_{t-1}^2  + \rho^{T-t}\frac{7\alpha^2 (\varsigma + \sigma)^2\tilde{A}^2}{\workers^2} + \rho^{T-t}7\alpha^2 \workerexamples^2 \norm{\nabla f(\vw_t)}_2^2,\quad \forall t \in \{2,\dots,T\}, 
\end{split}
\label{eq:deltacritical'}
\end{align}

where we have used $\alpha \le \frac{2}{9\workerexamples\mu}$ so that $\rho = 1 - \frac{\alpha \workerexamples \mu}{2} \ge (2/3)(4/3)$. 

Next, we sum the bounds in (\ref{eq:delta1'}) and (\ref{eq:deltacritical'}) for $\rho^{T-t} \Delta_t$ for all $t \in [T]$, and we get 
\begin{align*}
\rho^{T-1}\Delta_1^2  + \sum_{t=2}^T \rho^{T-t} \Delta_t^2 &\le \frac{3}{4}\sum_{t=2}^T\rho^{T-(t-1)}\Delta_{t-1}^2 +  \rho^{T-1}3\alpha^2 \workerexamples^2 \sigma^2 + \sum_{t=1}^T\rho^{T-t}\frac{7\alpha^2 (\varsigma + \sigma)^2\tilde{A}^2}{\workers^2} + \\
& \quad 7\alpha^2 \workerexamples^2 \sum_{t=1}^T\rho^{T-t}\norm{\nabla f(\vw_t)}_2^2.
\end{align*}

We can rewrite the right-hand side as
\begin{align*}
&\le \frac{3}{4}\sum_{t=1}^T\rho^{T-t}\Delta_{t}^2 +  \rho^{T-1}3\alpha^2 \workerexamples^2 \sigma^2 + \frac{7\rho\alpha^2 (\varsigma + \sigma)^2\tilde{A}^2}{(1-\rho)\workers^2} + 7\alpha^2 \workerexamples^2 \sum_{t=1}^T\rho^{T-t}\norm{\nabla f(\vw_t)}_2^2\\
&\le 12\rho^{T-1}\alpha^2 \workerexamples^2 \sigma^2 + \frac{28\rho\alpha^2 (\varsigma + \sigma)^2\tilde{A}^2}{(1-\rho)\workers^2} + 28\alpha^2 \workerexamples^2 \sum_{t=1}^T\rho^{T-t}\norm{\nabla f(\vw_t)}_2^2.\\
\end{align*}

Lastly, we use $\alpha \le \frac{1}{\sqrt{56} \workerexamples L_{2,\infty}}$ to get:
\begin{align*}
\sum_{t=1}^T \rho^{T-t} \Delta_t^2
&\le 12\rho^{T-1}\alpha^2 \workerexamples^2 \sigma^2 + \frac{28\rho\alpha^2 (\varsigma + \sigma)^2\tilde{A}^2}{(1-\rho)\workers^2} + \frac{1}{2L_{2,\infty}^2}\sum_{t=1}^T\rho^{T-t}\norm{\nabla f(\vw_t)}^2.\\
\end{align*}
\end{proof}

%% file: section/99-appendix/32-cdgrab/340-app-cdgrab-proof-thms.tex
\subsection{Proof of Theorems~\ref{thm:dgrab:smooth} and~\ref{thm:dgrab:PL}}\label{app:thm:proof}

Using the Lemmas above, we next prove our main results, presented in Section~\ref{sec:cdgrab:theory}.

\begin{proof}[Proof of Theorem \ref{thm:dgrab:smooth}]
The given learning rate $\alpha$ satisfies the constraints of Lemma~\ref{lem:jens} and Lemma~\ref{lem:Delta-1}. 

Therefore, 
\begin{align*}
\frac{1}{T}\sum_{t=1}^{T} \norm{\nabla f(\vw_t)}_2^2 &\le \frac{2F_1}{\alpha \workerexamples T} + L_{2,\infty}^2\left(\frac{21(\alpha(\varsigma + \sigma)\tilde{A})^2}{\workers^2} + \frac{9(\alpha \workerexamples \sigma)^2}{T} + 21\alpha^2 \workerexamples^2 \frac{1}{T}\sum_{t=1}^T\norm{\nabla f(\vw_t)}_2^2\right)\\
&\le \frac{4F_1}{\alpha \workerexamples T} + \frac{42L_{2,\infty}^2(\alpha(\varsigma + \sigma)\tilde{A})^2}{\workers^2} + \frac{18L_{2,\infty}^2(\alpha \workerexamples \sigma)^2}{T},\\
\end{align*}
due to $\alpha \le \frac{1}{\sqrt{42} \workerexamples L_{2,\infty}}$. 

We next derive the convergence rate. Let $\Gamma = \frac{42(L_{2,\infty}(\varsigma + \sigma)\tilde{A})^2}{\workers^2} + \frac{18L_{2,\infty}^2 \workerexamples^2 \sigma^2}{T}$. Then,
\begin{align*}
\frac{1}{T}\sum_{t=1}^{T} \norm{\nabla f(\vw_t)}_2^2 \le \frac{4F_1}{\alpha \workerexamples T} + \Gamma \alpha^2.
\end{align*}
We then set $\alpha \le \left(\frac{4F_1}{\workerexamples \Gamma T}\right)^{1/3}$. So we will have $\alpha = \min\left\{\frac{1}{16L_{2,\infty} (2\workerexamples + \tilde{A}/\workers)}, \left(\frac{4F_1}{\workerexamples\Gamma T}\right)^{1/3}\right\}$ or
\begin{align*}
\frac{1}{\alpha} = \max\left\{16 L_{2,\infty} (2\workerexamples + \tilde{A}/\workers), \left(\frac{4F_1}{\workerexamples\Gamma T}\right)^{-1/3}\right\}.
\end{align*}
Substitute $\alpha$:
\begin{align*}
\frac{1}{T}\sum_{t=1}^{T} \norm{\nabla f(\vw_t)}_2^2 &\le \frac{4F_1}{\workerexamples T} \left\{16 L_{2,\infty} (2\workerexamples + \tilde{A}/\workers) + \left(\frac{4F_1}{\workerexamples\Gamma T}\right)^{-1/3}\right\} + \Gamma \left(\frac{4F_1}{\workerexamples\Gamma T}\right)^{2/3}\\
&\le \left(\frac{4F_1}{\workerexamples T}\right)^{2/3}\Gamma^{1/3} + \frac{64F_1 L_{2,\infty} (2 + \tilde{A}/(\workers \workerexamples))}{T}\\
&\le \left(\frac{4F_1}{\workerexamples T}\right)^{2/3}\left(\frac{(\sqrt{42}L_{2,\infty}(\varsigma + \sigma)\tilde{A})^{2/3}}{\workers^{2/3}} + \frac{(\sqrt{18}L_{2,\infty} \workerexamples \sigma)^{2/3}}{T^{1/3}}\right)\\
&\quad + \frac{64 F_1 L_{2,\infty} (2 + \tilde{A}/(\workers \workerexamples))}{T}\\
&\le \frac{(4\sqrt{42} F_1 L_{2,\infty}(\varsigma + \sigma)\tilde{A})^{2/3}}{(\workers \workerexamples T)^{2/3}} + \frac{(72 F_1 L_{2,\infty}\sigma)^{2/3}}{T}\\
&\quad + \frac{64F_1 L_{2,\infty} (2 + \tilde{A}/(\workers \workerexamples))}{T},\\
\end{align*}
Since $(4\sqrt{42})^{2/3} < 9$, the above is

\begin{align*}
&\le \frac{9(F_1 L_{2,\infty}(\varsigma + \sigma)\tilde{A})^{2/3}}{(\workers \workerexamples T)^{2/3}} + \frac{(72 F_1 L_{2,\infty}\sigma)^{2/3} + 64F_1 L_{2,\infty} (2 + \tilde{A}/(\workers \workerexamples))}{T}, 
\end{align*}

in which the leading term (slowest in terms of $T$) is $\tilde{O}((\workers \workerexamples T)^{-2/3})$, proving the claim.
\end{proof}

\begin{proof}[Proof of Theorem \ref{thm:dgrab:PL}]
With the P.L. assumption (Assumption~\ref{ass:PL}), we use Lemma~\ref{lem:jens-pl} and Lemma~\ref{lem:Delta-2}. We show that their constraints are satisfied later) to get 
\begin{align*}
F_{T+1} &\le \rho^T F_1 + \frac{\alpha \workerexamples L_{2,\infty}^2}{2}\sum_{t=1}^T\rho^{T-t}\left(\Delta_t^2 - \frac{1}{2L_{2,\infty}^2} \norm{\nabla f(\vw_t)}_2^2\right)\\
&\le \rho^T F_1 + \frac{\alpha \workerexamples L_{2,\infty}^2}{2}\left(12\rho^{T-1}\alpha^2 \workerexamples^2 \sigma^2 + \frac{28\rho\alpha^2 (\varsigma + \sigma)^2\tilde{A}^2}{(1-\rho)\workers^2}\right)\\
&\le \rho^{T} F_1 + \rho^{T-1}6\alpha^3 \workerexamples^3 L_{2,\infty}^2\sigma^2 + \frac{28\rho\alpha^3 \workerexamples L_{2\infty}^2 (\varsigma + \sigma)^2\tilde{A}^2}{\alpha \workerexamples \mu \workers^2}\\
&\le \rho^{T} F_1 + \rho^{T} 7 \alpha^3 \workerexamples^3 L_{2,\infty}^2\sigma^2 + \frac{28\rho\alpha^3 \workerexamples L_{2\infty}^2 (\varsigma + \sigma)^2\tilde{A}^2}{\alpha \workerexamples \mu \workers^2}\\
&\le \rho^{T} (F_1 + \sigma^2/L_{2,\infty}) + \frac{28\alpha^2 L_{2,\infty}^2 (\varsigma + \sigma)^2\tilde{A}^2}{\mu \workers^2}\\
&\le (F_1 + \sigma^2/L_{2,\infty})\exp(- T\alpha \workerexamples \mu/2) + \frac{28\alpha^2 L_{2,\infty}^2 (\varsigma + \sigma)^2\tilde{A}^2}{\mu \workers^2},
\end{align*}
where we have further constrained $\alpha \le \frac{2}{9 \workerexamples \mu}$ so that $\rho \le 9/8$ in the forth inequality and  $\alpha \le \frac{1}{7^{1/3} \workerexamples L_{2,\infty}}$ in the fifth inequality. By setting the derivative w.r.t $\alpha$ of the RHS to 0, the minimizer $\alpha$ under the constraint that $0 < \alpha \le \min\left\{\frac{2}{9 \workerexamples \mu}, \frac{1}{16 L_{2,\infty}(2 \workerexamples + \tilde{A}/\workers)}\right\}$ (required by the lemmas) is:
\begin{align*}
\alpha = \frac{2}{T \workerexamples \mu} W_0(T^2 \workers^2 \workerexamples^2 C_3),
\end{align*}
as long as 
\begin{align*}
T &\ge 1 + \frac{2}{\workerexamples \mu}\max\{(9/2)\workerexamples \mu, 16 L_{2,\infty}(2\workerexamples+\tilde{A}/\workers)W_0(T^2\workers^2\workerexamples^2C_3)\}\\
&= 10 + \frac{1}{\mu}32 L_{2,\infty}(2+\tilde{A}/(\workers \workerexamples))W_0(T^2\workers^2\workerexamples^2C_3),
\end{align*}
where $C_3 = \frac{(F_1+\sigma^2/L_{2,\infty})\mu^2}{224L_{2,\infty}^2(\varsigma + \sigma)^2\tilde{A}^2}$. 

What we did here was to set $T$ just large enough so that the minimizer $\alpha$ is the same with or without the constraint. 

Denoting $\tilde{W} = W_0(T^2m^2n^2C_3) = \tilde{O}(1)$, we get 
\begin{align*}
F_{T+1} &\le  \frac{(F_1 + \sigma^2/L_{2,\infty})\tilde{W}}{T^2\workers^2\workerexamples^2C_3} + \frac{112L_{2,\infty}^2(\varsigma + \sigma)^2\tilde{A}^2\tilde{W}^2}{T^2\workers^2\workerexamples^2\mu^3}\\
&\le \frac{1}{T^2\workers^2\workerexamples^2}\left(\frac{(F_1 + \sigma^2/L_{2,\infty})\tilde{W}}{\tilde{C}_3} + \frac{112L_{2,\infty}^2(\varsigma + \sigma)^2\tilde{A}^2\tilde{W}^2}{\mu^3}\right),\\
\end{align*}
which shows rate the convergence rate in the P.L. case is $\tilde{O}((\workers \workerexamples T)^{-2})$.
\end{proof}

%% file: section/99-appendix/32-cdgrab/400-app-cdgrab-experiments.tex
\definecolor{codegreen}{rgb}{0,0.6,0}
\definecolor{codegray}{rgb}{0.5,0.5,0.5}
\definecolor{codepurple}{rgb}{0.58,0,0.82}
\definecolor{backcolour}{rgb}{0.95,0.95,0.92}

\lstdefinestyle{mystyle}{
    backgroundcolor=\color{backcolour},   
    commentstyle=\color{codegreen},
    keywordstyle=\color{magenta},
    numberstyle=\tiny\color{codegray},
    stringstyle=\color{codepurple},
    basicstyle=\ttfamily\footnotesize,
    breakatwhitespace=false,         
    breaklines=true,                 
    captionpos=b,                    
    keepspaces=true,                 
    numbers=left,                    
    numbersep=5pt,                  
    showspaces=false,                
    showstringspaces=false,
    showtabs=false,                  
    tabsize=2
}

\lstset{style=mystyle}

\section{Experiment Details}

Here we provide more extensive details on our empirical results. This includes background information on our experimental setup in the main paper (Appendix~\ref{sec:appendix-model-dataset}), an additional simulation experiment on pre-training and fine-tuning Tiny GPT-2 (Appendix~\ref{appendix-gpt2}), and an additional simulation experiment that investigates \dgrab{} with different learning rates (Appendix~\ref{appendix-lr}). 
Our source code can be found \href{https://github.com/GarlGuo/CD-GraB}{here}.

\subsection{Additional details on setup for main paper experiments}\label{sec:appendix-model-dataset}

\subsubsection{Distributed experiments}

We provide additional details on the experiments shown in Figure~\ref{fig:exp}.

\custompar{Hardware and software} 
We use a single machine with 128 GiB memory, 1 CPU, and 4 Nvidia GeForce 2080ti GPUs for the HMDA mortgage application, M4, and WikiText-2 tasks. 
We first discard the remainder $\examples \mod B$, and then randomly partition $\workerexamples$ to each worker.
Our experiments are all implemented with the PyTorch library. 

\custompar{Datasets and models}

\begin{itemize}

    \item \textbf{Logistic regression on mortgage application (NY 2017 subset)}: 
    The US Home Mortgage Disclose Act (HMDA) makes available US national data regarding mortgage applications, which has recently been packaged up for easy ML research use~\citep{cooper2024variance}. 
    We use the binary classification version of the task, which classifies features as either ``grant loan'' or ``deny loan,'' for the New York (NY) 2017 subset of the dataset, which includes 244107 examples with 18 features. 
    We model this problem using logistic regression, for which we first perform a random 80/20 train/test split on the raw dataset, and then we discard $\examples \mod B$ ($B$ is the aggregated minibatch size) examples to ensure that each worker receives exactly $\workerexamples$ examples. 
    We use 1 worker per GPU, and in total we have $\workers=4$ workers, and use NCCL~\citep{nccl} as the distributed communication backend; $\workers=4, \; \workerexamples=48816, \; d=18, \; B = 16$. 
    We report test accuracy as our evaluation metric. 
    
    \item \textbf{LSTM on WikiText-2}: We follow the settings in Lu et al.~\cite{lu2022grab} and train a 2-layer LSTM with an embedding size of 32 and dropout set to 0. 
    We use backpropagation through time, for which we set the sequence length to 35. We also adopt the word-vector-classifier-weight-sharing strategy inspired by Inan et al.~\cite{inan2016tying}. 
    WikiText-2 \citep{stephen2017pointer} has 600 articles in the train set, with more than 2M tokens and 30K vocabulary; 
    the validation and test sets each have 60 articles. We adapt our training script from \href{https://github.com/pytorch/examples/tree/main/word_language_model}{PyTorch's official Word Language Modeling Github repository}. 
    We use 4 workers in total, with each GPU hosting 1 worker, and use NCCL as the distributed communication backend; $\workers = 4, \; \workerexamples= 3728, \; d = 1081760, \; B = 16$.  
    We report test perplexity as the evaluation metric, and we follow the HuggingFace's approach of computing perplexity as the exponentiated average negative log-likelihood of a sequence.\footnote{\url{https://huggingface.co/docs/transformers/perplexity}}
    
    \item \textbf{Autoregressive MLP on M4 Weekly Dataset}: 
    We build a 3-layer autoregressive MLP with a hidden dimension of 64. We set input sequence length to be 20 and the output sequence length to be 6. 
    M4 is a time series dataset composed of 100,000 time series for yearly, quarterly, monthly, weekly, daily and hourly data~\citep{MAKRIDAKIS202054}, which is drawn from a random sample of ForeDeCk database \citep{spiliotis2020forecasting}. 
    We use the weekly data in our experiment. We use 32 workers, where each of the 4 GPUs hosts 8 process workers. We use GLOO as the distributed communication backend. $\workers = 32, \; \workerexamples = 3355, \; d = 5569, \; B = 32$.  
    We report test symmetric mean absolute percentage error (SMAPE) as the evaluation metric. We follow the formula of SMAPE in \citep{MAKRIDAKIS202054} as follows: 
\begin{equation*}
    \text{SMAPE} \triangleq \dfrac{2}{h} \sum_{t = n + 1}^{n + h} \dfrac{|Y_t - \hat{Y}_t|}{|Y_t| + |\hat{Y}_t|} * 100\%
\end{equation*}
where $Y_t$ is the reference time series value at timestep $t$, $\hat{Y}_t$ is the forecast time series value at timestep $t$, and $h$ is the forecasting horizon and $n$ is the number of datapoints. 

\end{itemize}

\custompar{Hyperparameter optimization} 
For all tasks, we tune the learning rate $\alpha$ for \dshuffle{} first, and then use the selected learning rate for \dgrab{}. 
Therefore, a performance improvement here implies we would have in-place substitution benefits via switching from \dshuffle{} to \dgrab{} with identical learning rate and experiment setups. We use SGD with momentum as the optimizer for all tasks. 
The hyperparameters for each task are as follows:

\begin{itemize}
    \item \textbf{Logistic regression on mortgage application (NY 2017 subset)}: $\alpha = $ 5e-3 $\in \{$1e-2, 5e-3, 1e-3$\}$, momentum: 0.9, weight decay: 0, $B$: 16.

    \item \textbf{LSTM on WikiText-2}: $\alpha = 5 \in \{$5, 10$\}$ and decays by 0.1 per 10 epochs, momentum: 0.9, weight decay: 0, $B$: 16. 

    \item \textbf{Autoregressive MLP on Weekly M4 Dataset} $\alpha = $ 1e-3 $\in \{$1e-2, 1e-3, 1e-4$\}$, momentum: 0.9, weight decay: 0, $B$: 32.
\end{itemize}

\subsubsection{Memory Overhead of \dgrab{} in LSTM on WikiText-2 Task} \label{sec:appendix-memory}

\begin{figure}[!ht]
\vspace{-.2cm}
  \centering
    \includegraphics[width=\columnwidth]{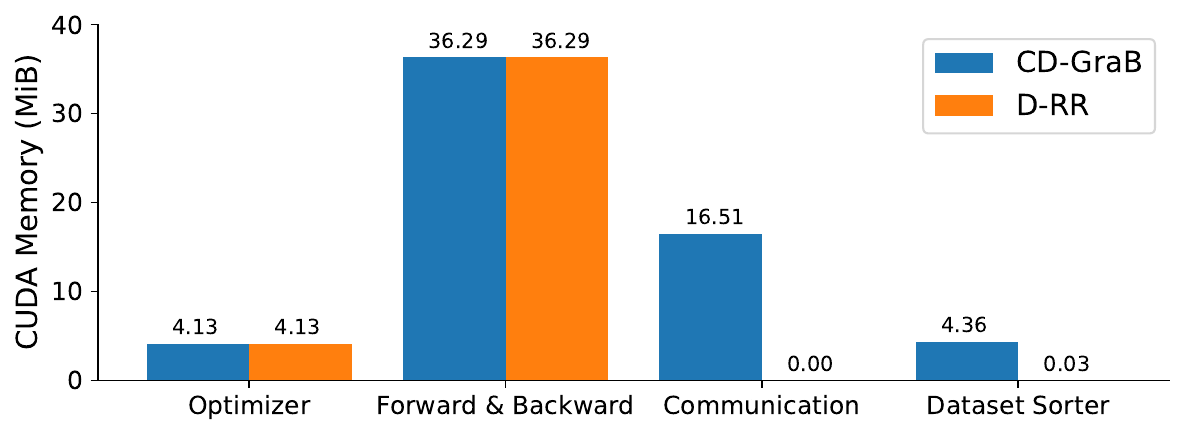}
    \caption{CUDA Memory Overhead of \dgrab{} and \dshuffle{} in LSTM on WikiText-2 Task. 
    \looseness=-1} 
  \label{fig:cuda-memory-lstm}
\end{figure}

We profile the CUDA memory usage for the LSTM on WikiText-2 Task with \dgrab{} and \dshuffle{} to understand the memory overhead of both data permutation algorithms. 
This memory analysis is both task and implementation dependent, but still serves to illustrate the overarching point that \dgrab's memory overhead is not so significant. 
The additional overhead comes from two sources for \dgrab: communication and example sorting (Figure~\ref{fig:cuda-memory-lstm}). 

In more detail: 
in our LSTM experiment, each local worker will share its gradients with all other workers at every optimization step. 
To reduce the communication burden of \dgrab, we make each local worker function as an order server.\footnote{An ideal location for a dedicated order server is on a network node that has large input bandwidth and memory buffer to host all gradients while not blocking the normal optimization stages. 
    Since we do not have enough computational resources to host a dedicated order server, we make each worker an order server.} 
The memory consumption of forward, backward, and optimizer states between \dgrab{} and \dshuffle{} should be (at least approximately) identical. 
The model size of LSTM is roughly 4 MiB. 
We use 4 workers, and as each worker (functioning as an order server) needs to \textit{all-gather} gradients, the memory overhead for \textit{all-gather} communication is roughly tensor\_size $\times$ \# workers = 4 MiB $\times$ 4 = 16 MiB for \dgrab (we observe 16.51 MiB in practice, Communication in Figure~\ref{fig:cuda-memory-lstm}), while \dshuffle{} only needs to \textit{all-reduce} the gradients (yielding no memory overhead; the communication buffer for \textit{all-reduce} is reusing the same gradient tensor). 
The $\mathsf{PairBalance}$ algorithm (Algorithm~\ref{alg:pairbalance}) internally needs a model-sized accumulator as the running sum $\vr$, and both computing inner product between $\vr$ and $\vg_1 - \vg_2$ and updating $\vr$ with $\vr + s \vc$ takes virtually no space with a memory-efficient implementation. 
Therefore, the memory consumption for $\mathsf{PairBalance}$ is still roughly 4 MiB (Data Sorter in Figure~\ref{fig:cuda-memory-lstm}). 

The main memory overhead of \dgrab{} will be dominated by the communication buffer size on the order server side: 
the order server have to gather the gradient (differences) from all workers, and sequentially apply the $\mathsf{PairBalance}$ algorithm. 
This memory bottleneck would similarly be found in \grab{} as \grab{} also needs per-example gradients to perform $\mathsf{Balance}$ sequentially. 

A future algorithmic improvement to the general gradient balancing framework would be finding a balancing algorithm that does not need per-example gradients to achieve comparable convergence guarantees. 
However, we still notice that $\mathsf{PairBalance}$ is more memory-efficient than $\mathsf{Balance}$ as $\mathsf{Balance}$ needs to store 3 model-sized tensors: 
1 for the balancing accumulator, 1 for running-average gradients for last epoch, and 1 for the running-average for current epoch. 
In contrast, $\mathsf{PairBalance}$ only needs 1 model-sized tensor as the balancing accumulator. 

\subsubsection{Simulated ablation study using LeNet on CIFAR-10 \label{appendix:workers}}

In the experiment shown on Figure~\ref{fig:nodes}, we select the same learning rate, momentum, and weight decay as the LeNet experiment in Lu et al.~\cite{lu2022grab}. 
We use 3 different random seeds to control 3 different initialization and the randomness in random reshuffling. 
The aggregated minibatch size $B$ is 64 for all runs. 
We implement this ablation study by using 1 GPU with up to $m=64$ workers (processes). 
As above, we discard $\examples \mod B$ examples and partition the remaining examples evenly on each worker. 

$\alpha = $ 1e-3 $\in \{$1e-2, 5e-3, 1e-3, 5e-4, 1e-4$\}$, momentum: 0.9, weight decay: 1e-2, $B$: 64. 

We do not implement this via distributed environment due to the fact that we do not have access to 64 GPUs, but expect the simulation results to be a good reflection of the results we would obtain in a multi-GPU setting. 

\begin{figure}[!t]
  \centering
    \includegraphics[width=\columnwidth]{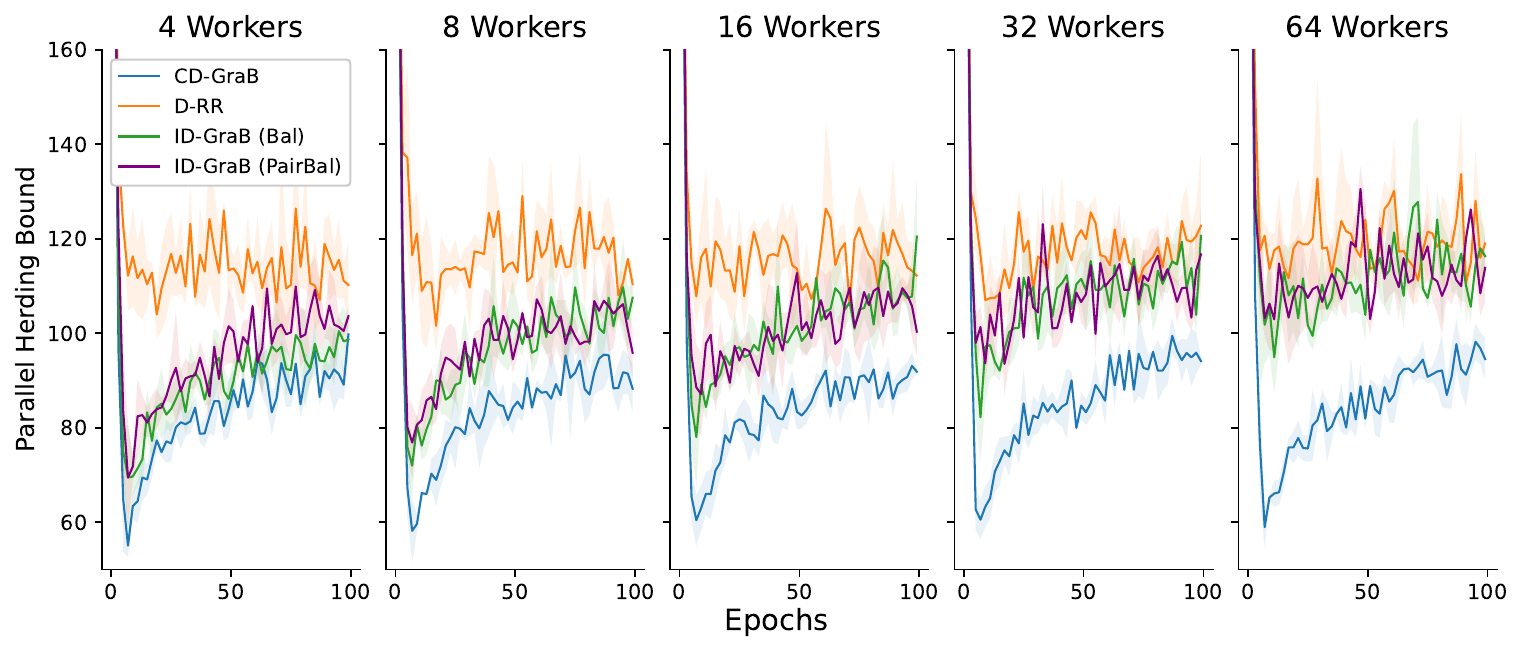}
    \vspace{-.5cm}
    \caption{Empirical parallel herding bounds of gradients for each algorithm in LeNet experiment. 
    We plot the mean as the curve and standard deviation across 3 random seeds. 
    \looseness=-1} 
  \label{fig:parallel-herding-bound-lenet}
\end{figure}

\custompar{Parallel herding bound} 
We further investigate the empirical parallel herding bounds (\ref{equ:paraherding:objective}) for the LeNet experiment for the different ordering methods. 
We plot the results in Figure~\ref{fig:parallel-herding-bound-lenet}. 
We observe that as the number of workers increases, the empirical parallel herding bounds of both \textbf{ID-\grab{} (Bal)} and \textbf{ID-\grab{} (PairBal)} also increase, and eventually exhibit little difference with \dshuffle{}. \dgrab, in contrast, exhibits a consistently lower bound. 

\begin{figure}[!t]
  \centering
    \includegraphics[width=\columnwidth]{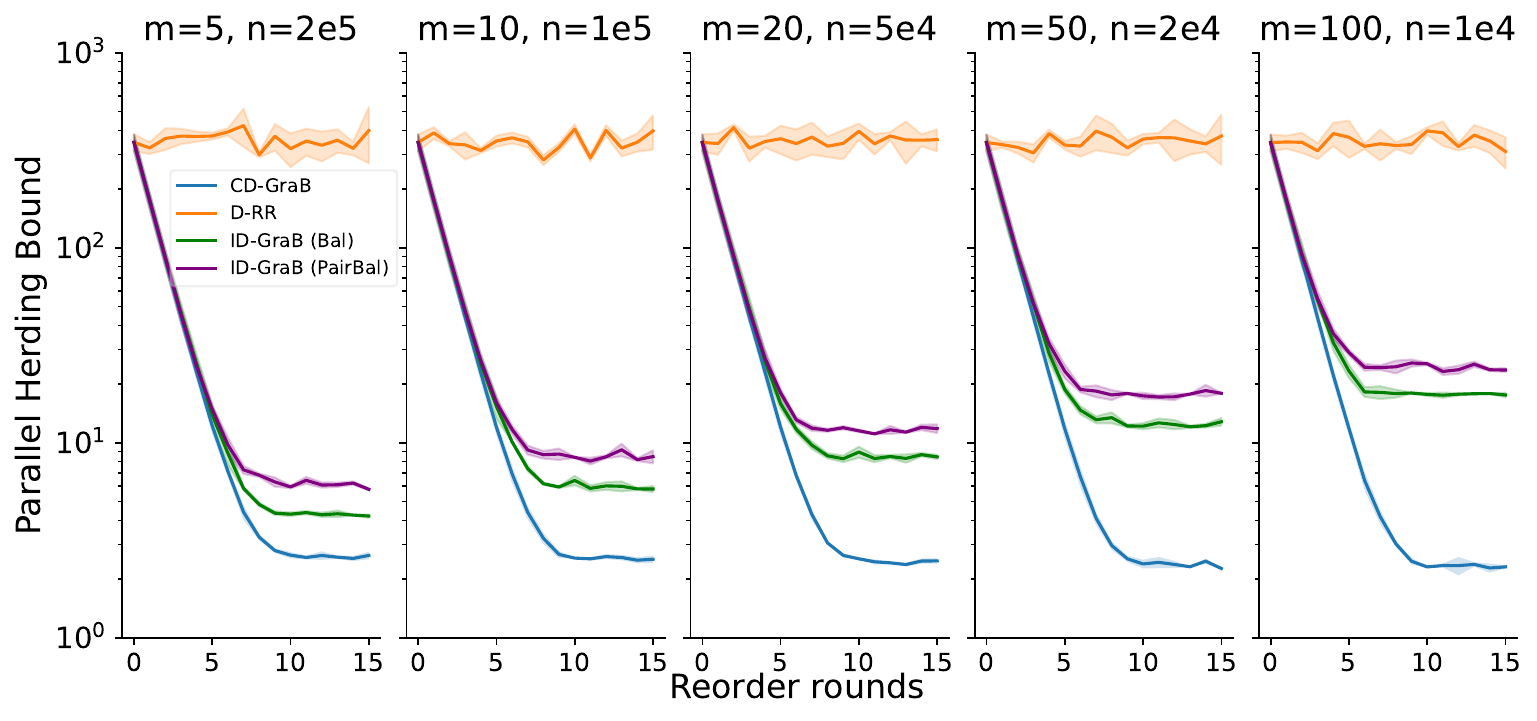}
    \caption{Parallel herding bounds for different example ordering algorithms on $\examples$=1 million random vectors.
    We use 3 random seeds, plot the mean and standard deviation across each random seed as the shaded area.
    \looseness=-1} 
  \label{fig:parallel-herding-bound-sim}
\end{figure}

For comparison, we also run a simulation experiment on synthetic data to investigate the behavior of the parallel herding bound. 
We include these below, in Figure~\ref{fig:parallel-herding-bound-sim}. 

We randomly initialize 1 million random vectors $\exij$ from a uniform distribution between 0 and 1 with 16 dimensions as $\exij \sim \mathsf{Unif}(0, 1)^{16}$, and then we zero-center this set of 1 million vectors and normalize them to all have $L_2$ norm as 1. 
We then evenly partition this set of 1 million random vectors to $\{5, 10, 20, 50, 100 \}$ workers and run each example ordering algorithm. 

In Figure~\ref{fig:parallel-herding-bound-sim}, we run \dgrab, \dshuffle, \textbf{ID-\grab{} (Bal)}, \textbf{ID-\grab{} (PairBal)} on these random vectors, and compute the parallel herding bounds (\ref{equ:paraherding:objective}). 
From left to right in Figure~\ref{fig:parallel-herding-bound-sim}, we observe that as the number of workers $\workers$ increases, the parallel herding bound of \textbf{ID-\grab{} (Bal)}, \textbf{ID-\grab{} (PairBal)} becomes larger. 
This shows the importance of coordination when we have a large number of workers.

These results for random vectors cohere with our above results for LeNet on CIFAR-10.

\subsection{An additional simulation experiment: pre-training and fine-tuning Tiny GPT-2}\label{appendix-gpt2}

We perform an end-to-end simulation experiment involving pre-training and fine-tuning Tiny GPT-2 on WikiText-103, which we document below. 

\subsubsection{Pre-training}

We adapt the training script from the \href{https://github.com/huggingface/transformers/blob/main/examples/pytorch/language-modeling/run_clm_no_trainer.py}{HuggingFace's PyTorch casual language modeling code} to train the GPT-2 architecture~\citep{radford2019language}. 
We set the maximum sequence length to 128 and token and positional embedding dimension to 128; use 2 hidden layers in the transformer encoder and 2 attention heads; and disable dropout. 
This model configuration corresponds to the following Python code snippet:

\begin{lstlisting}[language=Python]
from transformers import GPT2Config, GPT2LMHeadModel, GPT2Tokenizer

tokenizer = GPT2Tokenizer.from_pretrained('gpt2')
config = GPT2Config.from_pretrained('gpt2')
config.n_embd = 128
config.n_ctx = 128
config.n_layer = 2
config.n_head = 2
config.n_positions = 128
config.summary_first_dropout = 0
config.attn_pdrop = 0
config.resid_pdrop = 0
model = GPT2LMHeadModel(config)
\end{lstlisting}

We train our Tiny GPT-2 model from scratch on WikiText-103~\citep{stephen2017pointer}.
WikiText-103 is a standard language modeling benchmark that has 28,475 articles in the train set, and 60 for both the validation and test sets, with more than 100M tokens and 267K vocabulary inside the train set. 
We use the original GPT-2 tokenizer, and use maximum sequence length 128. 
We note that this is much smaller than the default maximum sequence length for GPT-2, which is 1024, which was too large to use given our computational budget. 
Nevertheless, 128 is still a reasonable sequence length for the initial phrase of pre-training; 
BERT uses a sequence length of 128 for the first 90\% of pre-training steps to speedup the experiment~\cite{devlin2019bert}. 
We tune the learning rate for \dshuffle{} with the grid $\{$ 5e-3, 1e-3, 5e-4, 1e-4 $\}$ (the final learning rate is 5e-4), and use AdamW optimizer~\cite{loshchilov2017decoupled}. 
We use 3 random seeds. Before the training, we simulate 64 workers, and similarly divide the training dataset evenly across them by discarding $\examples \mod B$ examples. 
Our hyperparameter optimization space is listed below:

\custompar{Pretraining Hyperparameters} $\alpha = $5e-4 $\in \{$5e-3, 1e-3, 5e-4, 1e-4$\}$, weight decay: 1e-4, $B$: 64.

\begin{figure}[!t]
  \centering
    \includegraphics[width=\columnwidth]{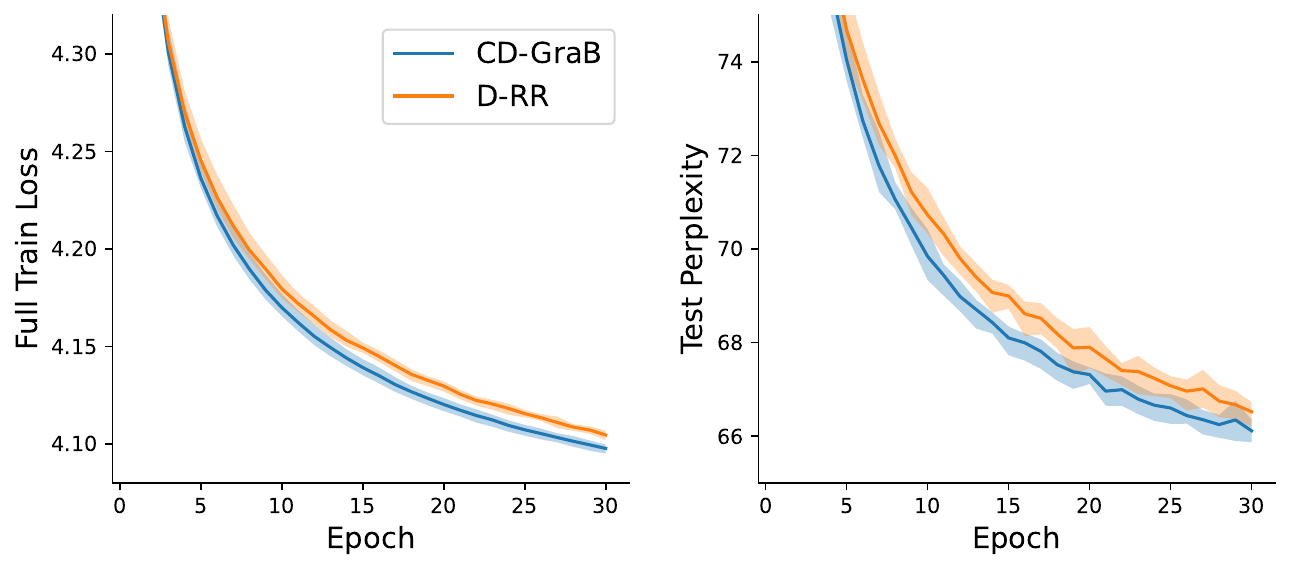}
    \vspace{-.5cm}
    \caption{Pre-training Tiny GPT-2 on WikiText-103 from scratch: Convergence for \dgrab{} and \dshuffle{} with $\workers = 64$ workers. 
    The aggregated minibatch size per update is 64. 
    We use 3 random seeds, and plot the mean and standard deviation. 
    \looseness=-1} 
  \label{fig:gpt2}
\end{figure}

We document convergence for pre-training in Figure~\ref{fig:gpt2}, and use test perplexity as our evaluation metric.

\subsubsection{Fine-tuning}

We then fine-tune the pre-trained Tiny GPT-2 model on downstream tasks. 
For each task, we load the pre-trained foundation model weights obtained at the end of 30 epochs of each example ordering algorithm after pretraining, and use the same example ordering algorithm to perform supervised fine-tuning. 
We focus on the largest 4 GLUE tasks~\cite{wang-etal-2018-glue}: MNLI, QQP, QNLI, and SST2. 
We tune the learning rate for \dshuffle{} with the AdamW optimizer, and for each run we report the best validation accuracy. We then take an average results of each run and summarize them in Table~\ref{tab:gpt2-glue}. 
Our training script is adapted from the \href{https://github.com/huggingface/transformers/blob/main/examples/pytorch/text-classification/run_glue_no_trainer.py}{HuggingFace's PyTorch GLUE fine-tuning example codes}.

\textbf{Fine-Tuning Hyperparameters}
\begin{itemize}
    \item \textbf{MNLI} $\alpha = $5e-4$ \in \{$5e-3, 1e-3, 5e-4, 1e-4$\}$, Weight decay: 1e-4, $B$: 32, epochs: 10, linear learning rate scheduler

     \item \textbf{QQP} $\alpha = $5e-4$ \in \{$5e-3, 1e-3, 5e-4, 1e-4$\}$, Weight decay: 1e-4, $B$: 32, epochs: 10, linear learning rate scheduler

     \item \textbf{QNLI} $\alpha = $5e-4$ \in \{$5e-3, 1e-3, 5e-4, 1e-4$\}$, Weight decay: 1e-4, $B$: 32, epochs: 10, linear learning rate scheduler

     \item \textbf{SST2} $\alpha = $5e-4$ \in \{$5e-3, 1e-3, 5e-4, 1e-4$\}$, Weight decay: 1e-4, $B$: 32, epochs: 10, linear learning rate scheduler
\end{itemize}

\begin{table*}[!ht] 
 \centering 
 \setlength{\tabcolsep}{2pt}
 \footnotesize

 \begin{tabular}{c|ccccc}
    \toprule
    
    & \textbf{MNLI (Matched)} & \textbf{MNLI (Mismatched)}  & \textbf{QQP} & \textbf{QNLI} & \textbf{SST2}  
    \\
    \midrule
    
    \textbf{\dgrab}
    & 65.91 $\pm$ 0.46 \%
    & 64.36 $\pm$ 2.03 \%
    & 82.25 $\pm$ 0.21 \%
    & 62.11 $\pm$ 0.70 \%
    & 82.65 $\pm$ 0.39 \%
    \\

    \textbf{\dshuffle} 
    & 65.42 $\pm$ 0.36 \%
    & 63.93 $\pm$ 1.63 \%
    & 81.74 $\pm$ 0.33 \%
    & 61.87 $\pm$ 0.67 \%    
    & 82.68 $\pm$ 0.57 \%
    \\

    \bottomrule
    \end{tabular}

    \caption{GLUE fine-tuning datasets: Validation accuracy of \dgrab{} in comparison to  \dshuffle{}, reporting mean and standard deviation of best results for each run. 
    There are 3 runs for each example ordering algorithm.}
    \label{tab:gpt2-glue}
\end{table*}

We include these fine-tuning results in part to support our claim in Section~\ref{sec:cdgrab:conclusion} that \dgrab{} exhibits its benefits more clearly when there are more training epochs. 
Our pre-training results suggest that \dgrab{} would confer benefits to pre-training large models over multiple epochs; 
however, \dgrab{} will not necessarily be useful for short runs of fine-tuning (as indicated in Table~\ref{tab:gpt2-glue}, for which the results for both ordering algorithms are effectively identical). 

\subsection{Ablation simulation study: The impact of learning rate $\alpha$}\label{appendix-lr}

\begin{figure}[!t]
  \centering
    \includegraphics[width=\columnwidth]{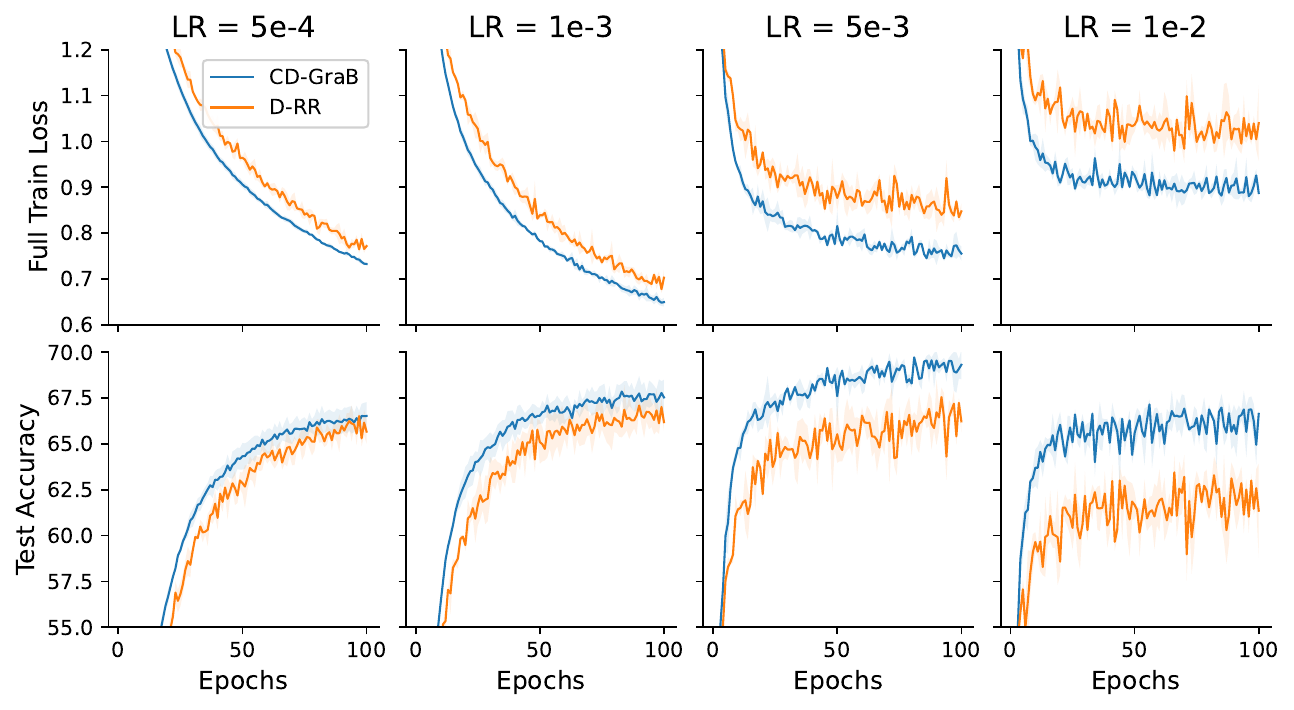}
    \vspace{-.5cm}
    \caption{Convergence for \dgrab, \dshuffle{}  training LeNet on CIFAR-10, with $\workers = 64$ workers. The aggregated minibatch size per update is 64. We use 3 random seeds, and plot the mean values across random seeds as the curve, the standard deviation as the shaded area.\looseness=-1} 
  \label{fig:lenet-lr}
\end{figure}

In the experiment shown on Figure~\ref{fig:lenet-lr}, we select the same momentum and weight decay as the LeNet experiment for 3 random seeds as in Appendix~\ref{appendix:workers}. 
The aggregated minibatch size is still 64 for all runs, and we use 64 workers.

We find that when we increase the learning rate from 1e-3 to 1e-2, \dgrab{} still maintains relatively better performance than \dshuffle. 
The best learning rate for \dshuffle{} is 1e-3, in terms of achieving the best test accuracy.
We did not tune the learning rate for \dgrab, and we expect that it is possible to use a higher learning rate and still maintain better empirical performance than \dshuffle{} and even faster convergence. 
We defer such empirical investigations to future work. 
Altogether, these preliminary empirical results confirm that it is possible to use higher learning rate for \dgrab{}, given that online $\mathsf{PairBalance}$ does not need to use a stale mean (Section~\ref{sec:cdgrab:dgrab:solution}), which would make larger learning rates perform poorly.

%% file: section/99-appendix/42-commoncanvas/00-app-commoncanvas.tex
\chapter{Appendix for \modelname}\label{chapter:app:cc}

\input{section/99-appendix/42-commoncanvas/10-app-commoncanvas-method}
\input{section/99-appendix/42-commoncanvas/20-app-commoncanvas-datasets}
\input{section/99-appendix/42-commoncanvas/30-app-commoncanvas-telephone}
\input{section/99-appendix/42-commoncanvas/40-app-commoncanvas-sys}
\input{section/99-appendix/42-commoncanvas/50-app-commoncanvas-figs}

%% file: section/99-appendix/42-commoncanvas/10-app-commoncanvas-method.tex
\section{Details on Data Scarcity Analysis}\label{app:sec:cc:data-scarce} 

\subsection{Hypothesis: Diffusion models are too small}
\label{app:sec:cc:diff-too-small}

A back-of-the-envelope calculation provides some insight on why this is the case. Consider a training dataset consisting of $N$ images with resolution $H\times W$ and $c$ channels. To completely memorize the training data, the model must be capable of storing $c\times H \times W\times N$ numbers. Given a number of trainable parameters $N_p$, it is natural to assume that on average each parameter is capable of storing roughly enough information to reconstruct a single number from the training dataset. Under this assumption, complete memorization is only possible if the size of the training dataset is at or below a critical size $N_c$ ($N\leq N_c$) with $N_c$ given by $N_c=\frac{N_p}{cHW}$. Note that this critical size assumes the data cannot be further compressed, which is obviously not the case for natural images. However, SD2 and SDXL are latent diffusion models, which first use a pretrained encoder to compress images by a factor of $8$ in both $H$ and $W$, and so when we train LDMS like SD2 and SDXL, we are training on data that has been significantly compressed already. 

In our experiments, $c=4$ and $H=W=32$, corresponding to $256\times256$ resolution RGB images in the SD2 and SDXL latent space. The SD2 UNet has $N_p=866\times10^6$ trainable parameters, and SDXL's UNet has $N_p=2567\times10^6$. So we calculate $N_c\approx0.2\times10^6$ for SD2 and $N_c\approx0.6\times10^6$ for CommonCanvas-Large; both of these numbers are several orders of magnitude below the size of our YFCC derived datasets, and so even with significant additional data compression we expect that our \datasetname{} datasets should be sufficient to train both SD2 and SDXL. Additionally, this argument predicts that we should only begin to see significant overfitting in these models for datasets of size $N\sim10^6$. These estimates are resolution dependent, and as image resolution increases we expect that $N_c$ will decrease as more information is provided per image. 

\subsection{Increasing model capacity}
\label{app:sec:cc:largeboi}
We also train a variant of SD2 with more trainable parameters, taking the UNet from SDXL. We refer to this model as CommonCanvas-LNC. We adapt the SDXL UNet architecture to SD2 by changing the cross-attention dimensionality to match that of the SD2 text encoder hidden state dimensionality (1024 for SD2 vs. 2048 for SDXL). SDXL also retrains the VAE component in their model, and we use this improved performance VAE as well. Except for these changes, the architecture is identical to that of SD2.

%% file: section/99-appendix/42-commoncanvas/20-app-commoncanvas-datasets.tex
\section{Training Dataset and Model Details}\label{app:sec:cc:data}

\subsection{LAION-2B}\label{app:sec:cc:laion}

The fact that LAION is not a stable benchmark can lead to multiple reproducability and security issues. Data poisoning attacks would be difficult to detect at the scale of 2 billion parameters. While this could be mitigated by using hash values of the images, then any time the a site decide to re-encode the image, those images would now need to be excluded from the dataset. Furthermore, targeted data poisoning attacks for diffusion models are no longer just academic conjecture. Last year after the release of Stable Diffusion, a protest was launched on ArtStation that had uses upload images that said ``NoAI'' to taint future training data for generative models after artists felt as though their work had been unfairly used to train the models. With the high degree of link rot, targeted attacks are fairly easy. 
Furthermore, reproduction of the experiments becomes virtually impossible. This means any benchmarks that use copies of LAION as ground truth are are likely using differing subsets of the full dataset.

\begin{table}[t!]
\centering
    \caption{CC licenses in YFCC100M.}\vspace{.1cm}
    \label{tab:yfcc100m-count-table}
    \footnotesize
    \begin{tabular}{lrr}
    \toprule
    \textbf{CC License} & \textbf{\# Images} & \textbf{\% Captioned} \\\midrule
\rowcolor{palepink}
CC-BY-NC-ND-2.0 
& 25,790,117 & 33.52\%  \\\midrule
\rowcolor{palepink}
CC-BY-ND-2.0 
&  4,827,970 & 30.23\% \\\midrule
\rowcolor{paleblue}
CC-BY-NC-2.0 & 12,468,229 & 31.39\% \\\midrule
\rowcolor{paleblue}
CC-BY-NC-SA-2.0 & 28,314,685 & 31.57\%  \\\midrule
\rowcolor{paleyellow}
CC-BY-SA 2.0
& 9,270,079 &  34.05\% \\\midrule
\rowcolor{paleyellow}
CC-BY 2.0 
& 16,962,338 & 28.96\% \\\bottomrule
    \end{tabular}
\end{table}

\begin{table}[t]
    \centering
    \setlength{\itemwidth}{0.3\linewidth}
    \caption{Randomly sampled images from the YFCC~\citep{thomee2016yfcc100m} training set. Our synthetic BLIP2 captions are also provided below. }
    \newcolumntype{C}[1]{>{\centering\arraybackslash}p{#1}}
    
    \begin{tabular}{C{\itemwidth} C{\itemwidth} C{\itemwidth}}
 \makebox[\itemwidth]{%
                     \includegraphics[width=\itemwidth]{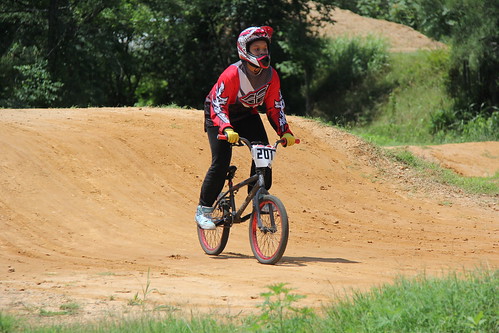}
                    }
         & \includegraphics[width=\itemwidth]{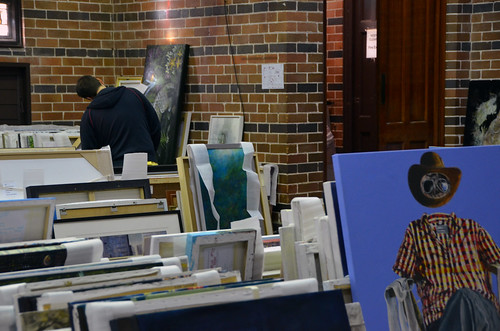}& \includegraphics[width=\itemwidth]{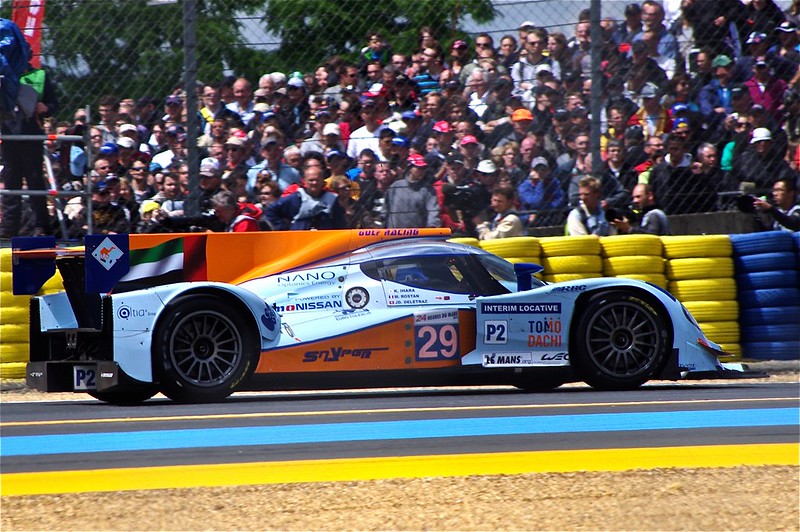} \\ 
         a person riding a bike on a dirt road & a paintings on the wall & an orange and blue race car driving on a track
    \end{tabular}
    
    \label{tab:yfcc-examples}
\end{table}

\subsubsection{Sourcing Creative-Commons images}\label{app:sec:cc:catalog:images}

We source these images from YFCC100M.
ND means derivative works are not licensed or the license doesn't allow the user to create derivative works. NC means images cannot be used in commercial contexts. \datasetname-C only contains data from the bottom two (yellow) rows, reflecting images licensed for commercial contexts (i.e., roughly 25 million images). \datasetname-NC contains \datasetname-C, and additionally includes the middle two (blue) rows, reflecting images licensed for non-commercial purposes. We do not include the roughly 30 million images in the top two (pink) rows in \datasetname, as they are non-derivative licenses. We do not train on these images. We do, however, produce BLIP-2 captions for them and release those captions as an evaluation set.

\begin{table*}[t]
\caption{Top 10 highest frequency captions in the YFCC dataset. The most common captions are not user generated and are not very descriptive of the corresponding image.
\vspace{.2cm}
}
\centering
\begin{tabular}{|p{0.9\linewidth}|c|}
\hline
\textbf{YFCC Original Caption} & \textbf{Count} \\
\hline
OLYMPUS+DIGITAL+CAMERA & 184889 \\
SONY+DSC & 123128 \\
Exif\_JPEG\_PICTURE & 104480 \\
Barclays+Center+Arena\%0AAtlantic+Yards\%0A6th+and+Atlantic+A
& 68832 \\
Olympus+digital+camera & 54805 \\
Effortlessly+uploaded+by \href{http://www.eye.fi}{Eye-Fi} & 48388 \\
. & 43227 \\
-+Camera+phone+upload+powered+by \href{http://www.shozu.com/?utm\_source=upload&utm\_medium=graphic&utm\_campaign=upload\_graphic}{ShoZu} & 38856 \\
Sony+dsc & 32709 \\
Photo+by \href{http://twitter.com/Kmeron}{@Kmeron} \href{http://www.facebook.com/musicfromthepit}{|Facebook page is this way|} & 23754 \\
\hline
\end{tabular}

\end{table*}

\begin{table}[t]
    \centering
        \caption{Number of usable captions from OpenAI's YFCC14M dataset~\citep{radford2021clip}. This table is actually a subset from~\ref{tab:yfcc100m-count-table} for which either the user description or image title were deemed usable. These figures provide an estimate on how many images in each category are actually potentially usable as captions.}

    \begin{tabular}{lr}
    \hline
                                  License Name &    count \\
    \hline
        CC-BY 2.0 &  2448002 \\
        CC-BY-ND 2.0 &   682273 \\
        CC-BY-NC  2.0 &  1925854 \\
        CC-BY-NC-ND  2.0 &  4058817 \\
        CC-BY-NC-SA  2.0 &  4146113 \\
       CC-BY-SA 2.0 &  1568336 \\
    \hline
\end{tabular}

    \label{tab:openai-usable}
\end{table}

\subsection{Model Architecture}

We follow the model architecture and training recipe of Stable Diffusion 2 as closely as we can to best reproduce the model for CC-Small. The model has an identical number of params and structure as the original model. In fact, we can even load SD2's model weights into our framework due to the identical architecture and naming scheme. We are able to achieve virtually identical performance with SD2 in a much shorter training time with less data. We use the same VAE, tokenizers, and UNet archicture as SD2 except for reducing the precision of the normalization layers.

Our CC-Large model takes SD2's model and replaces the UNet with the SDXL architecture~\citep{podell2023sdxl}. Like CC-Small, we also replace the normalization layers with their low-precision version. The replacement of all the normalization layers is handled automatically by MosaicML's Composer library~\citep{mosaicml2022composer}. We perform all dataloading through MosaicML's streaming library~\citep{mosaicml2022streaming}.

\subsection{Release and documentation} 
We release \datasetname{} and information about how to download \modelname{} at \url{https://github.com/mosaicml/diffusion/blob/main/assets/common-canvas.md}, with an associated data sheet. 

%% file: section/99-appendix/42-commoncanvas/30-app-commoncanvas-telephone.tex
\section{\capcaptionmethod}

We dub our solution for handling the lack of captions in CC images as \emph{\captionmethod{}}, a type of transfer learning (Figure~\ref{fig:telephoning}). 
\capcaptionmethod{} assumes the existence of a large labeled dataset $\mathcal{D}_1 = \{(x^{(i)}, y^{(i)})\}_{i=1}^n$, consisting of pairs of high-dimensional $x^{(i)}$ (e.g., images, audio) that map to a compact, structured label $y^{(i)}$ (e.g., caption, audio transcript). \capcaptionmethod{} trains a forward model $q(y | x)$ on $\mathcal{D}_1$ to learn the mapping of $y$ given $x$ via maximum likelihood learning $\max_{q \in \mathcal{Q}} \sum_{i=1}^n \log q(y^{(i)} | x^{(i)})$. It then uses $q$ as training signal for a reverse model $p(x|y)$ trained on a {\em separate} dataset $\mathcal{D}_2 = \{x^{(i)}\}_{i=1}^m$ by maximizing $\sum_{i=1}^m \mathbb{E}_{y \sim q(y|x^{(i)})} [\log p(x^{(i)} | y^{(i)})]$, the likelihood of the data $\mathcal{D}_2$ and the predicted label $y$ under $q$. This forms a type of knowledge transfer from the forward labeling task defined by $\mathcal{D}_1$ to the reverse task of inverting $x$ from $y$ on a separate $\mathcal{D}_2$.

While \captionmethod{} can be viewed as a type of synthetic labeling, it becomes particularly interesting when $x$ is a type of protected modality (e.g., a copyrighted image), while $y$ is a compact representation of $x$ that does not encode sensitive aspects of $y$ (e.g., a generic caption).
Effectively, \captionmethod{} performs a type of ``lossy compression'' or ``distillation" from a high-dimensional  or information-rich $x$ (e.g., an image of Snoopy) to a low-dimensional or information-poor $y$ that loses the sensitive content in $x$ (e.g., the visual characteristics of Snoopy). 
Because this compression step is ``lossy'', a reconstruction $x'$ of $x$ from $p(x|y)$ via $y$ often does not remotely resemble the original input, just like in a game of telephone~\citep{telephone}.  
We derive the term \captionmethod{} from the above intuition, and employ it as useful shorthand to denote instances of transfer learning that solve data-scarcity problems in multimodal generative modeling.  

\paragraph{\capcaptionmethod{} for text-to-image modeling.}
In this work, we apply \captionmethod{} to the image and text domains, where CC images are the high-dimensional inputs $x$, and we use a pre-trained BLIP-2 model~\cite{li2023blip2} for ``lossy compression'' to short-text captions $y$ (Figure~\ref{fig:telephoning}a). 
Together, these CC-image-caption pairs comprise  the \datasetname{} dataset, which we use to train our \modelname{} T2I models (Figure~\ref{fig:telephoning}b). 
Even though BLIP-2 was pre-trained on LAION-400M~\citep{laion400}, \datasetname{} and \modelname{} never have direct access to LAION-400M or, importantly, anything that is similar to the images that BLIP-2 was trained on. 
Instead, we only have access to the mapping in the model, which, given an image input, produces lossy output text that inherently does not literally resemble its image counterpart (Figure~\ref{fig:telephoning}c).

We draw on the example of Snoopy from~\cite{sag2023safety}. Figure~\ref{fig:telephoning}'s Snoopy is CC-licensed~\citep{snoopypic}.

%% file: section/99-appendix/42-commoncanvas/40-app-commoncanvas-sys.tex
\section{Details on Efficiency Optimizations}\label{app:sec:cc:mlsys}

In this section we provide additional details on the optimizations we implemented to achieve SD2 training speedups. We also report the approximate cost of training our implementation of SD2 on various hardware configurations in Table~\ref{table:speeduptable}.

\begin{table*}[t]
\centering
\caption{Performance (throughput) and approximate cost of training SD2 UNet with our optimizations. Depending on the number of GPUs used, the cost to train the same models without these optimizations range from \$90,000-\$140,000}
\label{table:speeduptable}
\resizebox{\textwidth}{!}{%
\small 
\begin{tabular}{|c|c|c|c|c|c|}
\hline
\textbf{Number of A100s} & \textbf{256x256 (img/s)} & \textbf{512x512 (img/s)} & \textbf{512x512 with EMA (img/s)} & \textbf{Days to Train} & \textbf{Cost (\$)} \\
\hline

8  & 1100 & 290  & 290  & 101.04 & \$38,800.00 \\
\hline

16 & 2180 & 585  & 580  & 50.29  & \$38,630.00 \\
\hline

32 & 4080 & 1195 & 1160 & 25.01  & \$38,420.00 \\
\hline

64 & 8530 & 2340 & 2220 & 12.63  & \$38,800.00 \\
\hline

128& 11600& 4590 & 3927 & 6.79  & \$41,710.00 \\
\hline
\end{tabular}%
}
\end{table*}

\paragraph{Flash Attention.} Cross attention operations are a very expensive part of training that occurs in dozens of layers in diffusion model UNets~\citep{rombach2022diffusion}. Flash Attention is an efficient implementation that is optimized to work well with reduced precision and GPU hardware~\citep{dao2022flashattention}, which was implemented using the XFormers library~\citep{xFormers2022}, allowing us to save compute and memory usage.

\paragraph{Precomputing latents.} 
Each forward pass of SD2 requires computing a latent representation of the input image, as well as transforming the caption into a text embedding. 
Instead of computing the latents for each example during training, we can precompute latents for the entire dataset, amortizing the cost. 
Doing so speeds up training of the model, especially at lower resolutions, in exchange for a one-time fixed cost of precomputing all the latents over 1 epoch. 

\paragraph{Reduced-precision GroupNorm and LayerNorm.} 
Most layers in SD2 are implemented in \textsf{float16} precision, but GroupNorm and LayerNorm are implemented in \textsf{float32}, in part because it was assumed to be necessary for training stability. The resulting, frequent upcasting causes a major bottleneck in training speed. Recent work shows that it is safe to implement LayerNorm using \textsf{float16} precision~\citep{portes2023mosaicbert}, and we found the same to be true of GroupNorm.
We thus cast all GroupNorm and LayerNorm operators to \textsf{float16} and are able to further reduce total memory consumption and accelerate training.

\paragraph{Fully-Sharded Data Parallelism (FSDP).} 
FSDP is a variant of data-parallel training that shards the models parameters, gradients and optimizer state across multiple devices. When training data batches  do not fit into memory, we do several forward and backward passes on smaller microbatches, followed by a single gradient update. At GPU scale, there may only be a single microbatch, so the time for the gradient update can become a significant bottleneck. In standard data distributed training, each GPU communicates all its gradients to every other GPU, and then each GPU updates its local copy of the model. Instead, we use a different paradigm inspired by~\citep{xu2020automatic} where each GPU only gets the gradients and updates the weights for a small part of the model before sending the updated weights for that part of the model to all of the other GPUs. By dividing the update step across all the GPUs, we can ensure that the amount of work per GPU decreases as we increase the number of GPUs, helping us achieve linear scaling. To tackle this problem, we use PyTorch's experimental support for Fully Sharded Data Parallelism (FSDP), specifically, FSDP’s SHARD\_GRAD\_OP mode.  

\paragraph{Scheduled Exponential Moving Average (EMA).}
SD2 uses EMA, which maintains an exponential moving average of the weights at every gradient update for the entire training period. This can be slow due to the memory operations required to read and write all the weights at every step. Since the old weights are decayed by a factor of 0.9999 at every batch, the early iterations of training only contribute minimally to the final average. We decide to only apply EMA for the final 50K steps (about 3.5\% of the training period), and are able to avoid adding overhead and still achieve a nearly equivalent EMA model.

%% file: section/99-appendix/42-commoncanvas/50-app-commoncanvas-figs.tex
\section{Additional Figures}

We provide some additional details on training and qualitative results.

\begin{figure}[t]
    \centering
    \includegraphics[width=\linewidth]{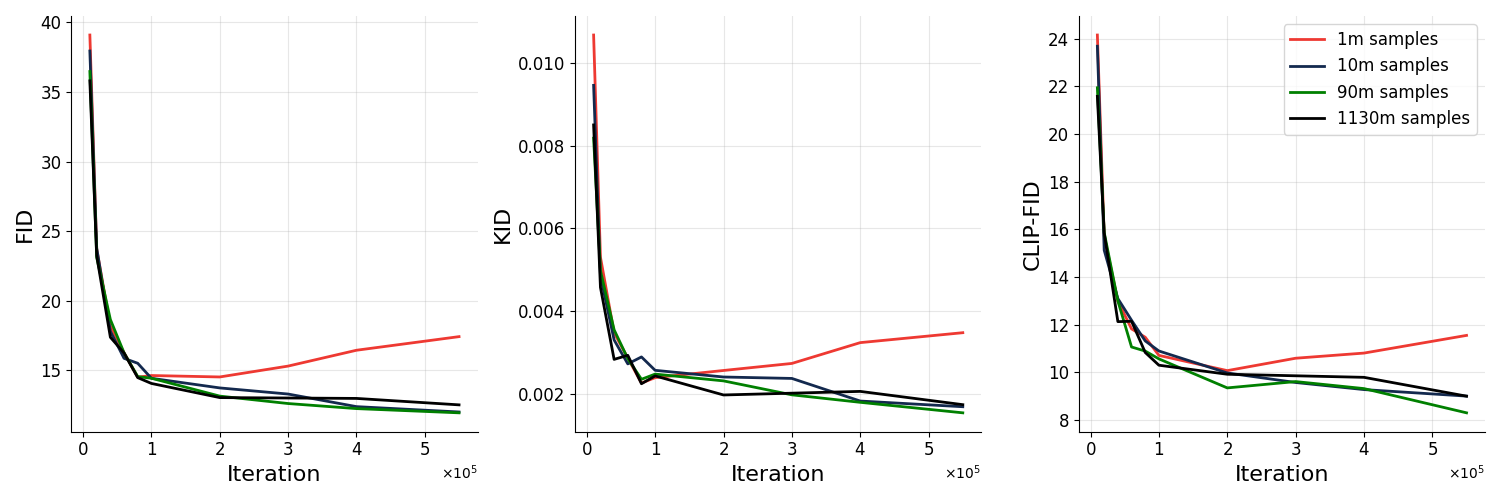}
    \caption{MS COCO metrics over training duration for various dataset sizes. We investigate how reducing the size of the training dataset affects training dynamics, and find that performance is largely unchanged until dropping below 10 million samples. We show that the FID of the eval set remains stable as training progresses. However, reducing the number of samples in our training dataset to 1 million leads to divergence. This finding suggests that only 10 million to 1 million synthetic image caption pairs are needed for good performance on MS COCO.}
    \label{fig:eval-over-time-less-data}
\end{figure}

\begin{figure*}
    \centering 
    \setlength{\tabcolsep}{1pt}
    \setlength{\itemwidth}{0.2\linewidth}
    \newcolumntype{M}[1]{>{\centering\arraybackslash}m{#1}}
    \begin{tabular}{M{\itemwidth}M{\itemwidth}M{\itemwidth}M{\itemwidth}M{\itemwidth}}
      Prompt & SD2 & \modelname-SC & \modelname-SNC & \modelname-LNC \\

    a 3D CAD model of an airplane &
    \includegraphics[width=\itemwidth]{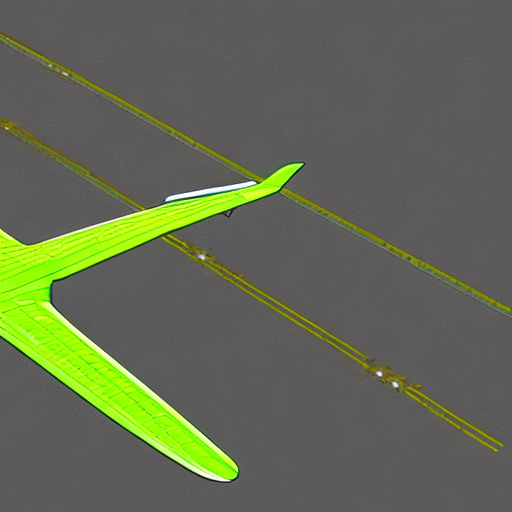}  &
    \includegraphics[width=\itemwidth]{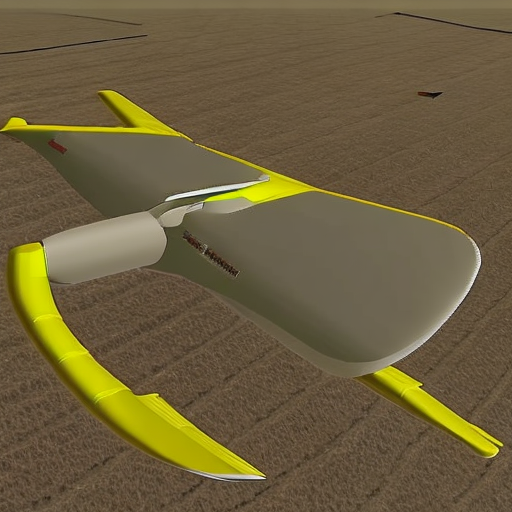} &
    \includegraphics[width=\itemwidth]{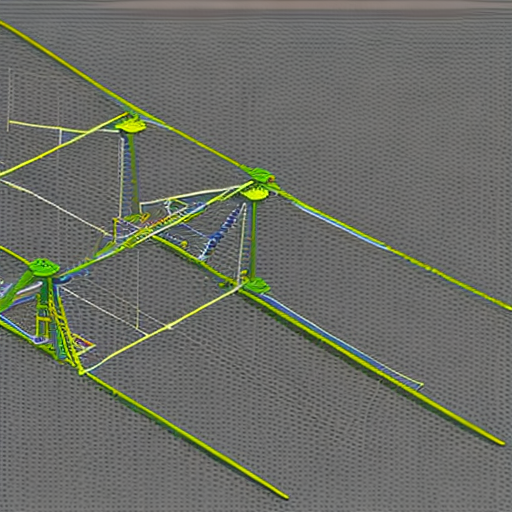} &     
    \includegraphics[width=\itemwidth]{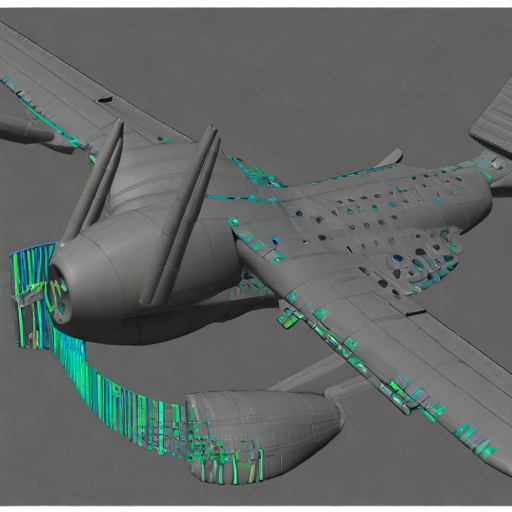} \\

    a bear and a fox in the forest &
    \includegraphics[width=\itemwidth]{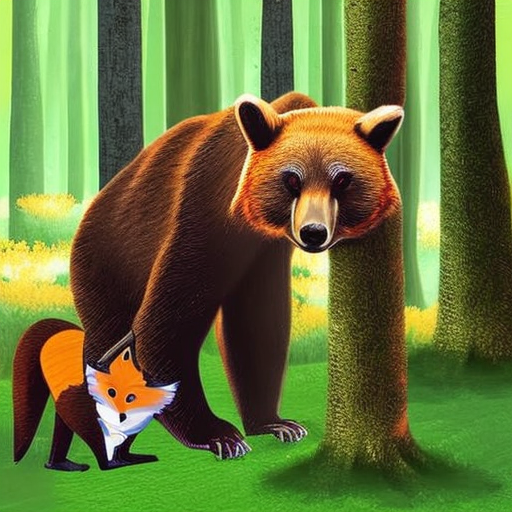}  &
    \includegraphics[width=\itemwidth]{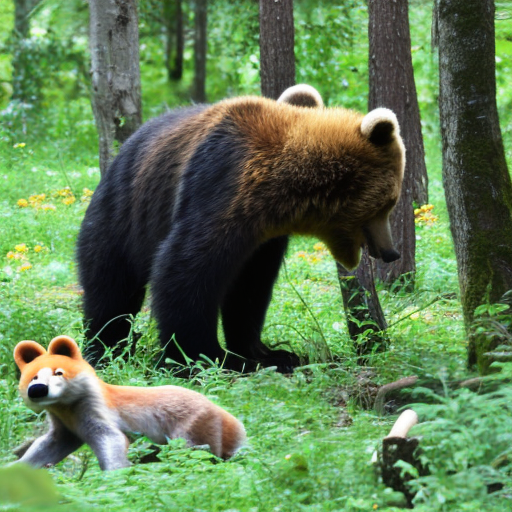} &
    \includegraphics[width=\itemwidth]{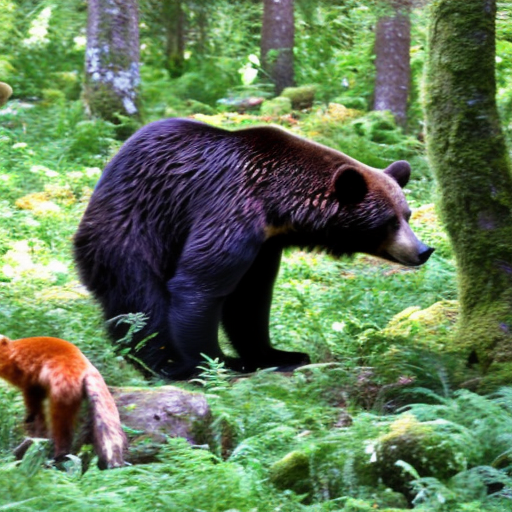} &     
    \includegraphics[width=\itemwidth]{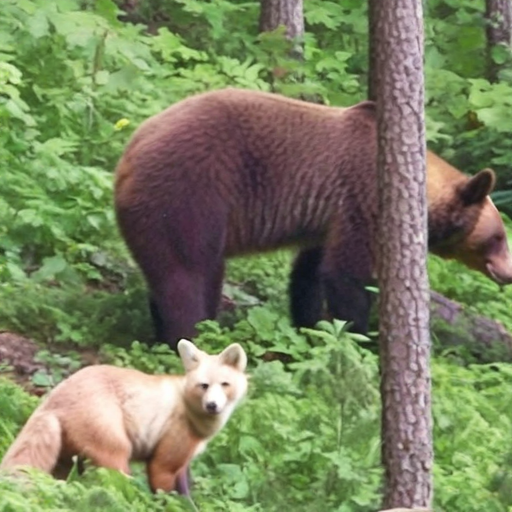} \\

    a klein bottle &
    \includegraphics[width=\itemwidth]{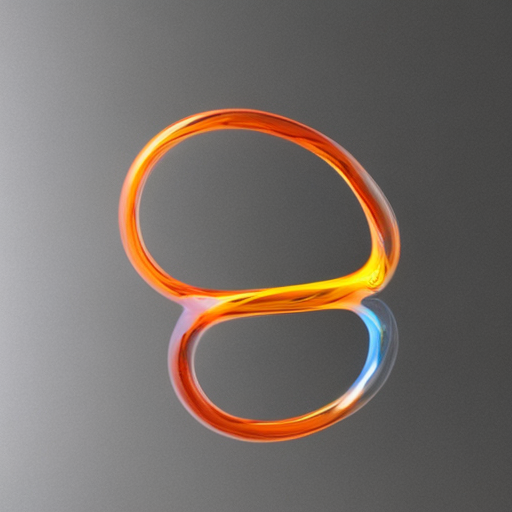}  &
    \includegraphics[width=\itemwidth]{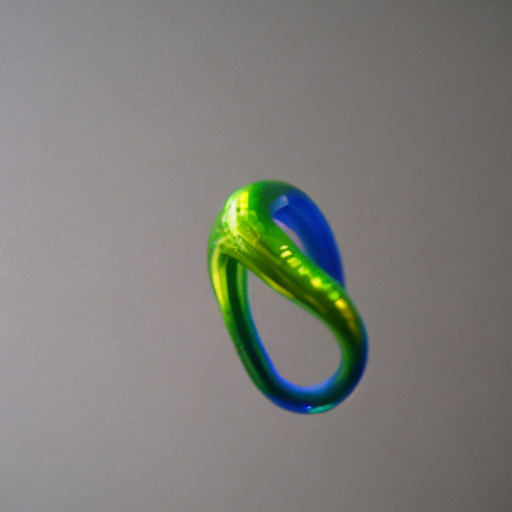} &
    \includegraphics[width=\itemwidth]{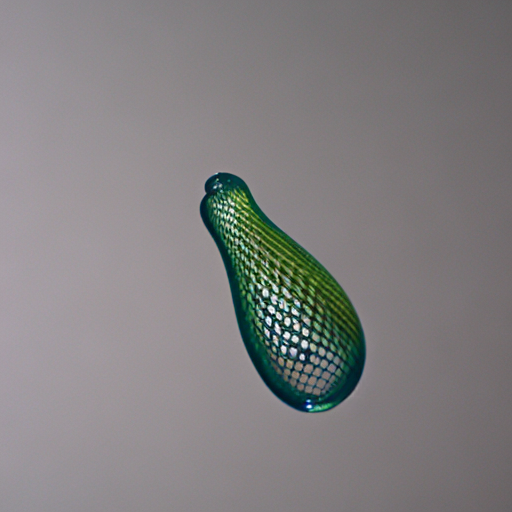} &     
    \includegraphics[width=\itemwidth]{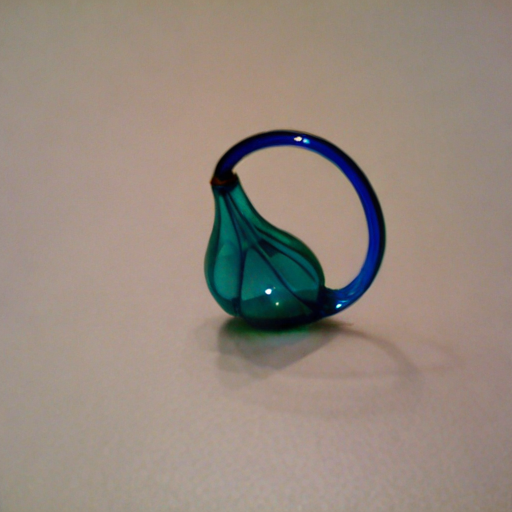} \\

    a partially cut birthday cake with pink and blue frosting &
    \includegraphics[width=\itemwidth]{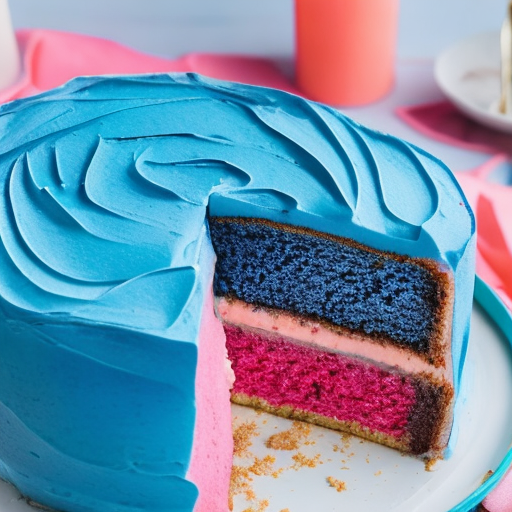}  &
    \includegraphics[width=\itemwidth]{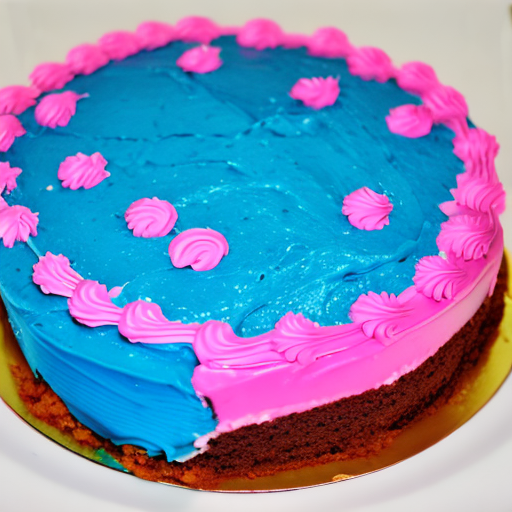} &
    \includegraphics[width=\itemwidth]{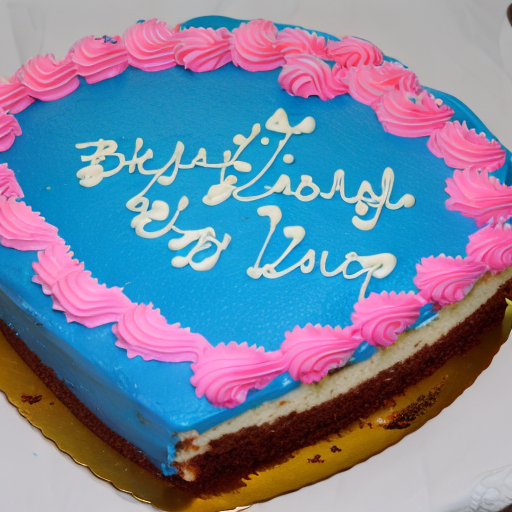} &     
    \includegraphics[width=\itemwidth]{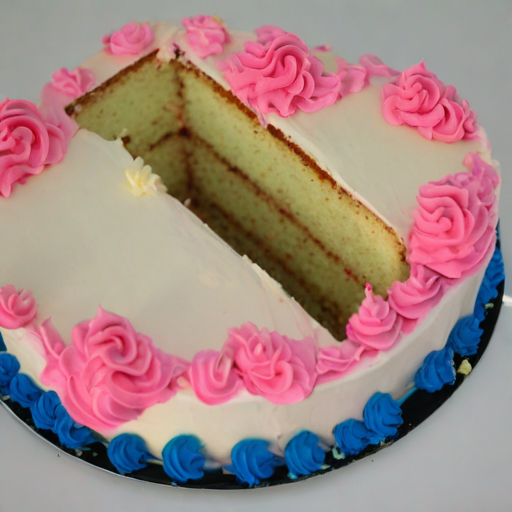} \\

    two hummingbirds and a squirrel in a bird bath &
    \includegraphics[width=\itemwidth]{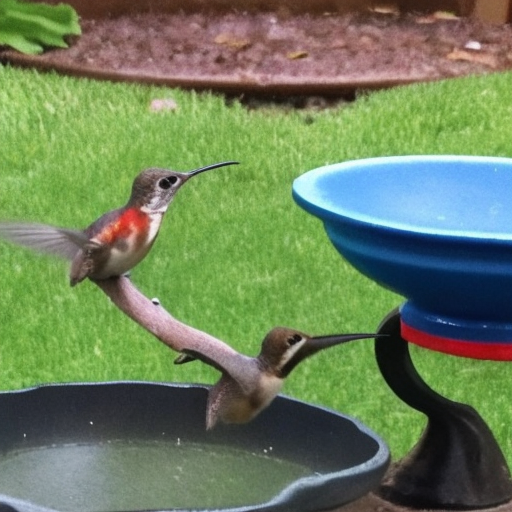}  &
    \includegraphics[width=\itemwidth]{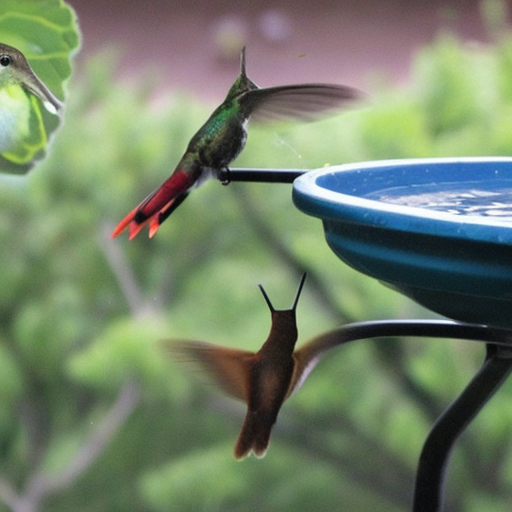} &
    \includegraphics[width=\itemwidth]{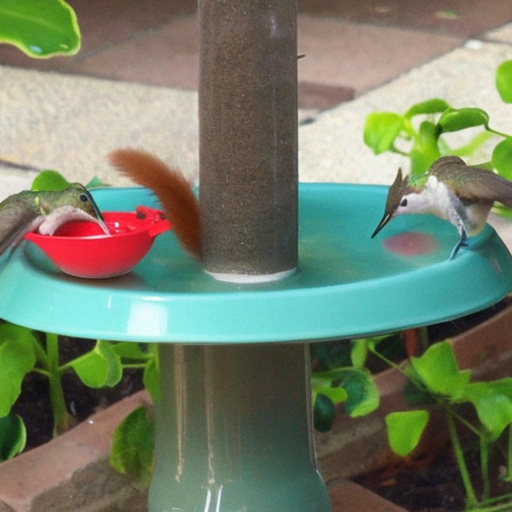} &     
    \includegraphics[width=\itemwidth]{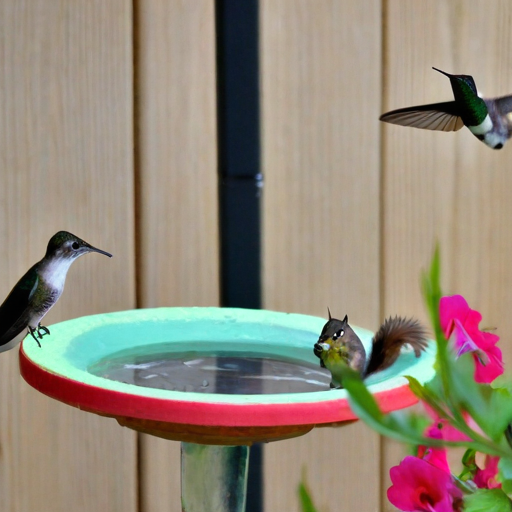} \\
    
    \end{tabular}

    \caption{Additional qualitative examples comparing SD2 to our model trained on the commerical split (\modelname-SC), non-commerical split (\modelname-SNC), and the larger UNet model trained on the non-commercial (\modelname-LNC).}
    \label{fig:additional-qual-exs}
\end{figure*}

\begin{figure*}
    \centering
    \setlength{\tabcolsep}{1pt}
    \setlength{\itemwidth}{0.24\linewidth}
    \begin{tabular}{c c c c c}
        \includegraphics[width=\itemwidth]{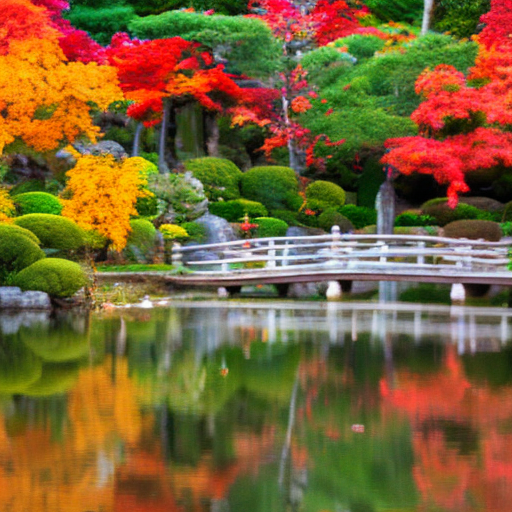}  & 
        \includegraphics[width=\itemwidth]{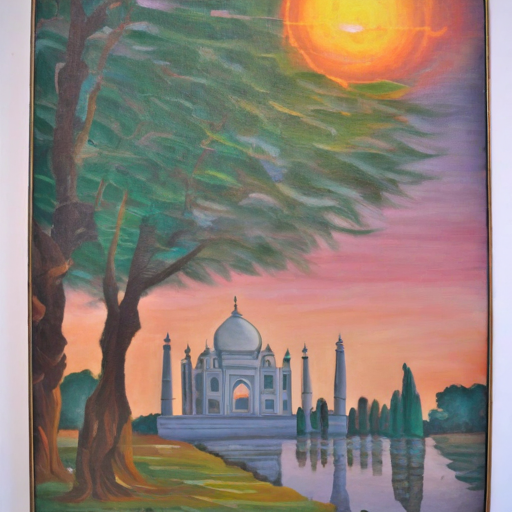}  &         &         \includegraphics[width=\itemwidth]{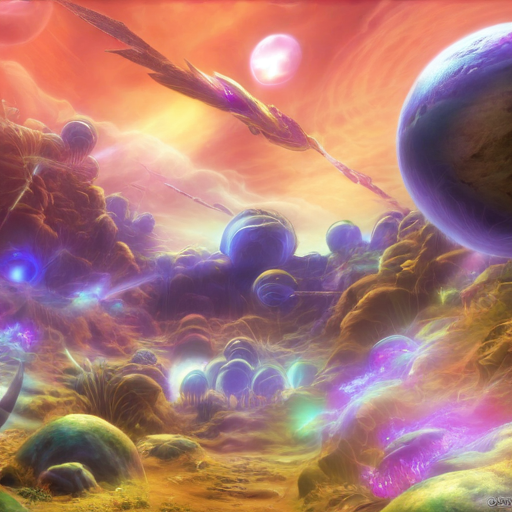}  &         \includegraphics[width=\itemwidth]{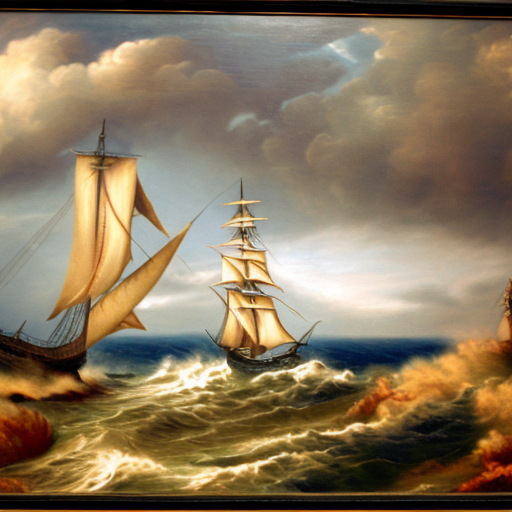}  
        \\ 
    \end{tabular}
    \caption{Additional qualitative examples of our \modelname{} models.}
    \label{fig:enter-label}
\end{figure*}

\begin{figure*}
    \centering 
    \setlength{\tabcolsep}{1pt}
    \setlength{\itemwidth}{0.2\linewidth}
    \newcolumntype{M}[1]{>{\centering\arraybackslash}m{#1}}
    \begin{tabular}{M{\itemwidth}M{\itemwidth}M{\itemwidth}M{\itemwidth}M{\itemwidth}}
      Input for BLIP2 & BLIP2 Caption & SD2 & \modelname-SNC & \modelname-SC \\

        \makebox[\itemwidth]{%
            \includegraphics[width=\itemwidth,height=\itemwidth,keepaspectratio]{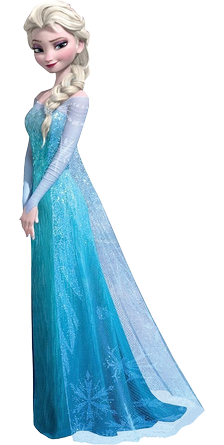}%
        }%
    &
          an image of elsa from frozen  &
    \includegraphics[width=\itemwidth]{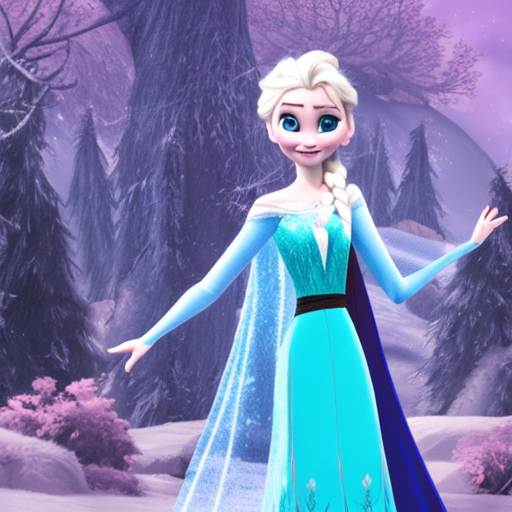} &
    \includegraphics[width=\itemwidth]{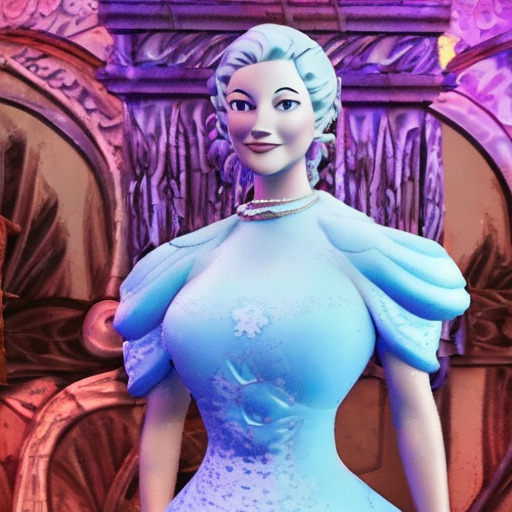} &     \includegraphics[width=\itemwidth]{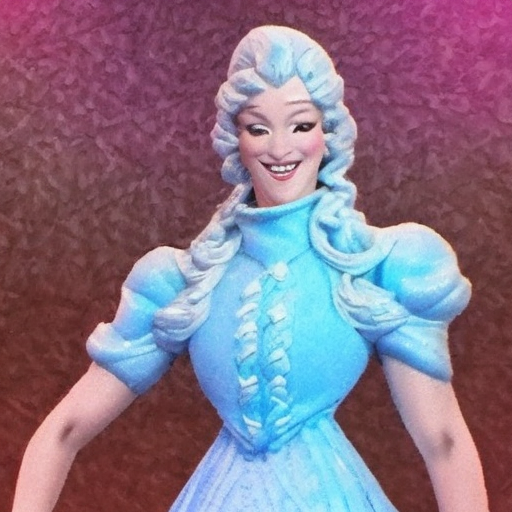} \\

    \makebox[\itemwidth]{%
            
    \includegraphics[width=\itemwidth,height=\itemwidth,keepaspectratio]{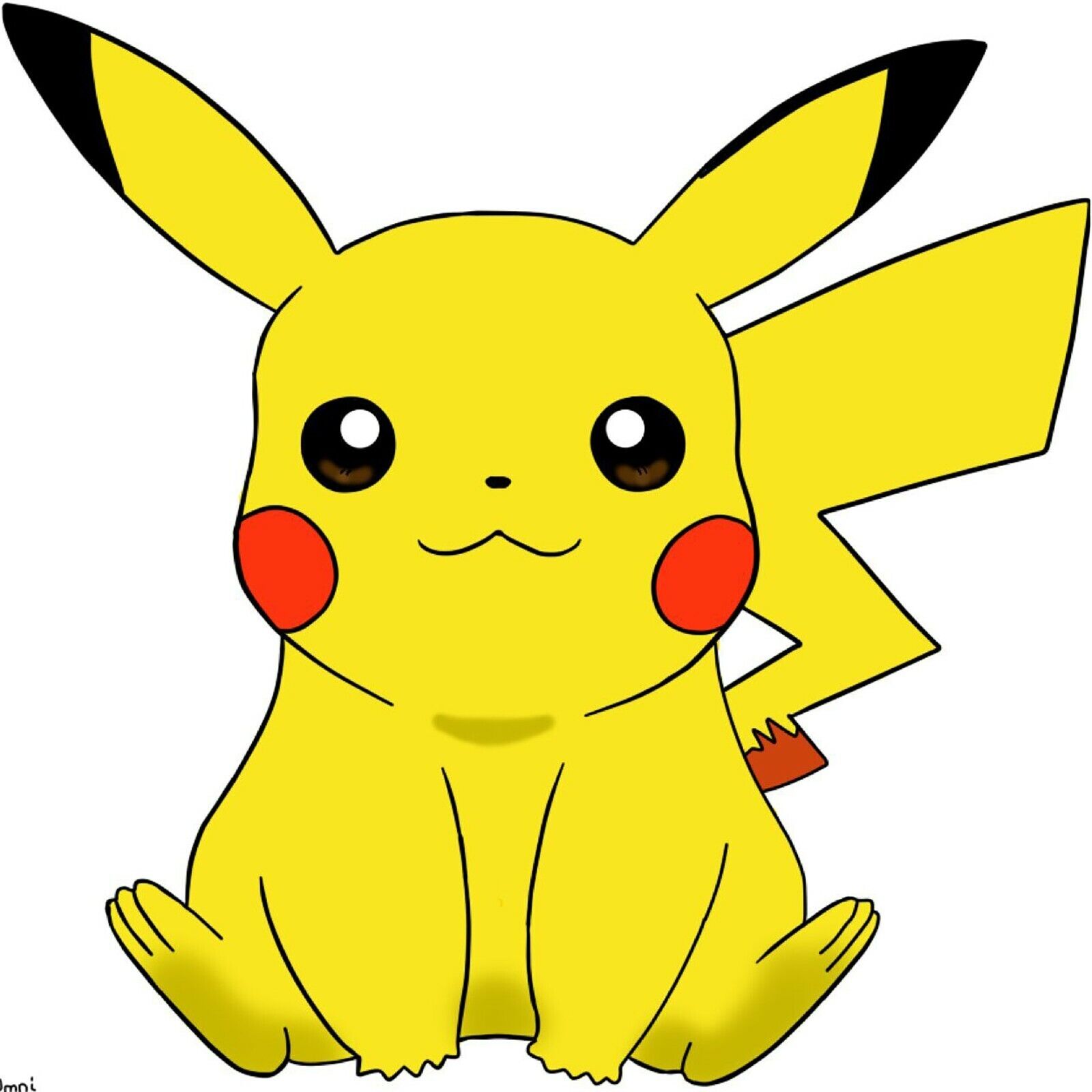}  
    } & 
        pikachu pikachu pikachu pikachu pikachu pikachu pikachu pikachu pikachu pikachu &
    \includegraphics[width=\itemwidth]{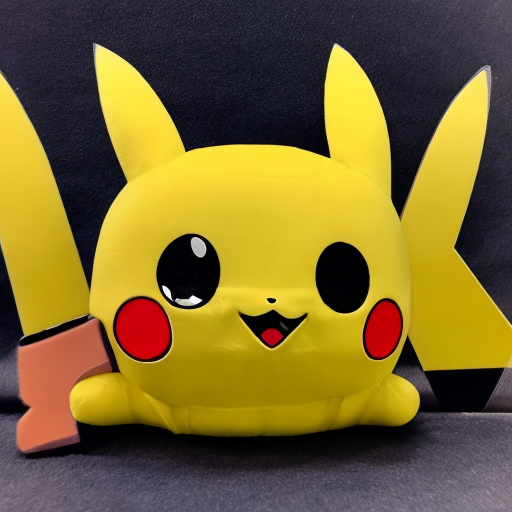} &
    \includegraphics[width=\itemwidth]{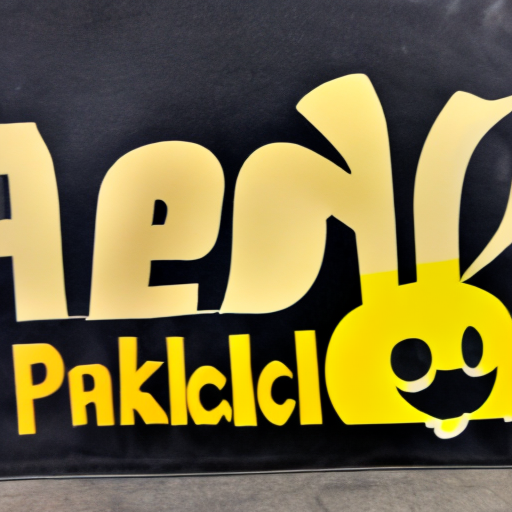} &     \includegraphics[width=\itemwidth]{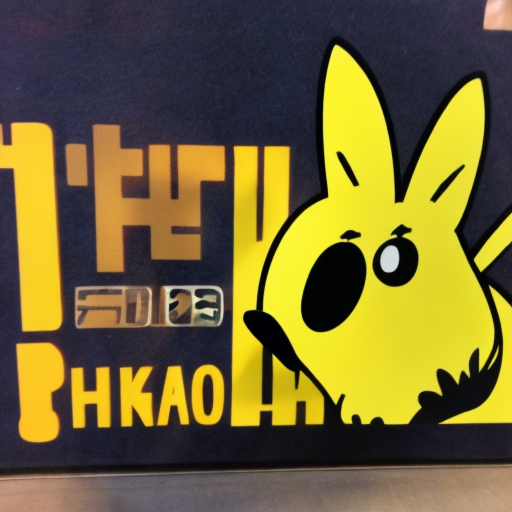} \\
        \makebox[\itemwidth]{%
            
    \includegraphics[width=\itemwidth,height=\itemwidth,keepaspectratio]{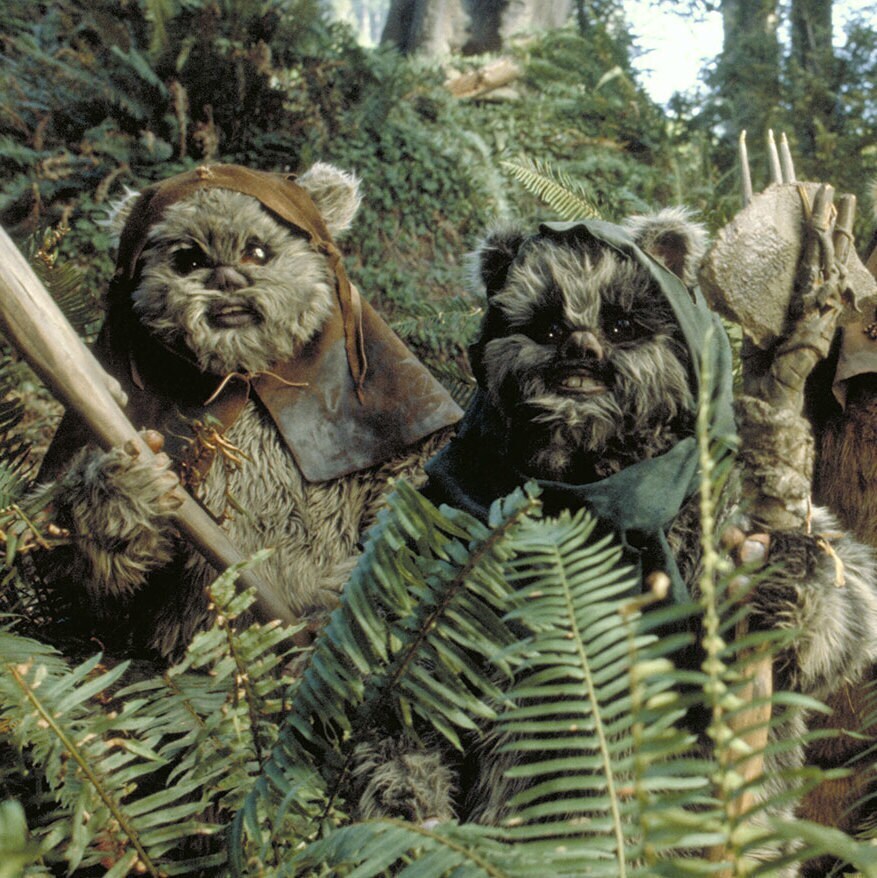}  
    } & 
    three characters dressed like bears, standing in the forest &
    \includegraphics[width=\itemwidth]{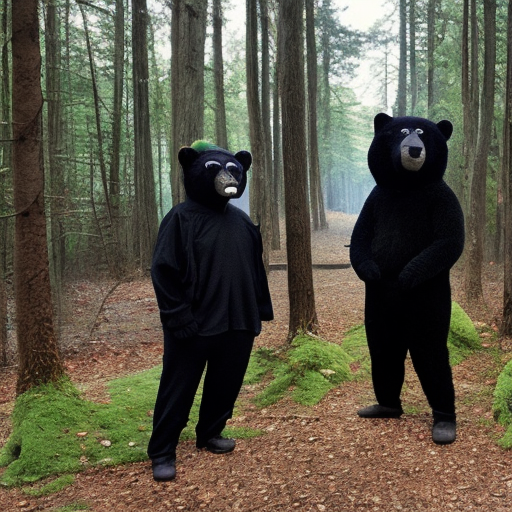} &
    \includegraphics[width=\itemwidth]{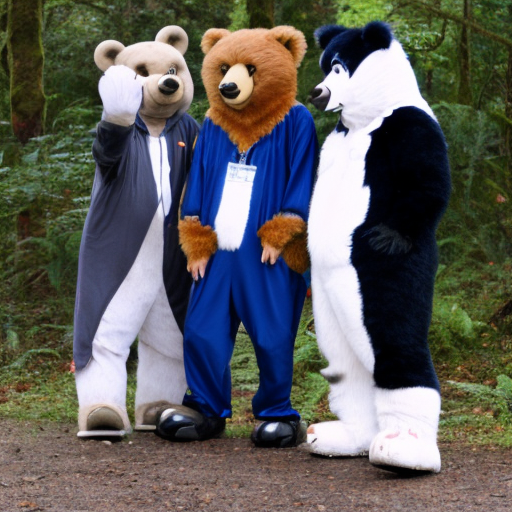} &     
    \includegraphics[width=\itemwidth]{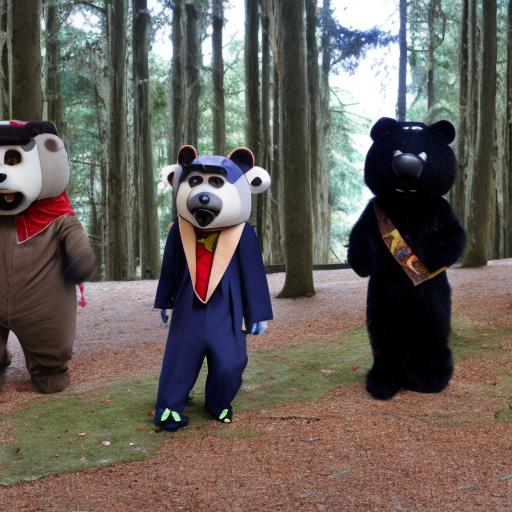} \\
    \end{tabular}

    \caption{Additional qualitative examples comparing our \modelname{} models to SD2, given synthetic BLIP2 captions as prompts. While not perfect, our models are better at avoiding generating potentially problematic data. }
    
    \label{fig:blip-roundtrip}
\end{figure*}

%% file: section/10-intro/12-accountability/12-accountability-main.tex
\chapter{Accountability in an Algorithmic Society: Relationality,
Responsibility, and Robustness in Machine Learning}\label{chapter:accountability}

We begin with some additional framing for the rest of this dissertation.
The work in this chapter represents ideas around accountability and machine-learning based systems that started taking shape in late 2017/ early 2018.\footnote{Indeed, the genesis of these ideas grounded my decision to go to graduate school in computer science, and to study topics at the intersection of machine learning and law.}
The contents of this chapter reflect the synthesis of these ideas in early 2022.
Of course, the world changed significantly just months later with the public launch of ChatGPT. 
Nevertheless, 
the core points articulated in this piece hold up, and in a sense presaged, contemporary challenges of accountability and generative-AI systems.
Arguably, this piece is even more salient now than it was at the time of writing. 
We will address this in future work.\\ 

\noindent\textbf{Chapter summary}: 
In 1996, \emph{Accountability in a Computerized Society}~\cite{nissenbaum1996accountability} issued a clarion call concerning the erosion of accountability in society due to the ubiquitous delegation of consequential functions to computerized systems.  
Nissenbaum~\cite{nissenbaum1996accountability} described four barriers to accountability that computerization presented: 
1) \textit{many hands}, the problem of attributing moral responsibility for outcomes caused by many moral actors; 
2) \textit{``bugs''}, the way software developers might shrug off responsibility by suggesting software errors are unavoidable; 
3) \textit{computer as scapegoat}, the shifting of blame to computer systems as if they were moral actors; 
and 4) \textit{ownership without liability}, the free pass given to the tech industry to deny responsibility for the software they produce
We revisit these barriers in relation to the ascendance of data-driven algorithmic systems --- i.e., machine learning (ML) or artificial intelligence (AI) --- to uncover 
new challenges for accountability that these systems present. 
Nissenbaum's original paper grounded discussion of the barriers in moral philosophy; 
we bring this analysis together with recent scholarship on relational accountability frameworks and discuss how the barriers present difficulties for instantiating a unified moral, relational framework in practice for data-driven algorithmic systems. 
We conclude by discussing ways of weakening the barriers in order to do so.\\

\noindent This chapter is a licensed derivative copy of work published at \emph{FAccT 2022}~\cite{cooper2022accountability}. 

\input{section/10-intro/12-accountability/1210-accountability-intro}
\input{section/10-intro/12-accountability/1230-accountability-framing}
\input{section/10-intro/12-accountability/1240-accountability-barriers}
\input{section/10-intro/12-accountability/1250-accountability-beyond}
\input{section/10-intro/12-accountability/1260-accountability-conclusion}

%% file: section/10-intro/12-accountability/1210-accountability-intro.tex
\section{Introduction}

In 1996, writing against the backdrop of the meteoric rise of the commercial Internet~\cite{leiner2009briefhistory}, Nissenbaum~\cite{nissenbaum1996accountability} warned of the erosion of accountability due to four barriers inimical to societies increasingly reliant on computerized systems.
These barriers are: \textit{many hands}, to refer to the problem of attributing moral responsibility for outcomes caused by multiple moral actors; 
``\textit{bugs},'' the way software developers might shrug off responsibility by suggesting software errors are unavoidable; 
\textit{computer as scapegoat}, the shifting of blame to computers as if they were moral actors; 
and \textit{ownership without liability}, the free pass to the software industry to deny responsibility, particularly via shrink-wrap and click-wrap Terms of Service agreements. 
Today, twenty-five years later, significant work has been done to address the four barriers through developments in professional practices of computer science~\cite{vandorp2002tracking, jarke1998requirements}, organizational management~\cite{javed2014systematic}, and civil law~\cite{mulligan2018governance, eu2017civil}; 
however, the effort to restore accountability remains incomplete. 
In the interim, the nature of computerized systems has been radically transformed by the ascendance of data-driven algorithmic systems\footnote{Since rule-based software systems are also ``algorithmic,'' we specify which of the meanings we intend in settings where the context does not disambiguate.} 
--- e.g., machine learning (ML) and artificial intelligence (AI) --- which either have replaced or complemented rule-based software systems, or have been incorporated within them as essential elements~\cite{carbin2019overparameterization, zhang2003machine,barstow1988artificial, mulligan2019ml, kroll2017aa}.

The resurgent interest in accountability is therefore timely for a world in which data-driven algorithmic systems are ubiquitous.\footnote{We adopt the term ``Algorithmic Society'' as used in Balkin~\cite{balkin2018society}.} 
In domains as varied as finance, criminal justice, medicine, advertising, entertainment, hiring, manufacturing, and agriculture, these systems are simultaneously treated as revolutionary, adopted in high-stakes decision software and machines~\cite{angwin2016machine, ajunwa2021auditing, kleinberg2018human, aaj2017avs, moradi2020ai}, and as novelties~\cite{kleeman2016watson}. 
The failure to comprehensively establish accountability within computational systems through the 1990s and 2000s has thus left contemporary societies just as vulnerable to the dissipation of accountability, with even more at stake. 
We remain in need of conceptual, technical, and institutional mechanisms to assess how to achieve accountability for the harmful consequences of data-driven algorithmic systems --- mechanisms that address both \emph{whom} to hold accountable and \emph{how} to hold them accountable for the legally cognizable harms of injury, property loss, and workplace hazards, and the not-yet-legally-cognizable harms increasingly associated with these systems, such as privacy violations~\cite{citron2022privacy}, manipulative practices~\cite{agarwal2019deepfake, kreps2020gpt}, and automation-driven discrimination~\cite{ajunwa2021auditing}. 

In light of growing concerns over accountability in computing, our paper revisits Nissenbaum's ``four barriers to accountability'' to assess whether insights from that work remain relevant to data-driven algorithmic systems, and to consider how the ascendance of such systems complicates, challenges, and demands more of sociotechnical, philosophical, and regulatory work. 
We first provide context on recent developments in standards of care, law and policy, and computer science that are necessary for our analysis (Section~\ref{sec:acc:rw}). 
Equipped with this background, we recapitulate the elements of moral philosophy on which Nissenbaum~\cite{nissenbaum1996accountability} depended (Section~\ref{sec:acc:rw:philosophy}), and discuss how this moral conception of accountability can be unified with Bovens's relational definition of accountability in political theory~\cite{bovens2007analysing}, which has drawn recent attention in AI ethics scholarship. 
In particular, we contend that moral and relational accountability can be brought together to illuminate the necessary parameters of an accountability framework for data-driven algorithmic systems --- determining \emph{who} is accountable, \textit{for what}, \textit{to whom}, and \emph{under which circumstances} (Section~\ref{sec:acc:rw:narrow}). 
To instantiate such a framework, however, requires recognizing the ways in which data-driven algorithmic systems specifically make determining these parameters challenging. 
We therefore update Nissenbaum's four barriers to accountability in relation to these systems, and clarify the ways that each barrier obscures and complicates realizing a moral, relational accountability framework in practice (Section~\ref{sec:acc:barriers}). Finally, we conclude by suggesting ways of weakening these barriers to accountability, thereby strengthening accountability practices for the entire field (Section~\ref{sec:acc:beyond}).

\subsection{Technological Interventions in Accountability}\label{sec:acc:rw}

Re-visiting the four barriers requires engaging with the significant body of work on accountability produced in the interim. Rather than comprehensively reviewing existing literature---an 
undertaking already addressed in, e.g., Wieringa~\cite{wieringa2020account} and Kohli et al.~\cite{kohli2018translation}---we highlight three areas of work that we find useful for our analysis
:

\paragraph{Standards of care.} These play a crucial role in building a culture of accountability --- establishing best practices and formal guidelines for ensuring that concrete practices align with agreed-upon values (e.g., safety). 
In engineering, standards of care dictate the behaviors and outputs expected of sound work. 
For data-driven algorithmic systems in particular,  they have taken the form of annotations~\cite{beretta2021detecting}, audits~\cite{ajunwa2021auditing}, and frameworks  concerning the appropriate use of data and other artifacts, which are often developed and used in the production of AI/ML systems~\cite{mcmillan2021reusable, boyd2021datasheets, mitchell2019model, shen2021value, hutchinson2021towards, gebru2021datasheets}. 
Taken together, these standards of care support accountability by making the intentions and expectations around such systems concrete; 
they provide a baseline against which one can evaluate deviations from expected behavior and, accordingly, are used to review and contest the legitimacy of specific applications of data-driven techniques.  
Some scholars have re-framed such standards around harmed and vulnerable parties~\cite{raji2020closing, metcalf2021algorithmic}. This work makes clear that standards of care, while important for developing actionable notions of accountability, do not guarantee accountability on their own~\cite{vecchione2021algorithmic, diakopoulos2020accountability}. 
Algorithmic impact assessments attempt to fill this gap~\cite{moss2021assembling}. 
They task practitioners with assessing new technologies in terms of their anticipated impacts~\cite{selbst2021institutional, metcalf2021algorithmic}, and they formalize accountability relationships in ways that may systematically address and correct algorithmic harms.

\paragraph{Law and policy.} 
Literature on data-driven algorithmic systems generally concerns AI/ML-related harms and corresponding interventions. 
Work on liability spans both anticipated harms related to new or forthcoming data-driven technology, including autonomous vehicles and robotics~\cite{abraham2019responsibility, aaj2017avs, surden2016avs, elish2019crumple, cooper2022fast}, and not-yet-legally-cognizable harms, such as unfair discrimination due to demographically\--imbalanced, biased, or otherwise-discredited training data~\cite{hellman2021data, okidebe2022data, waldman2019power}, privacy violations~\cite{crawford2014harms, citron2022privacy, kaminski2019algorithmic}, and manipulation~\cite{kreps2020gpt}. 
Regulatory and administrative scholarship tends to analyze data-driven algorithmic systems in relation to legislation and policy that predates many AI/ML technological developments~\cite{shah2018algorithmic, viljoen2021relational, whittington2015push, sadowski2021everyone, mulligan2019ml, delgado2022uncommontask}. 
That said, recent regulatory interventions, including GDPR (the nascent, yet wide-reaching data-privacy policy in the EU~\cite{kang2020algorithmic, wachter2017transparent, hamon2021impossible}) and the California Consumer Privacy Act of 2018~\cite{barrett2019eu}, which have also been applied to AI/ML systems, are increasingly represented within the law and policy literature. 

Law and policy approaches tend to focus on transparency, which is of broad import in democratic governance and is intimately connected to accountability~\cite{mulligan2018governance}. 
Transparency is necessary for identifying responsible parties (in order to attribute harms to those who are responsible for them), and necessary for identifying the sources of these harms and potential mitigations~\cite{diakopoulos2020accountability}. 
Work in this area spans a range of urgent concerns surrounding lack of transparency in data-driven algorithmic systems. These include the obfuscation of data provenance~\cite{wachter2018bigdata, levy2016transparency}, particularly caused by the concentration of data ownership within data brokers~\cite{ftc2014databrokers,lambdan2019databrokers, young2019beyond}, and insufficient transparency of algorithms and models, which contributes to the inscrutability of automated decisions~\cite{cofone2019secrecy, lehr2017legalml, kroll2017aa}. 
Critics have argued that outsourcing legal decisions to automated tools, particularly data-driven tools that obscure underlying decision logic, can create a crisis of legitimacy in democratic decision-making~\cite{mulligan2019ml, gao2021framework, calo2021modeling, citron2020workingpaper}.

\paragraph{Computer science.} Research in AI/ML has increasingly treated accountability as a topic for scholarly inquiry. 
In updating Nissenbaum's barriers, we address cases in which researchers explicitly recognize the relationship between their work and accountability~\cite{kim2021accountability} --- namely, in auditing and transparency --- and work on \textit{robustness}, which we identify as having significant implications for accountability, even when this  work itself does not explicitly make the connection. 
Recent work on audits underscores the importance of being able to analyze algorithmic outputs to detect and correct for the harm of unfair discrimination~\cite{adler2018auditing, raji2020closing}. 
Transparency tends to be treated as a property of models, particularly whether a model is interpretable or explainable to relevant stakeholders~\cite{doshivelez2018interpretability, freitas2014interpretability,bhatt2021uncertainty}. 
More recently, computational work has begun to take a more expansive view of transparency, applying it to other parts of the ML pipeline, such as problem formulation, data provenance, and model selection choices~\cite{forde2021model, kroll2021traceability, sivaprasad2020hpo, sloane2020participation, cooper2021hpo}. 

Lastly, often overlooked, \textit{robustness} draws attention to whether a model behaves as expected under likely, unlikely, anomalous, or adversarial conditions. 
Designing for and evaluating robustness implicates accountability, as it requires researchers to define their expectations of model performance rigorously; 
this in turn encourages inquiry into how to prevent deviations from those expectations, and to identify (and ideally correct for) such deviations. 
Robustness thus encompasses work in AI/ML that aims to achieve theoretical guarantees in practice~\cite{yang2019correctness, meinke2019provable, zhang2020tunamh}, and work that, even in the absence of such guarantees, produces models with reproducible empirical behavior~\cite{bouthillier2019reproducibility, raff2019reproducibility}. 
Robustness also includes the ability for models to generalize beyond the data on which they were trained~\cite{neyshabur2017generalization, hu2020generalization}, ranging from natural cases of distribution shift~\cite{ovadia2019shift, koh2021shift} to handling the presence of adversaries that are trying to game model outputs~\cite{goodfellow2014adversarial, papernot2015adversarial,szegedy2013adversarial}.

%% file: section/10-intro/12-accountability/1230-accountability-framing.tex
\section{Conceptual Framing} \label{sec:acc:framing}

The conceptual framing of accountability for this paper draws from two sources of scholarship: 
1) moral philosophy, which construes accountability as a relationship between and among multiple actors; 
and 2) political theory and the social sciences, largely focusing on work by Mark Bovens, whose framework for identifying accountability relationships has been particularly influential in contemporary scholarship on  ``algorithmic accountability''\footnote{We discuss concerns with this phrase in Section~\ref{sec:acc:barriers:scapegoat} (\textit{scapegoat}).}~\cite{wieringa2020account,kroll2021traceability,kacianka2021accountable}.

\subsection{Accountability in Moral Philosophy}\label{sec:acc:rw:philosophy}

Numerous efforts in moral philosophy have sought to develop a rigorous conception of accountability. 
We focus on two threads in the literature, \emph{blameworthiness} and \emph{relationships between moral actors}, and correspondences between the two.\looseness=-1

\paragraph{Blame.}
Nissenbaum~\cite{nissenbaum1996accountability} anticipated a problem of diminishing accountability as societies become increasingly dependent on computerized systems. 
She attributed this likelihood to the emergence of barriers to accountability in computerized society, and turned to philosopher Joel Feinberg's work to explain how and why these barriers are prone to arise: 
Blame, defined in terms of \emph{causation} and \emph{faultiness}, is assigned to moral agents for harms they have caused due to faulty actions~\cite{feinberg1970doing,feinberg1985sua}.\footnote{Neither of these elements is straightforward --- in fact, they are both the subjects of centuries of philosophical and legal thinking. 
    Faultiness, e.g., presumes free agency --- a concept whose metaphysical character and role in moral attribution has been the subject of centuries' 
    long debate --- and is a basic concept in all legal systems that informs judgements of legal liability (categorizing harmful actions as intentional, reckless, and negligent)~\cite{feinberg1970doing,feinberg1968collective}.
} 

Following Feinberg, Nissenbaum conceives of actors as accountable when they step forward to answer for harms for which they are blameworthy. Her concern was that in computerized societies too many circumstances would arise where no one would step forward to acknowledge blame for harm, whether due to genuine puzzlement or intentional avoidance. 
Accordingly, the barriers to accountability that she identifies arise because the conditions of accountability are systematically obscured, due, at some times, to circumstances surrounding computerization and, at other times, to a societal breakdown in confronting willful failures.  
\textit{Many hands} obscures lines of causal responsibility (Section~\ref{sec:acc:barriers:manyhands}); 
\textit{``bugs''} obscures the classification of errors as instances of faulty action (Section~\ref{sec:acc:barriers:bugs}); 
\textit{scapegoating computers} obscures answerable moral actors by misleadingly or mistakenly attributing moral agency to non-moral causes (Section~\ref{sec:acc:barriers:scapegoat}); 
and \textit{ownership without liability} bluntly severs accountability from blame (Section~\ref{sec:acc:barriers:liability}).\looseness=-1

\paragraph{Relationality.}
An alternative conception of accountability expands the focus to consider responsibility in light of the relationships between moral actors. 
Watson~\cite{watson1996two}, for example, argues that responsibility should cover more than \textit{attributablility}, a property assigned to an actor for bringing about a given outcome \cite{talbert2019sep}. 
A second dimension, which he calls \textit{accountability}, situates responsibility in a \emph{relationship among actors}. 
For Watson, ``Holding people responsible is not just a matter of the relation of an individual to her behavior; it also involves a social setting in which we demand (require) certain conduct from one another and respond adversely to another's failures to comply with these demands"~\cite[p. 229]{watson1996two}. 
Other work, including T.M. Scanlon's theory of responsibility, provides accounts of both \textit{being responsible} and \textit{being held responsible}, where the latter describes situations when parties violate relationship-defined norms \cite{scanlon2000we, shoemaker2011attributability}.
Accordingly, the characteristics of a harmed party might dictate whether, or what, accountability is needed. 
For instance, if one causes harm in self defense, there may be no moral imperative to hold them accountable.

\vspace{.5cm}\noindent 
This work attempts to situate accountability in the social, political, institutional, and interpersonal relationships in which we are enmeshed. 
Accordingly, the relationship-defined obligations we have to one another --- as spouses, citizens, employees, friends, etc. --- may dictate what it is we are responsible for, as well as the types and degrees of accountability we can expect. 
By situating accountability not just as \textit{attributability} between action and actor, but instead within a social framework, some of what has come out of the so-called ``narrow'' notion of accountability in political theory (discussed below in Section \ref{sec:acc:rw:narrow}) can be derived from the vantage of a more ``pure'' moral philosophy. 
Rather than formally pursuing this derivation here, we instead simply suggest that these notions of accountability need not be framed as alternatives to one another. 
Moral philosophy offers concepts through which a given relational framing --- be it interpersonal, institutional, or political --- can be said to be legitimate and ethically viable. 
Similarly, for practitioners holding a variety of organizational positions (in relation to one another), the moral responsibilities that individuals hold can shape the ethical obligations and specific forms of accountability at play.

\subsection{Accountability in Political Theory and the Social Sciences}\label{sec:acc:rw:narrow}

The work in moral philosophy discussed above aligns with work on accountability as a property of social structures~\cite{garfinkel1984studies}, which holds it to be relational --- not merely as a requirement on an accountable party to ``own up'' to blameworthy action as an obligation \textit{to} another. 
In the past few years, ``algorithmic accountability'' has attracted growing interest in approaches that are institutional or structural in character. 

The work of political scientist Mark Bovens, particularly what he has labeled, a ``narrow definition''~\cite{bovens2007analysing,bovens2014public}, has informed recent literature on accountability for ``algorithmic systems''~\cite{wieringa2020account}. 
Prompted by a concern that newly formed governmental structures and public authorities in the European Union lack ``appropriate accountability regimes''~\cite[p. 447]{bovens2014public}, Bovens proposed that accountability obtains between two key roles: 
an \emph{accountable actor} and a \emph{forum}. 
Under certain conditions, or in the wake of certain incidents, accountability exists when an accountable actor has an enforceable obligation to a forum to explain and justify itself --- to address a forum's questions and judgments and possibly suffer sanctions. 
Bovens calls this a ``relational'' definition because it locates accountability in a social relation between those occupying one role (e.g. governmental department, a public authority, or a person acting in an official capacity) and another (e.g., a different governmental entity, oversight committee, or even an individual acting in a relevant capacity, e.g. journalist).  
We read Bovens as gesturing toward  four key parameters in any relational accountability framework for which appropriate values need to be specified:

\textit{Who is accountable?}: \emph{Accountable actors} may include those who are not directly responsible for harm (e.g., engineers) but are designated as accountable (or liable) because of their deep pockets, capacities to render explanations, or positions in organizational hierarchies, such as corporate officers or government procurers of data-driven systems.

\textit{For what?}: Beyond legally-cognizable harms (e.g., bodily injury, property damage, pecuniary losses), harms particularly associated with data-driven algorithmic systems include privacy violations~\cite{citron2022privacy}, automation-driven unfair discrimination~\cite{ajunwa2021auditing}, autonomy losses due to manipulation~\cite{agarwal2019deepfake,kreps2020gpt}, and any number of emergent harms associated with novel technologies and their deployment. 

\textit{To whom?}: The members of the \emph{forum} may not just include those who are themselves harmed (or placed in harm's way through heightened risk). 
They may also include those deputized to represent and advocate on behalf of vulnerable parties, such as lawyers and public or special interest advocacy groups. 
Beyond direct advocates, these may include groups and individuals in oversight capacities such as journalists, elected officials, government agencies, professional societies, or the many publics which coalesce around particular matters of concern~\cite{metcalf2022relationship}.

\textit{Under which circumstances?}: This concerns the nature of the obligation --- what \emph{accountable actors} may owe to the forum (to explain, be judged, and address questions and challenges). 
For example, Moss et al.~\cite{moss2021assembling} describes an array of components that constitute accountability within impact assessment frameworks, noting that the specific 
obligations an actor owes to a forum depend on the norms of that relationship. 

\paragraph{Bringing together the moral and the relational.} 
Proponents of Bovens's relational framework claim that it illuminates the sociopolitical stakes of transparency and explainability, showing why these concepts are necessary for any accountability framework for data-driven algorithmic societies, even though they are ultimately not sufficient to constitute accountability in and of themselves~\cite{wieringa2020account}. 
Moreover, by defining actors' roles and capacities in terms of the respective sociopolitical structures in which we live, Bovens's framework is \emph{not} directed at the rights and obligations we have to one another as bare moral actors. 
We note that bringing together Bovens's relational definition with the moral conception of accountability can help clarify the scope of possible values for the framework's parameters: 
those who have caused or contributed to harm through faulty action are contenders for the class of \emph{accountable actors}, and those who have suffered harm (and/or their representatives) deserve a place among the members of the \emph{forum}. 

This point shows a confluence between accountability as answerability for blameworthy action, and accountability as a social arrangement. 
Being blameworthy for harm is (almost always) a sufficient condition for being designated an accountable actor; 
being harmed through blameworthy action is (almost always) a sufficient condition for being designated a member of the forum, empowered to demand explanations. 
These two conceptions of accountability --- the moral and the relational --- do not stand against one another as alternative solutions to the same problem; 
they are solutions to different problems that intersect in constructive ways.

\vspace{.5cm}\noindent 
Nevertheless, hard work remains to explain and justify concrete, appropriate values for these parameters, and to construct pervasive structures for accountability through context-bound contestation~\cite{metcalf2021algorithmic}. 
In Section~\ref{sec:acc:barriers} below, we demonstrate how data-driven algorithmic systems heighten the barriers to accountability by further obscuring conditions of responsibility and fault, which in turn presents challenges for instantiating the four parameters of a moral, relational accountability framework.  

%% file: section/10-intro/12-accountability/1240-accountability-barriers.tex
\section{Revisiting the Four Barriers to Accountability}\label{sec:acc:barriers}

In a typical scenario in which software is integrated into a functional system --- fully or partially displacing groups of human actors --- accountability could be displaced along with human actors who are its bearers. 
The cumulative effect of such displacements is the increasing incidence of harmful outcomes for which no one answers, whether these outcomes are major or minor, immediate or long-term, or accrue to individuals or to societies. 
Resuscitating accountability is no simple task because computerization sets up particularly troublesome barriers to accountability: 
\textit{Many hands}~(\ref{sec:acc:barriers:manyhands}), \textit{``Bugs''}~(\ref{sec:acc:barriers:bugs}), \textit{The computer as scapegoat}~(\ref{sec:acc:barriers:scapegoat}), and \textit{Ownership without liability}~(\ref{sec:acc:barriers:liability})~\cite{nissenbaum1996accountability}. 
These interdependent barriers are not necessarily an essential quality of computer software. 
Rather, they are a consequence of how software is produced, integrated into institutions, and embedded within physical systems; 
they are a function of the wonderment and mystique that has grown around computerization, and the prevailing political economy within which the computer and information industries have thrived. 
In the sections that follow, we revisit the barriers to accountability with an eye turned toward their implications amid the massive growth and adoption of data-driven algorithmic technologies. 
We provide examples of the barriers in action and defer discussion of how the barriers can be weakened to Section~\ref{sec:acc:beyond}.

\input{section/10-intro/12-accountability/1241-accountability-barriers-hands}
\input{section/10-intro/12-accountability/1242-accountability-bugs}
\input{section/10-intro/12-accountability/1243-accountability-scapegoat}
\input{section/10-intro/12-accountability/1244-accountability-liability}

%% file: section/10-intro/12-accountability/1241-accountability-barriers-hands.tex
\subsection{The Problem of \textit{Many Hands}}\label{sec:acc:barriers:manyhands}

The barrier of \textit{many hands} arises due to the large number of actors often involved in the design, development, and deployment of complex computerized systems. 
When such systems cause harm, it may be difficult to isolate the component(s) at its source and the agents responsible:  
``Where a mishap is the work of `many hands,' it may not be obvious who is to blame because frequently its most salient and immediate causal antecedents do not converge with its locus of decision making''~\cite[p. 29]{nissenbaum1996accountability}. 
Nissenbaum further analyzes the difficulty of  \textit{many hands} by showing how it operates at four different levels: 
1) software is produced in institutional, often corporate, settings in which there is no actor responsible for all development decisions; 
2) within these settings, multiple, diffuse groups of engineers contribute to different segments or modules of the overall deployed system, which additionally often depends on software implemented by other actors (in today's landscape, this may result in licensed or freely-available open-source software); 
3) individual software systems often interact with or depend on other software systems, which themselves may be unreliable or present interoperability issues; 
4) hardware, not just software, often contributes to overall system function, particularly in cyber-physical systems, and it can be difficult to pinpoint if harms occur due to issues with the code, the physical machine, or the interface between the two. 
Any and all of these four levels of \textit{many hands} problems can operate simultaneously, further obscuring the source of blame.  

These difficulties at the heart of the \textit{many hands} problem persist, further complicated in numerous ways now that computer systems are ubiquitous rather than merely ascendant. 
We focus on how data-driven algorithmic systems complicate this barrier with novel challenges using two illustrative (though necessarily non-exhaustive) examples: 
1) The \emph{ML pipeline} --- the multi-stage process by which machine-learned models are designed, trained, evaluated, deployed, and monitored; 
2) Reliance of contemporary data-driven algorithmic systems on the \emph{composability of openly-available ML toolkits and benchmarking suites}; 
these toolkits, often developed and maintained by large tech companies, tend to be advertised as general- or multi-purpose, and are frequently (mis)used in specific, narrow applications.

\paragraph{The ML pipeline.} 
The ML pipeline is a dynamic series of steps, each of which can involve multiple groups of actors, including designers, engineers, managers, researchers, and data scientists. 
The pipeline typically starts with problem formulation and, in commercial settings, results in the deployment and continued monitoring of a trained model~\cite{passi2019formulation}. 
Problem formulation involves the collection, selection, or curation of a dataset, followed by the operationalization of a concrete task to learn, such as classifying loan-granting decisions or generating natural-language text. 
The actors responsible for formulation may hand off their work to others responsible for implementation --- choosing the type of model and the learning procedure to use for model training. 
In selecting the type of model, these actors may custom-design their own model architecture, or may defer to a pre-existing one, such as an off-the-shelf neural network, which has been designed by others, possibly at another company or institution. 

Thereafter, training and evaluation begin, in which a group of developers run training many times, perhaps with multiple combinations of model types, training procedures, and hyperparameter values. 
These developers compare trained models, from which they select some ``best''-performing model (or ensemble of models), where ``best'' is informed by a quantitative metric they have adopted, such as mean overall test accuracy. 
These stages, from formulation to evaluation, are often repeated dynamically: until the model passes the threshold of developer-specified performance criteria, the process can cycle from re-modeling to tuning. 
If the model is deployed in practice, there is yet another set of actors who monitor the model's ongoing behavior, ensuring that its behavior aligns with expectations developed during training and evaluation.

Each stage of the ML pipeline involves numerous actors --- in fact, potentially \emph{indefinitely} many actors if the pipeline employs third-party model architectures or ML toolkits, which we discuss below.\footnote{Participatory design further expands the set of \textit{many hands} to end-user stakeholders~\cite{sloane2020participation}, illustrating an additional manifestation of the barrier: 
    when harms occur, it is possible to shift blame to harmed end-users who were involved in the ML pipeline.} 
Thus, in practice, if a trained model causes harms, it can be extremely challenging to tease out particular actors who should answer for them. 
For example, harms could originate from how actors operationalize the learning task at the start of the pipeline~\cite{cooper2021emergent}, move from high-level abstraction to concrete implementation~\cite{selbst2019abstraction}, or select hyperparameters or random seeds during model selection~\cite{forde2021model,sivaprasad2020hpo,cooper2021hpo}. 
Blame could lie with actors in any part of the pipeline, or some combination thereof whose faulty actions may have been causally responsible for harm. 
Bias, for example, could creep in early, from the choice of dataset, and accumulate and become magnified further downstream during model selection. 
In other words, the diffuse and dynamic nature of the pipeline makes locating accountability extremely challenging. 
This can be understood as an issue of transparency --- beyond the specific the problem of model interpretability --- 
concerning \emph{who} is responsible \emph{for what}, and \emph{how} this can be related to overarching accountability with respect to a model's ultimate use in practice~\cite{kroll2021traceability}.\footnote{This indicates why transparency in the form of model interpretability may be important, but is ultimately not sufficient, for identifying actors accountable for harms.} 

\paragraph{Multi-purpose toolkits.} 
Practitioners and researchers often do not code model architectures or optimization algorithms from scratch. 
Just as Nissenbaum highlighted the integration of third-party software modules as the indefinite expansion of \textit{many hands}, we note here that builders of data-driven algorithmic systems often rely on toolkits produced by others. 
To decrease the amount of time and money spent iterating the ML pipeline, these actors depend on the investment of tech companies with vast resources and large, concentrated pools of technical talent to develop and release efficient, correct, comprehensive, and user-friendly libraries of algorithm implementations, model architectures, and benchmark datasets~\cite{abadi2016tf,paszke2019pytorch, mattson2020mlperf}. 

Unlike more traditional modules, which only tend to contain reusable software algorithms, ML toolkits often also include large-scale, pre-trained models. 
Large companies train and release such models, like BERT~\cite{devlin2019bert}, which smaller companies and individuals can use out-of-the-box or fine-tune for particular use cases. 
Since these pre-trained models are often intended for downstream use by users different from their developers, they are designed for a multiplicity of applications (i.e., to be general-purpose). 
However, users employ pre-trained models in specific domains; 
there is a gap between general design goals and specific deployment intentions, which has been shown can bring about bias-related harms. 
Determining blame for these types of harms is far from simple. 
For example, if intended use is under-specified, blame could lie at least partially with the pre-trained model's creator. 
Compounding this problem is the fact that ML presents a recursive turn in the \textit{many hands} problem Nissenbaum highlighted, in that many ML systems incorporate pre-trained components that are, themselves, the product of \textit{many hands}. 
Nevertheless, tracing such harms presents an addressable technical challenge, not an insurmountable epistemological barrier.

\vspace{.5cm}\noindent
\textbf{In relation to a moral, relational accountability framework, this barrier obscures \ldots} 

\textit{Who is accountable}:  
\textit{Many hands} is central to identifying an accountable actor within Bovens's framework~\cite{bovens2007analysing}. 
This problem has long characterized challenges in holding corporate actors, institutions, and organizations accountable, and while it certainly constituted a barrier to accountability in 1996~\cite{nissenbaum1996accountability}, it has only become more difficult to understand who is accountable in a data-driven algorithmic society. 
Code reuse --- taken as a virtue in software development ---
has now been extended to model reuse, in turn generating a host of problems for equity and reliability by making it difficult to identify all the actors who contributed to components of an ML pipeline. 
Knowing who is responsible for these components as they are repurposed, as well as who ought to be responsible for incorporating those components into a downstream system, becomes prohibitively difficult for a forum to ascertain on its own, let alone for it to demand any explanations or changes in actors' behavior. 

\textit{For what}: The problem of \textit{many hands} extends the above question to determining \textit{what} an actor might be accountable for in relation to harms, in that it is hard to isolate which part of an ML pipeline actually contributes to an error or harm. 
Repurposed models may introduce dataset imbalances and proxies for protected categories without adequate scrutiny (or even the opportunity for scrutiny) by those assembling downstream components of a system. 
This raises  questions of appropriate use, wherein it is difficult to tease apart the responsibility of those who produced a component to adequately stipulate the limits of its appropriate use and the responsibility of those who use that component to ensure it is appropriate for the uses to which they are putting it. 

\textit{To whom}:  
\textit{Many hands} is primarily a barrier to knowing \emph{who} is accountable, but it is also a barrier to knowing \emph{to whom} those accountable actors are accountable where, for example, a differential error rate may exist for some population $\mathcal{P}$, but a specific harm occurs for an individual $p \in \mathcal{P}$. 
In such a case, it is difficult to determine whether accountability ought to be rendered to $\mathcal{P}$, because of the heightened risk of harm to which the entire population has been exposed, or only to $p$, who suffered harm because of their membership in $\mathcal{P}$. 
This is a \emph{many hands} problem because of the difficulty in knowing where within the ML pipeline risk was produced for the group, e.g., through training or dataset imbalances, and where it was produced for individuals, e.g., through implementation choices. 

\textit{Under which circumstances}: The problem of 
\textit{Many hands} presents a barrier even when standards of care exist, as it is difficult for actors to know precisely whom they should exercise that care toward (see ``to whom'' above). 
Standards of care, which are grounded in normative assumptions about appropriate component (re)use, are less straightforward to develop where \textit{many hands} are involved, as social practices which link actors together are obscured throughout the ML pipeline. 

%% file: section/10-intro/12-accountability/1242-accountability-bugs.tex
\subsection{\textit{``Bugs''}}\label{sec:acc:barriers:bugs}

Nissenbaum uses the term \textit{``bug''} to cover a variety of issues common to software, including ``modeling, design, and coding errors.'' 
\textit{``Bugs''} are said to be ``inevitable,'' ``pervasive,'' and ``endemic to programming,'' ``natural hazards of any substantial system''~\cite[p. 32]{nissenbaum1996accountability}. 
Even with software debuggers and verification tools that can assure correctness, \textit{``bugs''} emerge and cause unpredictable behavior when software systems are deployed and integrated with each other in the real world~\cite{smith1985limits, mackenzie2001proof}. 
The rhetorical power of \textit{``bugs''} is that they are predictable in their unpredictability; 
they serve as a barrier to accountability because they cannot be helped (except in obvious cases), and therefore are often treated as an accepted ``consequence of a glorious technology for which we hold no one accountable''~\cite[p. 34]{nissenbaum1996accountability}. 

What we consider to be the ``inevitable'' can change over time as technology evolves, with certain types of \textit{``bugs''} spilling over into the avoidable. 
For example, evolving norms and new debugging tools can rebrand the ``inevitable'' to be sloppy or negligent implementation, at which point programmers can be held to account for such errors. 
Similarly, the advent of data-driven algorithmic systems has indicated that this malleability also extends in the other direction: 
new technological capabilities can both contract and expand what we consider ``inevitable'' \textit{``buggy''} behavior. 
That is, while these systems contain \textit{``bugs''} of the ``modeling, design, and coding'' varieties that Nissenbaum describes for rule-based programs, the statistical nature of data-driven systems presents additional types of harm-inducing errors, which may present an additional barrier to accountability.\footnote{Of course, statistical software is not new to ML; 
    however, the proliferation of data-driven algorithmic systems has clarified the prevalence of such errors.}  
Where misclassifications, statistical error, and nondeterministic outputs cause harm --- and are presented as inevitable and unavoidable --- may impede the attribution of blame.

In 1996, it may have been evident that labeling certain errors as \textit{``bugs''} was a mere ploy to dodge blame. 
Today, certain types of errors are more plausibly asserted to be an inherent part of ML, attributable to its statistical nature. 
Misclassification, statistical error, and nondeterminism seem to turn the notion of \textit{``bug''} on its head: 
indeed, many experts would as readily call these \textit{features} of machine learning, not \textit{``bugs''}.\footnote{We return to this in Section~\ref{sec:acc:barriers:scapegoat} (\textit{scapegoat}) and is why we leave \textit{``bugs''} in quotes.} 
Nevertheless, regardless of where one attempts to draw the line, these errors share common elements with the \textit{``bugs''} Nissenbaum describes --- namely, they undermine our ability to reason, conclusively, about causality and fault. 
Insofar as they are accepted as an ``inevitable,'' ``pervasive,'' and  ``consequence of a glorious technology,'' they constitute a barrier to accountability~\cite{nissenbaum1996accountability}. 
Below, we illustrate this point with concrete instances of \textit{``bugs''} deemed unavoidable in data-driven algorithmic systems.

\paragraph{Faulty modeling premises.} 
As discussed in Section~\ref{sec:acc:barriers:manyhands}, data\--driven algorithmic systems require significant modeling decisions prior to implementation. 
For example, choosing a model to learn necessarily involves abstraction and can have significant ramifications~\cite{selbst2019abstraction, passi2019formulation}. 
Assumptions during this stage of the ML pipeline can bias the resulting computational solution space in a particular direction~\cite{friedman1996bias}, for example, assuming a linear model is sufficient to capture patterns in data precludes the possibility of modeling non-linearities. 
When such biases involve over-simplified or faulty reasoning, they can result in model mis-specification and the introduction of ``modeling error bugs.'' 
Such mis-specifications may include the assumption that values like fairness and accuracy are correctly modeled as a trade-off to be optimized~\cite{cooper2021emergent}, and that physical characteristics can serve as legitimate classification signals for identifying criminals~\cite{wu2016automated} or inferring sexual orientation~\cite{stark2021pseudo, wang2018gay}. 
More generally, a common modeling error may arise from assuming, in the first place, that a problem is amenable to classification --- that it is possible to divide data examples into separable categories~\cite{sun2019gender, sloane2021silicon}. 
Even if it is possible to train mis-specified models like these to behave ``accurately'' (i.e., to return better-than-chance results after learning these tasks), conclusions drawn from false premises will be unsound~\cite{cooper2021emergent}. 
If modeling assumptions are unclear or elided, an actor may evade accountability by blaming inexplicable, unavoidable \textit{``bugs''} endemic to computer software instead of taking responsibility for otherwise opaque errors.

\paragraph{Individual errors.} 
Even if one's premises are not faulty, the ML pipeline can still produce models that cause harm. 
Trained ML models exhibit errors that can harm individuals if their effects, for example, violate privacy or cause manipulation~\cite{feldman2015disaparte, lovering2021predicting, nadeem2021stereoset, kreps2020gpt}. 
ML has several techniques to quantify and minimize error~\cite{botchkarev2019metrics,hardt2016equality}, and yet even the most robust, well-trained models report imperfect accuracy. 
In fact, a model that achieves $100\%$ accuracy is usually considered suspect, likely over-fit to the training data and to exhibit poor performance when presented with new examples~\cite{srivastava2014dropout, hastie2009statistical, recht2019imagenet}. 
Therefore, when individual errors occur, they can be treated as inevitable, just like the \textit{``bugs''} Nissenbaum describes, displacing responsibility for the harms such errors cause affected individuals.

\paragraph{Bad model performance.} 
Unexpectedly bad overall model performance can likewise be excused as a \textit{``bug,''} rather than a blameworthy error. 
Consider a hypothetical example of a (well-formulated) computer vision system used to detect skin cancer, whose training and evaluation indicate will have an accuracy rate of 94\%. 
Once deployed, if the model coheres with (or even out-performs) its promised performance, then developers can claim that any mis-classifications were expected.\footnote{Individual errors can pose additional challenges for accountability: 
    the model may still overall exhibit an expected degree of error (i.e., be within a margin of error), for which it is possible to scapegoat the statistical nature of ML (Section~\ref{sec:acc:barriers:scapegoat}). 
} 
Since expected accuracy is a probabilistic claim about what is likely to occur, deviations from expectation can and do occur. 
When monitoring a deployed model, over time, if this deviation yields a substantial decrease in expected accuracy, developers may dodge accountability by ascribing the failure to the amorphous category of \textit{``bug''}, instead of admitting that it resulted from human negligence, poor generalization, distribution shift, or other faulty behavior.\looseness=-1

\vspace{.5cm}\noindent\textbf{In relation to a moral, relational accountability framework, this barrier obscures \ldots}

\textit{Who is accountable}: 
Accountability for \textit{``bugs,''} even within the expanded definition of \textit{``bugs''} provided above, emerges from specific regulatory regimes, corporate compliance practices, and contracting relationships. 
Civil law has a crucial role in determining the relationship between forums and responsibilized actors, which is often inflected by those who have the capacity to intervene or have benefitted from a particular action. 
\textit{``Bugs''} present a particular challenge to determining who is accountable when they are seen as endemic to ML, or as produced by non-determinism inherent to the domain in which an algorithmic system is deployed (a challenge shared by the \emph{scapegoating} barrier).

\textit{For what}: \textit{``Bugs''} remain a barrier to accountability because of the difficulty they pose to actors and forums trying to specify whether individual errors, bad model performance, faulty assumptions, or other mistakes contributed to a harm.

\textit{To whom}: \textit{``Bugs''} may affect an entire class of individuals, a community, or all of society, but evidence of harm may only accrue at the level of a specific individual, presenting a barrier for actors and forums interested in knowing to whom accountability ought to be rendered.

\textit{Under which circumstances}: Algorithmic systems inevitably rely on some degree of abstraction and make specific assumptions about the underlying nature of the phenomena they model~\cite{selbst2019abstraction,cooper2021emergent}. 
Under circumstances of imperfect information about every possible aspect of a data-driven algorithmic system (which is most of the circumstances outside the lab), \textit{``bugs''} of the character described above may exist and contribute to this barrier to accountability.

%% file: section/10-intro/12-accountability/1243-accountability-scapegoat.tex
\subsection{\textit{The Computer as Scapegoat}}\label{sec:acc:barriers:scapegoat}

Blaming a computer may pose a barrier to accountability, because ``having found one explanation for an error or injury, the further role and responsibility of human agents tend to be underestimated" \cite[p. 34]{nissenbaum1996accountability}\cite{elish2019crumple}. 
To explain why people could plausibly blame computers for  wrongdoing, Nissenbaum cites the role computers may play in ``tasks previously performed by humans in positions of responsibility;'' 
whereas before the human would be indicated as the blameworthy party, the computer has now taken up that role. 
And yet, even as computer systems have become immediate causal antecedents to an increasing number of harms, they lack moral agency and thus cannot be the bearers of moral blame~\cite{nissenbaum1996accountability}. 
In this section, we discuss how \textit{scapegoating the computer} has become even more complicated in the landscape of ubiquitous data-driven algorithmic systems. 
In the examples below, the system is made to bear the sins of the responsible party, the individual or the institution that has agency and is capable of carrying moral blame.

\paragraph{Moral agency.}
As data-driven algorithmic systems have become pervasive in life-critical contexts, there has been a corresponding tendency to anthropomorphize and view technological processes as akin to human cognition~\cite{turkle2005second, pradhan2019phantom, birhane2020robot}.  
These systems are described by their developers and commentators as intelligent, implying that they have agency as autonomous actors and thus rhetorically positioning them as blameworthy for error. 
However, directing blame toward data-driven algorithmic systems effectively imbues them with moral agency, ascribing them the ability to act intentionally~\cite{schlosser2019sep}. 
Nissenbaum likens blaming a computer to blaming a bullet in a shooting: While the bullet can be said to play an active, causal role, it cannot be said to have been intentional in its behavior. 
In the same vein, a data-driven algorithmic system may play a central role in life-critical decisions, and may even be said to \textit{make a choice} in a particular task, but a choice lacking deliberate intention, a precondition for moral agency~\cite{schlosser2019sep}.\footnote{This is consistent with scholarship in legal theory concerning AI, algorithms, agency, and personhood~\cite{veliz2021moral, birhane2020robot, himma2009artificial, bryson2017and, calo2015robotics, balkin2015robotics, lemley2019remedies}.
} 

\paragraph{``Accountable algorithms.''} 
This popular banner-phrase makes \textit{algorithms} the subject of accountability~\cite{kroll2017aa}, even though algorithms are not bearers of moral agency and, by extension, moral responsibility. 
It places responsibility on technology, not its developers, owners or operators, and it reduces accountability to a piecemeal, procedural quality that can be inferred from technology, rather than a normative concept that has to do with the moral obligations that people have toward one another. 
The phrase further occludes proper attribution of accountability by fixating attention on algorithms rather than on systems that are deployed in practice, within and through which algorithms function~\cite{cooper2021eaamo}. 
When, for example, studies of fairness in AI/ML-assisted judicial bail decisions fixate on respective algorithms, they fail to capture key inequities that are systemic in complex sociotechnical systems, of which AI/ML techniques are just one part~\cite{barabas2020studying}.

\paragraph{Mathematical guarantees.} 
Directing blame away from people and corporations can be either strategic or inadvertent. 
In some cases, a group of harmed individuals does not know whom to blame (\textit{many hands}) and settles on blaming the system. 
In others, scapegoating the system can be a way by which a moral actor dodges and dissipates public ire, for example, in the now-canonical example of Northpointe exhibiting bias in its risk-assessment tool~\cite{angwin2016machine}. 
Rather than attributing this bias to a mistake or \textit{``bug,''} Northpointe blamed the fundamental incompatibility of different algorithmic operationalizations of fairness as the source of the problem (and pointed to a specific measure, for which bias was not detectable, as evidence of blamelessness). 
Reliance on mathematical guarantees can reinforce barriers to accountability and divert attention away from its appropriate subjects. 
One can see this when a given system has a theoretically-guaranteed (and empirically-verified) upper bound on its error. 
If the system behaves within its guaranteed margin of error, it becomes possible to treat that margin as an immutable attribute of the system (rather than, more appropriately, the result of human-made decisions), and to scapegoat the system for any particular errors that fall within this margin. 

Let us consider the same case we discussed for the problem of individual errors in \textit{``Bugs:''} 
The engineers show that a system is 94\% accurate for tumor detection, and validate that this is in fact the case in practice. 
Above, we talked about this example in terms of individual errors, for which responsibility for harm could be excused due to \textit{``buggy''} behavior. 
Rather than analyzing behavior at this level of individual decisions, one can also examine the behavior \emph{of the model overall}. 
If the frequency of mis-classifications is within the model's guaranteed error rate, the engineers could attempt to excuse all resulting harms by gesturing to the fact that the model is performing \emph{exactly as expected}. 
In short, satisfying mathematical guarantees can serve as a \textit{scapegoat} because pointing to mathematical claims satisfied at the model-level can serve to obscure the need to account for harms that occur at the individual-decision level.\footnote{One could see-saw back-and-forth between \textit{``bug''} and \textit{scapegoat} to evade accountability. 
    If satisfying guarantees at the overall model-level is rejected as a rationale for an individual harm, one could claim there is a \textit{``bug;''} 
    if calling an individual decision \textit{``buggy''} is rejected, and the model is classifying within its expected error, one could then displace blame by arguing that the model is behaving according to its specification.
}

\paragraph{Non-determinism.} 
When data-driven algorithmic systems err, their errors can be attributed to the stochastic or otherwise non-deterministic components of either the system itself or the phenomena the system is modeling. 
In particular, systems that involve ML involve randomization, for example, by shuffling the order in which training data examples are presented to an algorithm. 
While such features of ML algorithms may seem like technical minutiae, in fact, they introduce stochasticity into the outputs of machine-learned models: 
training the same model architecture on the same dataset with the same algorithm --- but changing the order in which the training data are supplied to the algorithm --- can yield models that behave differently in practice. 
For example, as Forde at al.~\cite{forde2021model} shows, changing the order that the data examples are presented to train a tumor-detection model can lead to surprisingly variable performance. 
The relationship between training-data-ordering and resulting variance in model performance is under-explored in the technical literature. 
Thus, such differences in model performance are often attributed to an inherent stochasticity in ML. 
The randomization used in ML systems --- randomization on which these systems depend --- becomes a \textit{scapegoat} for the harms it may cause, such as missed tumor detection. 
In attributing the harms to mathematical chance, attention is drawn away from appropriate accountable agents.

\vspace{.5cm}\noindent
\textbf{In relation to a moral, relational accountability framework, this barrier obscures \ldots}

\textit{Who is accountable}: 
Similar actors are accountable as those described in \textit{``Bugs,''} although the barrier presented by \textit{scapegoating} is embedded in its implicit suggestion that entities are accountable, rather than those who are the responsible actors (see the nebulousness of \textit{many hands}), or that no responsible actor can be found because a harm occurred through randomness or chance.

\textit{For what}: 
\textit{Scapegoating} produces barriers to understanding the \emph{for what} of accountability in identical ways as described above in \textit{``Bugs,''}. 
It also contributes an additional difficulty when mathematical guarantees are offered that allow for some minimal degree of undesirable behavior in a system, or the system is characterized as non-deterministic in ways that would indemnify otherwise responsibilized actors from accountability for outcomes stemming from such undesirable behaviors.

\textit{To whom} and \textit{under which circumstances}: Same as in Section~\ref{sec:acc:barriers:bugs}.

%% file: section/10-intro/12-accountability/1244-accountability-liability.tex
\subsection{Ownership without Liability}\label{sec:acc:barriers:liability}

Nissenbaum~\cite{nissenbaum1996accountability} highlights a dual trend in the computer industry: 
1) strengthening property rights and 2) avoiding liability. 
Behavioral trends that informed these assertions have persisted in the decades since, with lively public debates over the fit of traditional forms of intellectual property (i.e., copyrights, patents, and trade secrets) to digital products such as software, data, databases, and algorithms~\cite{cofone2019secrecy, fromer2019machines}, and subsequent expensive legal struggles among industry titans~\cite{harvard2021oracle}. 
Similarly, we have seen explicit denials of liability expressed in shrink-wrap licenses, carried over into so-called ``click-wrap'' licenses, and Terms of Service disclaimers accompanying websites, web-based services, mobile apps, Internet of Things devices, content moderation decisions, and the like~\cite{kosseff2022sec230, citron2022privacy, levy2014intimate, tereszkiewicz2018digital}.

Before addressing how we see these trends carry forward in the contemporary landscape, we need to qualify our observations. 
Property and liability are weighty legal concepts with long histories and rich meanings. 
Narrowing our view to digital technologies, even before Nissenbaum~\cite{nissenbaum1996accountability}, a robust literature had grown over questions of ownership --- questions that have persisted through numerous landmark court cases. 
Liability, too, is a core legal concept that is increasingly an issue in relation to the products and services of digital industries. 
It lies outside the scope of this paper to attempt meaningful insights into these concepts as they manifest in scholarship, law, and the courts. 
However, it is useful to observe broad patterns and anticipate the likely actions of stakeholders. 
For a start it is not difficult to see how the trends toward strong ownership and weak liability reinforce barriers to accountability, and also to understand why industry incumbents might support them: 
liability is costly and strong property rights enrich rights holders and empower them against competitors. 
Four lines of advocacy on behalf of industry interests are noted below, supplementary to those discussed in Nissenbaum~\cite{nissenbaum1996accountability}: 
\begin{itemize}
\item Third-party providers of data-driven algorithmic systems refuse to expose their systems to scrutiny by independent auditors on grounds of trade secrets~\cite{cofone2019secrecy, fromer2019machines}. 
As long as experts maintain that transparency is necessary to evaluate the ML pipeline and AI development, strong property rights that block scrutiny are barriers to accountability.

\item Manufacturers and owners of cyber-physical systems, such as robots, Internet of Things devices, drones, and autonomous vehicles, evade liability for harms by shifting blame to environmental factors or humans-in-the-loop~\cite{lemley2019remedies}. 
In this respect, the barrier of \textit{ownership without liability} for data-driven algorithmic systems suggests a twist on the problem of \textit{scapegoating} (Section~\ref{sec:acc:barriers:scapegoat}): treating ``the human user as scapegoat'' --- claiming the user has mis-used an AI- or ML-enabled system in order to obscure responsibility for unclear, under-specified, or deliberately misleading user interfaces or expected use, as has happened with Tesla and accidents concerning its (so-called) ``AutoPilot'' autonomous driving feature~\cite{boudette2021tesla}.

\item Almost without question the computer industry, having metamorphosed into the data industry, has assumed ownership over data passing through its servers~\cite{ftc2014databrokers,okidebe2022data, lambdan2019databrokers}. 
We still do not have clear rules of liability for industry actors when their servers, holding unimaginable quantities of data, are breached~\cite{sharkey2016can}. 
Nor do we have sufficient insight into the completeness, quality, or validity of data, or the means to hold anyone liable for its misuse.\looseness=-1

\item Technology companies hold unprecedented sway over regulation. 
Twenty-five years ago, the software industry was already a force to be reckoned with and successfully persuaded Congress that imposing legal constraints would stifle innovation --- that societal well-being depended on a nascent industry that could not flourish under excessive regulatory and legal burden. 
\end{itemize}

\vspace{.5cm}\noindent
\textbf{In relation to a moral, relational accountability framework, this barrier obscures \ldots}

\textit{Who is accountable}: Having already enumerated above the many difficulties these barriers pose for tracing relationships of accountability, they generally pertain to the problem of \textit{ownership without liability}, as well. 
Additionally, questions of how liability is adjudicated in practice may obscure who is liable, what kind of liability they hold, or what they are liable for, while leaving intact the ways in which the benefits of data-driven algorithmic systems accrue to their developers, designers, and operators.

\textit{For what}: 
\textit{Ownership without liability} affects the very contours of what an actor can be found liable for. 
However, this does not absolve that actor of their moral responsibility or obviate the need for them to be held accountable for the consequences of their actions, the systems they oversee, or from which they benefit.

\textit{To whom}: 
\textit{Ownership without liability} is a barrier to accountability for those who may stand as plaintiffs in civil cases and representatives of those affected.

\textit{Under which circumstances}: \textit{Ownership without liability} is a barrier to accountability where those who suffer a harm lack standing in a court of law. 
This may be because a harm is not cognizable to courts (see, e.g., Metcalf et al.~\cite{metcalf2022relationship}), the harmed party does not constitute a certifiable class, or the nature of the harm is obscured through the ways harms are foisted onto \textit{scapegoats} or dismissed as \textit{``bugs''}.  

%% file: section/10-intro/12-accountability/1250-accountability-beyond.tex
\section{Weakening the Barriers}\label{sec:acc:beyond}

Nissenbaum~\cite{nissenbaum1996accountability} warned of a waning culture of accountability --- harms befalling individuals, groups, even societies, were being cast merely as sufferers' bad luck. 
In the previous section, we revisited the four barriers in light of data-driven algorithmic systems and found that the framework still provides a useful lens through which to locate sources contributing to the dissipation of accountability. 
Weakening the barriers would clear the way for more sound attribution of blame, in turn setting up a stronger societal expectation for blameworthy parties to step forward and take account. 
But we have also argued that accountability in algorithmic societies involves more: 
stepping forward is a necessary component of accountability, but it is insufficient (Section~\ref{sec:acc:framing}). 
Because the barriers we have described may not all be weakened, even with a firm resolve to identify blameworthy parties, we need more than astute attention on a case-by-case basis. 
To build a lasting culture of accountability, a necessary supplement involves establishing persistent institutional frameworks for identifying accountable parties (i.e., individuals, groups, or organizations) and for calling them to answer. 
Simultaneously, such frameworks should invest others with the powers to call these parties to account.

Any technical interventions that the research community has already developed --- notably, those that we have emphasized concerning transparency, audits, and robustness --- would need to be folded into such a framework, and their use justified in these moral and relational terms. 
For example, any technical definition of transparency is unlikely to satisfy the needs of all those who comprise a forum and who may hold variable or inconsistent ideas about what it might mean for a model to be ``interpretable.'' 
Technical assertions of robustness say what expectations are, but leave unanswered the question of the conditions under which deviations from expectations ought to be expected or remedied.\footnote{Moreover, if assumptions underlying such assertions are voided when moving from theory to deployment, robustness estimates can degrade in practice.} 
Relational treatments of these issues, it would seem, require that the obligation be tuned to the various needs of all members of the forum.

\paragraph{Taking each barrier in turn.} 
A moral and relational accountability framework opens the aperture to addressing \textit{many hands} (Section~\ref{sec:acc:barriers:manyhands}). 
In principle, many, if not all, of the \textit{many hands} could be designated as accountable actors. 
Deliberate consideration of the \textit{many hands} problem is clearly called for by those who develop licensing agreements relying on normative assumptions about appropriate use and reuse within the ML pipeline, and in articulating engineering best practices empirically against theoretical assumptions of robustness. 
This includes dataset creators, model developers, decision and control systems designers, vendors, and operators of these systems. 
Developing rigorous standards of care could help mitigate the problems of inappropriate use of pre-trained models and unclear measures of quality control at different stages of the ML pipeline. 
For example, robust auditing mechanisms at each stage, rather than approaching audit as an end-to-end concern~\cite{raji2020closing}, or worse, as a purely \textit{post hoc} endeavor, could help clarify the relationship between stage-specific issues and resulting harms.

Addressing various harms, depending on how they are contextualized, can implicate either the barrier of \textit{``bugs''} or \textit{scapegoating the computer} (Sections~\ref{sec:acc:barriers:bugs} \&~\ref{sec:acc:barriers:scapegoat}). 
For example, we note that the computer science community could have either treated algorithmic harms due to unfair discrimination as a \textit{``bug''} or blamed them on intrinsic aspects of AI/ML --- and yet, it did not. 
Instead, unfairness has more often been ascribed to biased or imbalanced training data~\cite{kallus2018residual, fish2016fair} --- data that exhibits historical biases that are arguably ``pervasive'' and ``unavoidable.'' 
This community could have pursued some ``tolerable'' degree of unfavorable outcomes in the real world (ideally, in consultation with those adversely impacted), and developed ways of ensuring models met that more ``tolerable'' specification, under specific conditions. 
This, notably, would still have allowed developers to evade accountability by \textit{scapegoating} inherent properties of the model as instead deserving of blame. 

However, instead of treating unfairness as an aspect of accountability, much technical work on algorithmic fairness has attempted to address unfairness harms by developing training algorithms that are robust to biased input data. 
The field of algorithmic fairness therefore serves as an example that challenges the narrative of the invulnerability of the barriers. 
The technical community and its interlocutors have demanded more from ML modelers concerning the treatment of unfair discrimination.  
The community has set expectations concerning the necessity of interventions to root out and correct for unfairness, thereby weakening the barriers of \textit{scapegoating} or being attributed to \textit{``bugs''}. 
This example could, and we believe should, encourage similar treatment of other issues like robustness and its relationship to privacy violations, or adversarial ML and its relationship to manipulation. 

Lastly, \textit{being liable} is related but not identical to being accountable (Section~\ref{sec:acc:barriers:liability}). 
The latter is applied to blameworthy parties who step forward to answer, the former to parties who step forward to compensate victims of harm. 
Often liability is assigned to those who are found to be blameworthy. 
If lines of accountability are blurred, for example, as a consequence of the barriers we have discussed, harms due to AI/ML and other data-driven algorithmic systems will be viewed as unfortunate accidents; the cost of ``bad luck" will settle on victims. 
Instead, legal systems have developed approaches, such as strict liability, to compensate victims harmed in certain types of incidents even without a demonstration of faulty behavior. 
Strict liability assigned to actors who are best positioned to prevent harm is sound policy as it motivates these actors to take extraordinary care with their products. 
Barriers such as \textit{many hands} make the attribution of blame difficult. 
Strict liability for a range of harms that are produced by \textit{many hands} would shift the ``bad luck" from victims to those best positioned to mitigate and prevent such harms.

\vspace{.5cm}\noindent 
Eroding the barriers of accountability is a key societal challenge requiring multiple forms of expertise and, with respect to ML especially, the use of these tools needs to be justified. 
Just as mature political governance requires durable institutions and formal attributions of rights and duties, we have similar needs for the governance of producers, purveyors, and operators of data-driven algorithmic systems. 
That is, as we have contended throughout this paper, accountability is moral \emph{and} relational. 
It depends on social, legal, and political structures that provide legitimacy for the checks actors and forums place on each others' behavior; 
it depends on the way those checks are internalized as professional, personal, legal, and ethical duties that motivate actors' personal responsibility. 
Multi- and inter-disciplinary research on accountability, fairness, and transparency --- given its potential to bring together an array of expertise focused on themes of equity and justice --- is uniquely positioned to help develop a moral, relational accountability framework. 
Such structures provide legitimacy, as well as the professional codes and standards of care, disciplinary norms, and personal mores that tie moral and relational forms of accountability together. 
The future work of creating these structures, as noted earlier, is no small undertaking, it lies in the sociopolitical contestations, the hard, deliberative work of living within a pluralistic society, by the many constituencies implicated in any particular computational system.  

%% file: section/10-intro/12-accountability/1260-accountability-conclusion.tex
\section{Conclusion}\label{sec:acc:conclusion}
In this paper we revisited Nissenbaum's ``four barriers'' to accountability, with attention to the contemporary moment in which data-driven algorithmic systems have become ubiquitous in consequential decision-making contexts. 
We have drawn on conceptual framing from Nissenbaum's use of the concept of \emph{blameworthiness} and how it can be aligned with, rather than cast in opposition to, Bovens's work on accountability as a \emph{relational property of social structures}~\cite{bovens2007analysing,bovens2014public}. 
We have demonstrated how data-driven algorithmic systems heighten the barriers to accountability with regard to determining the conditions of blame, and have looked ahead to how one might endeavor to weaken the barriers. 
In particular, we have put forward the conditions necessary to satisfy a moral and relational accountability framework, discussed how the development of such a framework would weaken the barriers, and argued that an interdisciplinary and multidisciplinary research community is uniquely positioned to construct such a framework and to develop lines of inquiry to erode the barriers to accountability. 

Given our tender historical moment, addressing why these or those parties belong in the forum or in the set of accountable actors, why those obligations are justified, and, of course, evaluating the numerous permutations the relational nature of the approach demands is the provenance of future work. 
No easy formulations make sense until we have developed a rigorous approach to justification. 
In our view, this calls for expertise in relevant technologies, moral philosophy, the prevailing political economy of data and computing industries, organizational sociology, current political and regulatory contexts, domain area expertise, and more. It is not that all these are needed all the time; 
but any of them may be called in to develop linkages between proposed values and social welfare.